\documentclass[10pt,twocolumn,letterpaper]{article}

\usepackage{iccv}
\usepackage{times}
\usepackage{epsfig}
\usepackage{graphicx}
\usepackage{amsmath}
\usepackage{amssymb}
\usepackage{authblk}

\usepackage{array}
\usepackage{capt-of}
\usepackage{comment}
\usepackage{booktabs}
\usepackage{multirow}
\usepackage{multicol}
\usepackage[dvipsnames]{xcolor}
\usepackage{qtree}

\makeatletter
\let\BaseCaption\@makecaption
\makeatother

\usepackage{subcaption}

\makeatletter
\let\@makecaption\BaseCaption
\makeatother

\usepackage{appendix}

\newcommand{\topic}[1]{\vspace{1mm}\noindent\textbf{#1}}
\newcommand{\norm}[1]{\left\lVert#1\right\rVert}
\newcommand{\PreserveBackslash}[1]{\let\temp=\\#1\let\\=\temp}
\newcolumntype{C}[1]{>{\PreserveBackslash\centering}p{#1}}

\usepackage[pagebackref=true,breaklinks=true,letterpaper=true,colorlinks,bookmarks=false]{hyperref}

\iccvfinalcopy

\begin{document}

\title{StyleFusion: A Generative Model for Disentangling Spatial Segments \vspace{-2ex}}

\author[]{Omer Kafri}
\author[]{Or Patashnik}
\author[]{Yuval Alaluf}
\author[]{Daniel Cohen-Or}
\affil[]{The Blavatnik School of Computer Science, Tel Aviv University}

\twocolumn[{%
\renewcommand\twocolumn[1][]{#1}%
\vspace{-2em}
\maketitle
\vspace{-3em}
\begin{center}
    \includegraphics[width=\linewidth]{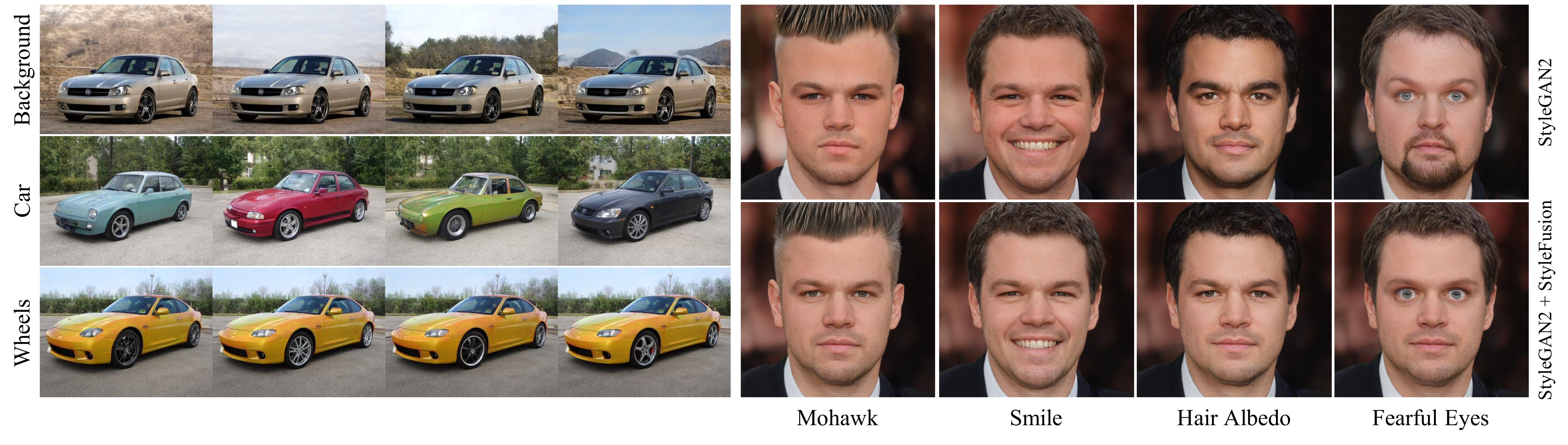}
    \captionof{figure}{StyleFusion learns a disentangled representation for composing an image by fusing several latent codes each corresponding to a single semantic spatial region of the generated image. 
    In doing so, StyleFusion provides users the ability to vary a target semantic region by modifying its corresponding latent code without altering other image regions, as shown to the left.
    Additionally, by pairing StyleFusion's learned representation with existing editing techniques, we obtain more precise edits which better preserve identity and other facial attributes.
    }
    \label{fig:teaser}
\end{center}%
}]

\begin{abstract}
\vspace{-0.35cm}
We present \textit{StyleFusion}, a new mapping architecture for StyleGAN, which takes as input a number of latent codes and fuses them into a single style code. 
Inserting the resulting style code into a pre-trained StyleGAN generator results in a single \textit{harmonized} image in which each semantic region is controlled by one of the input latent codes. 
Effectively, StyleFusion yields a disentangled representation of the image, providing fine-grained control over each region of the generated image. 
Moreover, to help facilitate global control over the generated image, a special input latent code is incorporated into the fused representation.
StyleFusion operates in a hierarchical manner, where each level is tasked with learning to disentangle a pair of image regions (e.g., the car body and wheels).
The resulting learned disentanglement allows one to modify both local, fine-grained semantics 
(e.g., facial features) 
as well as more global features (e.g., pose and background), providing improved flexibility in the synthesis process.
As a natural extension, StyleFusion enables one to perform semantically-aware cross-image mixing of regions that are not necessarily aligned.
Finally, we demonstrate how StyleFusion can be paired with existing editing techniques to more faithfully constrain the edit to the user's region of interest.
Code is available at:
\small{\url{https://github.com/OmerKafri/StyleFusion}}.
\end{abstract}
\begin{figure*}
    \centering
    \includegraphics[width=0.975\linewidth]{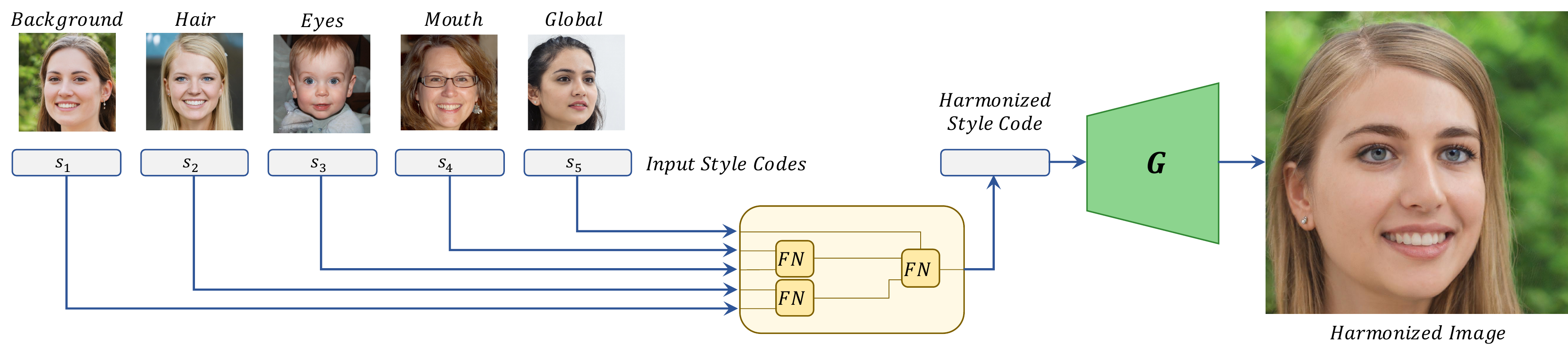}
    \vspace{-0.1cm}
    \caption{
    \textit{StyleFusion Overview.} 
    Given a set of style codes, StyleFusion generates a single \textit{harmonized} image in which each semantic region is controlled by one of the input latent codes. 
    This harmonized image is obtained via a pre-trained generator (e.g., StyleGAN) by \textit{learning} to fuse each of the input latent codes into a single unified style code.
    In a sense, this learned fusion results in a semantically-aware, disentangled representation of the image. We additionally introduce a latent code tasked with controlling global aspects of the generated image (e.g., the individual's pose). Observe that StyleFusion operates in a hierarchical manner and learns to disentangle the image regions in a coarse-to-fine fashion, providing users control over the learned disentanglement.
    }
    \label{fig:overview}
\end{figure*}

\section{Introduction}

Generative Adversarial Networks (GANs) \cite{Goodfellow2014GenerativeAN} are nowadays considered a strong and common method for synthesizing images of phenomenal realism. In particular, StyleGAN \cite{karras2019style, karras2020analyzing, Karras2020ada} has been established as the state-of-the-art synthesis network, producing images of high visual quality and fidelity. The recent surge of interest in StyleGAN should also be attributed to its learnt latent space $\mathcal{W}$, which has been shown to encode valuable semantic information~\cite{abdal2019image2stylegan, abdal2020image2stylegan++} and intriguing disentanglement properties \cite{yang2020semantic,shen2020interpreting,wu2020stylespace,collins2020editing}.

Disentangling various attributes is of utmost importance as it helps facilitate better and easier use of the generator and control over the synthesized images. Therefore, different latent spaces \cite{abdal2019image2stylegan, abdal2020image2stylegan++, wu2020stylespace, zhu2021improved} and architecture variants \cite{kwon2021diagonal, gal2021swagan,lewis2021vogue,kim2021stylemapgan} of StyleGAN have been explored, aiming to construct a more disentangled latent representation. 
To leverage the disentanglement of StyleGAN's latent space for manipulating specific segments of an image, recent works have demonstrated that specific channels of the latent code control spatially local attributes of the generated image \cite{wu2020stylespace, collins2020editing}. 
However, while there exist diverse editing techniques, they oftentimes struggle in achieving fine-grained local control over the generation as
as the single input latent code influences the synthesized image in its entirety.
As such, attaining accurate control requires a precise manipulation over the single latent representation. 

In this work, we leverage the expressiveness of the StyleGAN latent space to provide direct and intuitive control over local, possibly semantic, segments of the generated image. 
Specifically, we introduce a mapping mechanism that learns to generate a single \textit{harmonized} image from a \textit{set} of latent codes, each controlling a specific segment of the synthesized image. 
This mechanism results in a disentangled latent representation corresponding to disjoint semantic image regions.
As shall be shown, by learning a semantically-aware disentanglement, our proposed approach, \textit{StyleFusion}, facilitates easier control over local, fine-grained features (e.g., facial attributes) and global image features, see Figure~\ref{fig:teaser}. 
Specifically, StyleFusion provides an intuitive ``plug-and-play" interface for users to compose a single harmonized image formed by fusing multiple latent codes, see Figure~\ref{fig:overview}.
As shown, StyleFusion operates in a hierarchical fashion, where each level of the hierarchy, a \textit{FusionNet} learns to disentangle a pair of image regions. 

Observe that constructing a harmonized image from a set of latent codes is not trivial. For example, composing a single facial image where the background, identity, and hair are extracted from three different images is especially challenging when the three images are unaligned (e.g., of different poses) or contain different global features (e.g., lighting). 
To address this, we introduce an additional latent code tasked with aligning these global features of the source images before fusing the input latent codes. As a result, this learned alignment results in a more precise composition of semantic regions in the synthesized image.

\begin{figure}
    \setlength{\tabcolsep}{1pt}
    \centering
    \small{
        \begin{tabular}{C{0.18\linewidth} c C{0.18\linewidth} C{0.18\linewidth} C{0.18\linewidth} C{0.18\linewidth}}
        
        & & StyleCLIP & InterFace & GANSpace & GANSpace \\
        
        \includegraphics[width=\linewidth]{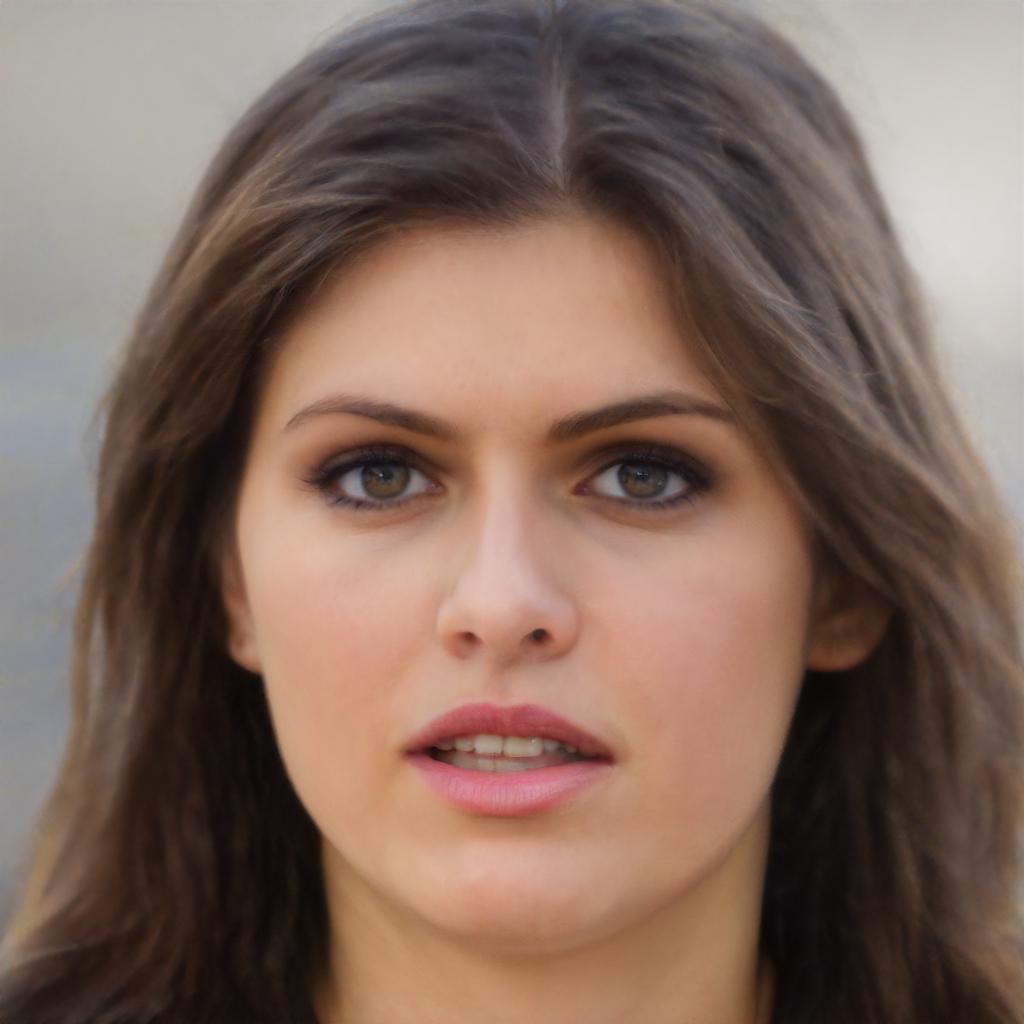} &
        \raisebox{0.2in}{\rotatebox[origin=t]{90}{\footnotesize StyleGAN2}} &
        \includegraphics[width=\linewidth]{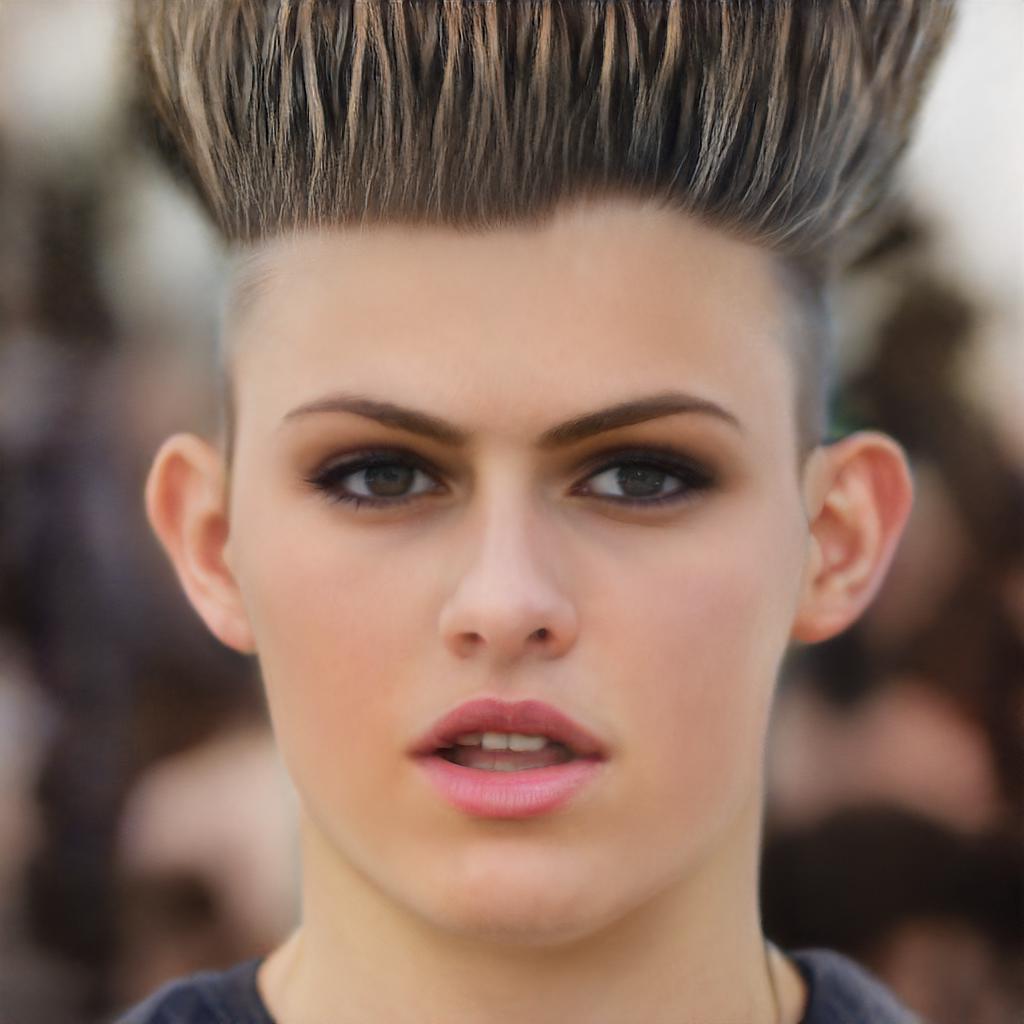} &
        \includegraphics[width=\linewidth]{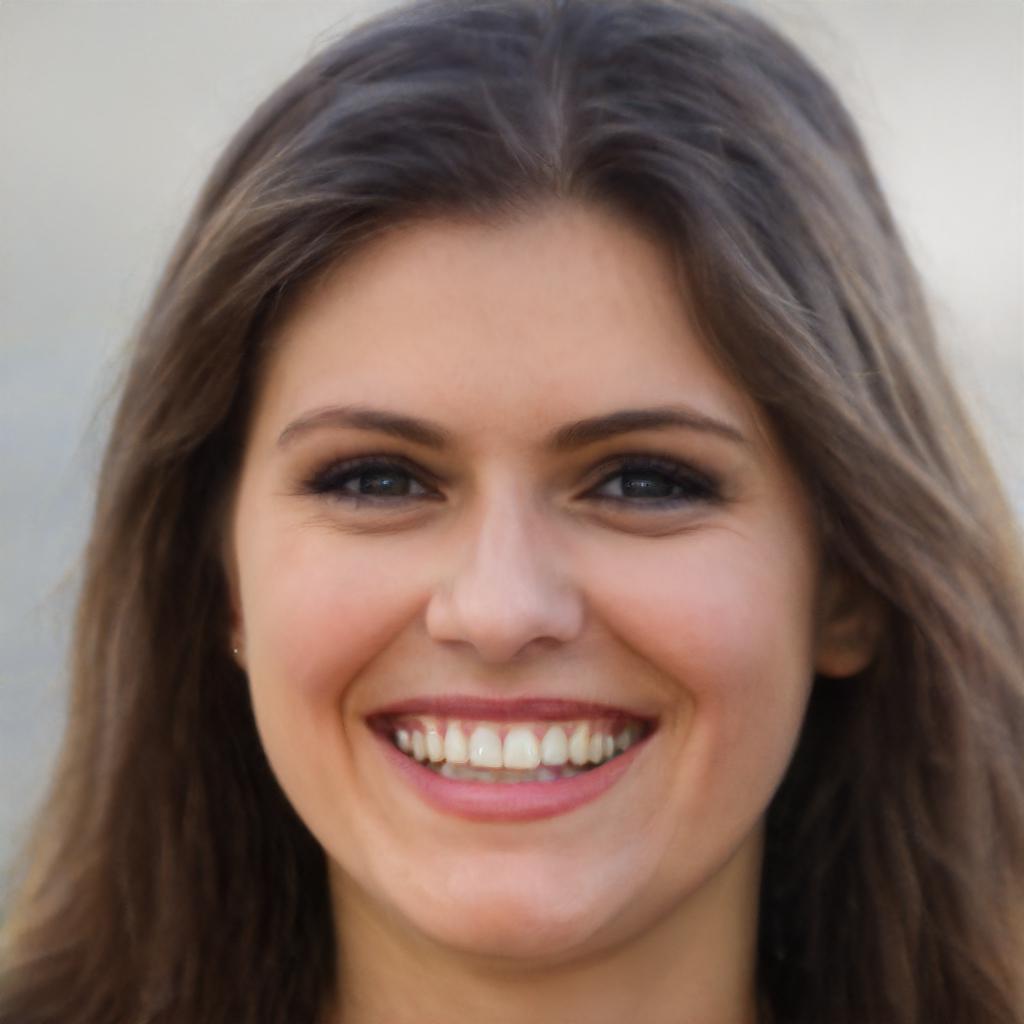} &
        \includegraphics[width=\linewidth]{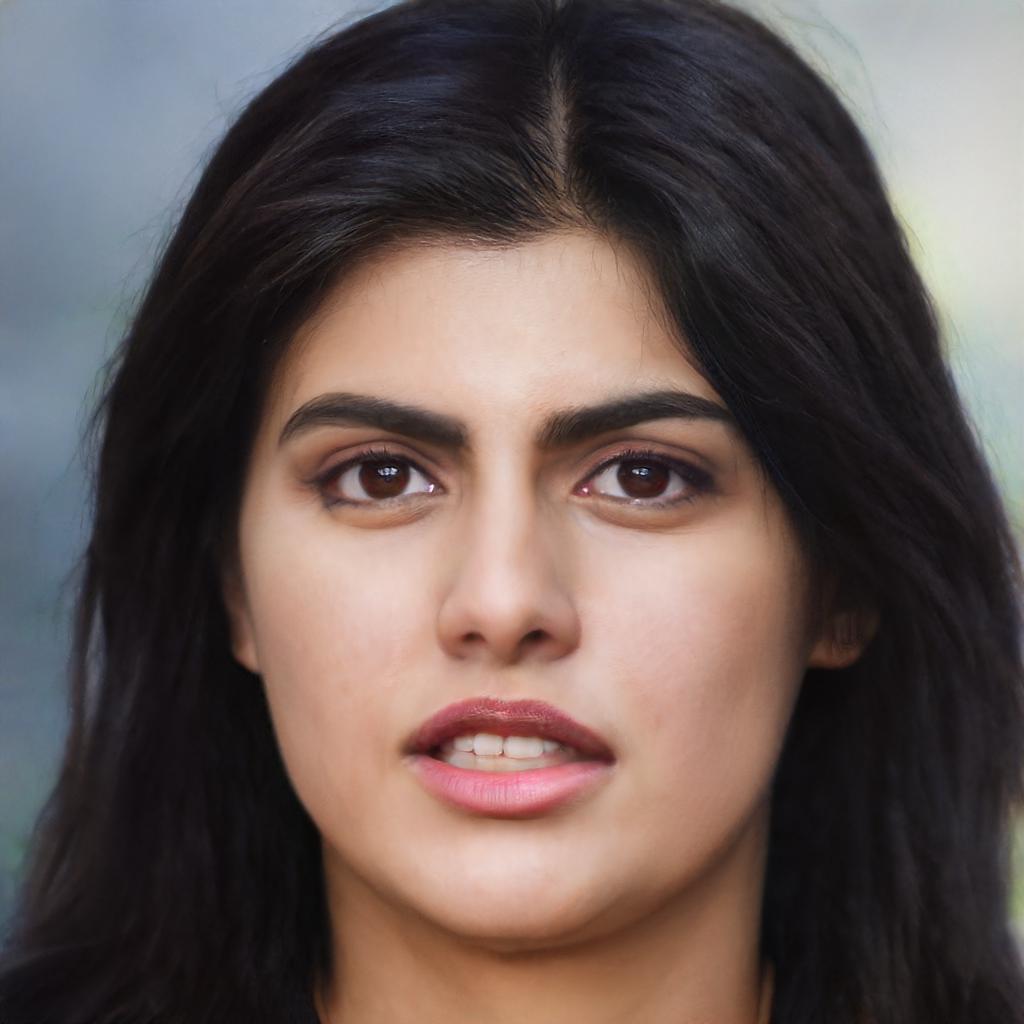} &
        \includegraphics[width=\linewidth]{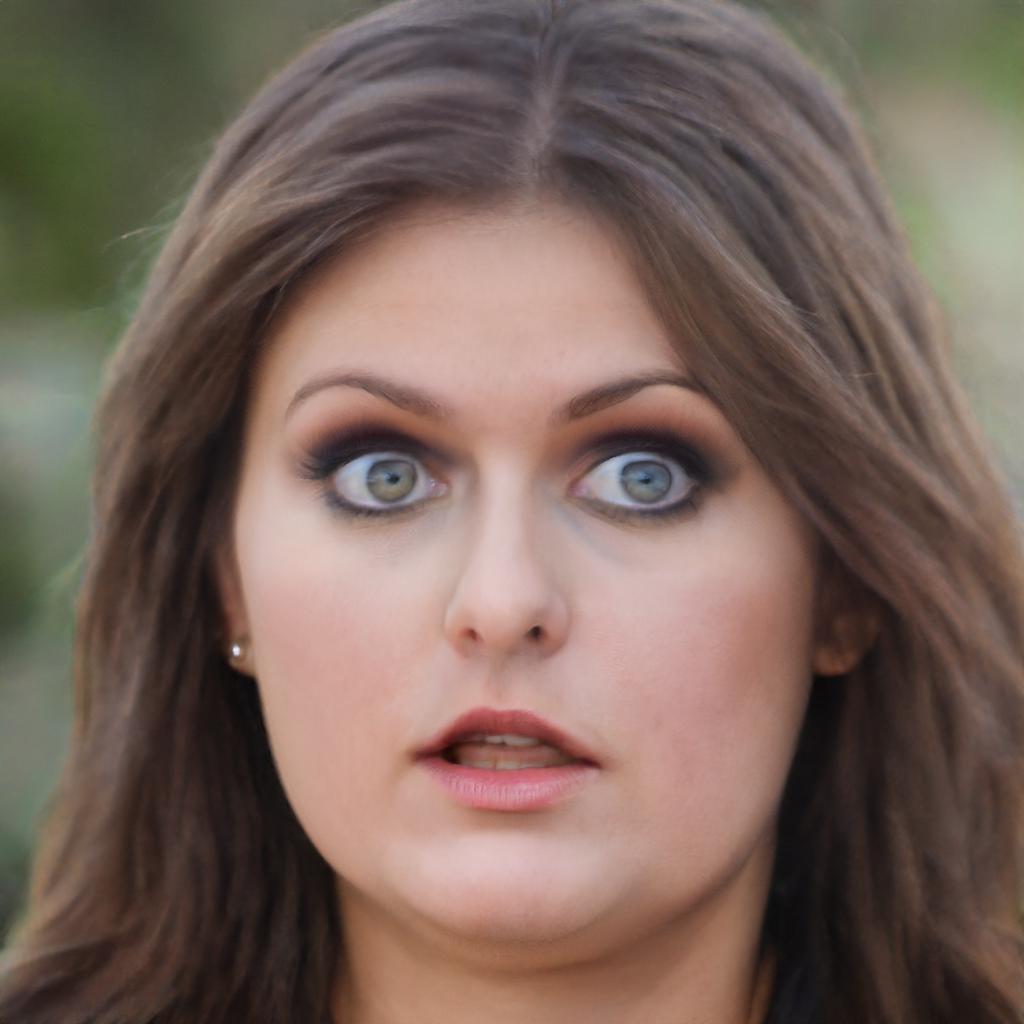} 
        \tabularnewline
        & \raisebox{0.22in}{\rotatebox[origin=t]{90}{\footnotesize StyleFusion}} &
        \includegraphics[width=\linewidth]{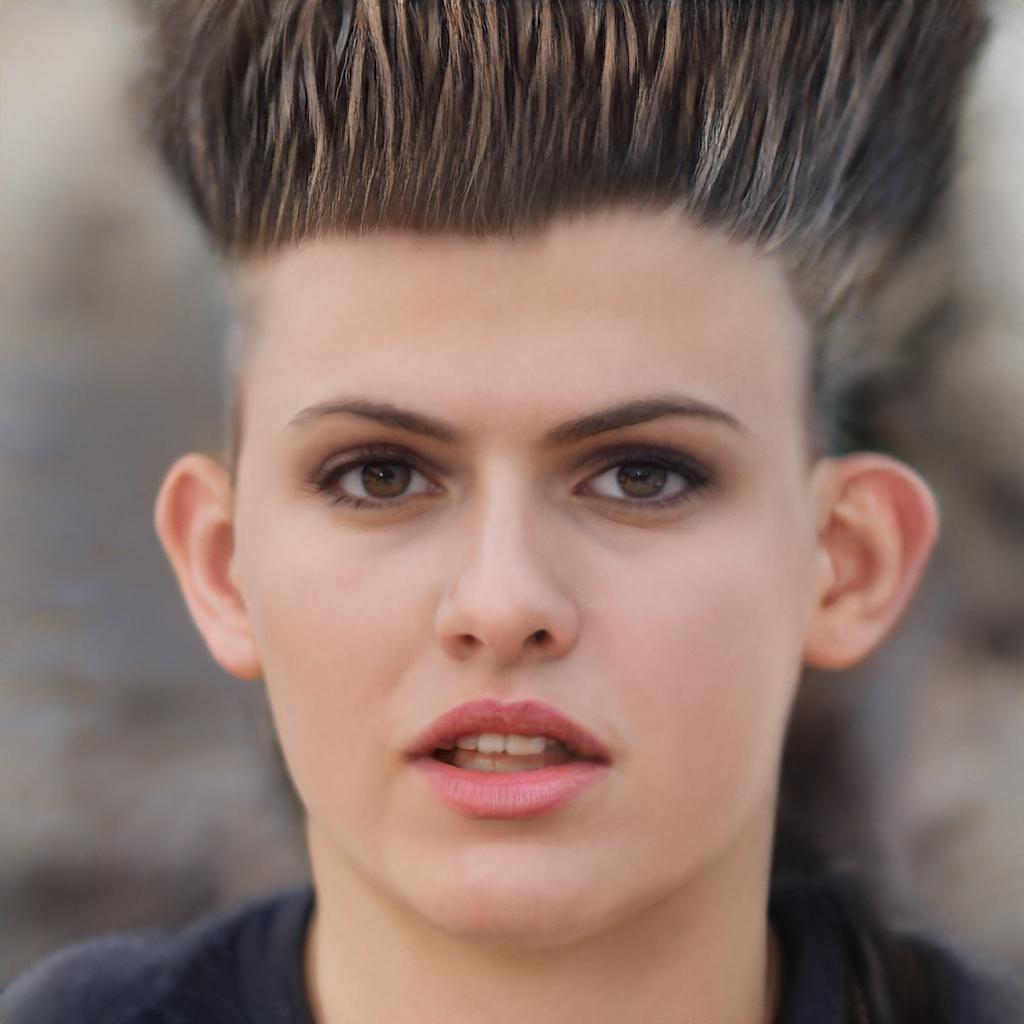} &
        \includegraphics[width=\linewidth]{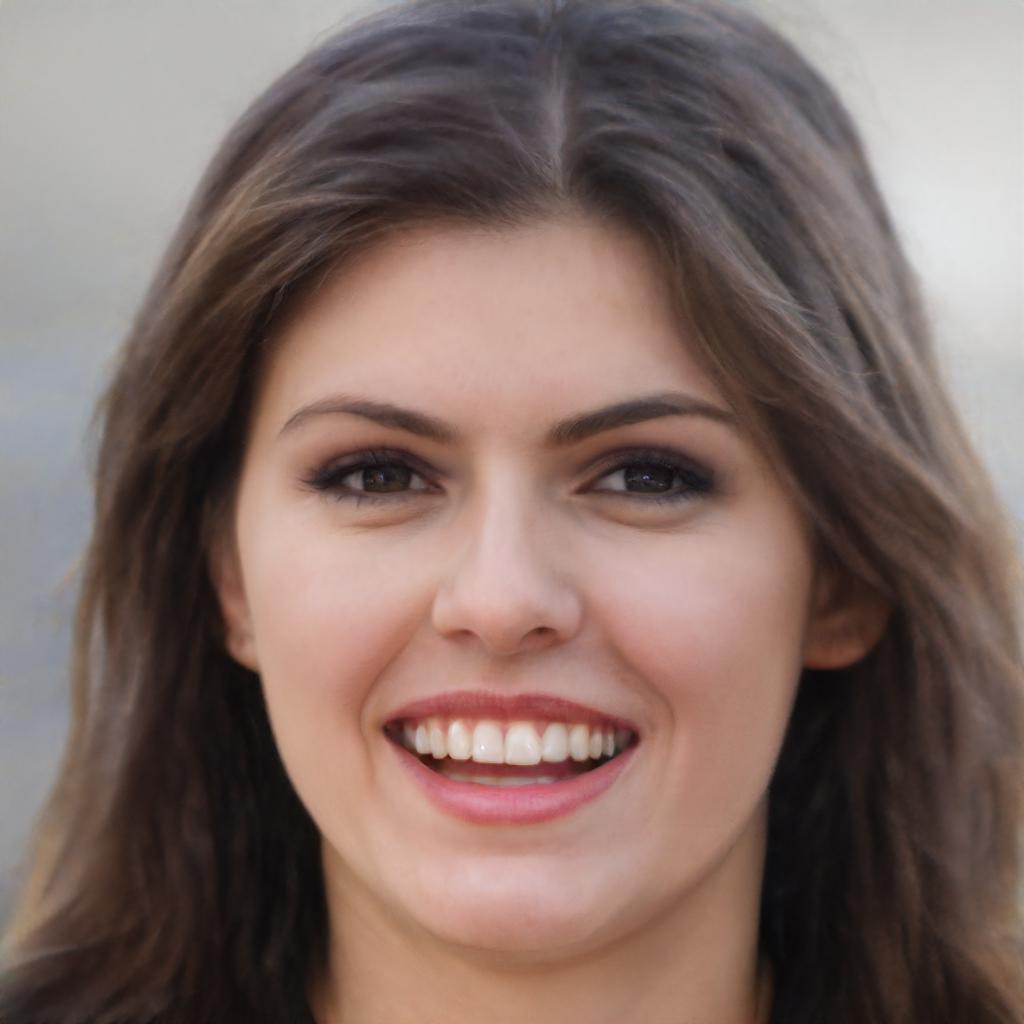} &
        \includegraphics[width=\linewidth]{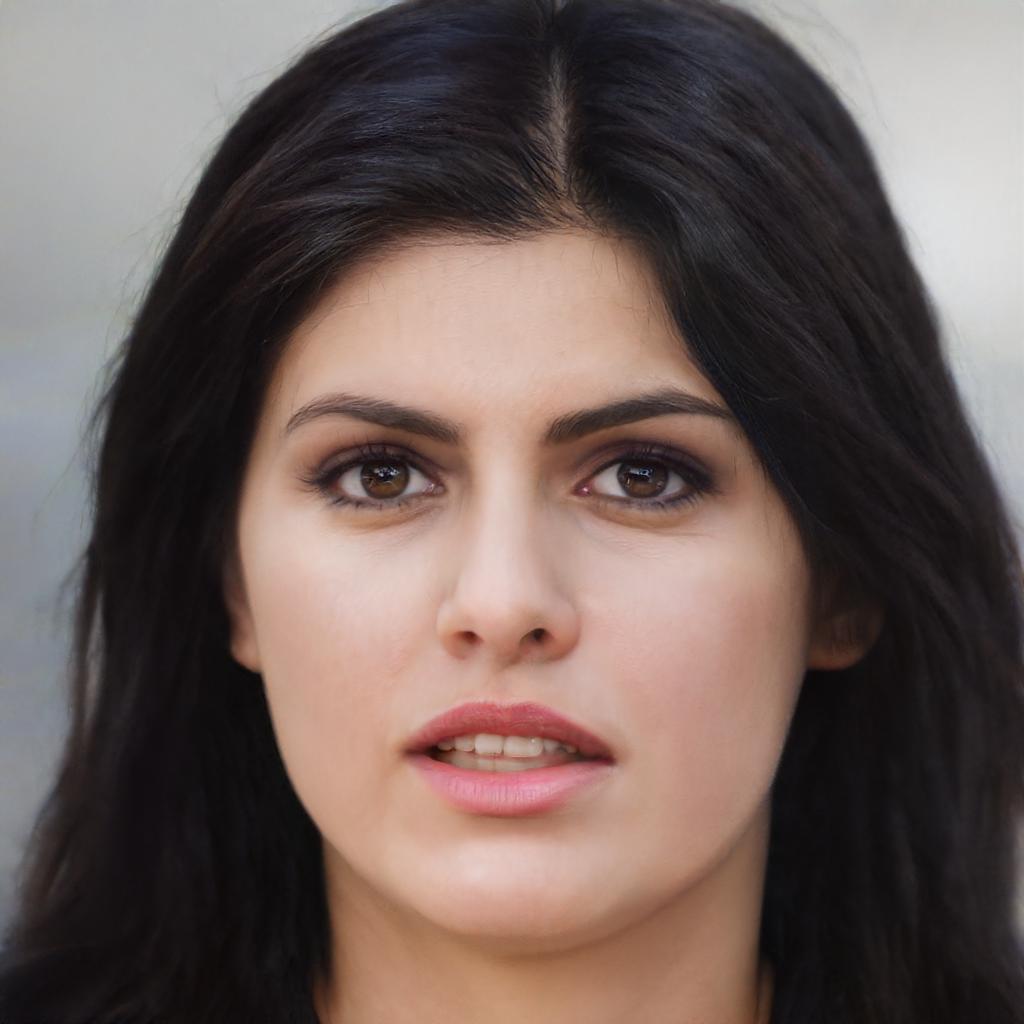} &
        \includegraphics[width=\linewidth]{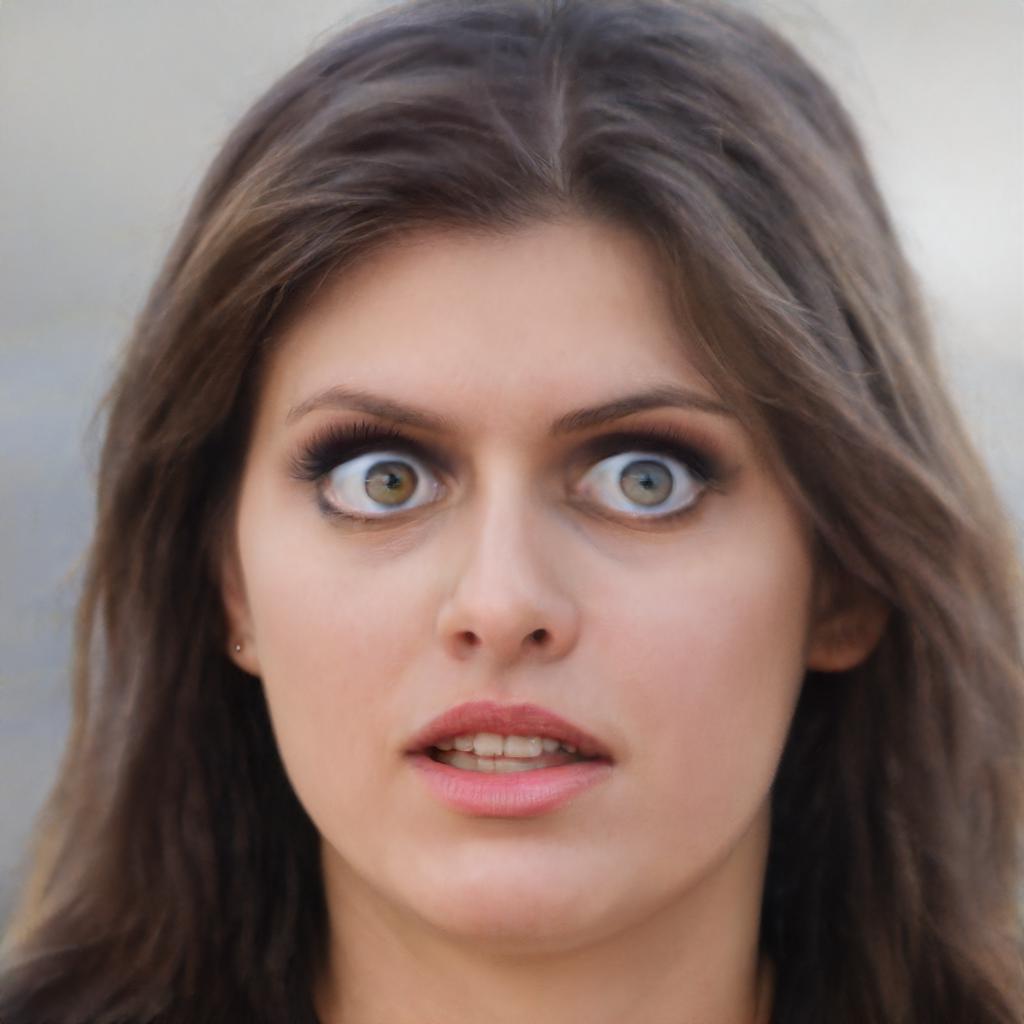} 
        \tabularnewline
        
        Input & & Mohawk & Smile & Hair Albedo & Fearful Eyes
        \tabularnewline
        \end{tabular}
        }
    \vspace{0.1cm}
    \caption{\textit{Latent traversal editing techniques performed with StyleFusion.} 
    As shall by demonstrated, pairing StyleFusion with existing latent editing techniques (e.g., InterFaceGAN~\cite{shen2020interpreting}, StyleCLIP~\cite{patashnik2021styleclip}) results in more precise image manipulations.}
    \label{fig:latent_editing_teaser}
\end{figure}

We perform an extensive analysis of our method to show the advantages of the learned disentangled representation StyleFusion offers on multiple domains. 
To demonstrate this, we show how StyleFusion allows one to generate images with explicit control over a  semantic target image region. 
Moreover, we show the role of the global latent codes in the synthesis process. 
Finally, we illustrate how our method complements other editing techniques. Specifically, as shown in Figure~\ref{fig:latent_editing_teaser}, our intuitive disentangled representations allow one to easily pair existing latent space editing techniques with our proposed representation, facilitating improved control over the manipulation of \textit{real} images.

\section{Background and Related Works}

\subsection{Learning a Disentangled Representation}
Learning a disentangled representation is of utmost importance in a multitude of Computer Graphics and Computer Vision tasks. A disentangled representation can be viewed as a representation that encodes data using independent factors of variation, allowing one to control a single factor without affecting other factors~\cite{locatello2019challenging}.

A key challenge in learning a disentangled representation is reducing the supervision needed to learn the desired disentanglement. For example, manually collecting paired data of the same scene in different styles is tedious and even impractical at times. To address this, diverse weakly-supervised and unsupervised methods have recently been explored.
InfoGAN~\cite{chen2016infogan} learns a disentangled representation in an unsupervised manner by maximizing the mutual information between the latent variables and the observation.
Lee \etal~\cite{lee2020highfidelity} pair a Variational Autoencoder (VAE) and a GAN for learning a more meaningful, disentangled latent representation. 
Others have explored additional means for learning disentangled representations including, but not limited to, adversarial training~\cite{mathieu2016disentangling}, cycle-consistency~\cite{jha2018disentangling}, and mixing sub-parts of different latent representations~\cite{karras2019style,hu2018disentangling,gal2020mrgan,hong2020low}.

Following the seminal work of Isola \etal~\cite{pix2pix2017}, numerous works~\cite{wang2018pix2pixHD,zhu2017unpaired,kotovenko2019content,lee2018diverse} learn to disentangle style and content for performing image-to-image translation between two domains. 
More recently, Park \etal~\cite{park2020swapping} train an autoencoder to encode real images into two independent components representing structure and texture in an unsupervised manner.
Kazemi~\etal~\cite{kazemi2018style} and Kwon \etal~\cite{kwon2021diagonal} modify the generator's architecture to explicitly learn a disentanglement of content and style.
Our work differs from these works in several key aspects. First, many works rely on human supervision such as collecting image pairs with the same style or appearance. Second, previous methods are typically more global in nature, mainly transferring texture, and may therefore struggle to model semantic-based local changes in the desired image. In contrast, our StyleFusion approach enables both local and global control via a learned disentanglement and fusion of multiple style codes.

\subsection{GAN-Based Image Editing}
While StyleGAN is able to synthesize images of phenomenal realism and diversity, one is often more interested in utilizing the trained GAN for performing image manipulations.

\paragraph{Latent Space Editing}
Most commonly, recent works perform a latent space traversal of the GAN's learned manifold for controlling a specific attribute of interest such as age, gender, and expression~\cite{abdal2020styleflow,goetschalckx2019ganalyze,harkonen2020ganspace,shen2020interpreting,shen2020closedform,jahanian2020steerability,wu2020stylespace,voynov2020unsupervised}.
There have also been numerous works exploring more diverse methods for manipulating latent representations. Tewari \etal~\cite{tewari2020stylerig} learn semantic face edits by employing pre-trained 3DMM models. Shen \etal~\cite{shen2020closedform} perform eigenvalue decomposition on the affine transformation layers of StyleGAN2 generators~\cite{karras2020analyzing} to learn versatile manipulation directions.
Xia \etal~\cite{xia2021tedigan} and Patashnik and Wu \etal~\cite{patashnik2021styleclip} manipulate images using a human-understandable text prompt providing a more intuitive image editing interface. 

Other works propose end-to-end approaches for learning a direct disentanglement and editing of real images over a specific attribute such as hairstyle~\cite{tan2020michigan,zhu2021barbershop}, facial identity~\cite{nitzan2020face,li2020faceshifter}, and other facial attributes~\cite{he2018attgan,hou2020guidedstyle,alaluf2021matter,lample2018fader}.

While the aforementioned works allow for extensive manipulations of images, most rely on the existence of a well-behaved, disentangled latent space, which is still difficult to achieve in practice. 
For example, it is not necessarily trivial to modify an individual's hairstyle by manipulating the image's latent representation without altering other facial features since these works operate over a single latent code controlling the entire image.
Therefore, a disentangled representation is of great importance for the success of these methods.

It is important to emphasize that our work is not designed to compete with the aforementioned editing techniques. Instead, one may view our work as complementing these existing approaches. For example, as shall be shown, pairing StyleFusion with GANSpace~\cite{harkonen2020ganspace} or StyleCLIP~\cite{patashnik2021styleclip} leverages their diverse manipulations while ensuring that the resulting edits alter only the desired semantic regions.

\paragraph{Editing Local Regions}
Some works propose methods specifically aimed at editing local regions of the image.
Collins \etal ~\cite{collins2020editing} blend two style codes to perform local semantic editing guided by a target spatial segment of the image.
Alhabri \etal~\cite{alharbi2020disentangled} achieve a grid-based spatial disentanglement by injecting structured noise into a GAN.
Chai \etal ~\cite{chai2021latent} design a latent regressor capable of fusing multiple image patches into a single coherent image. Yet, their approach requires manual region selection from each input image.
Bau \etal~\cite{bau2021paint} combine an input text and a mask to edit specific image regions.
These works suffer from the same drawbacks in that require a manual per-image marking of the region to be altered.
Conversely, our approach requires only specifying a semantic region from a pre-defined set, providing easier and more intuitive user control.
Finally, similarly to our approach, Zhu \etal~\cite{zhu2020sean} encode images using a set of style codes, each controlling a specific image region. However, their method requires collecting pairs of corresponding images and segmentation maps for learning this disentanglement. 

A particular task that has achieved increased attention is that of \textit{hair transfer} where we wish to transfer the hairstyle from one individual to another. Due to the entanglement of the hair region with other image attributes (e.g., pose, lighting), faithfully solving this task remains challenging. Tan \etal~\cite{tan2020michigan} design a conditional GAN for learning to disentangle hair into structure and appearance. Saha \etal~\cite{saha2021loho} perform latent vector optimization to explicitly disentangle hair attributes. More recently, Zhu \etal~\cite{zhu2021barbershop} perform mixing between the latent representations of a reference and target image, guided by a target segmentation map. 
However, all aforementioned works rely on external supervision (e.g., segmentation masks) or a specialized inpainting mechanism to guide the learned disentanglement and region transfer. 
Using our learned disentangled representation, we demonstrate StyleFusion's ability to faithfully transfer the hair region, even though it was not explicitly trained to do so.
\section{Preliminaries}

In recent years, StyleGAN \cite{karras2019style, karras2020analyzing, Karras2020ada} has become regarded as the state-of-the-art image generator.
In addition to the standard Gaussian latent space $\mathcal{Z}$, StyleGAN has a learnt latent space $\mathcal{W}$, which is derived from the initial latent space $\mathcal{Z}$ via a fully-connected mapping network. Many works have demonstrated that $\mathcal{W}$ is able to better encode semantic properties which provide control over the generated image~\cite{karras2019style,shen2020interpreting,collins2020editing}.
In the standard StyleGAN architecture, a latent code $w \in \mathcal{W}$ is additionally transformed via a different learned affine transformation at each input layer of the generator. 
The outputs of these affine transformations are then inserted into the generator's synthesis network through style modulation layers at different resolutions to generate the output image.

The space spanned by the output of the affine transformations, referred to as the StyleSpace or the $\mathcal{S}$ space, has recently been explored~\cite{wu2020stylespace,liu2020style}. Assuming a generator that outputs images at a $1024\times1024$ resolution, the $\mathcal{S}$ space has $9,088$ dimensions in total, where $6,048$ dimensions are applied
over the feature maps, and $3,040$ additional dimensions are applied over the tRGB blocks. We refer the reader to \cite{wu2020stylespace} for a more detailed description of the $\mathcal{S}$ space.
Wu \etal~\cite{wu2020stylespace} demonstrate that $\mathcal{S}$ is more disentangled than $\mathcal{W}$ and is therefore better suited for providing fine-grained control over the generated images. 

Collins~\etal~\cite{collins2020editing}, demonstrate that a per-channel linear interpolation between two style codes allows one to transfer local regions between the two corresponding images. In this work, we refer to this per-channel interpolation as \textit{fusion}. 
We employ two ideas presented in \cite{collins2020editing}. First, given an image generated by StyleGAN, it is possible to segment the image into semantic regions by clustering the activation vectors at a given layer.
Second, given a partitioning of an image into $K$ semantic regions, Collins~\etal~\cite{collins2020editing} define a matrix which measures the correspondence between a style code channel $c$ and a semantic region $k$ of the image. 

This matrix has been shown to be useful for various editing methods \cite{collins2020editing, lewis2021vogue}, and is defined as follows:
\begin{equation}
M_{k,c} = \frac{1}{NWH} \sum_{n, h, w} {A^2_{n, c, h, w} \cdot U_{n, k, h, w}}.
\end{equation}
Here, $N$ is the number of images over which the matrix is calculated, $A \in \mathbb{R}^{N \times C \times H \times W}$ is the activation tensor at a given layer of StyleGAN, and $U \in \{0, 1\}^{N \times K \times H \times W}$ specifies the region membership of each spatial location to a segmented region. Note that both $A^2_{n, c, h, w}$ and  $U_{n, k, h, w}$ are scalars and that $M_{k,c}$ is normalized such that 
$M_{k,c} = 1$ denotes that the influence of channel $c$ is fully contained in region $k$.
Similarly, $M_{k,c} = 0.5$ means that $c$ has equal influence over both $k$ and its complement.
Finally, $M_{k,c} = 0$ indicates that $c$ has no influence over the image region $k$.

\section{Method}

\begin{figure}
    \setlength{\tabcolsep}{1pt}
    \centering
        \begin{tabular}{c c c c c}
        \includegraphics[width=0.095\textwidth]
        {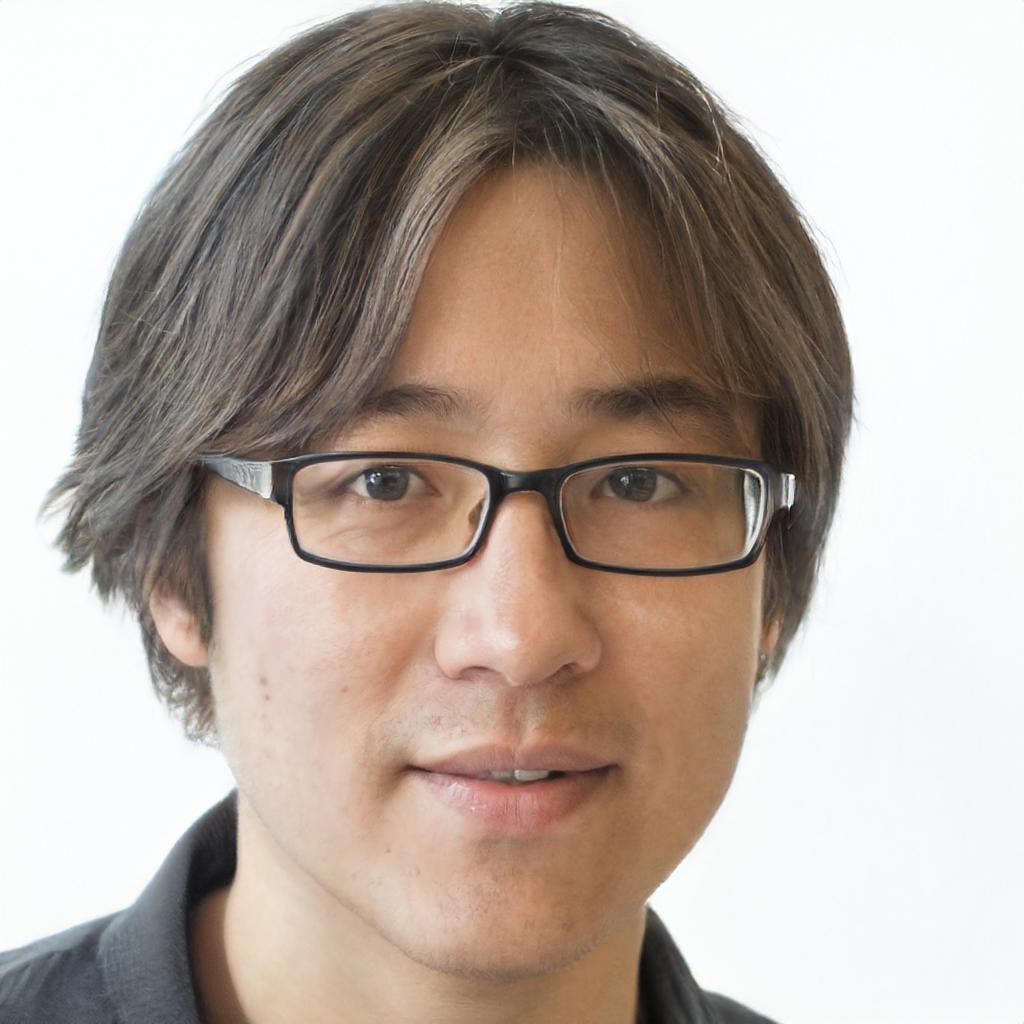} &
        \includegraphics[width=0.095\textwidth]
        {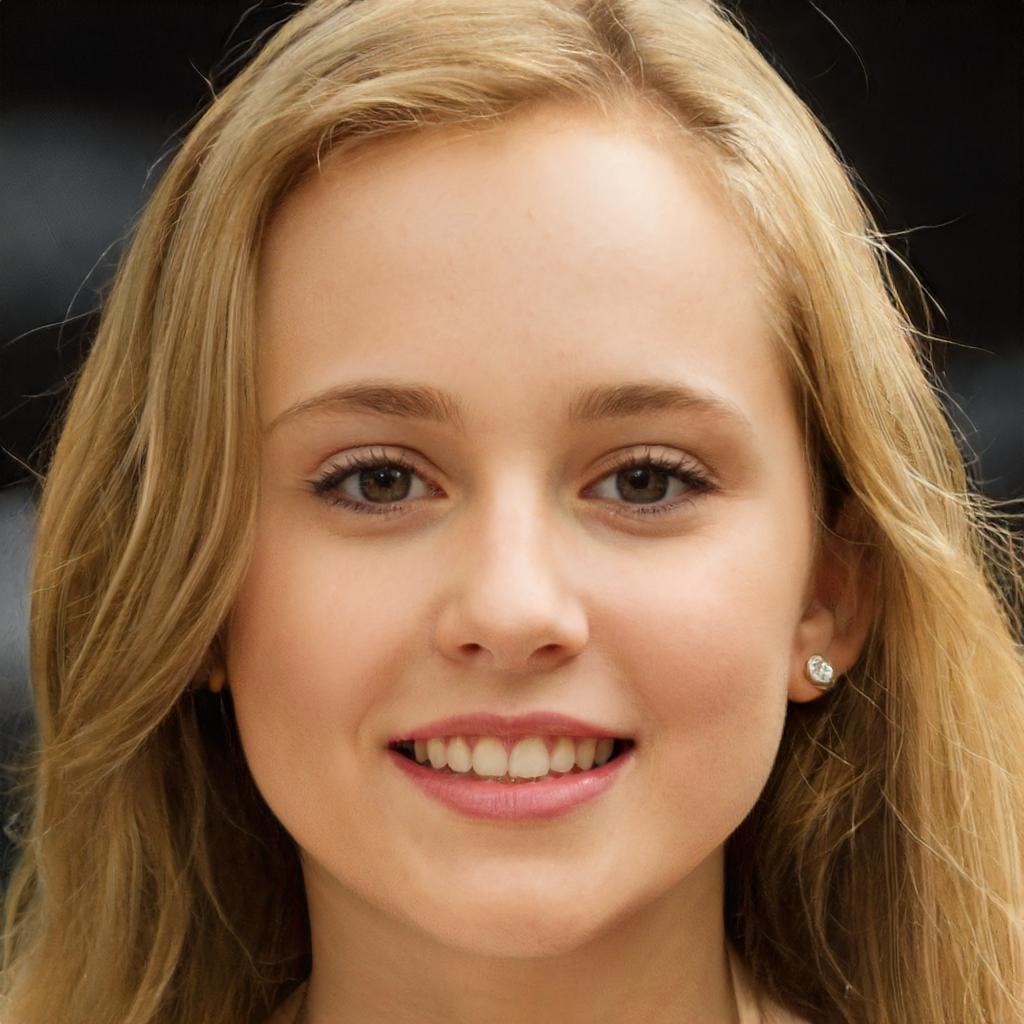} &
        \includegraphics[width=0.095\textwidth]
        {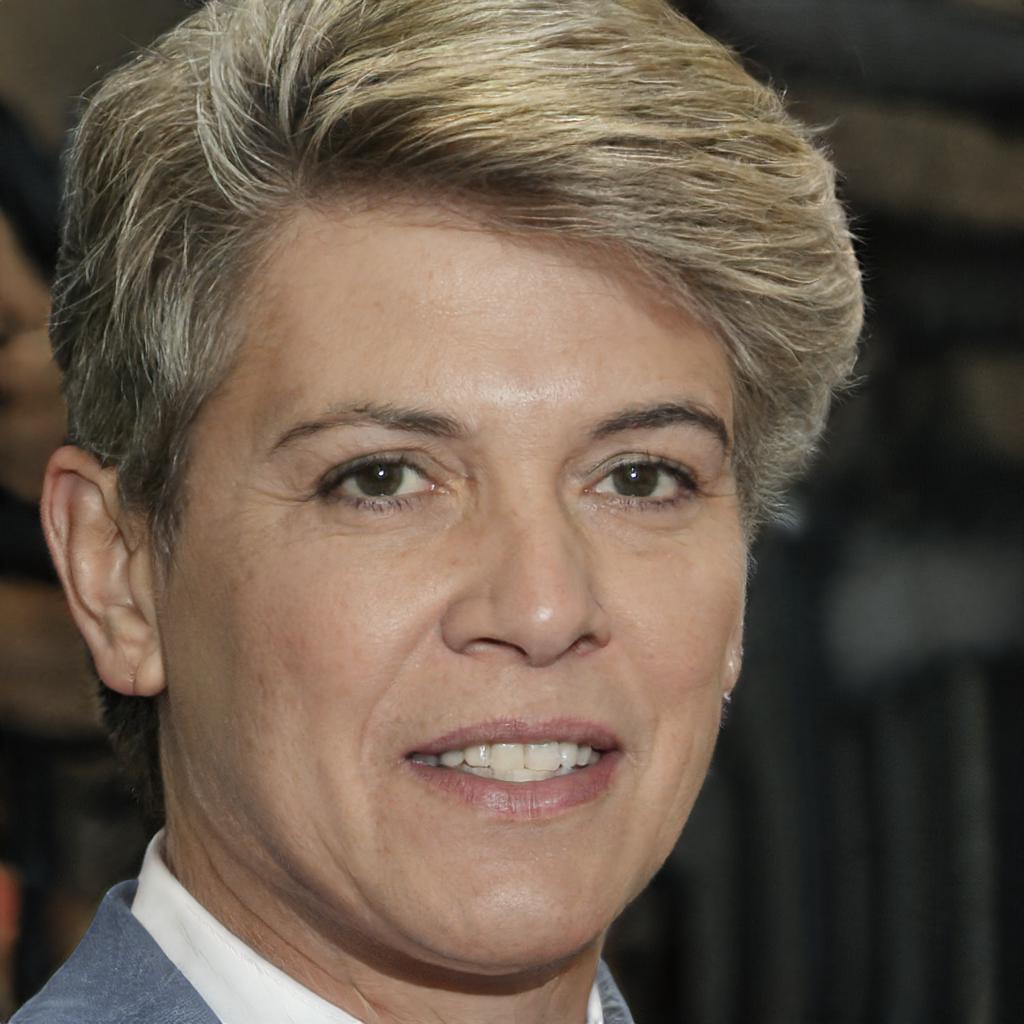} &
        \includegraphics[width=0.095\textwidth]
        {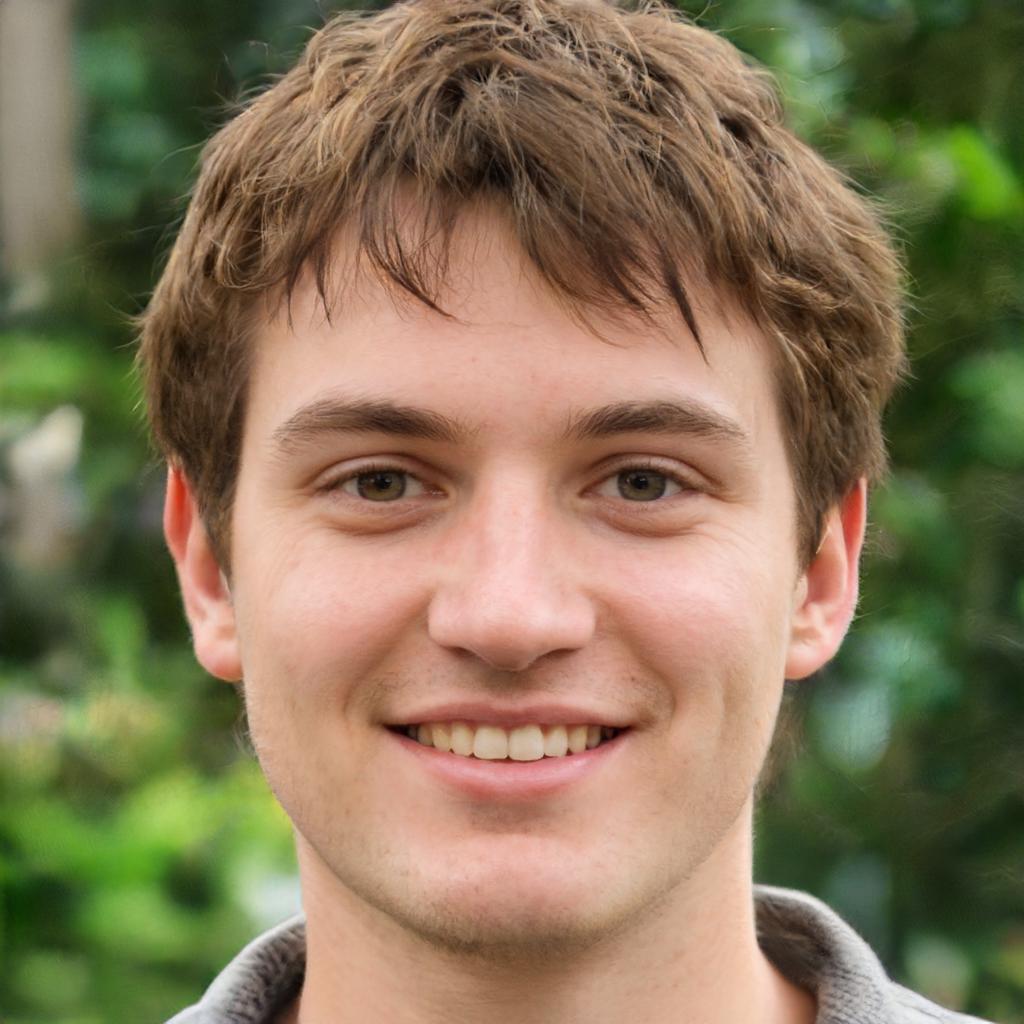} &
        \includegraphics[width=0.095\textwidth]
        {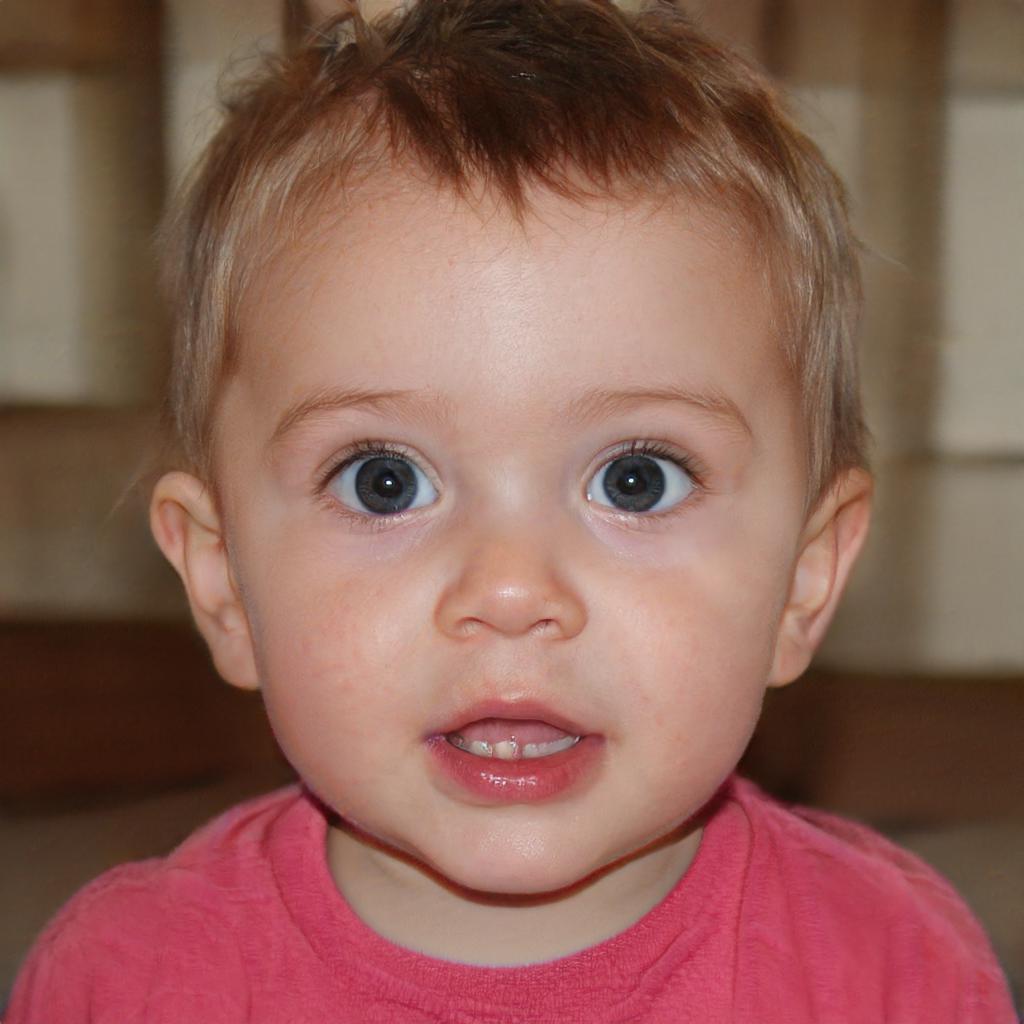}
        \tabularnewline
        \includegraphics[width=0.095\textwidth]
        {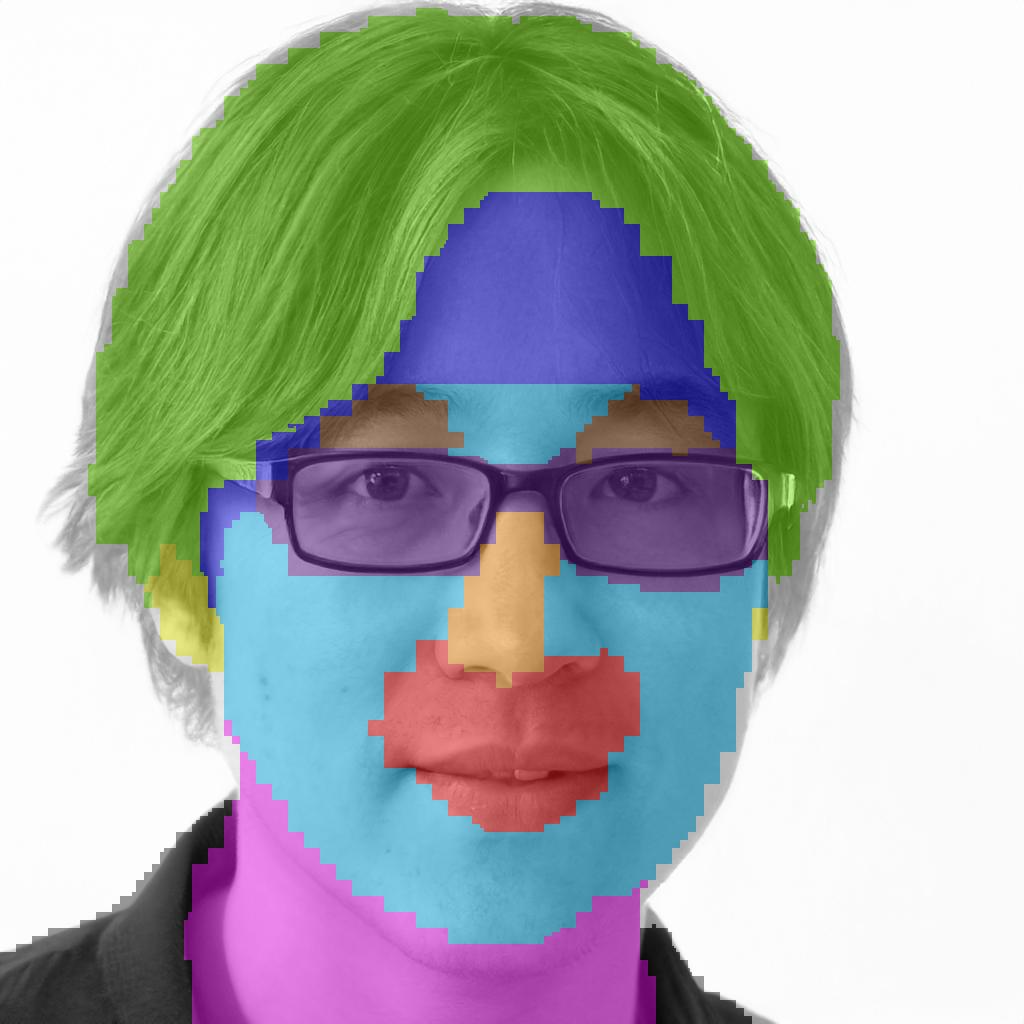} &
        \includegraphics[width=0.095\textwidth]
        {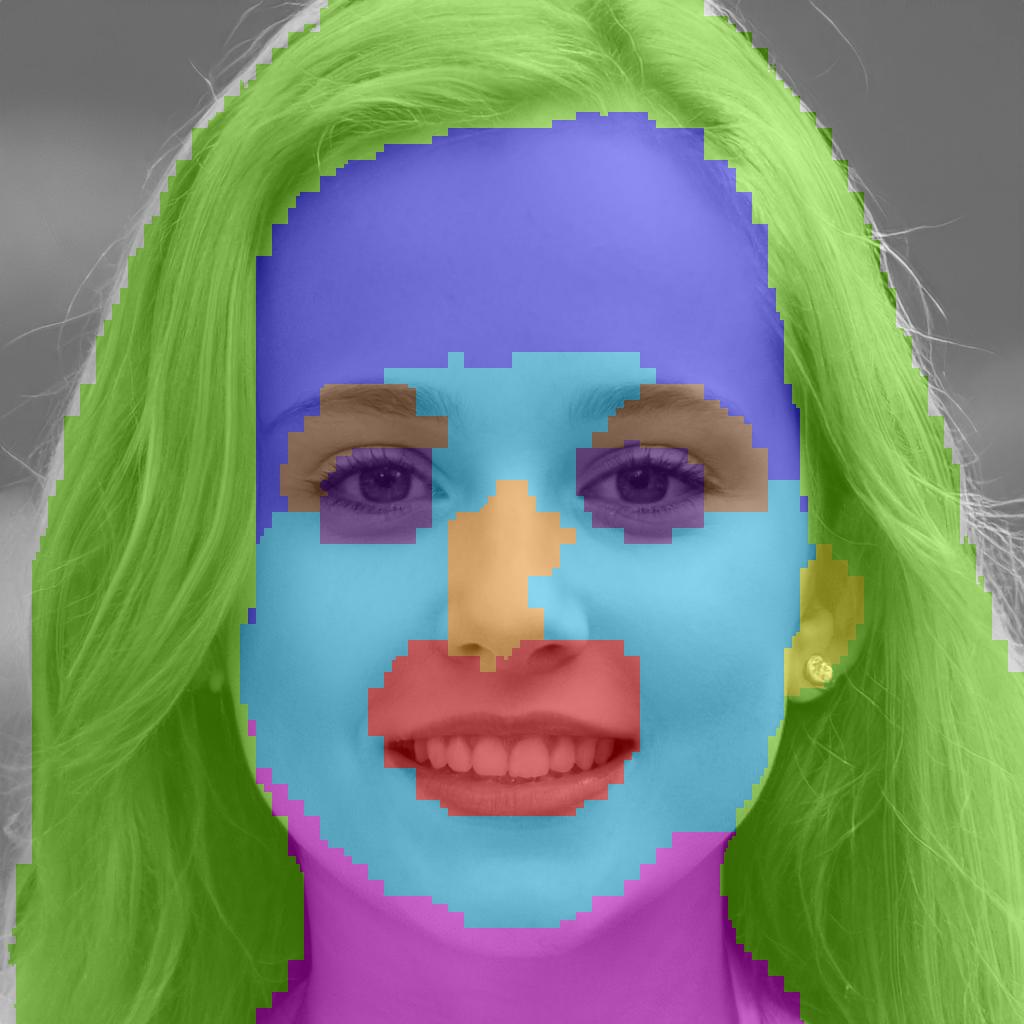} &
        \includegraphics[width=0.095\textwidth]
        {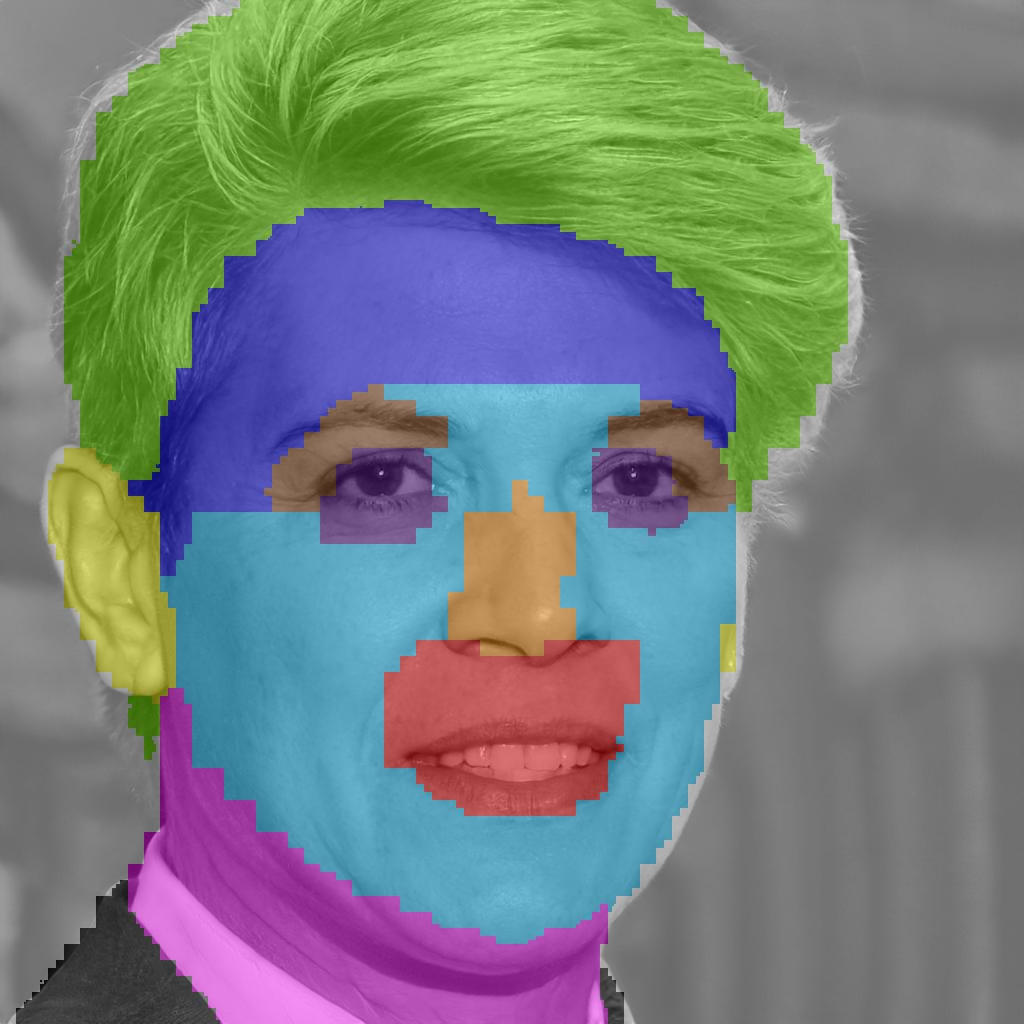} &
        \includegraphics[width=0.095\textwidth]
        {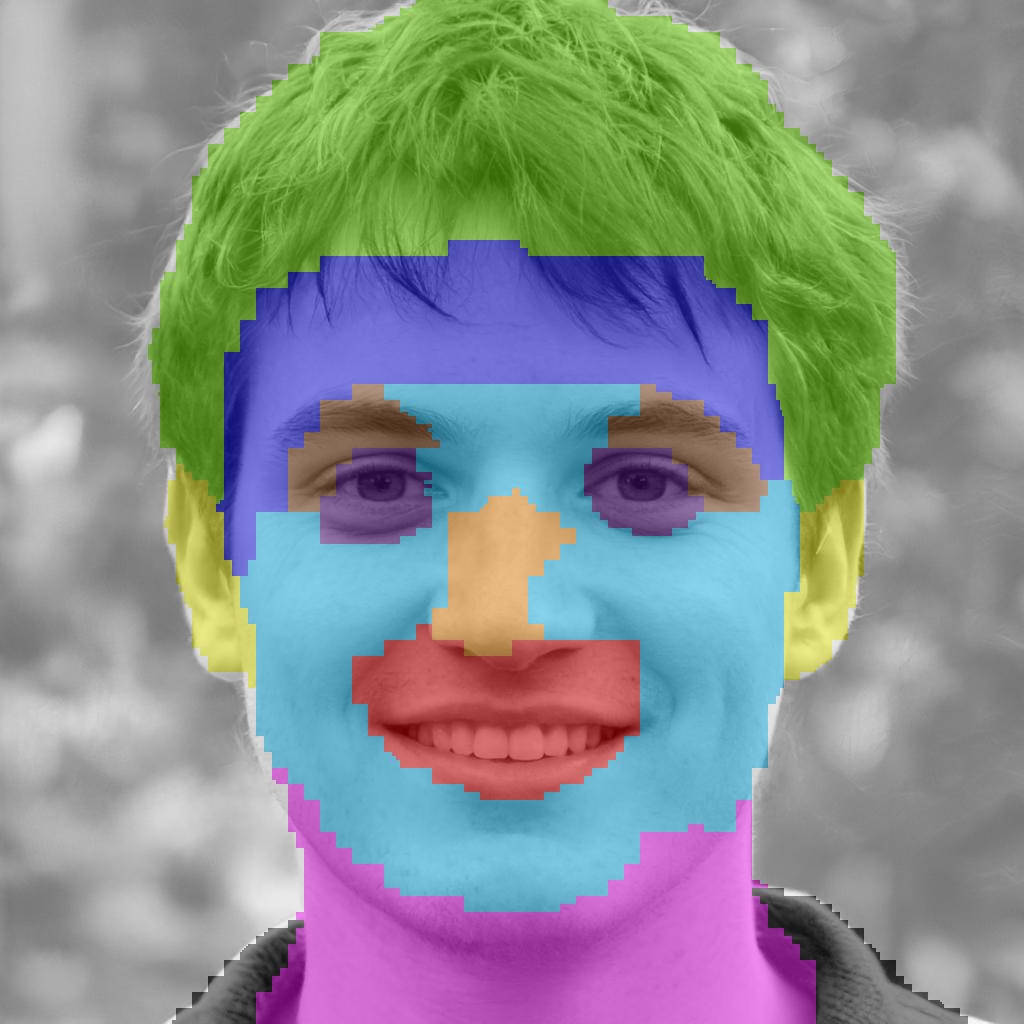} &
        \includegraphics[width=0.095\textwidth]
        {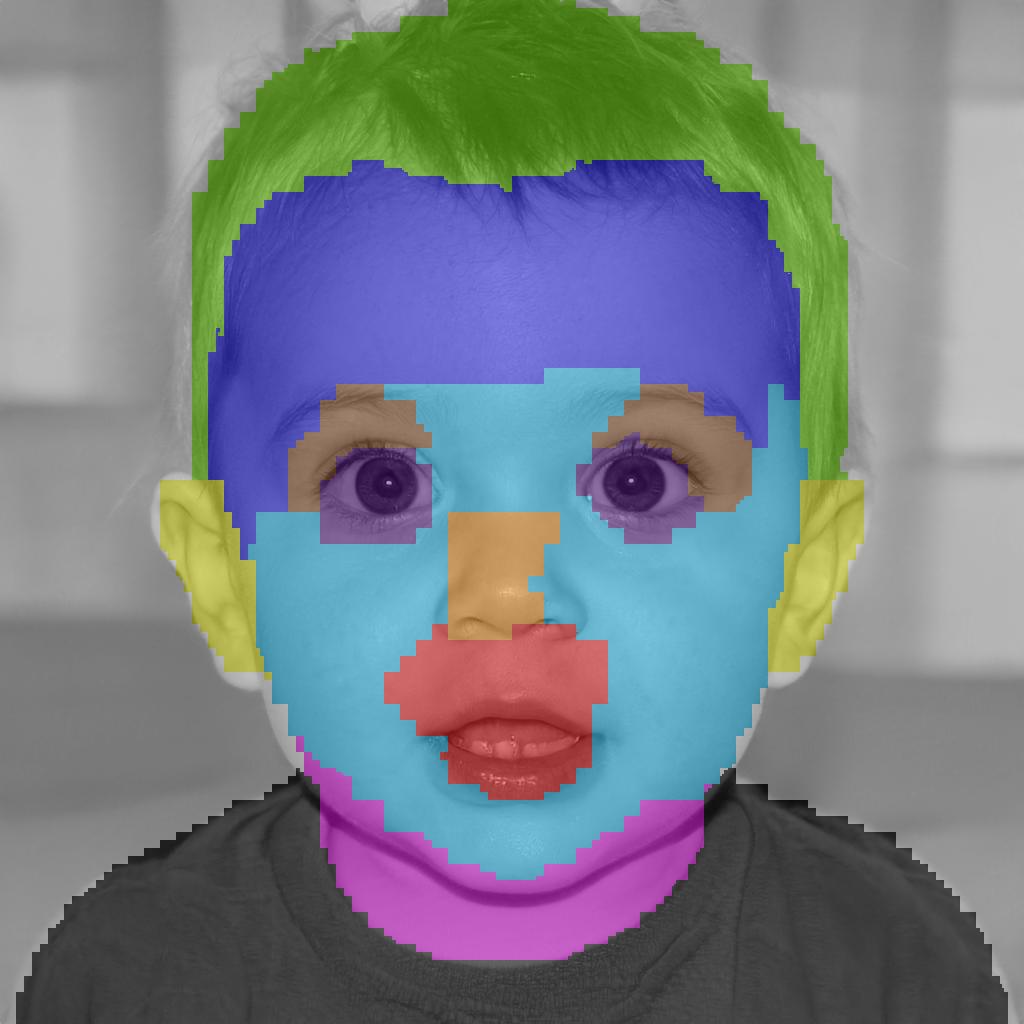}
        \tabularnewline
        \includegraphics[width=0.095\textwidth]
        {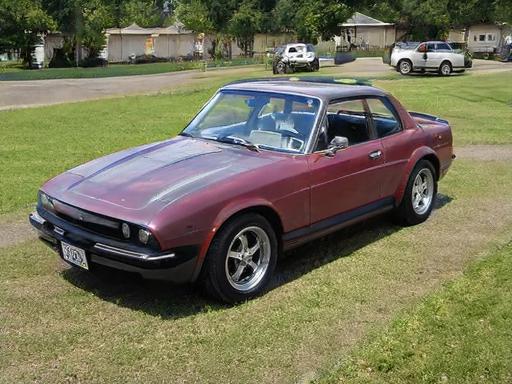} &
        \includegraphics[width=0.095\textwidth]
        {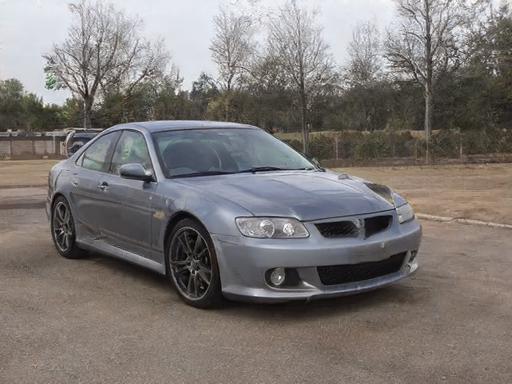} &
        \includegraphics[width=0.095\textwidth]
        {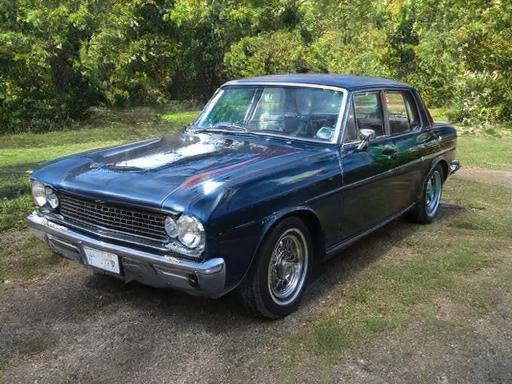} &
        \includegraphics[width=0.095\textwidth]
        {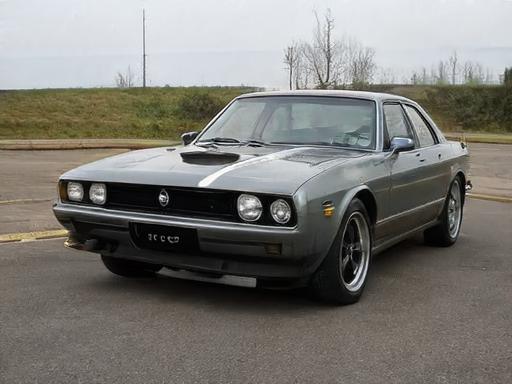} &
        \includegraphics[width=0.095\textwidth]
        {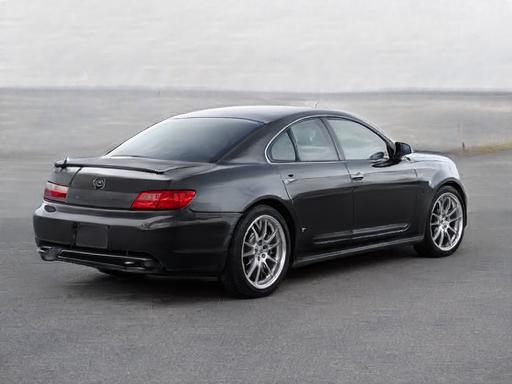}
        \tabularnewline
        \includegraphics[width=0.095\textwidth]
        {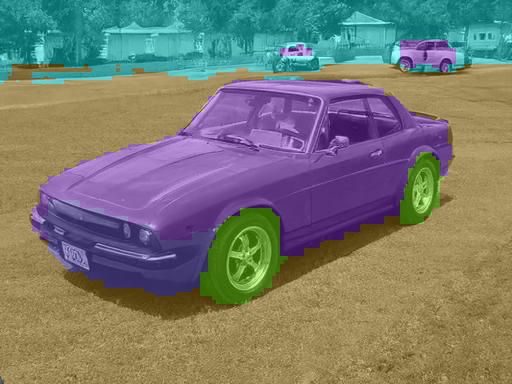} &
        \includegraphics[width=0.095\textwidth]
        {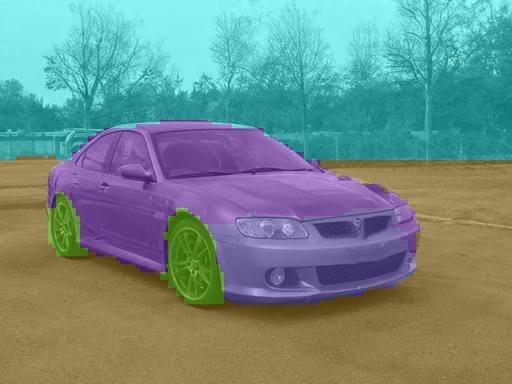} &
        \includegraphics[width=0.095\textwidth]
        {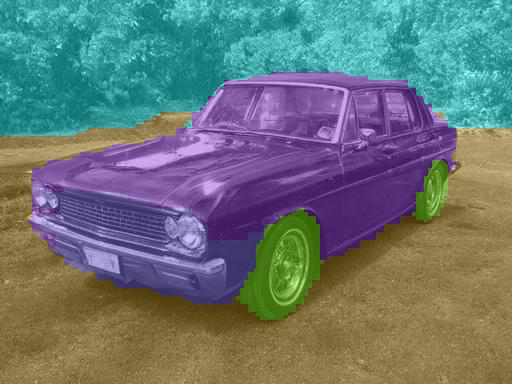} &
        \includegraphics[width=0.095\textwidth]
        {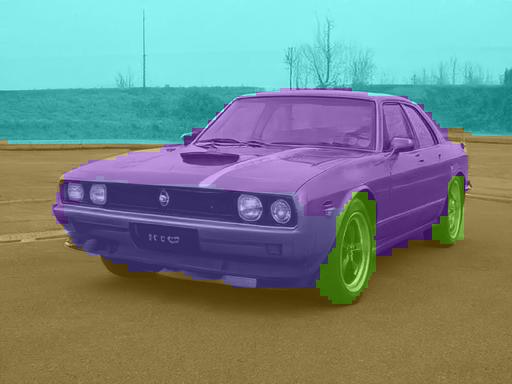} &
        \includegraphics[width=0.095\textwidth]
        {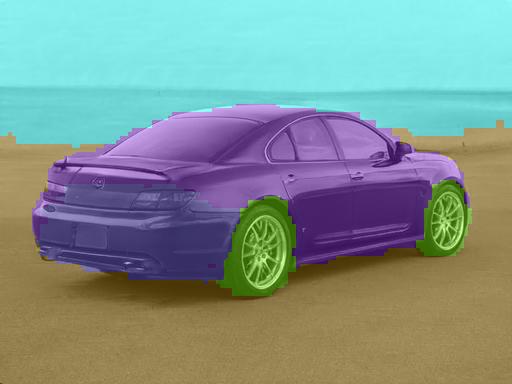}
        \tabularnewline
        \end{tabular}
    \caption{\textit{The semantic image regions of StyleFusion.} Here, we illustrate the different semantic image segments computed by StyleFusion to learn a disentangled representation on the human facial and cars domains. 
    }
    \label{fig:semantic_clusters}
\end{figure}

\begin{figure*}
    \centering
    \includegraphics[width=0.955\linewidth]{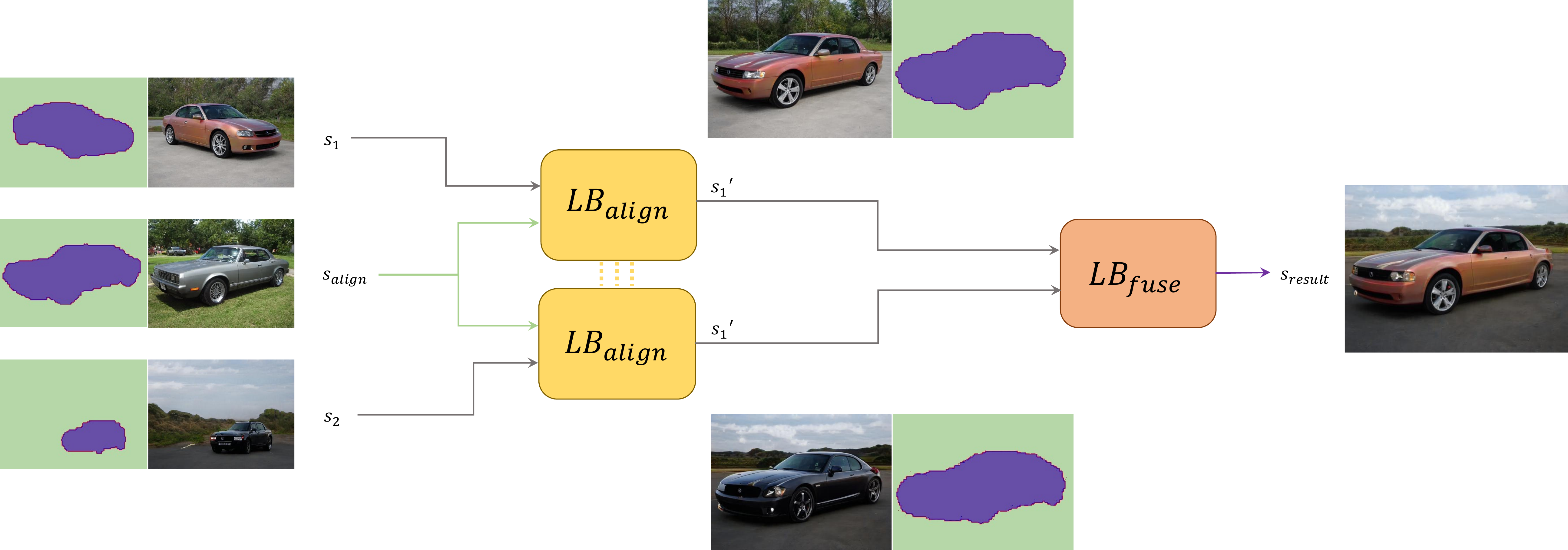}
    \vspace{0.1cm}
    \caption{
    \textit{The FusionNet architecture.} The FusionNet learns to disentangle between two specified semantic image regions (e.g., the car body and image background).
    The FusionNet, composed of a pair of building blocks named \textit{Latent Blender} (LB), receives as input three input latent codes $s_1,s_2,s_{align}\in\mathcal{S}$. 
    The first block, $LB_{align}$ aligns the images of $s_1$ and $s_2$ to match the spatial location of the image regions of $s_{align}$.
    Observe how the images of the outputted latent codes $s_1'$ and $s_2'$ match the spatial layout of $s_{align}$, namely the car pose and background layout, as illustrated by the segmentation mask shown to the right of each image.
    Observe that while $LB_{align}$ operates on $s_1$ and $s_2$ independently, its network weights are shared between the two.
    Having aligned the two images, the second block, $LB_{fuse}$, is then used to fuse the two semantic regions into a unified image. 
    Specifically, notice how the car body in the resulting image is taken from $s_1$, the background is taken from $s_2$, and the spatial layout and pose are controlled by $s_{align}$.
    }
    \label{fig:blendnet_architecture}
\end{figure*}

\subsection{Overview} 
We start by presenting a high-level overview of our approach for learning a disentangled latent representation of semantic image segments. 
Given a pre-trained StyleGAN generator, our goal is to generate an image by fusing a set of $K$ latent codes each of which controls a \textit{single} spatial, semantic segment (or region) of the generated image. 
As the generated image is defined by a learned fusion of multiple style codes, we name our approach \textit{StyleFusion}.
Intuitively, such a scheme would allow one to seamlessly control specific regions of the image without altering other regions.

We illustrate this idea in Figure~\ref{fig:overview} where five latent codes are randomly sampled from the $\mathcal{Z}$ latent space.
These latent codes are first passed through StyleGAN's mapping network and learned affine transformation layers and encoded into the generator's $\mathcal{S}$ space~\cite{wu2020stylespace}. 
The resulting generated image of each input style code is shown above as a reference to the reader. 

Having obtained the five $\mathcal{S}$-space representations of the input codes, our StyleFusion mechanism operates in a hierarchical fashion. Each component in this hierarchy, a \emph{FusionNet}, learns a spatial disentanglement of a pair of image regions by learning to disentangle and interpolate between their corresponding $\mathcal{S}$-space representations.
The output of the StyleFusion hierarchy is a single harmonized style code constructed from the five input style codes. Finally, passing this style code to the pre-trained StyleGAN generator returns the final harmonized image, shown to the right. Observe that in the harmonized image, each semantic region is controlled by a single input latent code, demonstrating StyleFusion's learned disentanglement of the regions. 

\subsection{Discovering the Semantic Image Regions}~\label{sec:segmentation}
To learn a disentanglement of the input latent codes into different semantic regions of interest, we must first segment the generated image into those segments.
We build on the segmentation approach introduced in Collins \etal~\cite{collins2020editing}. 
Consider all the activation tensors along StyleGAN's synthesis network. We first up-sample the tensors to match the size of the largest activation tensor, and then concatenate the activations to a single tensor composed of $C=6,048$ channels, corresponding to the $6,048$ channels of the $\mathcal{S}$ space.

Having obtained the joined representation, we then apply Ward's hierarchical clustering~\cite{wards1963hierarchical} where each spatial location is assigned a cluster membership.
Finally, we review the resulting cluster and label each cluster as corresponding to a certain semantic object.
In the case that multiple clusters correspond to a single attribute, we merged their regions. Notice that this process occurs one time per domain, requiring minimal human supervision (e.g., several minutes). That is, given a new generated image, the image's semantic regions can be computed without additional supervision by assigning each pixel to its closest semantic cluster.

Figure~\ref{fig:semantic_clusters} shows examples of the resulting clusters generated for the human facial and cars domains. As can be seen, the clustering process results in clusters that encode semantically-meaningful attributes (e.g., hair, eyes, mouth for the human facial domain and wheels, body, and background for the cars domain).

\subsection{FusionNet} 
The FusionNet component is tasked with learning a spatial disentangled representation over two semantic regions $R_1$ and $R_2$ (e.g., hair and face).
The FusionNet receives as input three style codes $s_1, s_2, s_{align} \in{\mathcal{S}}$ where $s_1$ and $s_2$ are designed to control the semantics associated with regions $R_1$ and $R_2$. Additionally, $s_{align}$ is designed to align the spatial location of $R_1$ and $R_2$ in the resulting fused image. Specifically, the spatial locations of the two image regions in the final fused image is determined by the segmentation of the image corresponding to $s_{align}$.

Given these three latent codes, the FusionNet is designed to output a single unified style code $s_{result}\in\mathcal{S}$ with the following properties: 
\begin{enumerate}
    \item Every semantic attribute in $R_1$ is determined by $s_1$.
    \item Every semantic attribute in $R_2$ is determined by $s_2$.
    \item The spatial location of each region and image features common to both input images are determined by $s_{align}$.
\end{enumerate}
Below we describe the FusionNet architecture followed by the training scheme and loss objectives employed for achieving these properties.

\topic{\textit{Architecture. }}
To generate a harmonized image from two latent codes $s_1$ and $s_2$, we introduce the FusionNet constructed from a pair of building blocks named \textit{Latent Blender}. The first blender, denoted $LB_{align}$ is trained to align $s_1$ and $s_2$ to a common spatial layout determined by the third latent code, $s_{align}$. 

Given the aligned latent codes, the second blender, $LB_{fuse}$ is tasked with learning to fuse the semantic regions of the two input latent codes. This idea is illustrated in Figure~\ref{fig:blendnet_architecture} where $s_1$ controls the car body and $s_2$ controls the background of the generated image, with the spatial layout (i.e., car pose) determined by $s_{align}$.
Note that while $LB_{align}$ and $LB_{fuse}$ share the same architecture (described below), they naturally differ in their role.

\begin{figure}
    \setlength{\tabcolsep}{1pt}
    \setlength{\belowcaptionskip}{-7.5pt}
    \centering
    \includegraphics[width=\linewidth]{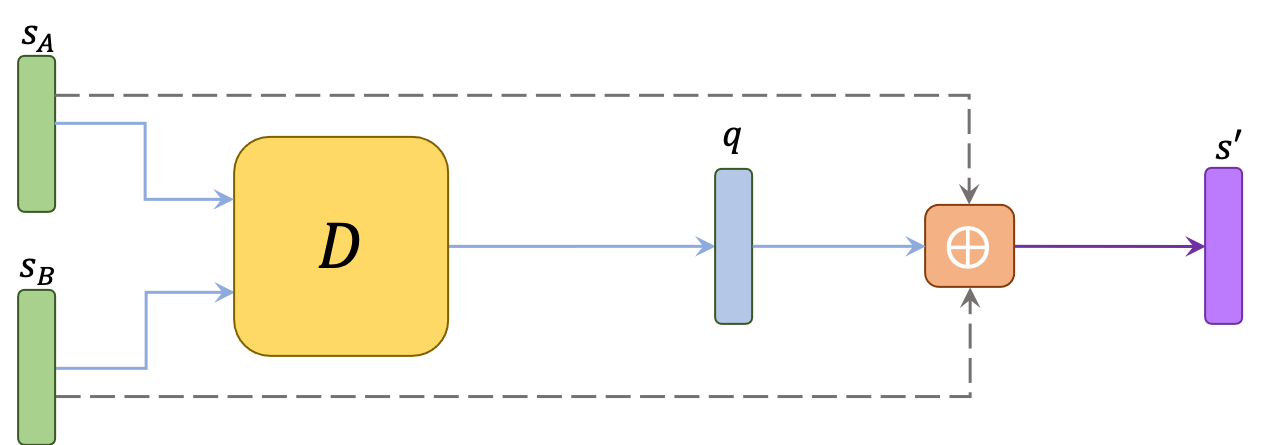}
    \vspace{-0.2cm}
    \caption{
    \textit{The Latent Blender.} Given two input latent codes $s_A,s_B\in\mathcal{S}$, the Latent Blender, composed of a fully-connected network $D$, learns a single fusion coefficient $q\in[0,1]^\mathcal{S}$ used to blend between the two input codes via a per-channel interpolation, resulting in a fused latent code $s'$. 
    }
    \label{fig:d_and_f}
\end{figure}

\topic{\textit{Latent Blender. }} 
As mentioned above, the core component of the FusionNet is the \textit{Latent Blender}, illustrated in Figure~\ref{fig:d_and_f}. Consider two input latent codes $s_A,s_B\in\mathcal{S}$, which are concatenated and passed to a simple fully-connected network, denoted $D$, followed by a sigmoid activation function. The output of this network is a single vector $q \in{[0,1]^\mathcal{S}}$, which we label the fusion coefficient vector.

This coefficient vector and the input style codes are then passed to the \textit{fusion} procedure which performs a per-channel interpolation between the two style codes, weighted by the fusion coefficient. Specifically, we compute, 
\begin{align}
\begin{split}
&s' =  q \odot s_A + (1 - q) \odot s_B,
\end{split}
\end{align}
where $\odot$ denotes a per-element multiplication.
That is, the fusion operation blends $s_A$ and $s_B$ according to $q$ and outputs a unified code $s'$.

Observe that, in practice, the size of the fully-connected network described above will be large --- its input will be of size $2|\mathcal{S}|$ and its output will be of dimension $|\mathcal{S}| = 9,088$. To address this, we propose using a layer encoding allowing the Latent Blender to consider each of the generator's input layers independently. We describe this design in detail in Appendix~\ref{sec:latent_blender_arch}.  

\subsection{Training}
Given a pre-trained generator $G$, the FusionNet is trained to disentangle between two target semantic image regions $R_1$ and $R_2$.
As a motivating example, let us consider the human facial domain and assume $R_1$ corresponds to the face region and $R_2$ corresponds to the hair region.

At each training iteration, we begin by randomly sampling four latent codes: $z_1, z_2, z_{align}, z_{rnd}\in\mathcal{Z}$.
We then pass the latent codes to StyleGAN's mapping function and affine transformation layers resulting in their corresponding style codes $s_1, s_2, s_{align}, s_{rnd}\in\mathcal{S}$. 

Given the style codes $s_1, s_2, s_{align}$, the FusionNet blends the three codes into a single style code $s_{result}$. Passing the code through StyleGAN's synthesis network, we obtain the corresponding image $I_{result}$. Finally, using the clustering procedure described in Section~\ref{sec:segmentation}, we compute the $M_{k,c}^1, M_{k,c}^2$ correlation matrices of the images of $s_1, s_2$, respectively.

Given the input $\mathcal{S}$-space representations and the outputted fused style code $s_{result}$, the FusionNet is trained using a weighted combination of several loss objectives described below. Note that during training the pre-trained generator remains fixed.

\vspace{0.20cm}
\topic{\textit{Mask Loss}}
The mask loss is tasked with aligning the two images corresponding to $s_1$ and $s_2$. This is done by demanding that the spatial regions in the fused output image will be aligned with the spatial regions of the image $I_{align}$ corresponding to $s_{align}$. 
Assume we have obtained the latent code $s_{result}$ corresponding to the generated image $I_{result}$.
We additionally define $m_1(I_{result}), m_2(I_{result})$ to denote the binary masks of image regions $R_1$ and $R_2$ in the outputted image $I_{result}$. Specifically, $m_{i}(I_{result})$ takes a value of $1$ in all pixels inside image region $R_{i}$ and $0$ elsewhere. 

The mask loss is then defined as,
\begin{equation*}
d_1 = \norm{ m_{1}(I_{result}) - m_1(I_{align}) }_1 \\
\end{equation*}
\begin{equation}
d_2 = \norm{ m_{2}(I_{result}) - m_{2}(I_{align}) }_1 \\
\end{equation}
\begin{equation*}
\begin{split}
\mathcal{L}_{mask} = {}&
\dfrac
    {d_1 + d_2}
    {\norm{ m_1(I_{align}) + m_2(I_{align}) }_1}.
\end{split}
\end{equation*}
Here, we align the masks corresponding to regions $R_1,R_2$ in the fused image $I_{result}$ according to the regions' spatial locations in the image $I_{align}$.
Consider our motivating example of fusing the face and hair regions. This loss encourages the FusionNet to align the face and hair segments of the two input latent codes according to the spatial location of the hair and face regions in the image $I_{align}$.

Observe that in order to compute this loss, one must calculate the masks in a differentiable manner. Details on doing so are provided in the Appendix~\ref{sec:mask_loss}.

\topic{\textit{Localization Loss. }}
The core idea behind the disentanglement process is to demand that the latent code $s_1$ does \textit{not} affect the semantic region $R_2$ of the generated image, and vice-versa. 
To encourage the FusionNet to only interpolate between style channels affecting the region of interest, we introduce a localization loss as follows. Given the styles $s_1$ and $s_2$, we insert small perturbations into the style codes according to $s_{rnd}$: 
\begin{align}
\begin{split}
\widetilde{s_1} = s_1 + \varepsilon\cdot(s_{rnd} - s_1), \\
\widetilde{s_2} = s_2 + \varepsilon\cdot(s_{rnd} - s_2),
\end{split}
\end{align}
where $\varepsilon$ is a scalar denoting the perturbation strength.
Given $\widetilde{s_1},\widetilde{s_2}$, we then generate two images in addition to $I_{result}$ introduced above:
\begin{enumerate}
    \item $I_{\widetilde{1}}$ by fusing the style codes $s_{\widetilde{1}}, s_2, s_{align}$ 
    \item $I_{\widetilde{2}}$ by fusing the style codes $s_1, s_{\widetilde{2}}, s_{align}$.
\end{enumerate}
We additionally generate the two binary masks $m_1(I_{result}), m_2(I_{result})$ as defined in the mask loss above. 
The localization loss is then computed as: 
\begin{align}
\begin{split}
\mathcal{L}_{local} = 
&\dfrac{
    \norm{ m_2(I_{result}) \cdot I_{result} - m_2(I_{result}) \cdot I_{\widetilde{1}} }_2 }
    {\norm{ m_2(I_{result}) }_1} \;\; + \\
&    
\dfrac{
    \norm{ m_1(I_{result}) \cdot I_{result} - m_1(I_{result}) \cdot I_{\widetilde{2}} }_2 }
    {\norm{ m_1(I_{result}) }_1}.
\end{split}
\end{align}
Observe that we divide the losses of the two regions by their spatial size to normalize the loss. 
Consider our motivating example of disentangling between the face region ($R_1$) and the hair region ($R_2$). This loss ensures that making small changes to the face style code $s_1$ does not change the hair region of the outputted image (i.e., $ m_2(I_{result}) = m_2( I_{\widetilde{1}})$). Likewise, making small changes to $s_2$ should not result in changes to the image's face region. 

\vspace{0.20cm}
\topic{\textit{Common Regularization Loss. }}
Consider the first Latent Blender, $LB_{align}$, within the FusionNet that is tasked with aligning the two input latent codes $s_1,s_2$ according to the alignment code $s_{align}$. 
Consider a scenario in which $LB_{align}$ outputs the all-ones vector for the fusion coefficient $q_{align}$. In such a case, the fusion procedure will then return $s_1' = s_2' = s_{align}$ for any input. 
As a result, the intermediate images $I_1'$ and $I_2'$ will both be equal to $I_{align}$ and the final fused image will also be equal to $I_{align}$ To avoid this trivial solution to the fusion procedure, we constrain the magnitude of the fusion coefficient of $LB_{align}$,
\begin{align}
\begin{split}
&\mathcal{L}_{alignReg} = \norm{ q_{align} }_1.
\end{split}
\end{align}

\vspace{0.20cm}
\topic{\textit{Fusion Loss. }}
To faithfully transfer each image region $R_i$ from its source image to the final fused output image, we define weight vectors $w_{R_1},w_{R_2} \in [0,1]^{|S|}$. For a given style channel $c$, $w_{R_i}(c)$ represents the contribution of channel $c$ to the region $R_i$. Using the weight vectors, we define the fusion loss as, 
\begin{equation}
\begin{split}
\small
\mathcal{L}_{fusion} = {}&
\norm{ w_{R_1} \odot (s_{result} - s_1) }_2 + \\
&\norm{ w_{R_2} \odot (s_{result} - s_2) }_2 
\end{split}
\end{equation}
where we define $w_{R_i} = [w_{R_i}(c_0), w_{R_i}(c_1), ..., w_{R_i}(c_{|S|})]$.
Back to our example of disentangling between the face region ($R_1$) and hair region ($R_2$), this loss is designed to ensure that channels in $s_1$ that influence the face region are transferred to the fused code $s_{result}$ while channels in $s_2$ that control the hair region are transferred to $s_{result}$.

We now turn to describe how each of the weight vectors $w_{R_i}$ is computed for region $R_i$ and corresponding code $s_i$. 
As described above, the correlation between a style code channel $c$ and an image region $k$ is encoded in a matrix $M_{k,c}$.
Specifically, $M_{k,c}$ measures what portion of the activations of channel $c$ are located in the semantic cluster corresponding to the image region $k$. 

Consider the image of the latent code $s_i$. We compute the image's corresponding correlation matrix for region $R_i$ (i.e., $M_{R_i,c}$ for all channels $c$). The weight vector for region $R_i$ is then defined by, 
\begin{equation}
w_{R_i}(c) = 2 \cdot \max(M_{R_i,c},0.5) - 1.
\end{equation}
Note that these weights rank the contribution of each channel $c$ to $R_i$. 
In words, if $c$ has a strong contribution over $R_i$ (i.e., $M_{R_i,c} > 0.5$), we set $w_{R_i}(c)>0$. Conversely, if $c$ does not contribute to the image region $R_i$, we set $w_{R_i}(c)=0$. 

\vspace{0.20cm}
\topic{\textit{Total Loss. }}
In summary, the complete loss objective is given by: 
\begin{equation}
\begin{split}
\mathcal{L} = {}& \lambda_{mask} \mathcal{L}_{mask} + \lambda_{local} \mathcal{L}_{local} + \\
& \lambda_{alignReg} \mathcal{L}_{alignReg} + \lambda_{fusion} \mathcal{L}_{fusion},
\end{split}
\end{equation}
where $\lambda_{mask},\lambda_{local},\lambda_{alignReg},\lambda_{fusion}$ denote the loss weights. 
Additional implementation details and hyper-parameters are provided in Appendix~\ref{sec:appendix_losses}.

\begin{figure*}
    \centering
    \includegraphics[width=\linewidth]{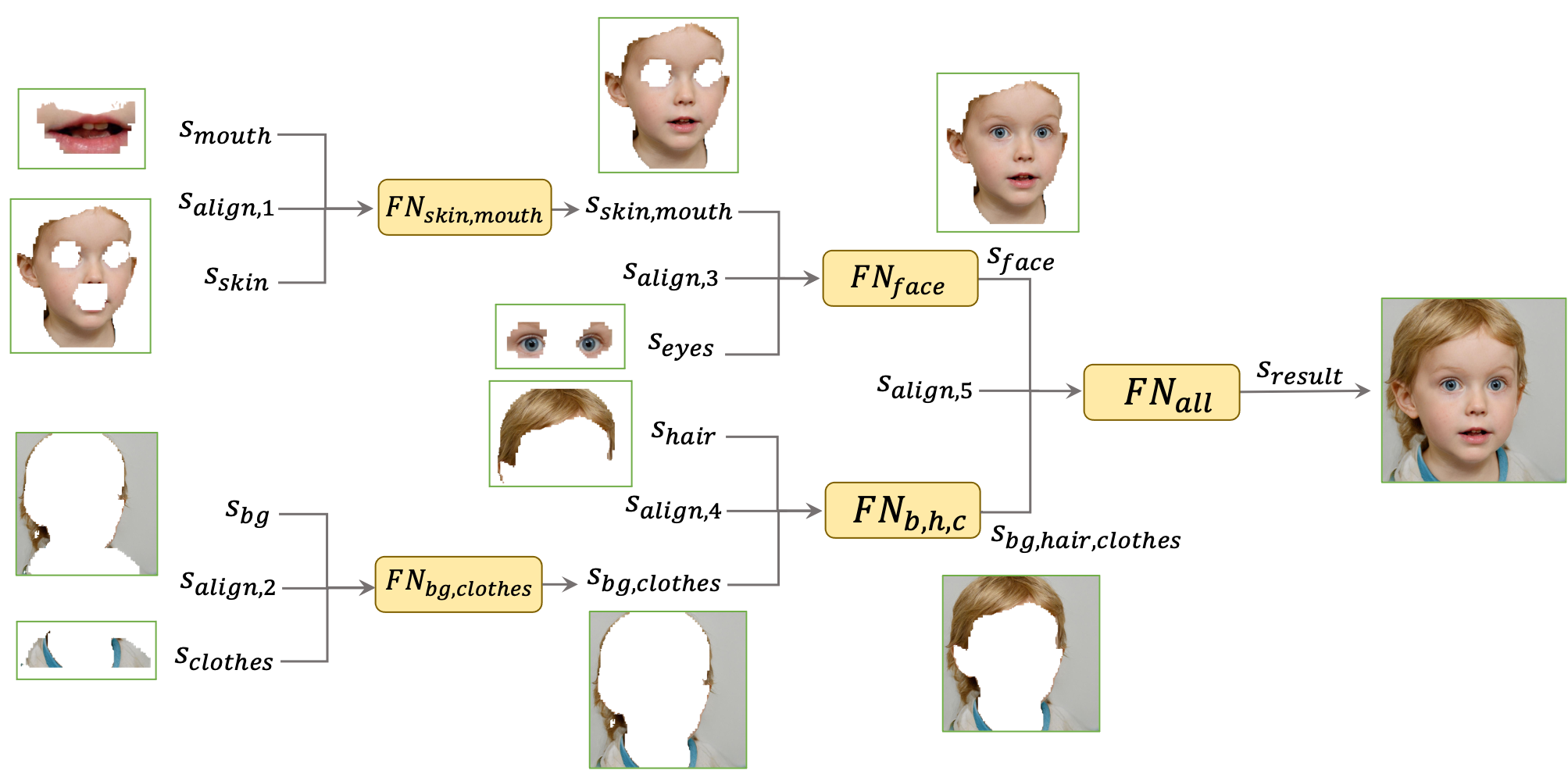}
    \caption{\textit{Hierarchical Multi-Fusion.} By constructing a tree of FusionNets (FN), StyleFusion is able to disentangle multiple image regions in a coarse-to-fine fashion. As shown, given a set of input codes, each FusionNet learns to disentangle then fuse two image regions which are passed as input to the next FusionNet in the hierarchy. The final output is a single unified image composed of all input codes. The hierarchies constructed for all domains are provided in Appendix~\ref{sec:fusionnet_hierarchies}.}
    \label{fig:hierarchy}
\end{figure*}

\vspace{0.20cm}
\topic{\textit{Multi-Step Training.}}~\label{multi-step-training}
To improve the results and provide a more stable training process, we use a multi-step training procedure. Specifically, we split the training process into three stages as follows: 
\begin{enumerate}
    \item Only the first Latent Blender component, $LB_{align}$, which is tasked in aligning the two input latent codes, is trained.
    In this stage, the FusionNet does not use the $LB_{fuse}$ component. We instead employ only $\mathcal{L}_{mask}$, $\mathcal{L}_{alignReg}$, and $\mathcal{L}_{fusion}$ on the intermediate outputs $s_1'$ and $s_2'$ and their corresponding generated images.
    \item Here, only the second Latent Blender component, $LB_{fuse}$ is trained. This stage focuses on identifying the attributes we wish to disentangle, given the aligned latent codes. To achieve this, we apply only $L_{fusion}$ on the intermediate latent codes $s_1'$ and $s_2'$.
    \item In the final stage, both components are trained simultaneously using $\mathcal{L}_{local}$, $\mathcal{L}_{alignReg}$, and $\mathcal{L}_{fusion}$.
\end{enumerate}

\subsection{Hierarchical Multi-Fusion}~\label{sec:hierarchical_fusion}
To obtain a disentanglement of multiple image regions, we build a hierarchy of FusionNets. Specifically, we model the spatial regions of the image as a binary tree, in which each node corresponds to a single region of the image.
The two children of each node correspond to the partition of the region into two sub-regions (e.g., the face to the eyes and mouth). Note that the root of the tree corresponds to the entire image. 

As such, we build a set of FusionNets, one for each node in the above tree. We then combine the FusionNets into a hierarchy as illustrated in Figure~\ref{fig:hierarchy}. Observe that the training of the hierarchy is done sequentially, starting with the root FusionNet. Once the parent node is trained, we begin training the child nodes with the parent node remaining fixed. Observe further that the tree need not be balanced, providing users the ability to control the granularity of the learned disentanglement into disjoint image regions. In a sense, this hierarchy provides users a plug-and-play interface for generating a unified image composed of multiple semantic regions.

For example, consider the human facial domain, shown in Figure~\ref{fig:hierarchy}. The first FusionNet, $FN_{all}$, is trained to disentangle between the face and the remaining image features (background, hair, and clothes) using input style codes $s_{face}$ and $s_{bg,hair,clothes}$. Once trained, $FN_{all}$ becomes fixed. We can then move to disentangling the face into multiple facial regions by introducing two additional codes $s_{eyes}$ and $s_{skin,mouth}$ and training an additional FusionNet, $FN_{face}$. 
Similarly, we can disentangle the image background and hair into two sub-parts using a third FusionNet, $FN_{b,h,c}$.
We can continue this process and move down the constructed tree with each level resulting in an additional learned disentanglement.

After training is complete, at inference time, we utilize a set of input latent codes for controlling each of the semantic regions (mouth, skin, eyes, hair, etc.) and additional latent codes (i.e., the global codes) for controlling the spatial layout of each region in the outputted facial image. 
Note that users may also control the granularity of the composition post-training. For example, users may take the mouth and skin from the same image by bypassing the FusionNet labeled $FN_{skin,mouth}$ and passing the desired input code directly to $FN_{face}$. 

The FusionNet hierarchies for all domains are provided in Appendix~\ref{sec:fusionnet_hierarchies}.
\begin{figure}
    \centering
    \setlength{\tabcolsep}{0pt}
    {\small
        \begin{tabular}{c c c c c c}
        \raisebox{0.30in}{\rotatebox[origin=t]{90}{Hair}} & 
        \hspace{1pt} &
        \includegraphics[width=0.23\linewidth]
        {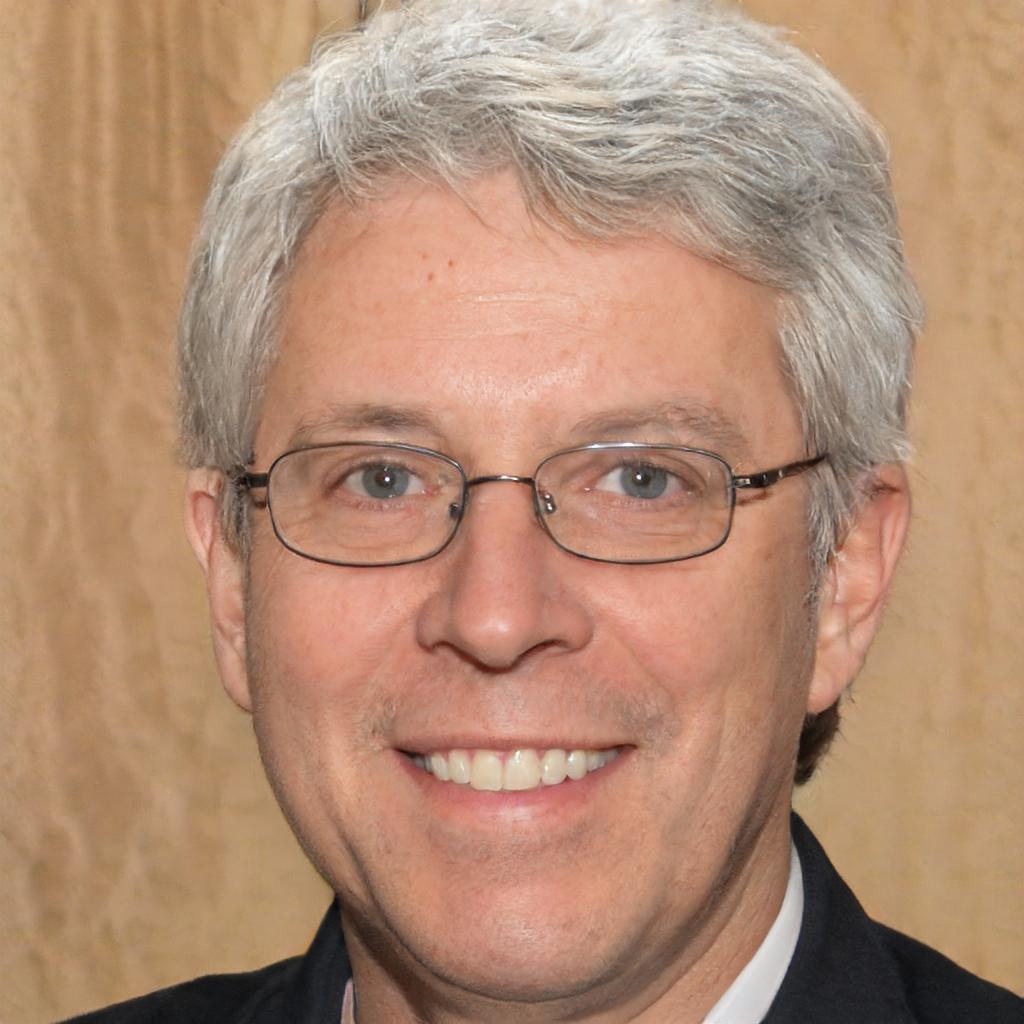} &
        \includegraphics[width=0.23\linewidth]
        {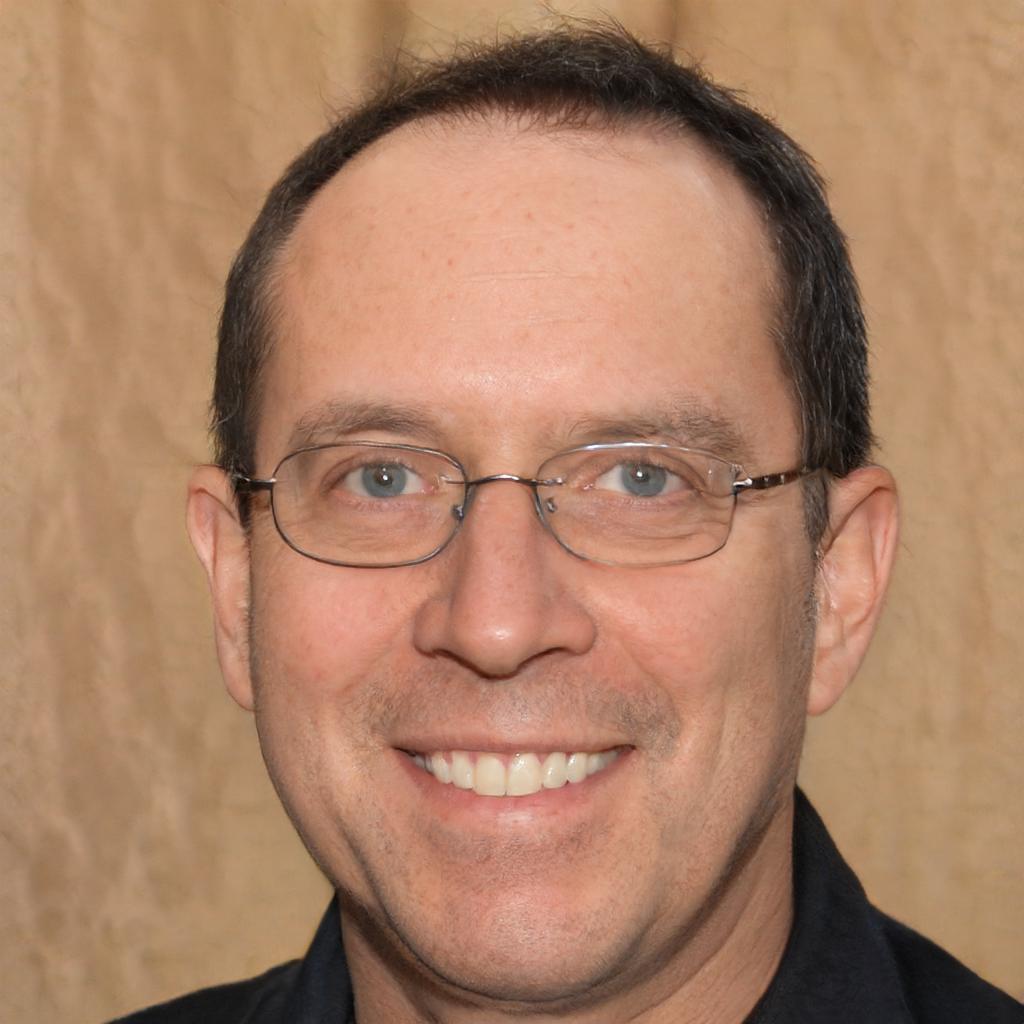} &
        \includegraphics[width=0.23\linewidth]
        {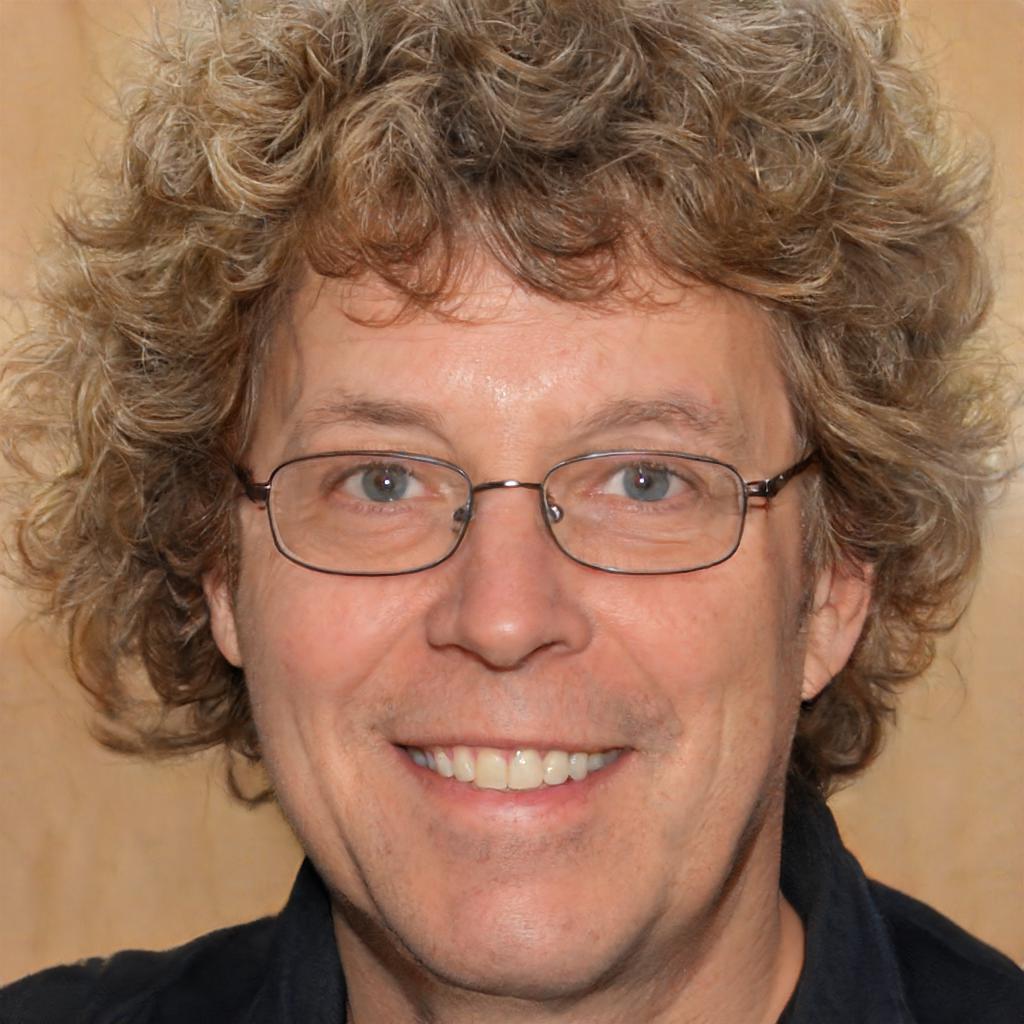} &
        \includegraphics[width=0.23\linewidth]
        {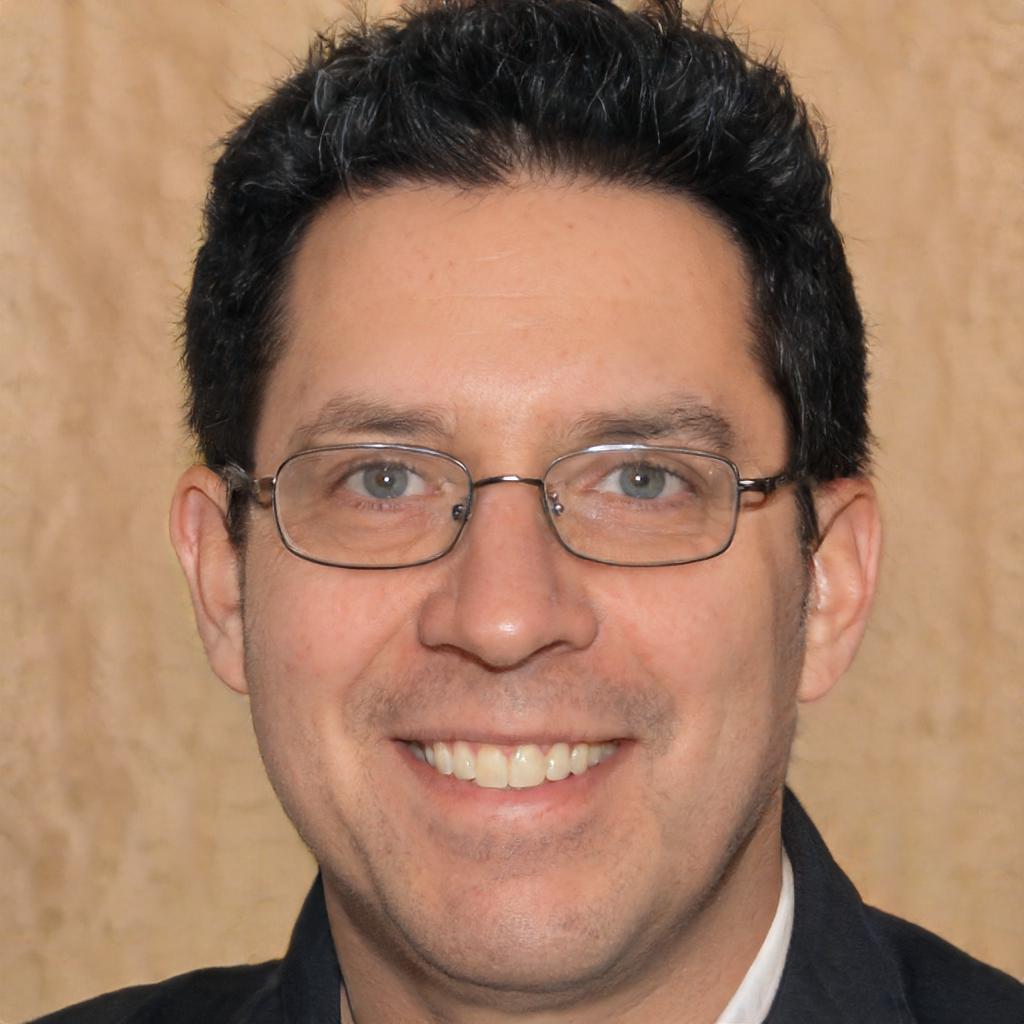}
        \tabularnewline
        \raisebox{0.30in}{\rotatebox[origin=t]{90}{Face}}& 
        \hspace{1pt} &
        \includegraphics[width=0.23\linewidth]
        {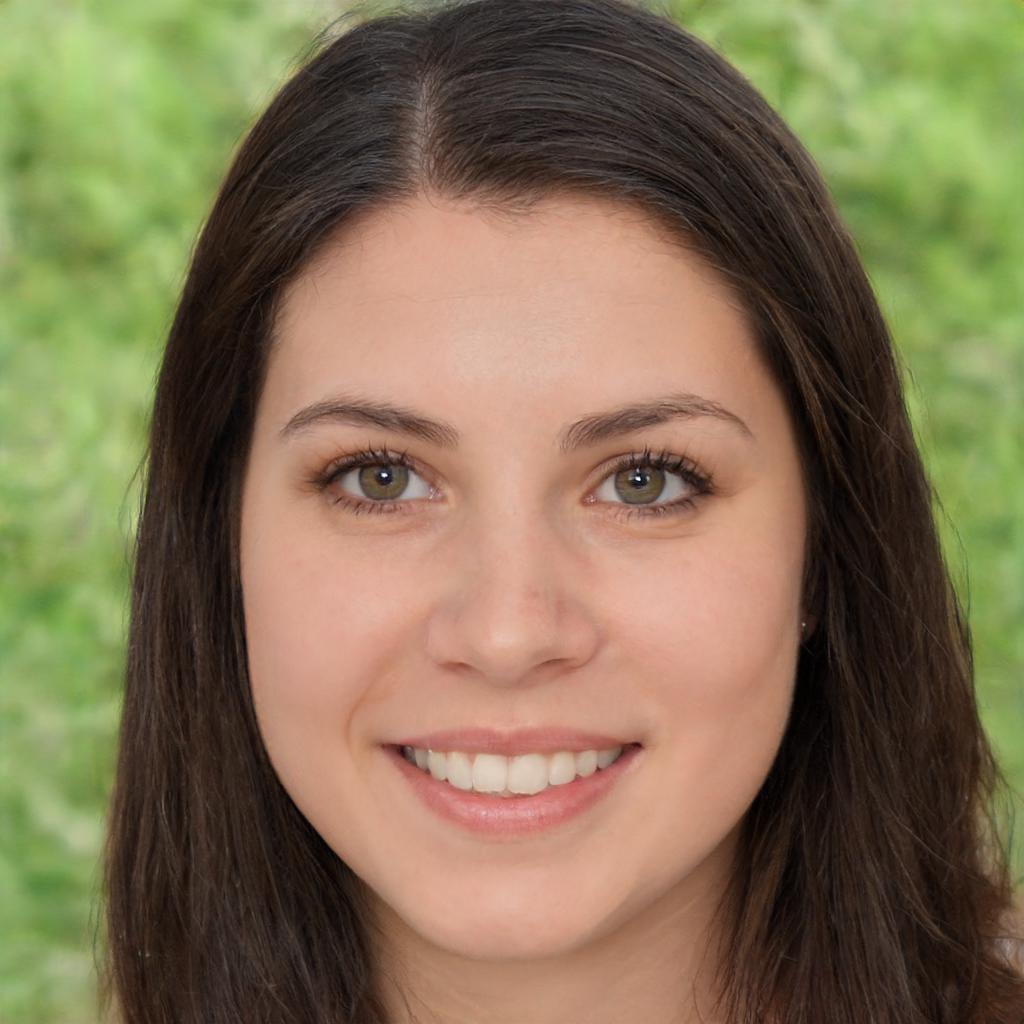} &
        \includegraphics[width=0.23\linewidth]
        {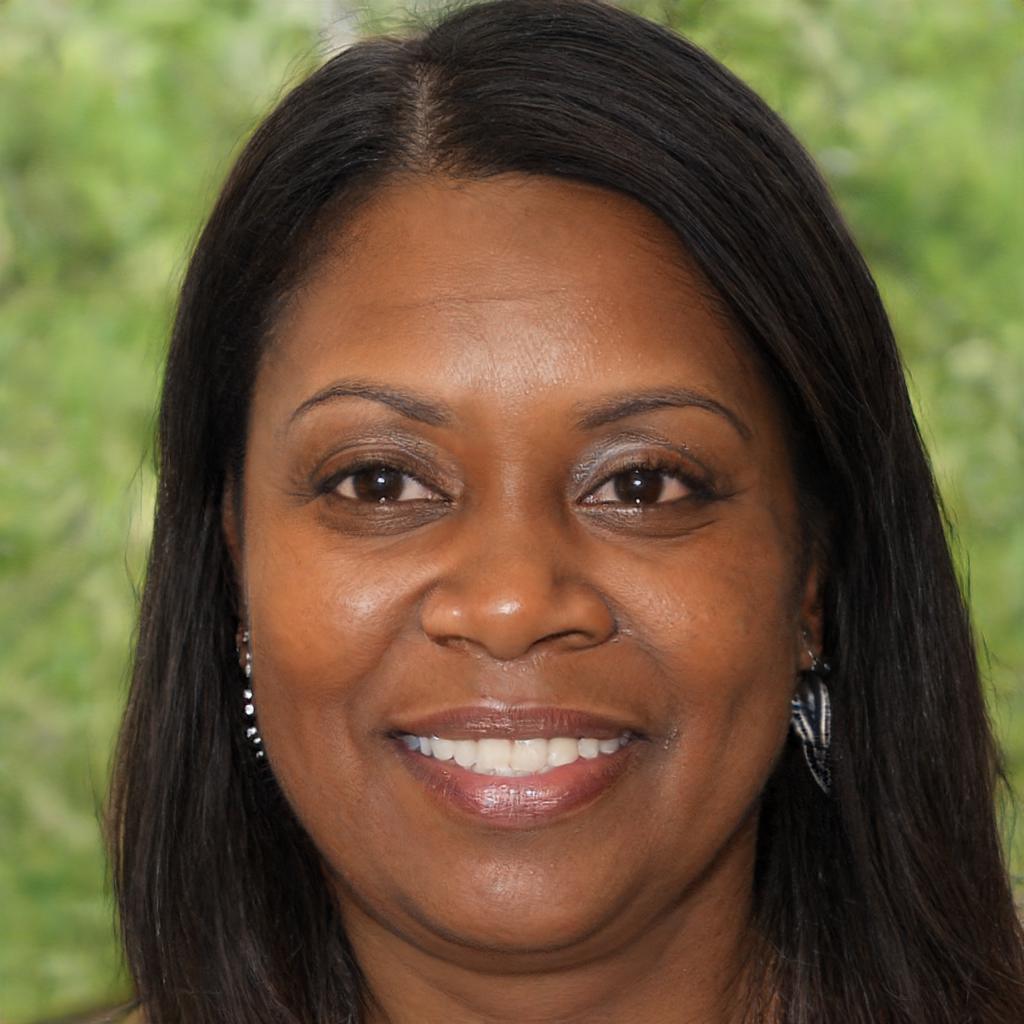} &
        \includegraphics[width=0.23\linewidth]
        {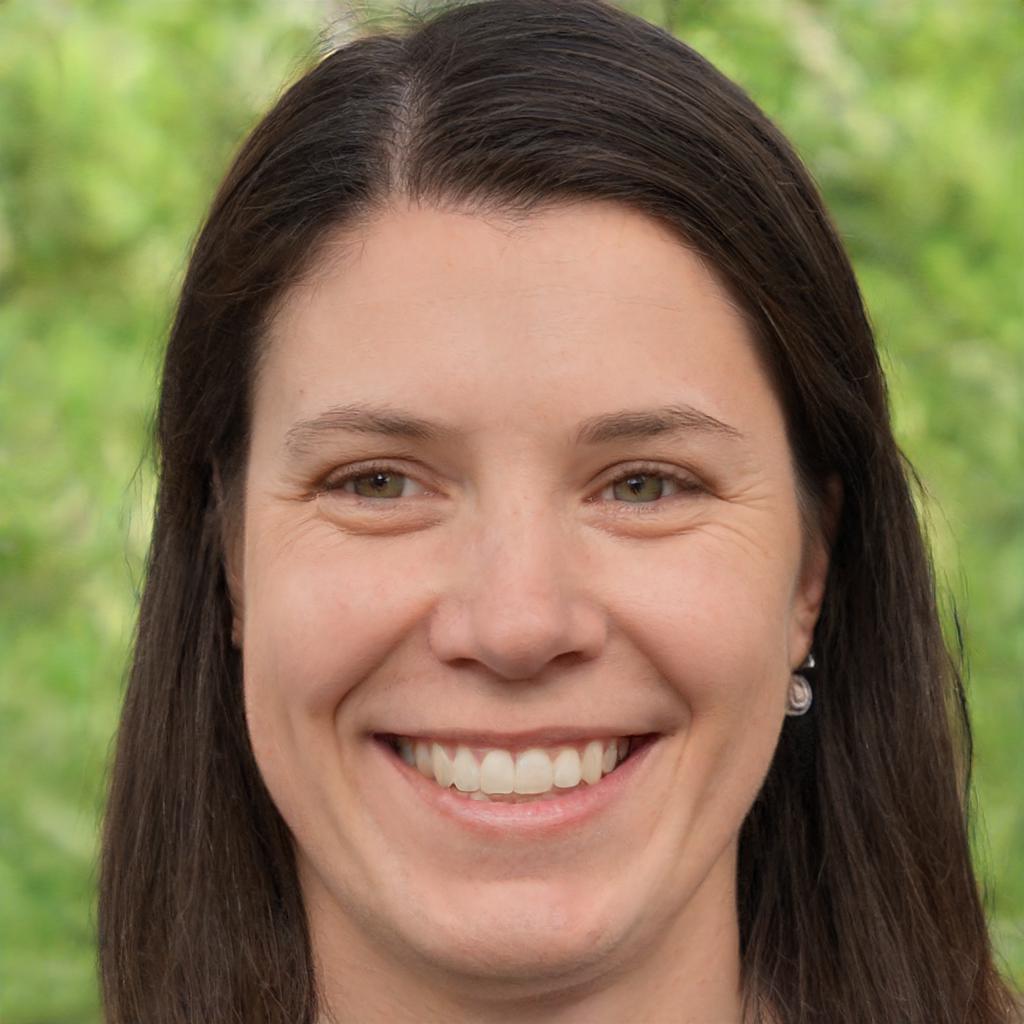} &
        \includegraphics[width=0.23\linewidth]
        {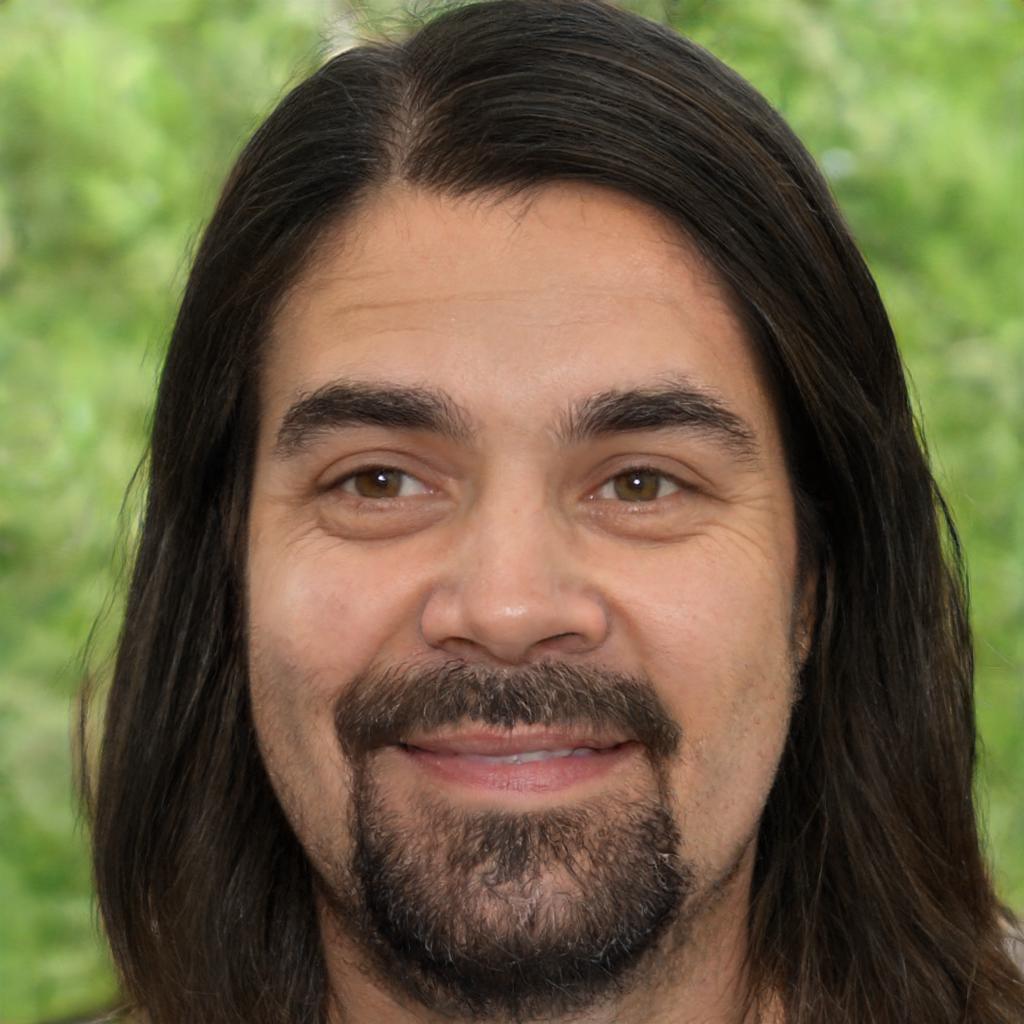}
        \tabularnewline
        \raisebox{0.30in}{\rotatebox[origin=t]{90}{Mouth}}& 
        \hspace{1pt} &
        \includegraphics[width=0.23\linewidth]
        {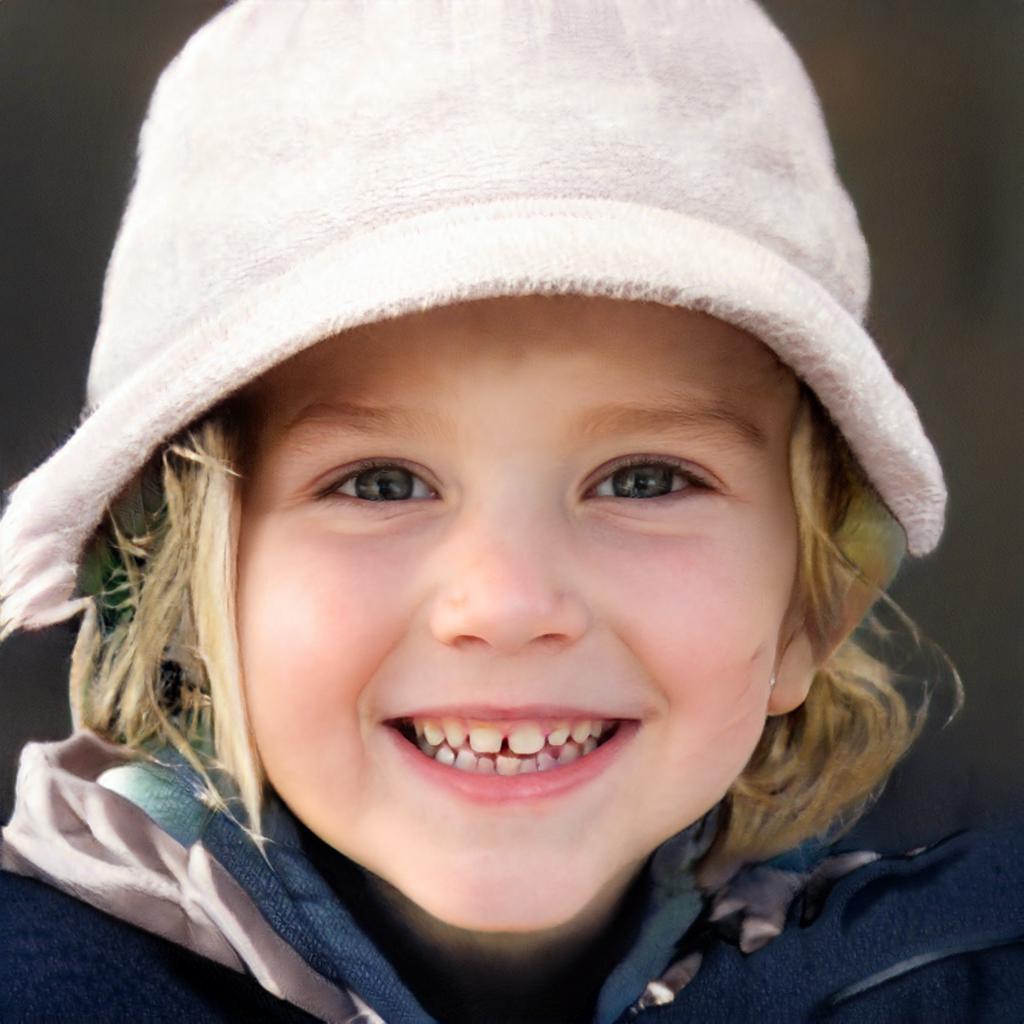} &
        \includegraphics[width=0.23\linewidth]
        {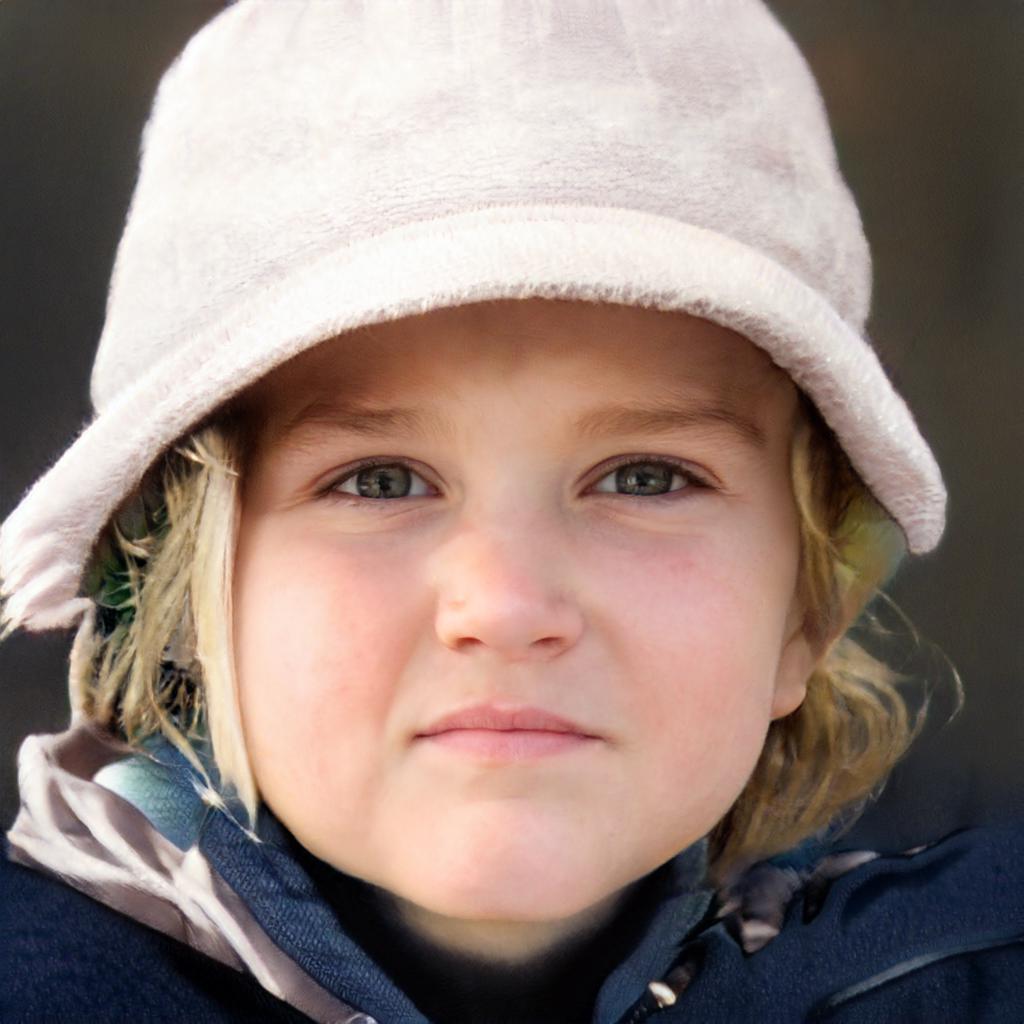} &
        \includegraphics[width=0.23\linewidth]
        {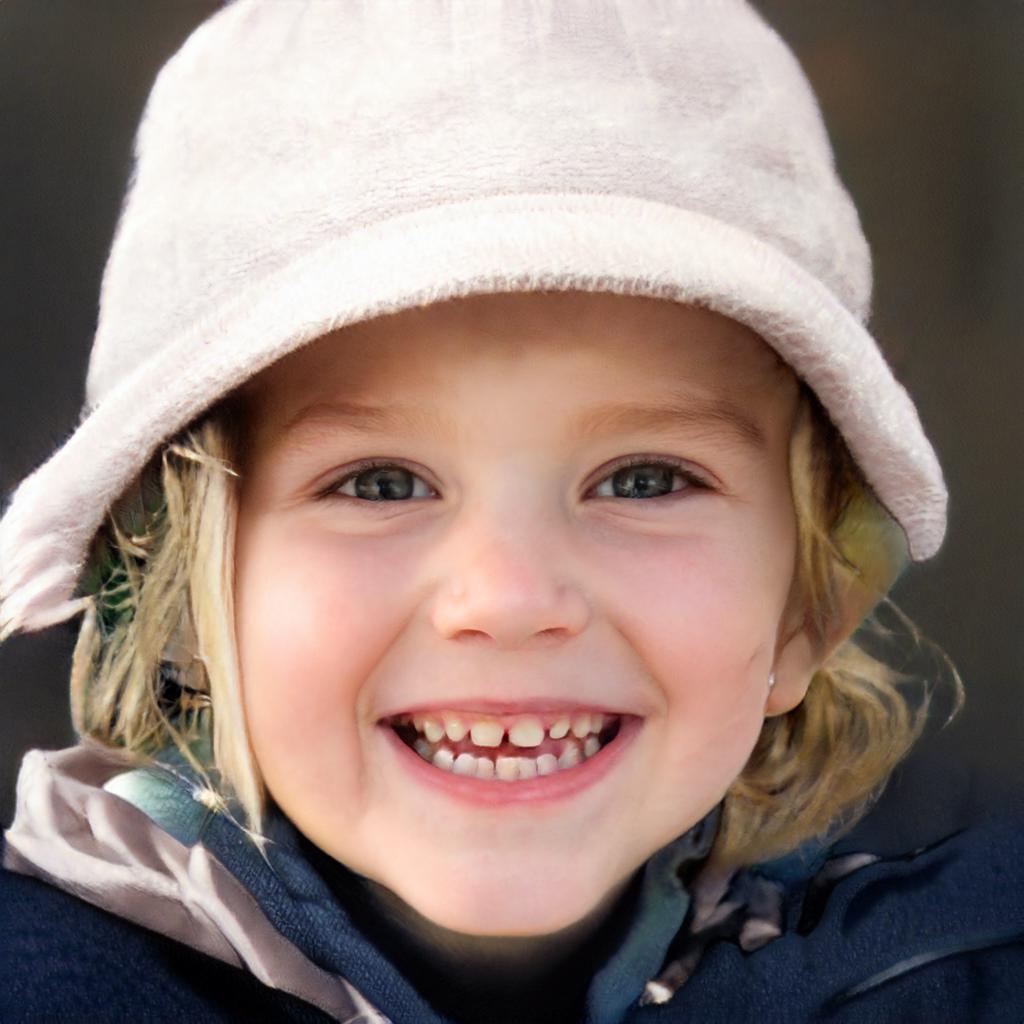} &
        \includegraphics[width=0.23\linewidth]
        {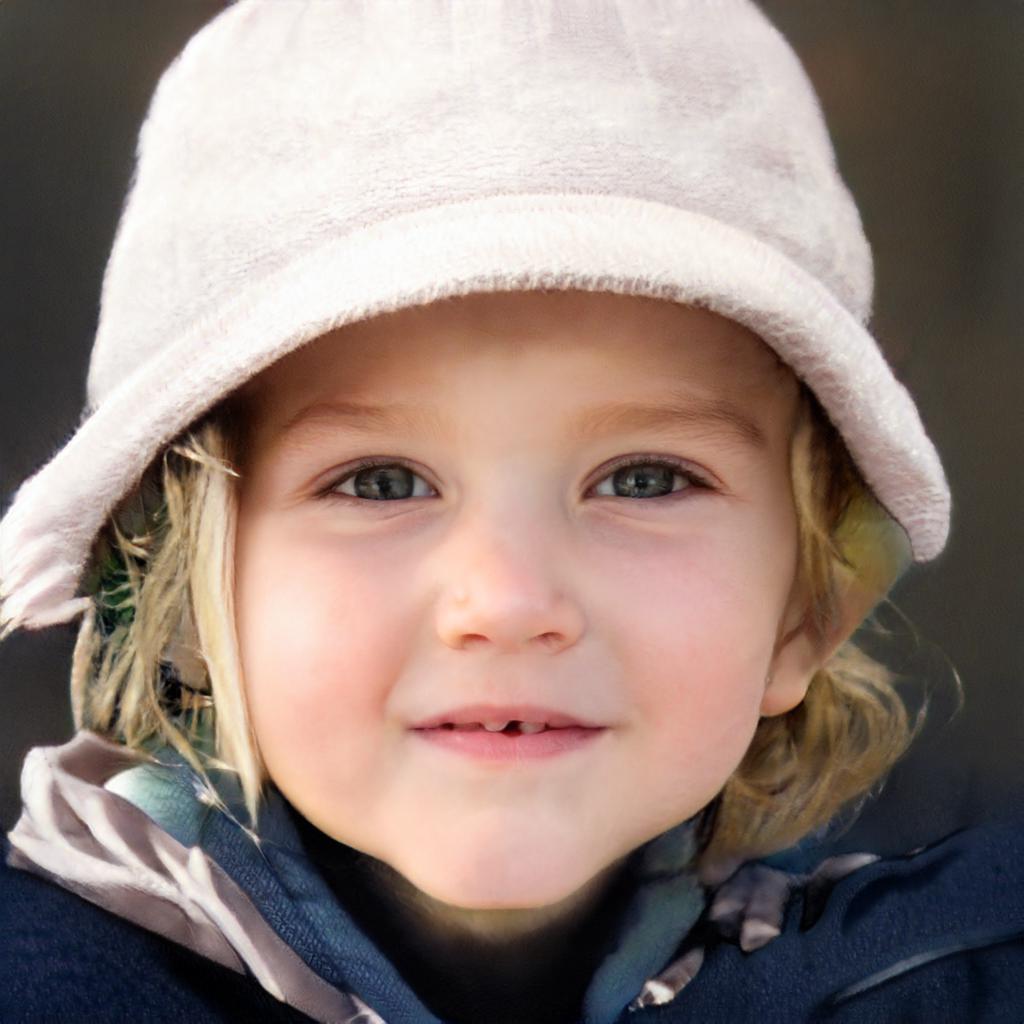}
        \tabularnewline
        \raisebox{0.30in}{\rotatebox[origin=t]{90}{Eyes}}& 
        \hspace{1pt} &
        \includegraphics[width=0.23\linewidth]
        {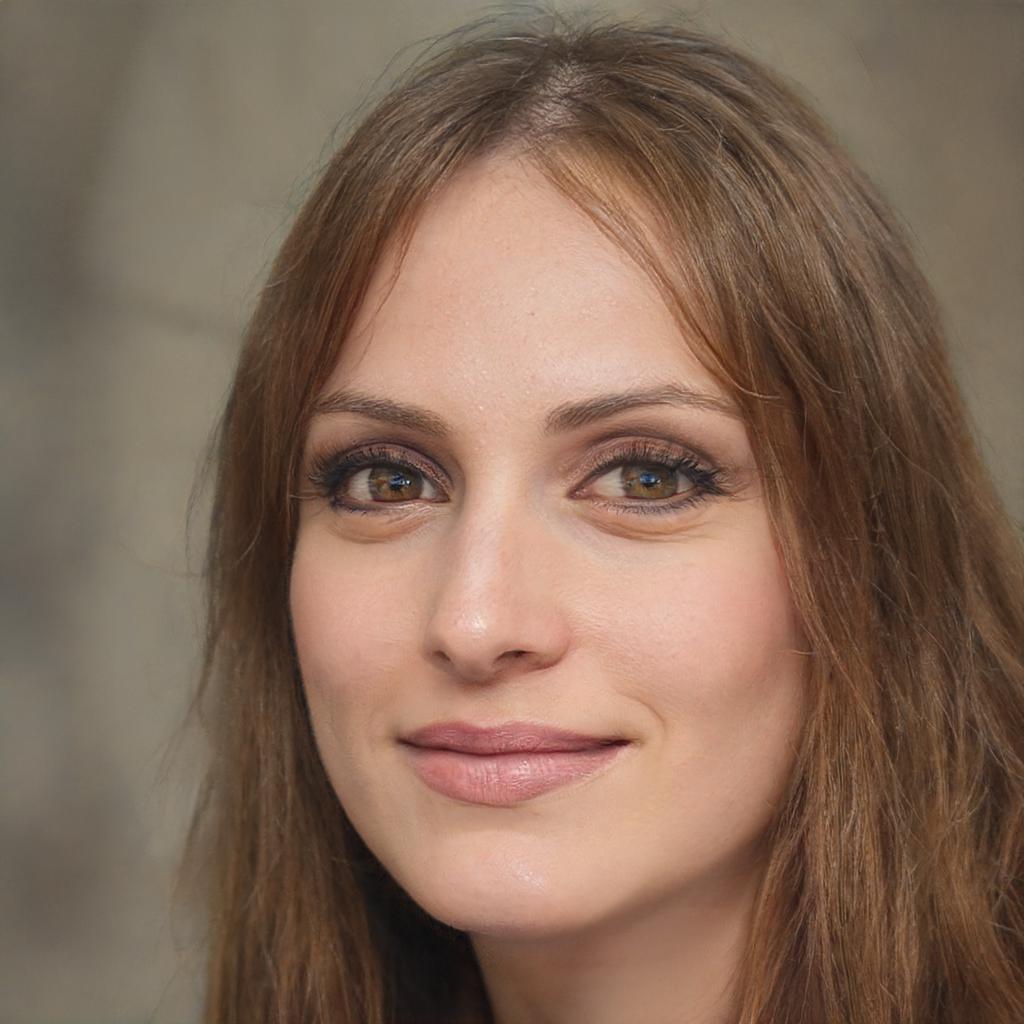} &
        \includegraphics[width=0.23\linewidth]
        {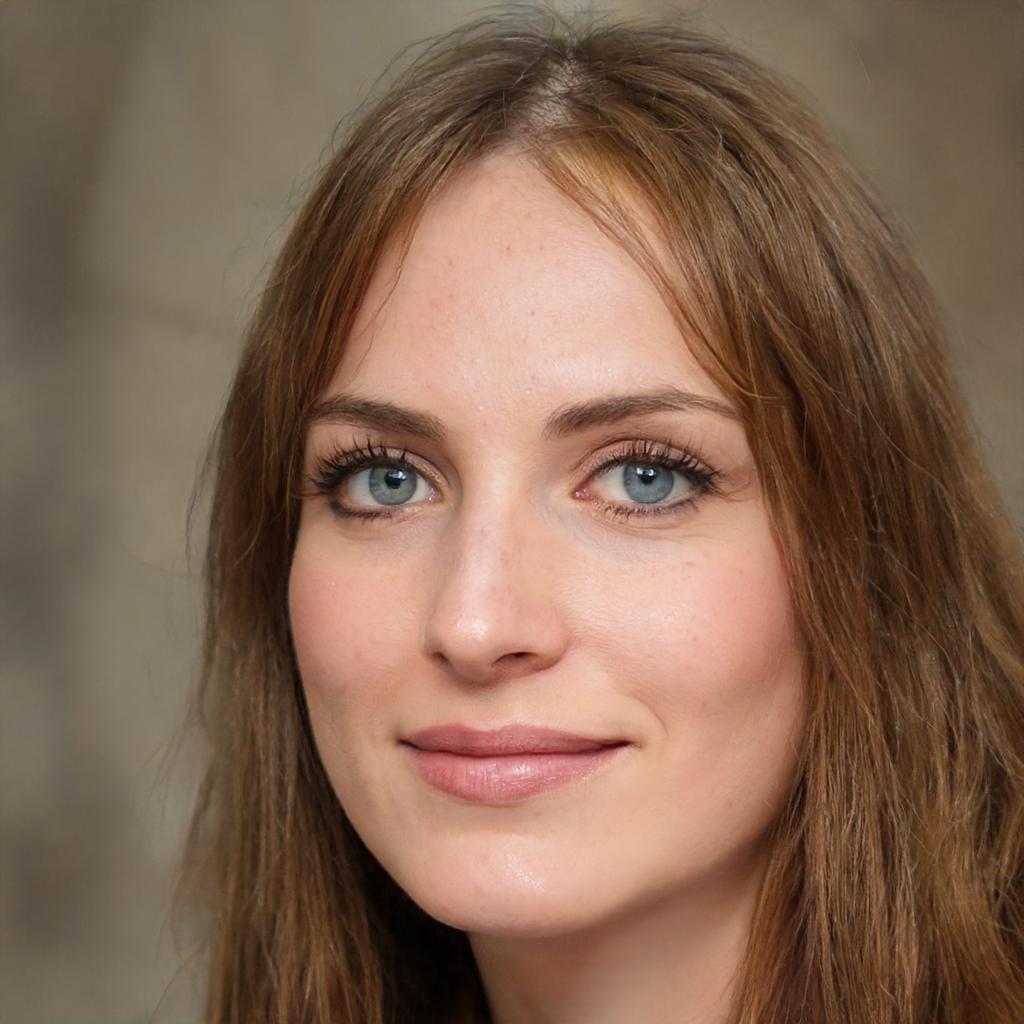} &
        \includegraphics[width=0.23\linewidth]
        {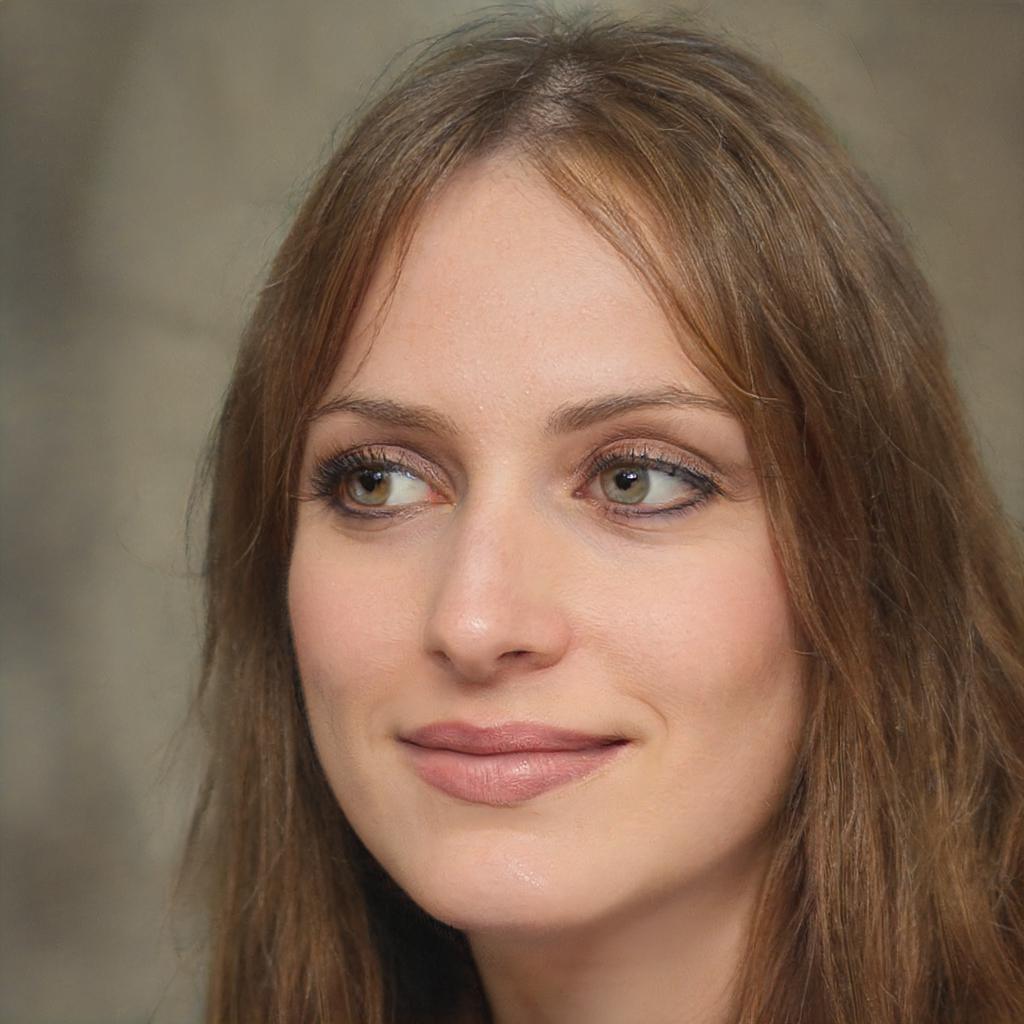} &
        \includegraphics[width=0.23\linewidth]
        {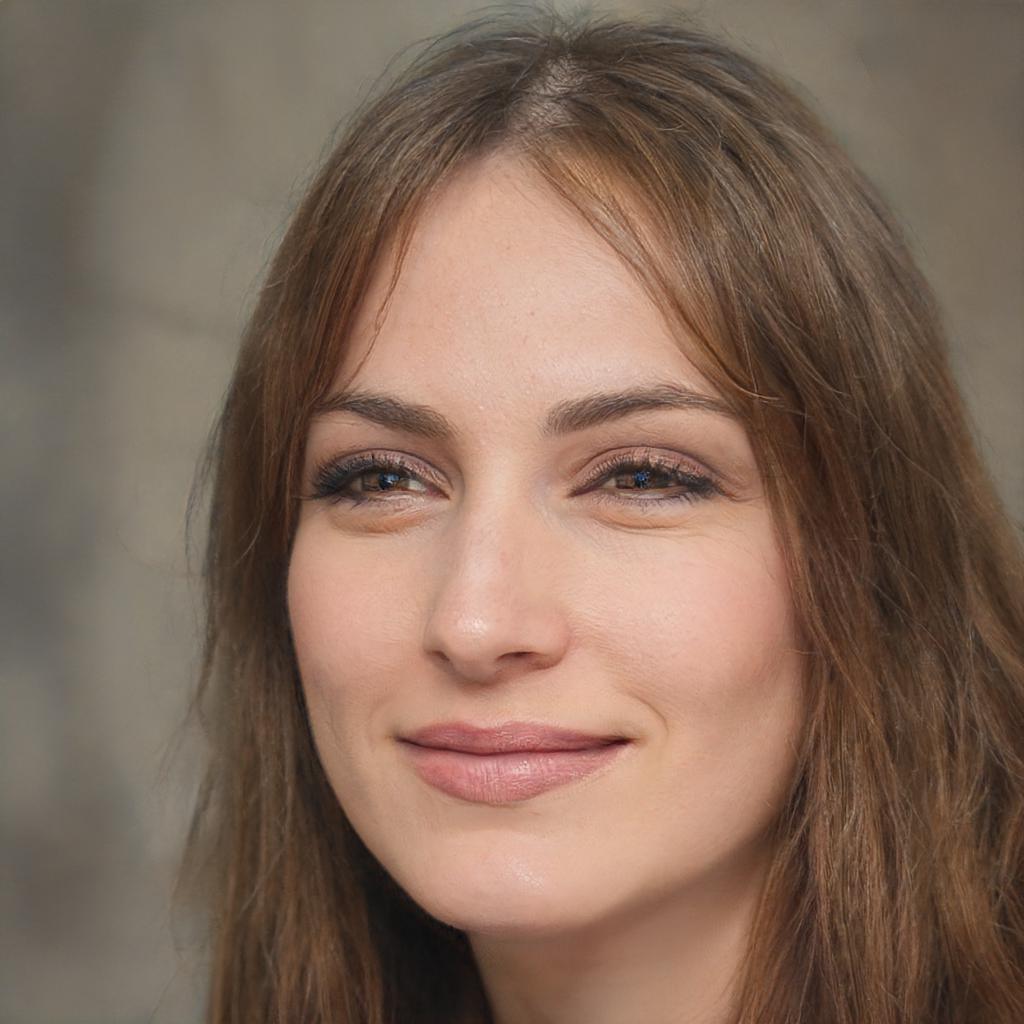}
        \tabularnewline
        \end{tabular}
    }
    \caption{\textit{Local semantic control on the human facial domain.} StyleFusion faithfully controls specific facial attributes without altering other attributes. Observe our ability to control both finer details (e.g., eyes), as well as more global attributes (e.g., hair) showing the versatility of StyleFusion's control.}
    \label{fig:local_control_faces}
\end{figure}

\begin{figure}
    \centering
    \setlength{\tabcolsep}{0pt}
    {\small
        \begin{tabular}{c c c c c c}
        \raisebox{0.215in}{\rotatebox[origin=t]{90}{Upper BG}} & 
        \hspace{1pt} &
        \includegraphics[width=0.23\linewidth]
        {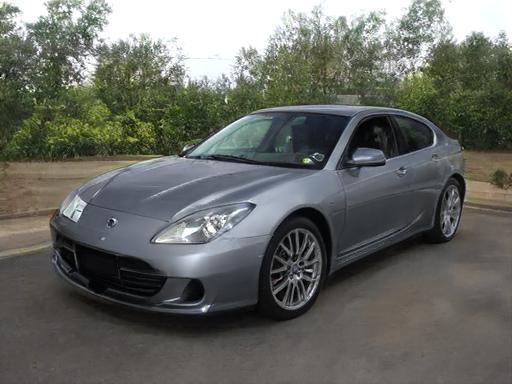} &
        \includegraphics[width=0.23\linewidth]
        {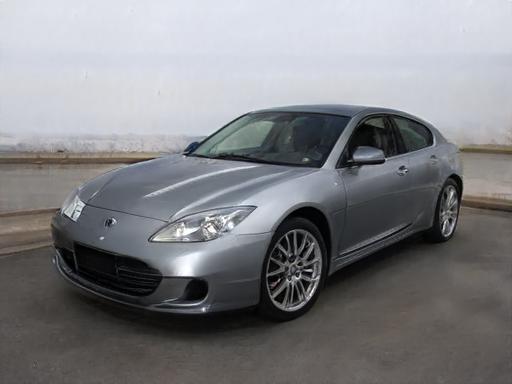} &
        \includegraphics[width=0.23\linewidth]
        {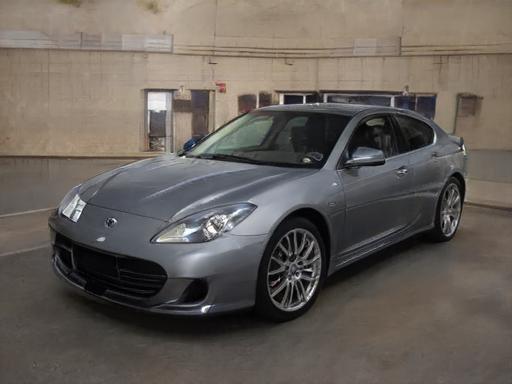} &
        \includegraphics[width=0.23\linewidth]
        {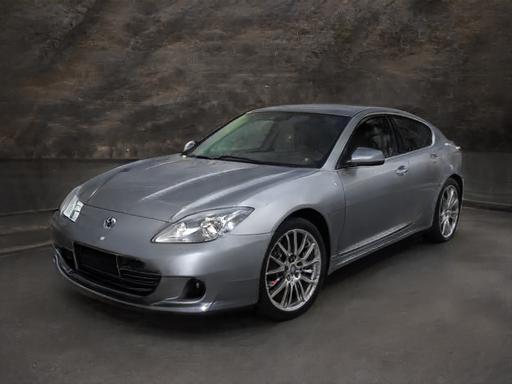}
        \tabularnewline
        \raisebox{0.2in}{\rotatebox[origin=t]{90}{Lower BG}}& 
        \hspace{1pt} &
        \includegraphics[width=0.23\linewidth]
        {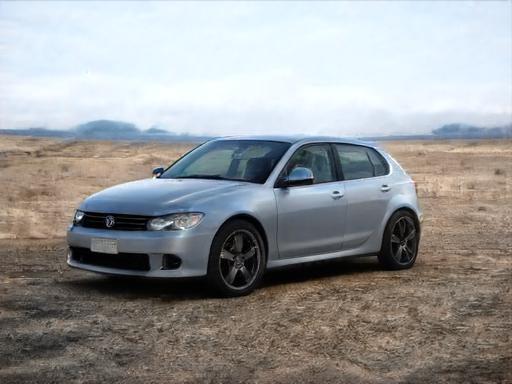} &
        \includegraphics[width=0.23\linewidth]
        {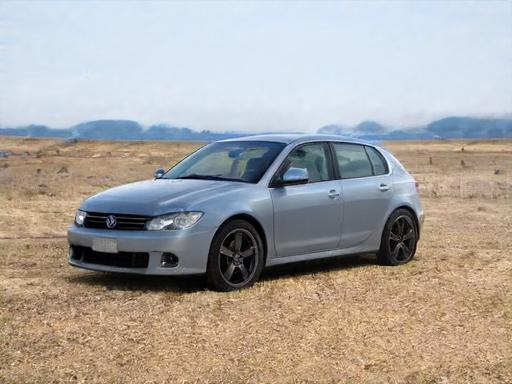} &
        \includegraphics[width=0.23\linewidth]
        {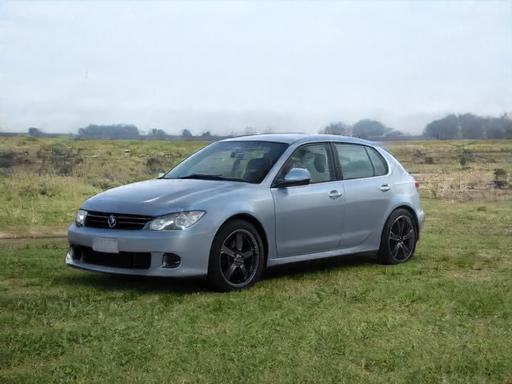} &
        \includegraphics[width=0.23\linewidth]
        {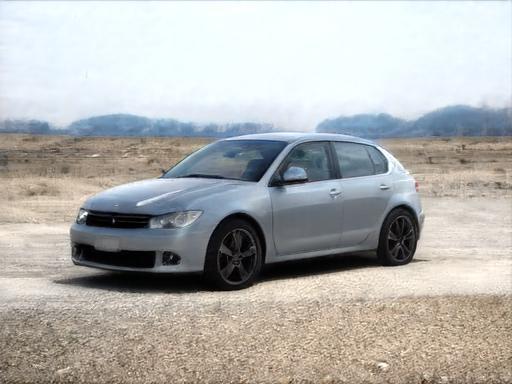}
        \tabularnewline
        \raisebox{0.2in}{\rotatebox[origin=t]{90}{Body}}& 
        \hspace{1pt} &
        \includegraphics[width=0.23\linewidth]
        {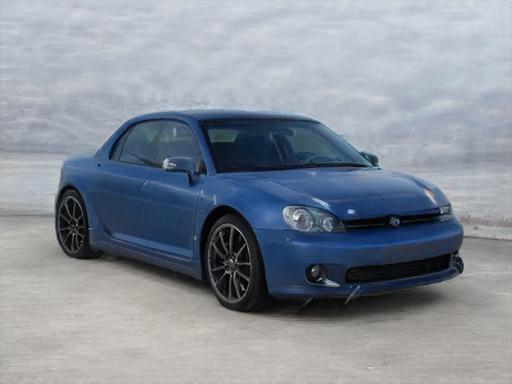} &
        \includegraphics[width=0.23\linewidth]
        {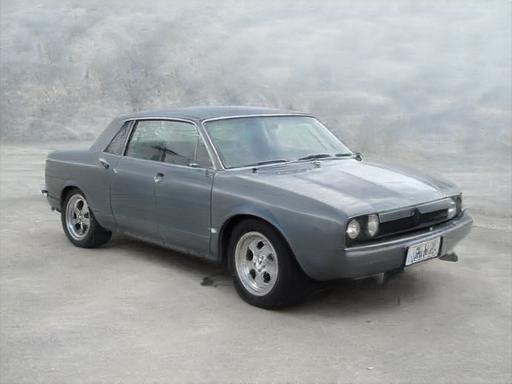} &
        \includegraphics[width=0.23\linewidth]
        {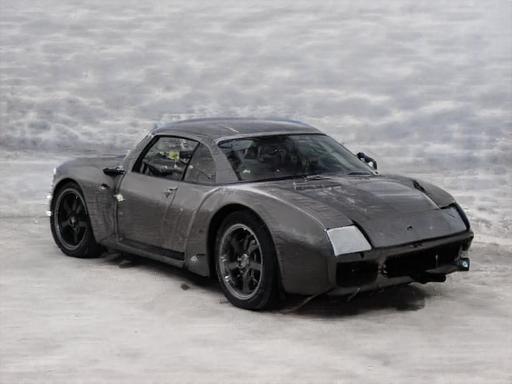} &
        \includegraphics[width=0.23\linewidth]
        {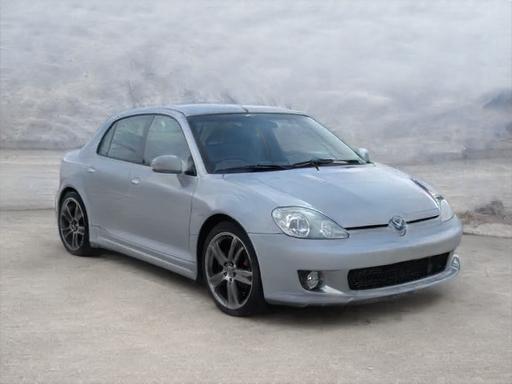}
        \tabularnewline
        \raisebox{0.215in}{\rotatebox[origin=t]{90}{Wheels}}& 
        \hspace{1pt} &
        \includegraphics[width=0.23\linewidth]
        {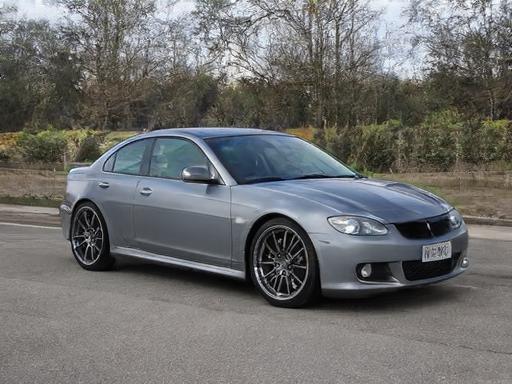} &
        \includegraphics[width=0.23\linewidth]
        {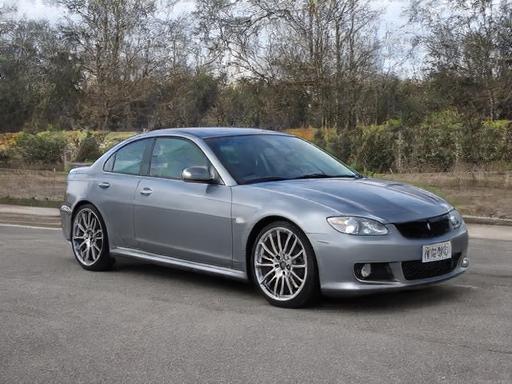} &
        \includegraphics[width=0.23\linewidth]
        {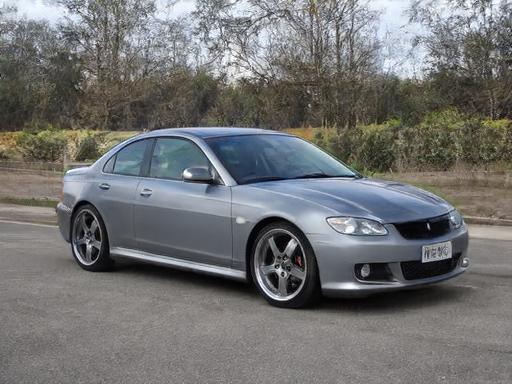} &
        \includegraphics[width=0.23\linewidth]
        {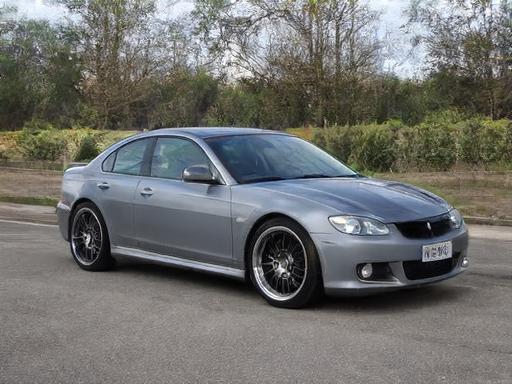}
        \tabularnewline
        \end{tabular}
    }
    \caption{\textit{Local semantic control on the cars domain.} StyleFusion faithfully controls specific attributes of the car image without altering other attributes. Similar to the facial domain, observe the control over finer details (e.g., wheels), as well as more global attributes (e.g., background).}
    \label{fig:local_control_cars}
\end{figure}

\begin{figure}
    \centering
    \setlength{\tabcolsep}{0pt}
    {\small
        \begin{tabular}{c c c}
        \includegraphics[width=0.33\linewidth]{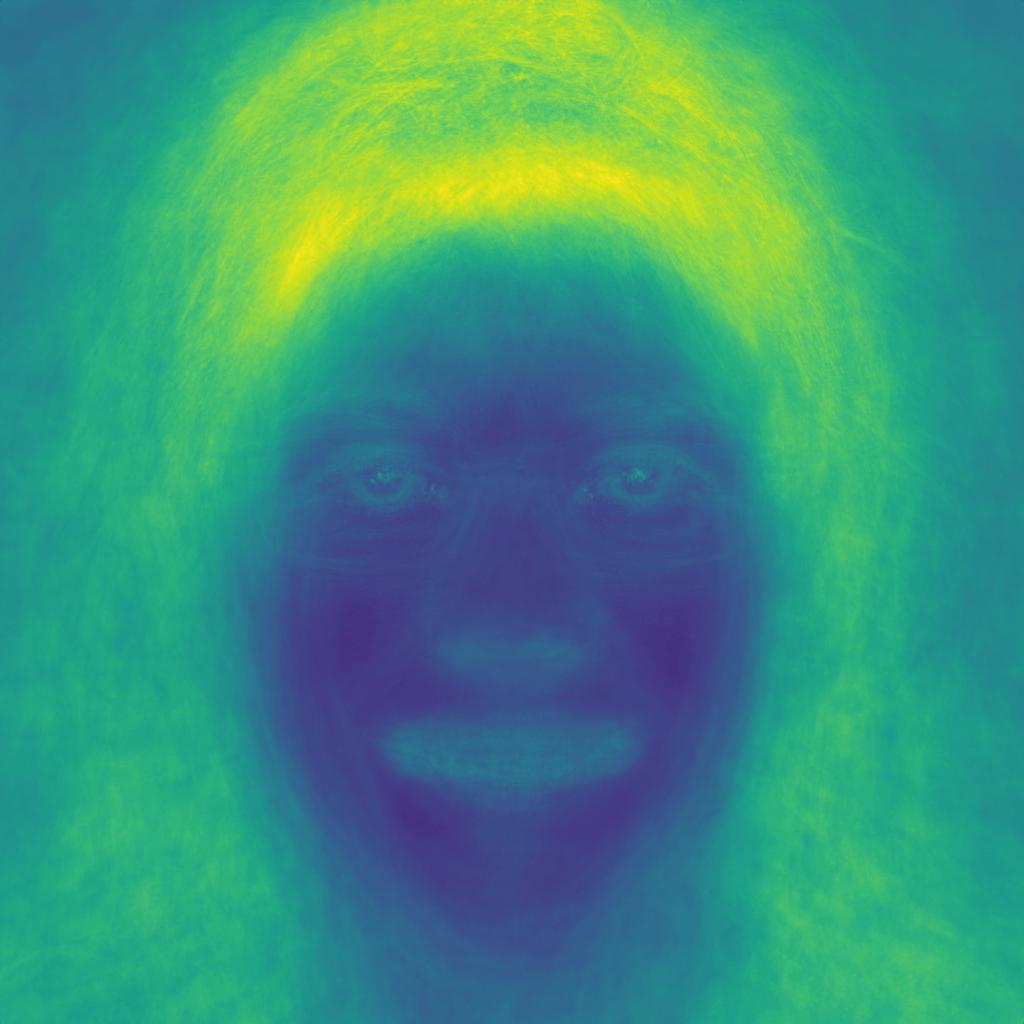} &
        \includegraphics[width=0.33\linewidth]{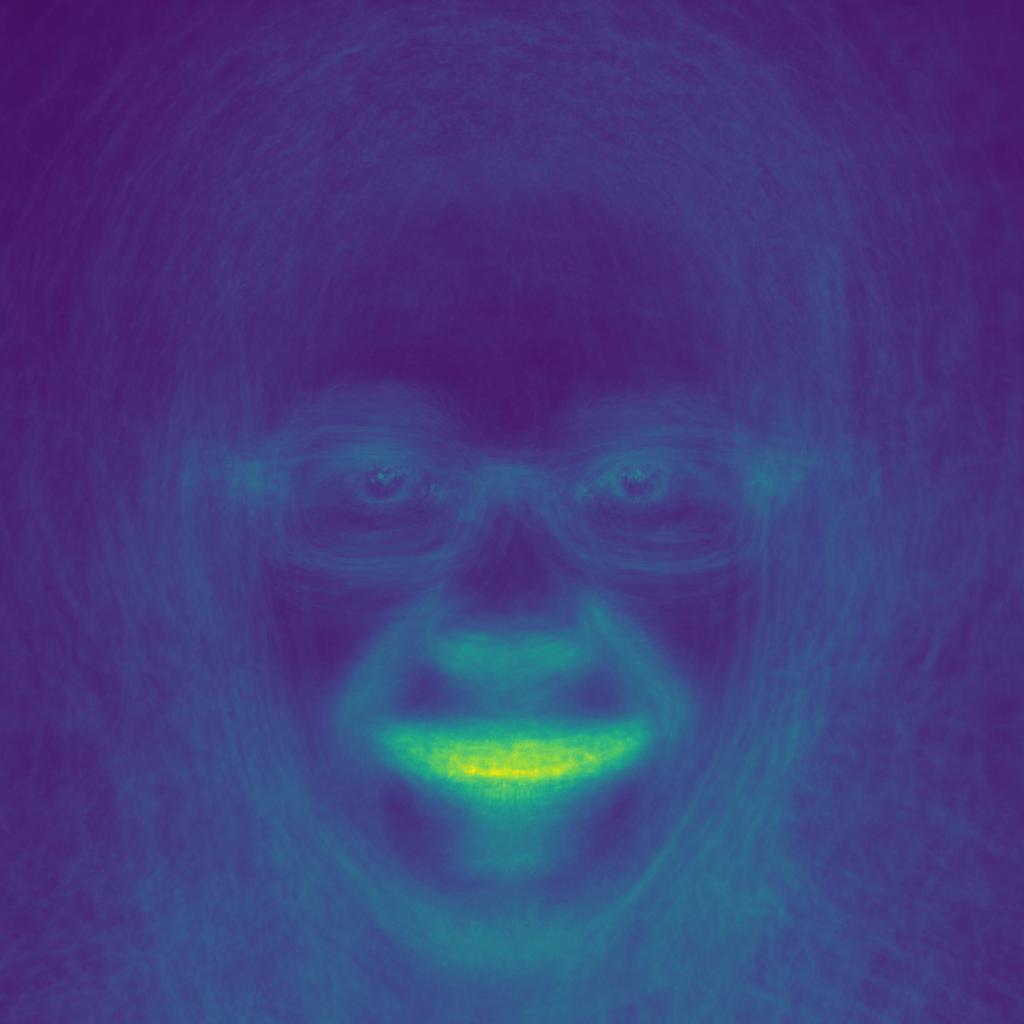} &
        \includegraphics[width=0.33\linewidth]{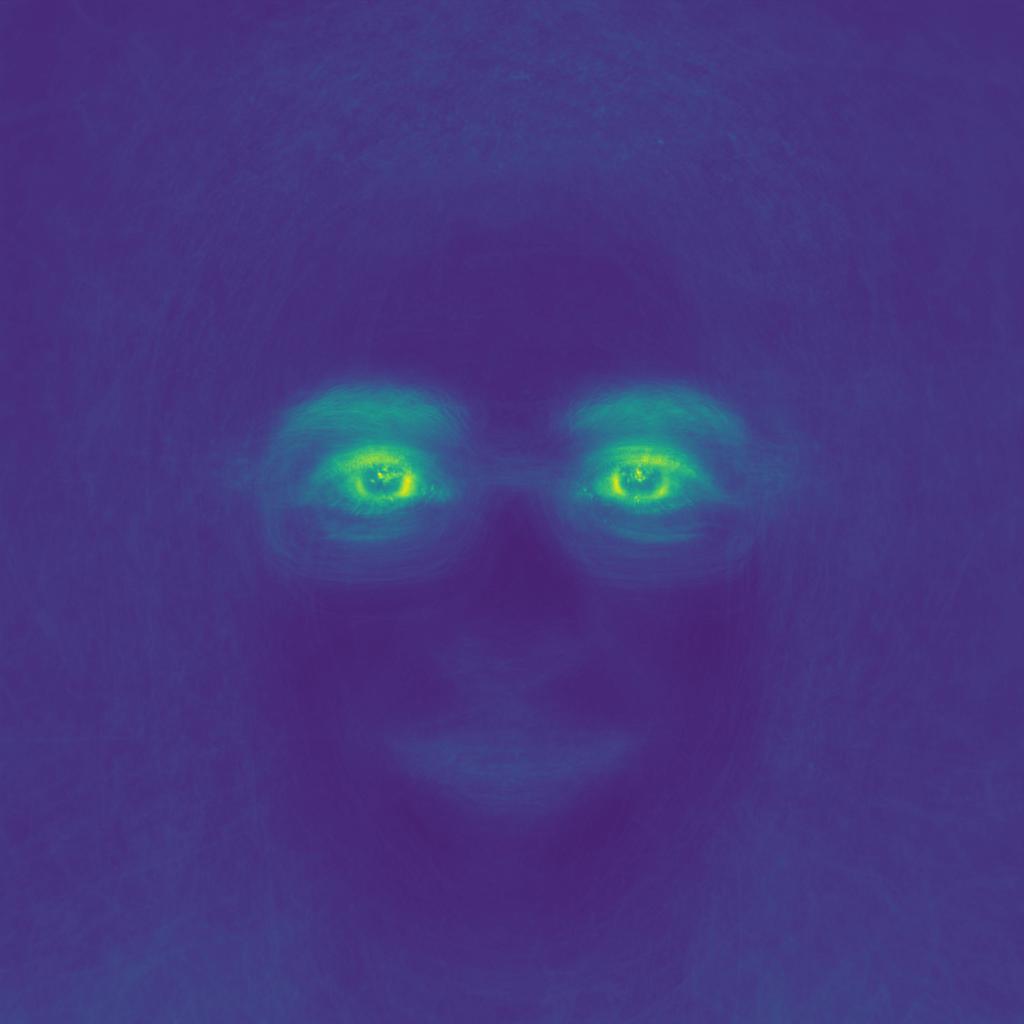} \\
        Hair & Mouth & Eyes
        \end{tabular}
    }
    \vspace{0.05cm}
    \caption{\textit{Visualizing the local changes.}
    For each attribute, we randomly sample $500$ images and alter the specified attribute. We then compute the per-pixel differences between the original and altered images and average across all $500$ images to obtained the displayed heatmaps. Observe that the changes are easily perceptible while mainly altering the specified region.}
    \vspace{-0.15cm}
    \label{fig:heatmap_face_average}
\end{figure}
\begin{figure}
    \centering
    \setlength{\tabcolsep}{0pt}
    {\small
        \begin{tabular}{c c c c}
        \includegraphics[width=0.25\linewidth]{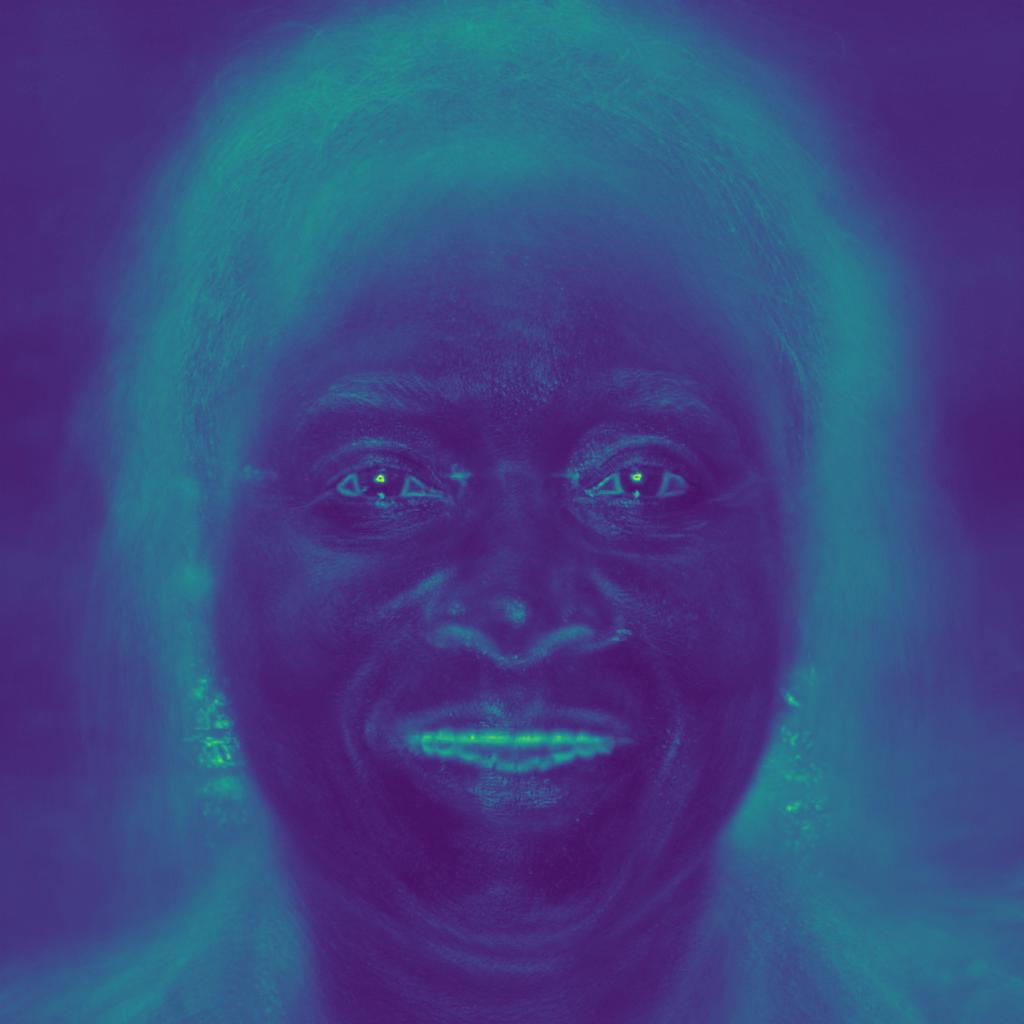} &
        \includegraphics[width=0.25\linewidth]{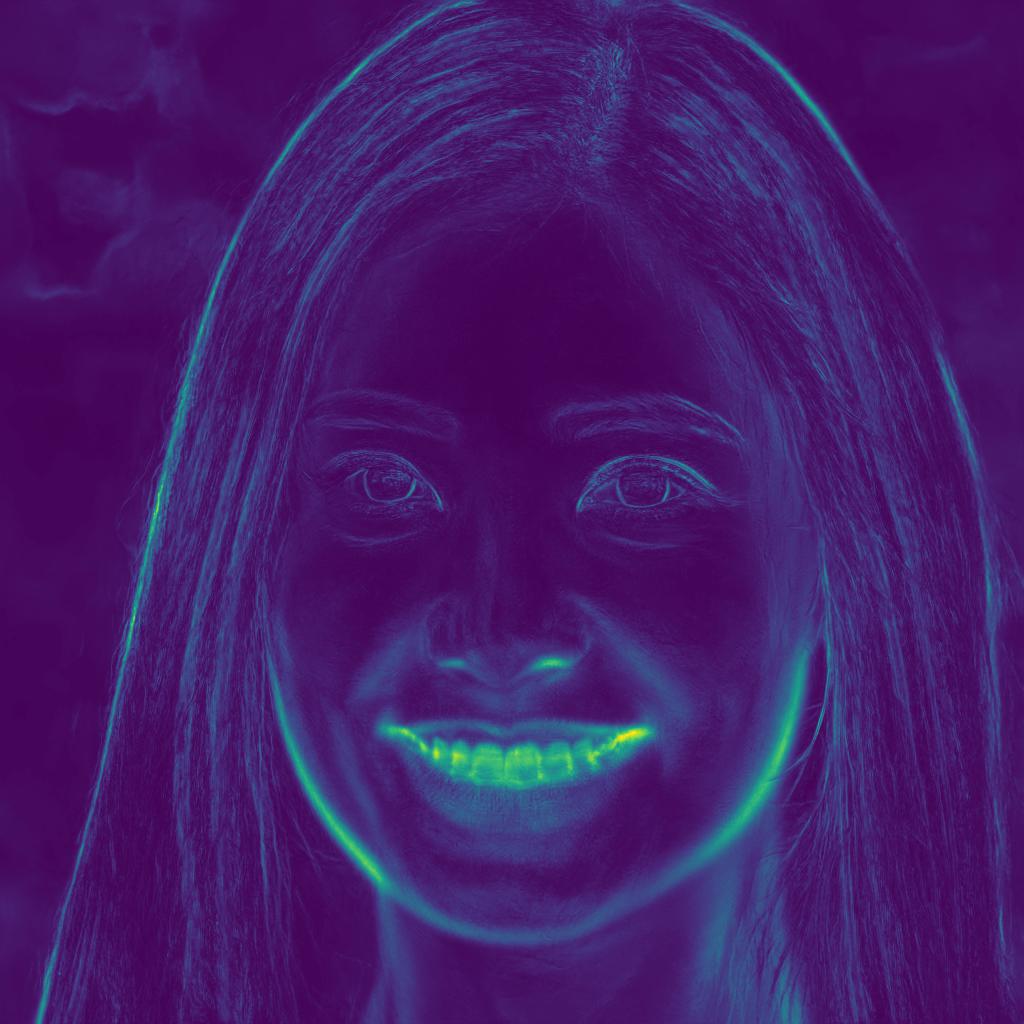} &
        \includegraphics[width=0.25\linewidth]{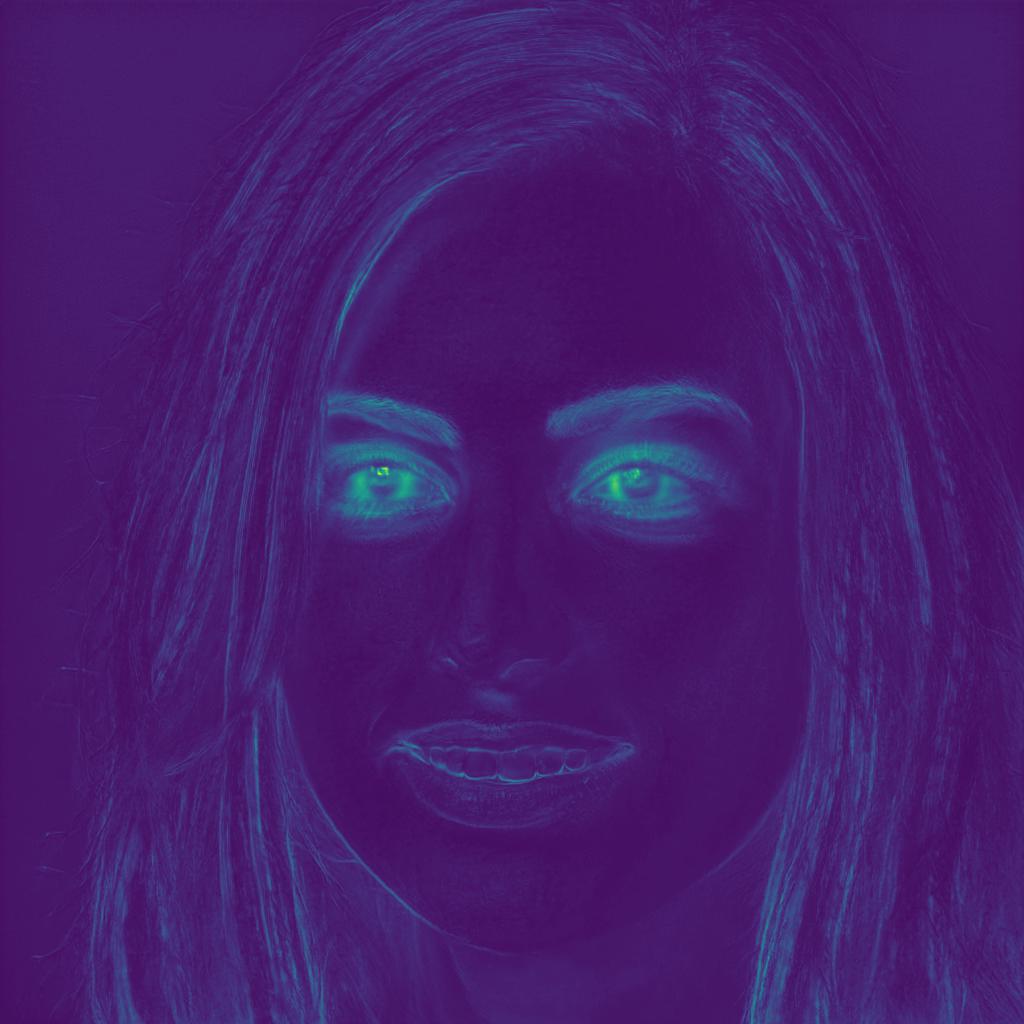} & 
        \includegraphics[width=0.25\linewidth]{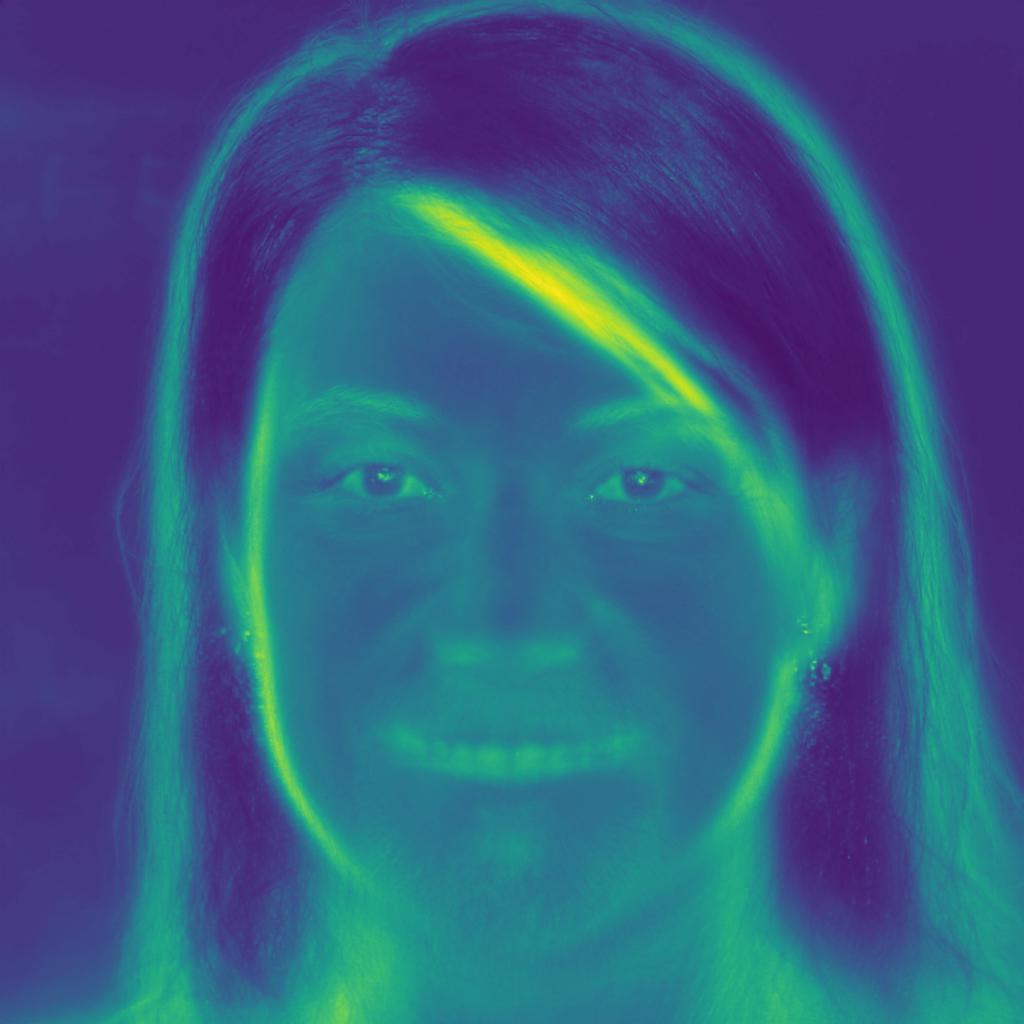} \\
        Hair & Mouth & Eyes & Face \\
        
        \includegraphics[width=0.25\linewidth]{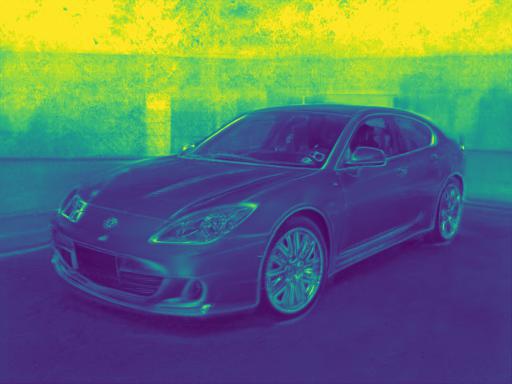} &
        \includegraphics[width=0.25\linewidth]{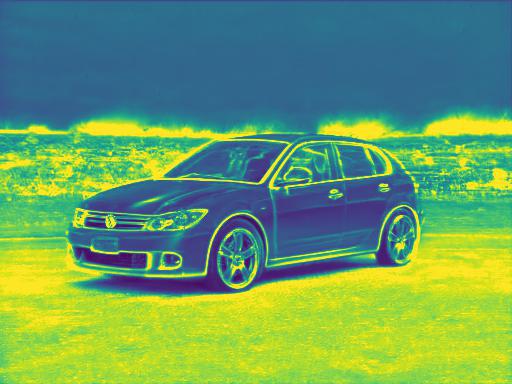} &
        \includegraphics[width=0.25\linewidth]{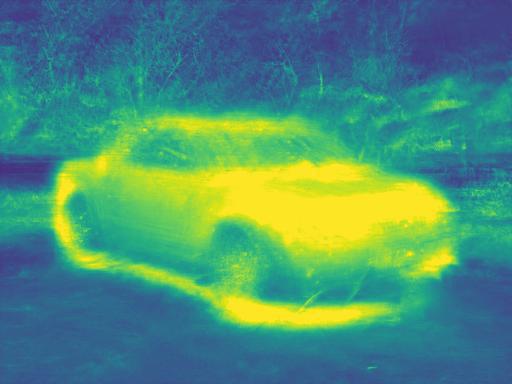} & 
        \includegraphics[width=0.25\linewidth]{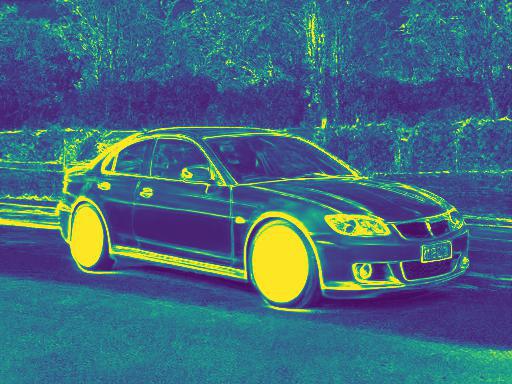} \\
        Upper BG & Lower BG & Body & Wheels
        \end{tabular}
    }
    \vspace{0.1cm}
    \caption{\textit{Visualizing local changes.}
    Similar to Figure~\ref{fig:heatmap_face_average} where instead of averaging over $500$ randomly sampled images, we average over $500$ random modifications of the specified attribute relative to a constant base image.} 
    \vspace{-0.1cm}
    \label{fig:heatmap_face_cars_single}
\end{figure}

\section{Results}~\label{sec:results}
To gain a stronger understanding of StyleFusion and attain key insights into its functionality, we now turn to analyze the various capabilities of our proposed system. Specifically, we begin by analyzing the locality of StyleFusion's control (e.g., altering the eyes does not affect other regions of the generated image). We then illustrate the role of the ``global" latent code in controlling global attributes such as pose and structure and show its usefulness in more challenging domains with large variability. Additional results can be found in Appendix~\ref{sec:appendix_results}.

\vspace{0.15cm}
\topic{\textit{Local Semantic Control. }}~\label{local_control}
The first natural evaluation of the learned disentanglement of StyleFusion can be done by generating multiple images sharing the same characteristics with the exception of a single semantic image region (e.g., images sharing the same face, hair, eyes, but all with different mouths). In other words, we would like to provide users with local control over the image attributes.
In Figure~\ref{fig:local_control_faces} we analyze StyleFusion's ability to provide such local control in the human facial domain. In the Figure, each row corresponds to a single semantic region which we wish to modify in the target image. 
Observe, for example, how StyleFusion is able to faithfully alter the hair in the top row and the mouth in the third row while preserving the other attributes and the individual's identity. Similarly, in Figure~\ref{fig:local_control_cars}, notice StyleFusion's ability to faithfully manipulate more global features such as the image background as well as smaller details such as the wheels of the car. 

To further validate the locality of StyleFusion's control, we refer the reader to Figures~\ref{fig:heatmap_face_average} and ~\ref{fig:heatmap_face_cars_single}. 
First, in Figure~\ref{fig:heatmap_face_average}, we demonstrate the area controlled by a target semantic region $R$. 
To do so, we sample one latent code, $s_{R^C}$, to control all of the regions other then $R$, and sample two latent codes, $s_{R_1}, s_{R_2}$ for controlling region $R$. 
We then measure the per-pixel difference between the image generated by $(s_{R_1}, s_{R^C})$ and the image generated by $(s_{R_2}, s_{R^C})$. We then average this difference over $500$ randomly chosen samples for $R_1, R_2,$ and $R_C$.
As can be seen, altering a specific target attribute results in variations specific to the desired region with very small changes to other facial attributes.

The above visualization is useful for the human facial domain since the semantic regions generally remain in the same spatial location across all samples. For domains with higher variability (e.g., cars) this assumption cannot be made. We therefore refer the reader to Figure~\ref{fig:heatmap_face_cars_single}. There, we illustrate the image regions controlled by a given semantic attribute, relative to a constant latent code for all regions other than $R$. 
In other words, we repeat the same process as above, but keep $s_{R^C}$ fixed over all of the samples. 
Observe that even in the more complex cars domain, StyleFusion is still able to faithfully alter the specified region, demonstrating the accuracy of the learned disentanglement.

\begin{figure}
    \centering
    \setlength{\tabcolsep}{0pt}
    {\small
        \begin{tabular}{c c c c c c}
        \raisebox{0.275in}{\rotatebox[origin=t]{90}{Head Pose}} &
        \hspace{1pt} &
        \includegraphics[width=0.23\linewidth]
        {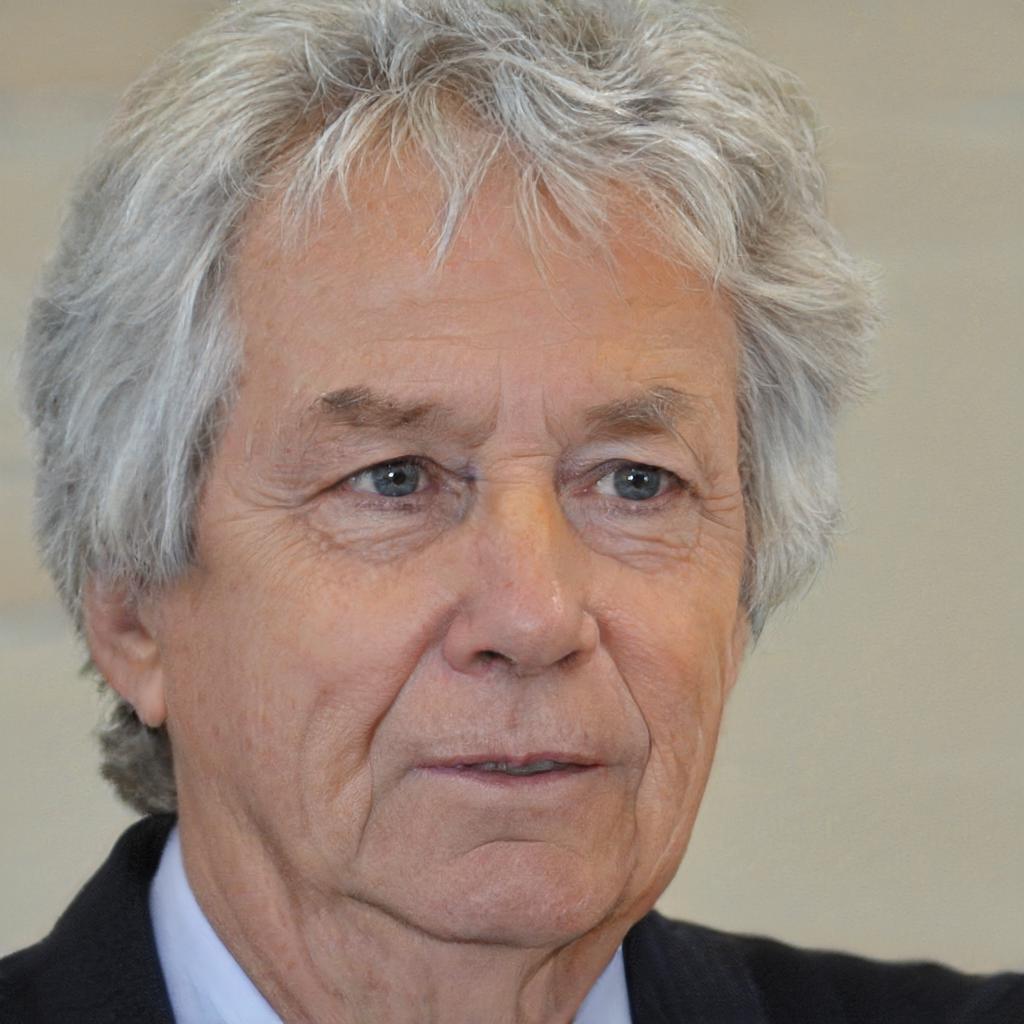} &
        \includegraphics[width=0.23\linewidth]
        {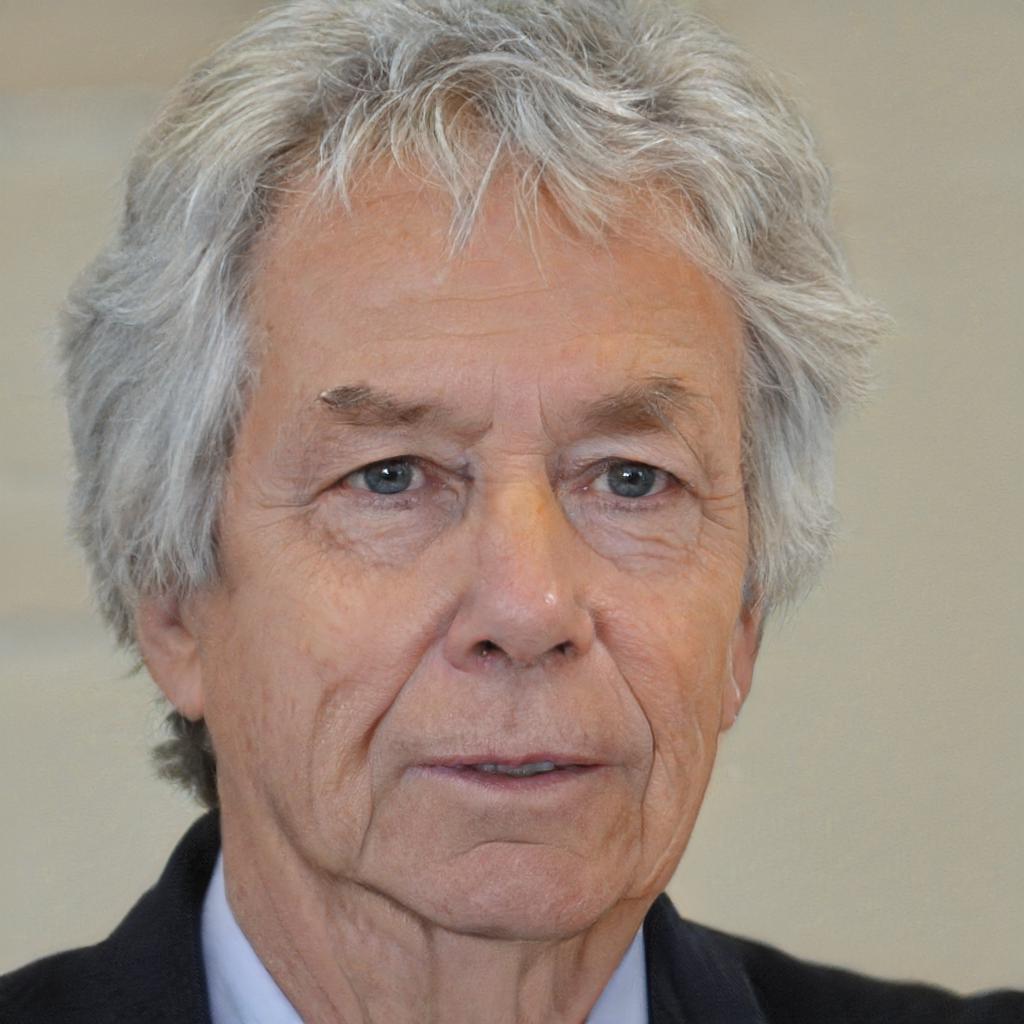} &
        \includegraphics[width=0.23\linewidth]
        {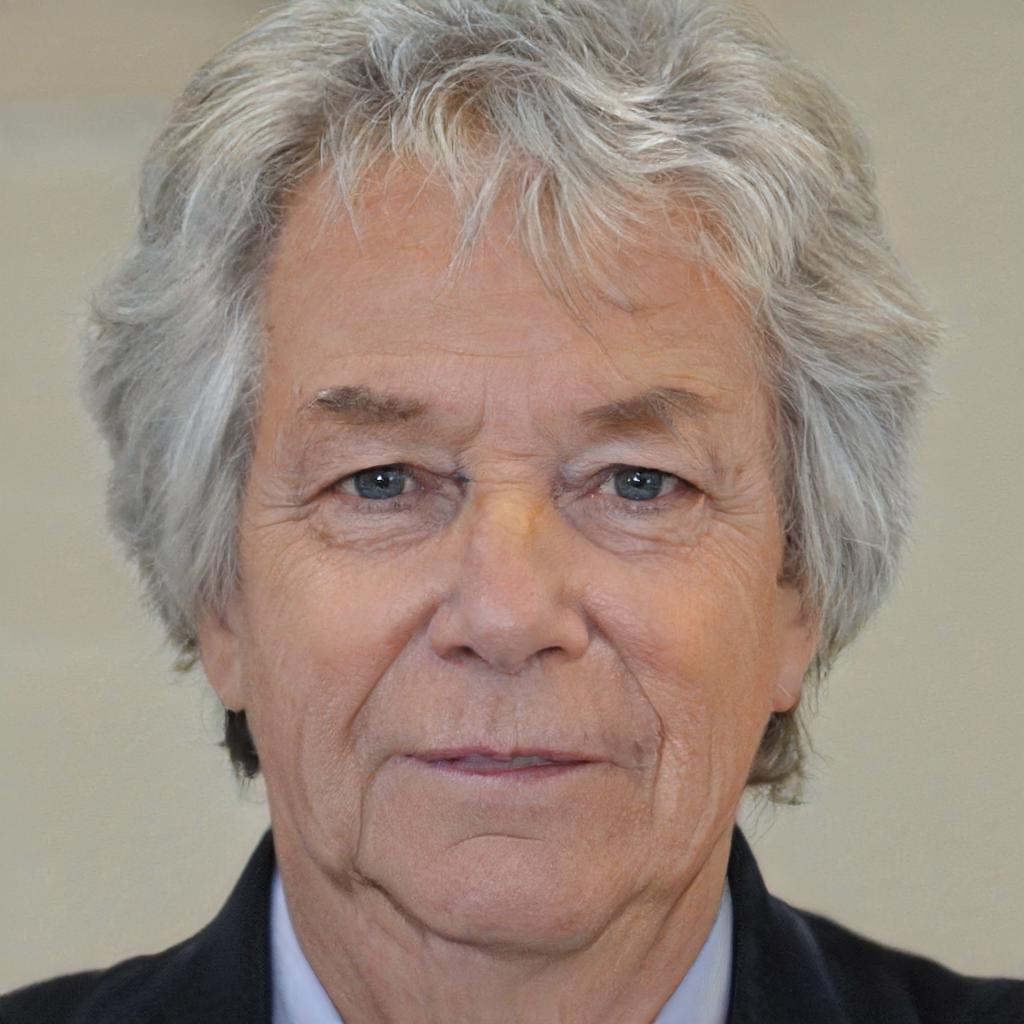} &
        \includegraphics[width=0.23\linewidth]
        {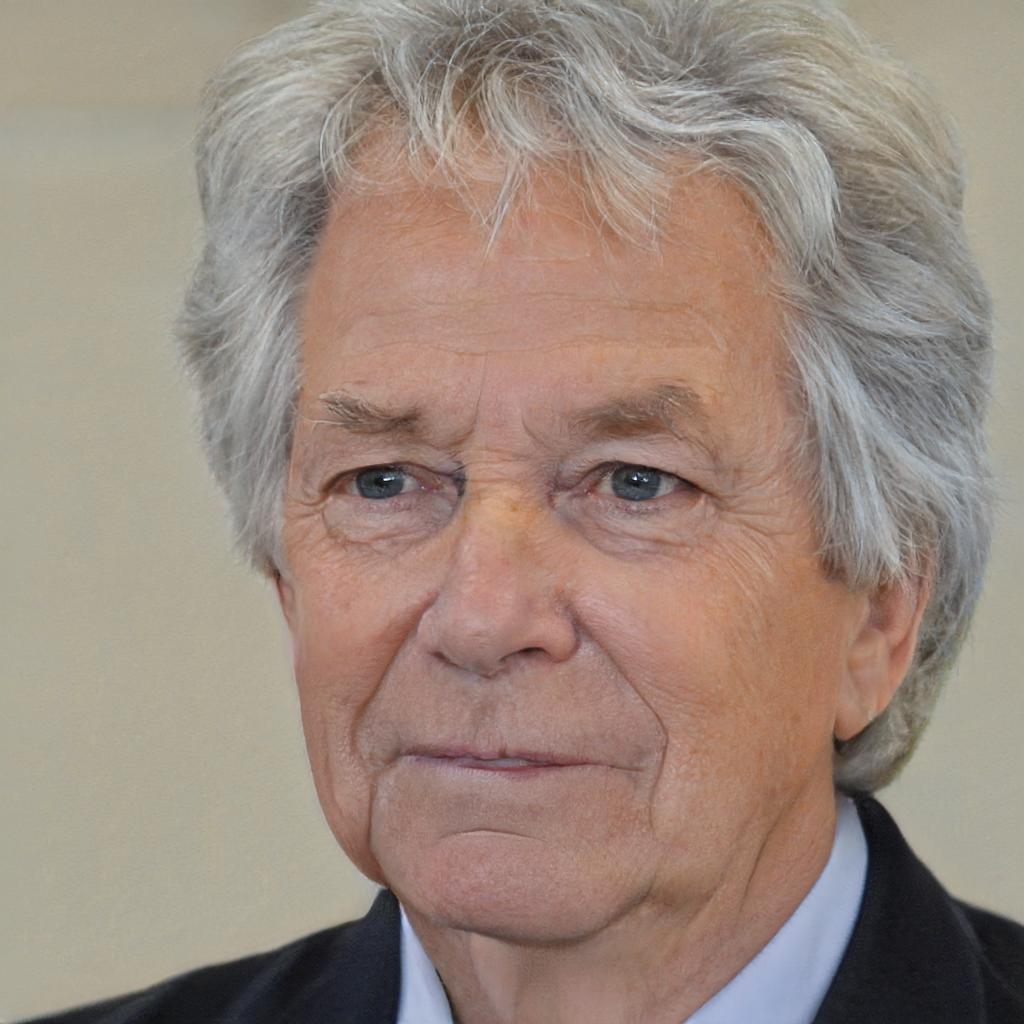}
        \tabularnewline
        \raisebox{0.275in}{\rotatebox[origin=t]{90}{Hair}} &&
        \includegraphics[width=0.23\linewidth]
        {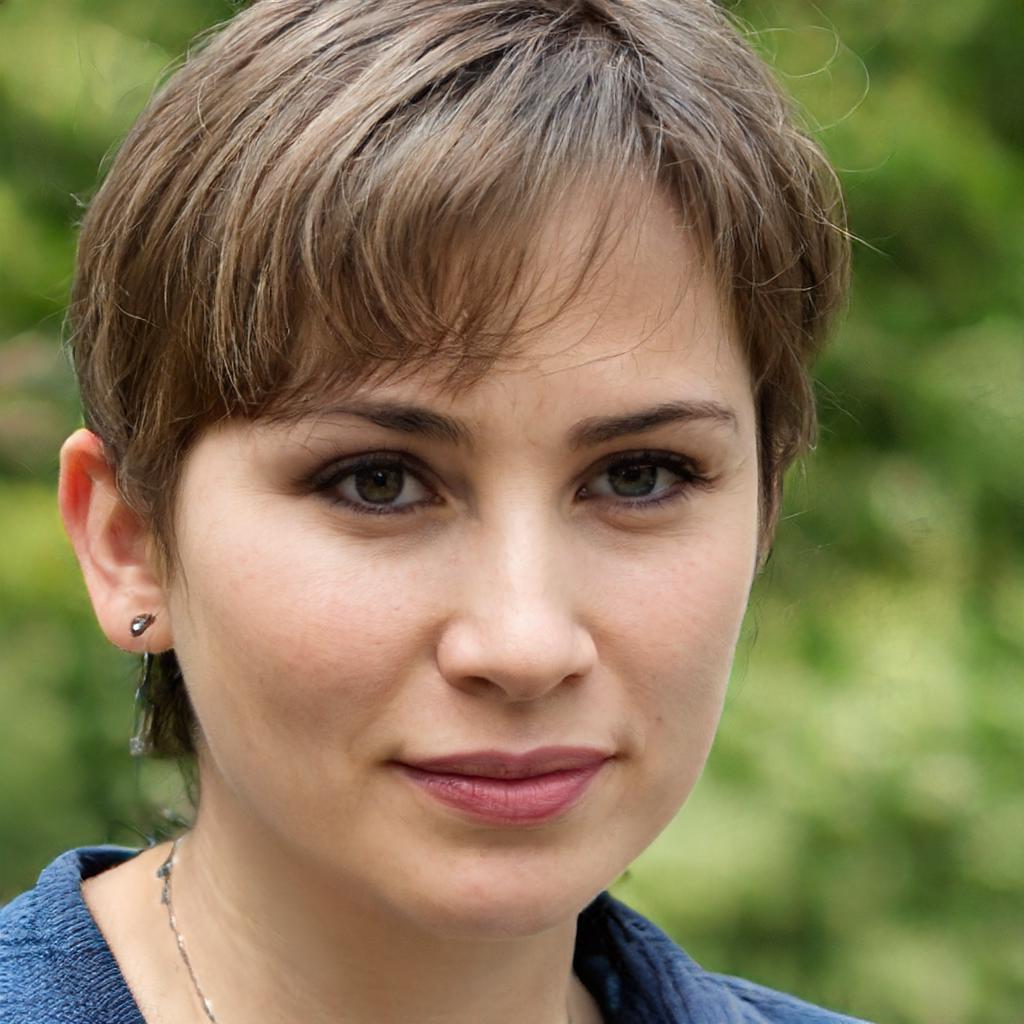} &
        \includegraphics[width=0.23\linewidth]
        {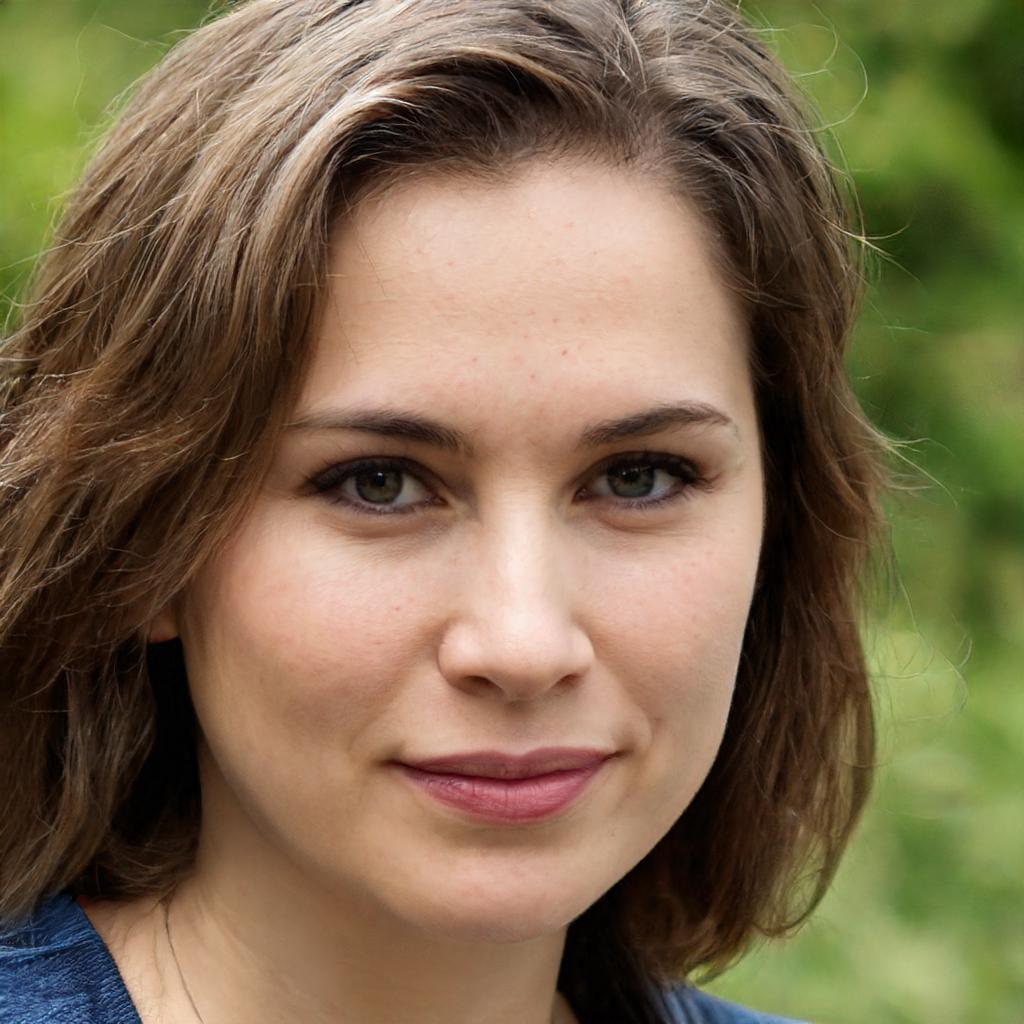} &
        \includegraphics[width=0.23\linewidth]
        {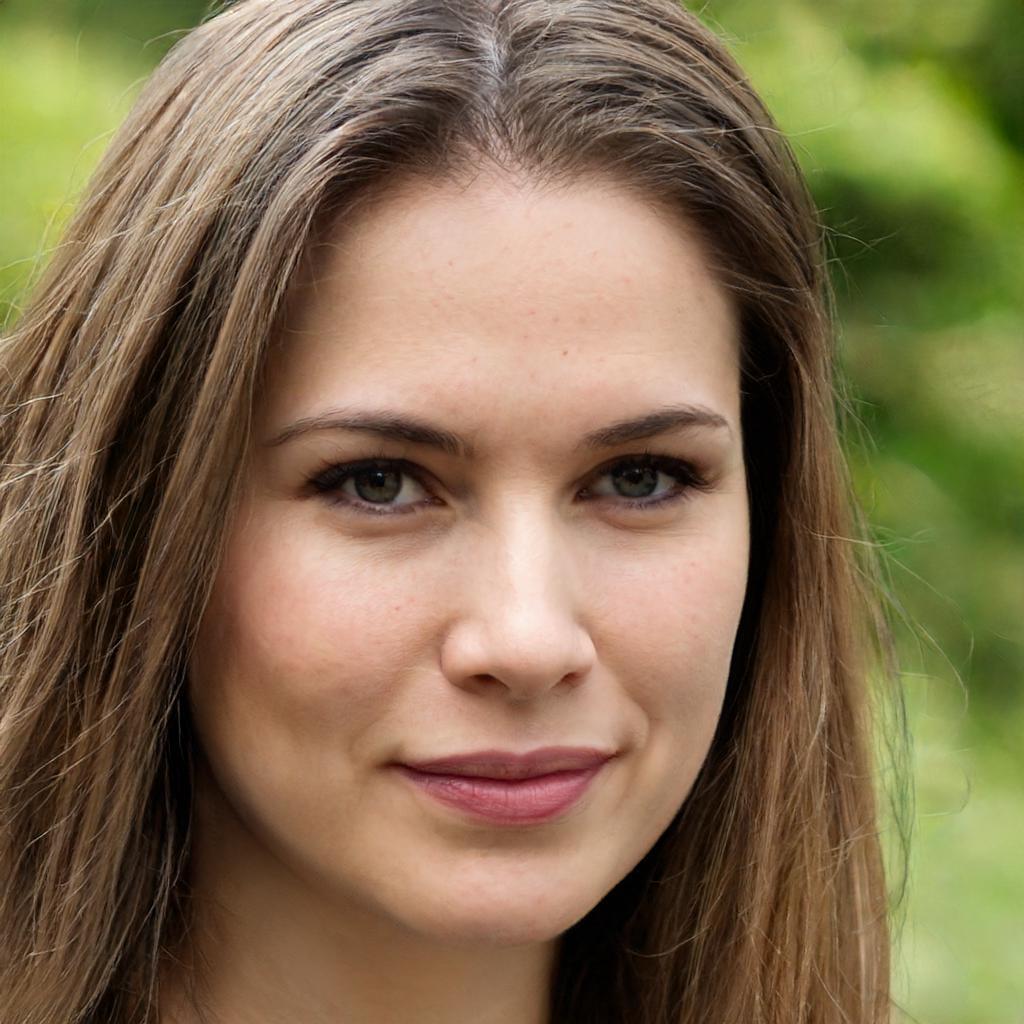} &
        \includegraphics[width=0.23\linewidth]
        {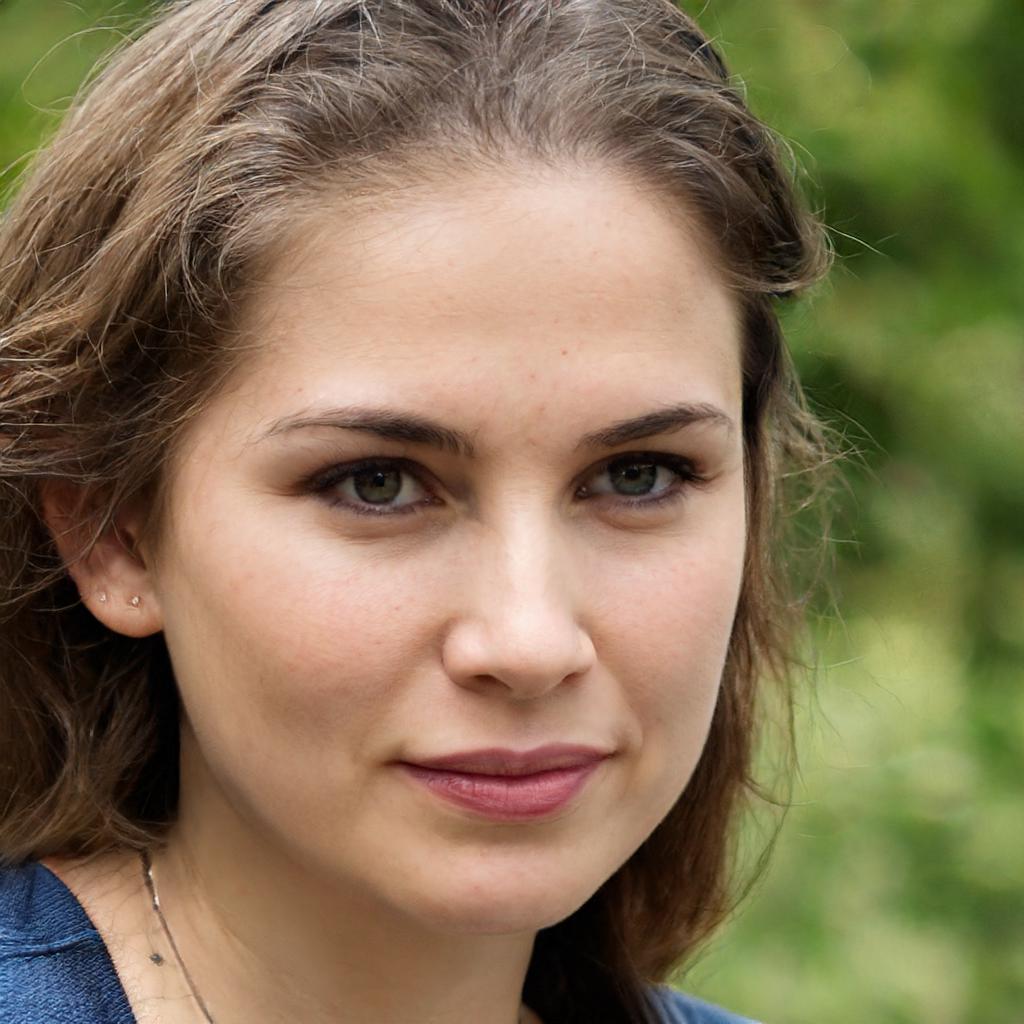}
        \tabularnewline
        \raisebox{0.215in}{\rotatebox[origin=t]{90}{Car Pose}} &&  
        \includegraphics[width=0.23\linewidth]
        {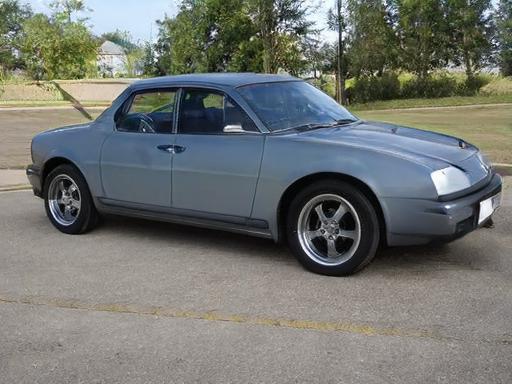} &
        \includegraphics[width=0.23\linewidth]
        {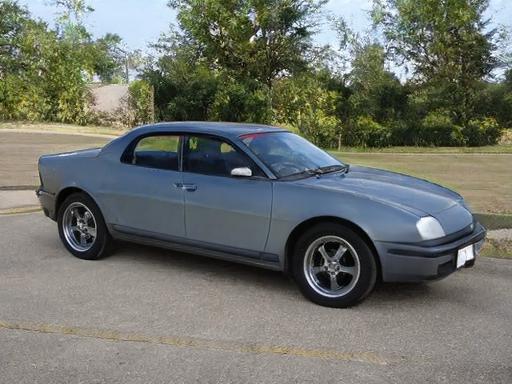} &
        \includegraphics[width=0.23\linewidth]
        {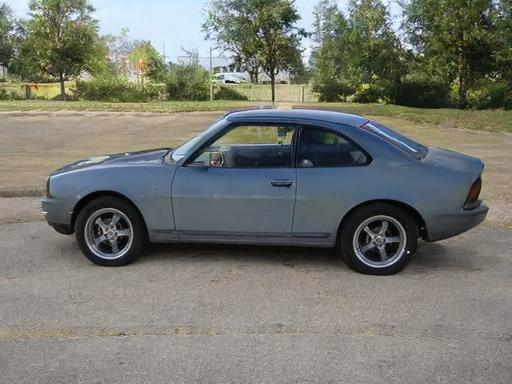} &
        \includegraphics[width=0.23\linewidth]
        {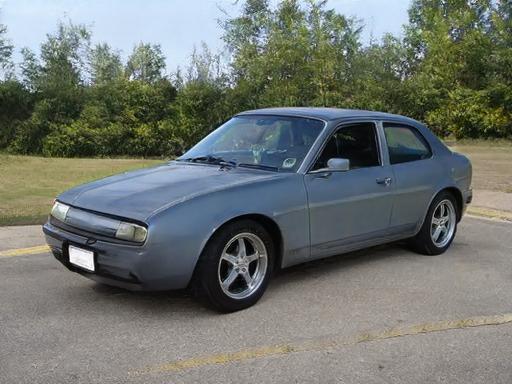}
        \tabularnewline
        \raisebox{0.275in}{\rotatebox[origin=t]{90}{Structure}} &&  
        \includegraphics[width=0.23\linewidth]
        {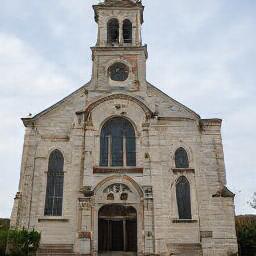} &
        \includegraphics[width=0.23\linewidth]
        {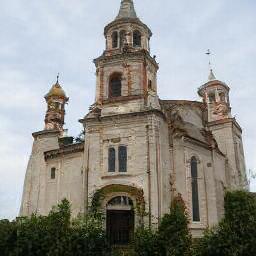} &
        \includegraphics[width=0.23\linewidth]
        {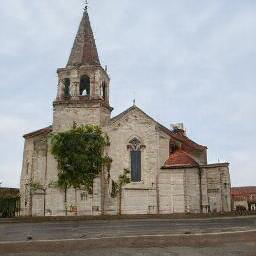} &
        \includegraphics[width=0.23\linewidth]
        {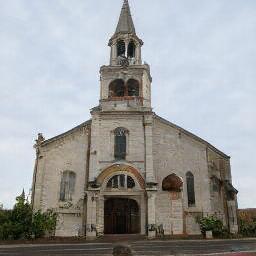}
        \tabularnewline
        \end{tabular}
    }
    \caption{\textit{Global control using StyleFusion.} We demonstrate the control gained over more global image attributes such as pose, hairstyle, and church structure by altering the global style code.
    Observe, for example, that altering these global features does not affect other details (e.g., identity or car wheels).}
    \label{fig:global_control}
\end{figure}

\begin{figure}
    \setlength{\tabcolsep}{0pt}
    \centering
        \begin{tabular}{c c c c c}
        \includegraphics[width=0.095\textwidth]
        {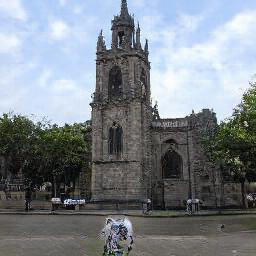} &
        \includegraphics[width=0.095\textwidth]
        {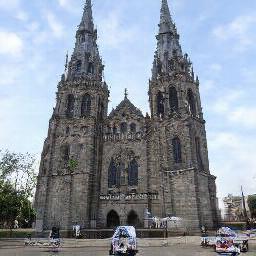} &
        \includegraphics[width=0.095\textwidth]
        {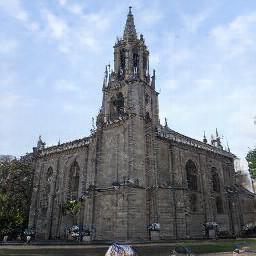} &
        \includegraphics[width=0.095\textwidth]
        {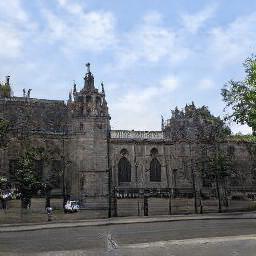} &
        \includegraphics[width=0.095\textwidth]
        {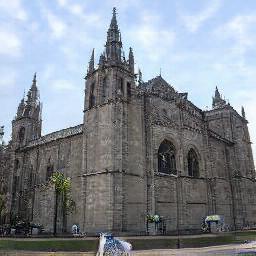}
        \tabularnewline
        \includegraphics[width=0.095\textwidth]
        {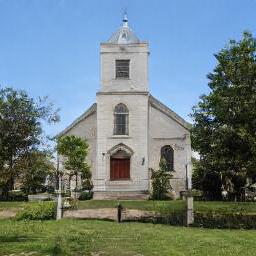} &
        \includegraphics[width=0.095\textwidth]
        {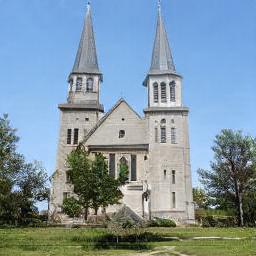} &
        \includegraphics[width=0.095\textwidth]
        {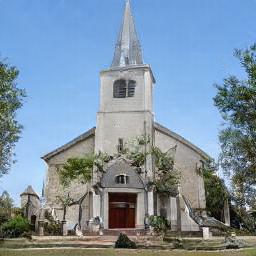} &
        \includegraphics[width=0.095\textwidth]
        {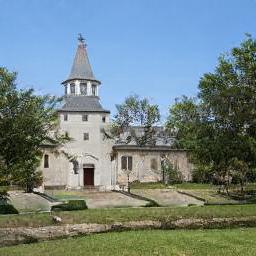} &
        \includegraphics[width=0.095\textwidth]
        {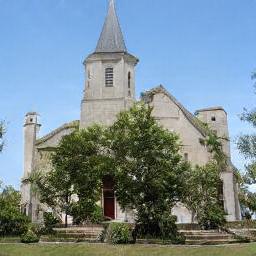}
        \tabularnewline
        \includegraphics[width=0.095\textwidth]
        {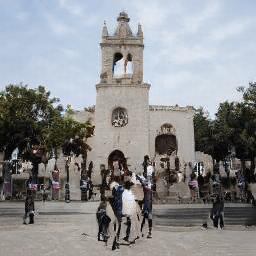} &
        \includegraphics[width=0.095\textwidth]
        {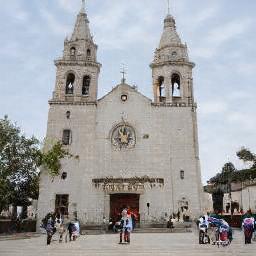} &
        \includegraphics[width=0.095\textwidth]
        {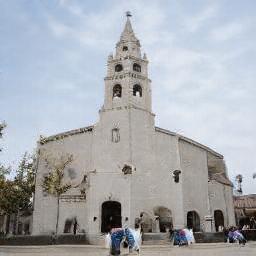} &
        \includegraphics[width=0.095\textwidth]
        {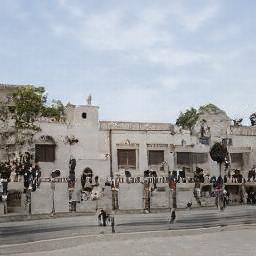} &
        \includegraphics[width=0.095\textwidth]
        {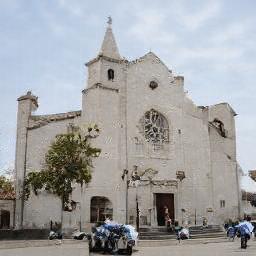}
        \tabularnewline
        \includegraphics[width=0.095\textwidth]
        {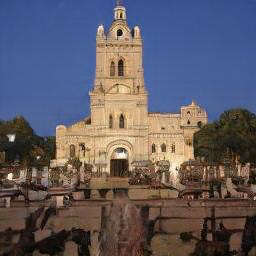} &
        \includegraphics[width=0.095\textwidth]
        {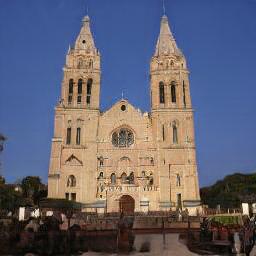} &
        \includegraphics[width=0.095\textwidth]
        {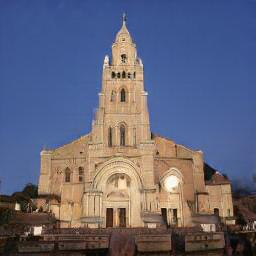} &
        \includegraphics[width=0.095\textwidth]
        {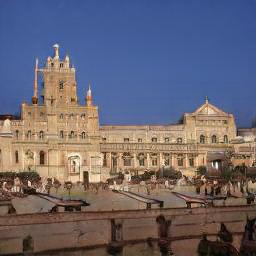} &
        \includegraphics[width=0.095\textwidth]
        {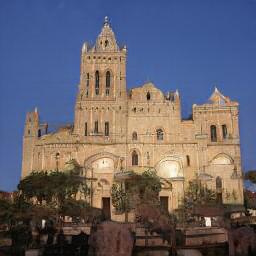}
        \tabularnewline
        \end{tabular}
    \caption{\textit{Global control using StyleFusion on the churches domain}. Similar to Figure ~\ref{fig:global_control}, we demonstrate that StyleFusion is able to generate multiple church buildings of the same style, but different structure. Notably, observe the similarity in the church layout for each of the columns shown.}
    \label{fig:global_control_churches}
\end{figure}

\topic{\textit{Global Manipulation.}}
We now turn to analyze the role of the global latent code when learning to disentangle the different semantic image regions. Recall that the global latent code is tasked with aligning the location of semantic regions of one image with those of a second image. Here, we demonstrate how StyleFusion provides users with control over global attributes such as head pose, car pose, and church structure by altering this global code. Specifically, consider Figure~\ref{fig:global_control} where we demonstrate control over such features. Here, different global variations are created by randomly sampling different global latent codes as input to StyleFusion (with the remaining input latent codes remain fixed). 
Observe StyleFusion's ability to change the pose of the car while faithfully preserving its car body and small details such as the wheels. Interestingly, StyleFusion is able to complete missing details when altering the car pose. This demonstrates the versatility of StyleFusion in that it facilitates control over both local and global aspects of the generated image, even in the absence of specific details.
Additionally, altering the global code allows one to generate church buildings of the same style but different structure, as shown in Figure~\ref{fig:global_control_churches}.

\section{Applications}

\begin{figure}
    \setlength{\tabcolsep}{1pt}
    \centering
    \small{
        \begin{tabular}{C{0.18\linewidth} c C{0.18\linewidth} C{0.18\linewidth} C{0.18\linewidth} C{0.18\linewidth}}
        
        & & StyleCLIP & InterFace & GANSpace & GANSpace \\
        
        \includegraphics[width=\linewidth]{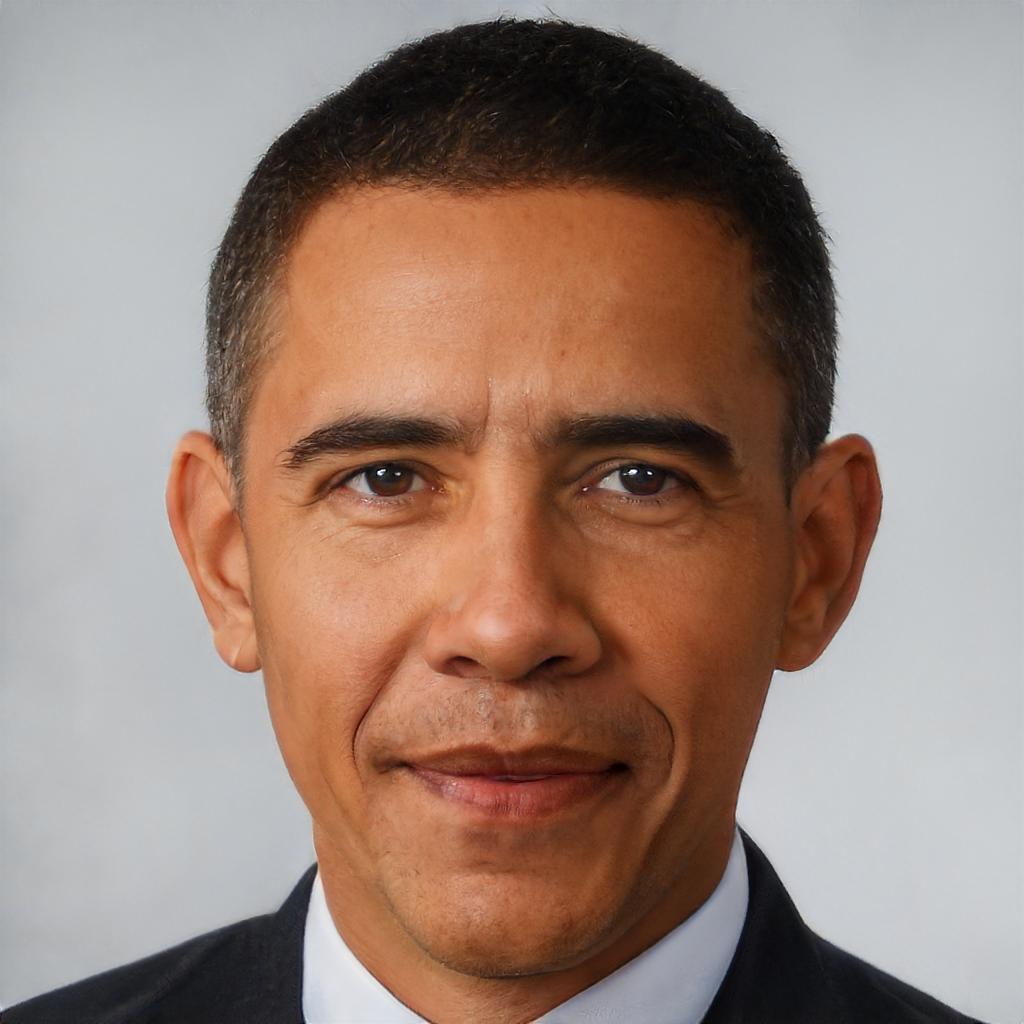} &
        \raisebox{0.22in}{\rotatebox[origin=t]{90}{\footnotesize StyleGAN2}} &
        \includegraphics[width=\linewidth]{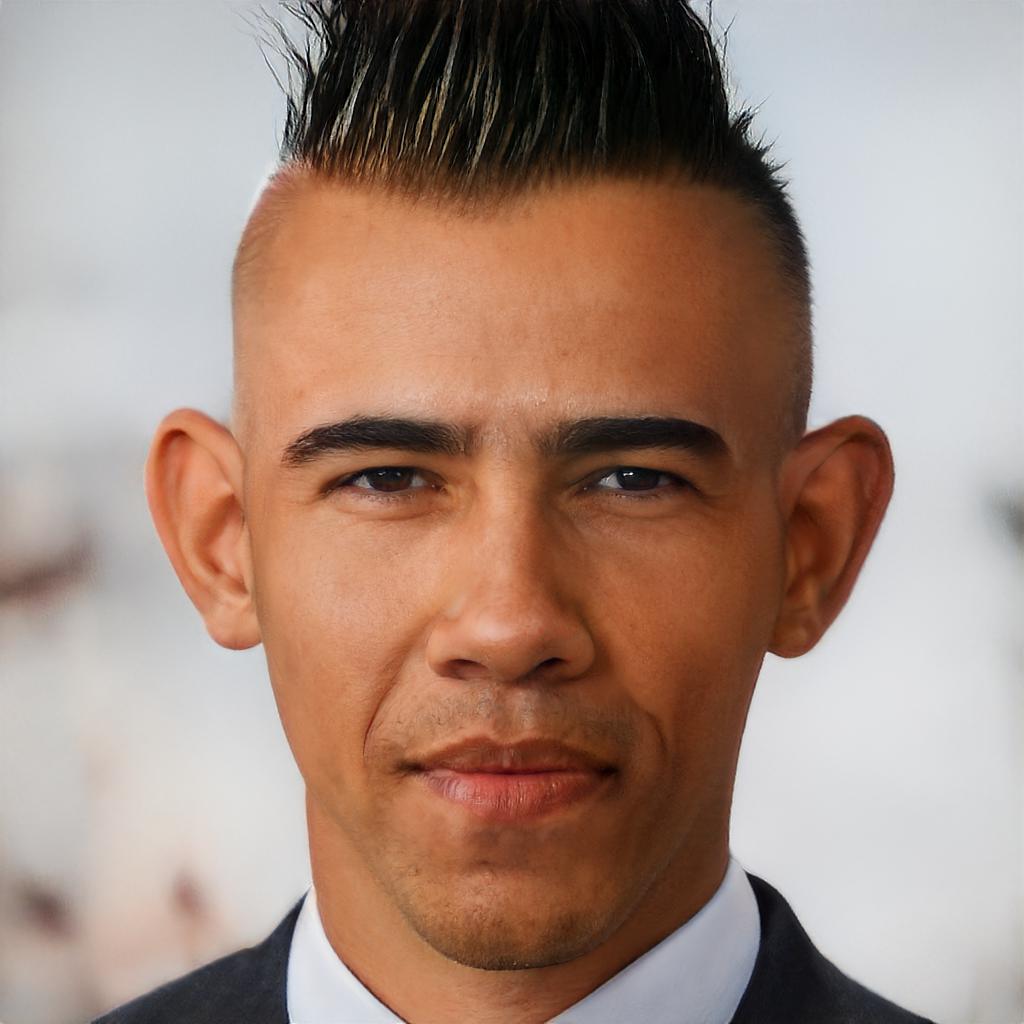} &
        \includegraphics[width=\linewidth]{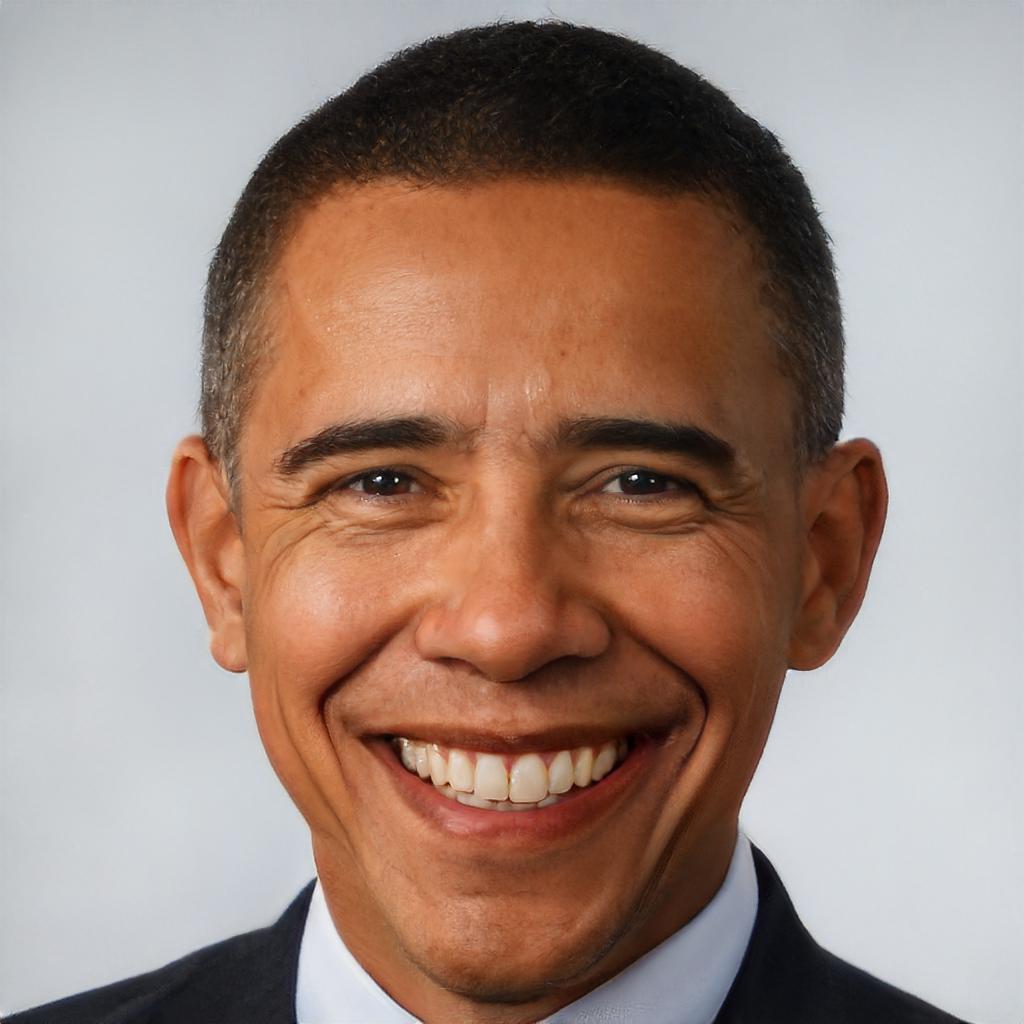} &
        \includegraphics[width=\linewidth]{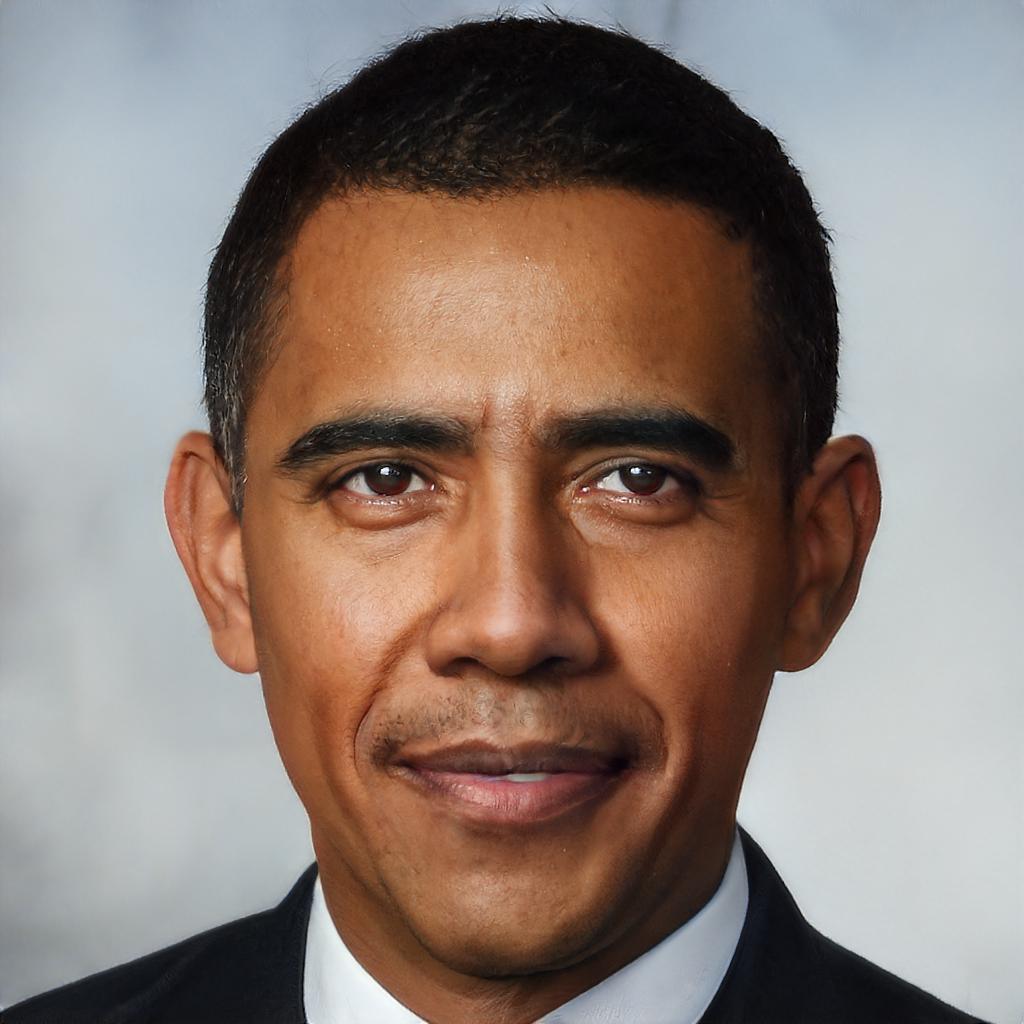} &
        \includegraphics[width=\linewidth]{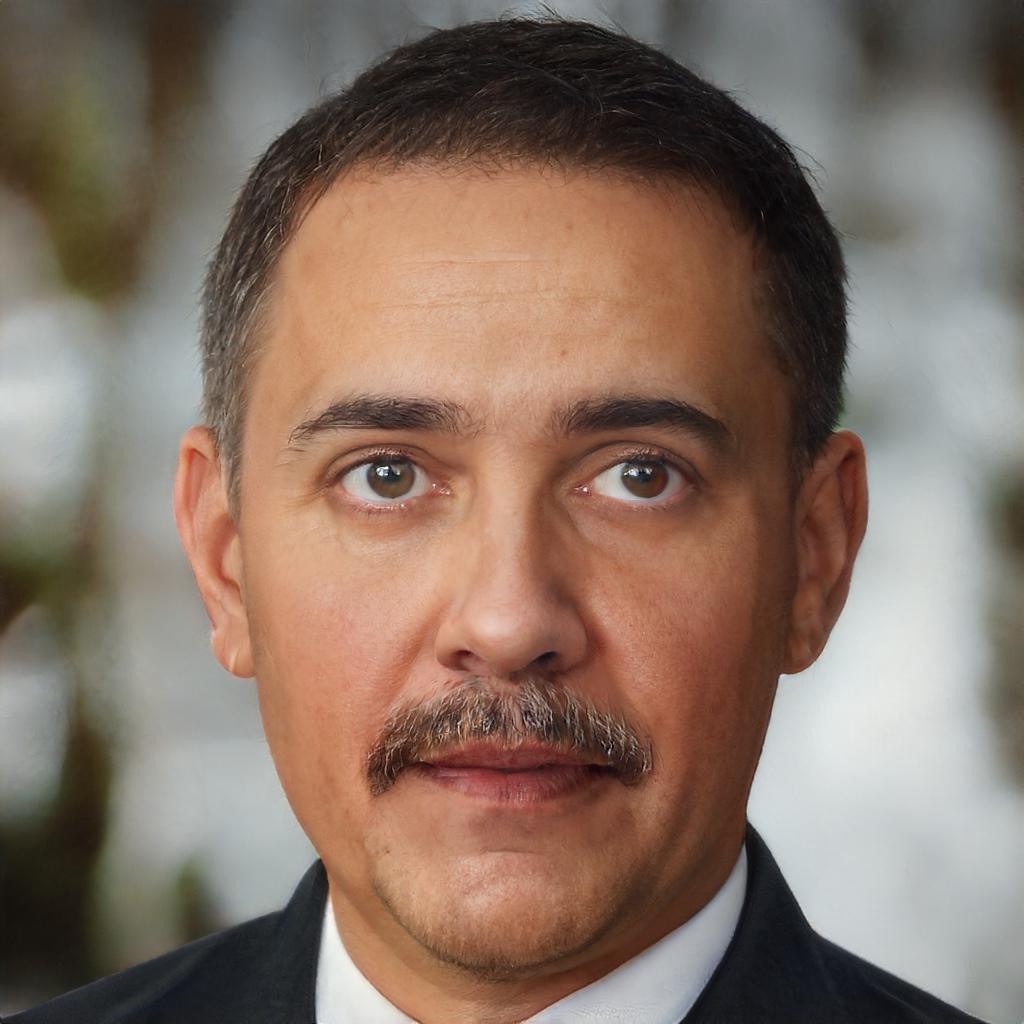} 
        \tabularnewline
        & \raisebox{0.22in}{\rotatebox[origin=t]{90}{\footnotesize StyleFusion}} &
        \includegraphics[width=\linewidth]{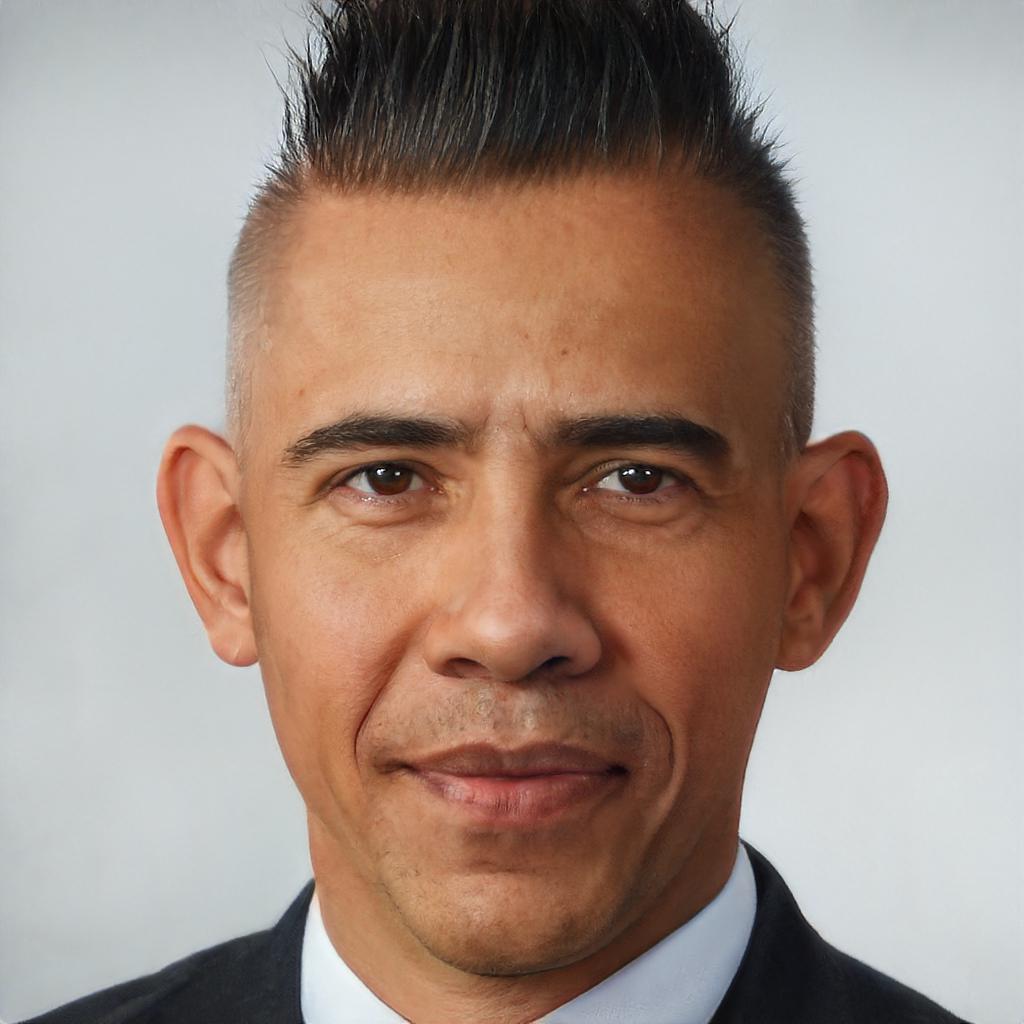} &
        \includegraphics[width=\linewidth]{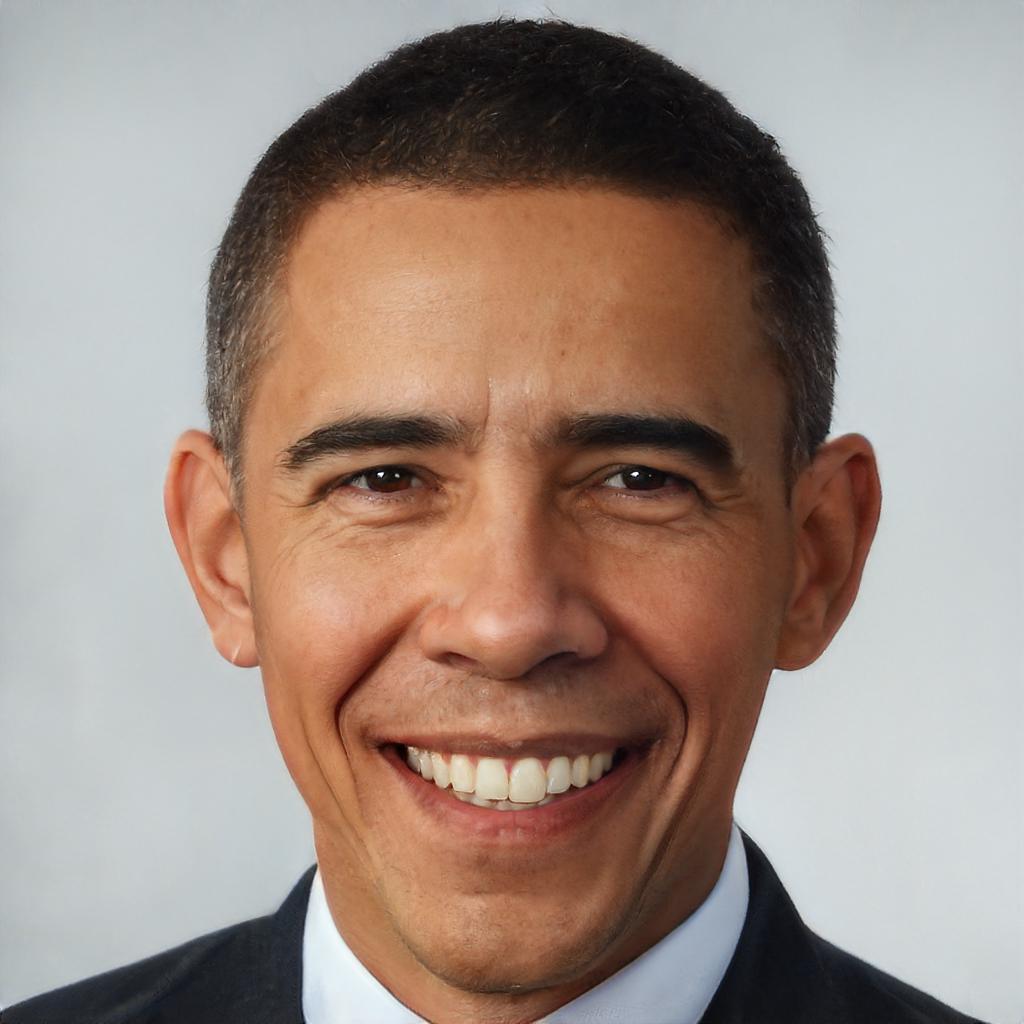} &
        \includegraphics[width=\linewidth]{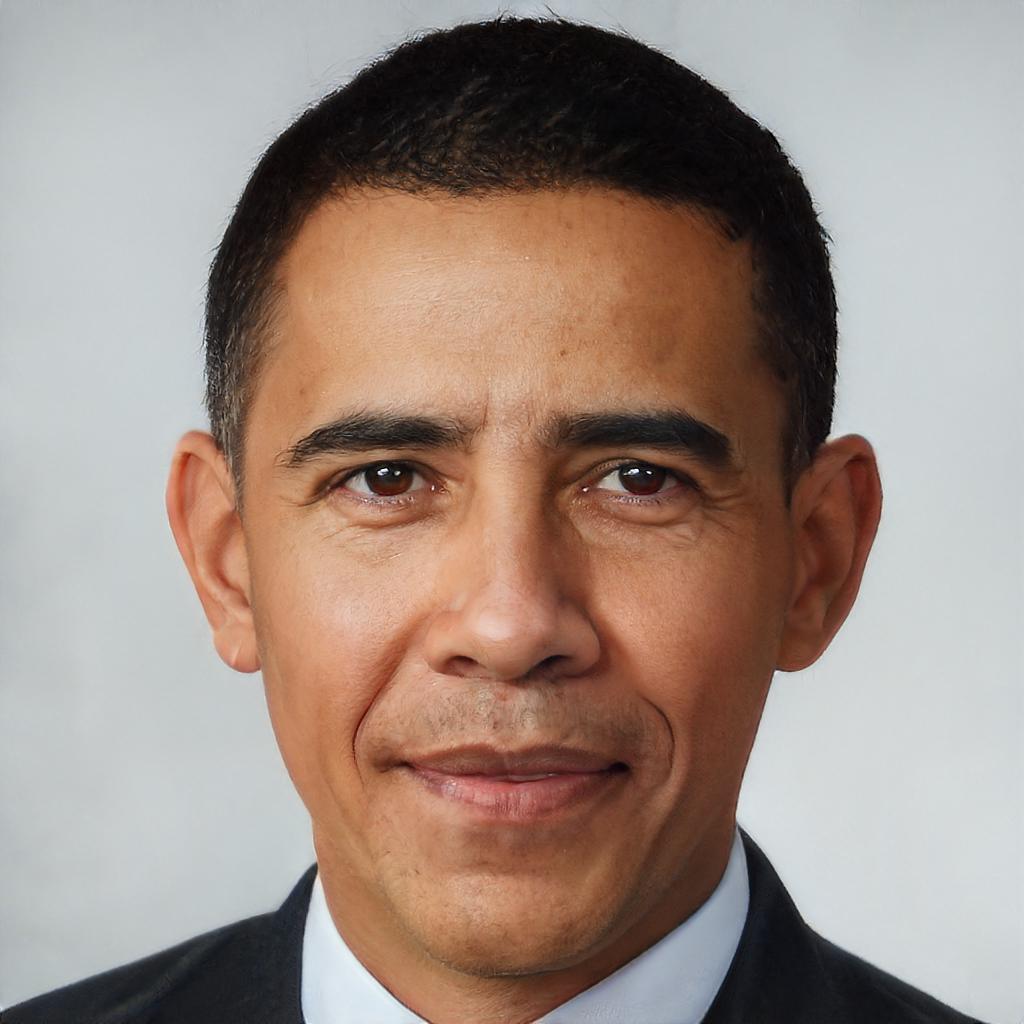} &
        \includegraphics[width=\linewidth]{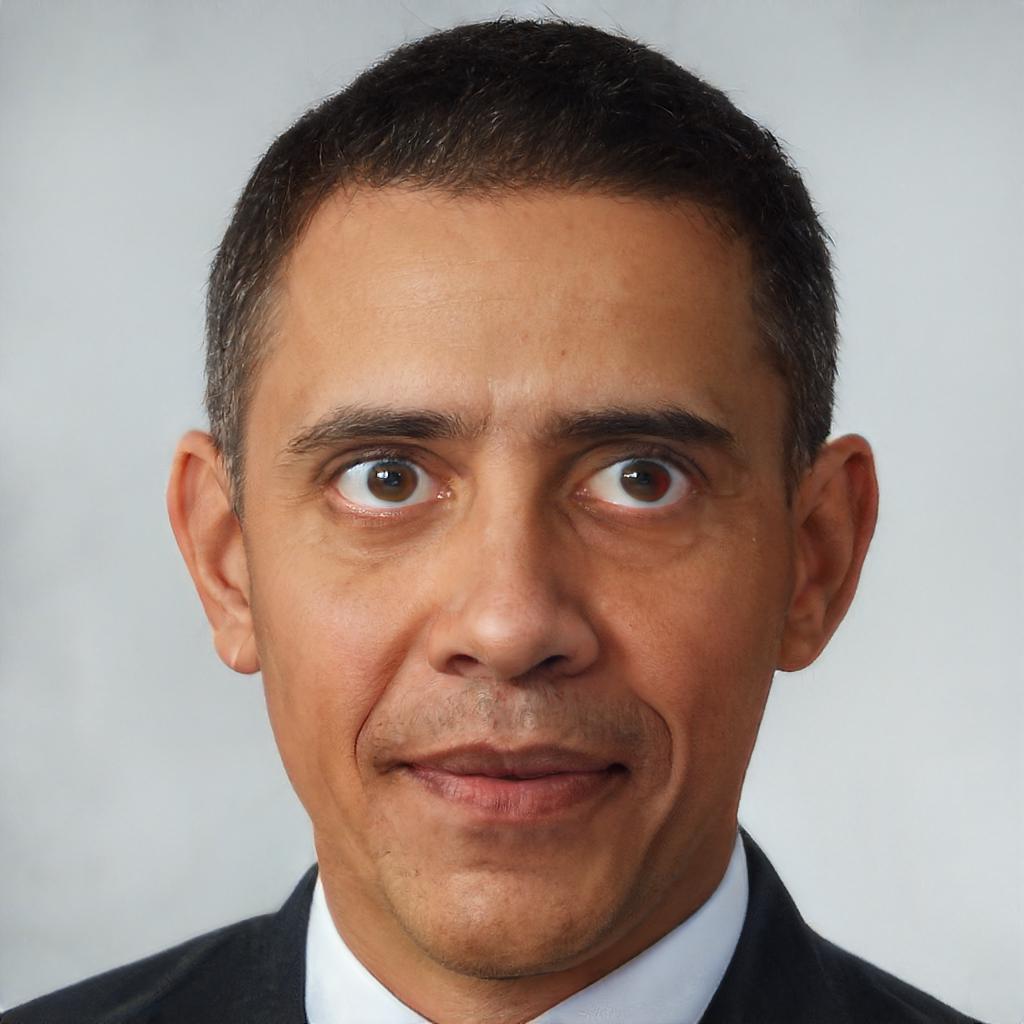} 
        \tabularnewline
        
        \includegraphics[width=\linewidth]{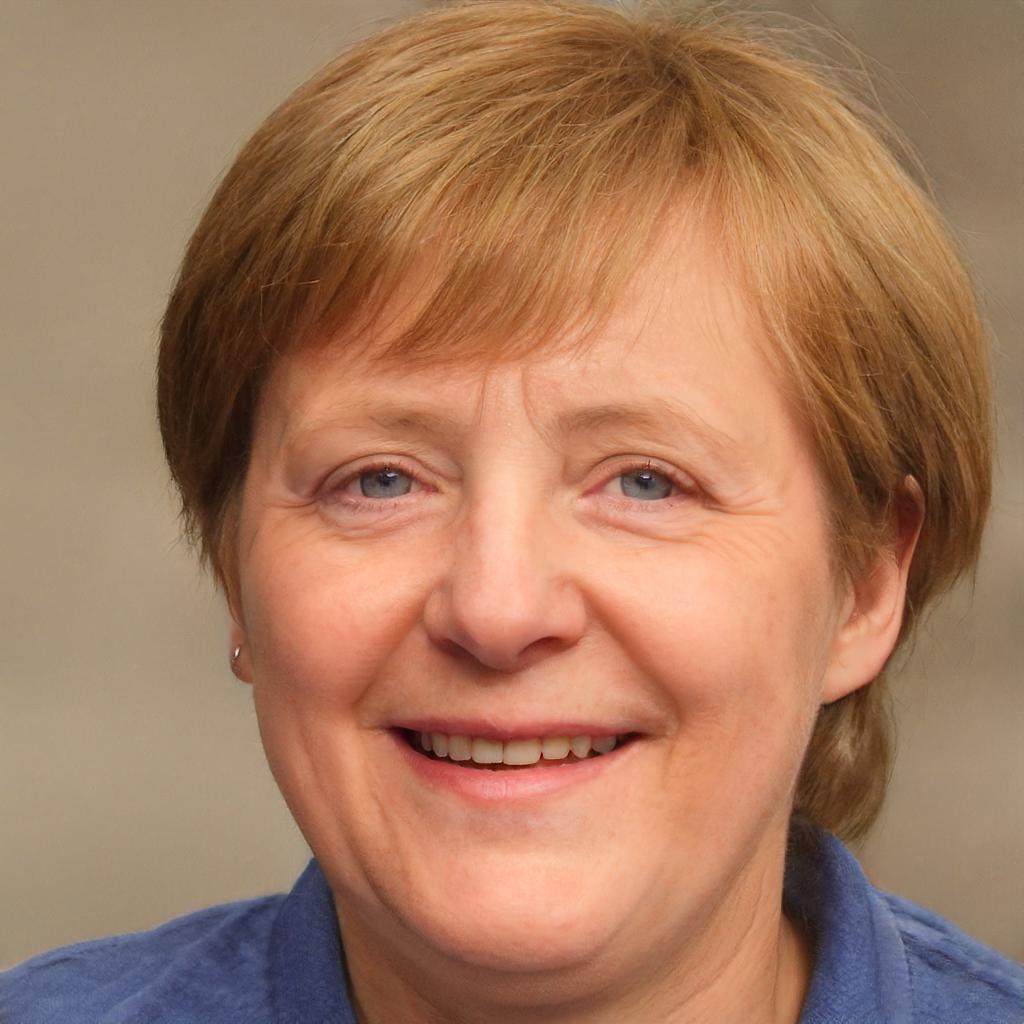} &
        \raisebox{0.22in}{\rotatebox[origin=t]{90}{\footnotesize StyleGAN2}} &
        \includegraphics[width=\linewidth]{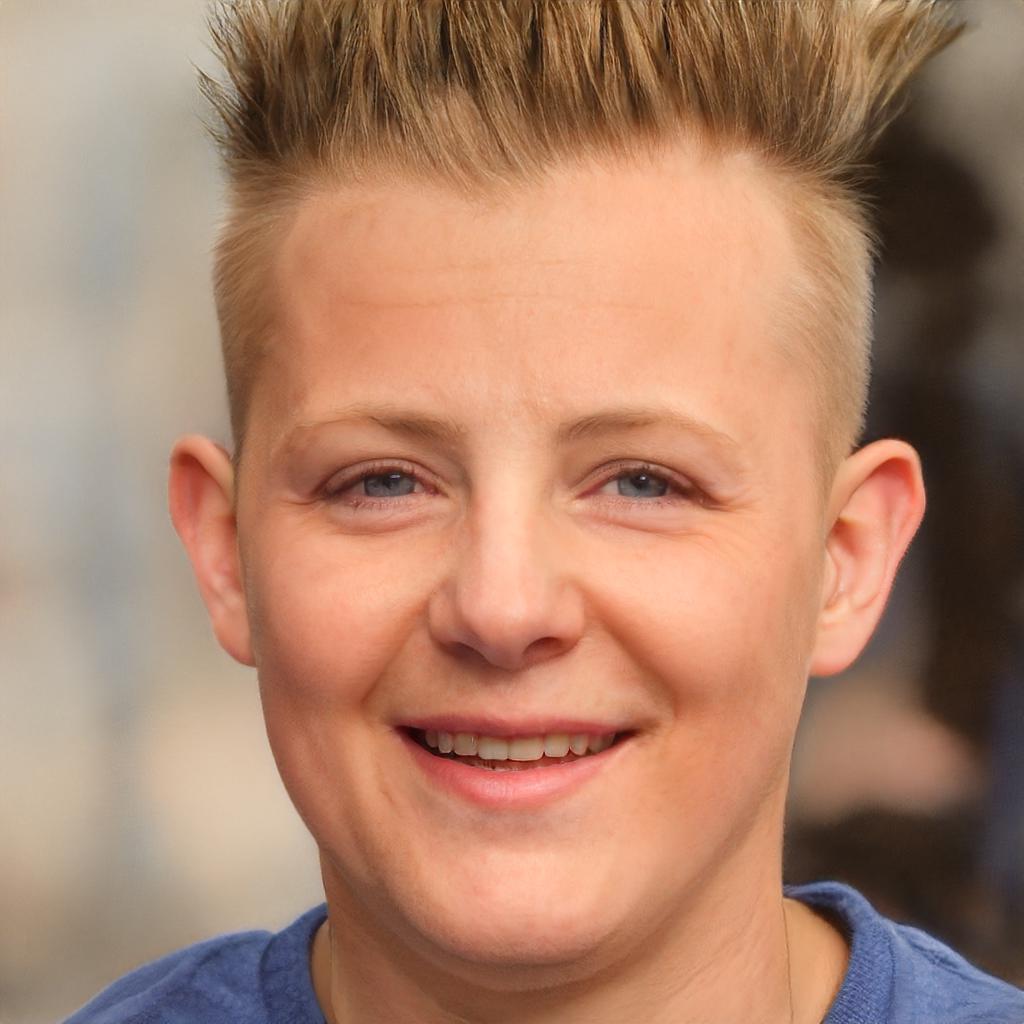} &
        \includegraphics[width=\linewidth]{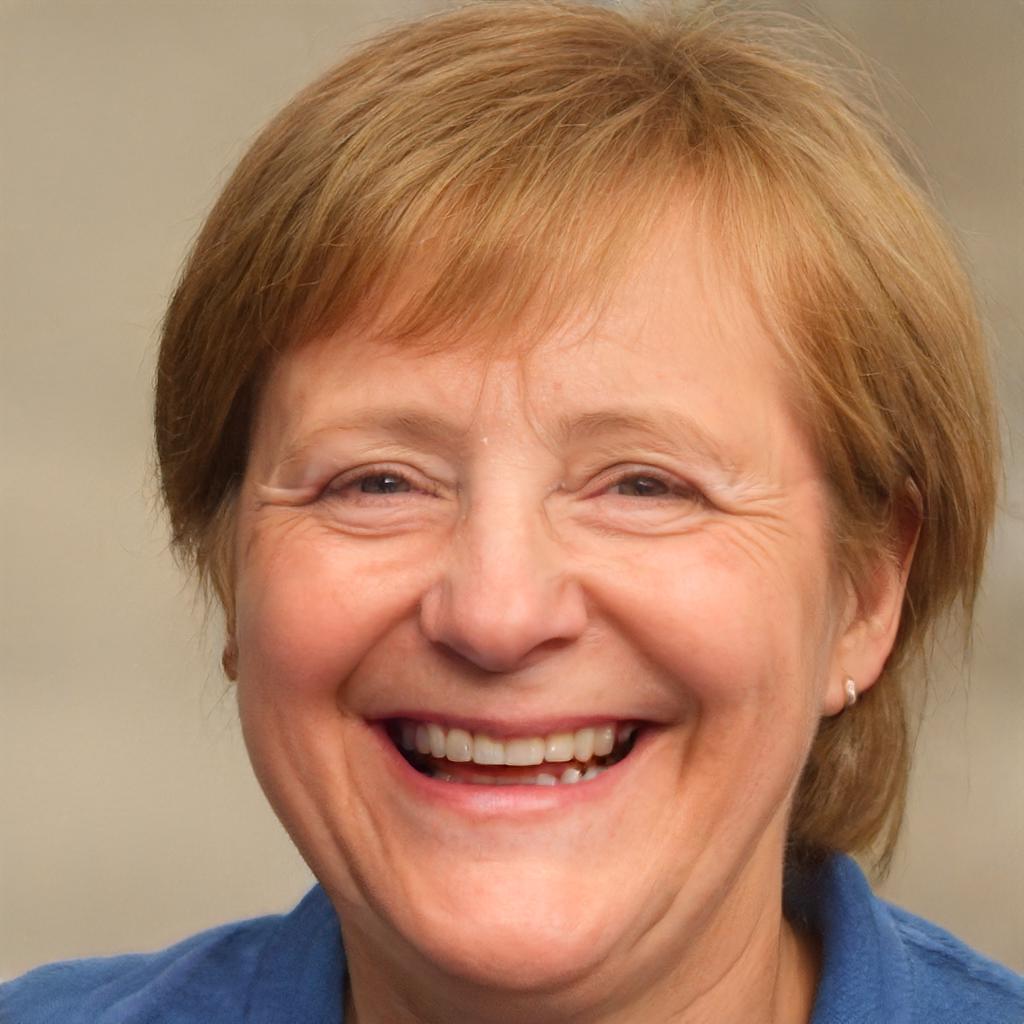} &
        \includegraphics[width=\linewidth]{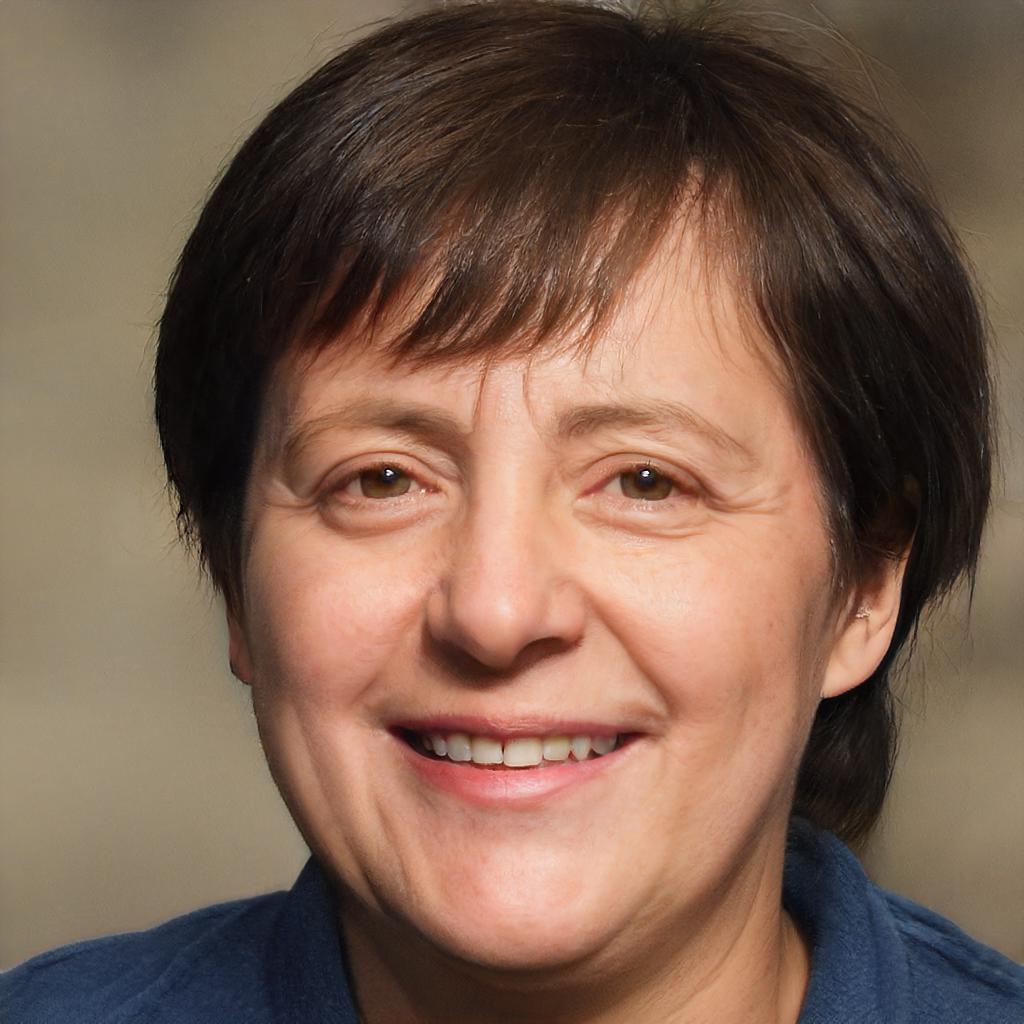} &
        \includegraphics[width=\linewidth]{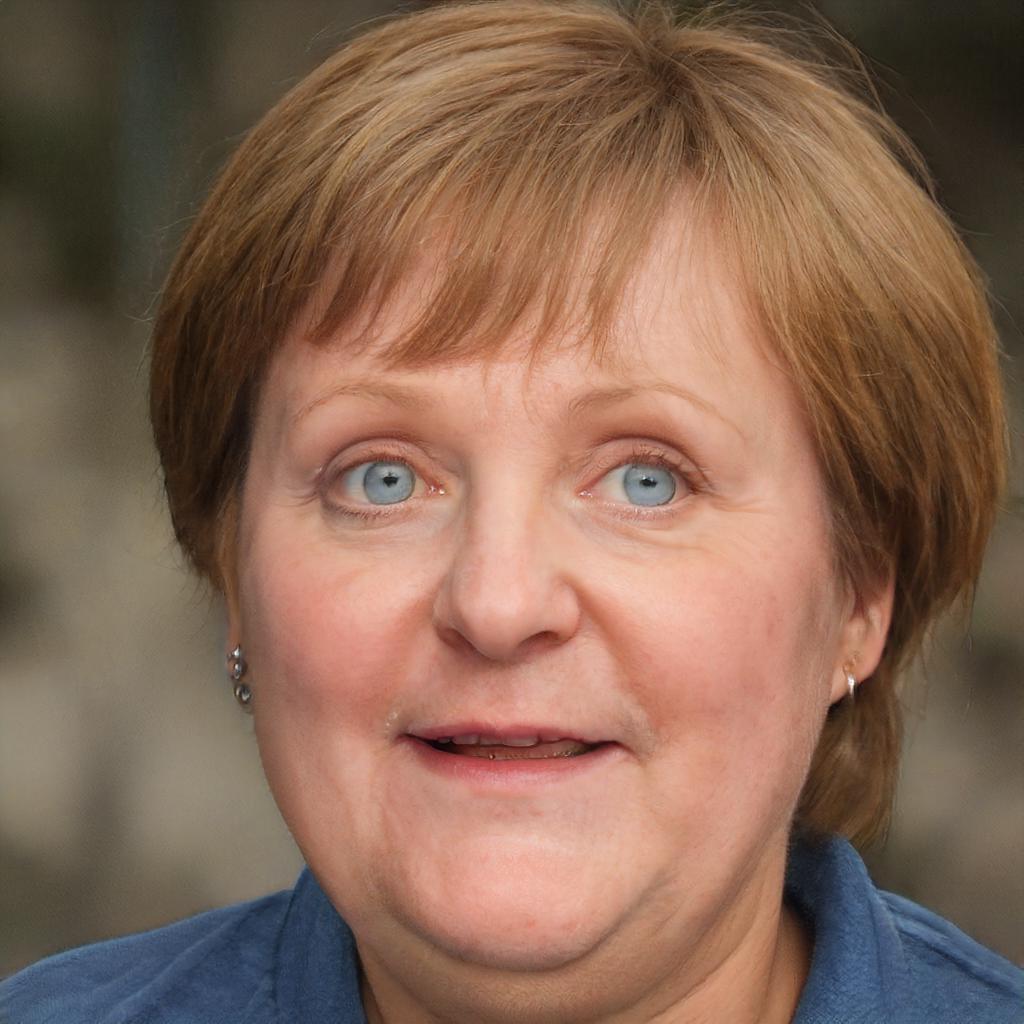} 
        \tabularnewline
        & \raisebox{0.22in}{\rotatebox[origin=t]{90}{\footnotesize StyleFusion}} &
        \includegraphics[width=\linewidth]{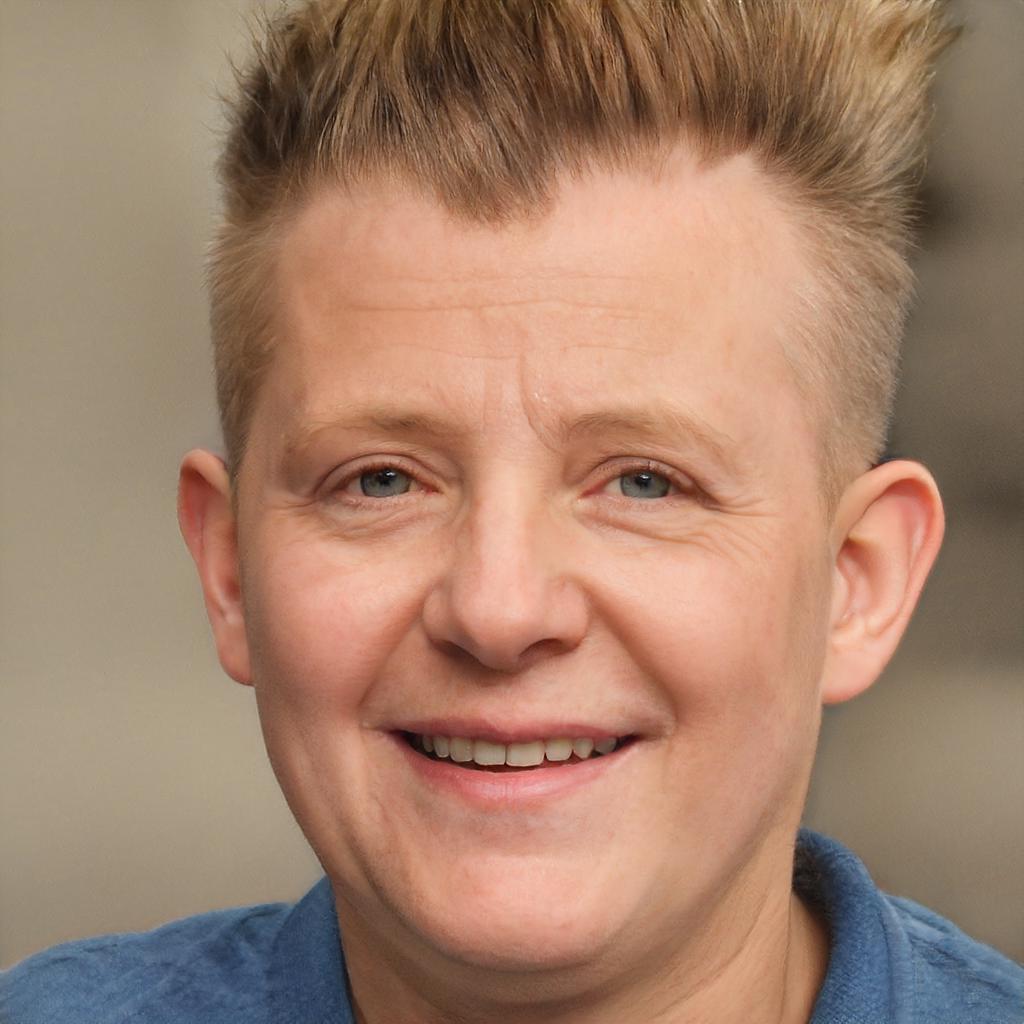} &
        \includegraphics[width=\linewidth]{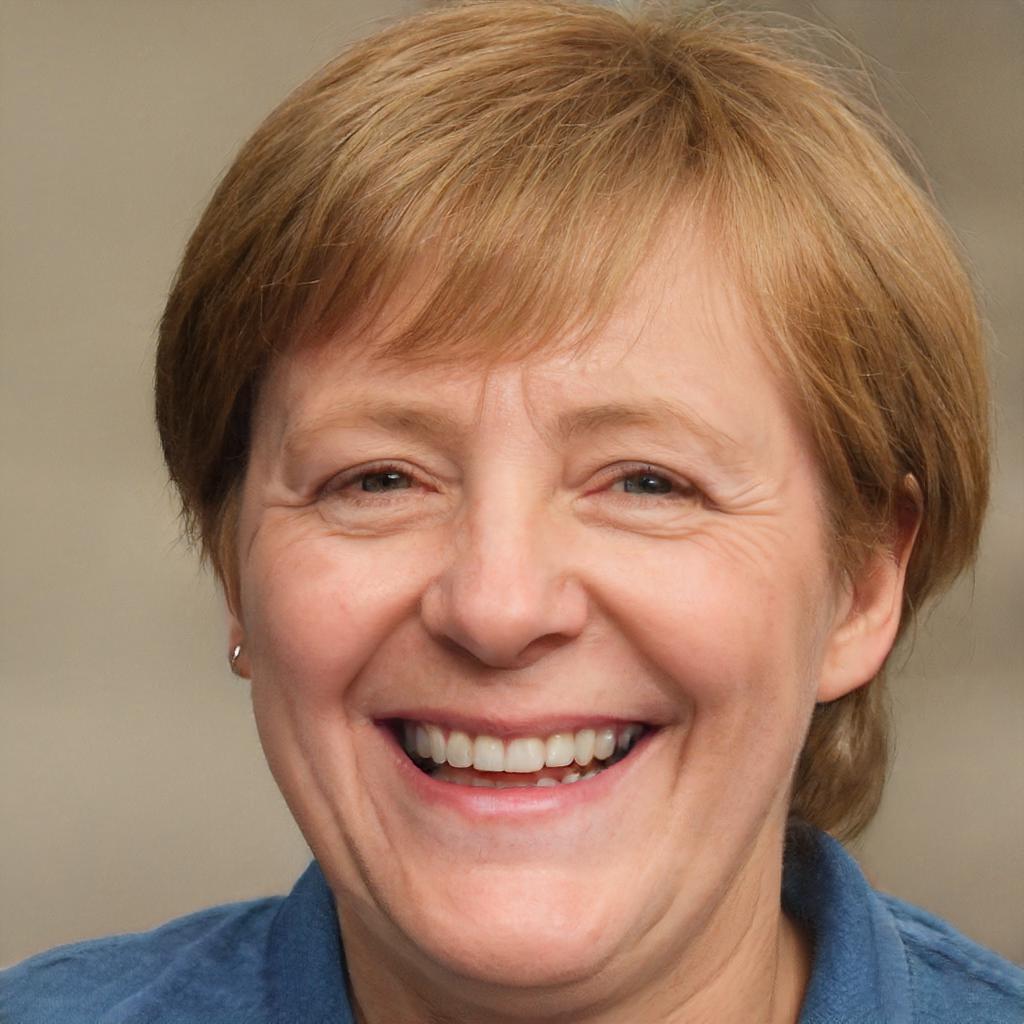} &
        \includegraphics[width=\linewidth]{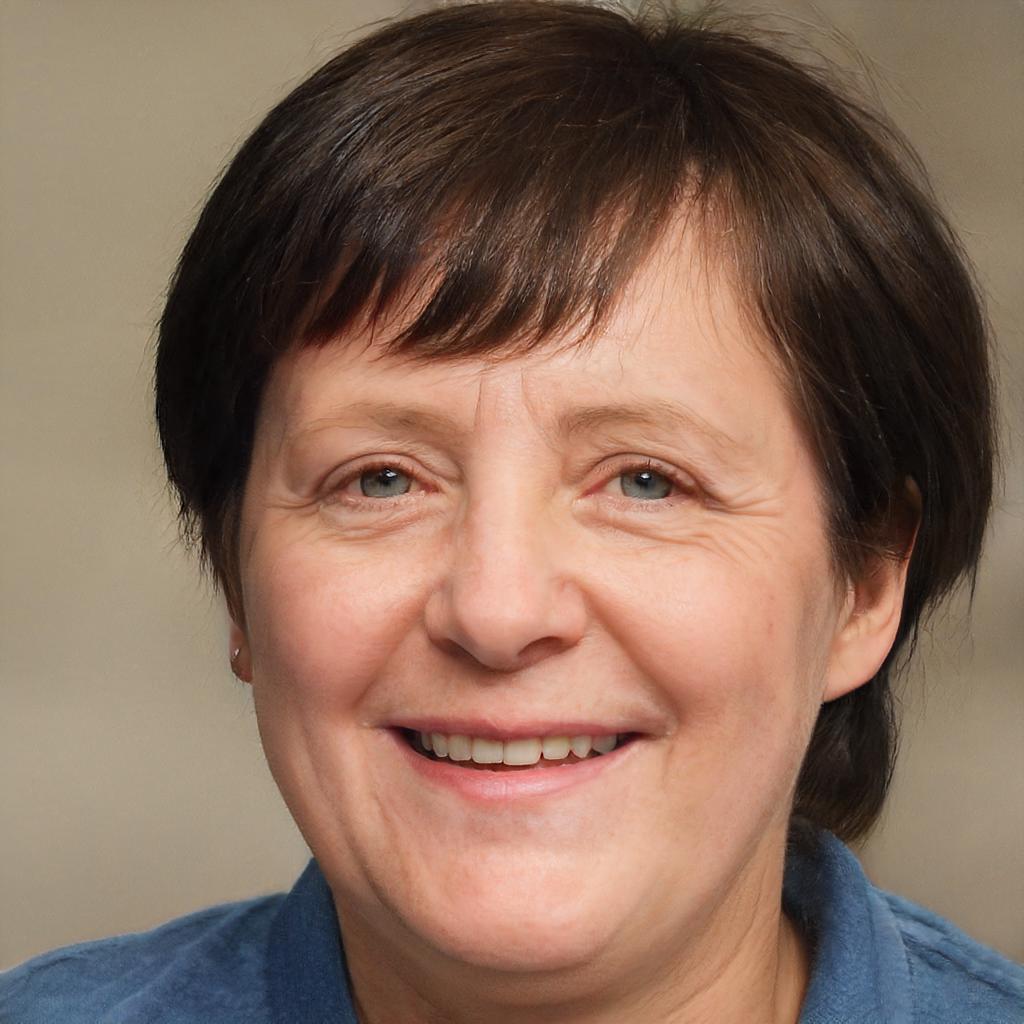} &
        \includegraphics[width=\linewidth]{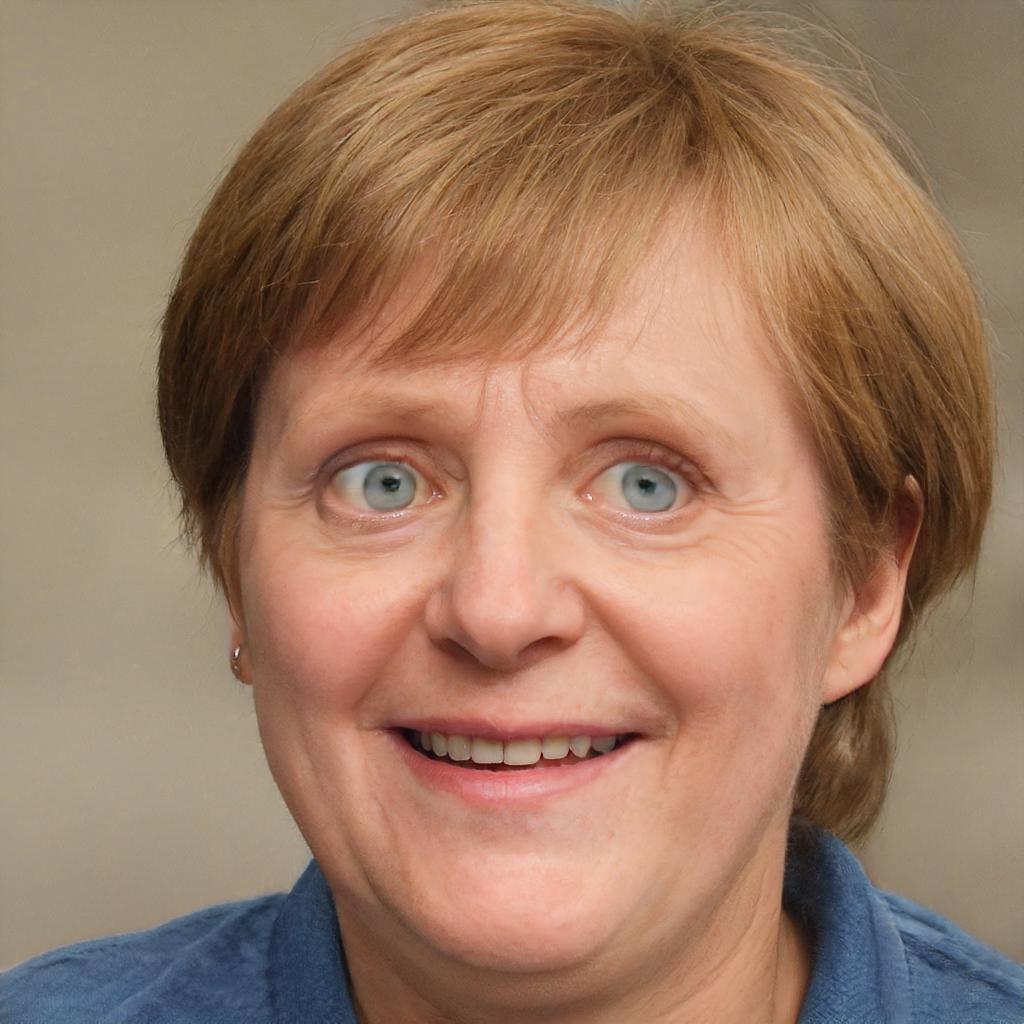}   
        \tabularnewline

        \includegraphics[width=\linewidth]{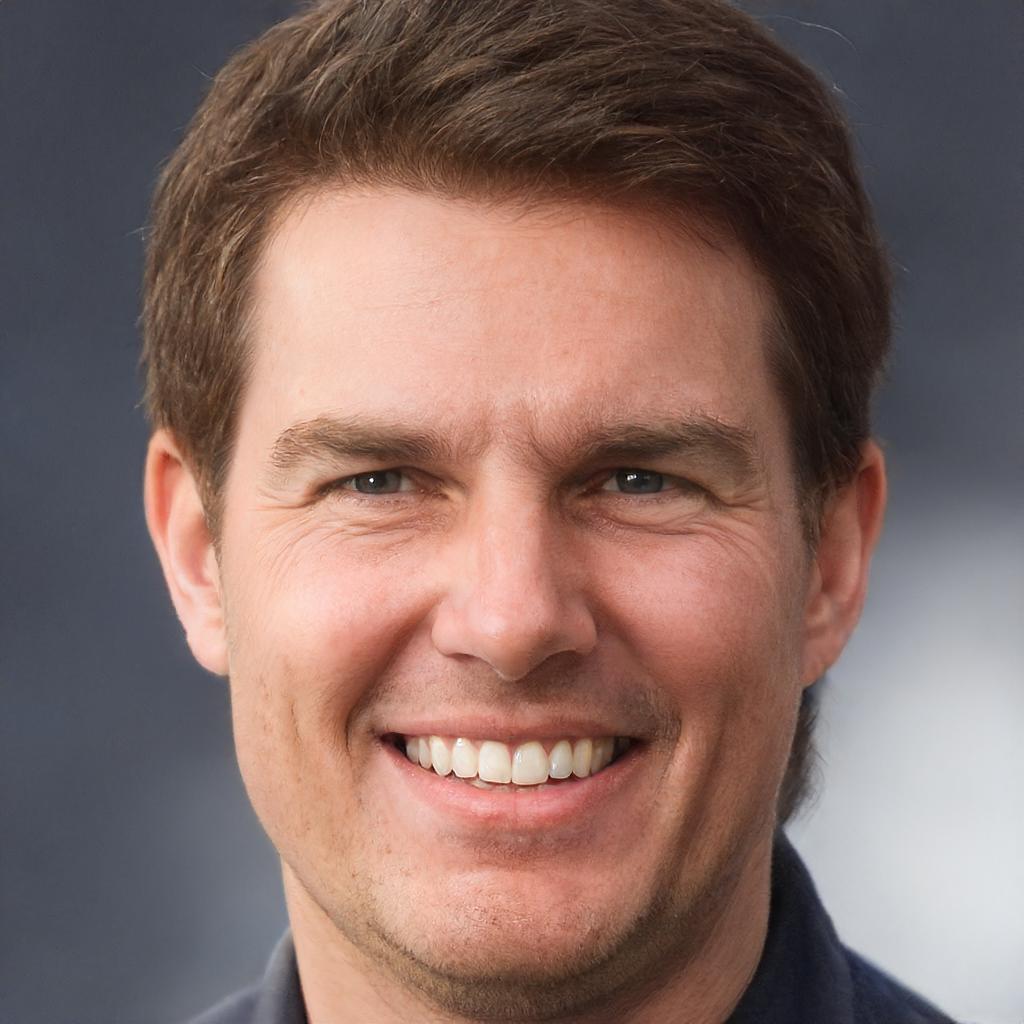} &
        \raisebox{0.22in}{\rotatebox[origin=t]{90}{\footnotesize StyleGAN2}} &
        \includegraphics[width=\linewidth]{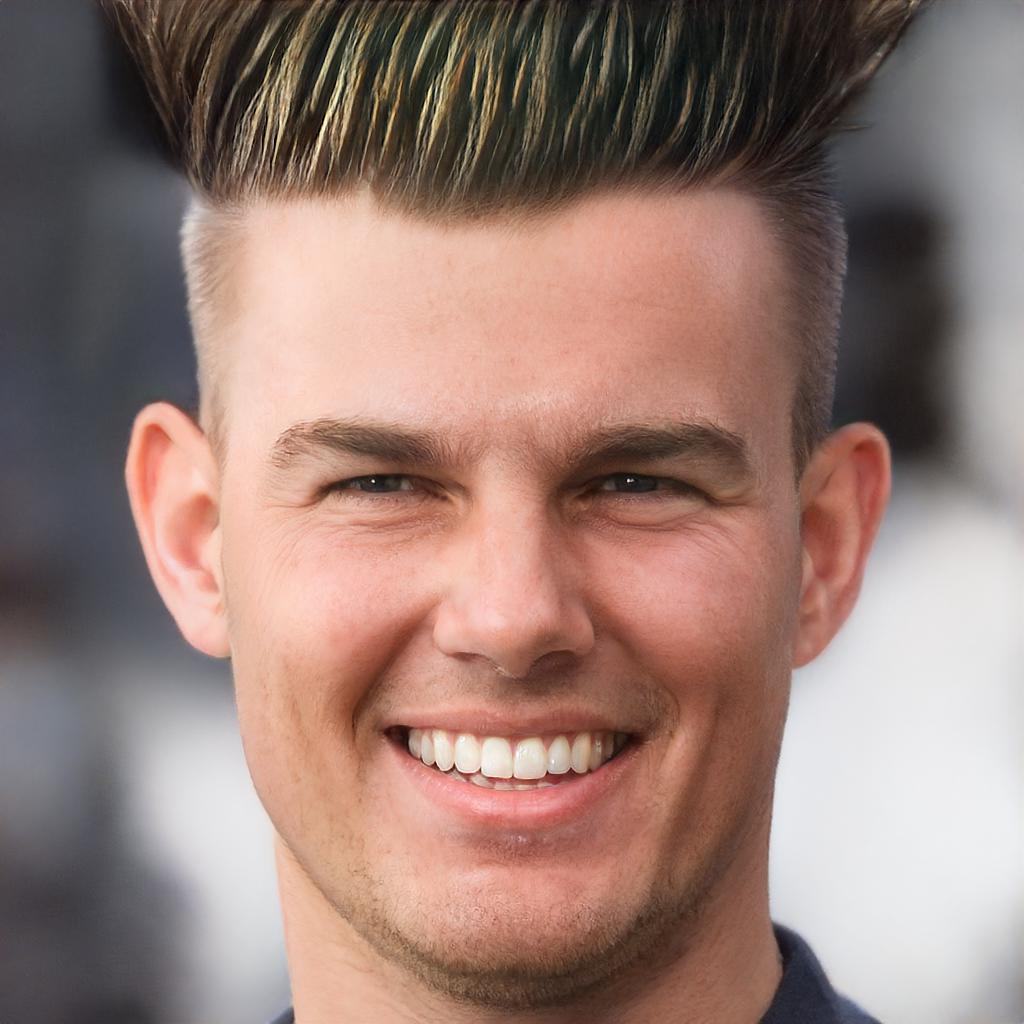} &
        \includegraphics[width=\linewidth]{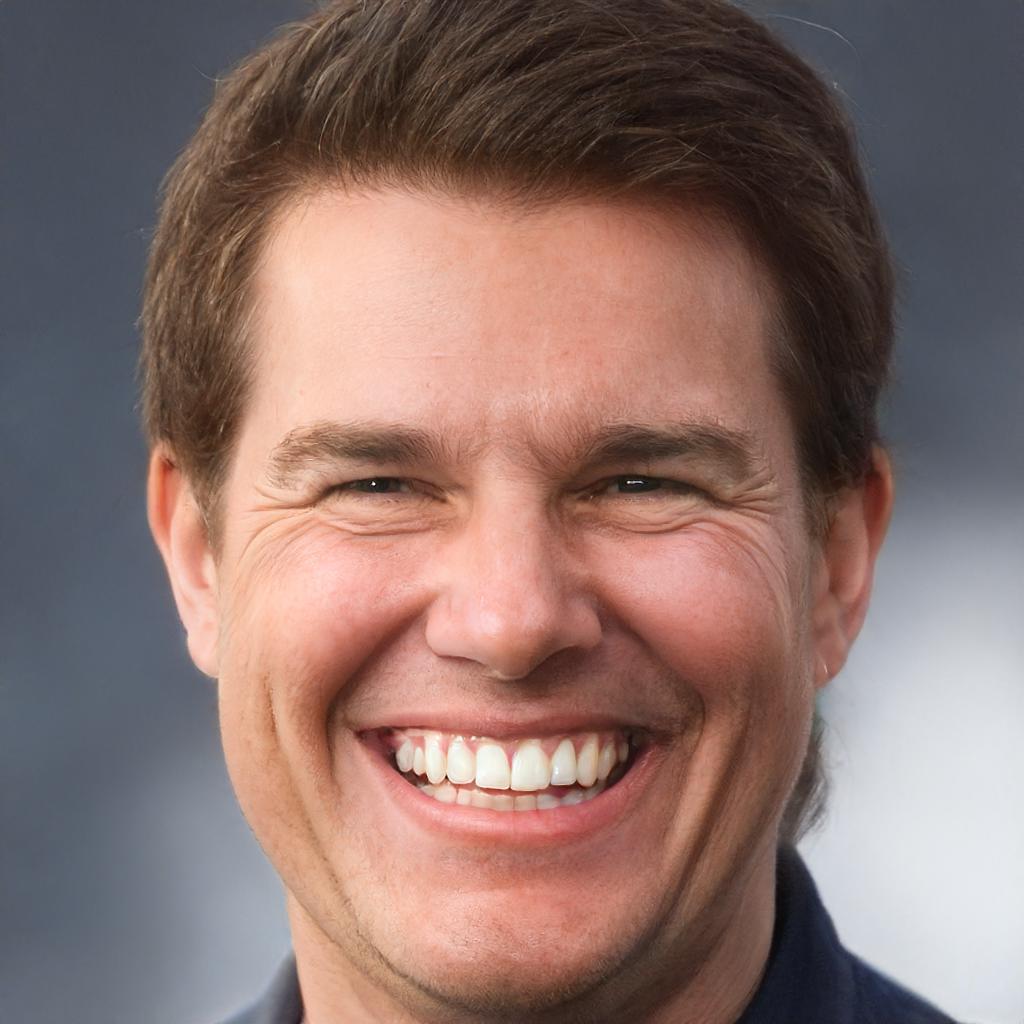} &
        \includegraphics[width=\linewidth]{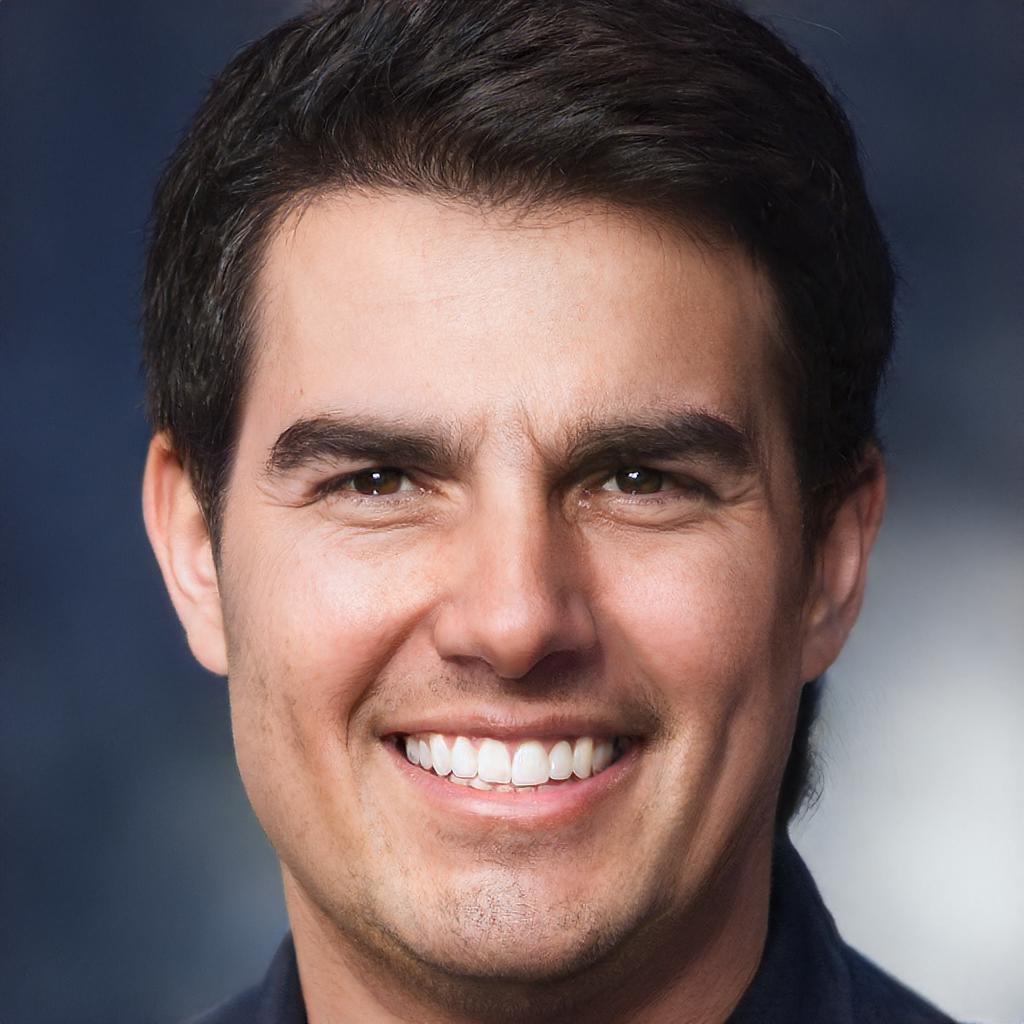} &
        \includegraphics[width=\linewidth]{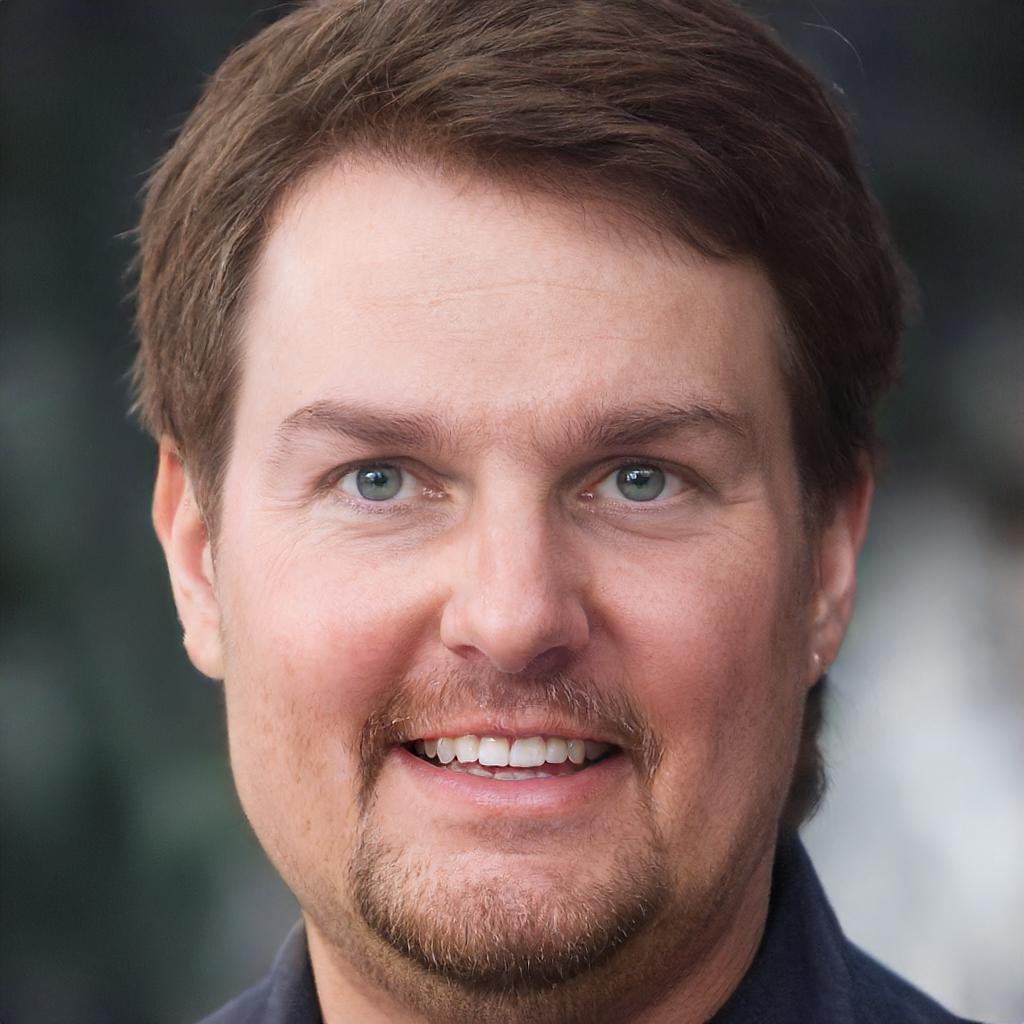} 
        \tabularnewline
        & \raisebox{0.22in}{\rotatebox[origin=t]{90}{\footnotesize StyleFusion}} &
        \includegraphics[width=\linewidth]{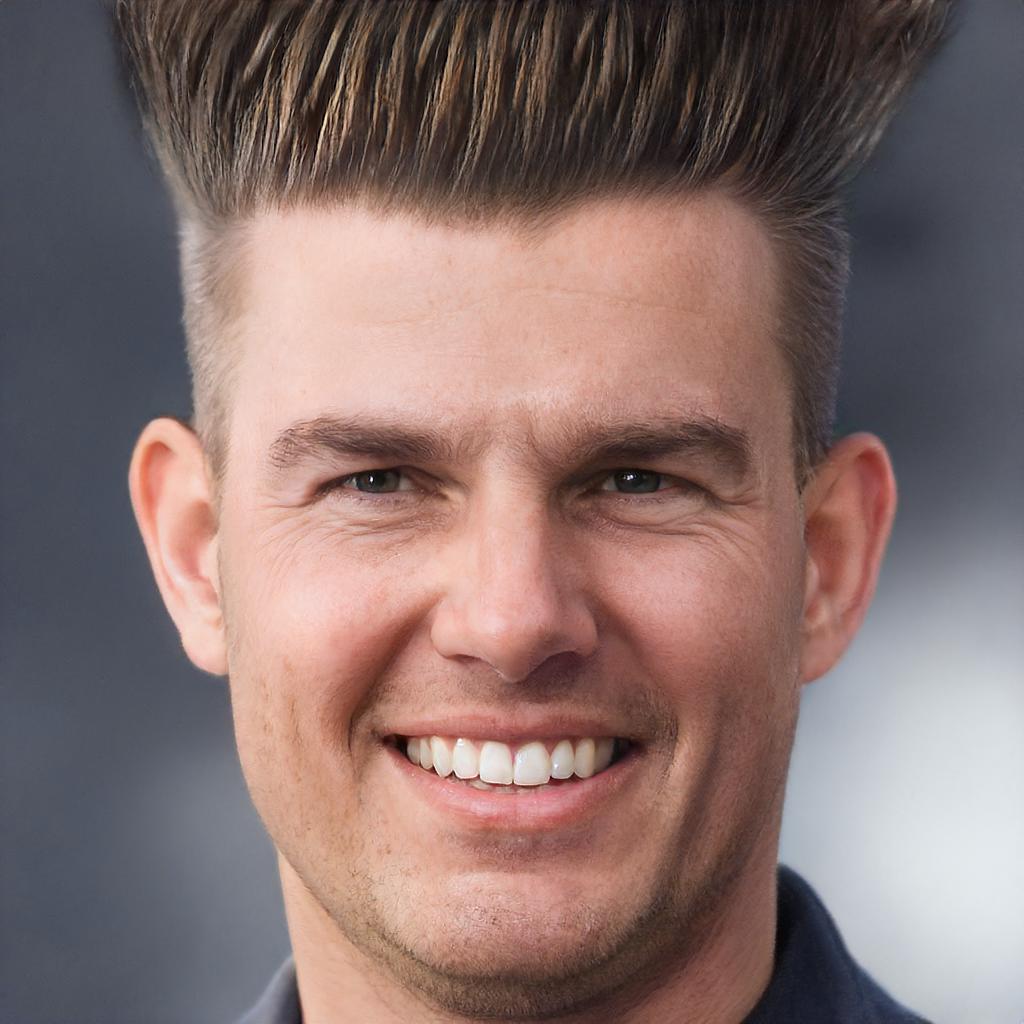} &
        \includegraphics[width=\linewidth]{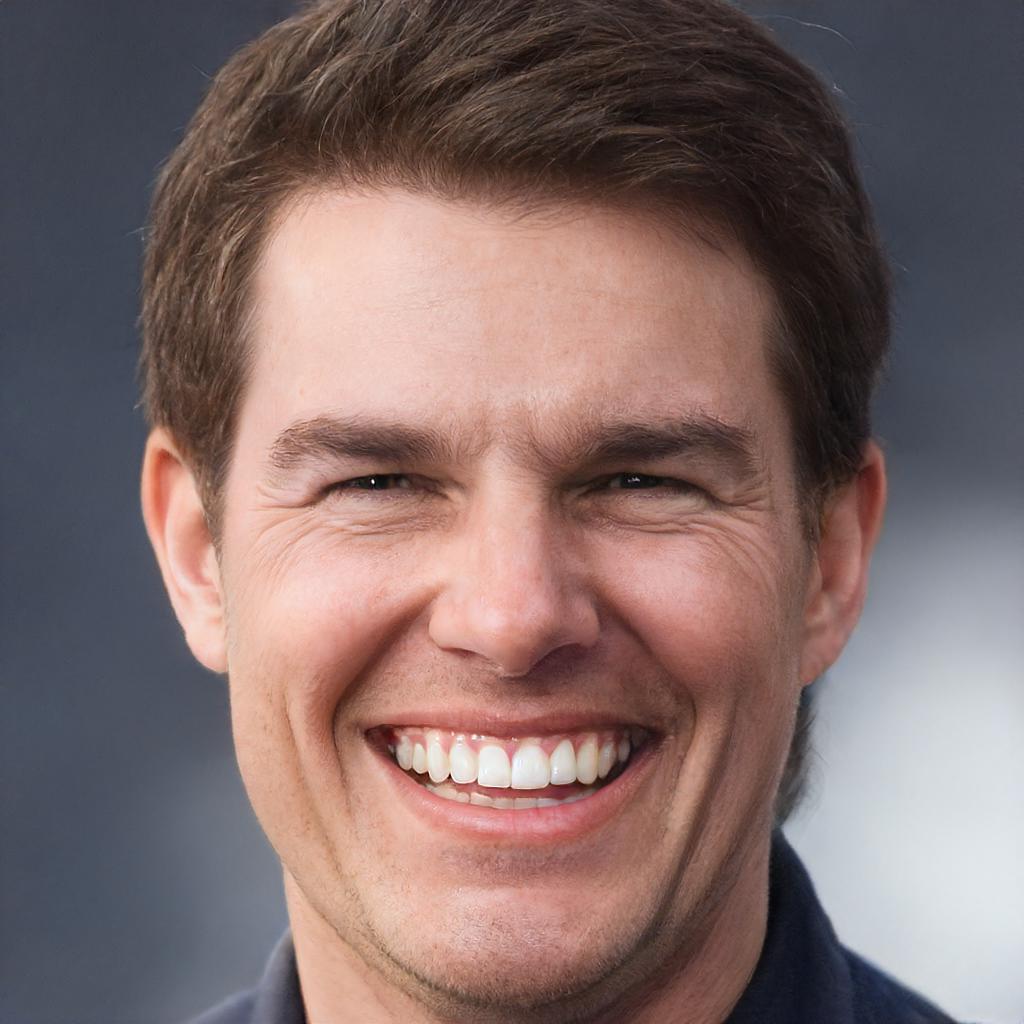} &
        \includegraphics[width=\linewidth]{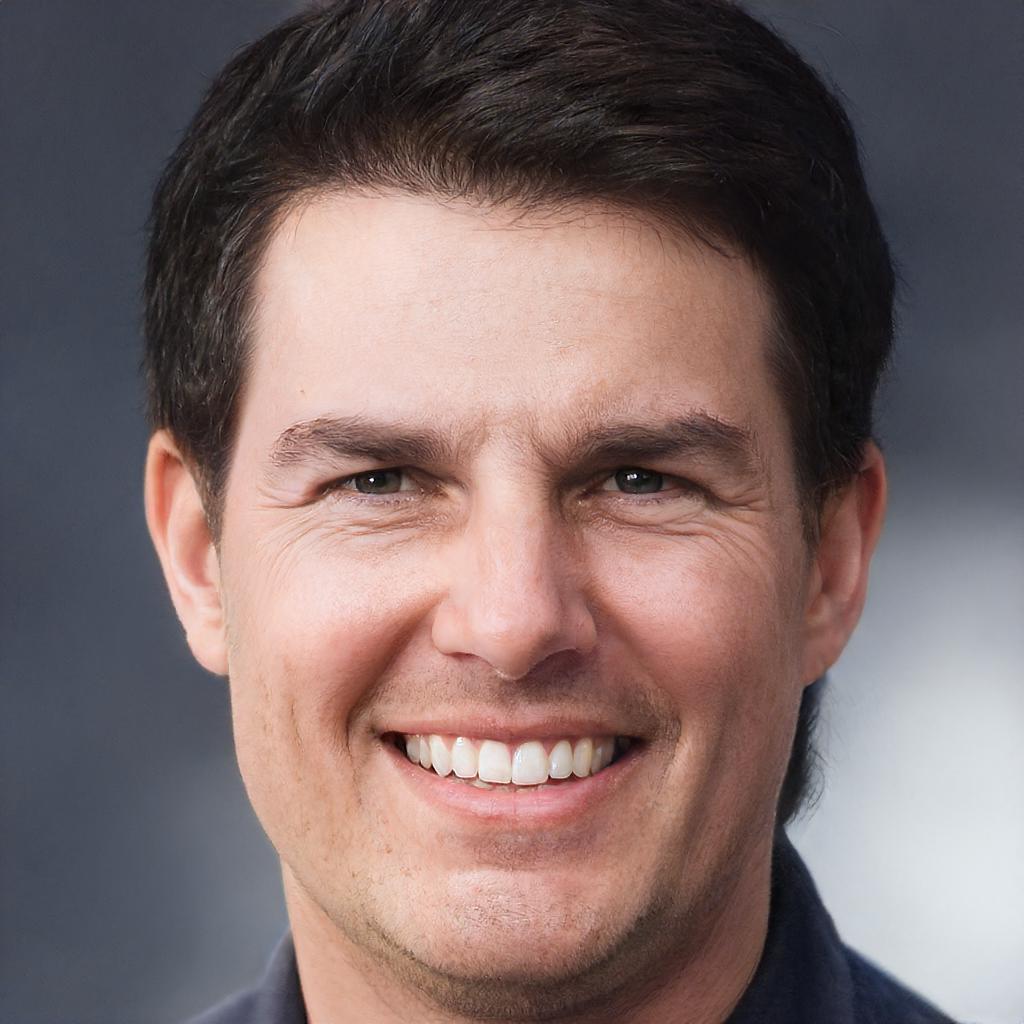} &
        \includegraphics[width=\linewidth]{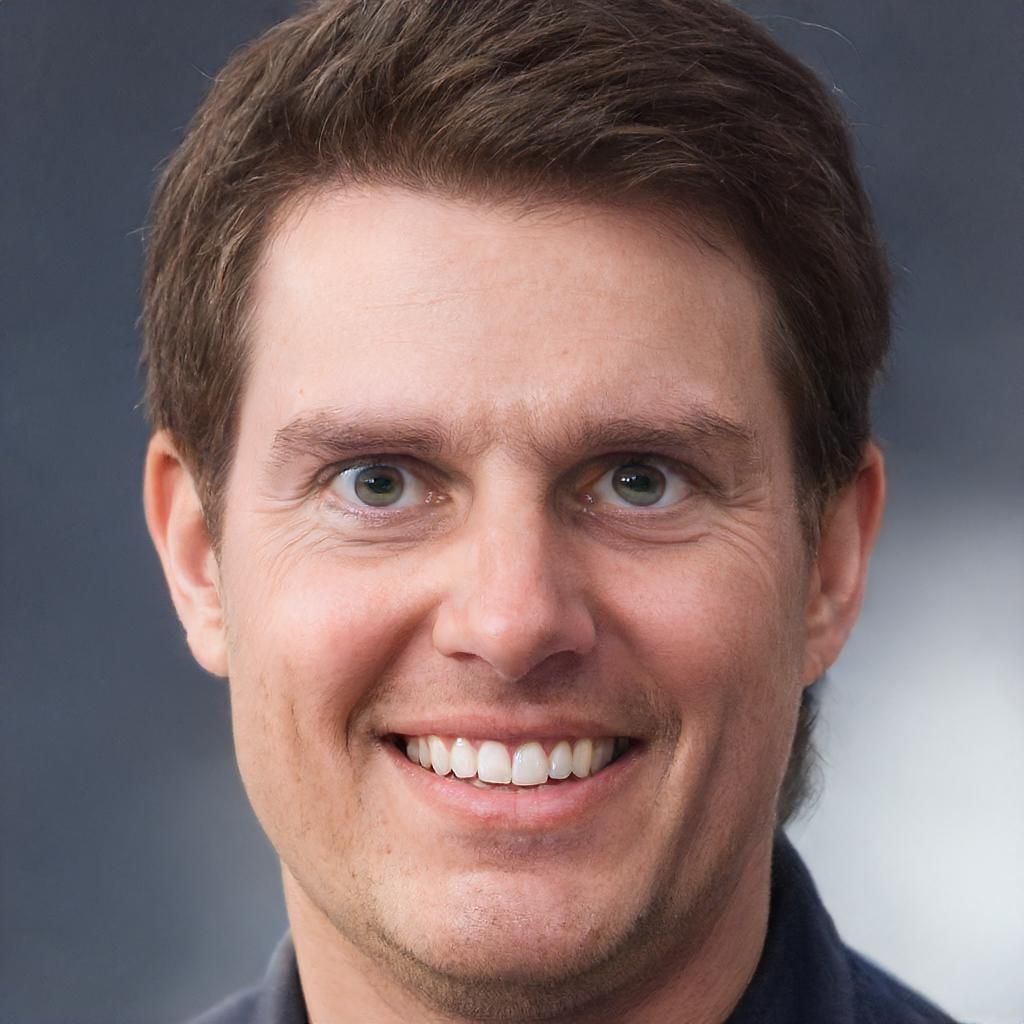} 
        \tabularnewline
        
        \includegraphics[width=\linewidth]{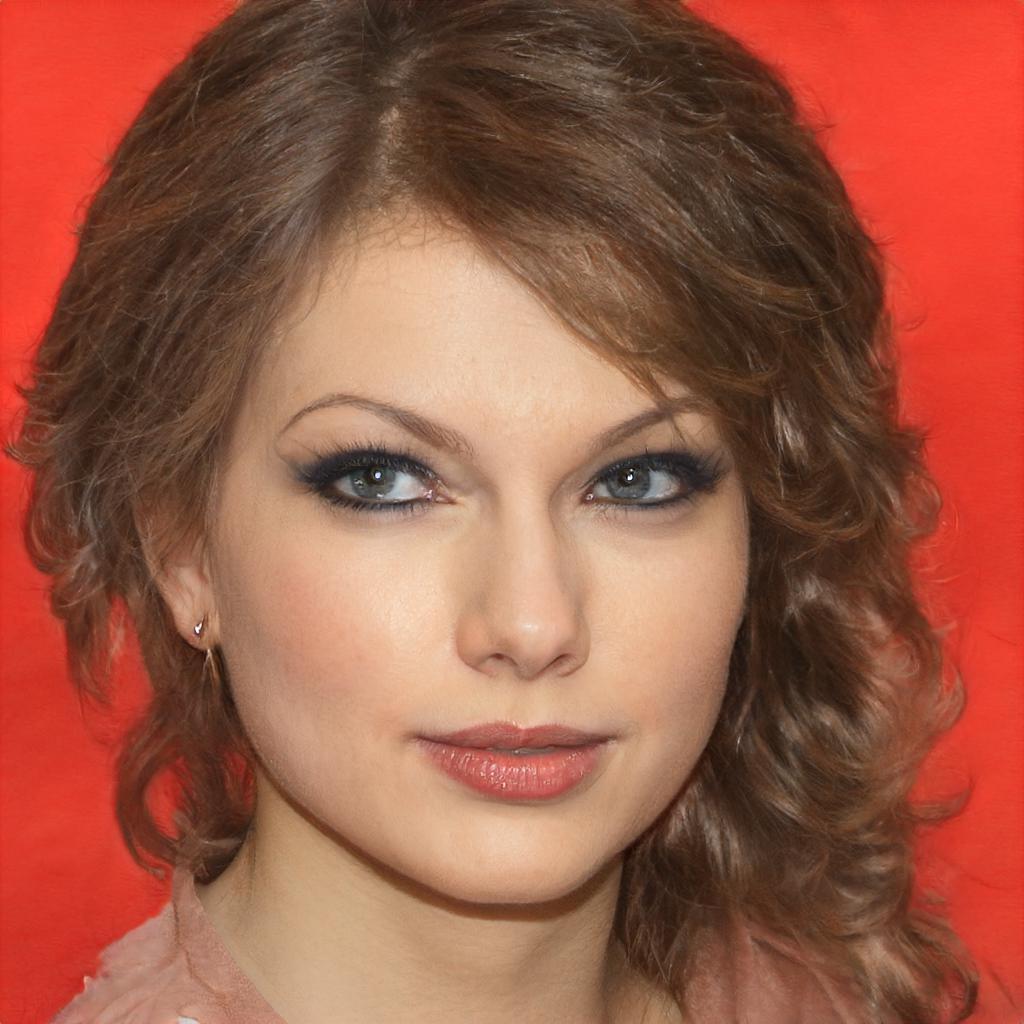} &
        \raisebox{0.22in}{\rotatebox[origin=t]{90}{\footnotesize StyleGAN2}} &
        \includegraphics[width=\linewidth]{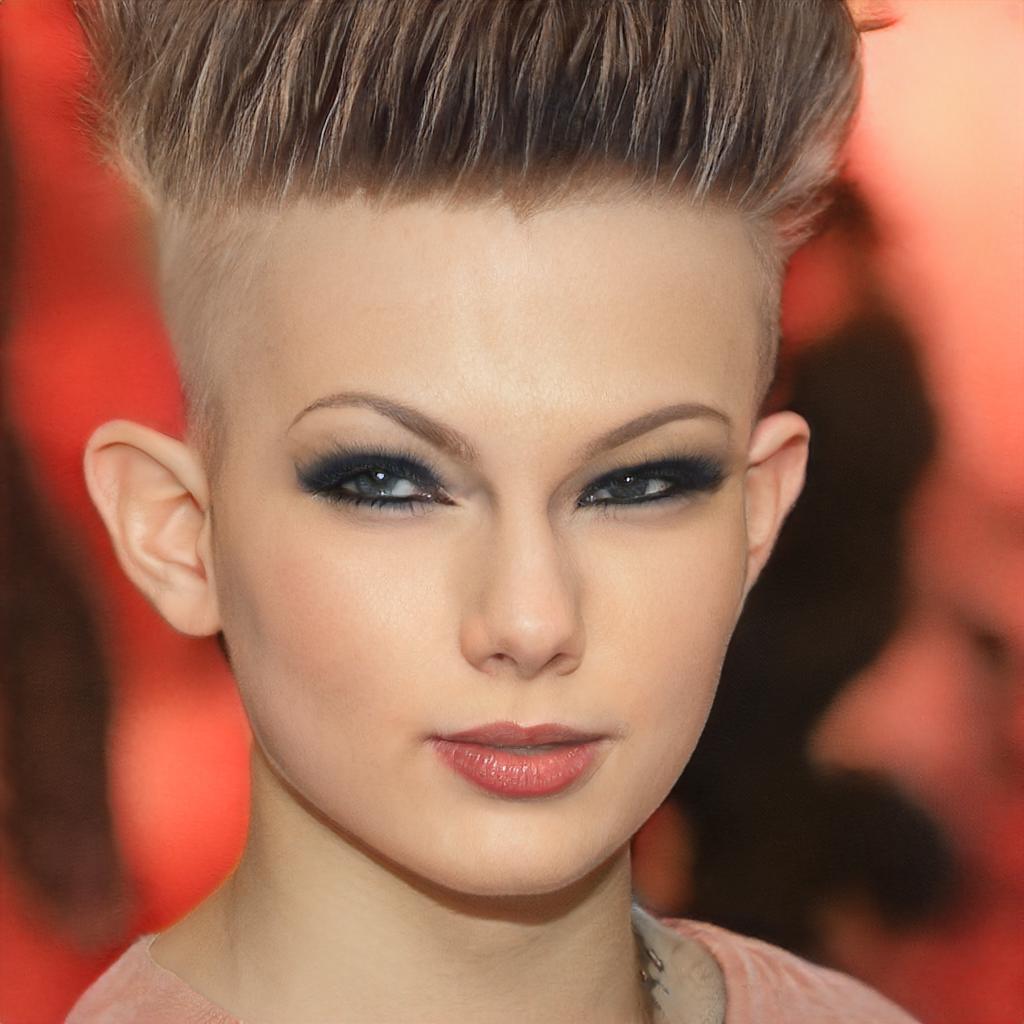} &
        \includegraphics[width=\linewidth]{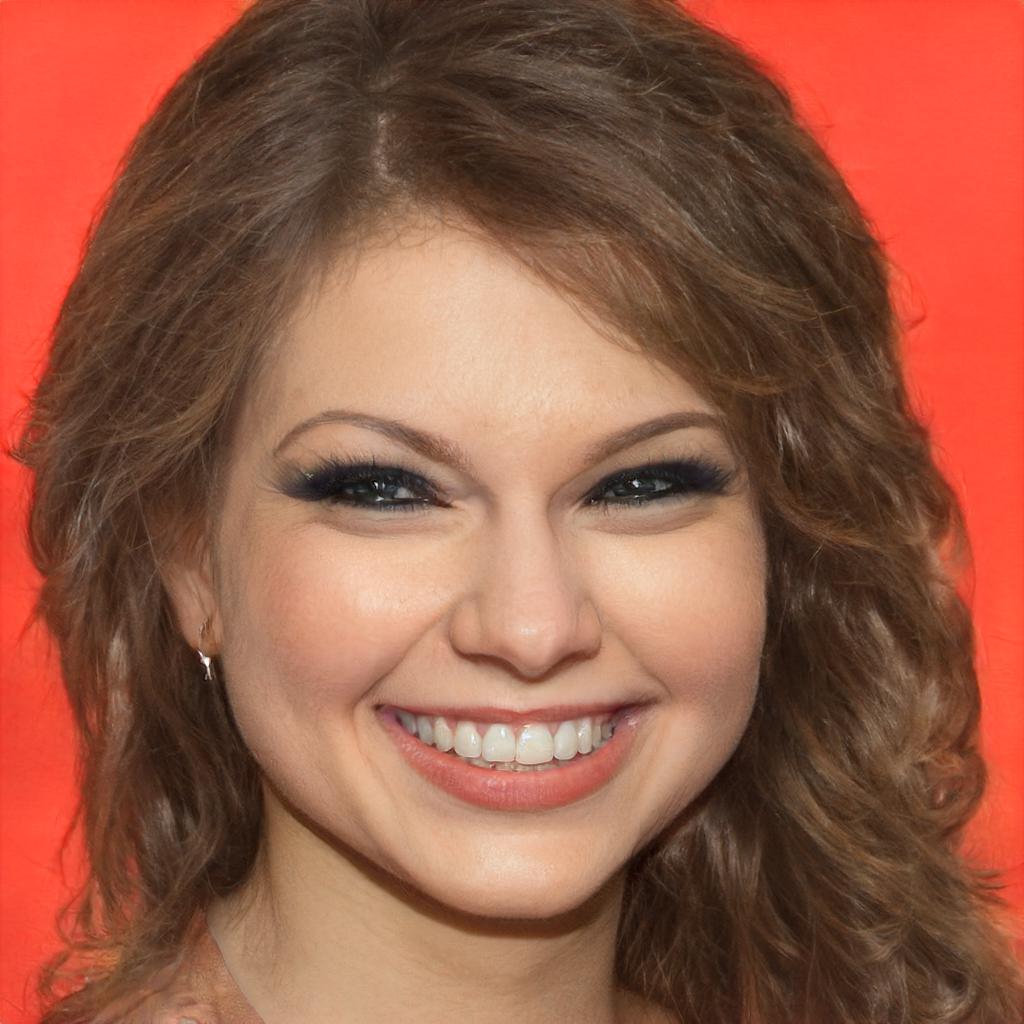} &
        \includegraphics[width=\linewidth]{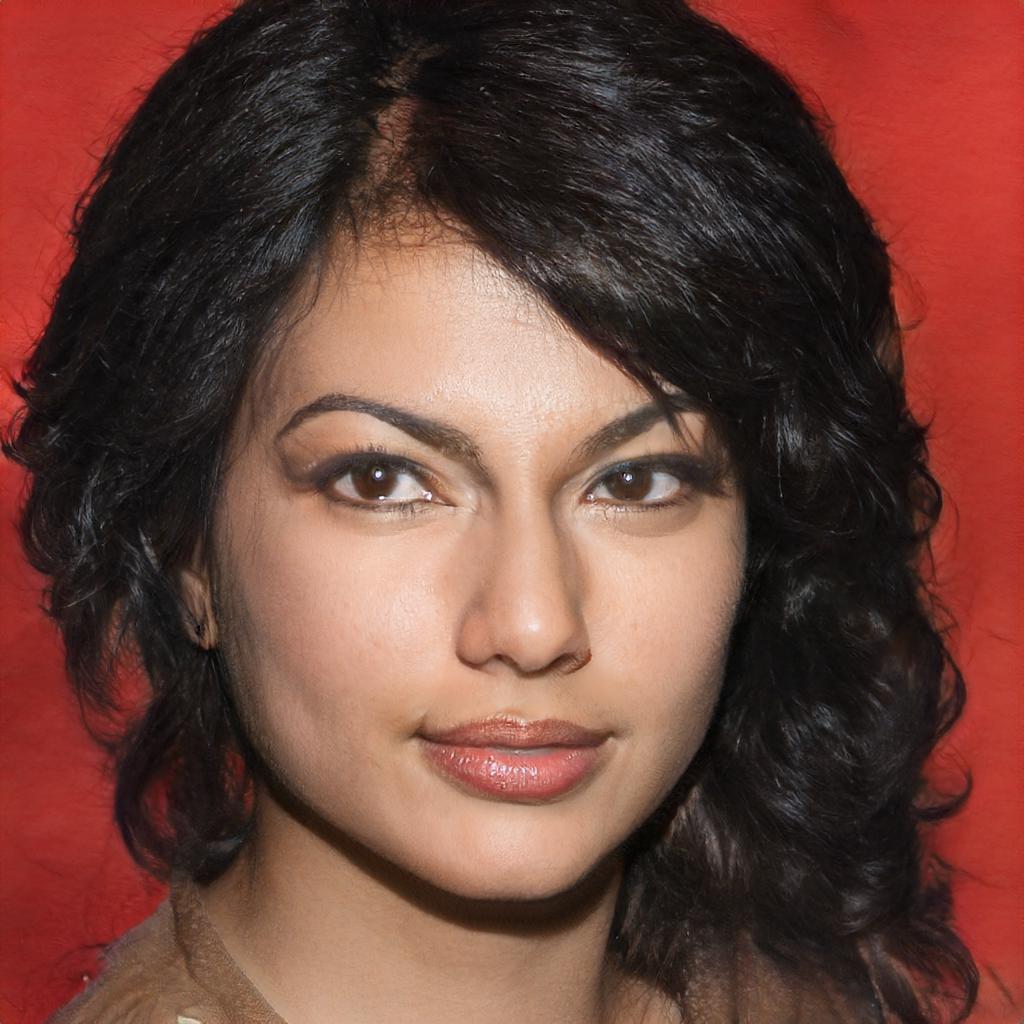} &
        \includegraphics[width=\linewidth]{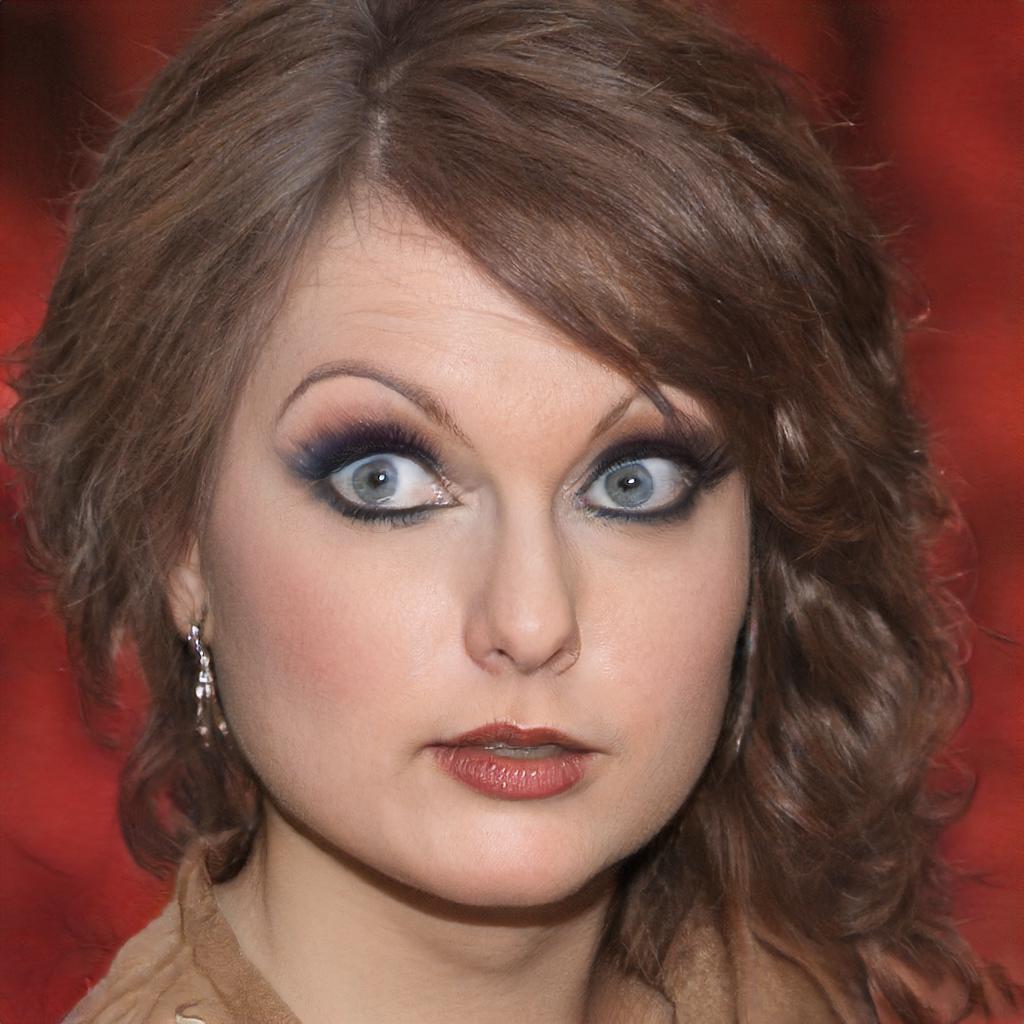} 
        \tabularnewline
        & \raisebox{0.22in}{\rotatebox[origin=t]{90}{\footnotesize StyleFusion}} &
        \includegraphics[width=\linewidth]{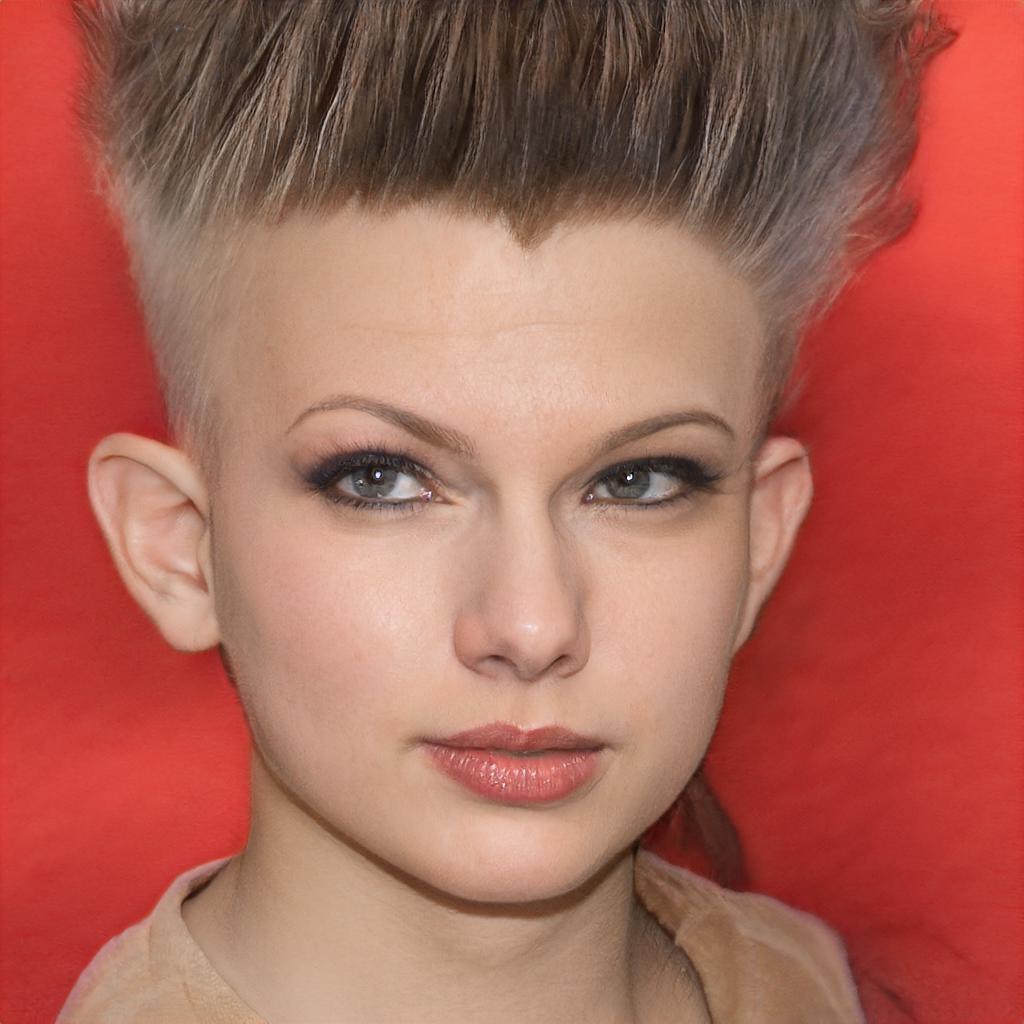} &
        \includegraphics[width=\linewidth]{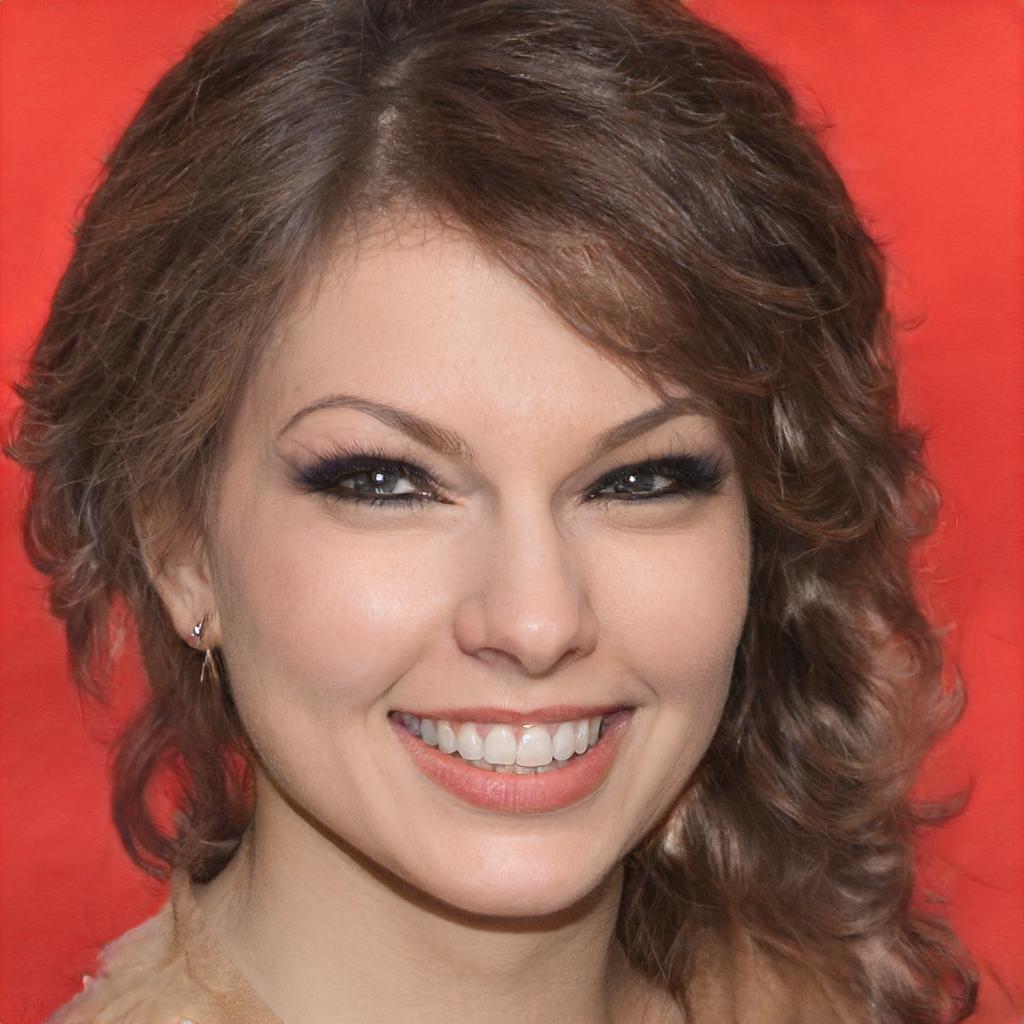} &
        \includegraphics[width=\linewidth]{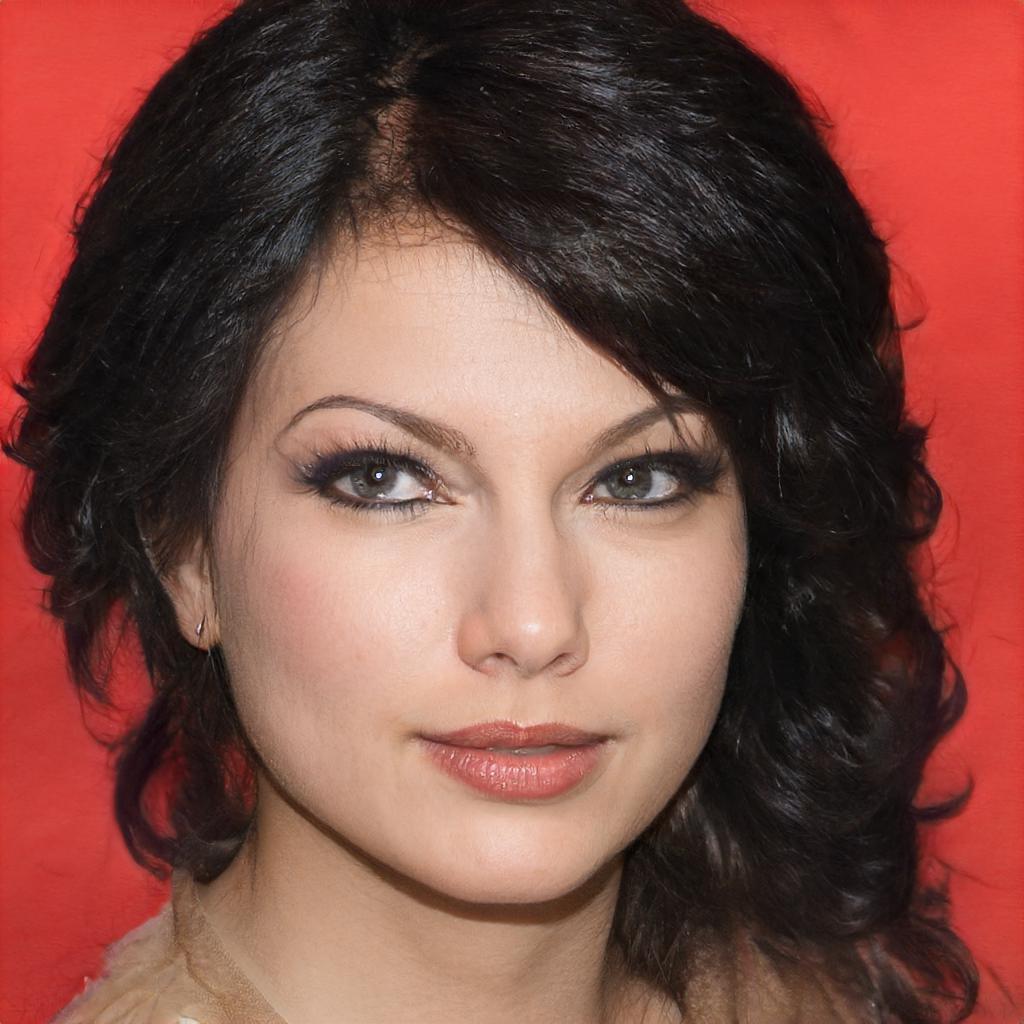} &
        \includegraphics[width=\linewidth]{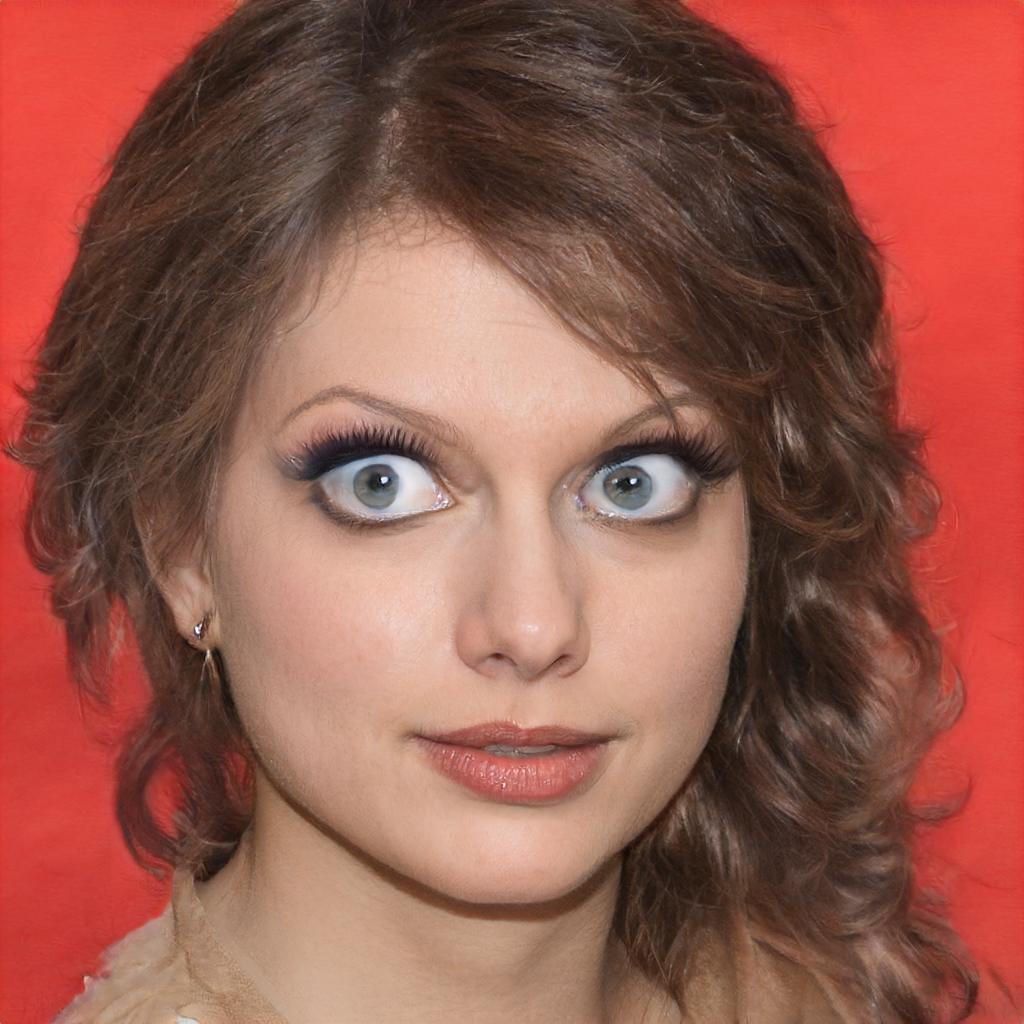} 
        \tabularnewline
        
        Input & & Mohawk & Smile & Hair Albedo & Fearful Eyes
        \tabularnewline
        \end{tabular}
        }
    \caption{\textit{Latent traversal editing techniques performed with and without StyleFusion.} 
    With StyleFusion, the techniques are applied only to the latent code that controls the relevant region(s).
    As can be seen, pairing a latent editing technique with StyleFusion leads to a more disentangled, precise edit.}
    \label{fig:latent_editing_comparison}
\end{figure}

\begin{figure}
    \centering
    \setlength{\tabcolsep}{0pt}
    \begin{tabular}{c c c c} &
        \includegraphics[width=0.25\linewidth]
        {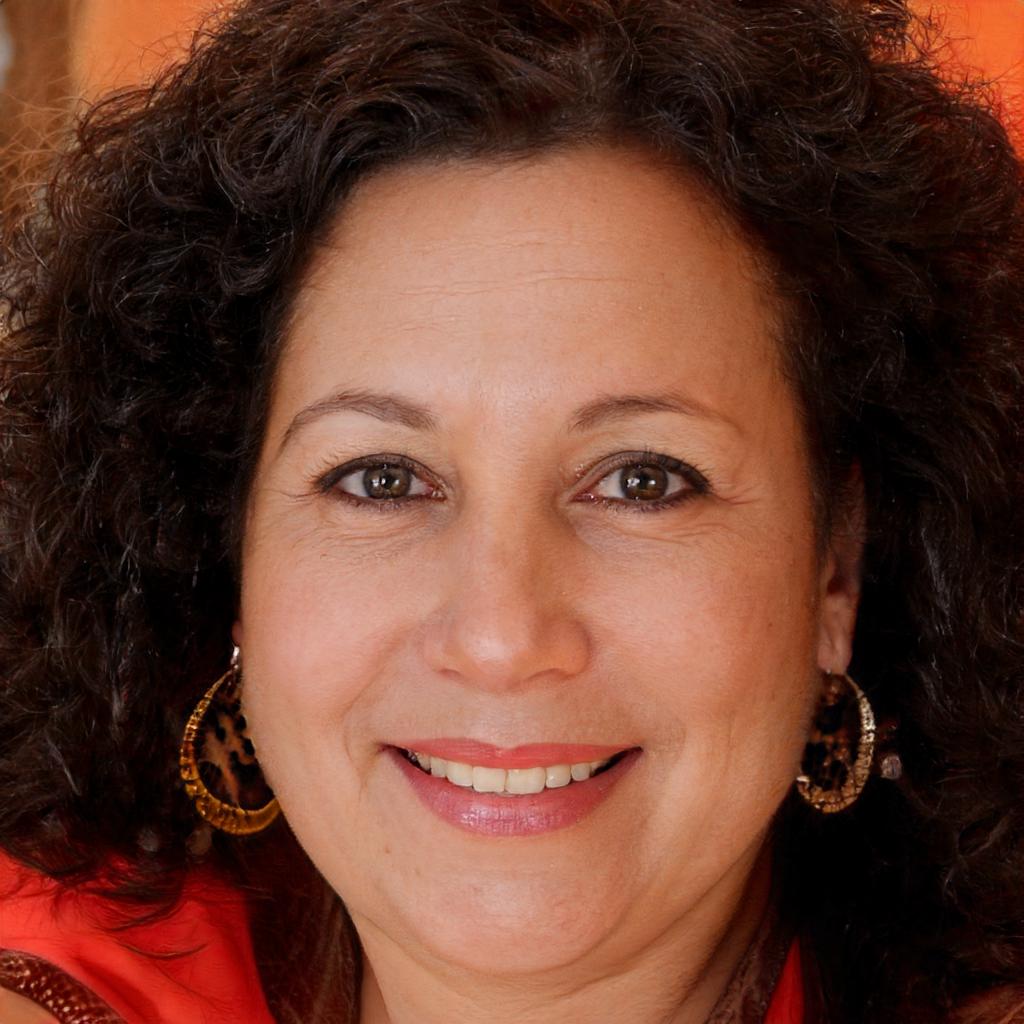} &
        \includegraphics[width=0.25\linewidth]
        {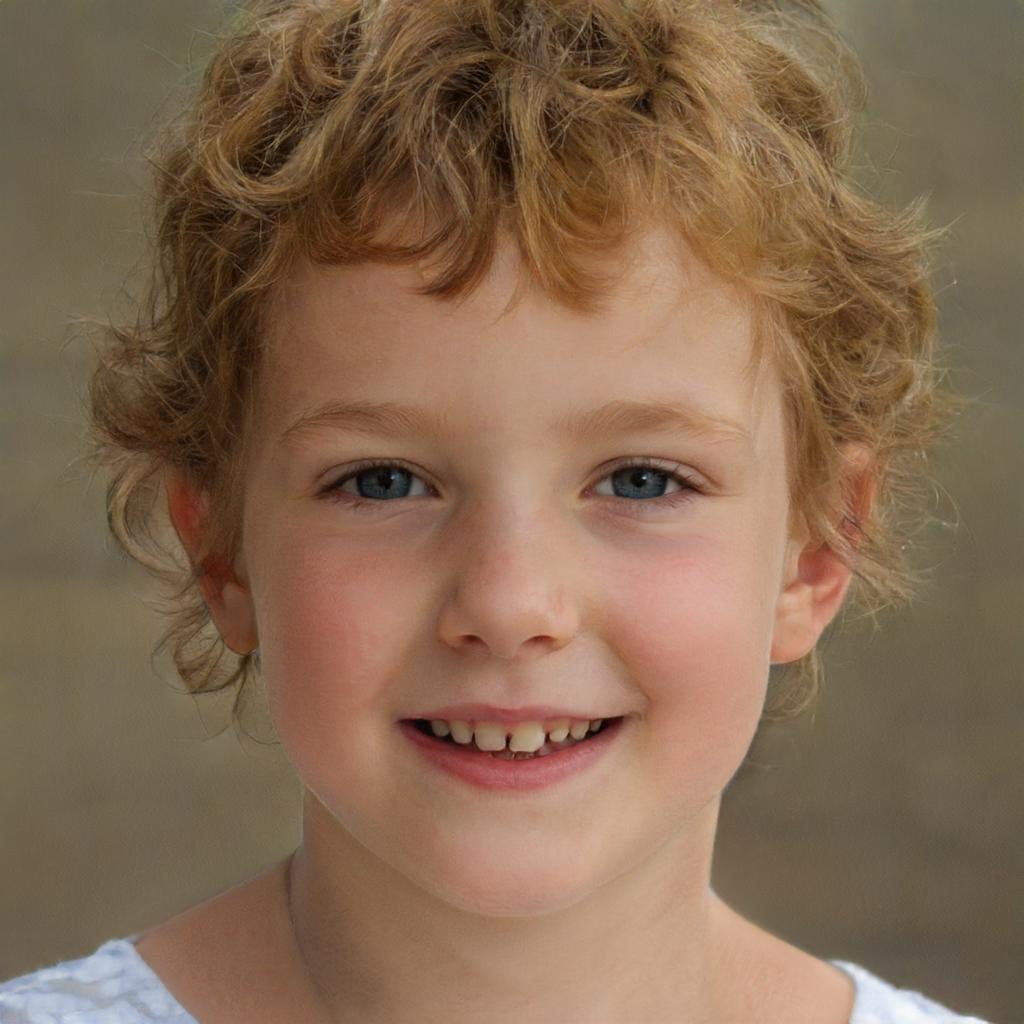} &
        \includegraphics[width=0.25\linewidth]
        {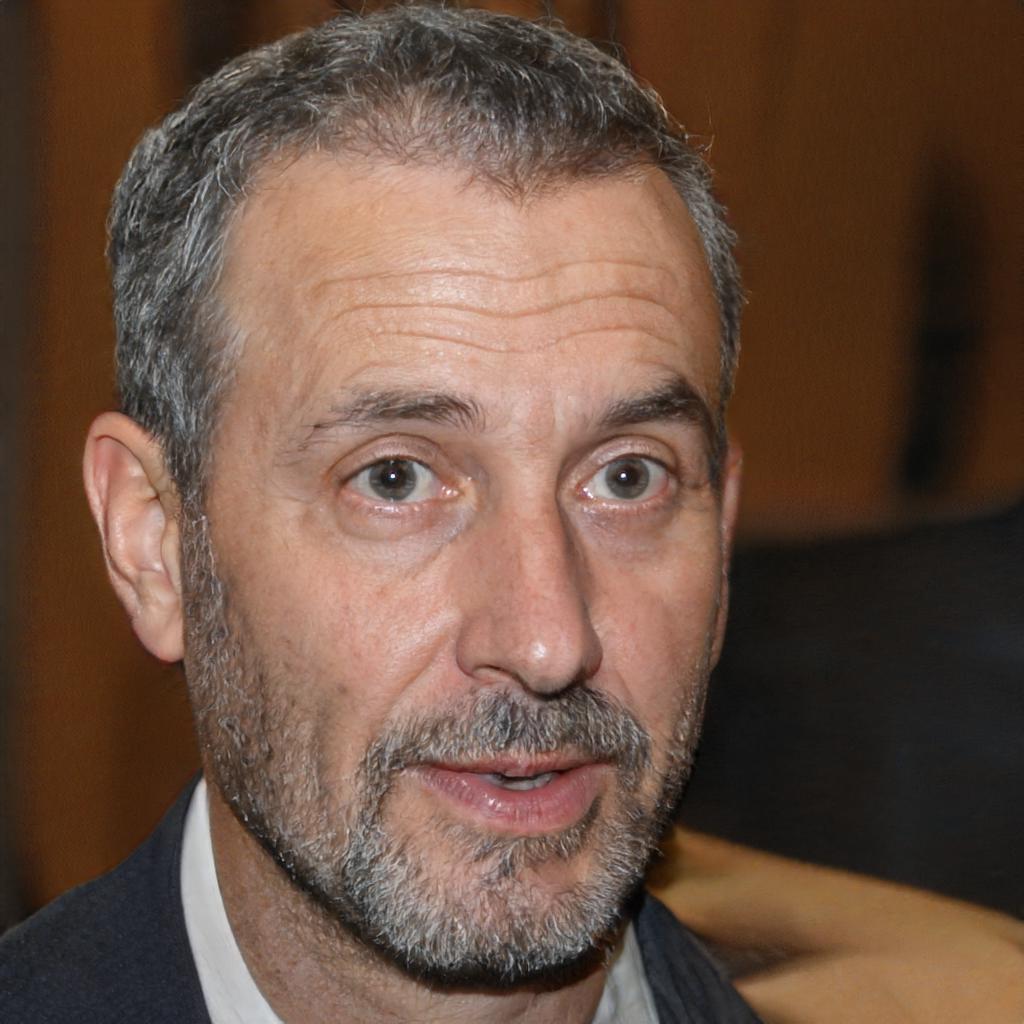}
        \tabularnewline
        \includegraphics[width=0.25\linewidth]
        {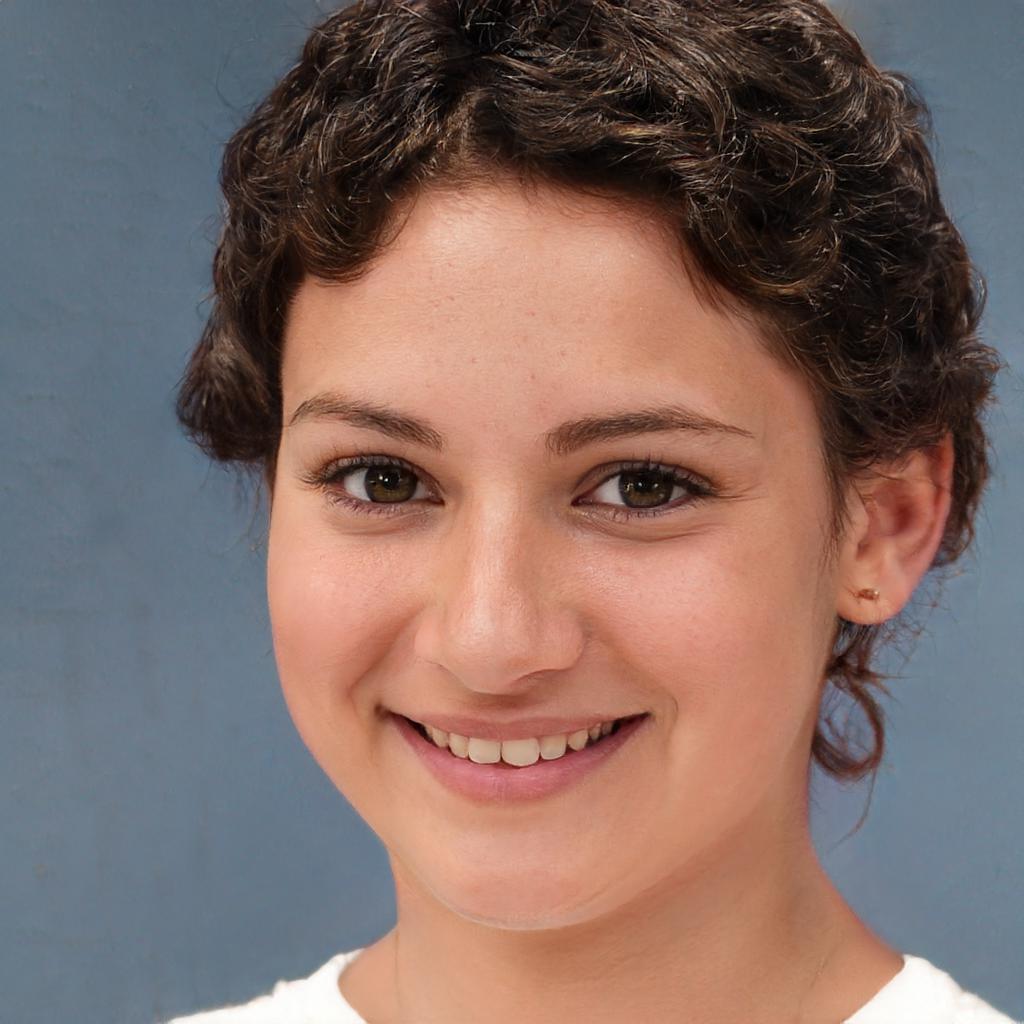} &
        \includegraphics[width=0.25\linewidth]
        {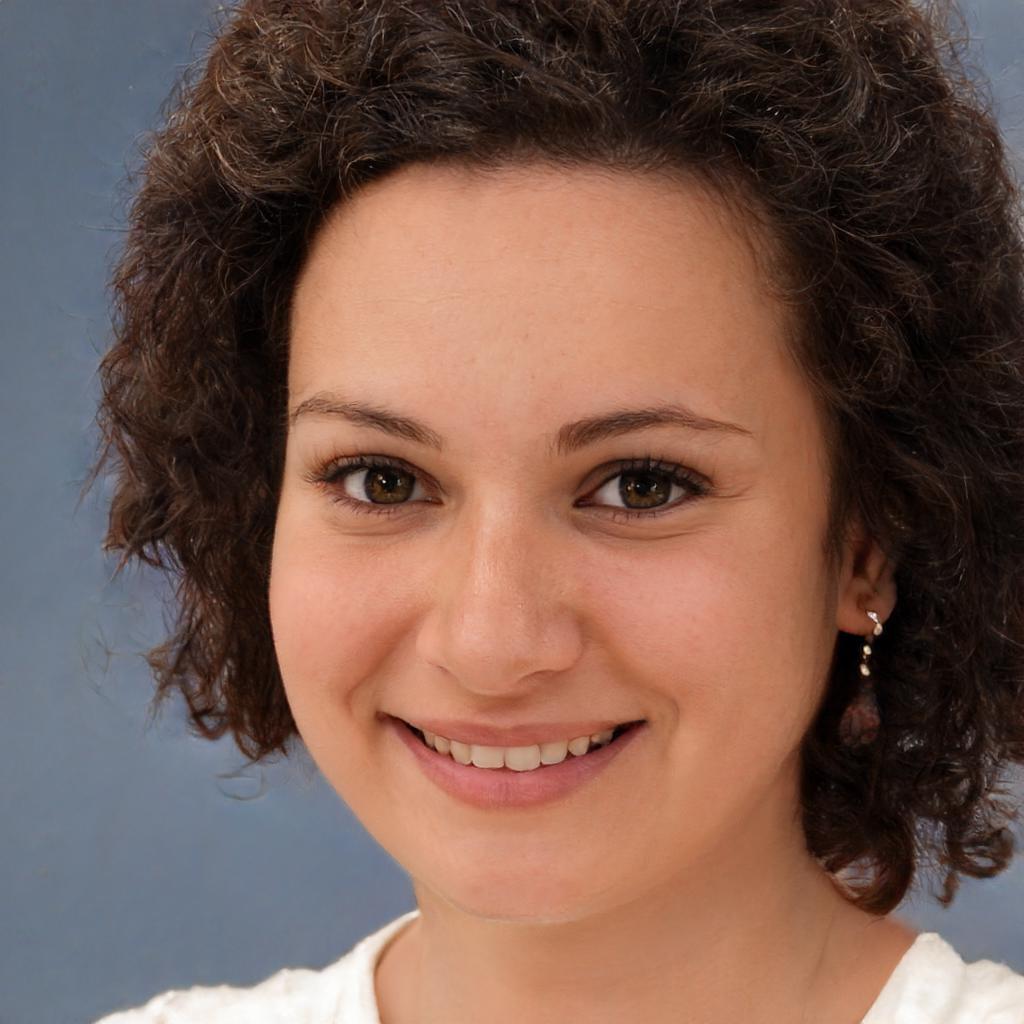} &
        \includegraphics[width=0.25\linewidth]
        {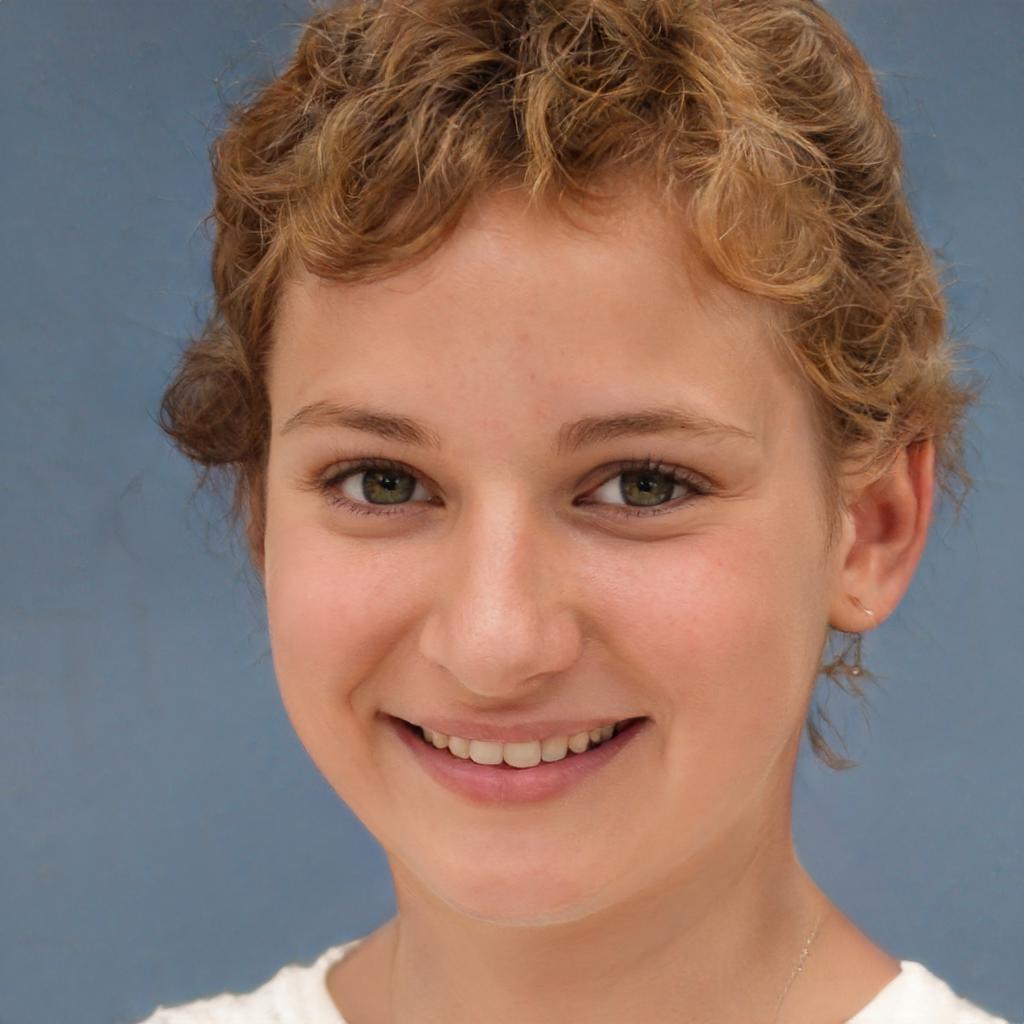} &
        \includegraphics[width=0.25\linewidth]
        {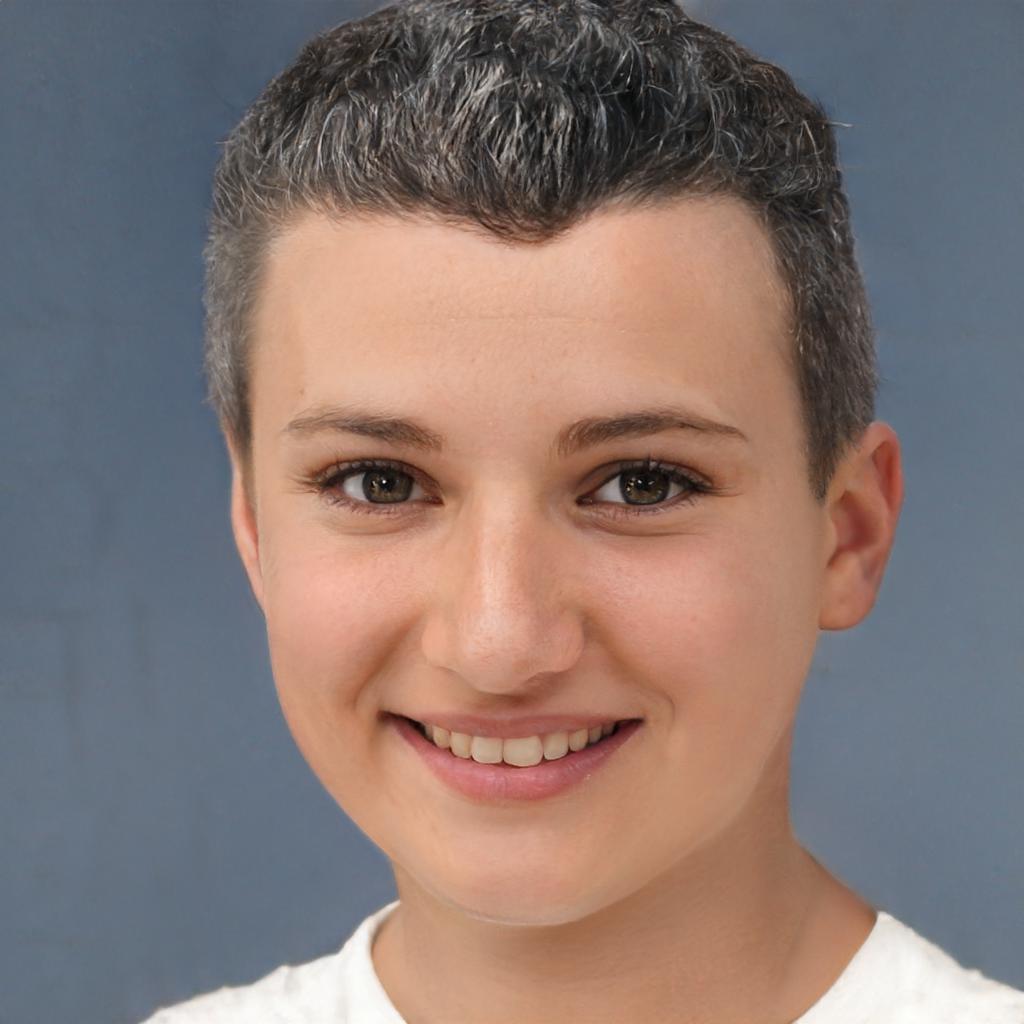}
        \tabularnewline
        \includegraphics[width=0.25\linewidth]
        {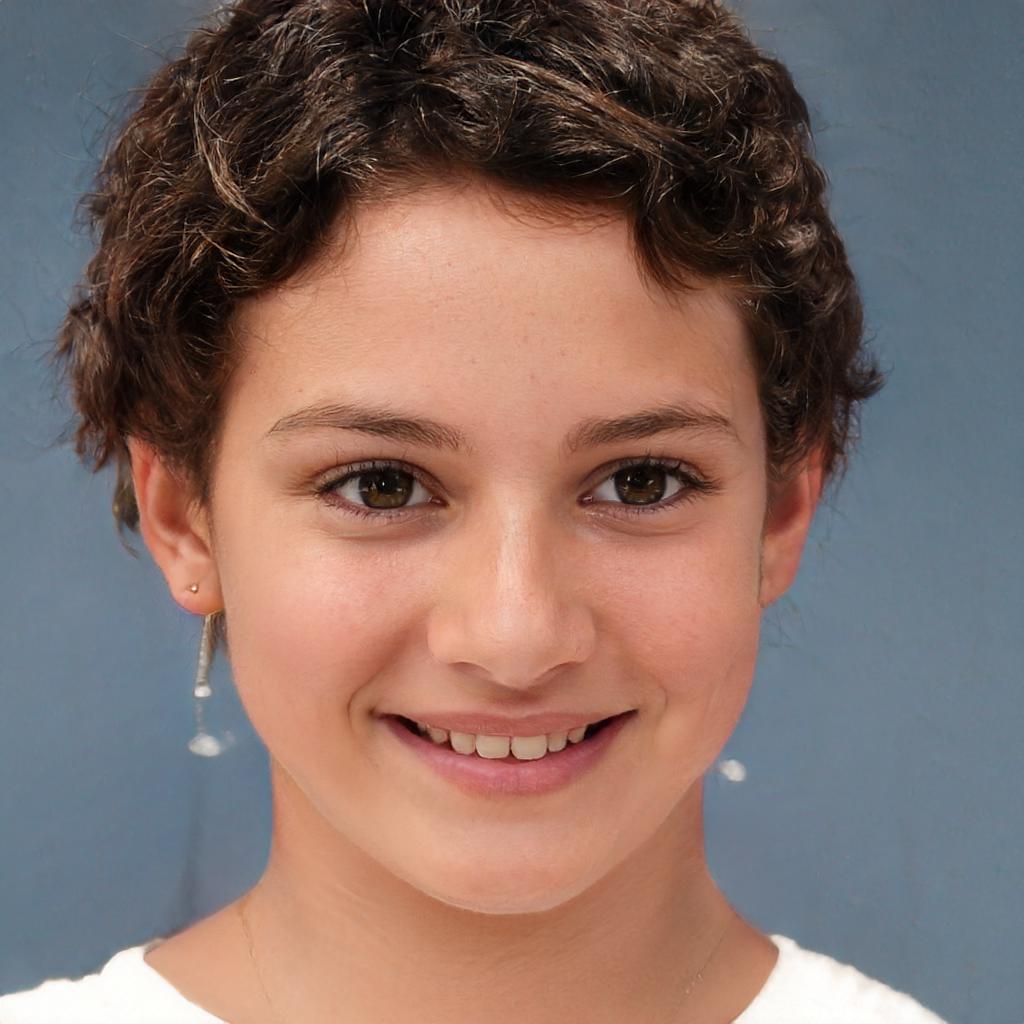} &
        \includegraphics[width=0.25\linewidth]
        {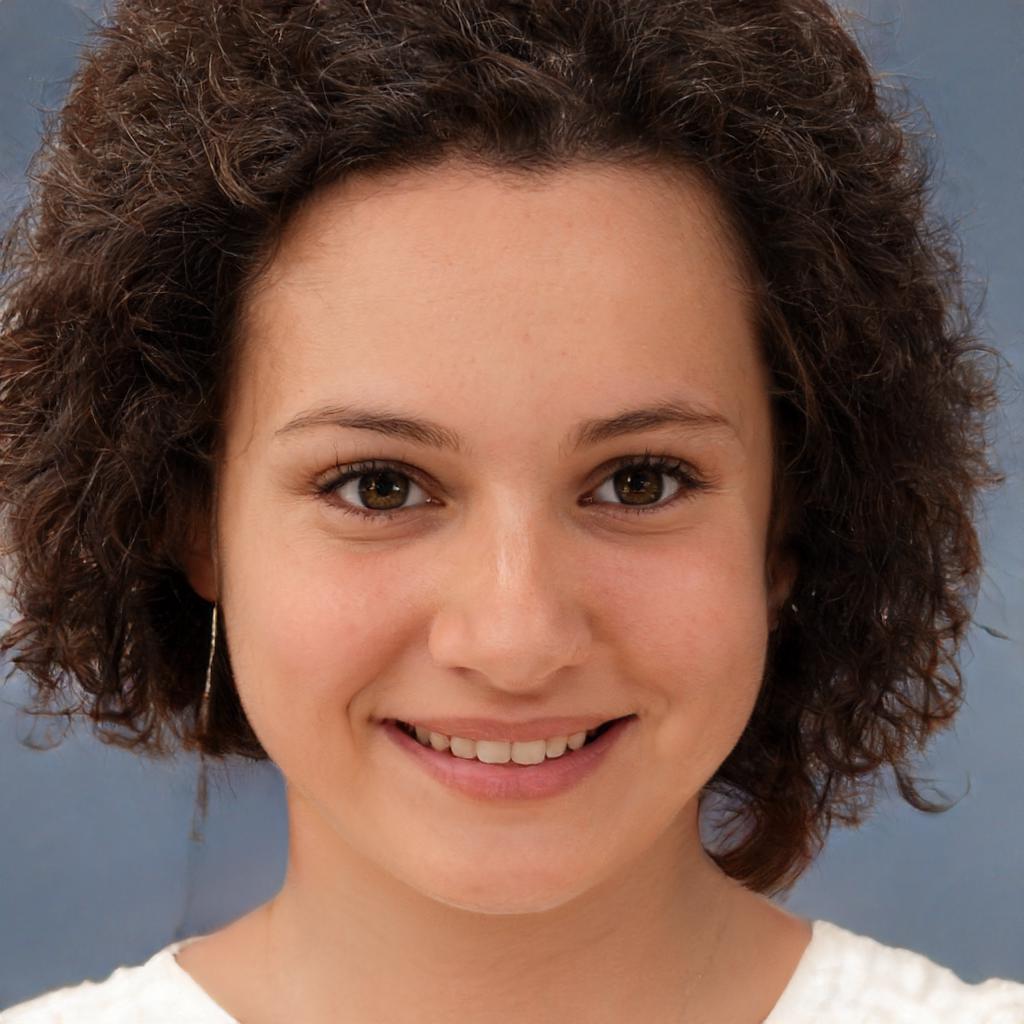} &
        \includegraphics[width=0.25\linewidth]
        {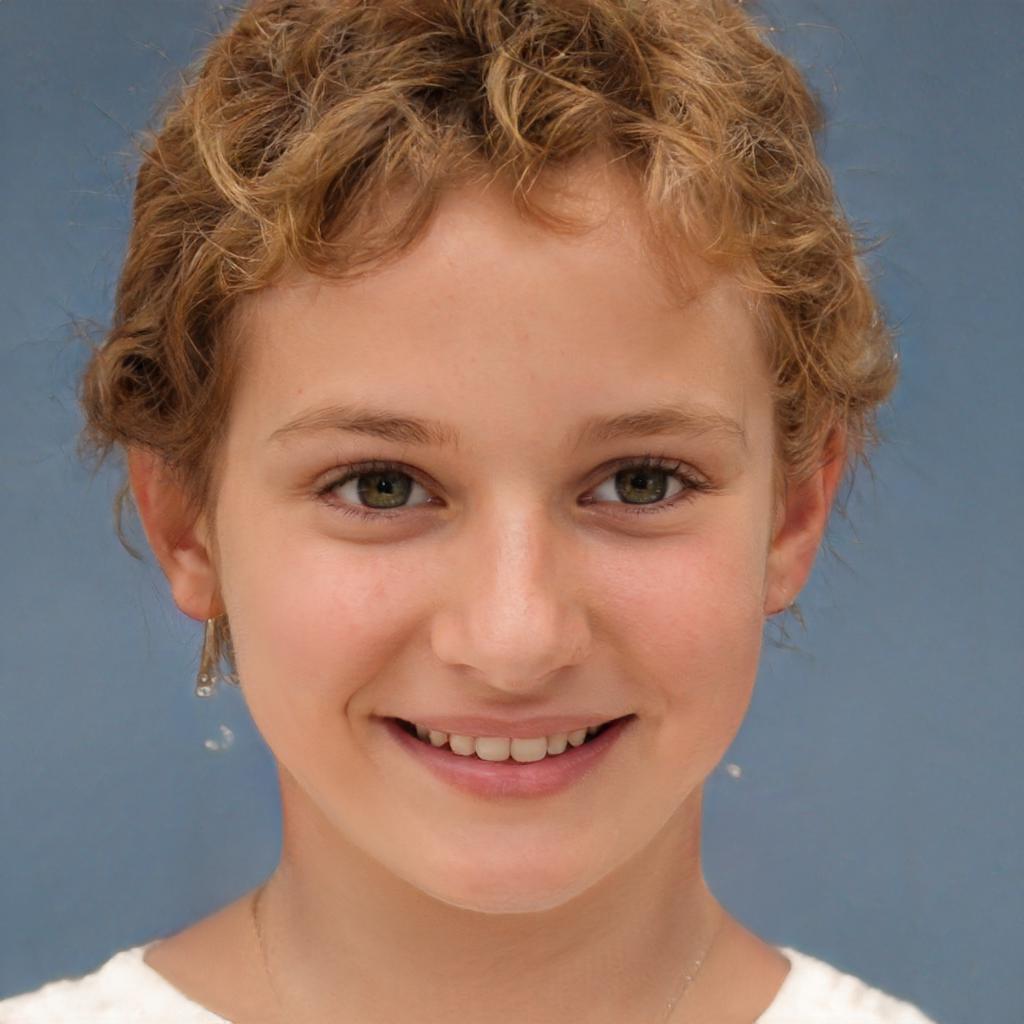} &
        \includegraphics[width=0.25\linewidth]
        {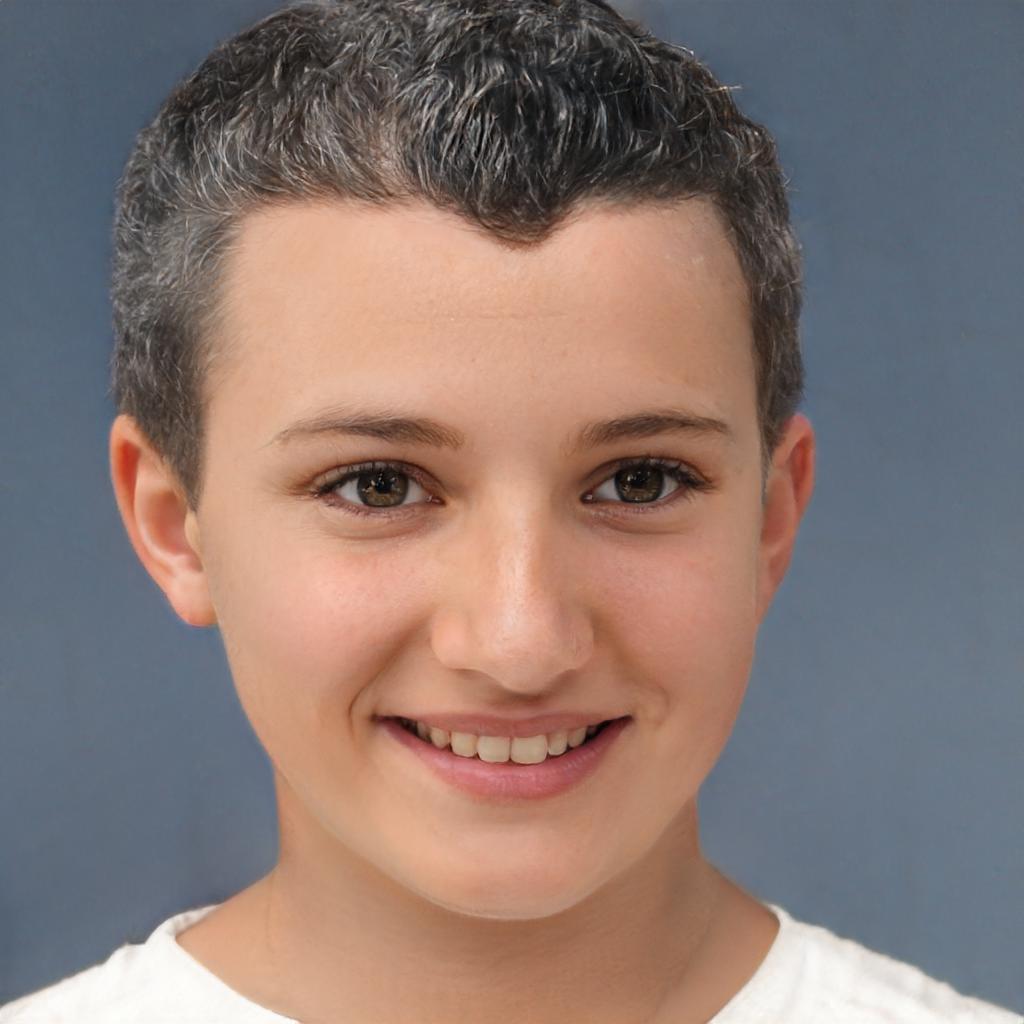}
        \tabularnewline
        \includegraphics[width=0.25\linewidth]
        {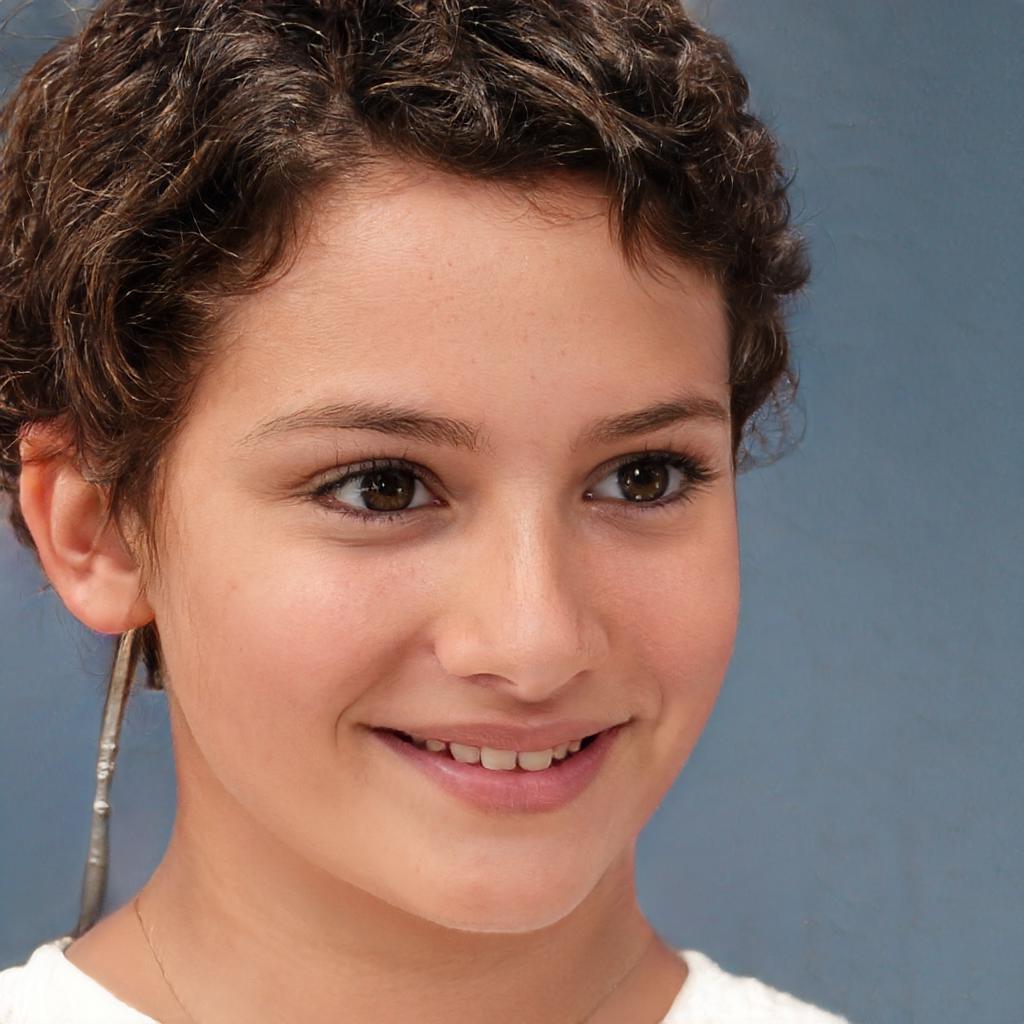} &
        \includegraphics[width=0.25\linewidth]
        {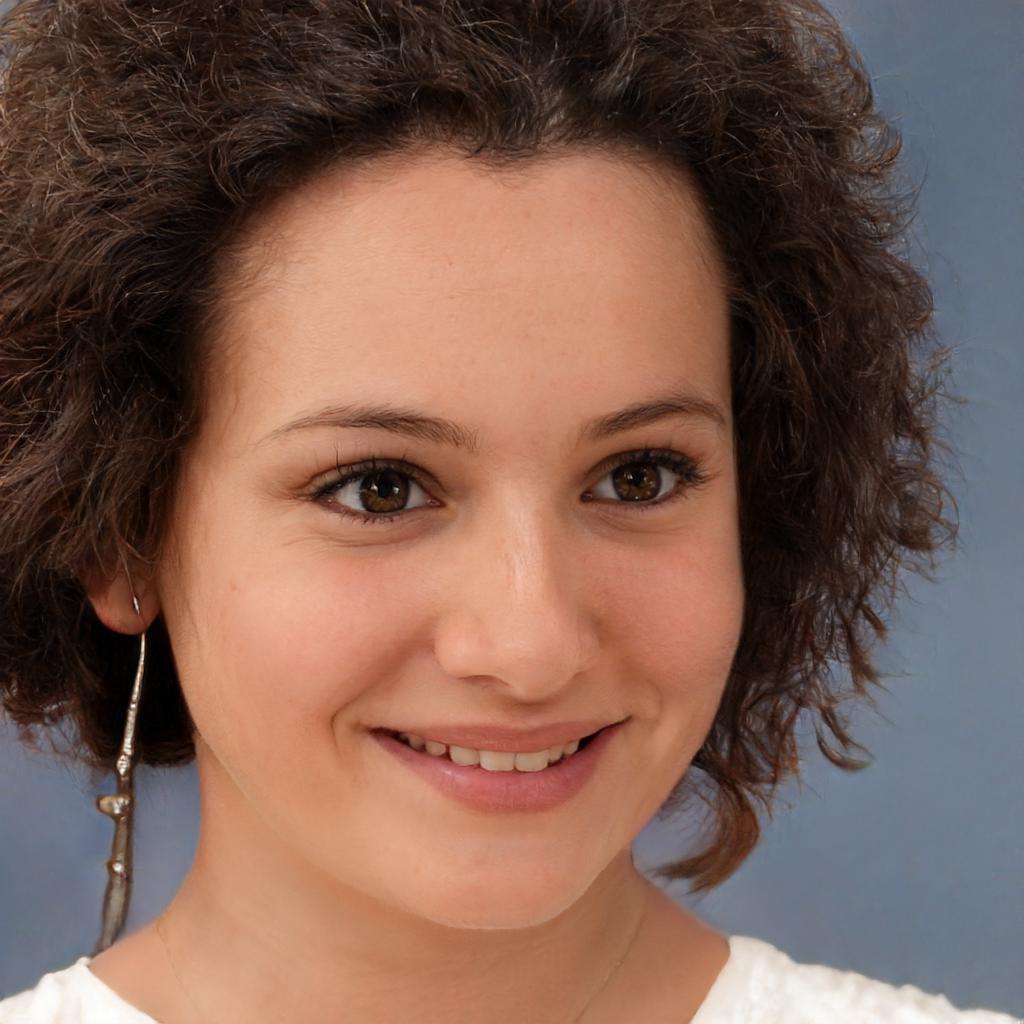} &
        \includegraphics[width=0.25\linewidth]
        {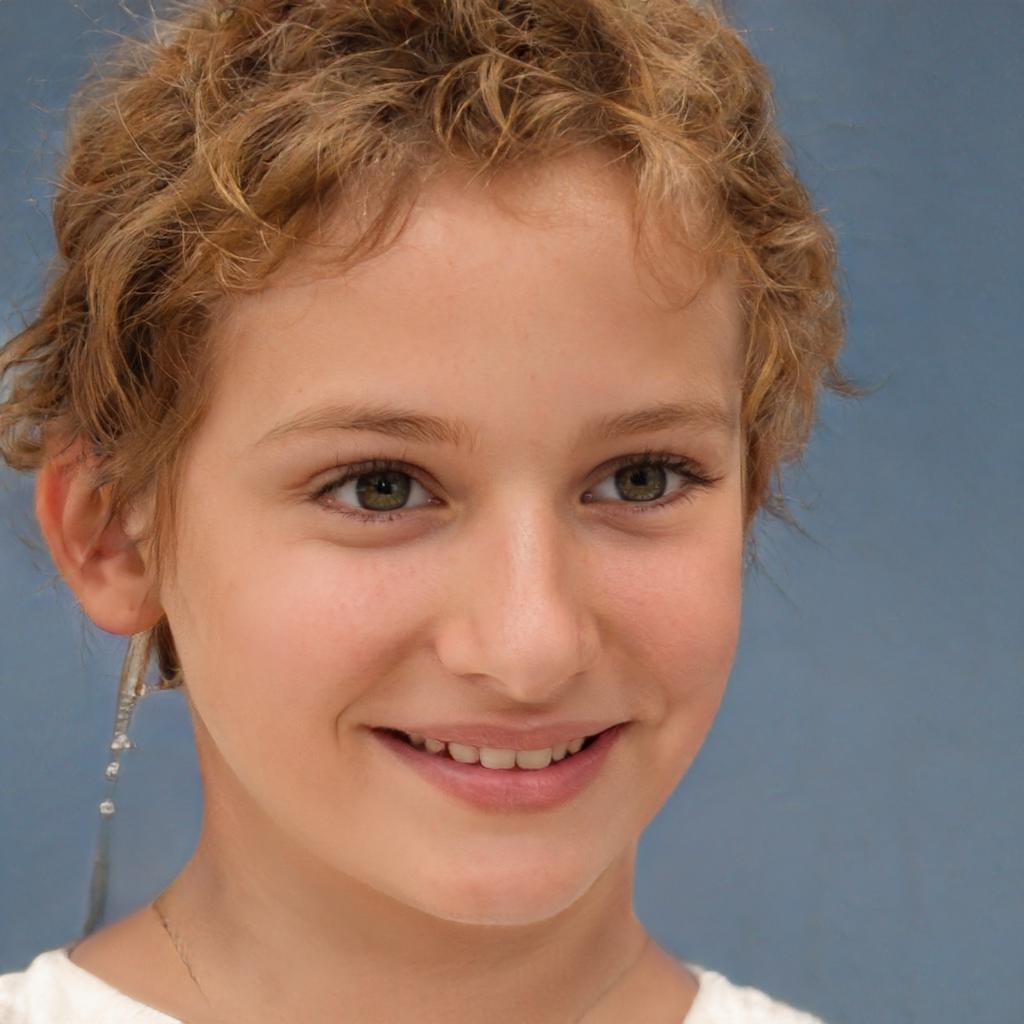} &
        \includegraphics[width=0.25\linewidth]
        {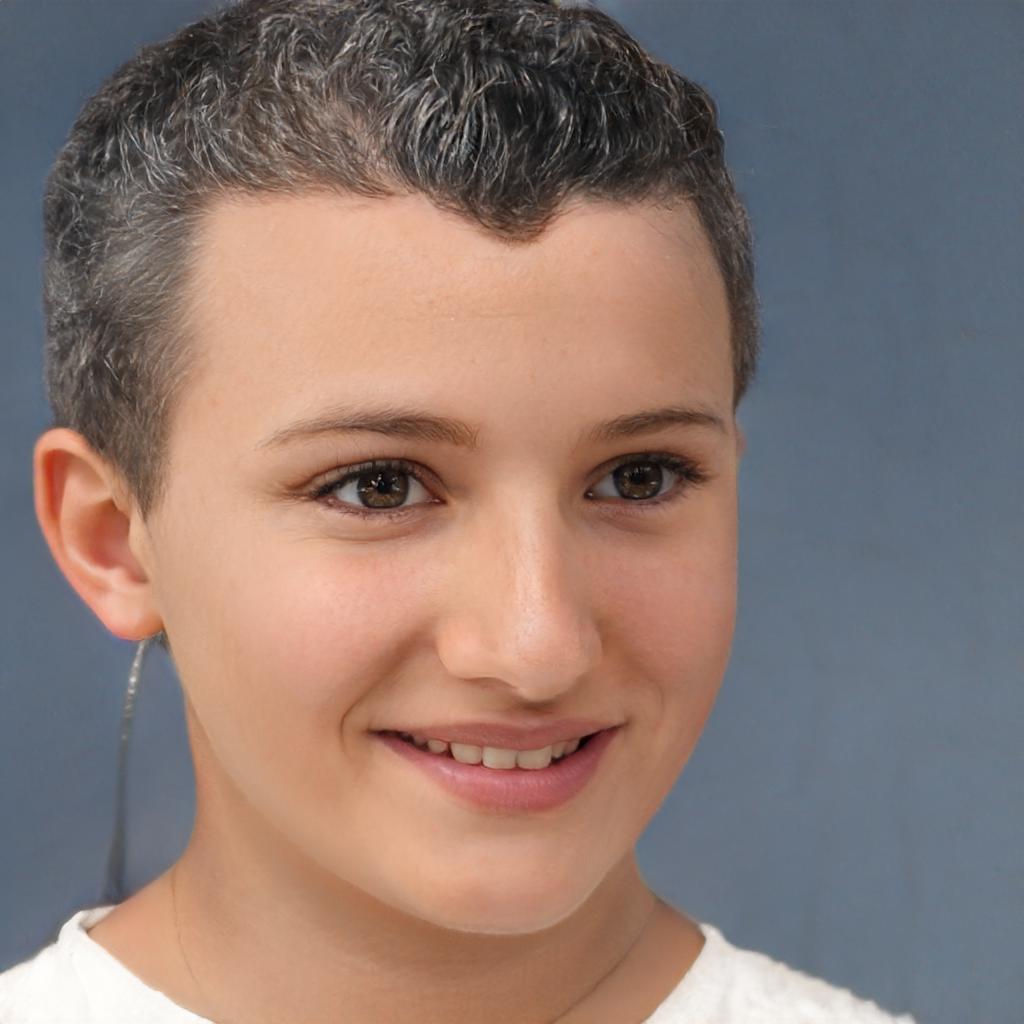}
        \tabularnewline
        \end{tabular}
    \caption{\textit{Hair transfer using StyleFusion.} 
    Given the same target individual (shown to the left) at different poses, StyleFusion is able to transfer the hairstyle from multiple reference images (shown in the top row). Observe the hairstyle consistency along each column and the pose consistency along each row.}
    \label{fig:cross_pose_hair}
\end{figure}

\subsection{Real Image Editing}
In the previous section, we demonstrated how StyleFusion's learned disentanglement provides users with both local and global control over the generated images. In this section, we extend this and illustrate how the representation learned by StyleFusion can be paired with existing latent space editing techniques to achieve improved editing control over \textit{real} images. 

To manipulate real images, it is first necessary to invert them into their latent code representations. This is typically done via per-image optimization~\cite{zhu2016generative,lipton2017precise,Bau_2019,creswell2018inverting,abdal2019image2stylegan,abdal2020image2stylegan++,karras2020analyzing,tewari2020pie,roich2021pivotal} or by training an encoder to learn a direct mapping from a given image to its corresponding latent code~\cite{zhu2016generative,perarnau2016invertible,zhu2020domain,pidhorskyi2020adversarial,richardson2020encoding,tov2021designing,alaluf2021restyle,chai2021latent}. For a comprehensive survey on GAN inversion, we refer the reader to Xia~\etal~\cite{xia2021gan}.
In this work, we adopt the e4e encoder from Tov \etal~\cite{tov2021designing} as it is designed to facilitate improved editing on the resulting inversions.

Many methods resort to manipulating images by traversing the latent space. 
A main challenge in doing so is editing a specific image region (e.g., hairstyle) by modifying a latent code that controls the entire image.
As shown above, StyleFusion allows one to manipulate a latent code controlling a specific target region(s) of the image. 
It is therefore natural to pair existing latent traversal methods with StyleFusion for manipulating only the latent code controlling the desired region. 

Given an input image, we first invert the image into its $\mathcal{W}+$ latent representation, which we denote by $w$. We then manipulate the resulting code to obtain the edited representation $w_{edit}$ (e.g., via a traversal along a latent path learned by InterFaceGAN or GANSpace). 
We then pass to StyleFusion's hierarchy the $\mathcal{S}$-representation of $w_{edit}$ in all regions relevant to the desired edit with $w$ being passed as input to the remaining FusionNets. For example, when performing the ``fearful eyes" edit we pass $w_{edit}$ as input to $FN_{face}$ while the original code $w$ is passed as input to the remaining networks.  

In Figure~\ref{fig:latent_editing_comparison} we show the advantage of using StyleFusion's disentangled representation when editing images using three latent traversal editing methods: InterFaceGAN~\cite{shen2020interpreting}, GANSpace~\cite{harkonen2020ganspace}, and StyleCLIP~\cite{patashnik2021styleclip}.
For InterFaceGAN and GANSpace we use their official implementation and latent directions while for StyleCLIP we adapt the official implementation to train a latent mapper which manipulates only the desired image region. 

As can be seen, pairing these editing methods with StyleFusion's disentangled representation leads to more accurate, local edits that more faithfully alter the desired image regions. For example, when StyleFusion is not used, observe the change in the face shape in the 
``smile'' edit across all examples or the wrinkles that are removed in the ``mohawk'' edit of the second example. Additionally, observe the change in skin color and eye color when editing the hair albedo in the fourth example. Overall, when StyleFusion is paired with existing editing techniques, the resulting edits are able to better preserve the original head shape and facial identity.  

\begin{figure*}[t]
    \begin{subfigure}{0.5\textwidth}
    \setlength{\tabcolsep}{1pt}
    \centering
    {\small
    	\setlength{\tabcolsep}{0pt}
        \begin{tabular}{c c c c c c}
        \raisebox{0.215in}{\rotatebox[origin=t]{90}{No Global}} & 
        \hspace{1pt} &
        \includegraphics[width=0.23\linewidth]
        {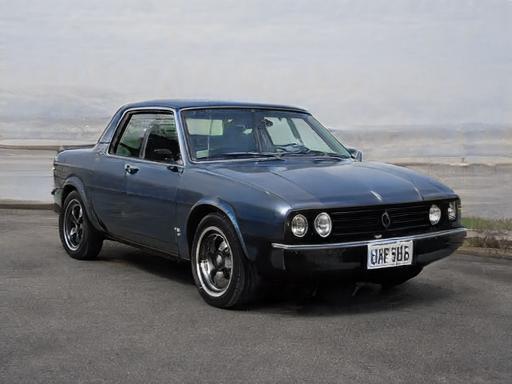} &
        \includegraphics[width=0.23\linewidth]
        {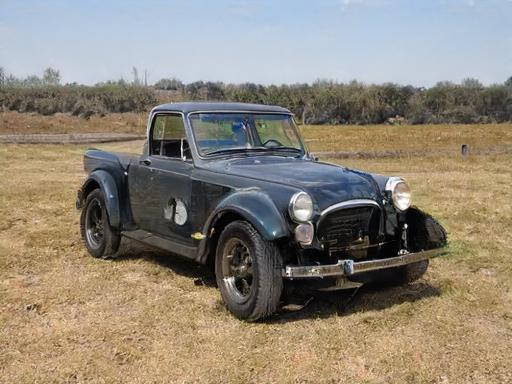} &
        \includegraphics[width=0.23\linewidth]
        {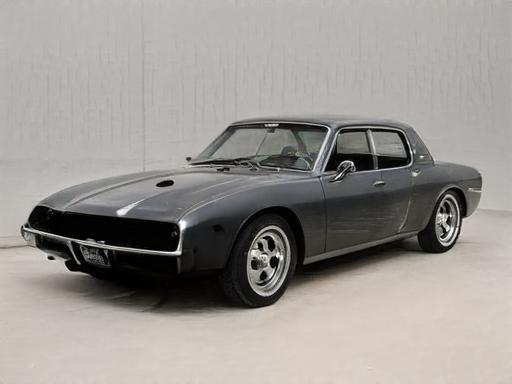} &
        \includegraphics[width=0.23\linewidth]
        {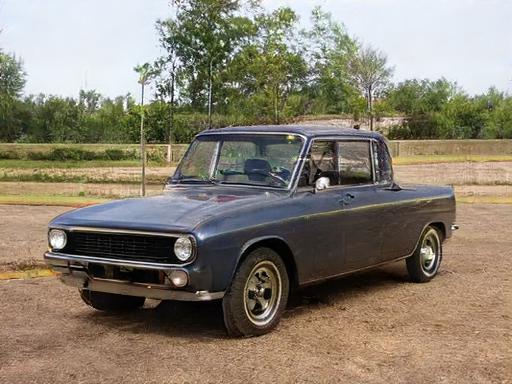}
        \tabularnewline
        \raisebox{0.215in}{\rotatebox[origin=t]{90}{No Stage 3}}& 
        \hspace{1pt} &
        \includegraphics[width=0.23\linewidth]
        {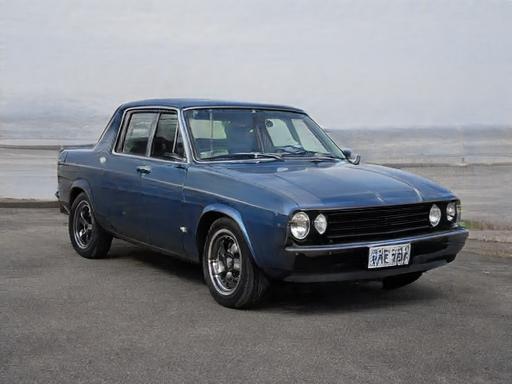} &
        \includegraphics[width=0.23\linewidth]
        {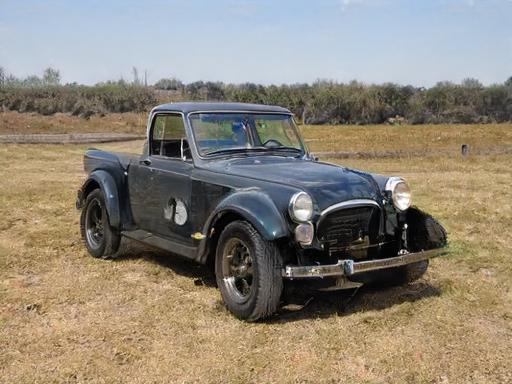} &
        \includegraphics[width=0.23\linewidth]
        {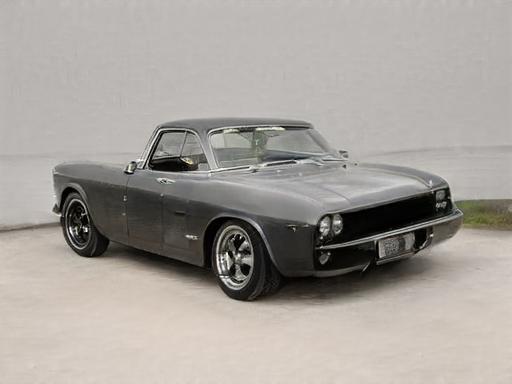} &
        \includegraphics[width=0.23\linewidth]
        {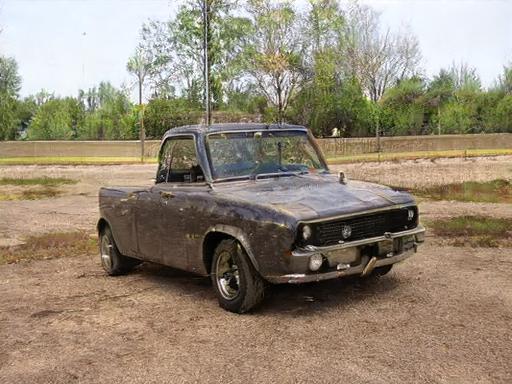}
        \tabularnewline
        \raisebox{0.215in}{\rotatebox[origin=t]{90}{No Stage 2}}& 
        \hspace{1pt} &
        \includegraphics[width=0.23\linewidth]
        {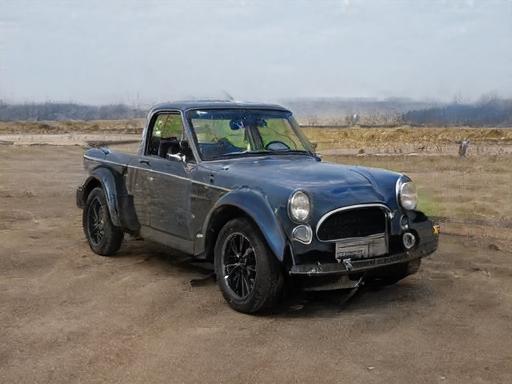} &
        \includegraphics[width=0.23\linewidth]
        {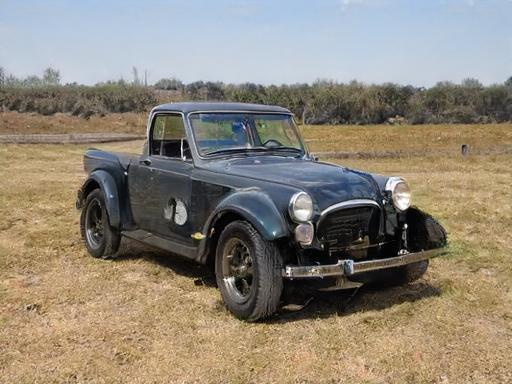} &
        \includegraphics[width=0.23\linewidth]
        {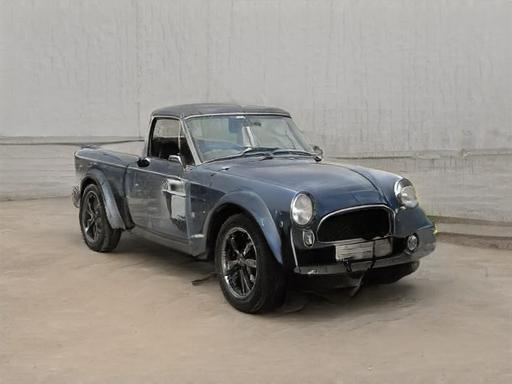} &
        \includegraphics[width=0.23\linewidth]
        {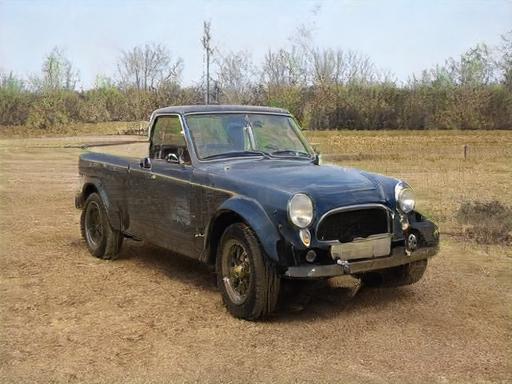}
        \tabularnewline
        \raisebox{0.215in}{\rotatebox[origin=t]{90}{No Stage 1}}& 
        \hspace{1pt} &
        \includegraphics[width=0.23\linewidth]
        {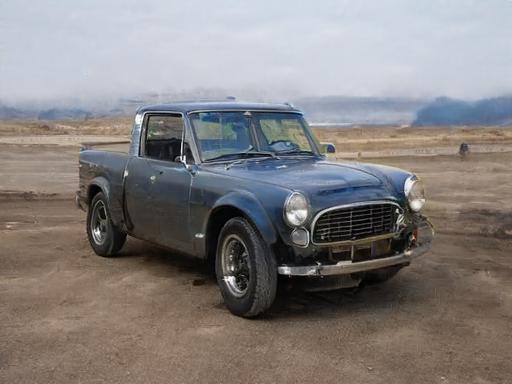} &
        \includegraphics[width=0.23\linewidth]
        {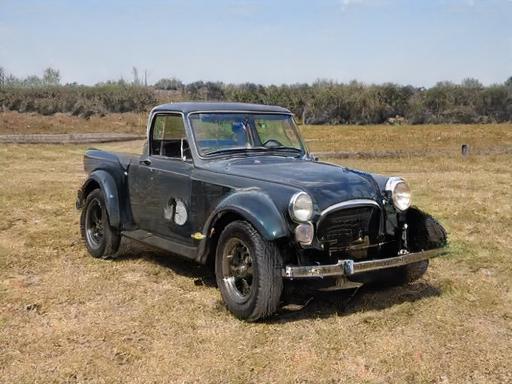} &
        \includegraphics[width=0.23\linewidth]
        {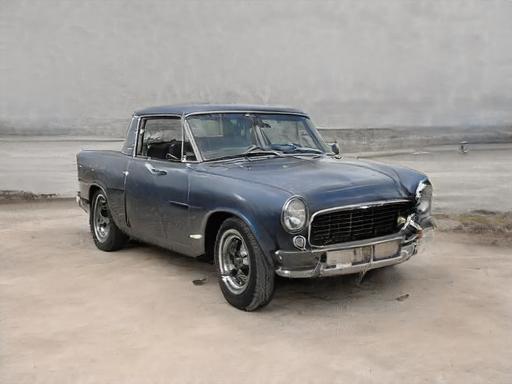} &
        \includegraphics[width=0.23\linewidth]
        {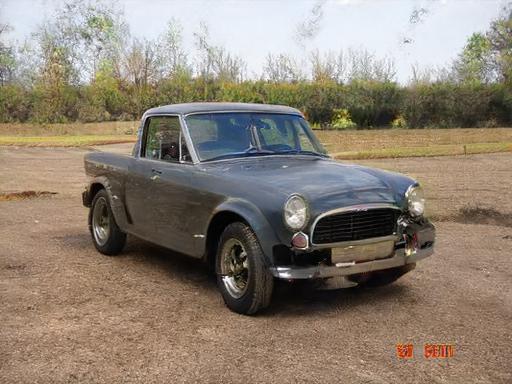}
        \tabularnewline
        \raisebox{0.215in}{\rotatebox[origin=t]{90}{StyleFusion}}& 
        \hspace{1pt} &
        \includegraphics[width=0.23\linewidth]
        {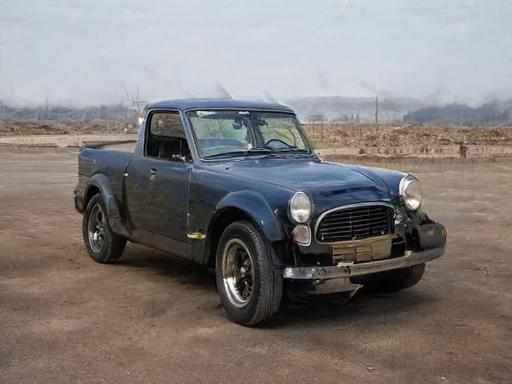} &
        \includegraphics[width=0.23\linewidth]
        {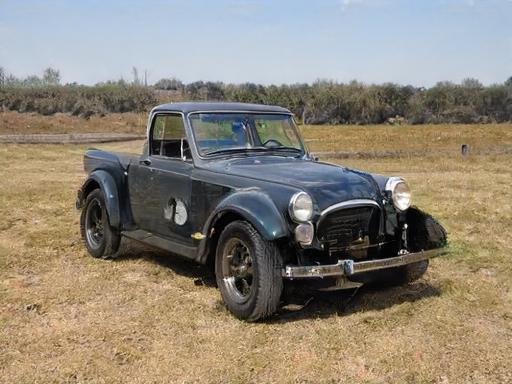} &
        \includegraphics[width=0.23\linewidth]
        {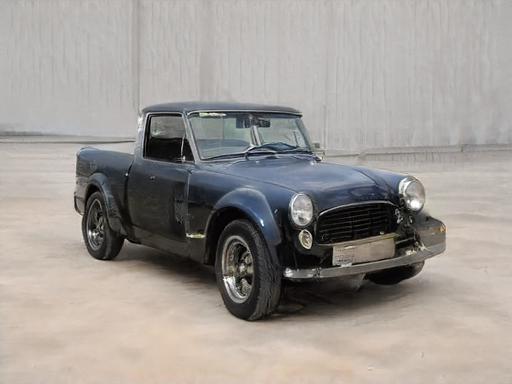} &
        \includegraphics[width=0.23\linewidth]
        {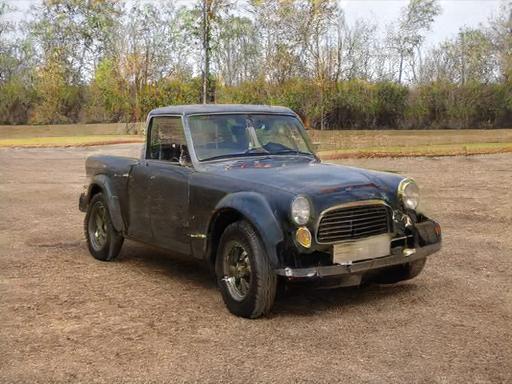}
        \tabularnewline
        \end{tabular}
    }
    \caption{}
    \label{fig:ablation_car}
    \end{subfigure}%
    \begin{subfigure}{0.5\textwidth}
    \setlength{\tabcolsep}{1pt}
    \centering
    {\small
    	\setlength{\tabcolsep}{0pt}
        \begin{tabular}{c c c c c c}
        \raisebox{0.215in}{\rotatebox[origin=t]{90}{No Global}} & 
        \hspace{1pt} &
        \includegraphics[width=0.23\linewidth]
        {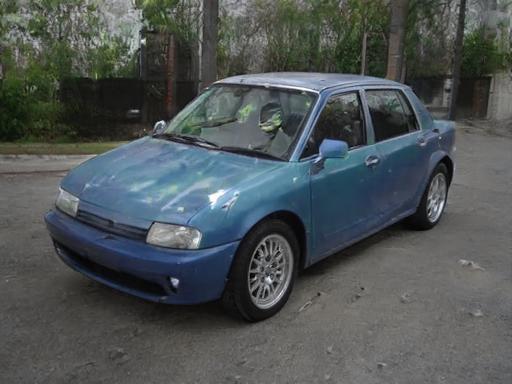} &
        \includegraphics[width=0.23\linewidth]
        {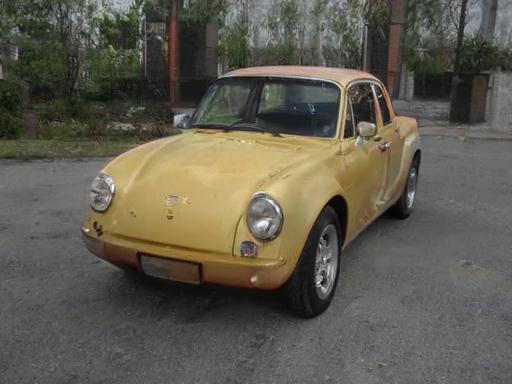} &
        \includegraphics[width=0.23\linewidth]
        {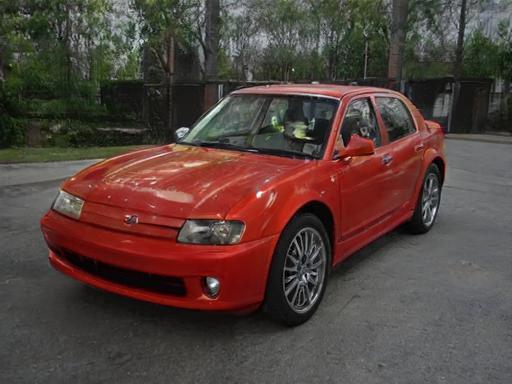} &
        \includegraphics[width=0.23\linewidth]
        {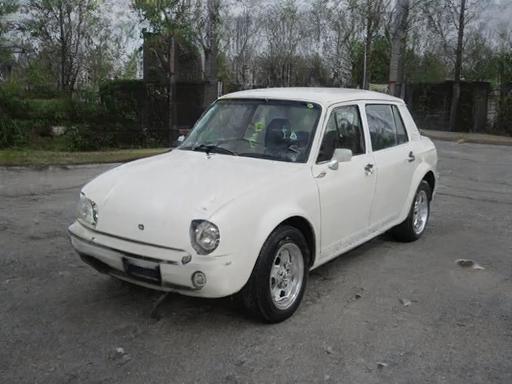}
        \tabularnewline
        \raisebox{0.215in}{\rotatebox[origin=t]{90}{No Stage 3}}& 
        \hspace{1pt} &
        \includegraphics[width=0.23\linewidth]
        {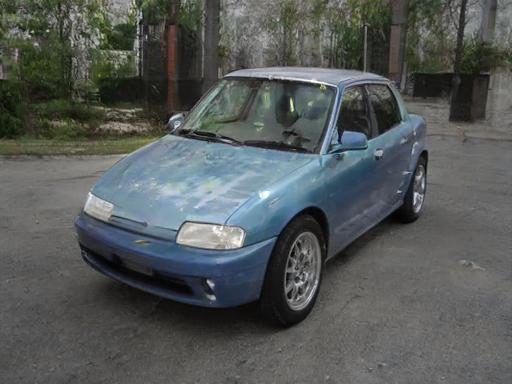} &
        \includegraphics[width=0.23\linewidth]
        {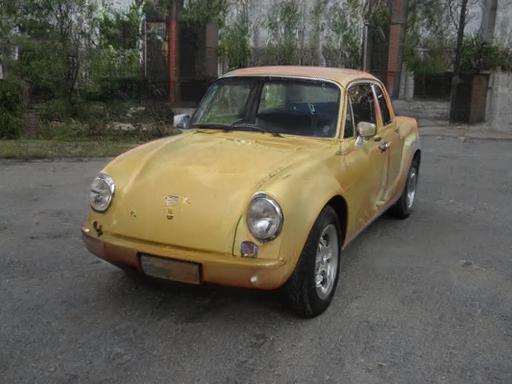} &
        \includegraphics[width=0.23\linewidth]
        {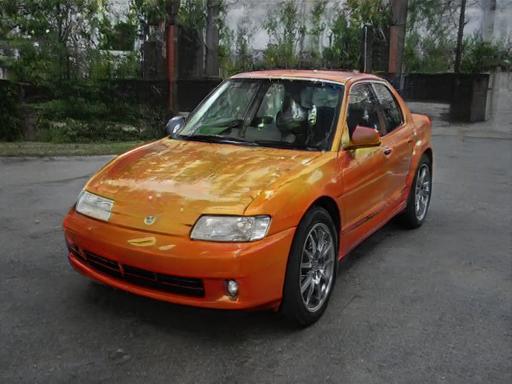} &
        \includegraphics[width=0.23\linewidth]
        {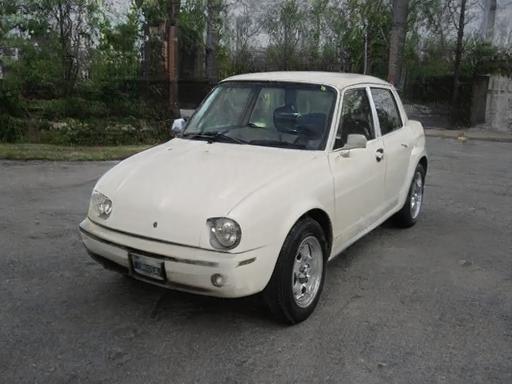}
        \tabularnewline
        \raisebox{0.215in}{\rotatebox[origin=t]{90}{No Stage 2}}& 
        \hspace{1pt} &
        \includegraphics[width=0.23\linewidth]
        {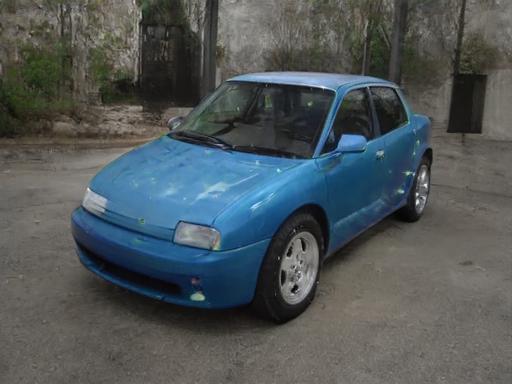} &
        \includegraphics[width=0.23\linewidth]
        {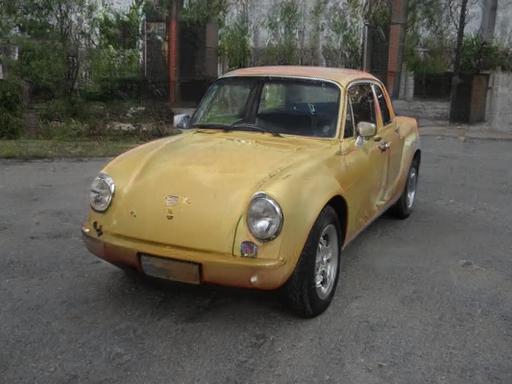} &
        \includegraphics[width=0.23\linewidth]
        {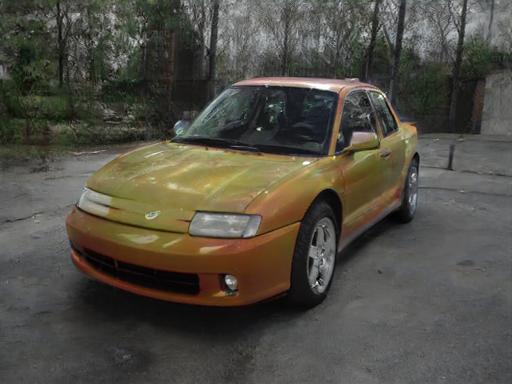} &
        \includegraphics[width=0.23\linewidth]
        {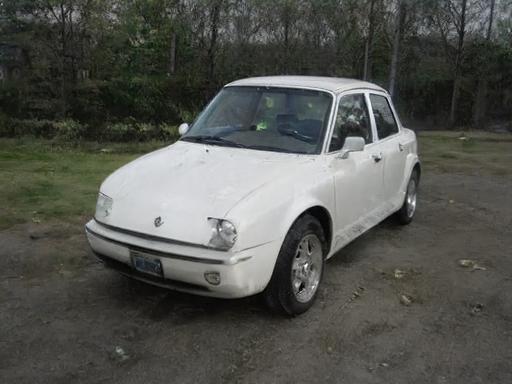}
        \tabularnewline
        \raisebox{0.215in}{\rotatebox[origin=t]{90}{No Stage 1}}& 
        \hspace{1pt} &
        \includegraphics[width=0.23\linewidth]
        {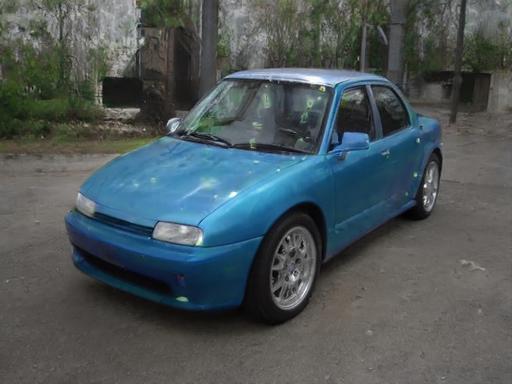} &
        \includegraphics[width=0.23\linewidth]
        {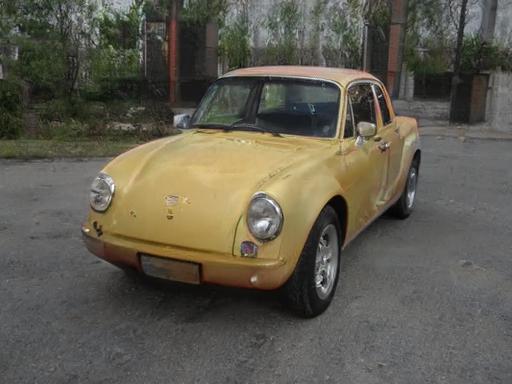} &
        \includegraphics[width=0.23\linewidth]
        {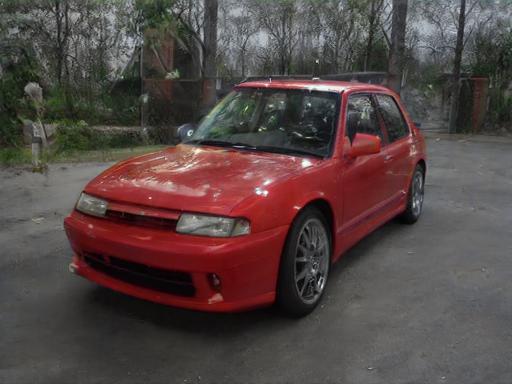} &
        \includegraphics[width=0.23\linewidth]
        {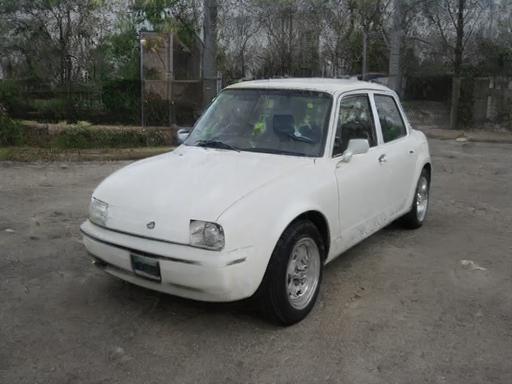}
        \tabularnewline
        \raisebox{0.215in}{\rotatebox[origin=t]{90}{StyleFusion}}& 
        \hspace{1pt} &
        \includegraphics[width=0.23\linewidth]
        {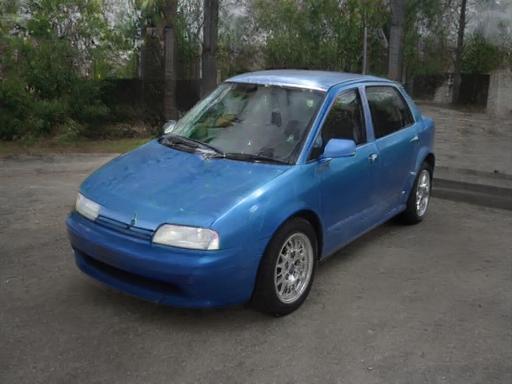} &
        \includegraphics[width=0.23\linewidth]
        {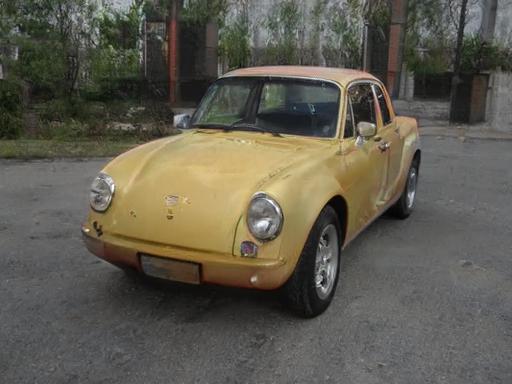} &
        \includegraphics[width=0.23\linewidth]
        {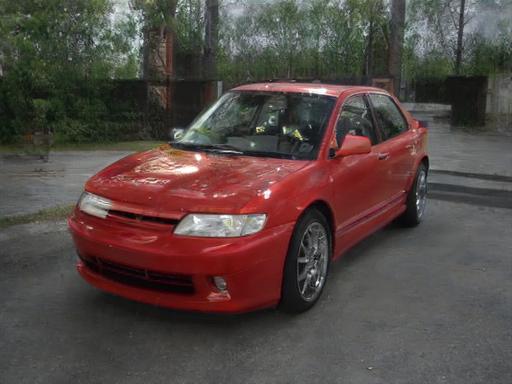} &
        \includegraphics[width=0.23\linewidth]
        {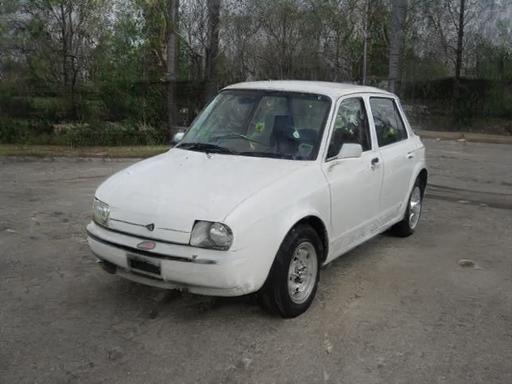}
        \tabularnewline
        \end{tabular}
    }
    \caption{}
    \label{fig:ablation_car_2}
    \end{subfigure}%
    \caption{\textit{Ablation Study.} We train a single FusionNet to disentangle between the car body and the image background. In each column of ~\ref{fig:ablation_car} we alter the image background, while keeping the car body and pose fixed. Conversely, in~\ref{fig:ablation_car_2}, we alter the car body, while keeping the background and pose fixed.
    }
\end{figure*}

\subsection{Hair Transfer}
A natural extension of the local and global editing presented in Section~\ref{sec:results} is allowing one to faithfully transfer a semantic region from a reference image to a given target image. An attribute that has received particular attention is the task of hair transfer~\cite{tan2020michigan,zhu2021barbershop} in which we wish to transfer the hair from a reference image onto a target image without altering the remaining image attributes.
Note that although we focus on the task of hair transfer due to its complexity, this reference-guided editing is applicable to other attributes.
Observe that doing so is especially challenging when the reference and target images are unaligned (e.g., have different poses) and of different ages and gender.

In Figure~\ref{fig:cross_pose_hair}, we demonstrate StyleFusion's ability to perform such a transfer even in the presence of such variations. In the Figure, each row depicts the same source individual at different poses with each column representing a different hairstyle we would like to transfer. Observe how StyleFusion is able to faithfully transfer the hairstyle to the different poses (i.e., the hairstyle remains fixed at each column). Notably, observe that the transfer remains accurate even when the reference image is of individuals of different ages and genders. Notice also that the identity of the source is well-preserved, showing the disentanglement between the hair and face image regions.
\section{Ablation Study}
In this section, we provide an ablation study to validate the design choices of StyleFusion. Specifically, we first demonstrate the importance of the global style code tasked with aligning the two input images. We additionally demonstrate the importance of the multi-step training scheme presented in Section~\ref{multi-step-training}. For our ablation study, we focus on the cars domains and train a FusionNet tasked with disentangling the car body and image background. 

We compare between our complete StyleFusion model and four variants:
\begin{enumerate}
    \item \textit{No Global}: Here, we do not use the additional global code used for aligning the two input images. As such, the FusionNet is composed only of the $LB_{fuse}$ Latent Blender component (see Figure~\ref{fig:blendnet_architecture}).
    \item \textit{No Stage 1}: Training is performed without the first stage. Observe, that in this setting, the \textit{mask loss} is not used for training the FusionNet.
    \item \textit{No Stage 2}: Training is performed without the second stage where only $LB_{fused}$ is trained. 
    \item \textit{No Stage 3}: Training is performed without the third stage where both $LB_{fuse}$ and $LB_{align}$ are trained simultaneously. Observe that the localization loss $L_{local}$ is utilized only in this stage.
\end{enumerate}

We begin our analysis by referring the reader to Figure~\ref{fig:ablation_car}. Here, the latent code controlling the car body and the global latent code remain fixed while the latent code controlling the image background is changed across each column. That is, each image along the columns should depict the same car model at the same pose, but with different backgrounds. Observe how when the global latent code is not utilized (in the top row), the car body varies greatly across the four outputs. 
Comparing our full model to the other variants, notice how StyleFusion is able to better preserve coarse details (e.g., the car build varies in row 2) while better preserving smaller details along the wheels (e.g., the car wheels vary in row 3). Finally, comparing our full results with those shown in the fourth row illustrates the effectiveness of the first stage of training in improving the preservation of small details.

In Figure~\ref{fig:ablation_car_2} we perform a similar process, but here, the background and global latent codes remain fixed while the latent code controlling the car body is varied. Here, notice the importance of the fusion and localization losses utilized in the second and third stages of training. Specifically, without these losses, the results suffer from a slight mode collapse, mainly with respect to the car's color. 

The results of the two evaluations demonstrate that StyleFusion is successful in making local, yet meaningful and easily perceptible changes to the generated images.

\section{Discussion and Conclusions}
We have presented a novel approach in which an image is generated from a set of disentangled latent codes, each controlling a single semantic region of the image. Specifically, we showed that such a generator can be built by pairing a pre-trained generator with a hierarchical mapping network that fuses a set of latent codes into a single one.

\begin{figure}
    \centering
    \setlength{\tabcolsep}{0pt}
    {\small
        \begin{tabular}{c c c c}
        \includegraphics[width=0.25\linewidth]
        {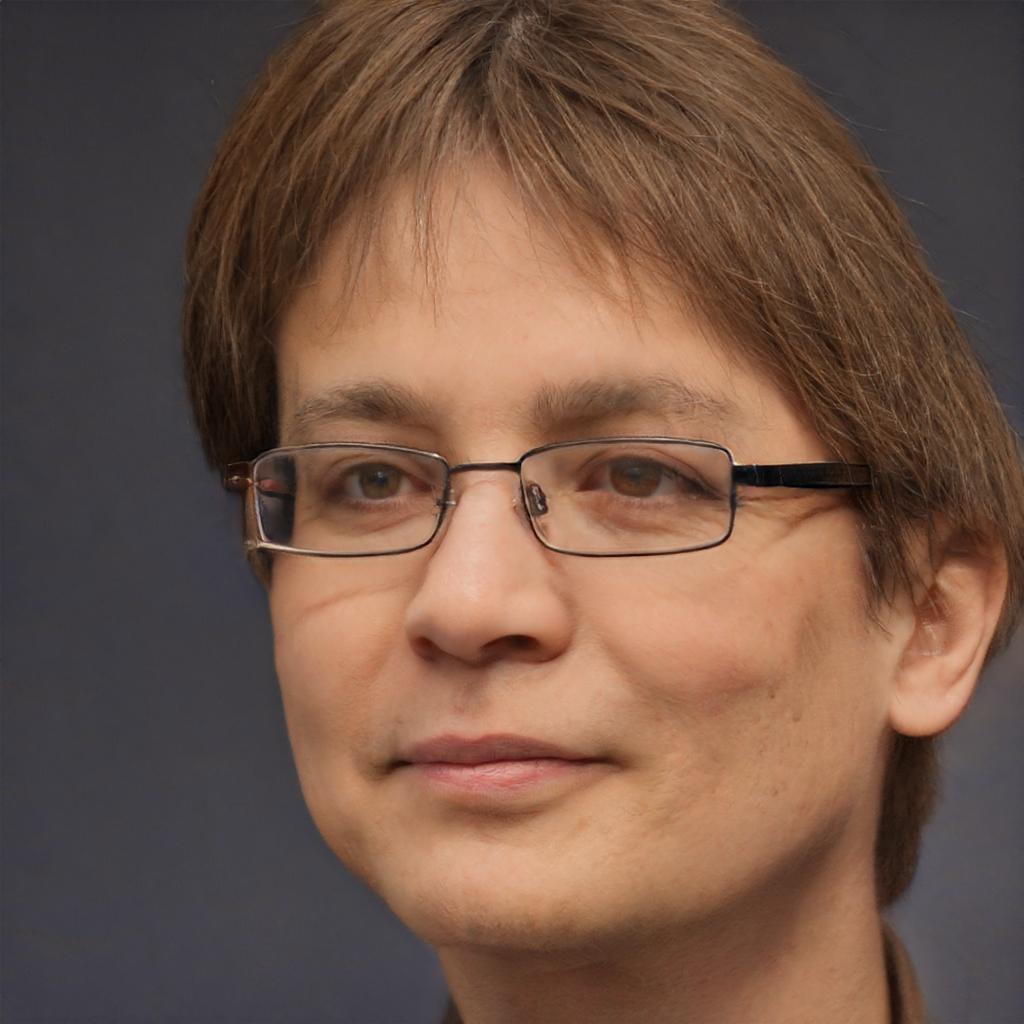} &
        \includegraphics[width=0.25\linewidth]
        {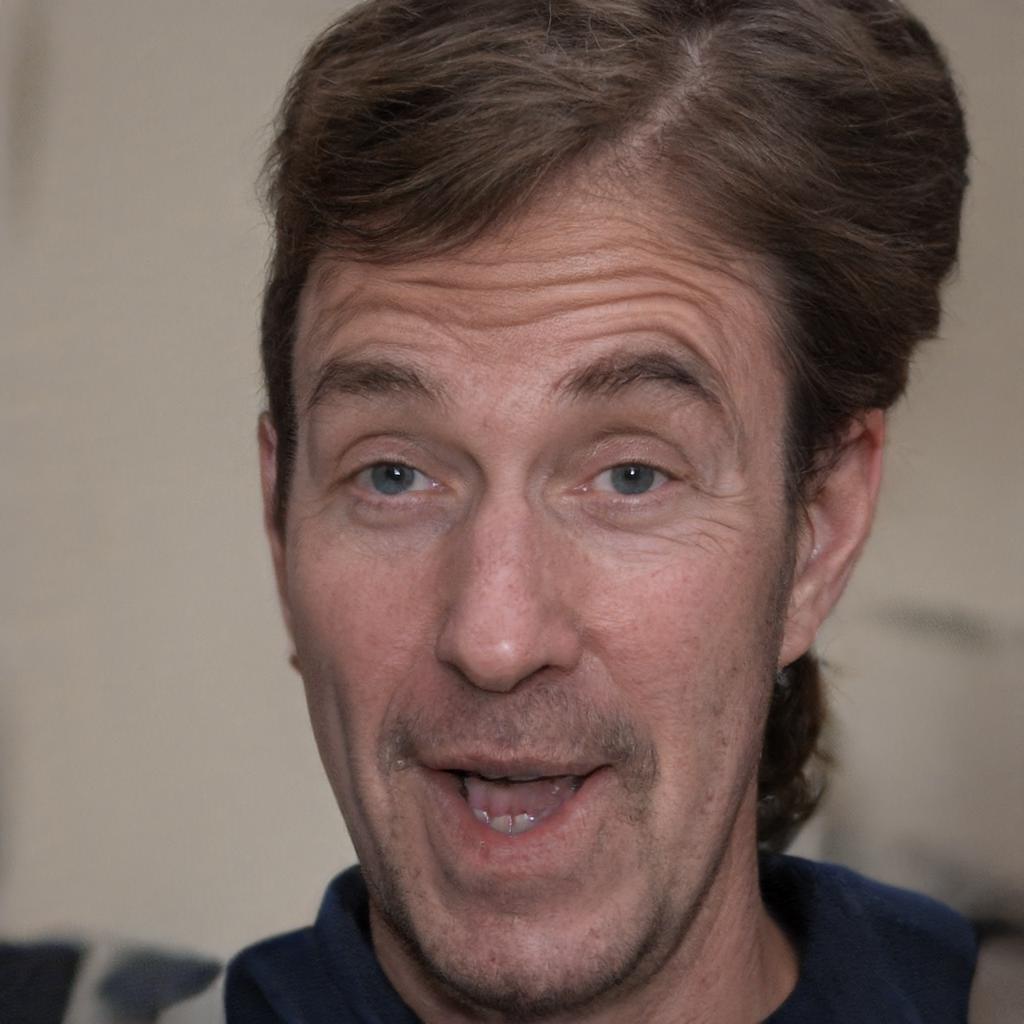} &
        \includegraphics[width=0.25\linewidth]
        {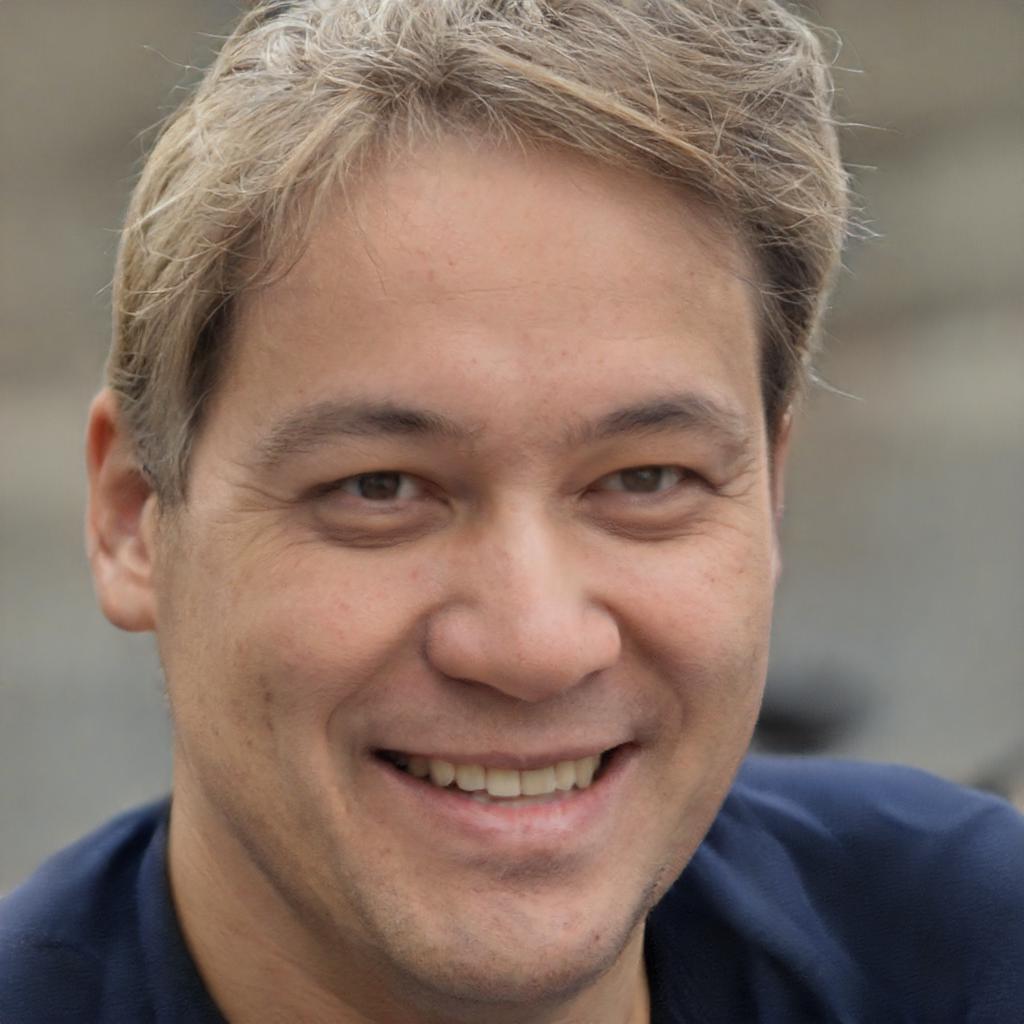} &
        \includegraphics[width=0.25\linewidth]
        {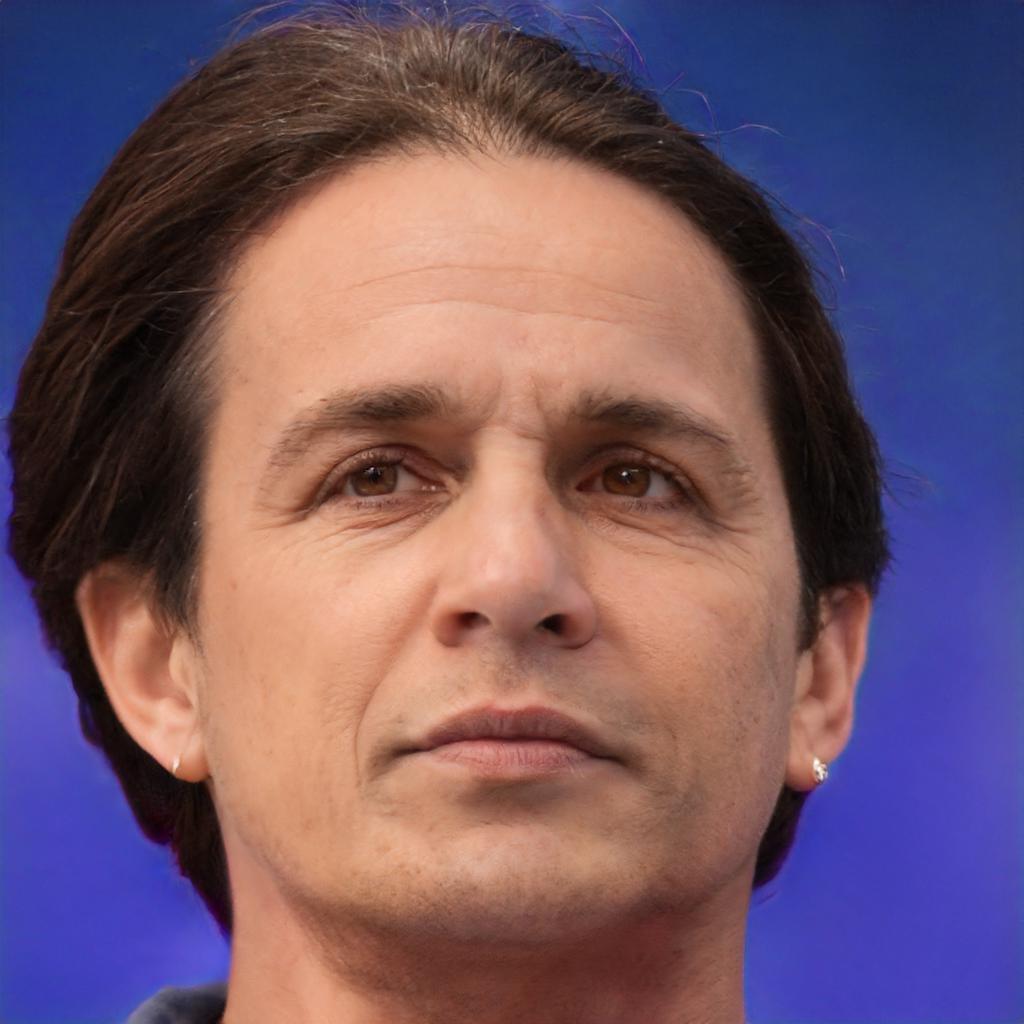}
        \tabularnewline
        \includegraphics[width=0.25\linewidth]
        {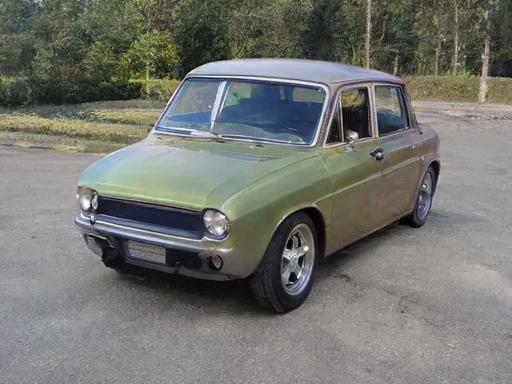} &
        \includegraphics[width=0.25\linewidth]
        {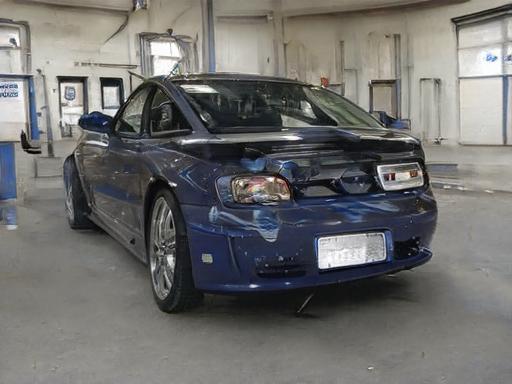} &
        \includegraphics[width=0.25\linewidth]
        {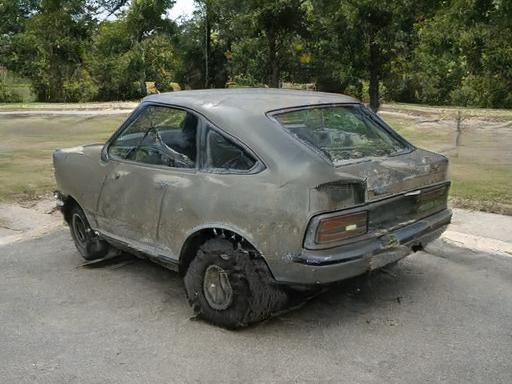} &
        \includegraphics[width=0.25\linewidth]
        {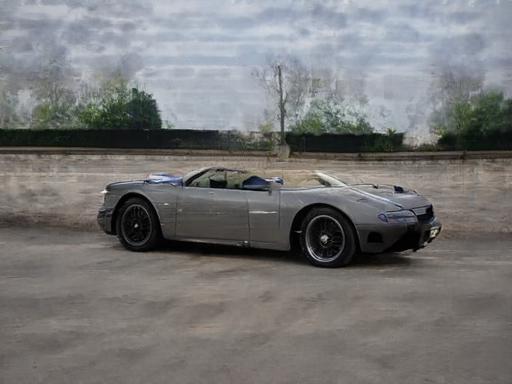}
        \end{tabular}
    }
    \caption{\textit{Limitations.} 
    Although StyleFusion is effective in aligning semantic regions, when given images of extreme poses, the resulting local and global control may result in unwanted artifacts. For example, in the first row we control the hair with a latent code that is strongly unaligned with the base latent code. 
    In the second row, in some cases, altering the global pose of the car may result in skewed and unrealistic car shapes.}
    \label{fig:limitations}
\end{figure}

While we have demonstrated the competence of StyleFusion in learning a semantically-aware disentanglement of images, there are several limitations that should be considered. First, while the global input code is effective in aligning images to the same spatial layout, when the global code represents an image with an extreme pose or under-represented layout, the alignment stage may result in unwanted artifacts, see Figure~\ref{fig:limitations}.

While we have demonstrated the locality of StyleFusion's control over semantic regions in Section~\ref{local_control}, there may still remain some level of entanglement between different image attributes. This is most prominent when editing more global features such as hairstyle, which may oftentimes be entangled with facial identity. 
 
Our fusion network is trained hierarchically, which provides control over the granularity of the learned disentanglement. This implies a sequential top-down training, where the disentanglement of finer features is dependent on the fidelity of the disentanglement of coarser semantics.

It should be noted that many of the results presented in the paper can possibly be achieved using previous editing techniques.
Yet, our StyleFusion mechanism itself is not an editing method, but a mapping network that facilitates performing local editing over semantic regions by refining existing latent space editing techniques.

The contributions of our work are two-fold. First, we introduce a new approach for designing generators. Instead of feeding the synthesis network a single latent code that controls the image in its entirety, we feed a latent code for each semantic region of the image.
The resulting fused, disentangled code facilitates better flexibility over the synthesis process.
Second, we propose implementing this design by training a new mapping function that takes a set of latent codes into the latent space of a pre-trained synthesis network. 

In a sense, our fusion network introduces a new disentangled latent space. 
It should be noted that using a pre-trained generator inherently bounds the disentanglement that can be achieved by our method.
A future research direction is to re-design the channels of the generator itself to achieve better disentanglement and control over semantic regions, or to train such a generator that receives several latent codes in an end-to-end fashion.

{\small
\bibliographystyle{ieee_fullname}
\bibliography{egbib}
}

\clearpage
\appendix
\appendixpage
In this appendix, we provide additional implementation details and provide additional results obtained using our proposed StyleFusion method.

\section{Implementation Details}

\subsection{FusionNet Hierarchies}~\label{sec:fusionnet_hierarchies}
The hierarchy of FusionNets for each domain is illustrated below. Observe that each leaf node corresponds to an input latent code with each parent node corresponding to a single FusionNet. For simplicity, we omit the global input latent code in the diagram. Note, however, that each node in the tree receives a \textit{different} input global latent code. For example, the global codes used to train the ``Face" and ``Skin,Mouth" may differ.

\vspace{0.25cm}
\textit{\textbf{Faces:}}

\Tree [.All [.Face [.{Skin, Mouth} Skin Mouth ] Eyes ] [.{BG, Hair, Clothes} [.{BG, Clothes} Background Clothes ] Hair ] ]

\vspace{0.25cm}
\textit{\textbf{Cars:}}

\Tree [.All [.Car Body Wheels ] [.Background {BG Top} {BG Bottom} ] ]

\vspace{0.25cm}
\textit{\textbf{Churches:}}

\Tree [.All Building Background ]

\subsection{Losses}~\label{sec:appendix_losses}
Each FusionNet can be trained with different configurations of loss lambdas. We use $3$ training configurations to train all of the FusionNets: \textit{Regular, Short, and Favored}. We described each configuration below:

\vspace{0.25cm}
\topic{\textit{Regular. }} Here, each FusionNet is trained for a total of $25,000$ steps with each stage trained with the following number of steps:
\begin{enumerate}
    \item Stage 1: $5,000$ steps
    \item Stage 2: $10,000$ steps
    \item Stage 3: $10,000$ steps
\end{enumerate}
The loss coefficients are set as follows:
For the mask loss, the loss coefficient is set as $\lambda_{mask} = 30,000$. For the localization loss we set $\varepsilon$ to $0.1$ with $\lambda_{local} = 15,000$. For the alignment regularization we set $\lambda_{alignReg} = 6$ in Stage 1 and $\lambda_{alignReg} = 1$ in Stage 3. Finally, we set $\lambda_{fusion}$ to $375$ in all stages.

\topic{\textit{Short. }} 
This configuration is identical to the \textit{Regular} configuration above, with the exception that Stage 3 is omitted here during training.

\topic{\textit{Favored. }} 
The human eye may be more susceptible to changes in certain image regions than other regions. For example, when considering the human facial domain, we may tend to notice changes along the face region rather than changes in the background. As a result, we design a configuration to allow the FusionNet to attend to one region over the other.
This configuration is the same as the \textit{Regular} configuration, but here, the coefficients $\lambda_{fusion}$ and $\lambda_{local}$ are not equal for the two regions. For the favored region we set $\lambda_{fusion} = 500$ and $\lambda_{local} = 20,000$. For the other region we use $\lambda_{fusion} = 250$ and $\lambda_{local} = 10,000$.

Below we list the configuration used to train each FusionNet across the three domains explored.

\topic{\textit{Faces: }}
\begin{itemize}
    \item All: Favored (toward the face region)
    \item Face: Short
    \item Skin, Mouth:  Regular
    \item BG, Hair, Clothes: Regular
    \item BG, Clothes:  Regular
\end{itemize}

\topic{\textit{Cars: }}
\begin{itemize}
    \item All: Favored (toward the car region)
    \item Car: Short
    \item Background: Short
\end{itemize}

\topic{\textit{Churches: }}
\begin{itemize}
    \item All: Favored (toward the building region)
\end{itemize}

\subsection{Latent Blender Architecture}~\label{sec:latent_blender_arch}

As described in the main paper, the network $D$ of the Latent Blender receives an input of size $2|\mathcal{S}|$ and returns a fusion coefficient $q$ of size $|\mathcal{S}|$. 
If the network $D$ were to be a fully-connected network, the input layer would be of size $|\mathcal{S}| = 9,088$ (for the human facial domain).
To reduce the overall size of $D$, we treat each of the generator's input layers independently. 
Specifically, consider two input latent codes $s_A,s_B\in\mathcal{S}$. 
The pre-trained StyleGAN generator receives a subset of the channels of $s_A$ and $s_B$ at each input layer. For some input layer $l$, we denote this sub-vector by $s_A^l,s_B^l$ respectively.
Observe that in the standard StyleGAN2 generator for the human facial domain, there are $18$ input layers and for all layers $l$, $|s_A^l|,|S_B^l|\in \{512, 256, 128, 64, 32\}$. 

Given the definition of $s_A^l$ and $s_B^l$, the network $D$ operates by receiving $s_A^l,s_B^l$ for each layer $l$ independently through the same fully-connected network, and concatenating each of the outputs. 
To encode which layer is given to the network we define a layer indication vector $v(l)$, which is a $1$-hot vector marking which layer is given as input.
In addition, some input layers in StyleGAN2 are specialized tRGB layers. As such, we additionally define an RGB indication bit $r(l)$ which is $1$ if the given layer is a tRGB layer and $0$ otherwise.
For each layer $l$, $s_A^l$ and $s_B^l$ are first up-sampled to dimension $512$ and then concatenated together with $v(l)$ and $r(l)$, both repeated $10$ times. 
The output of $D$ for each layer is max-pooled to the original layer dimension, which results in the concatenated result being the original dimension of $\mathcal{S}$. 

Assuming that there are $18$ layers in the pre-trained generator, the input of the resulting fully-connected network will be of length $1,214$ ($512\times 2 + (18 + 1) \times 10$) and the output will be of length $512$. As a result, the architecture presented here allows for a much smaller network than a naive fully-connected network.
Specifically, the fully-connected network $D$ is a $6$-layer network with the following dimensions:
[1214, 1536, 1536, 1536, 512, 512].

\subsection{$LB_{align}$ Modification}
In Karras \etal~\cite{karras2020analyzing}, the authors demonstrate the first input layers of the generator control coarse-level details (e.g., pose) while the final layers control the finer-level details (e.g., lighting). 
Motivated by this, we design the Latent Blender $LB_{align}$, tasked with aligning the input codes, to consider only the coarse input layers of the generator. 
Specifically, given two input latent codes $s$ and $s_{align}$, $LB_{align}$ only fuses the first $5$ layers of $s$ and $s_{align}$. For the remaining generator layers $l$, we set $s_{result}^l = s^l$. That is, we copy the remaining sub-vector of $s$ to the final fused latent code. 

\begin{figure}
    \centering
    \includegraphics[width=\linewidth]{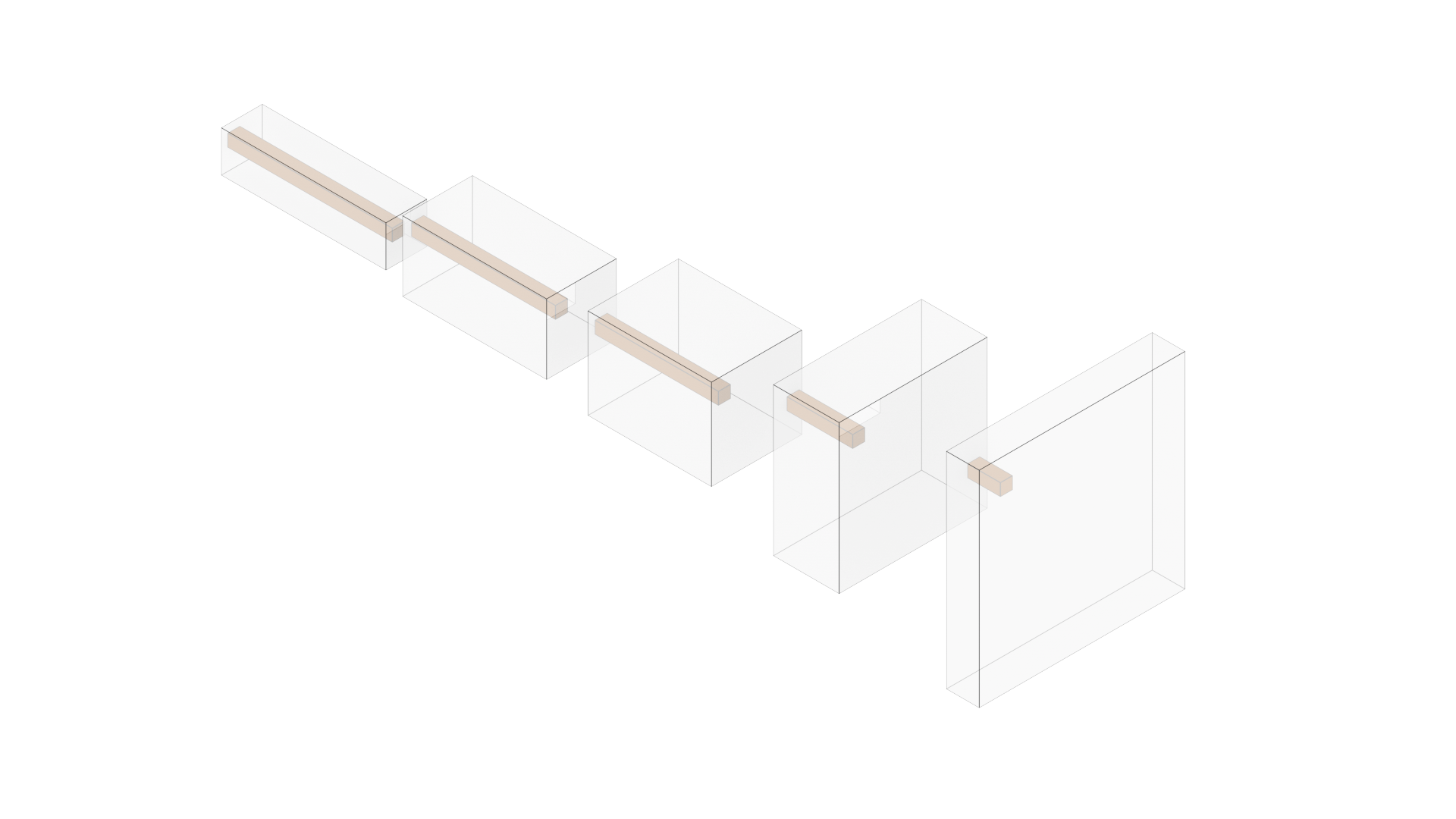}
    \vspace{-0.1cm}
    \caption{
    \textit{DeepPixel.} We define the DeepPixel at a spatial location $(x,y)$ as the $C$-dimensional vector obtained by  
    concatenating all activation tensors of StyleGAN's synthesis network when generating a given image.}
    \label{fig:deep_pixel}
\end{figure}

\subsection{Computing the Mask Loss}~\label{sec:mask_loss}
In this section, we describe how to compute the masks in a differentiable manner, which is needed for calculating $\mathcal{L}_{mask}$ in an end-to-end fashion. 
Specifically, we wish to calculate the mask for an image generated by our StyleGAN generator.
As described in Section~\ref{sec:segmentation}, we can compute a joined representation of the image by concatenating all the intermediate activation tensors of StyleGAN2's synthesis network. This results in a representation of the image of size $H \times W \times C$ where $|C|=6,048$ and $H \times W$ are the dimensions of the largest activation map. 
Consider some spatial location of this representation and the corresponding $C$-dimensional vector. We call this representation of the spatial location a \textit{DeepPixel}, see Figure~\ref{fig:deep_pixel}.
Now, consider two regions $R_1, R_2$ and a mask $m\in [0,1]^{W\times H \times 2}$. We define $m_{x,y,i}$ to represent the probability that the DeepPixel at location $(x,y)$ belongs to the region $R_i$. 
To compute $m_{x,y,i}$ we first calculate all the distances between the DeepPixel at $(x,y)$ and each of the segmentation clusters. 
We then apply soft-min over the distances and sum all the clusters labeled as region $R_i$ to obtain the value of $m_{x,y,i}$.

To reduce computation costs and memory usage, we compute the masks at a lower resolution of $64\times64$ and up-sample to dimension $H\times W$.

\subsection{Truncation}
During training, we sample latent codes by generating a random noise and passing it through the StyleGAN's pre-trained mapping network. We then use a truncation of $0.7$ for the human facial domain and $0.5$ for the cars and churches domains.

\section{Additional Results}~\label{sec:appendix_results}
\begin{figure*}
    \centering
    \setlength{\tabcolsep}{0pt}
    {\small
        \begin{tabular}{c c c c c c}
        \raisebox{0.65in}{\rotatebox[origin=t]{90}{Hair}} & 
        \hspace{1pt} &
        \includegraphics[width=0.23\linewidth]
        {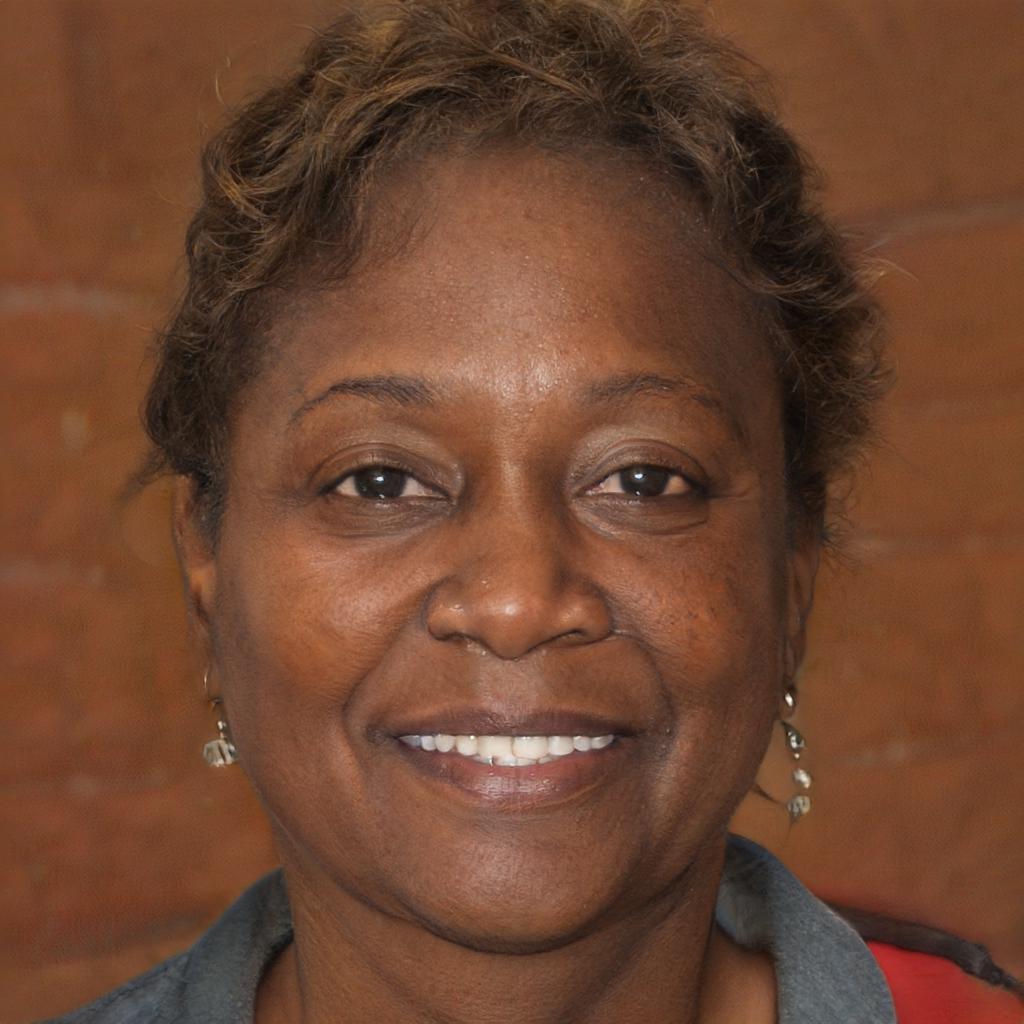} &
        \includegraphics[width=0.23\linewidth]
        {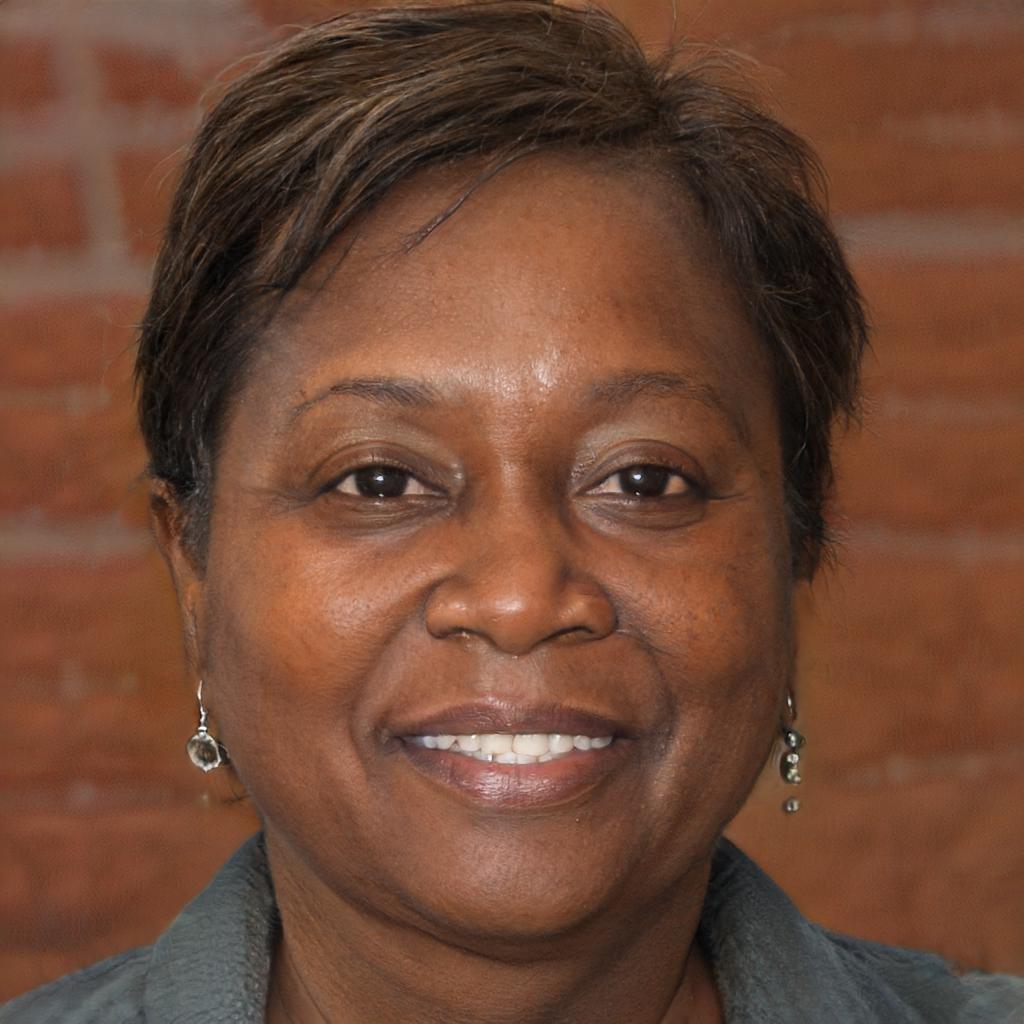} &
        \includegraphics[width=0.23\linewidth]
        {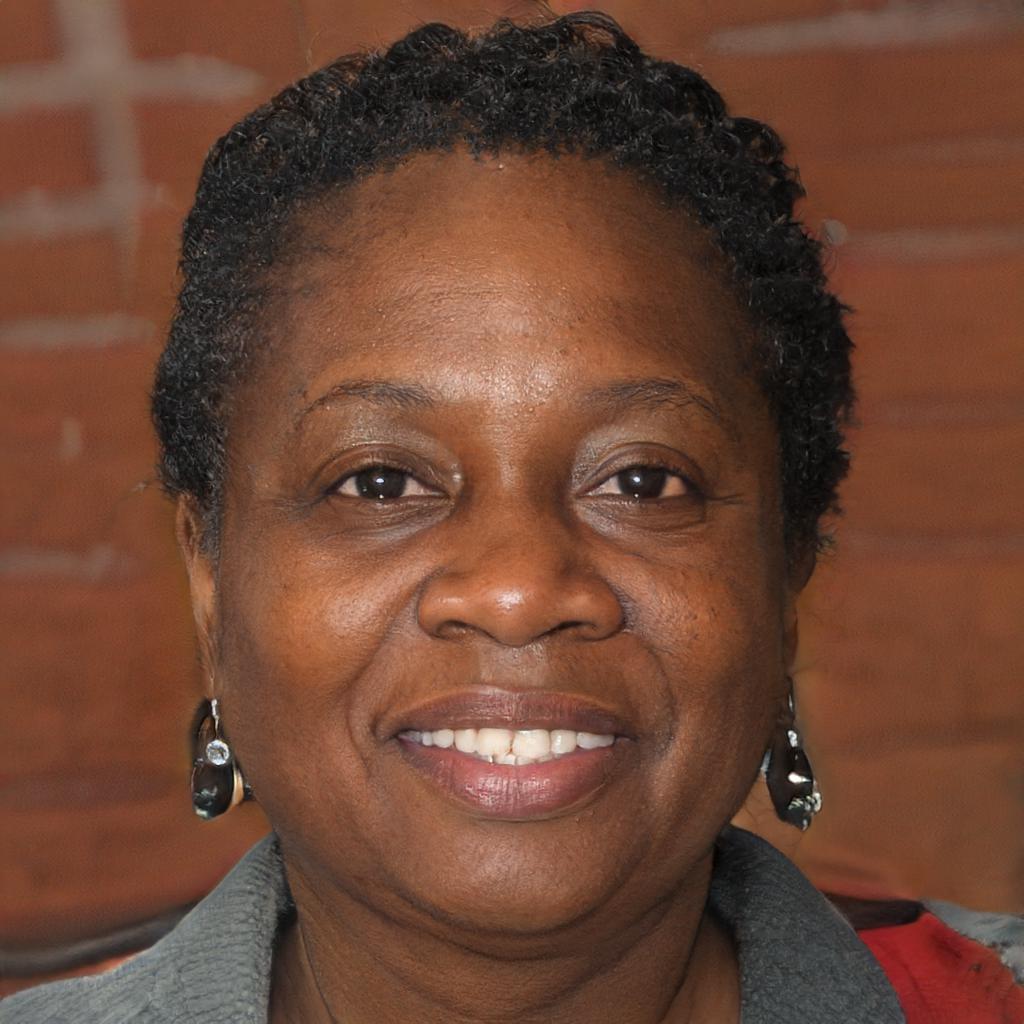} &
        \includegraphics[width=0.23\linewidth]
        {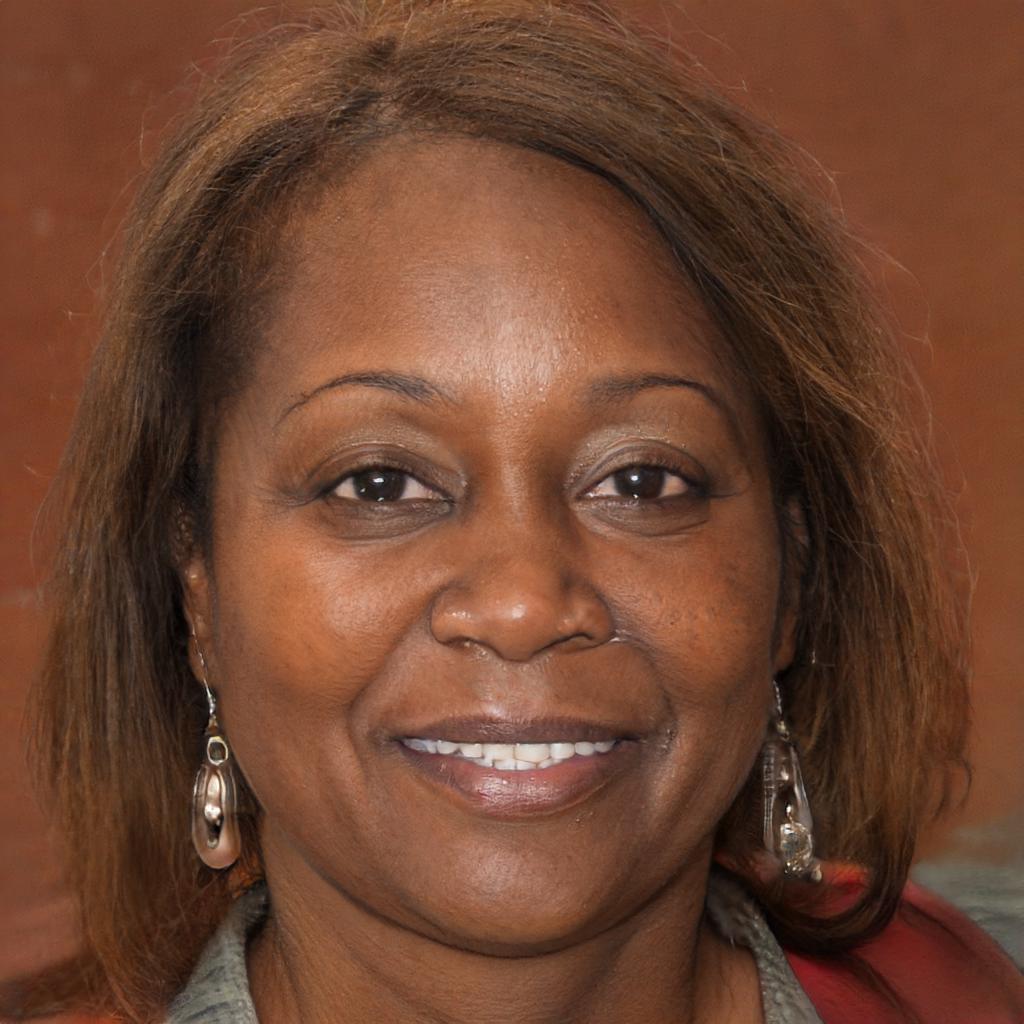}
        \tabularnewline
        \raisebox{0.65in}{\rotatebox[origin=t]{90}{Face}}& 
        \hspace{1pt} &
        \includegraphics[width=0.23\linewidth]
        {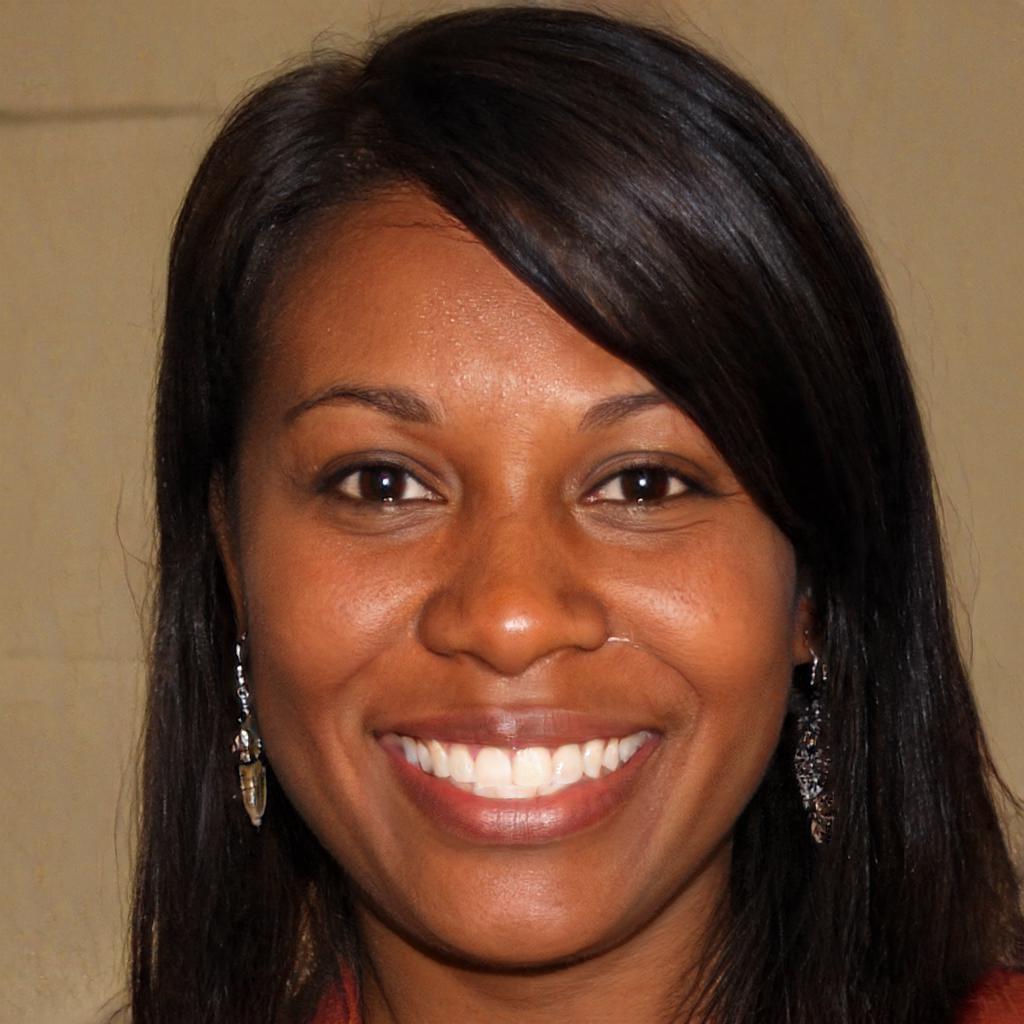} &
        \includegraphics[width=0.23\linewidth]
        {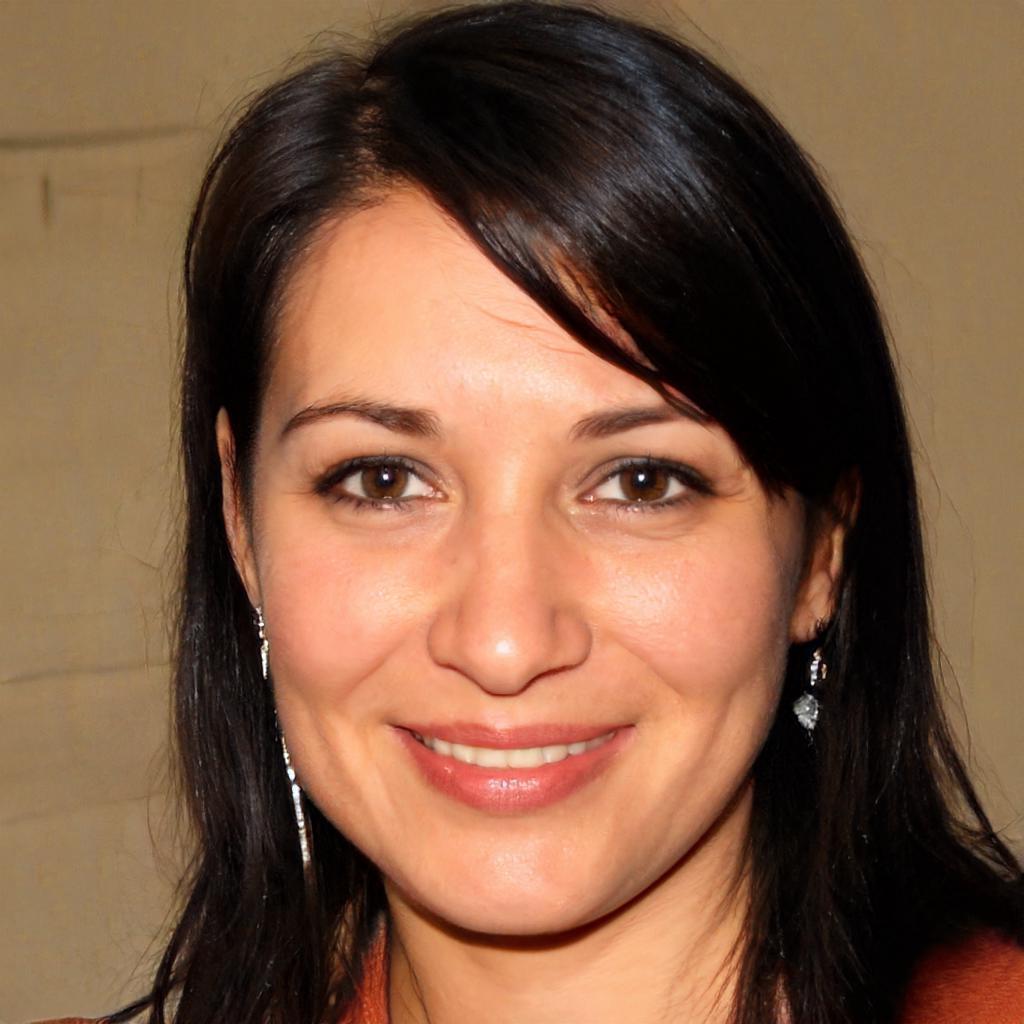} &
        \includegraphics[width=0.23\linewidth]
        {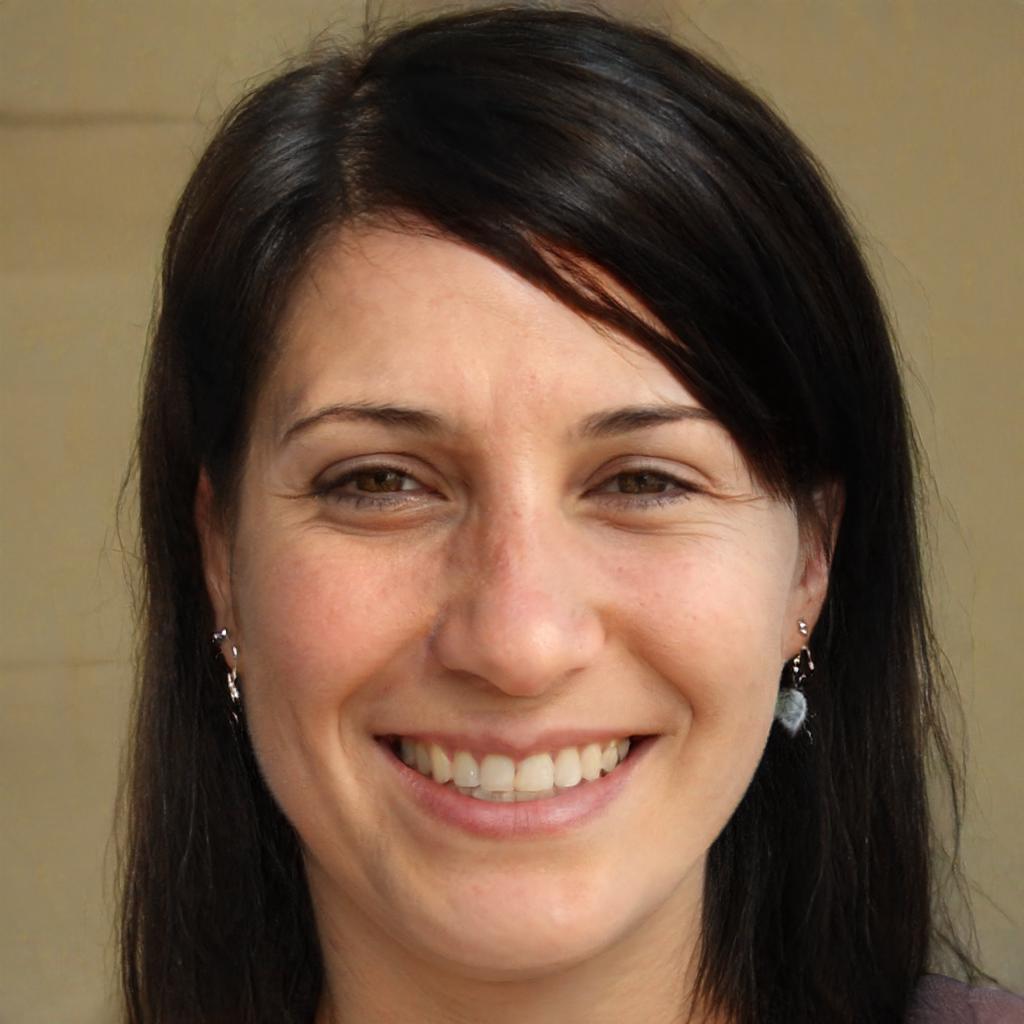} &
        \includegraphics[width=0.23\linewidth]
        {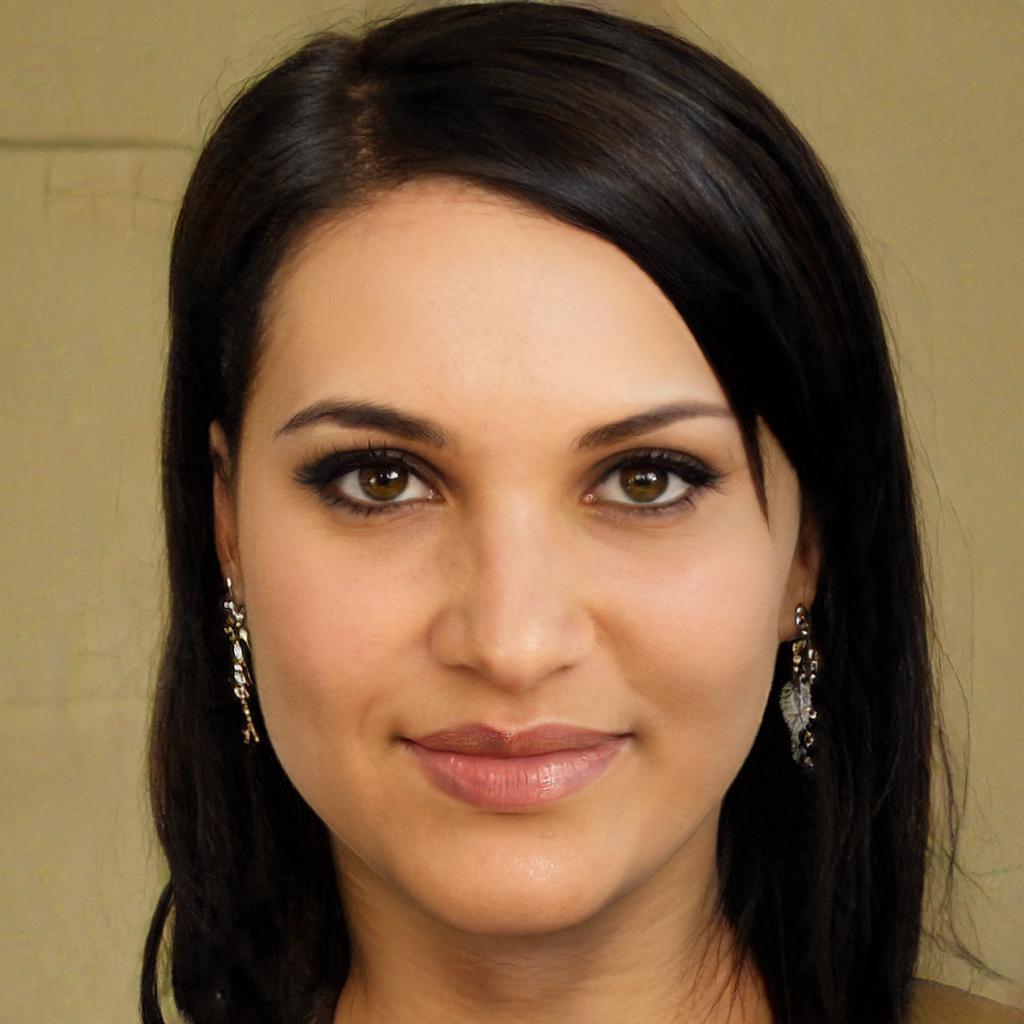}
        \tabularnewline
        \raisebox{0.65in}{\rotatebox[origin=t]{90}{Mouth}}& 
        \hspace{1pt} &
        \includegraphics[width=0.23\linewidth]
        {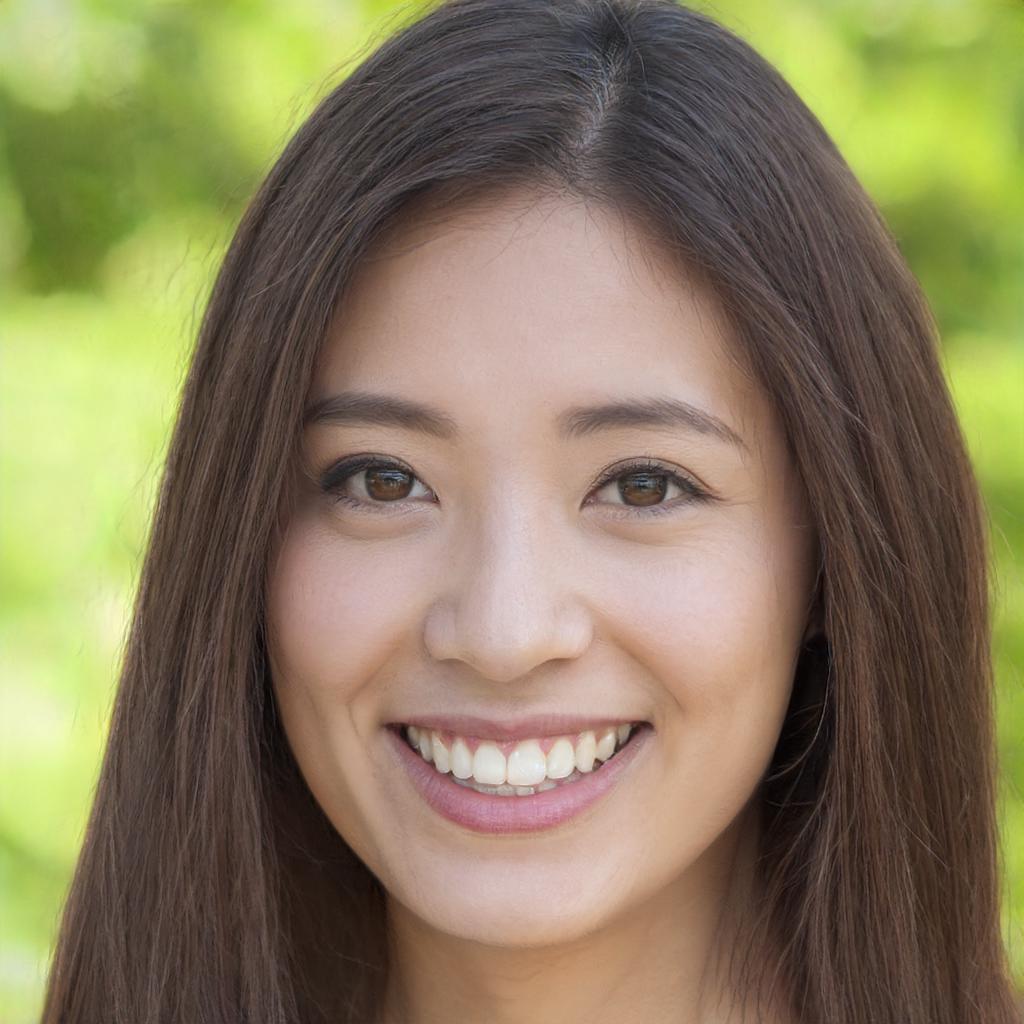} &
        \includegraphics[width=0.23\linewidth]
        {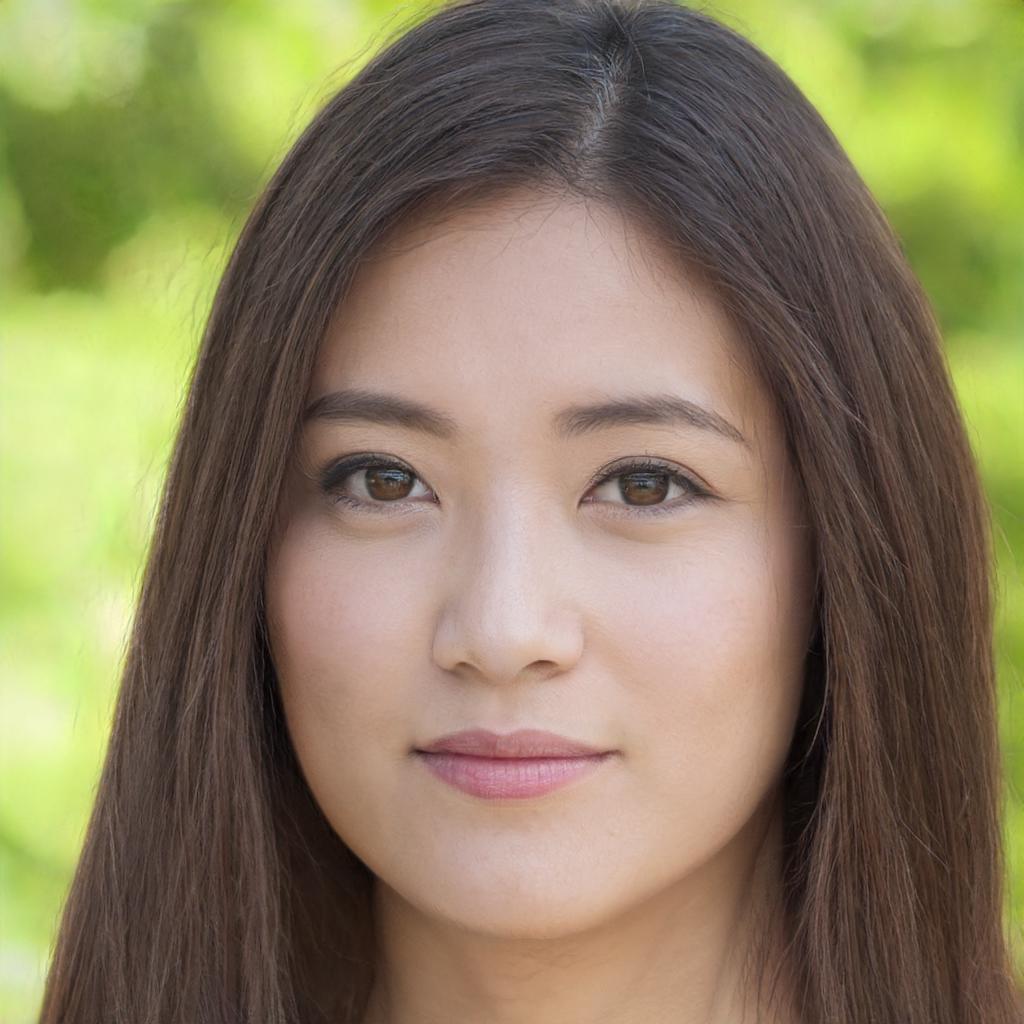} &
        \includegraphics[width=0.23\linewidth]
        {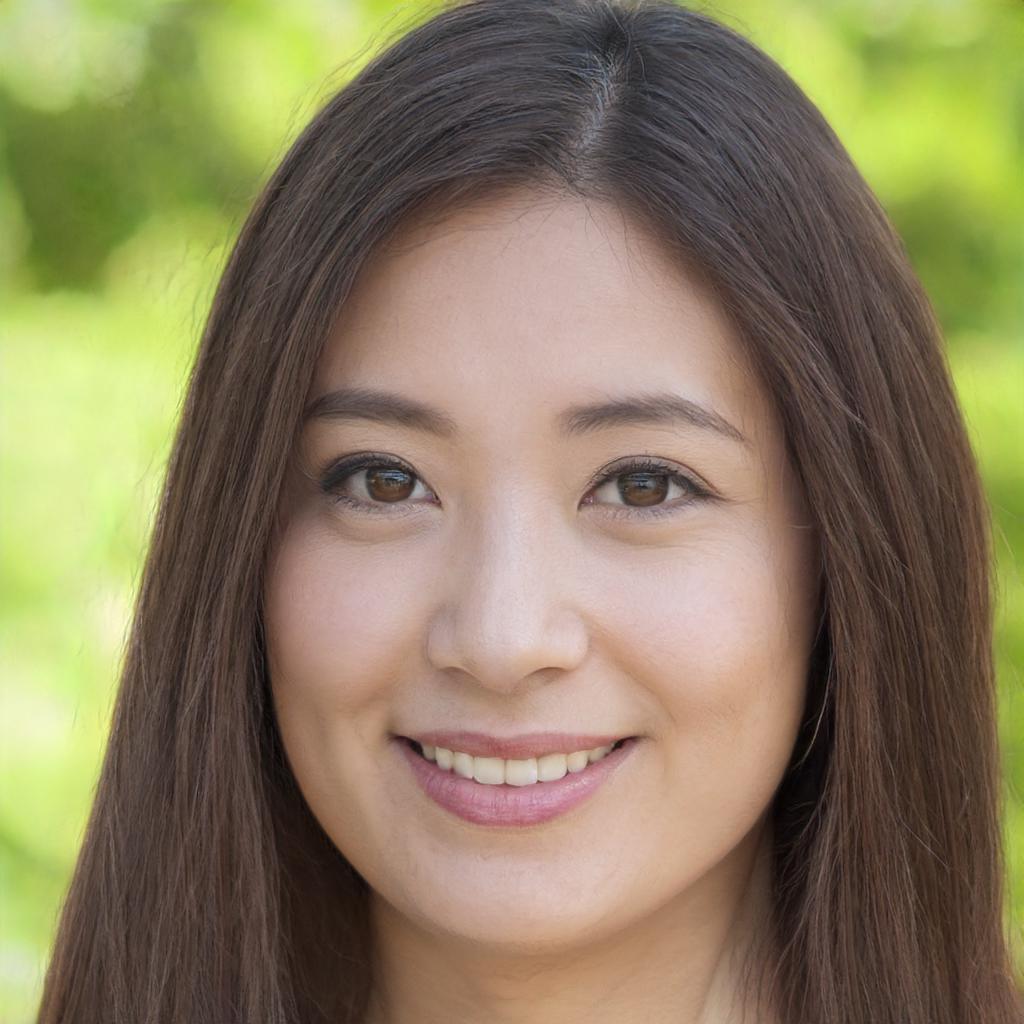} &
        \includegraphics[width=0.23\linewidth]
        {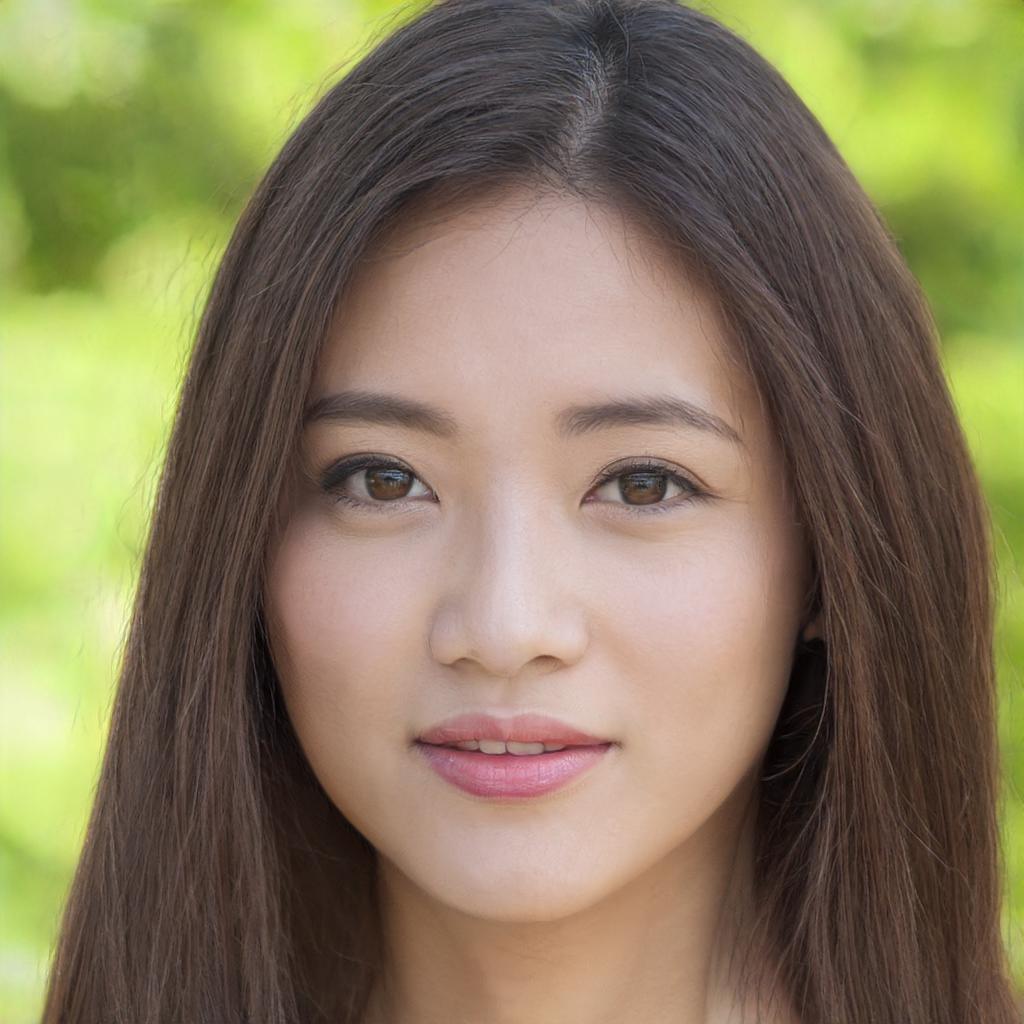}
        \tabularnewline
        \raisebox{0.65in}{\rotatebox[origin=t]{90}{Eyes}}& 
        \hspace{1pt} &
        \includegraphics[width=0.23\linewidth]
        {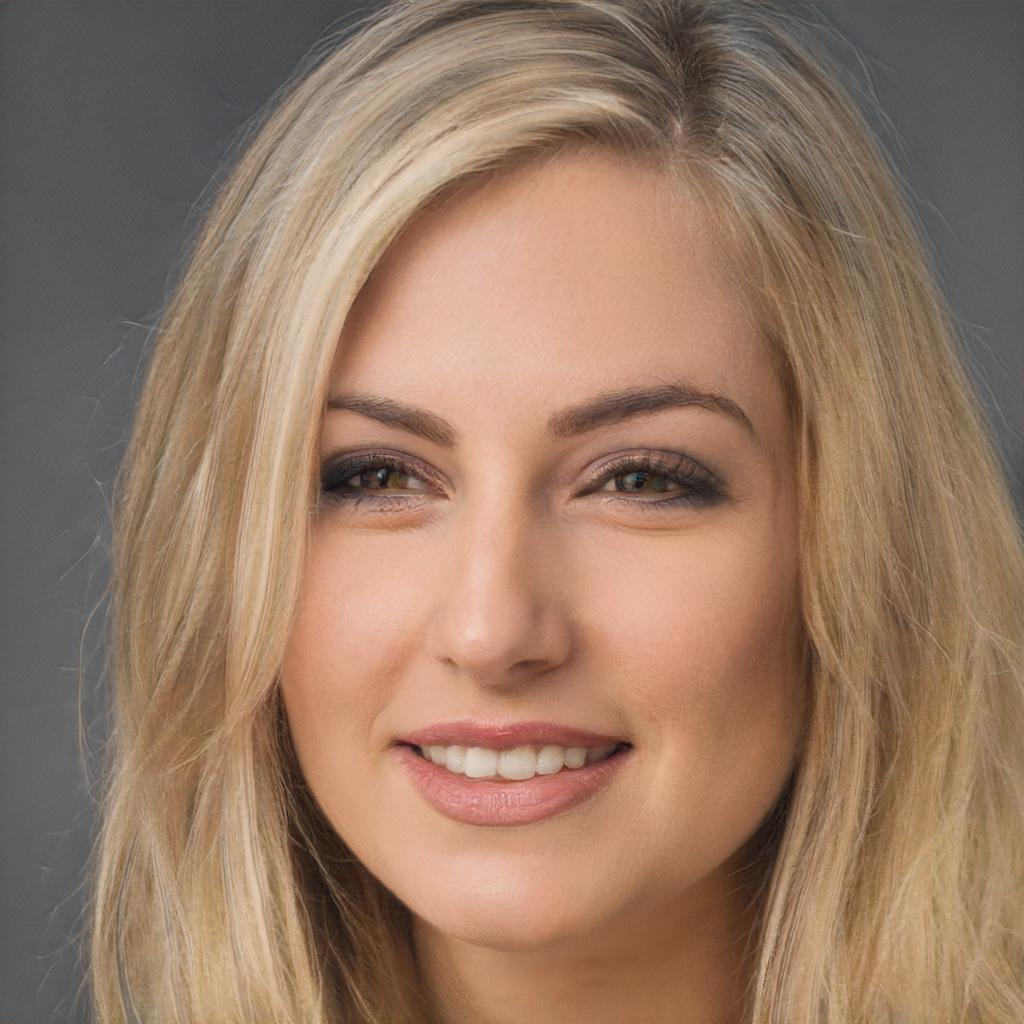} &
        \includegraphics[width=0.23\linewidth]
        {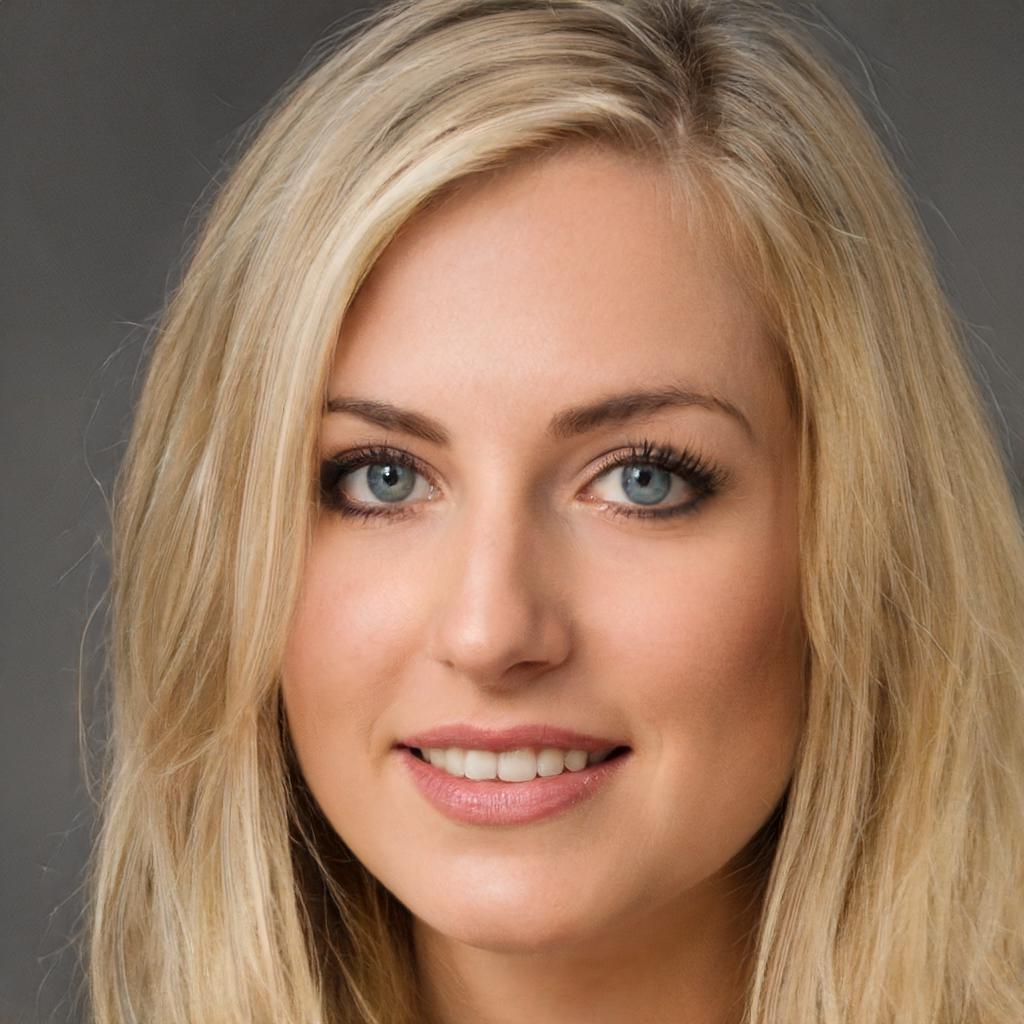} &
        \includegraphics[width=0.23\linewidth]
        {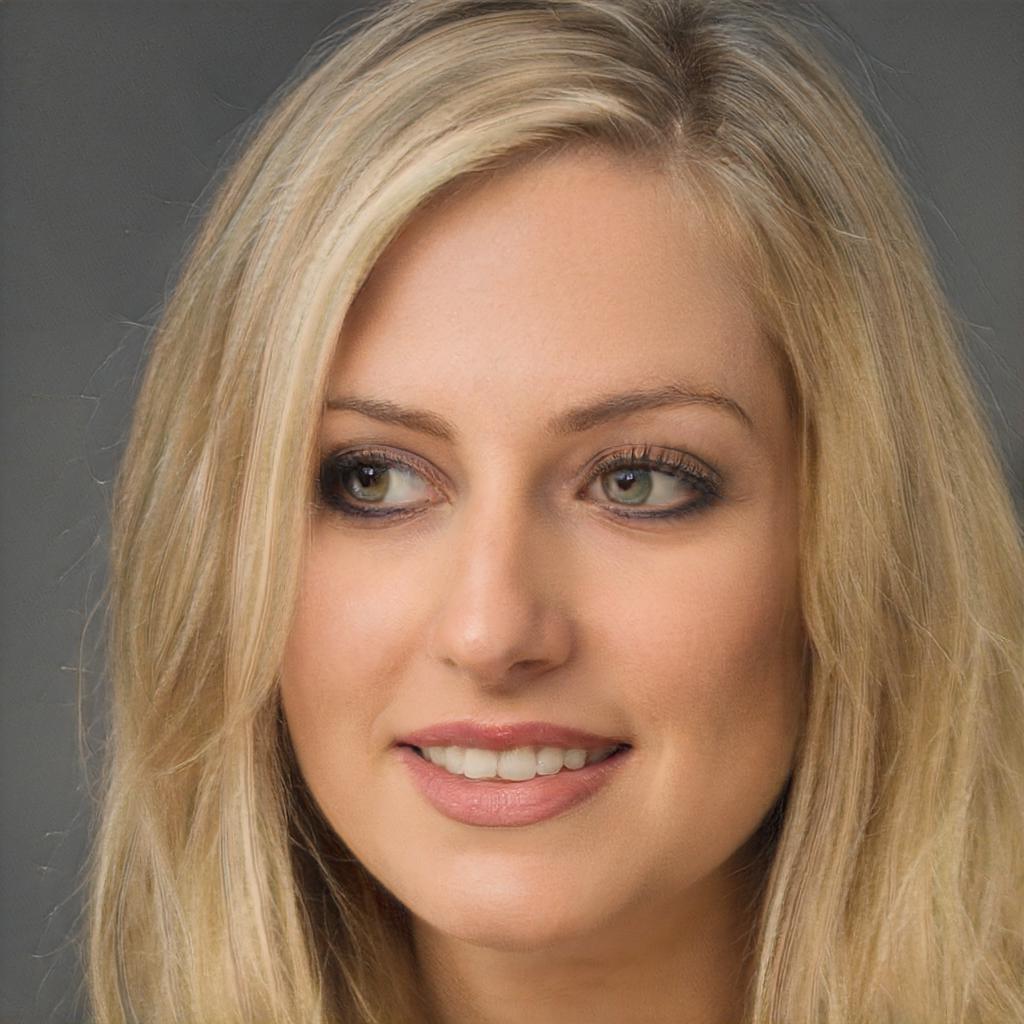} &
        \includegraphics[width=0.23\linewidth]
        {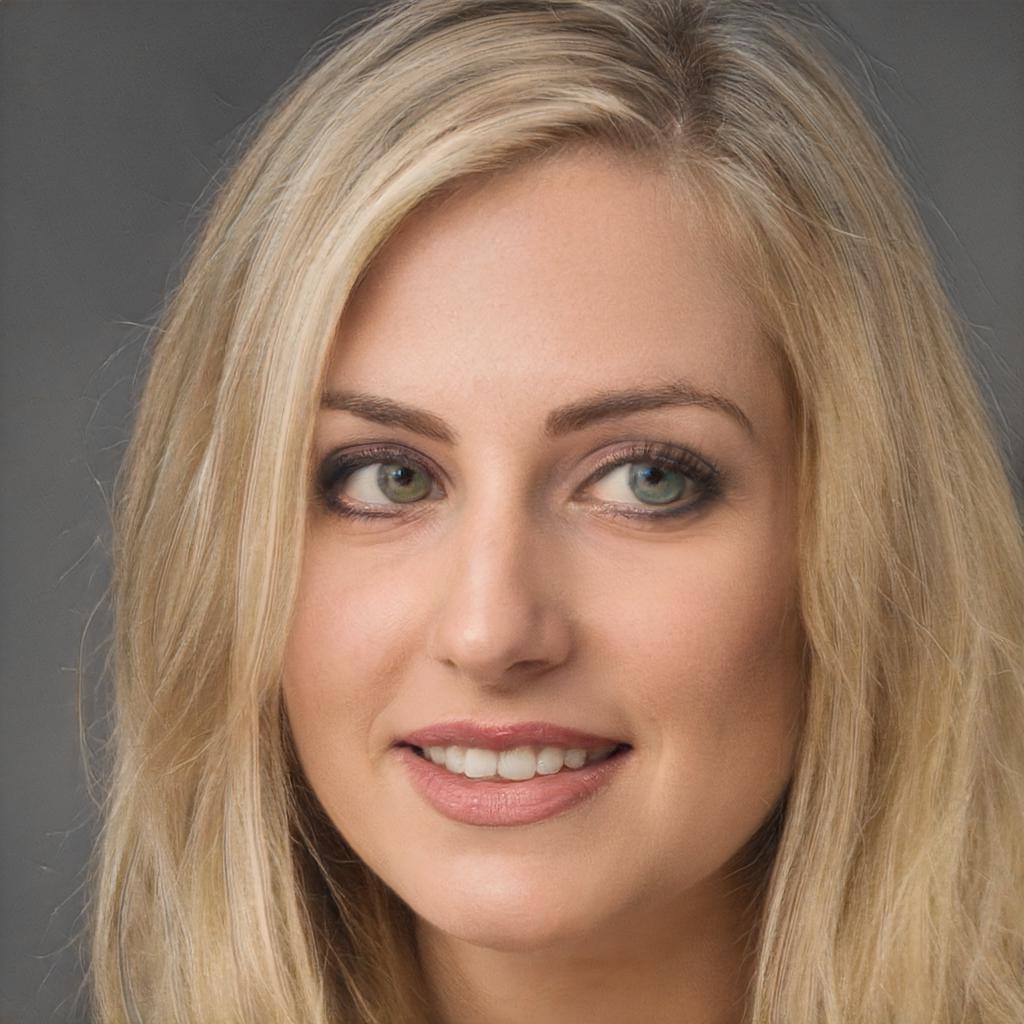}
        \tabularnewline
        \end{tabular}
    }
    \vspace{0.2cm}
    \caption{\textit{Local  semantic  control  on  the  human  facial  domain.} Additional results illustrating StyleFusion's ability to faithfully control specific facial attributes (both local and global) without altering other attributes.}
    \label{fig:local_control_faces2}
\end{figure*}

\begin{figure*}
    \centering
    \setlength{\tabcolsep}{0pt}
    {\small
        \begin{tabular}{c c c}
        \raisebox{0.65in}{\rotatebox[origin=t]{90}{Hair}} & 
        \hspace{1pt} &
        \includegraphics[width=0.94\linewidth]{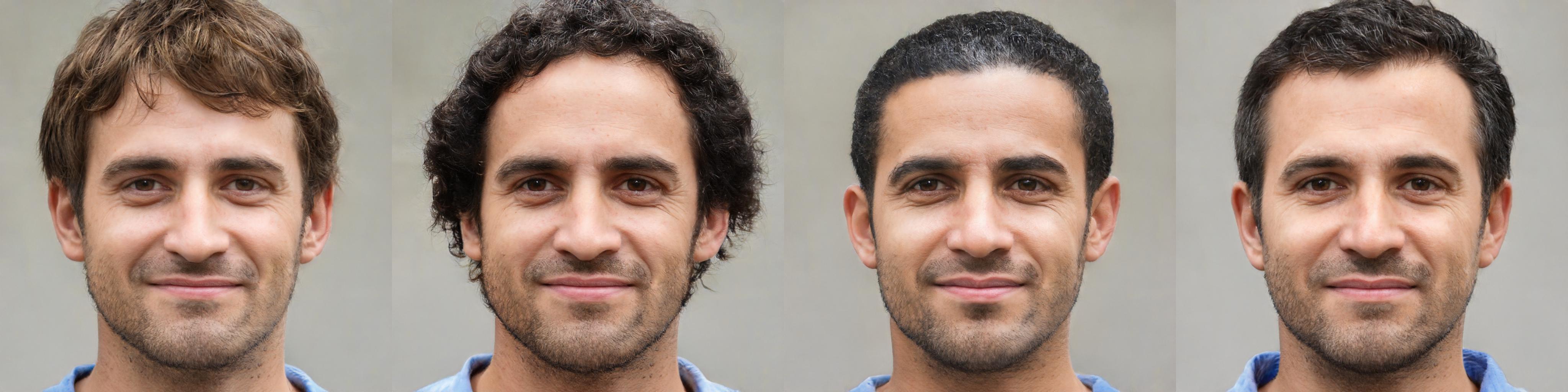} \\

        \raisebox{0.65in}{\rotatebox[origin=t]{90}{Face}} & 
        \hspace{1pt} &
        \includegraphics[width=0.94\linewidth]{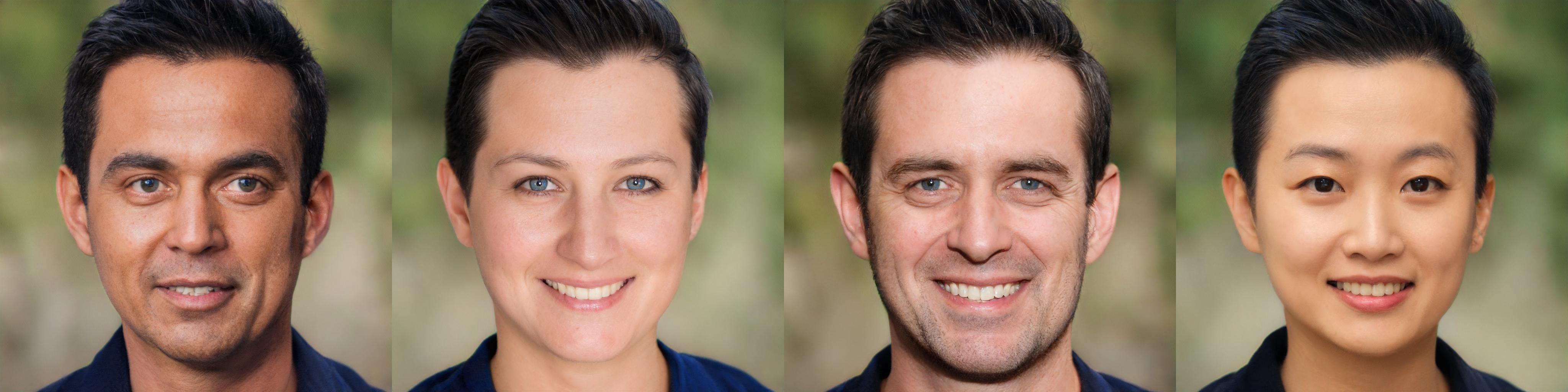} \\
        
        \raisebox{0.65in}{\rotatebox[origin=t]{90}{Mouth}} & 
        \hspace{1pt} &
        \includegraphics[width=0.94\linewidth]{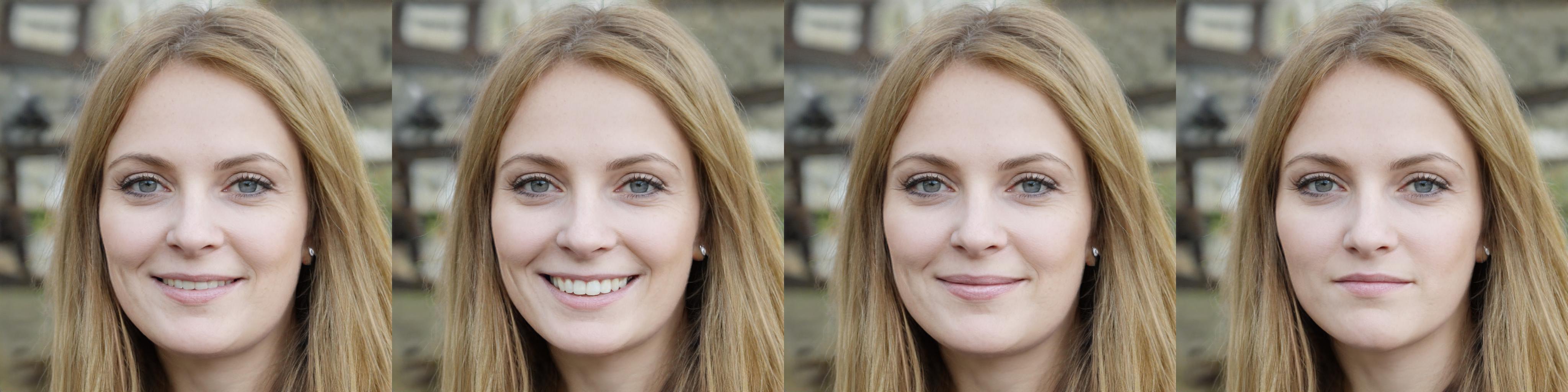} \\
        
        \raisebox{0.65in}{\rotatebox[origin=t]{90}{Eyes}} & 
        \hspace{1pt} &
        \includegraphics[width=0.94\linewidth]{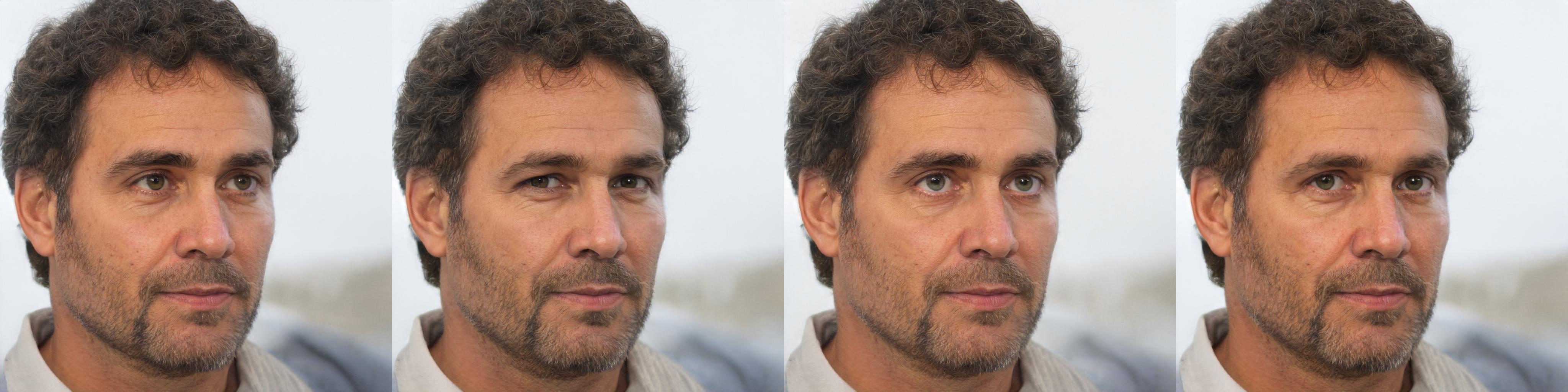} \\
        
        \end{tabular}
    }
    \vspace{0.2cm}
    \caption{\textit{Local  semantic  control  on  the  human  facial  domain.} As in Figure~\ref{fig:local_control_faces2}, we illustrate StyleFusion's ability to precisely control various facial attributes.}
    \label{fig:local_control_faces_3}
\end{figure*}

\begin{figure*}
    \centering
    \setlength{\tabcolsep}{0pt}
    {\small
        \begin{tabular}{c c c c c c}
        \raisebox{0.45in}{\rotatebox[origin=t]{90}{Upper BG}} & 
        \hspace{1pt} &
        \includegraphics[width=0.23\linewidth]
        {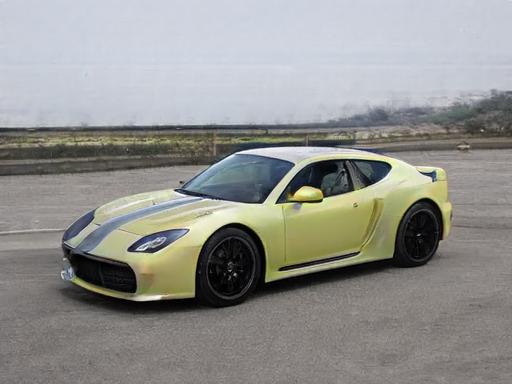} &
        \includegraphics[width=0.23\linewidth]
        {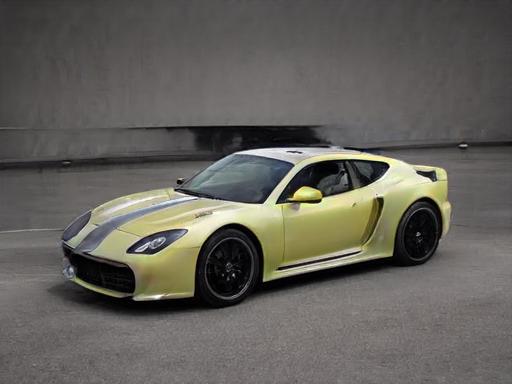} &
        \includegraphics[width=0.23\linewidth]
        {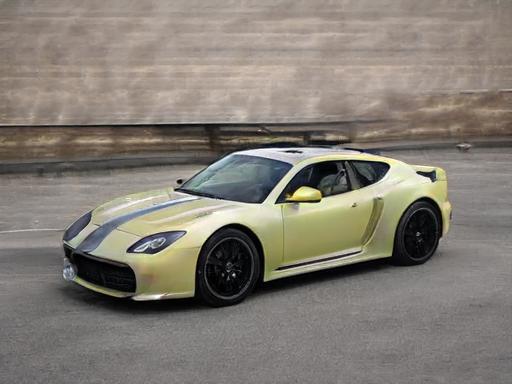} &
        \includegraphics[width=0.23\linewidth]
        {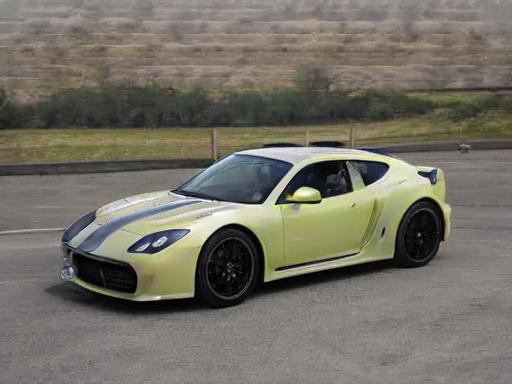}
        \tabularnewline
        \raisebox{0.45in}{\rotatebox[origin=t]{90}{Lower BG}}& 
        \hspace{1pt} &
        \includegraphics[width=0.23\linewidth]
        {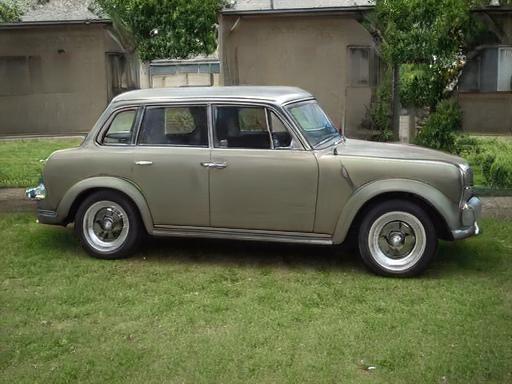} &
        \includegraphics[width=0.23\linewidth]
        {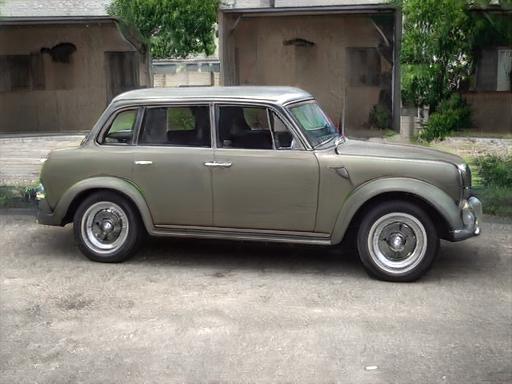} &
        \includegraphics[width=0.23\linewidth]
        {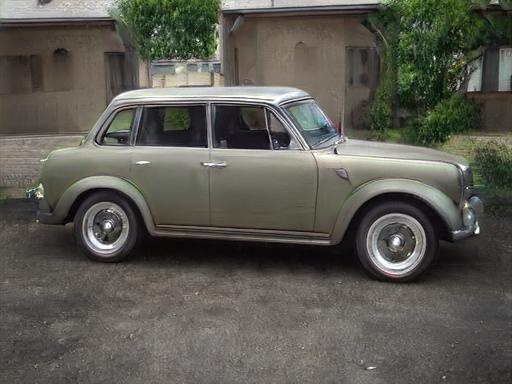} &
        \includegraphics[width=0.23\linewidth]
        {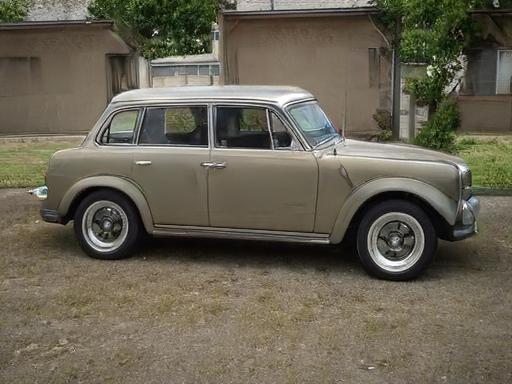}
        \tabularnewline
        \raisebox{0.45in}{\rotatebox[origin=t]{90}{Body}}& 
        \hspace{1pt} &
        \includegraphics[width=0.23\linewidth]
        {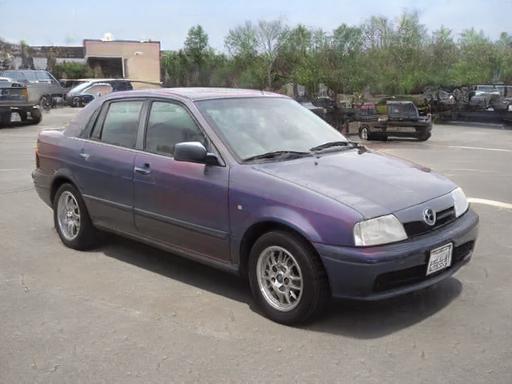} &
        \includegraphics[width=0.23\linewidth]
        {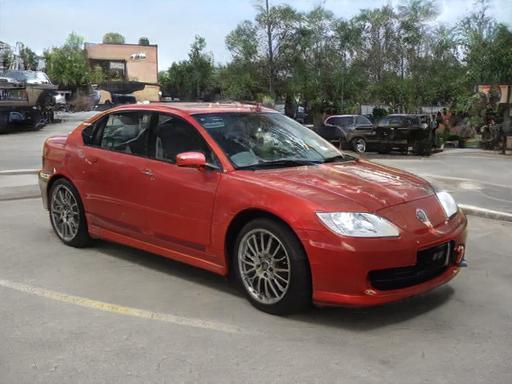} &
        \includegraphics[width=0.23\linewidth]
        {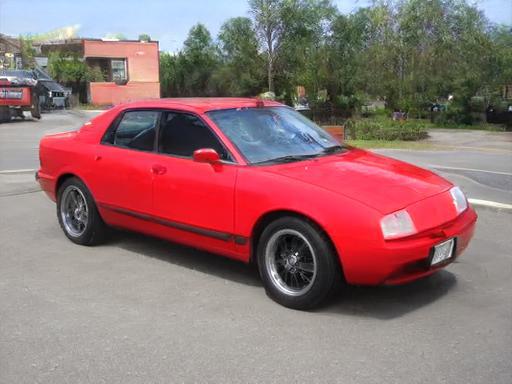} &
        \includegraphics[width=0.23\linewidth]
        {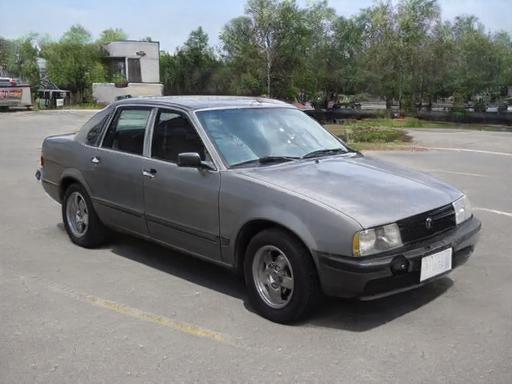}
        \tabularnewline
        \raisebox{0.45in}{\rotatebox[origin=t]{90}{Wheels}}& 
        \hspace{1pt} &
        \includegraphics[width=0.23\linewidth]
        {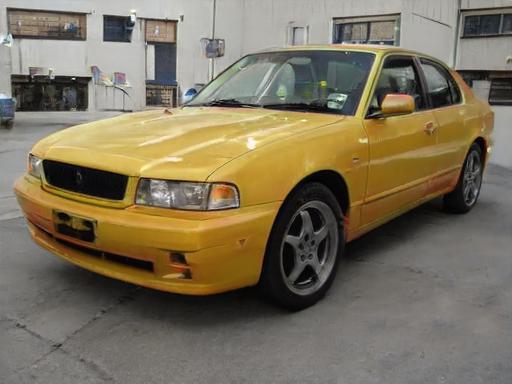} &
        \includegraphics[width=0.23\linewidth]
        {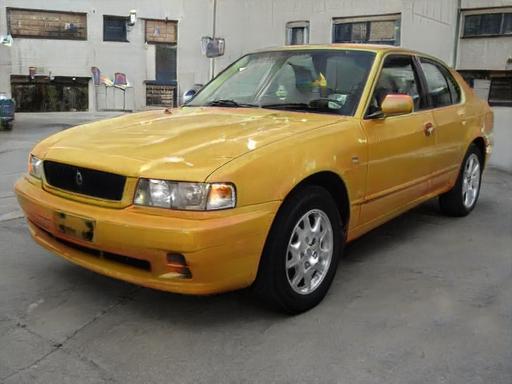} &
        \includegraphics[width=0.23\linewidth]
        {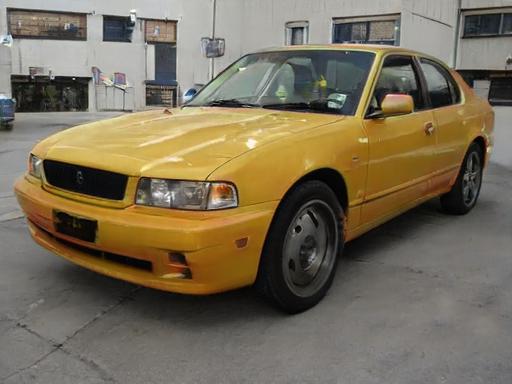} &
        \includegraphics[width=0.23\linewidth]
        {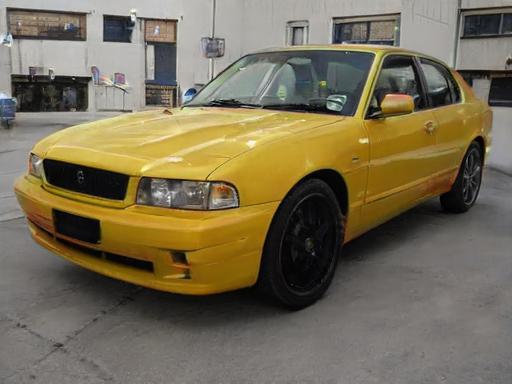}
        \end{tabular}
    }
    \vspace{0.2cm}
    \caption{\textit{Local  semantic  control  on  the  cars domain.} We demonstrate StyleFusion's ability to precisely control various local (e.g., wheels) and global (e.g., background) image attributes over the cars domain.}
    \label{fig:local_control_cars2}
\end{figure*}

\begin{figure*}
    \centering
    \includegraphics[width=\linewidth]{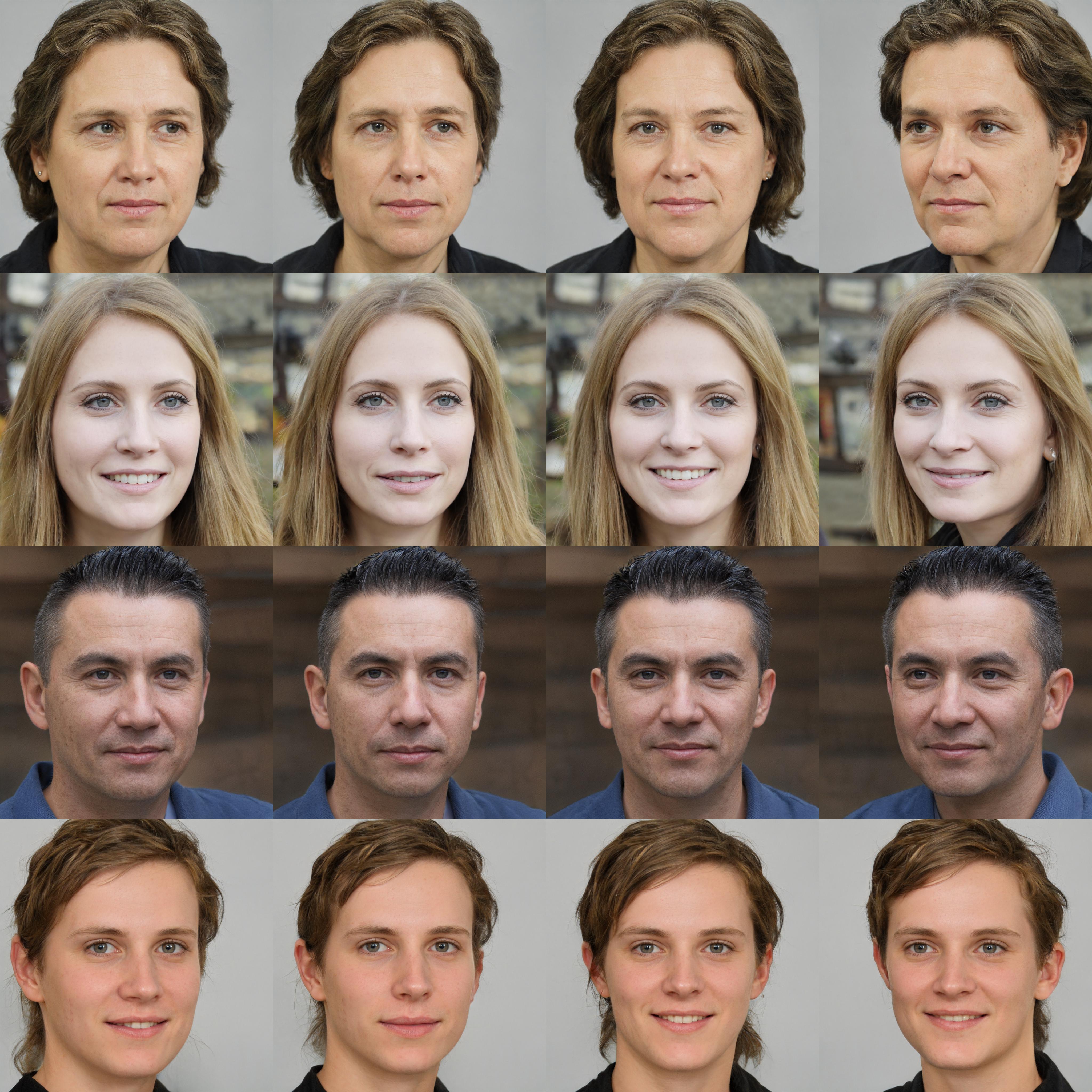}
    \caption{\textit{Global control using StyleFusion.} 
    We demonstrate StyleFusion's control over the global pose of a facial image by altering the global alignment style code. Observe how the identity and facial characteristics are well-preserved across the different poses.}
    \label{fig:appendix_ffhq_all_1}
\end{figure*}

\begin{figure*}
    \centering
    \includegraphics[width=\linewidth]{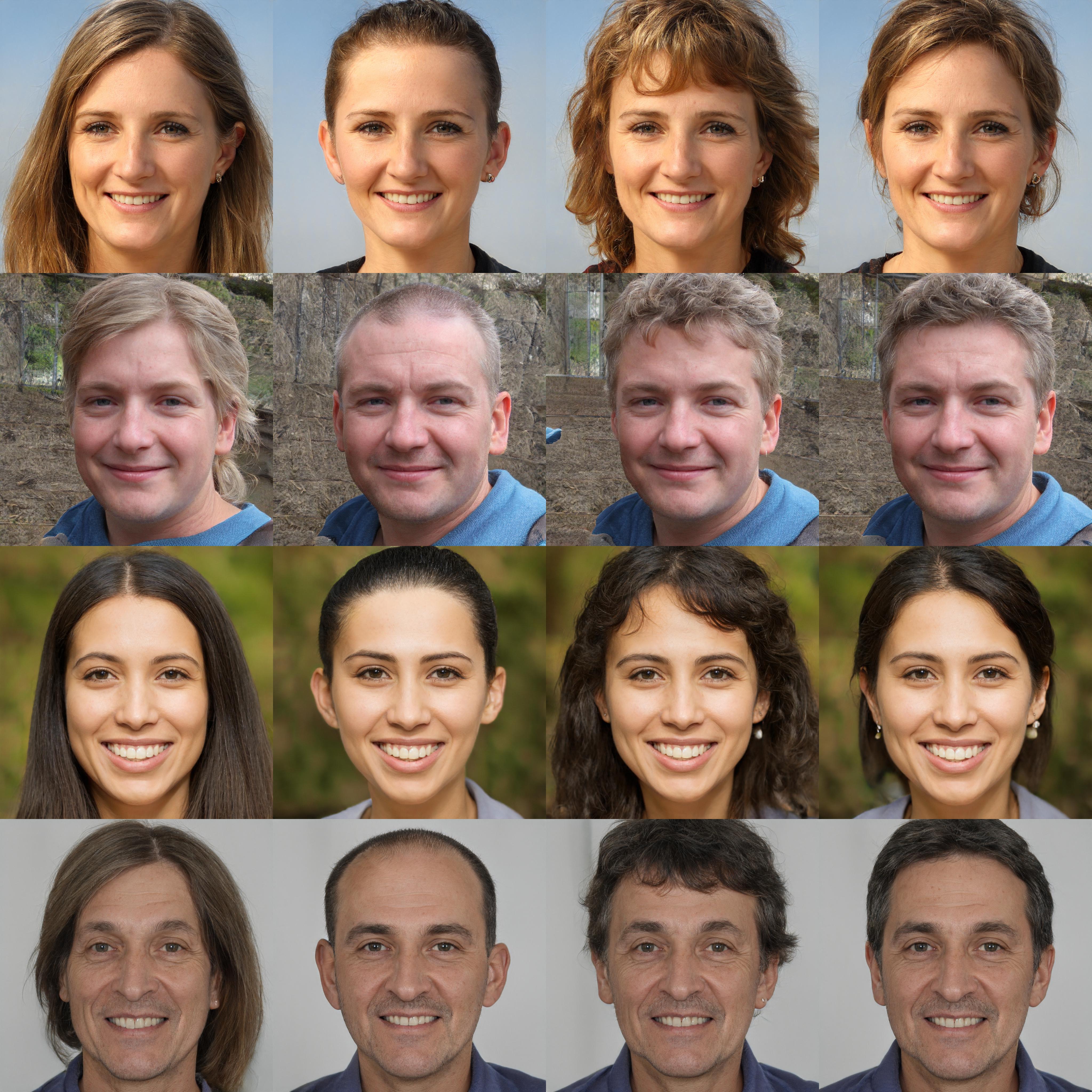}
    \caption{\textit{Global control using StyleFusion.} 
    As in Figure~\ref{fig:appendix_ffhq_all_1}, we demonstrate StyleFusion's control over the global hairstyle of a facial image.}
    \label{fig:appendix_ffhq_common_bg_hair}
\end{figure*}

\begin{figure*}
    \centering
    \includegraphics[width=\linewidth]{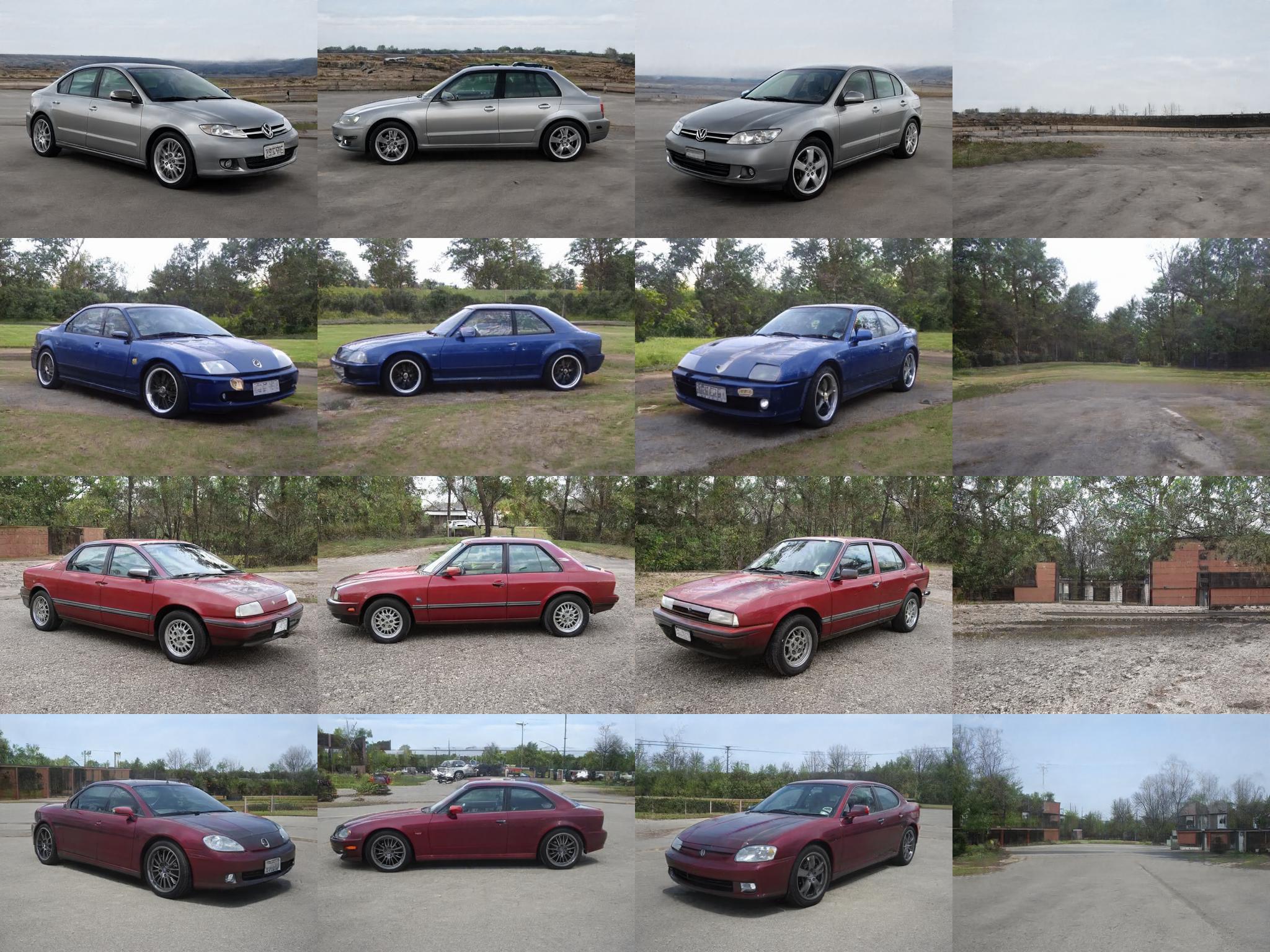}
    \caption{\textit{Global control using StyleFusion.} By applying different alignment latent codes, StyleFusion allows for global control over the car pose. Observe also that such control allows one to add or remove the car from a given image.}
    \label{fig:appendix_car_all_1}
\end{figure*}

\begin{figure*}
    \centering
    \includegraphics[width=\linewidth]{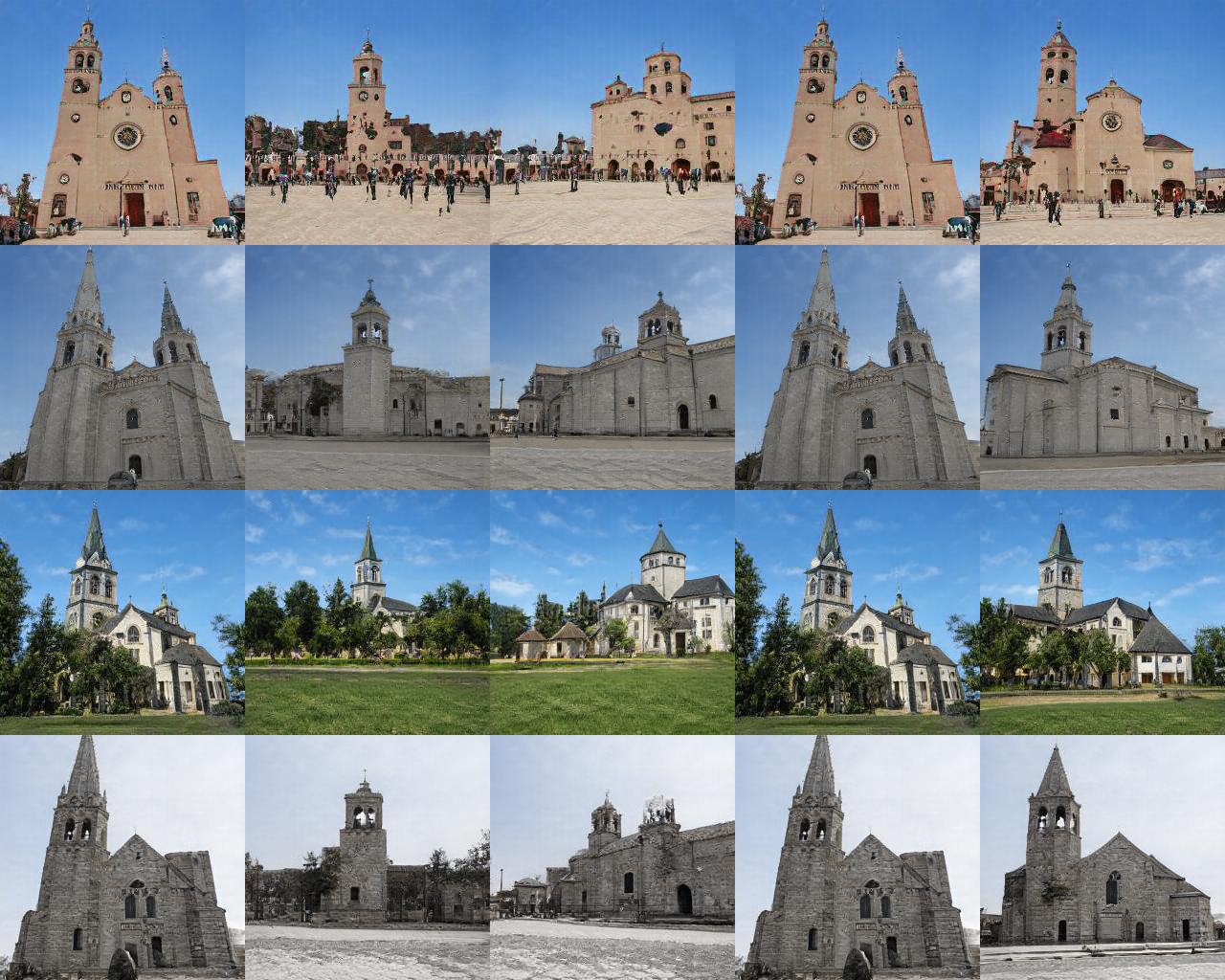}
    \caption{\textit{Global control using StyleFusion.} Altering the global latent code allows users to generate multiple church buildings of the same style, but of different global structure. Notably, observe the similarity in the church layout for each of the columns shown.}
    \label{fig:appendix_church_common_1}
\end{figure*}

\begin{figure*}
    \centering
    \includegraphics[width=\linewidth]{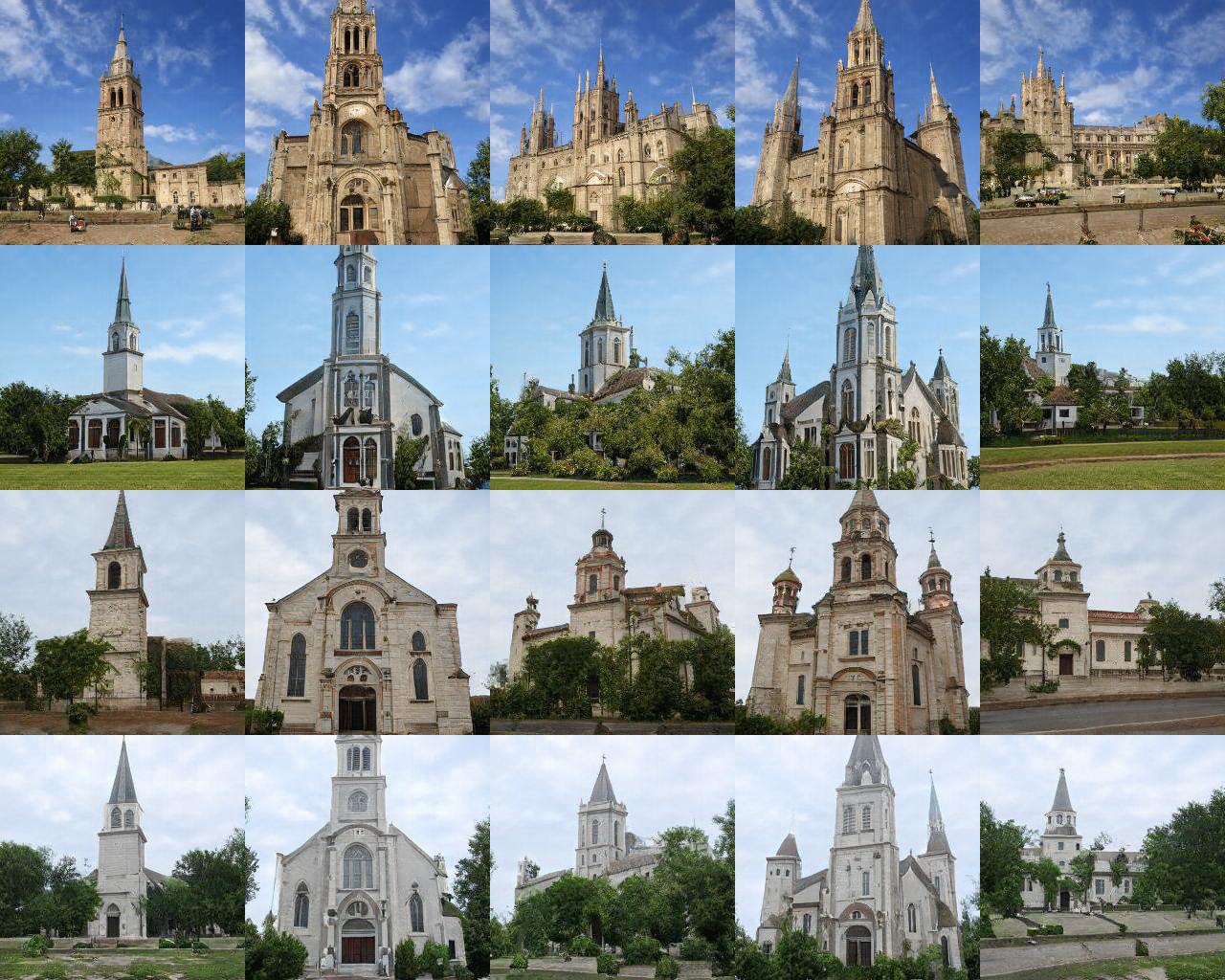}
    \caption{\textit{Global control using StyleFusion.} As in Figure~\ref{fig:appendix_church_common_1}, StyleFusion provides users with the ability to generate multiple church buildings of the same style, but of different structure.}
    \label{fig:appendix_church_common_2}
\end{figure*}

\begin{figure*}
    \centering
    \includegraphics[width=\linewidth]{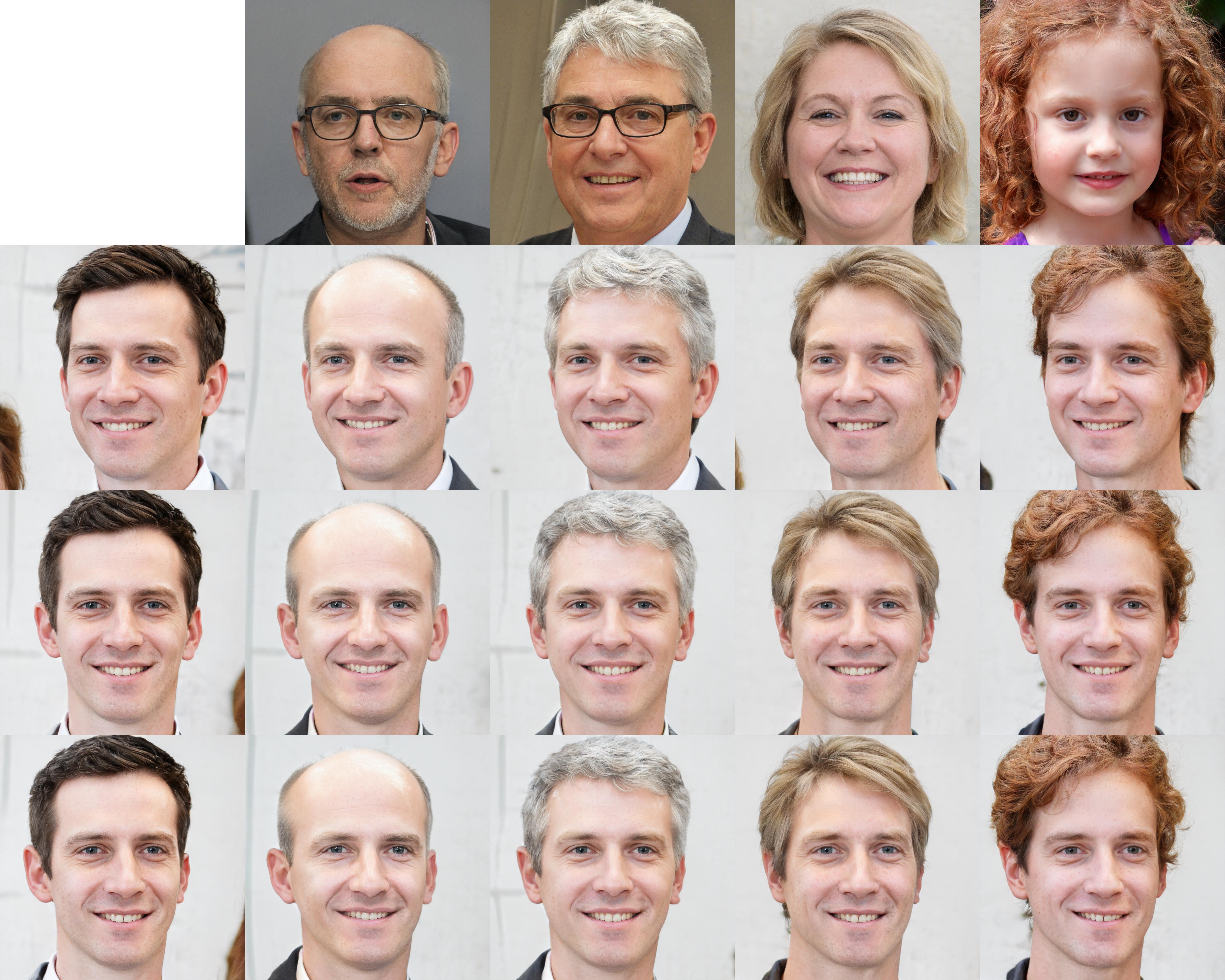}
    \caption{\textit{Hair transfer using StyleFusion.} 
    Given the same target individual (shown to the left) at different poses, StyleFusion is able to transfer the hairstyle from different reference images. Observe that the hair transfer provides realistic looking images that preserve the original target identity even when the two images are unaligned. 
    }
    \label{fig:appendix_cross_pose_hair1}
\end{figure*}

\begin{figure*}
    \centering
    \includegraphics[width=\linewidth]{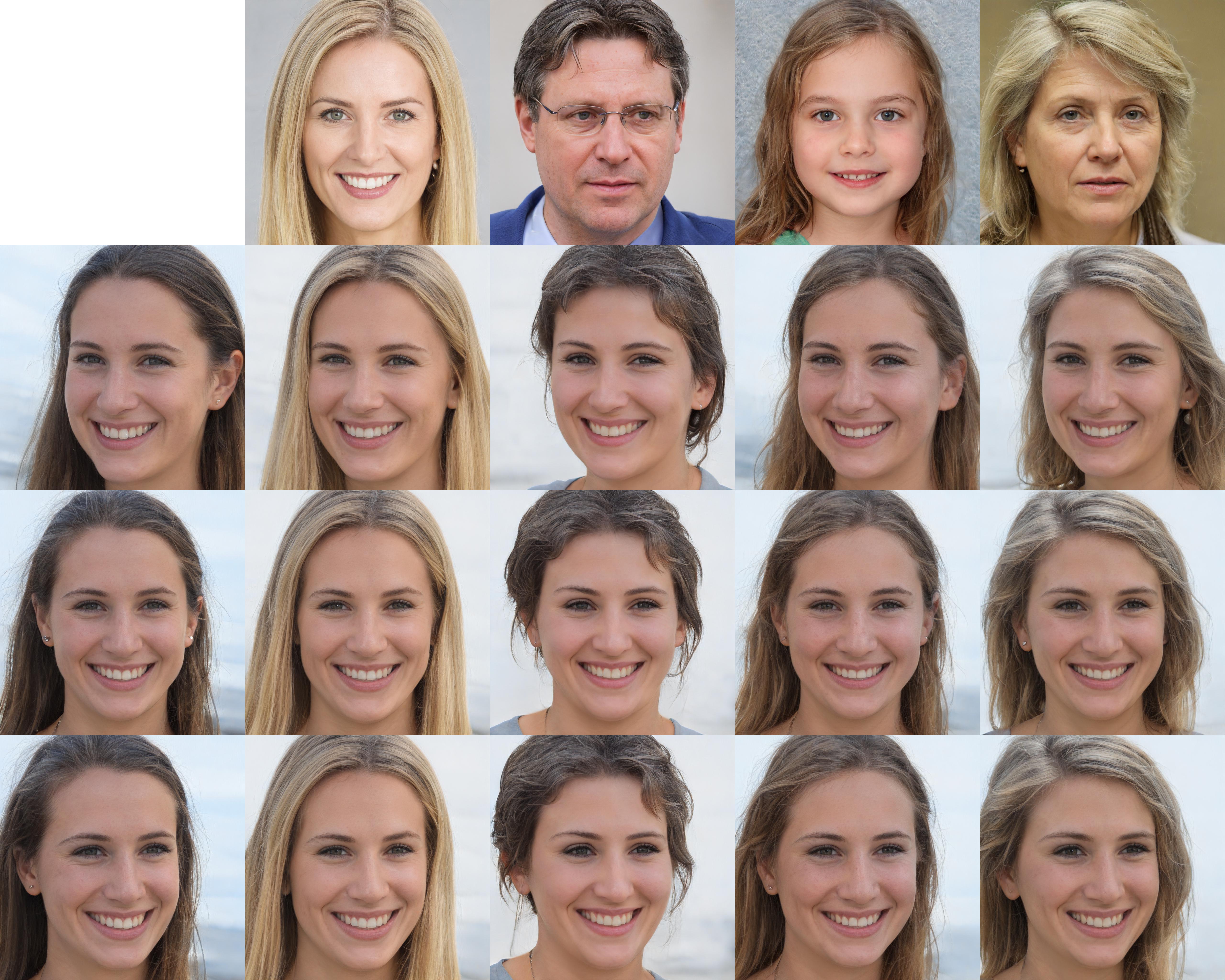}
    \caption{\textit{Hair transfer using StyleFusion.} As in Figure~\ref{fig:appendix_cross_pose_hair1}, we demonstrate StyleFusion's ability to perform cross-image hair transfer. Observe how the hairstyle remains consistent along each column while the pose remains consistent along each row.}
    \label{fig:appendix_cross_pose_hair2}
\end{figure*}

\begin{figure*}
    \setlength{\tabcolsep}{1.5pt}
    \centering
    \small{
        \begin{tabular}{c c c c c c c}
        
        &&& StyleCLIP & InterFace & GANSpace & GANSpace \\
        
        \includegraphics[width=0.1375\linewidth]{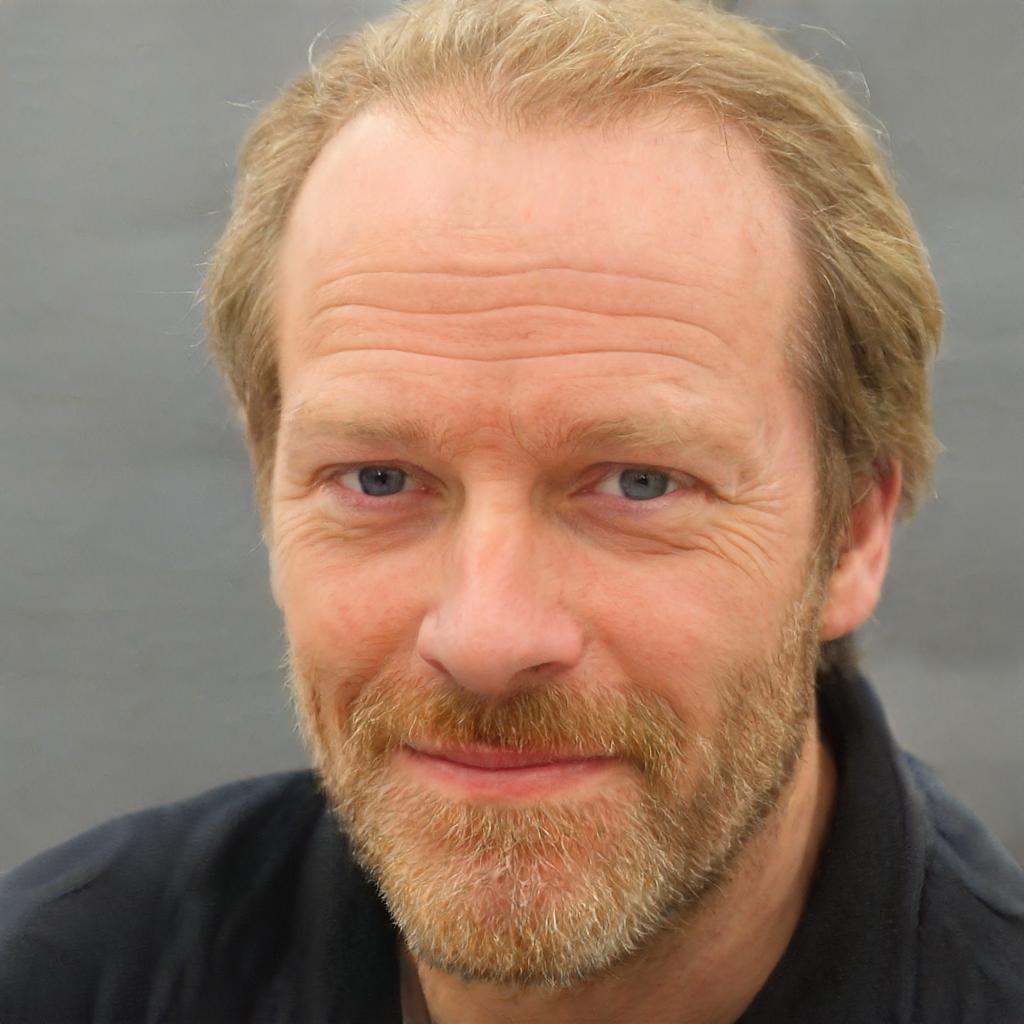} &&
        \raisebox{0.35in}{\rotatebox[origin=t]{90}{ StyleGAN2}} &
        \includegraphics[width=0.1375\linewidth]{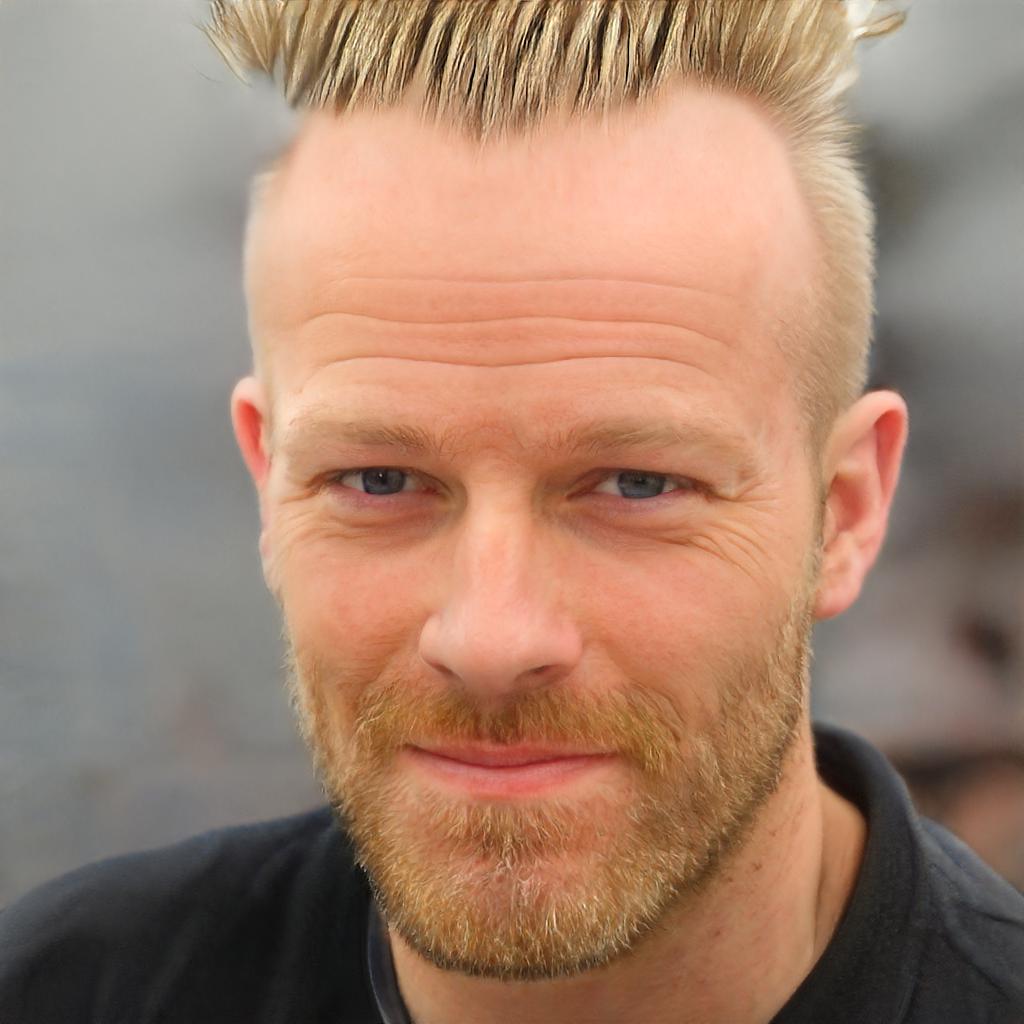} &
        \includegraphics[width=0.1375\linewidth]{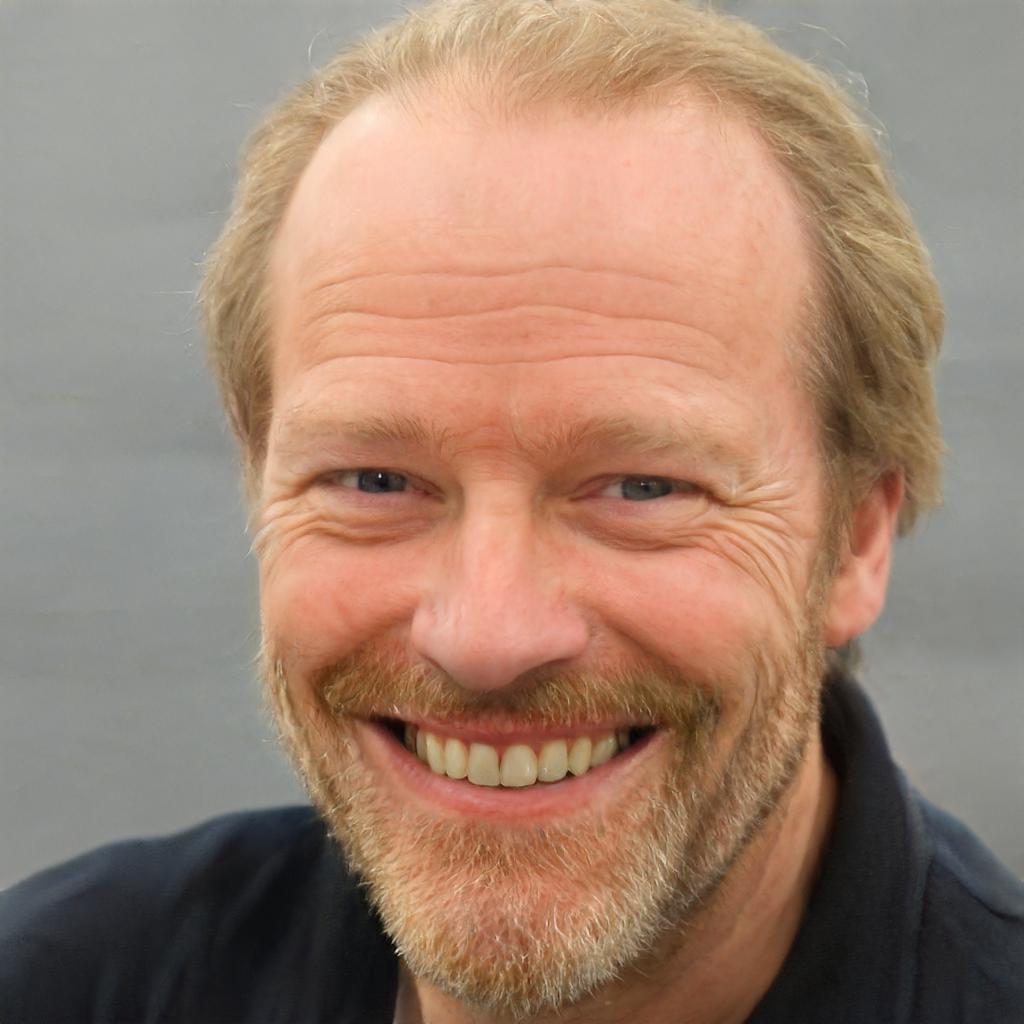} &
        \includegraphics[width=0.1375\linewidth]{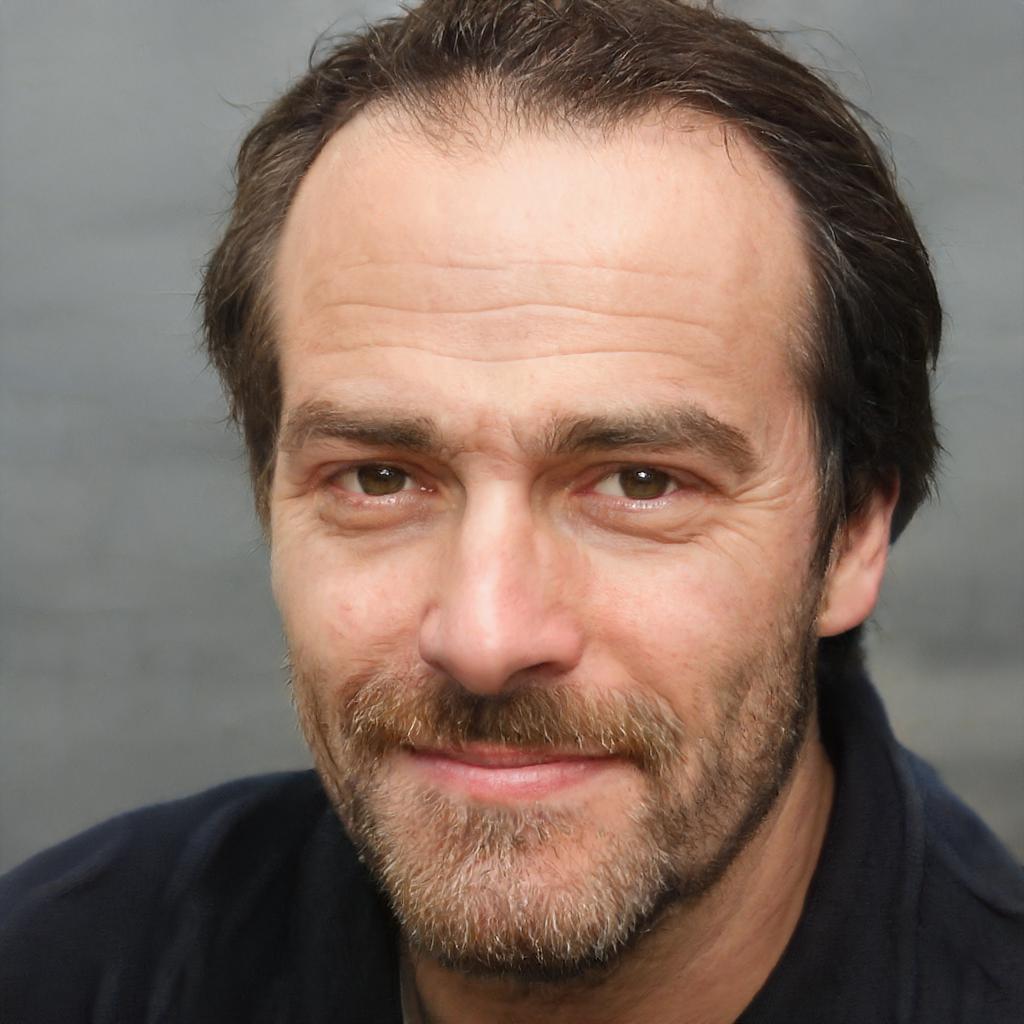} &
        \includegraphics[width=0.1375\linewidth]{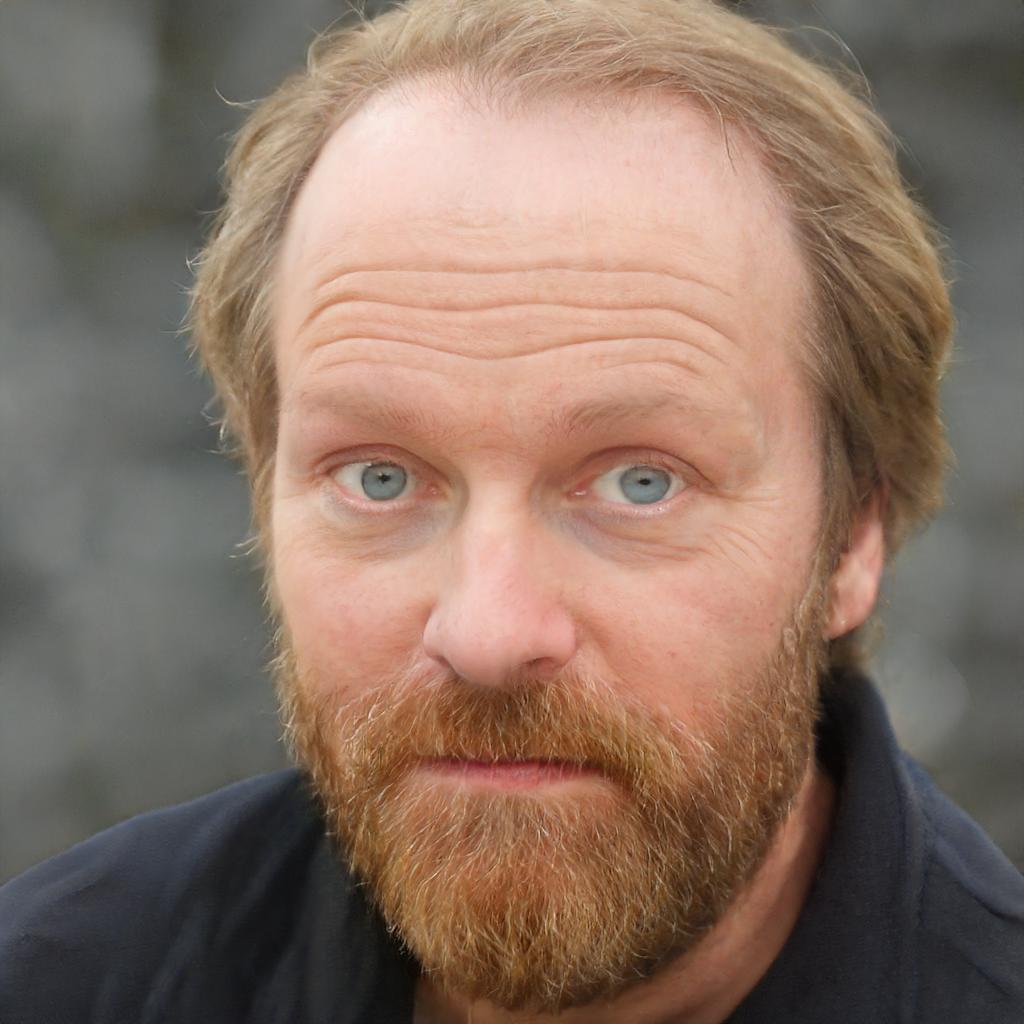} 
        \tabularnewline
        && \raisebox{0.35in}{\rotatebox[origin=t]{90}{ StyleFusion}} &
        \includegraphics[width=0.1375\linewidth]{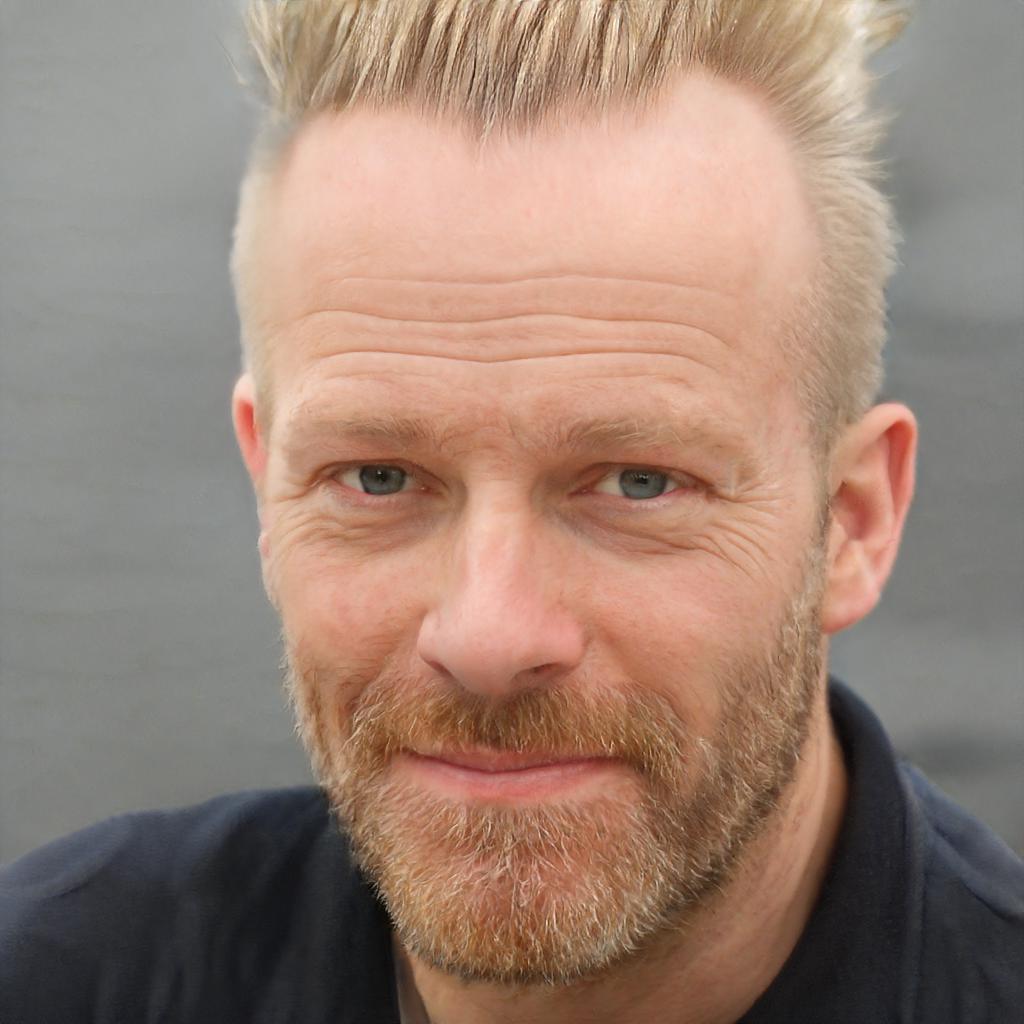} &
        \includegraphics[width=0.1375\linewidth]{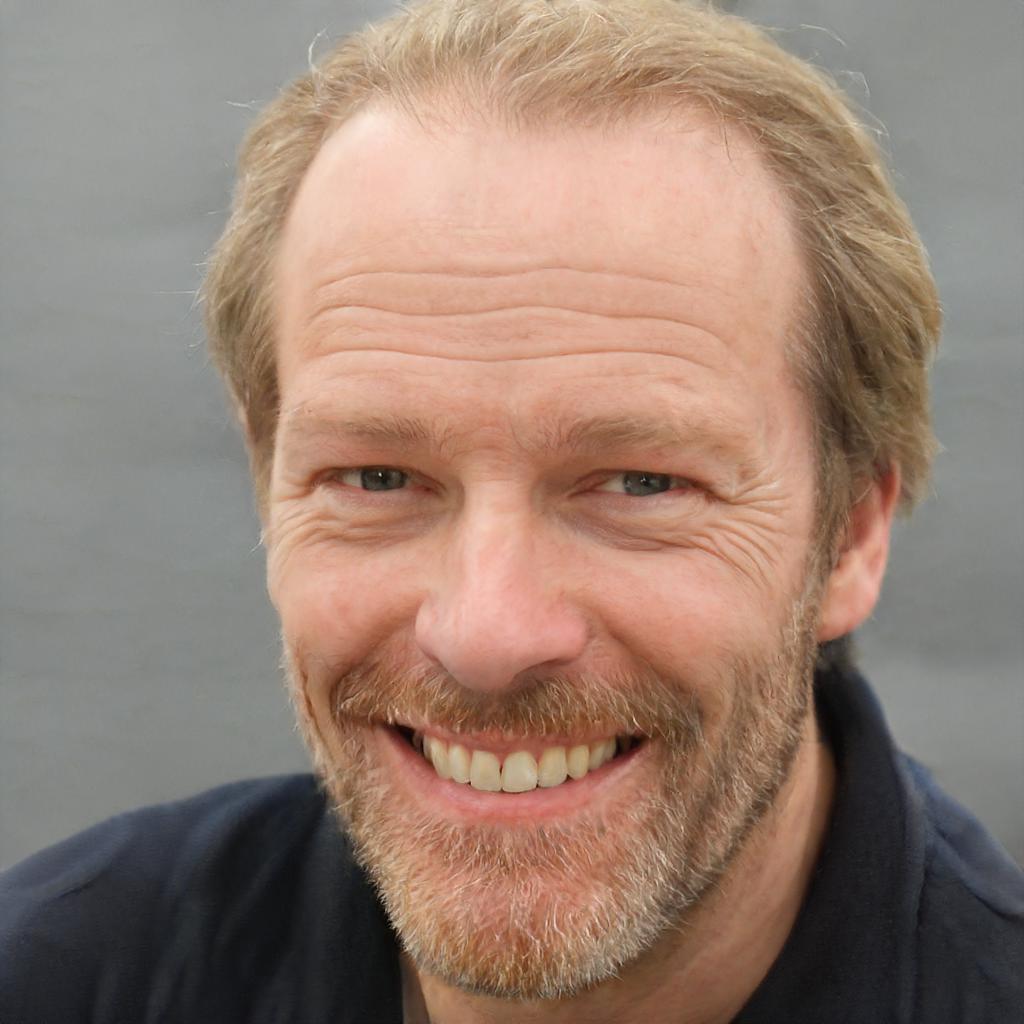} &
        \includegraphics[width=0.1375\linewidth]{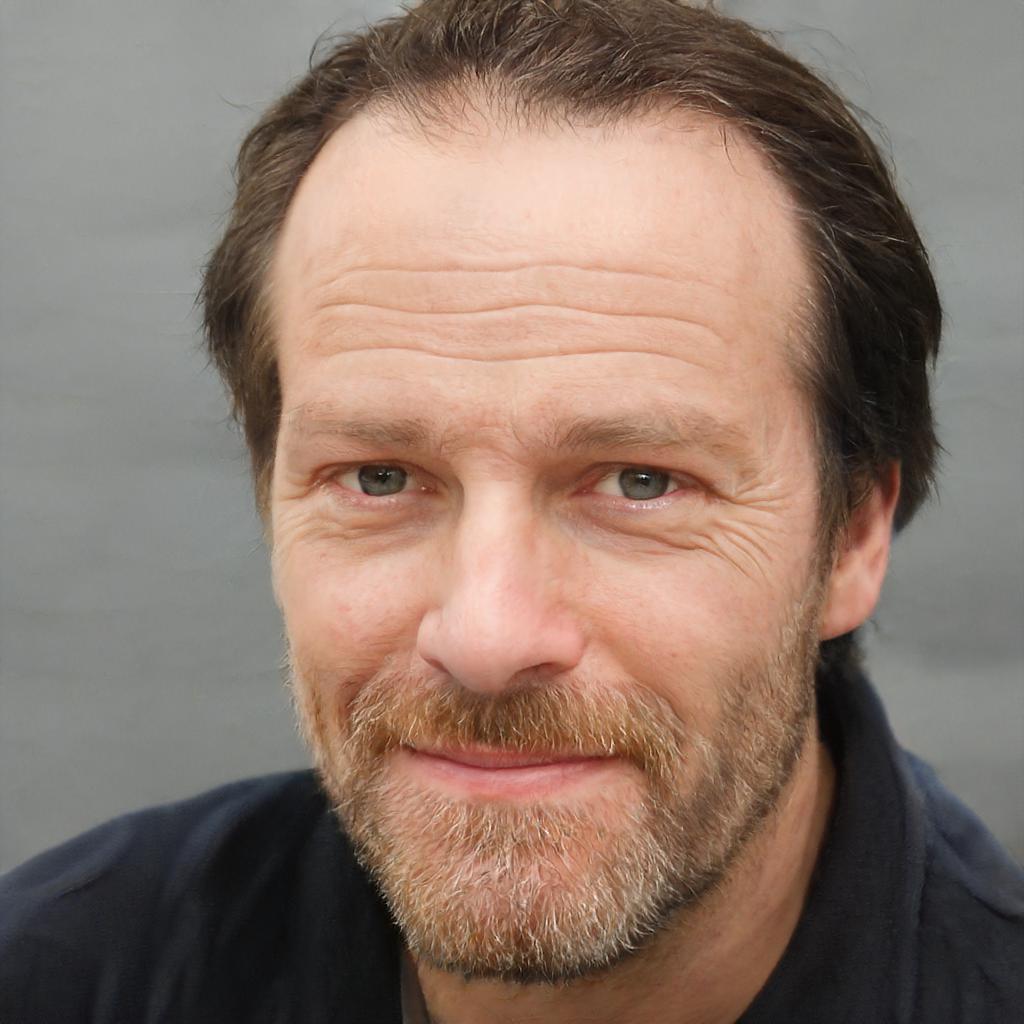} &
        \includegraphics[width=0.1375\linewidth]{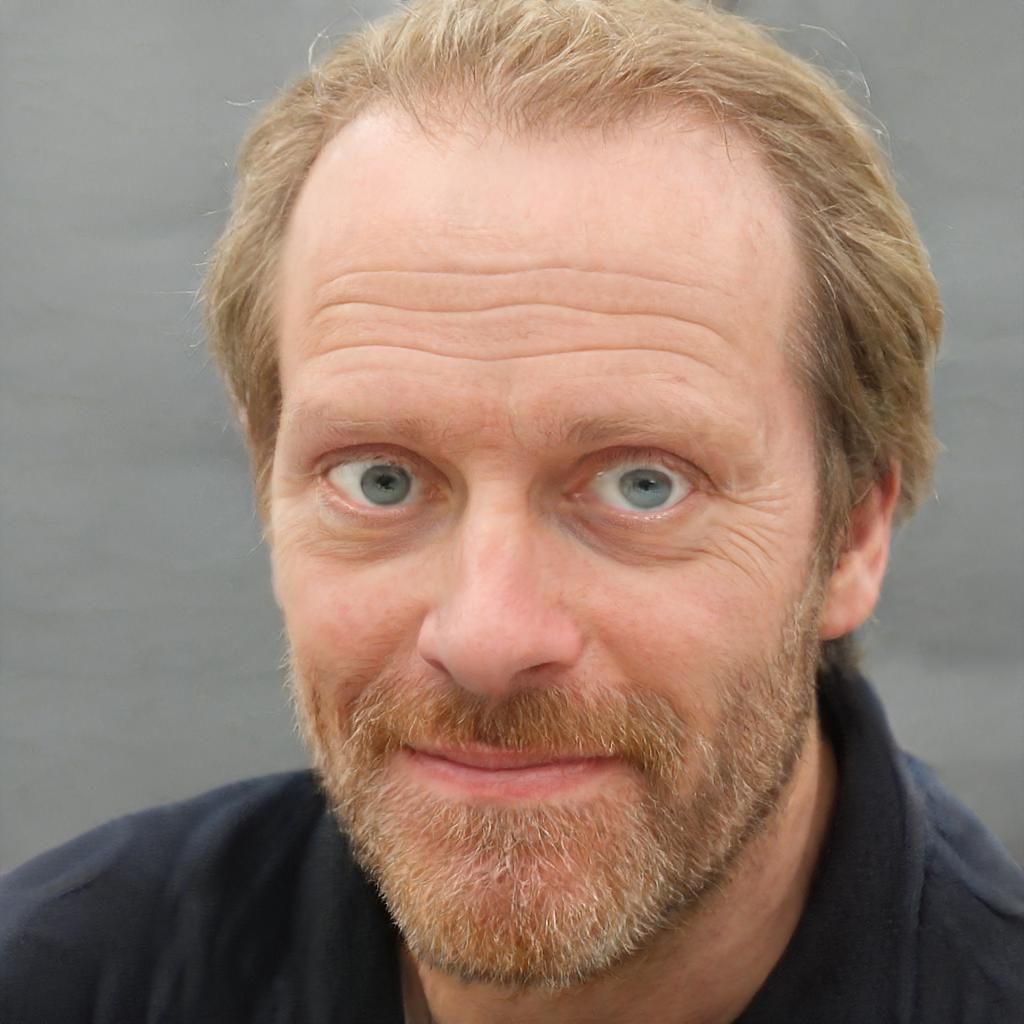} 
        \tabularnewline

        \includegraphics[width=0.1375\linewidth]{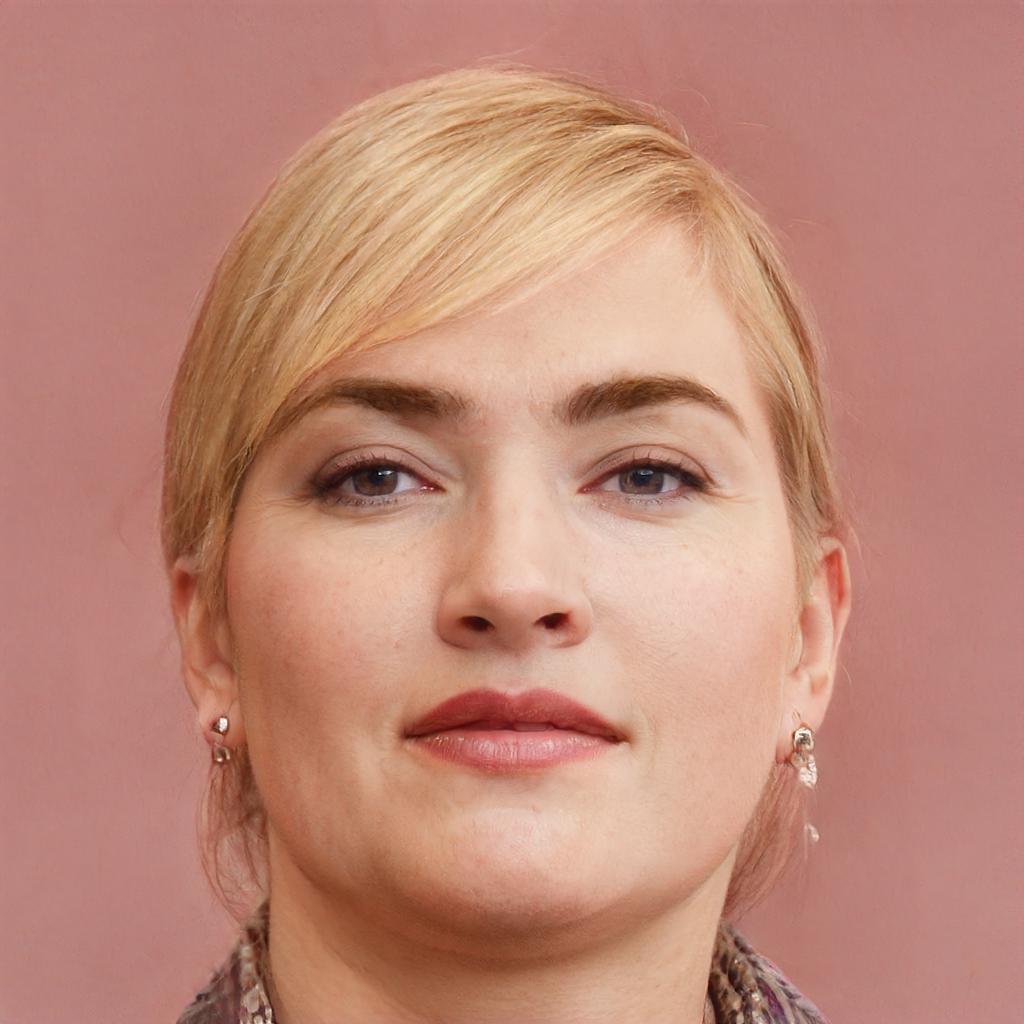} &&
        \raisebox{0.35in}{\rotatebox[origin=t]{90}{ StyleGAN2}} &
        \includegraphics[width=0.1375\linewidth]{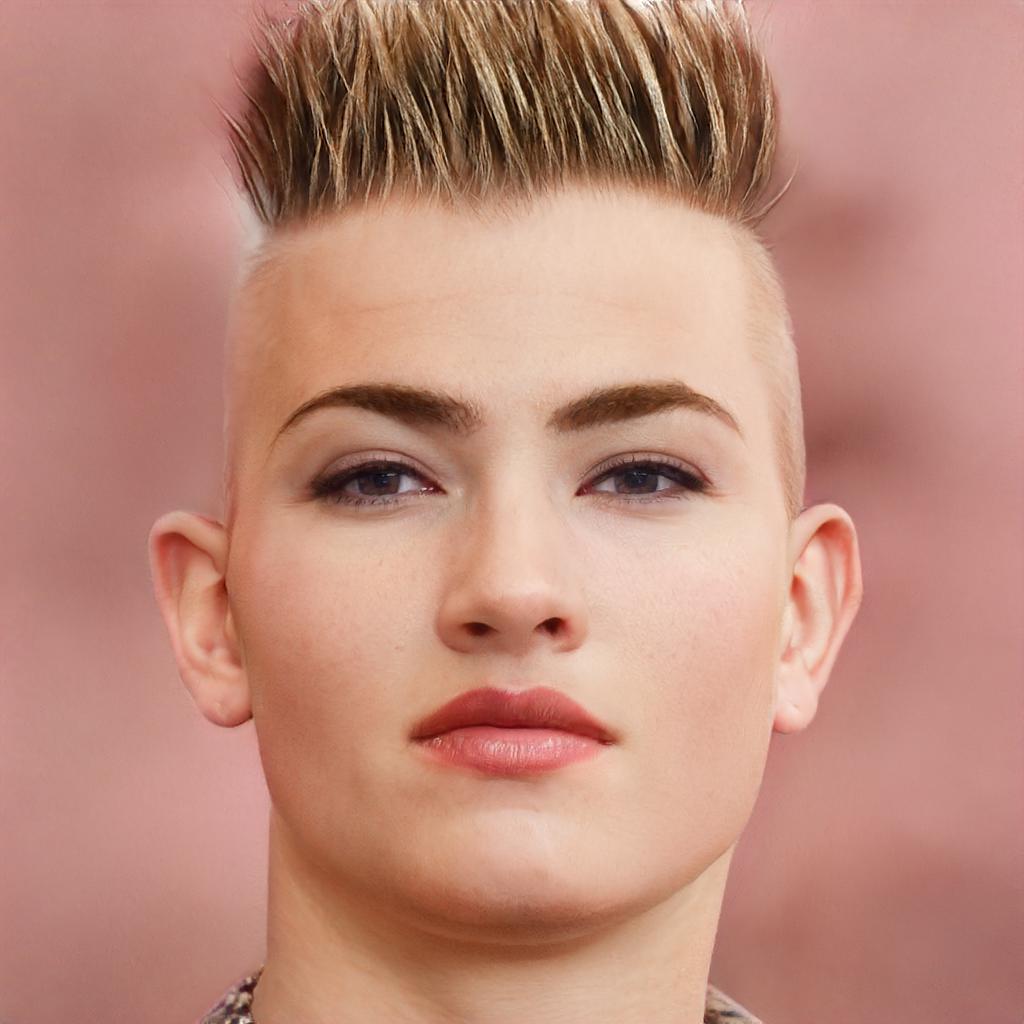} &
        \includegraphics[width=0.1375\linewidth]{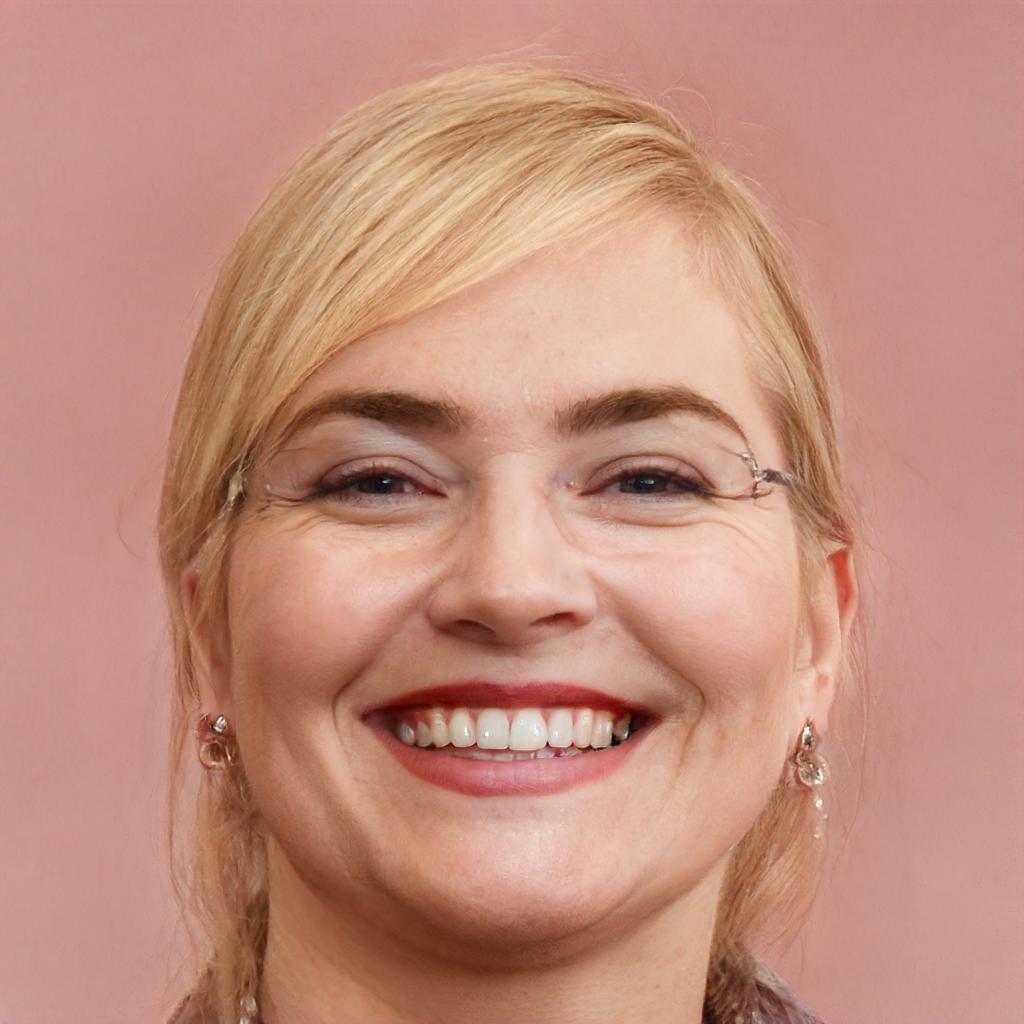} &
        \includegraphics[width=0.1375\linewidth]{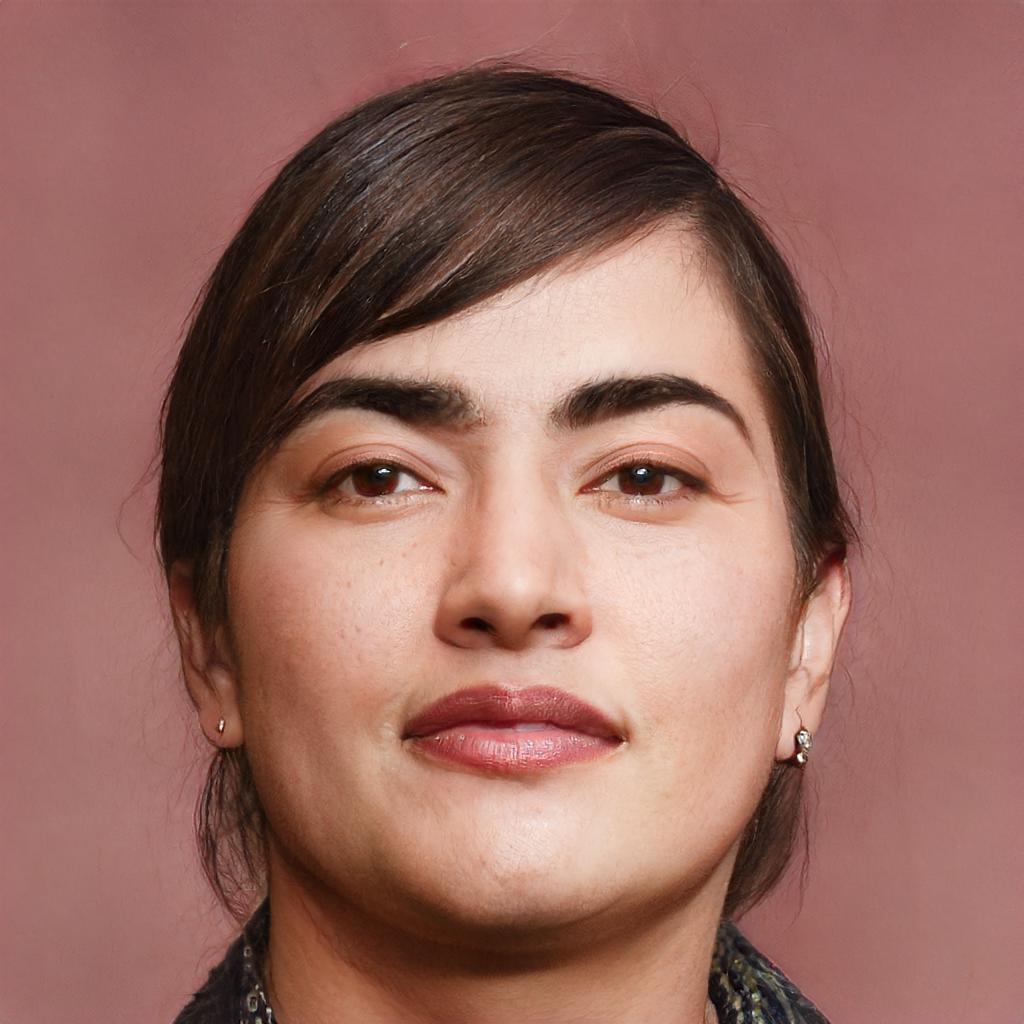} &
        \includegraphics[width=0.1375\linewidth]{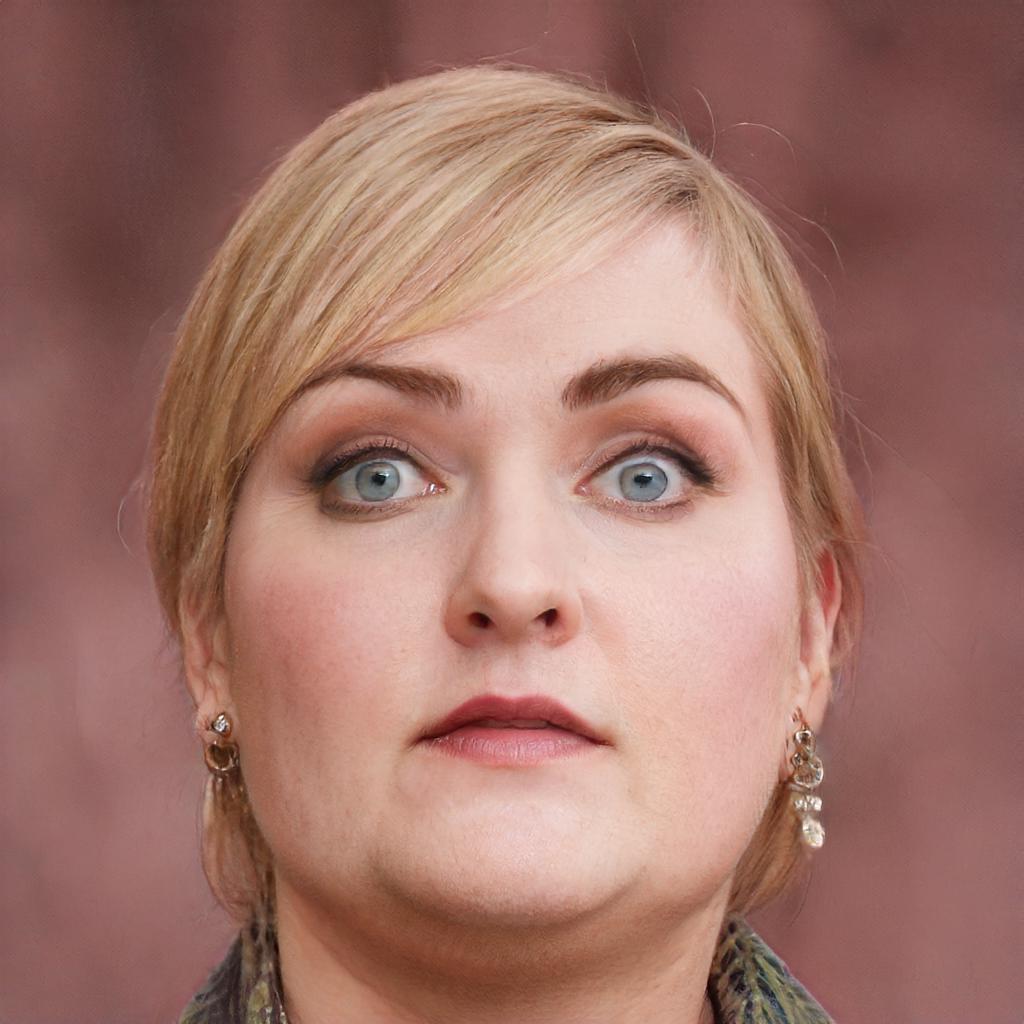}
        \tabularnewline
        && \raisebox{0.35in}{\rotatebox[origin=t]{90}{ StyleFusion}} &
        \includegraphics[width=0.1375\linewidth]{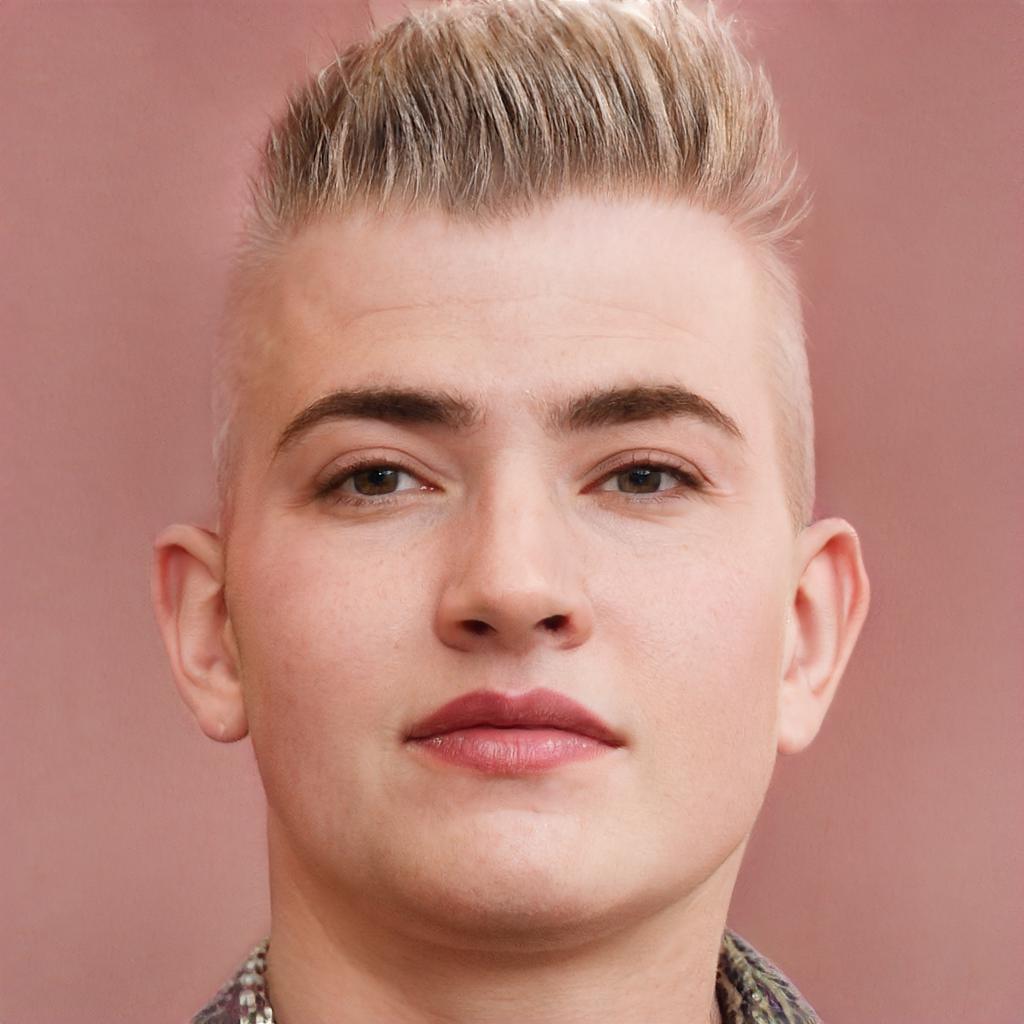} &
        \includegraphics[width=0.1375\linewidth]{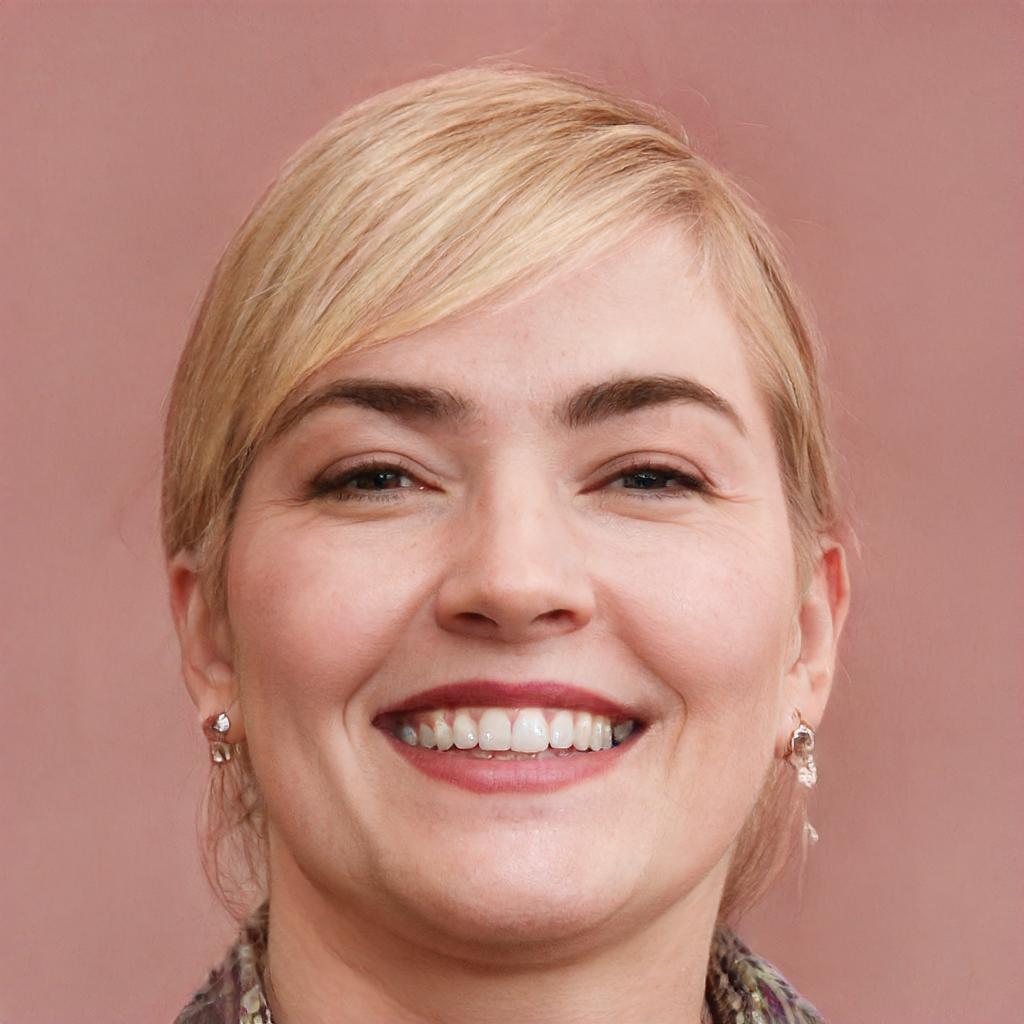} &
        \includegraphics[width=0.1375\linewidth]{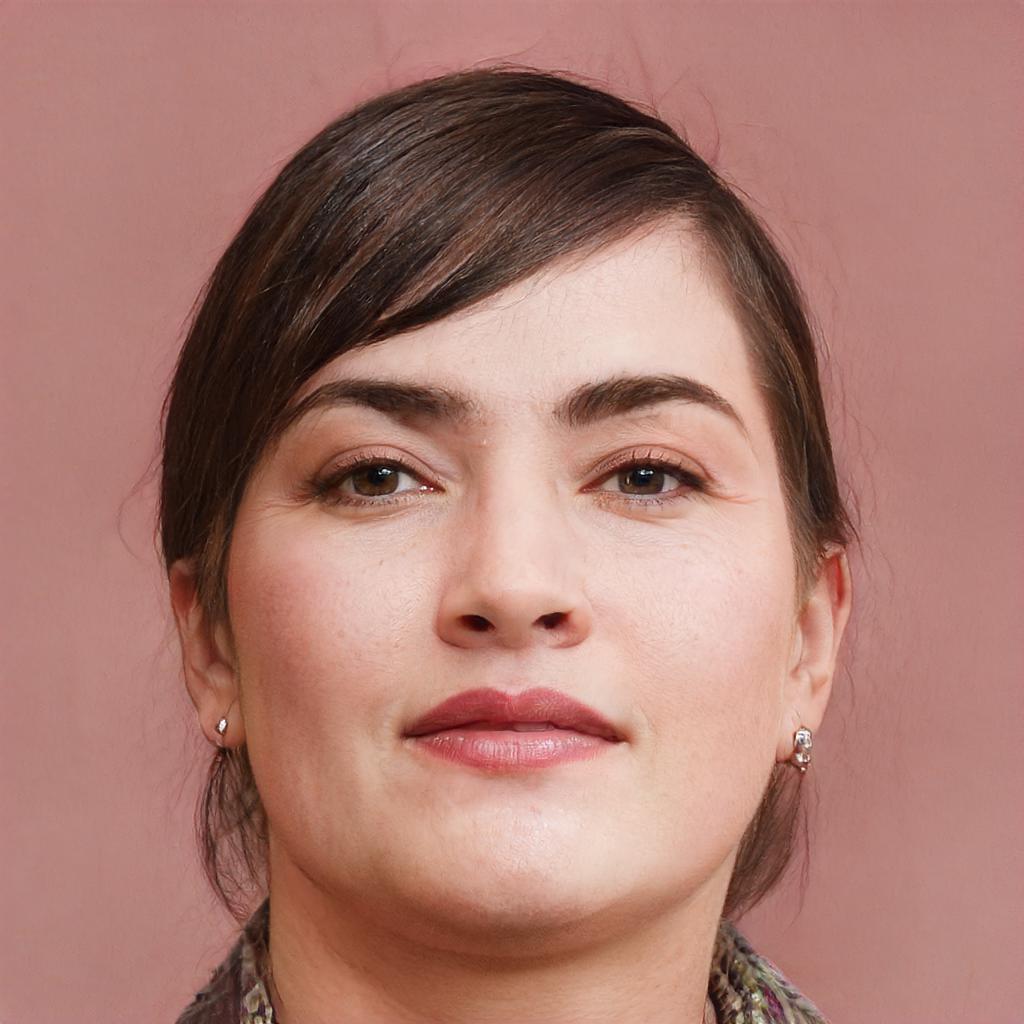} &
        \includegraphics[width=0.1375\linewidth]{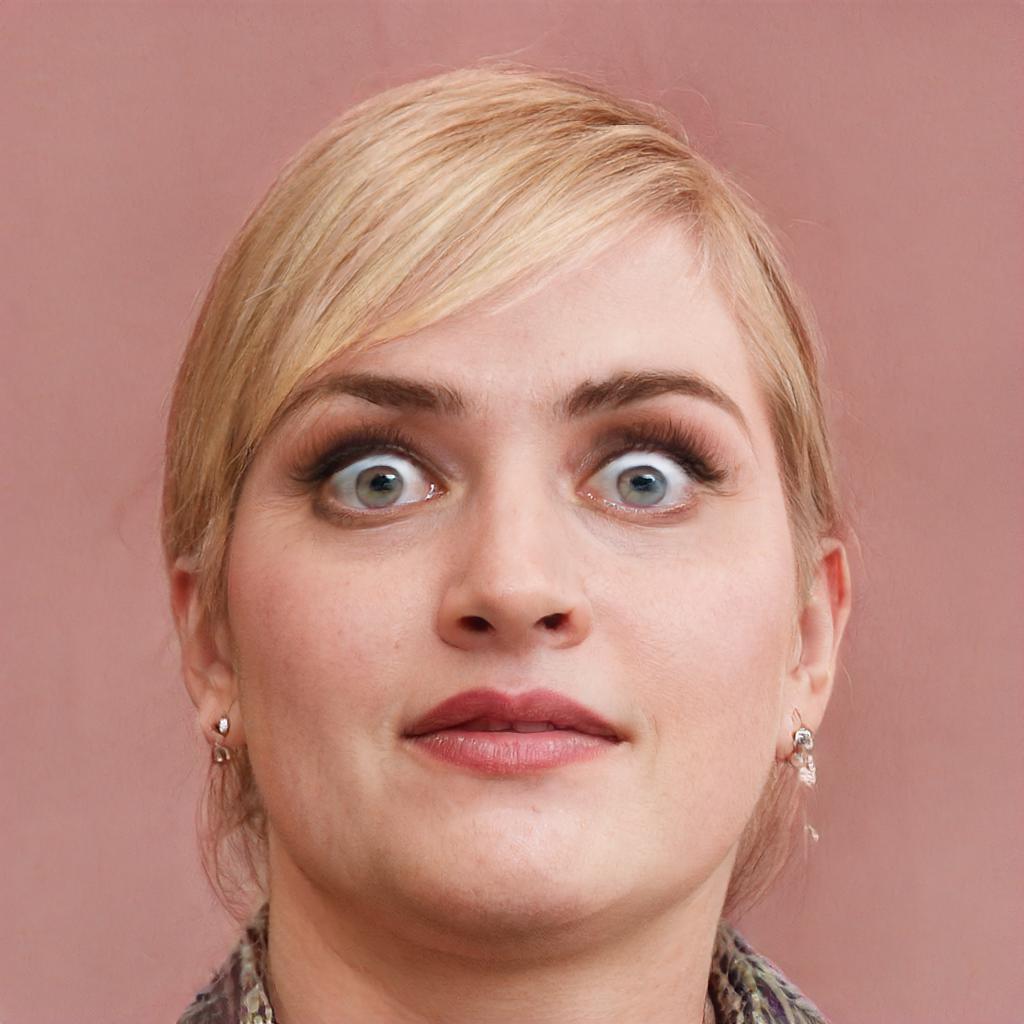} 
        \tabularnewline

        \includegraphics[width=0.1375\linewidth]{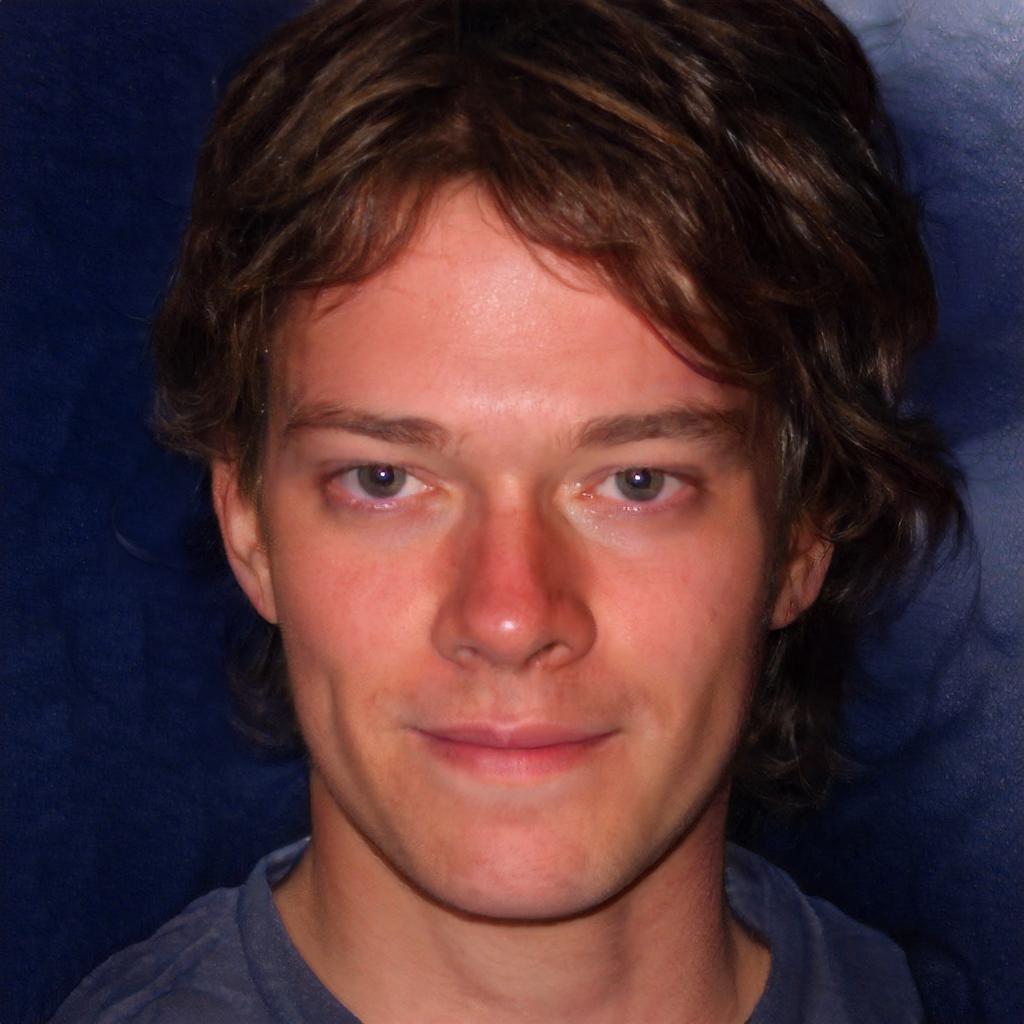} &&
        \raisebox{0.35in}{\rotatebox[origin=t]{90}{ StyleGAN2}} &
        \includegraphics[width=0.1375\linewidth]{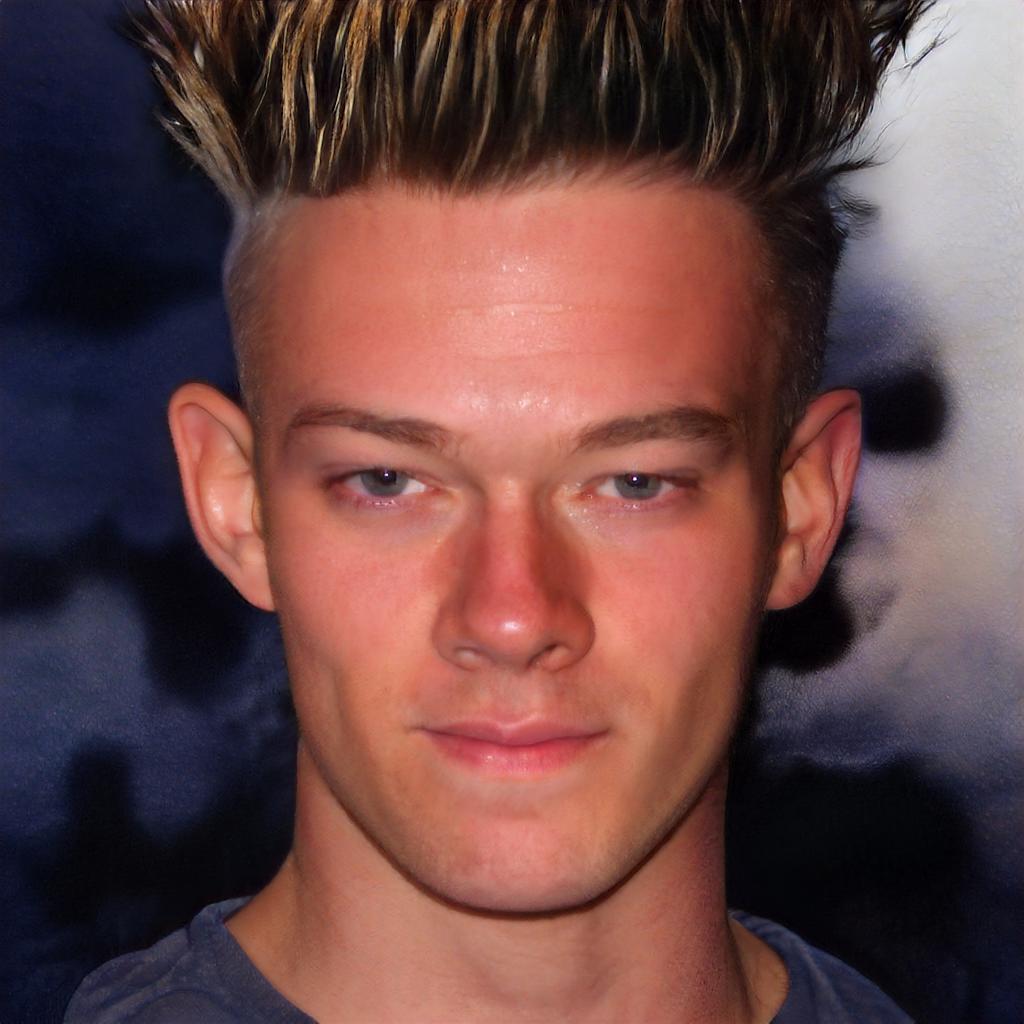} &
        \includegraphics[width=0.1375\linewidth]{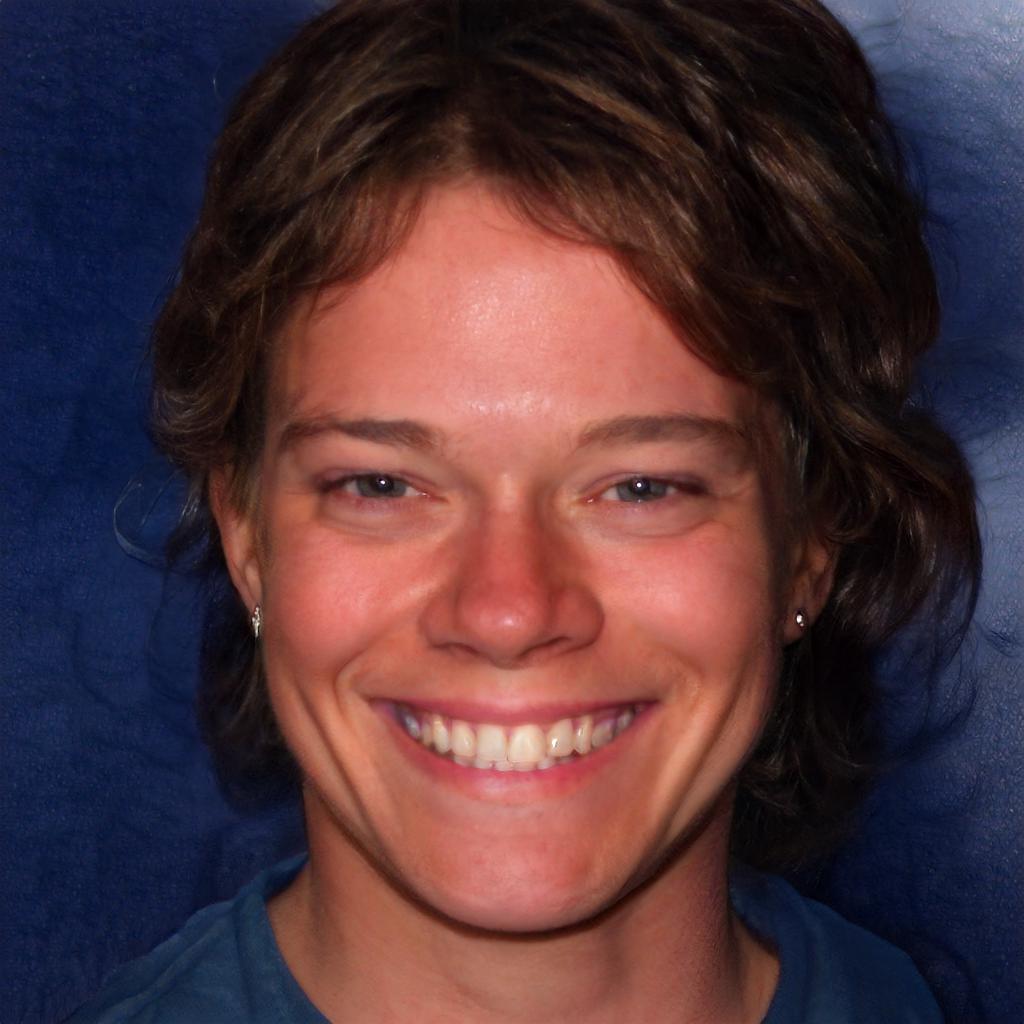} &
        \includegraphics[width=0.1375\linewidth]{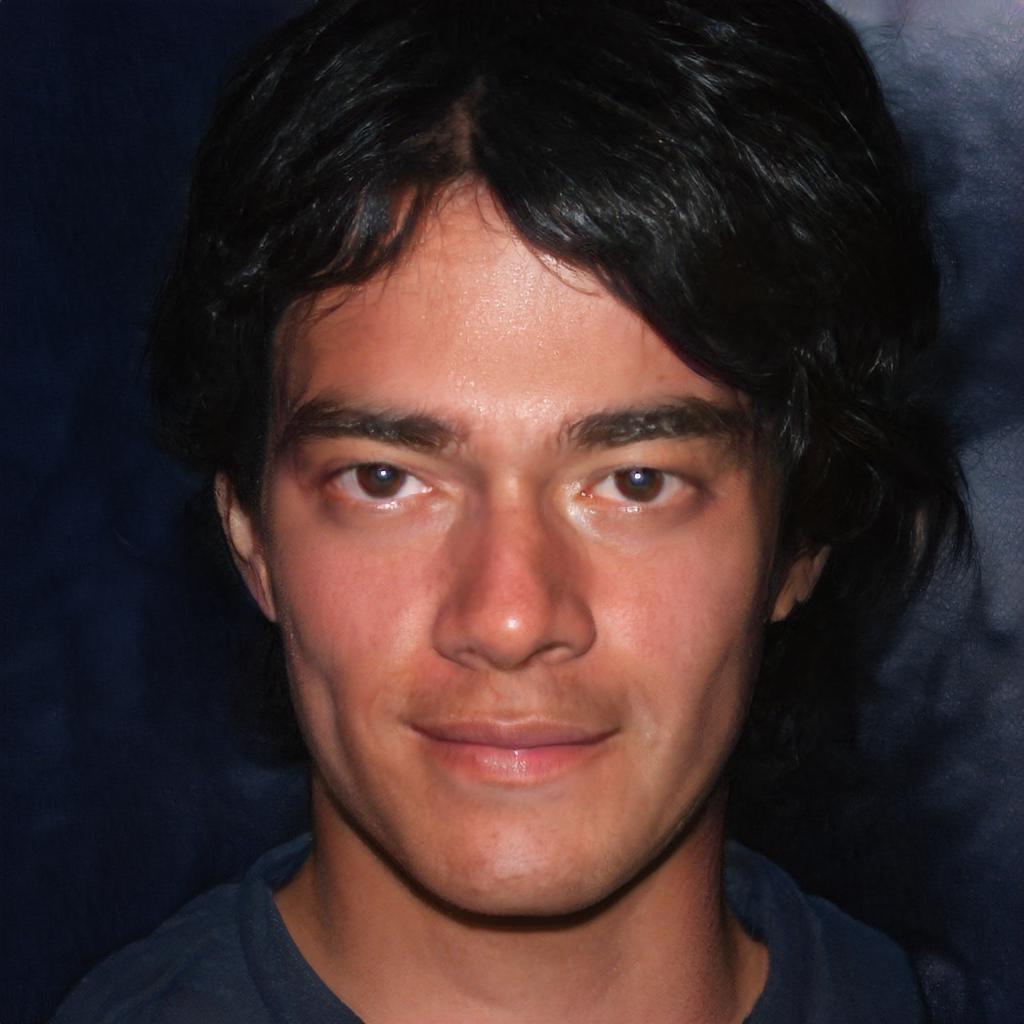} &
        \includegraphics[width=0.1375\linewidth]{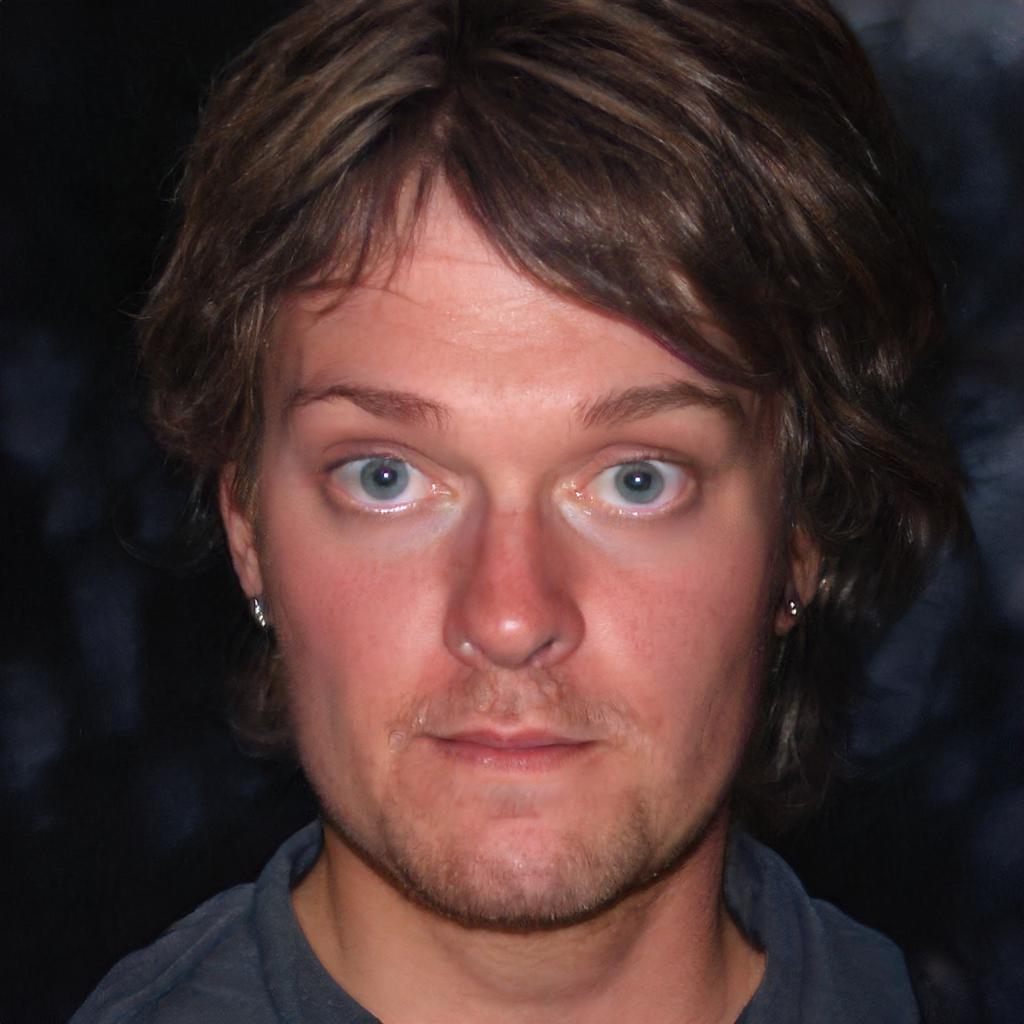} 
        \tabularnewline
        && \raisebox{0.35in}{\rotatebox[origin=t]{90}{ StyleFusion}} &
        \includegraphics[width=0.1375\linewidth]{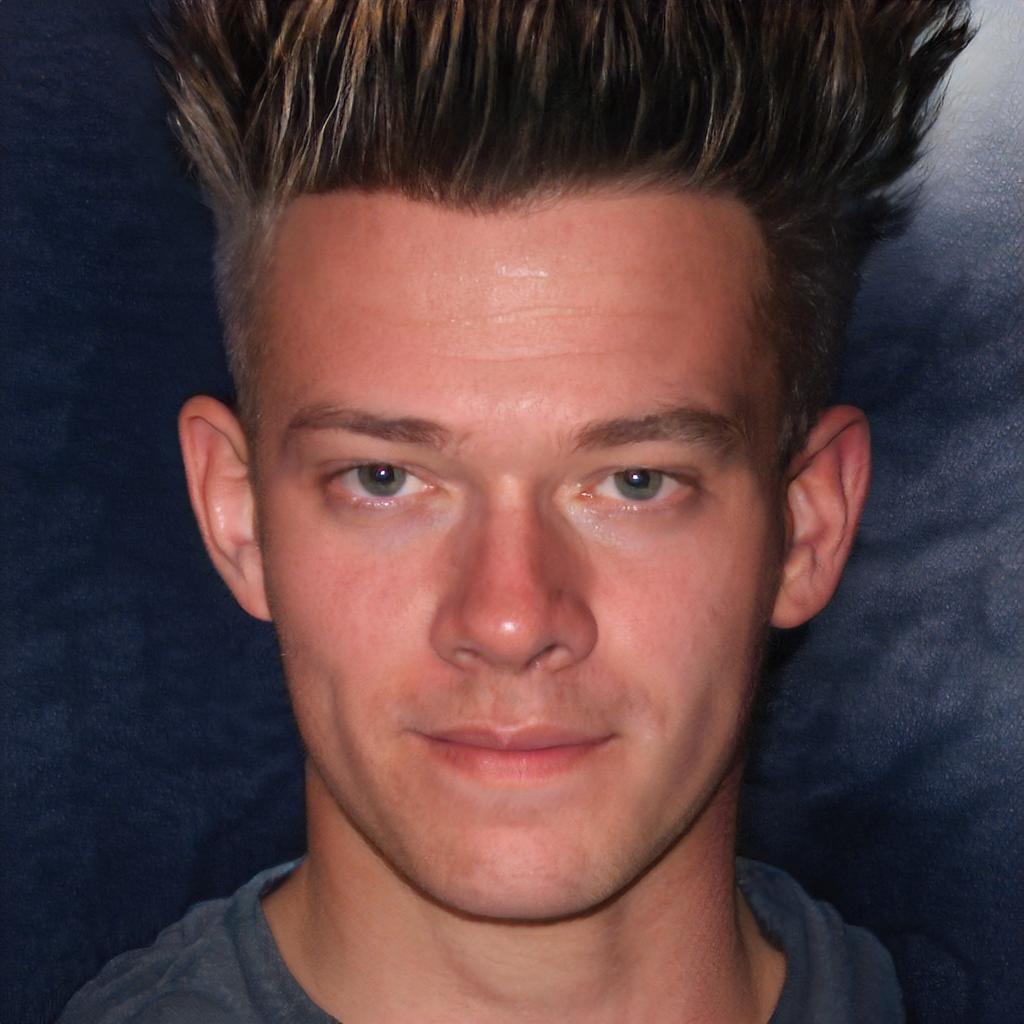} &
        \includegraphics[width=0.1375\linewidth]{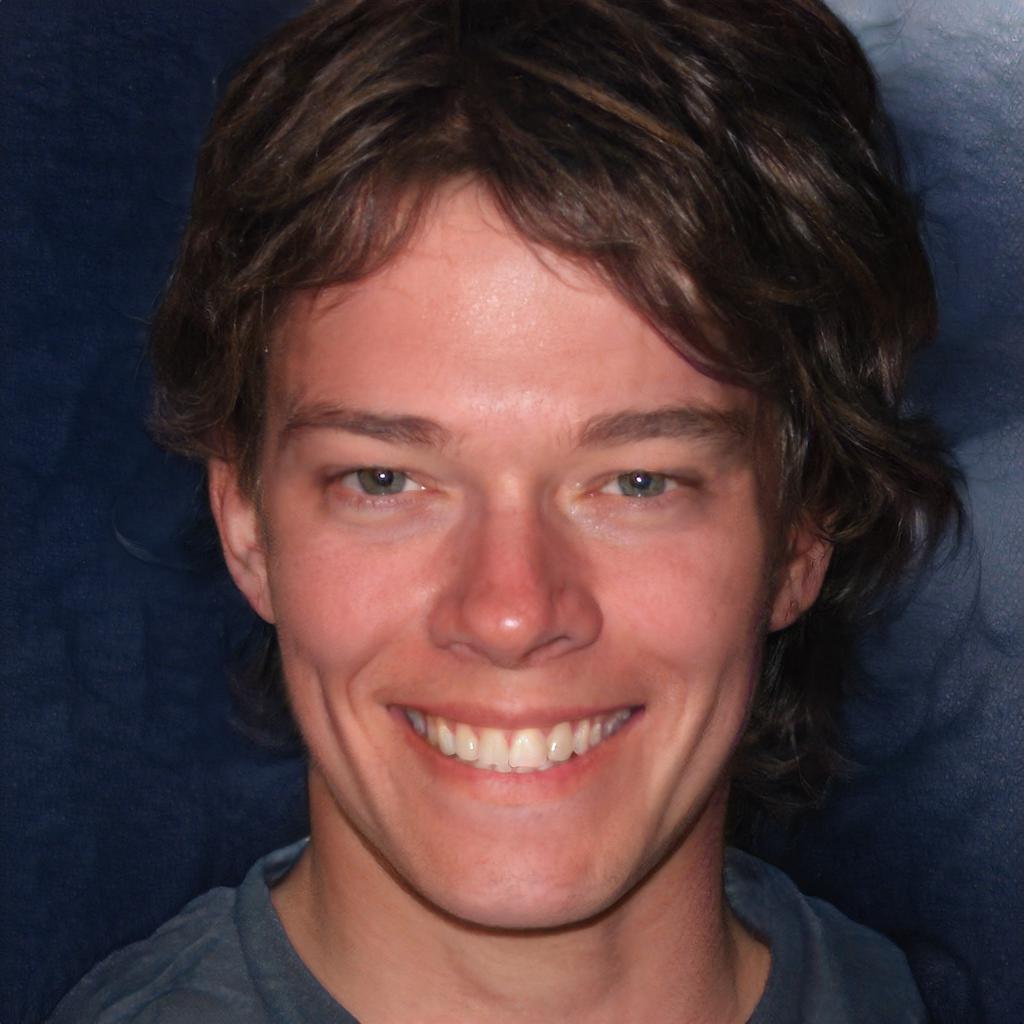} &
        \includegraphics[width=0.1375\linewidth]{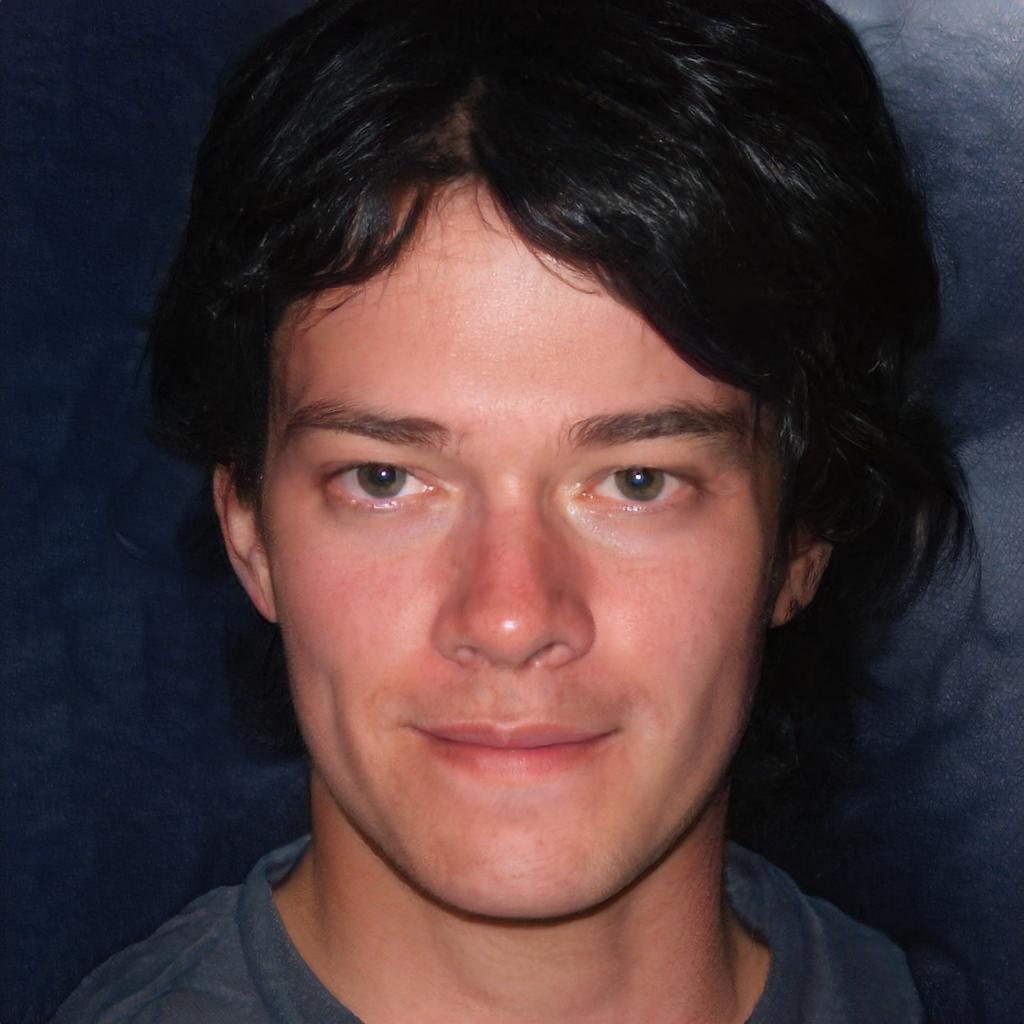} &
        \includegraphics[width=0.1375\linewidth]{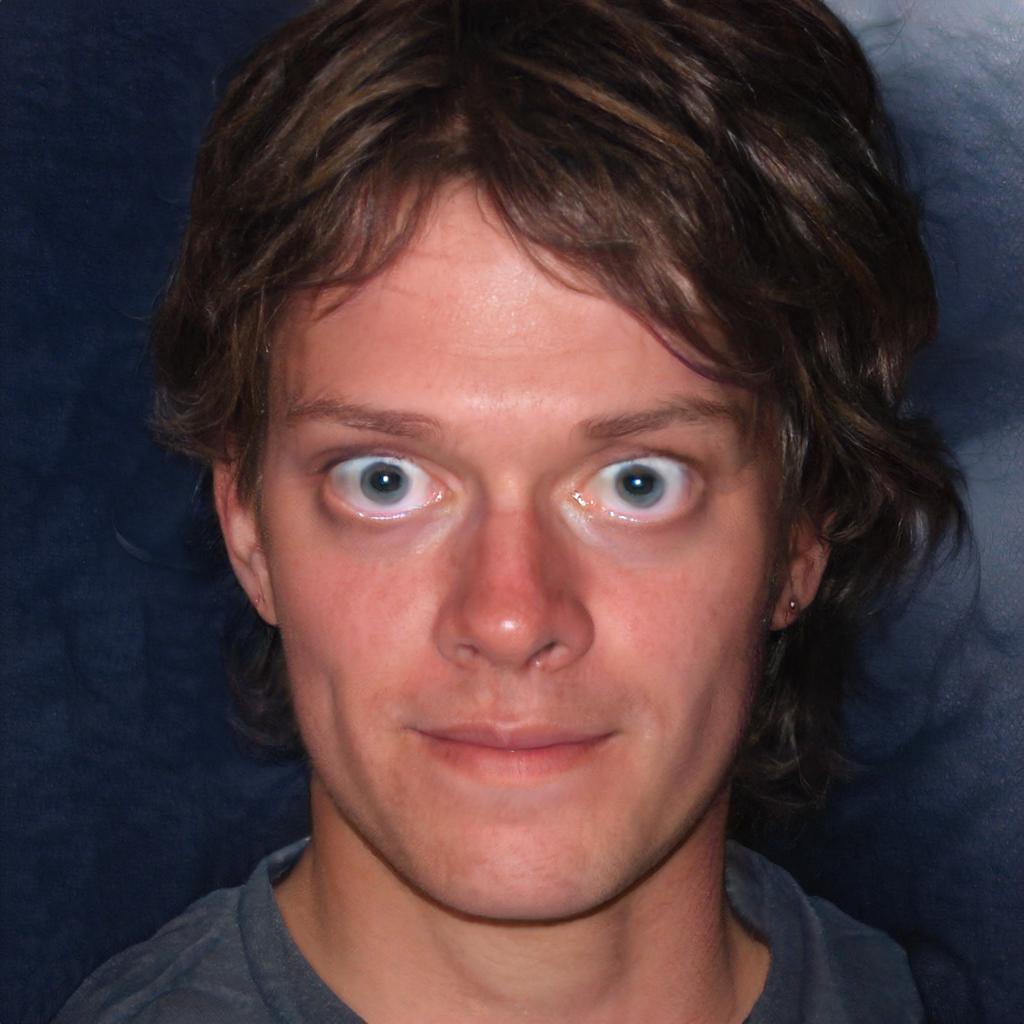} 
        \tabularnewline

        \includegraphics[width=0.135\linewidth]{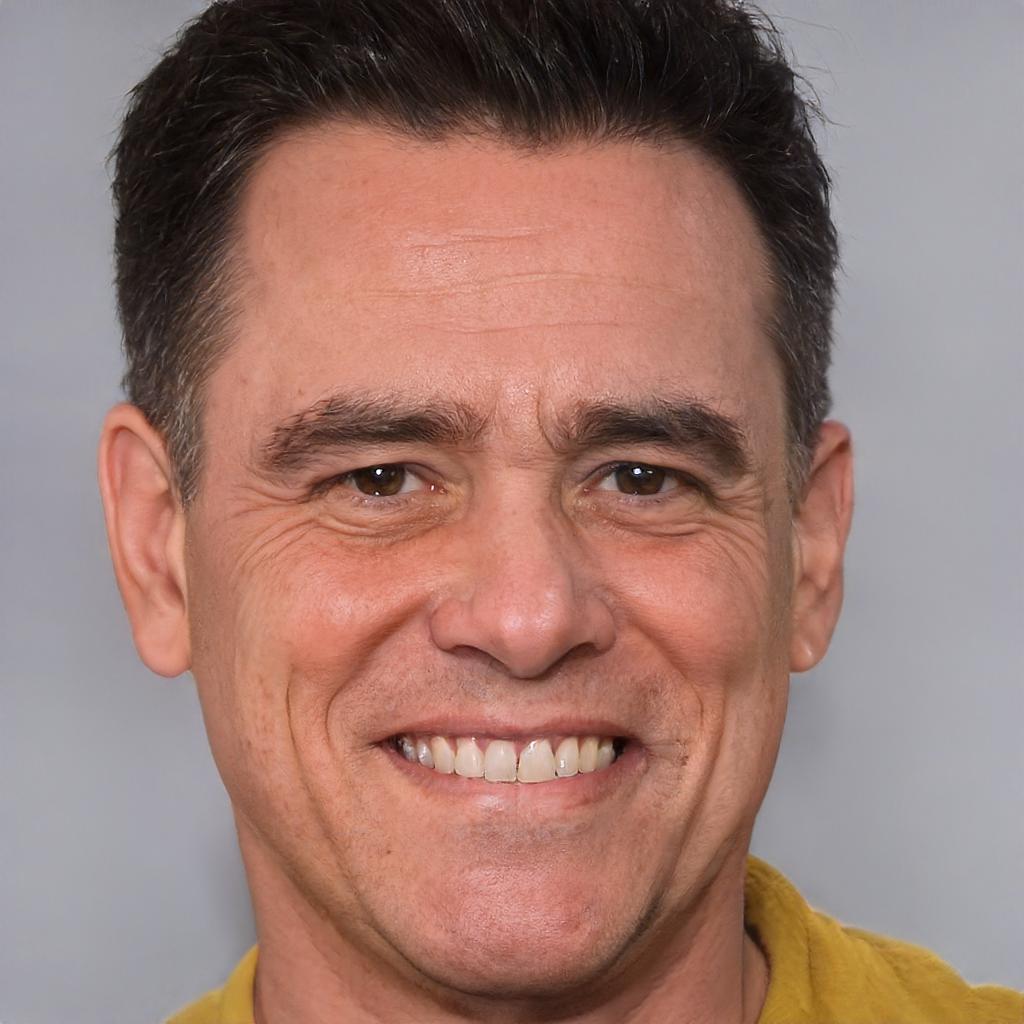} &&
        \raisebox{0.35in}{\rotatebox[origin=t]{90}{\footnotesize StyleGAN2}} &
        \includegraphics[width=0.135\linewidth]{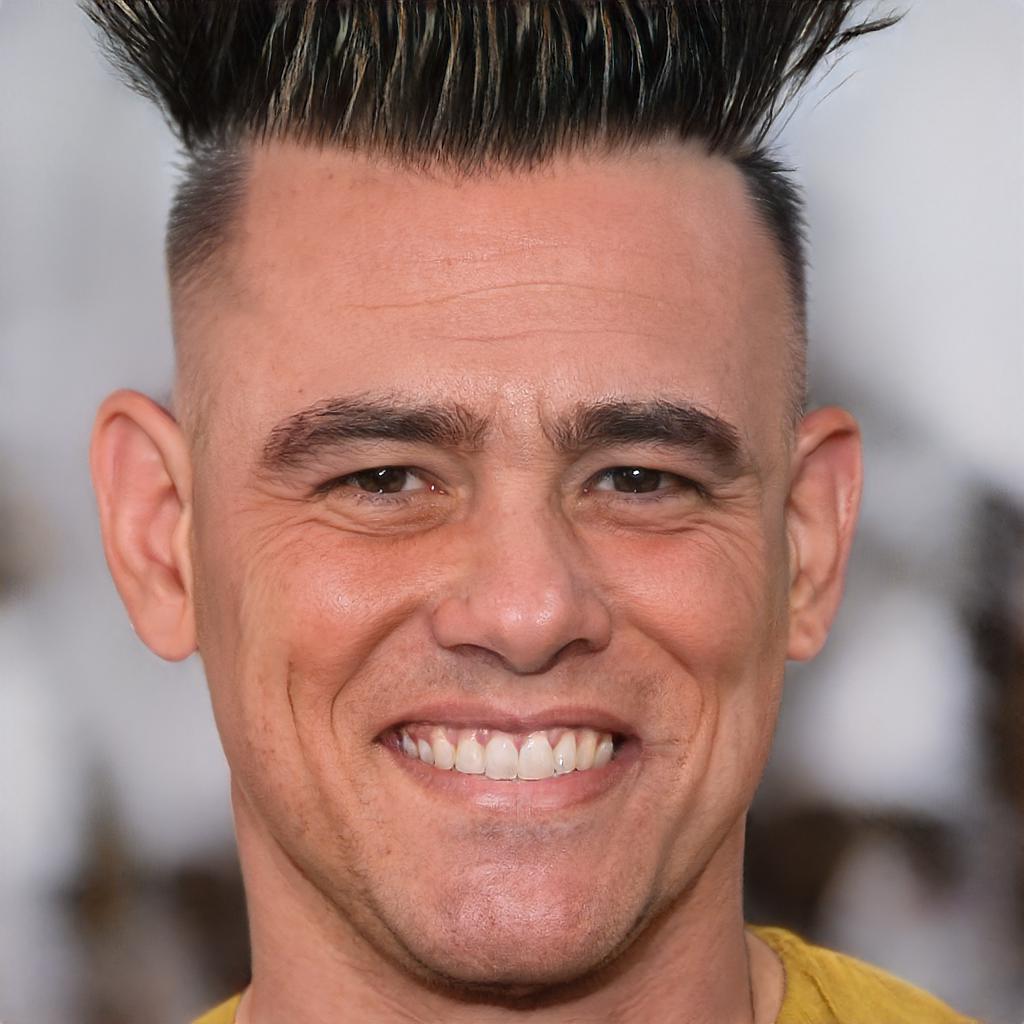} &
        \includegraphics[width=0.135\linewidth]{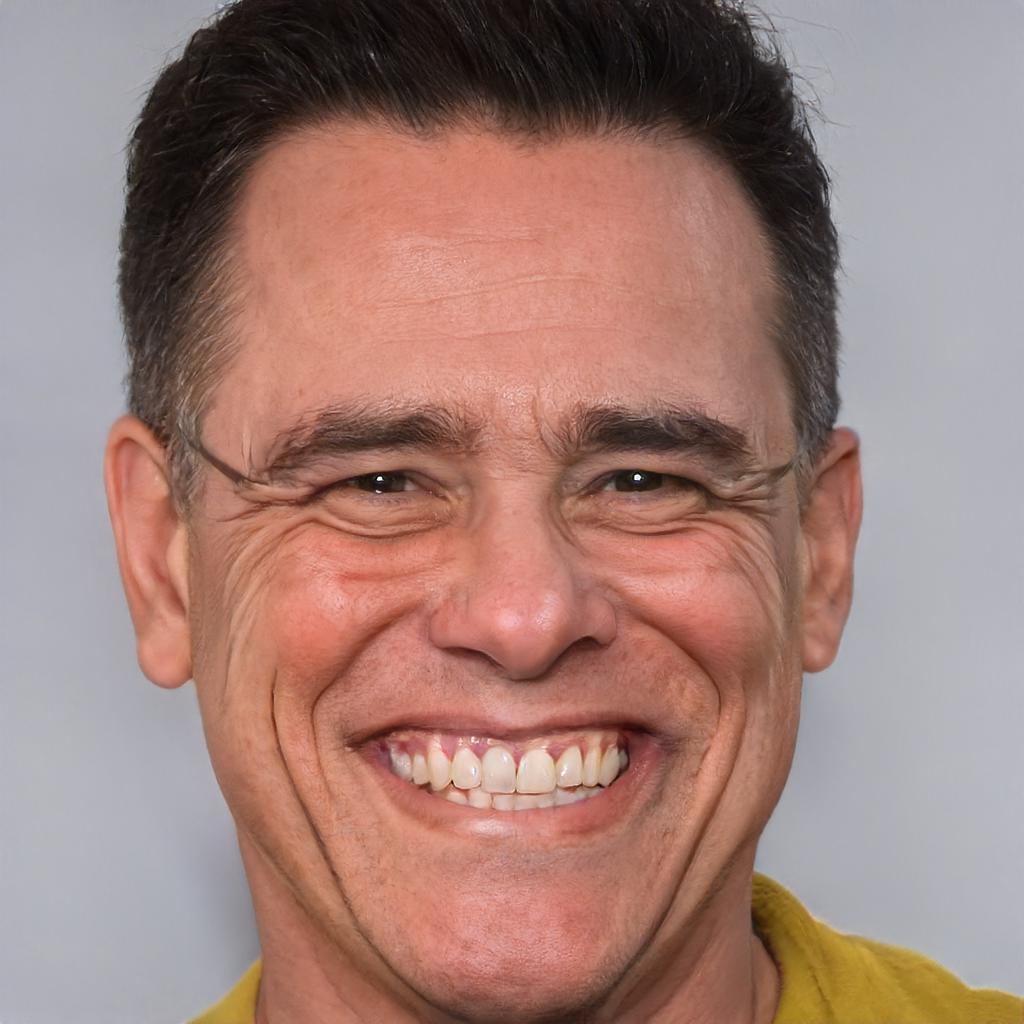} &
        \includegraphics[width=0.135\linewidth]{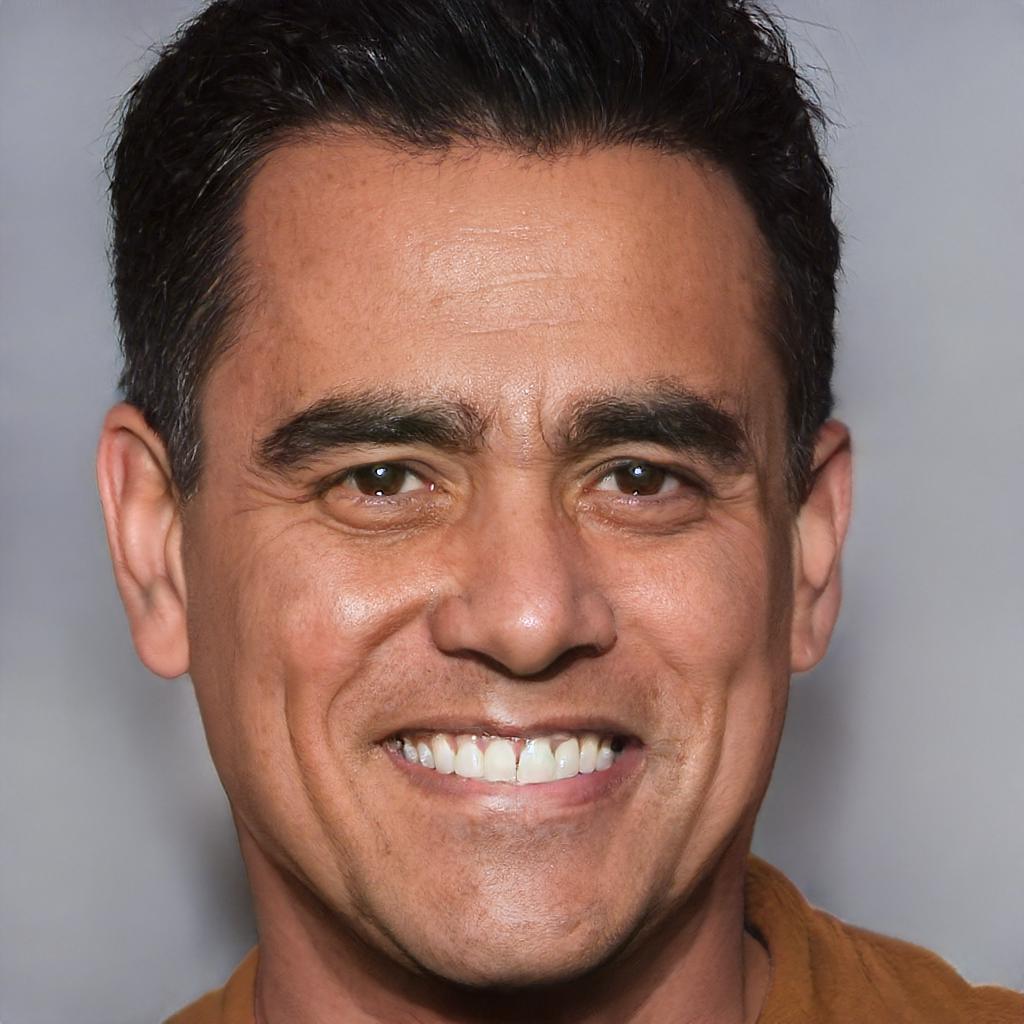} &
        \includegraphics[width=0.135\linewidth]{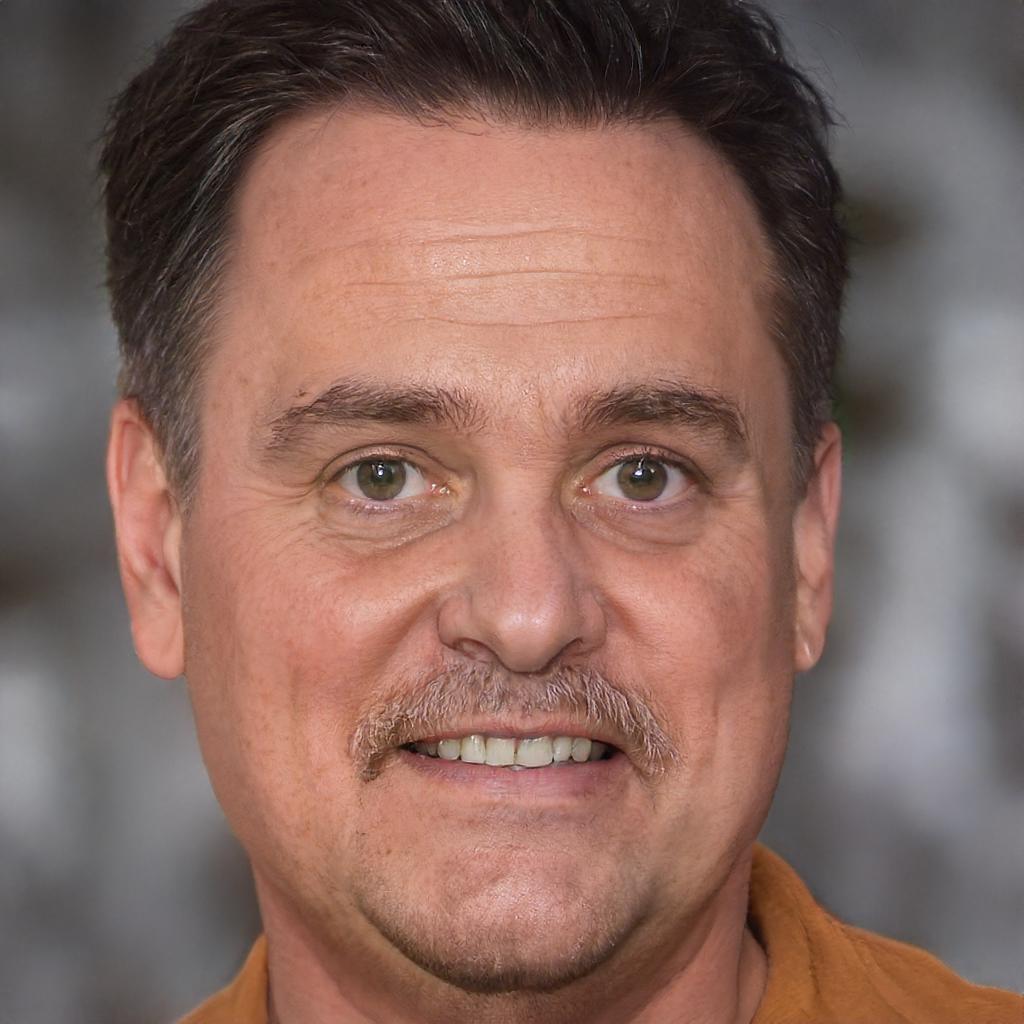} 
        \tabularnewline
        && \raisebox{0.35in}{\rotatebox[origin=t]{90}{\footnotesize StyleFusion}} &
        \includegraphics[width=0.135\linewidth]{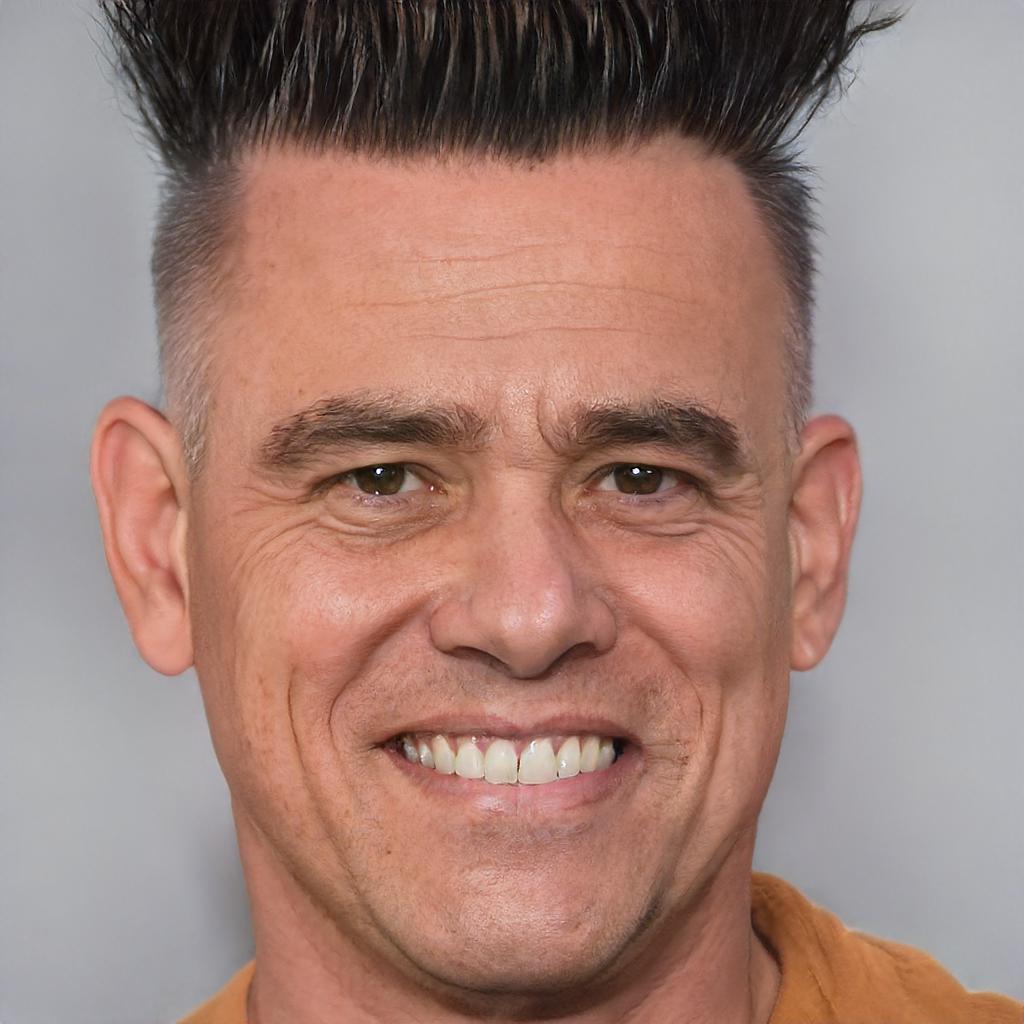} &
        \includegraphics[width=0.135\linewidth]{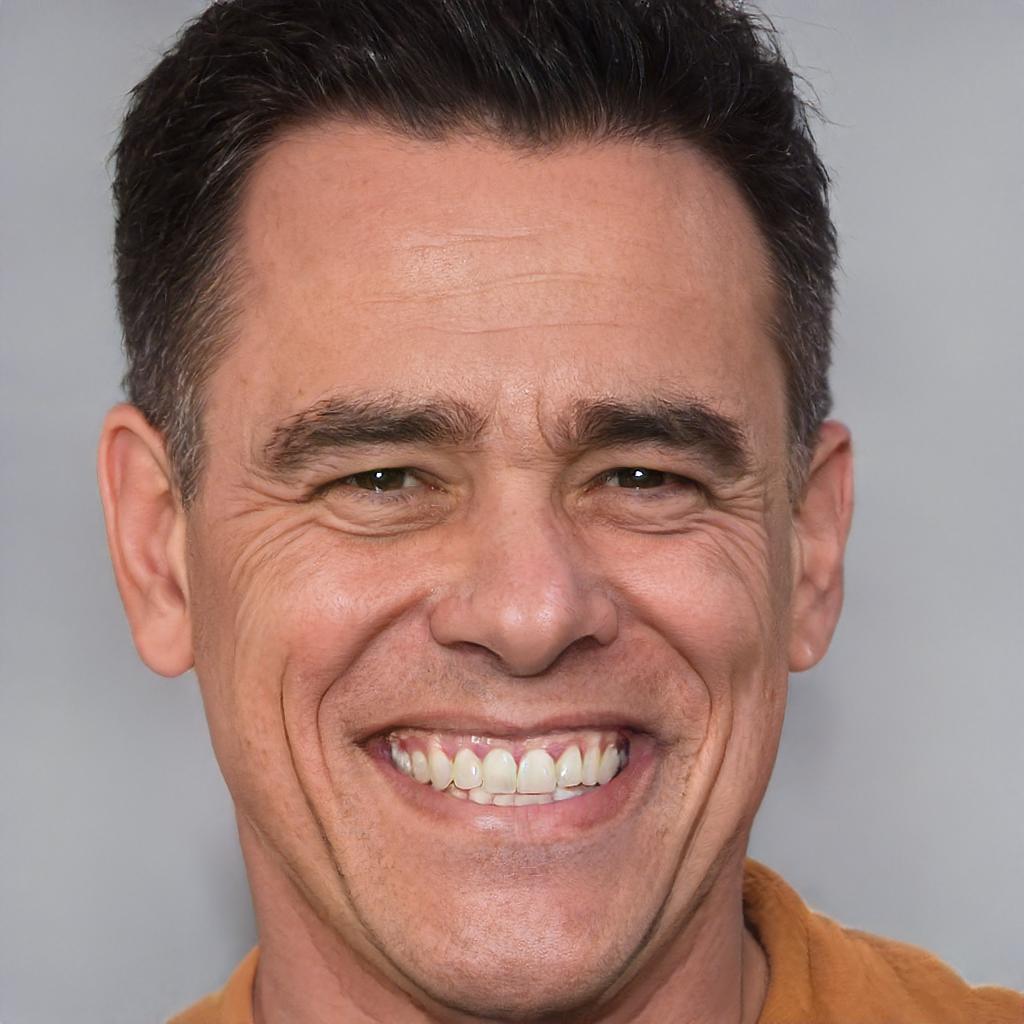} &
        \includegraphics[width=0.135\linewidth]{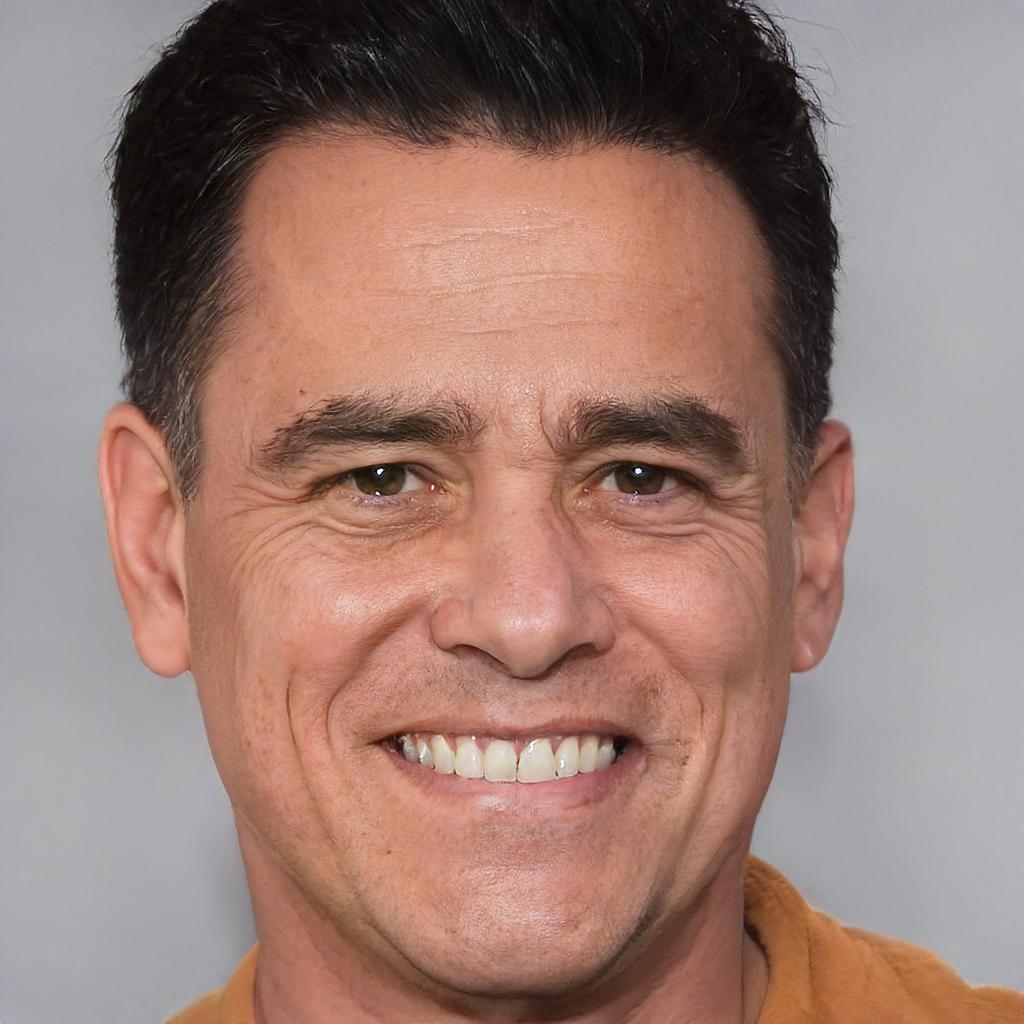} &
        \includegraphics[width=0.135\linewidth]{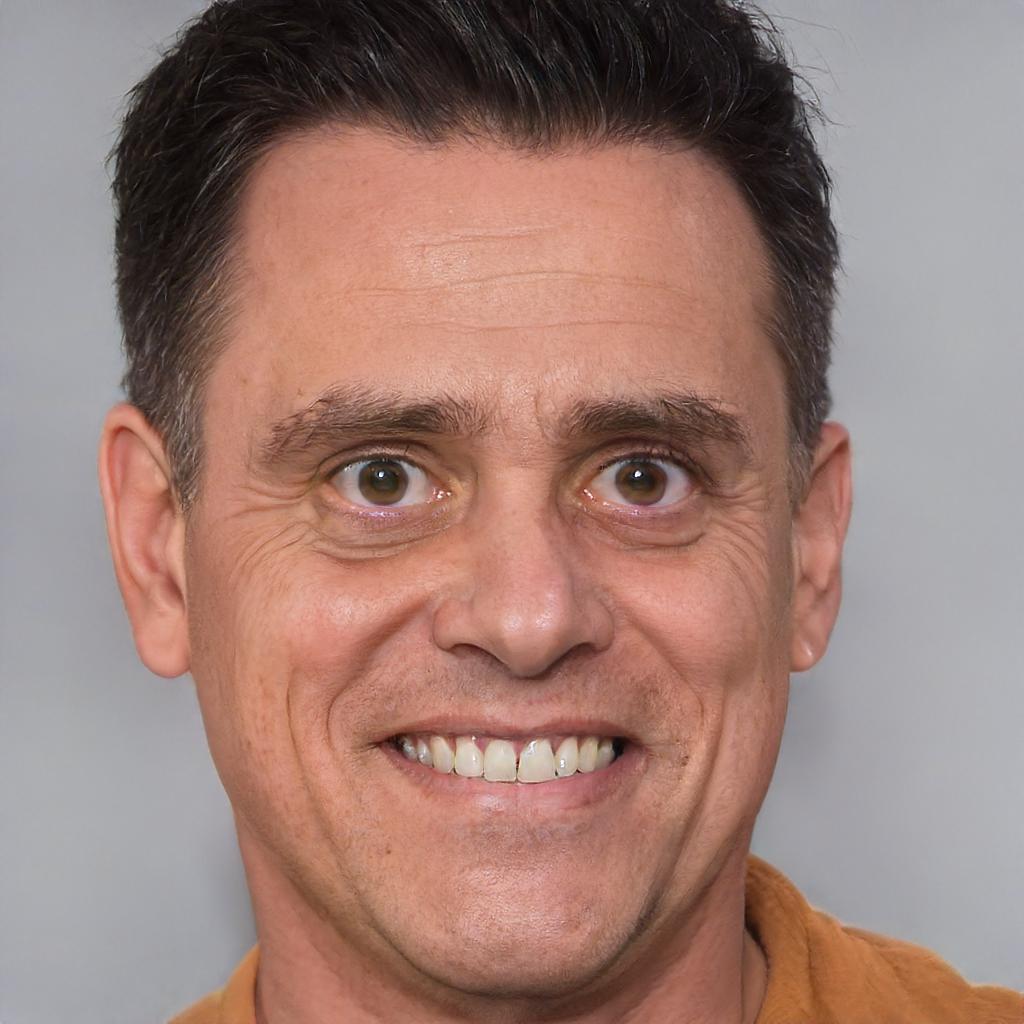} 
        \tabularnewline

        Input &&& Mohawk & Smile & Hair Albedo & Fearful Eyes
        \tabularnewline
        \end{tabular}
        }
    \vspace{0.1cm}
    \caption{Latent traversal editing techniques performed on real images with and without StyleFusion. Observe how StyleFusion helps facilitate more precise manipulation of the images by constraining the edit to the desired image region(s).}
    \label{fig:latent_editing_appendix_1}
\end{figure*}

\begin{figure*}
    \setlength{\tabcolsep}{1.5pt}
    \centering
    \small{
        \begin{tabular}{c c c c c c c}
        
        &&& StyleCLIP & InterFace & GANSpace & GANSpace \\
        
        \includegraphics[width=0.135\linewidth]{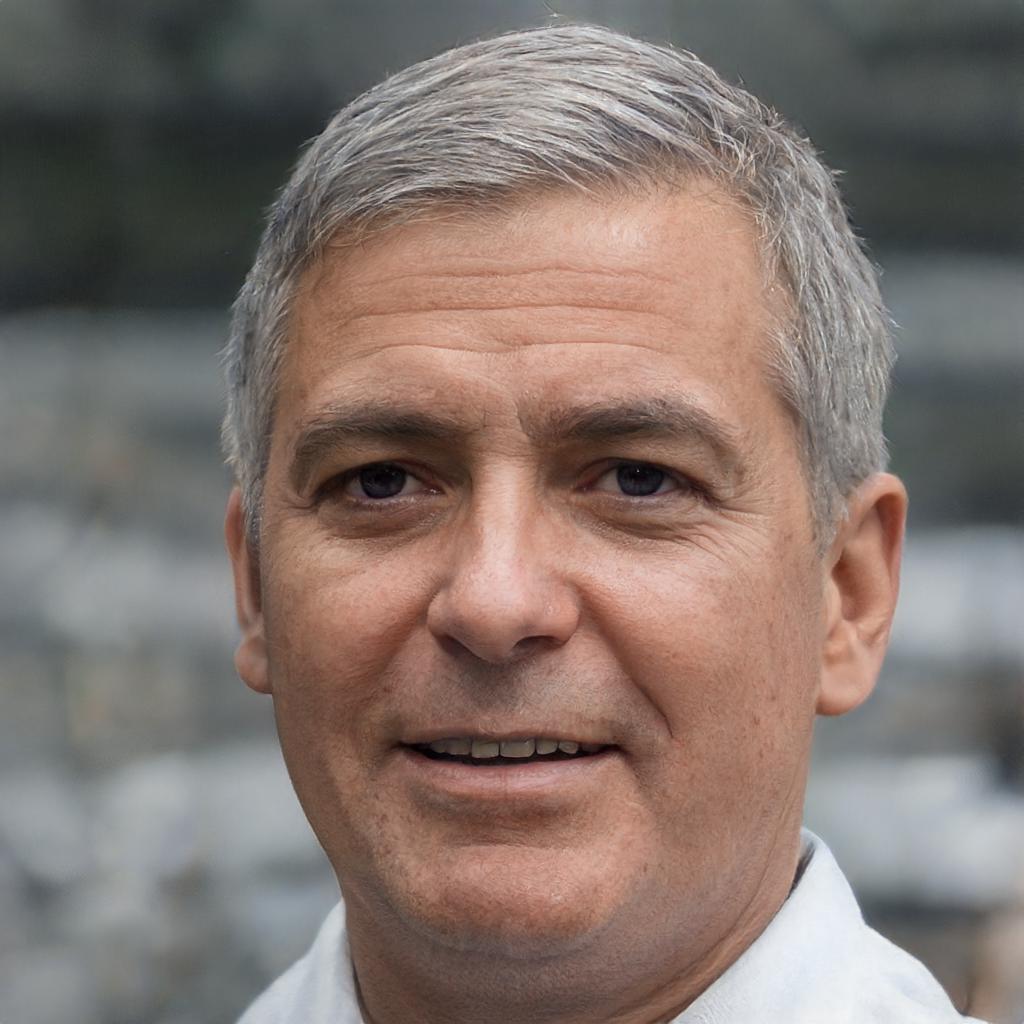} &&
        \raisebox{0.35in}{\rotatebox[origin=t]{90}{\footnotesize StyleGAN2}} &
        \includegraphics[width=0.135\linewidth]{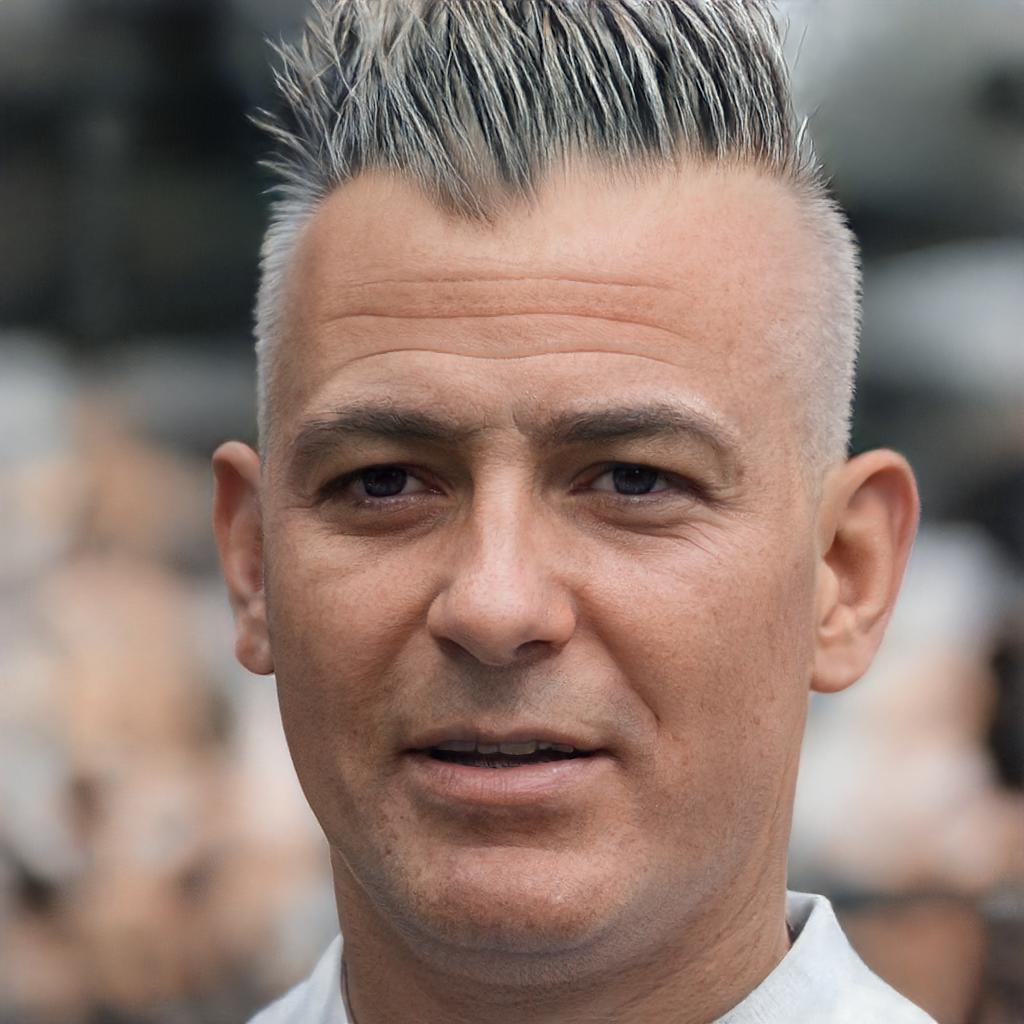} &
        \includegraphics[width=0.135\linewidth]{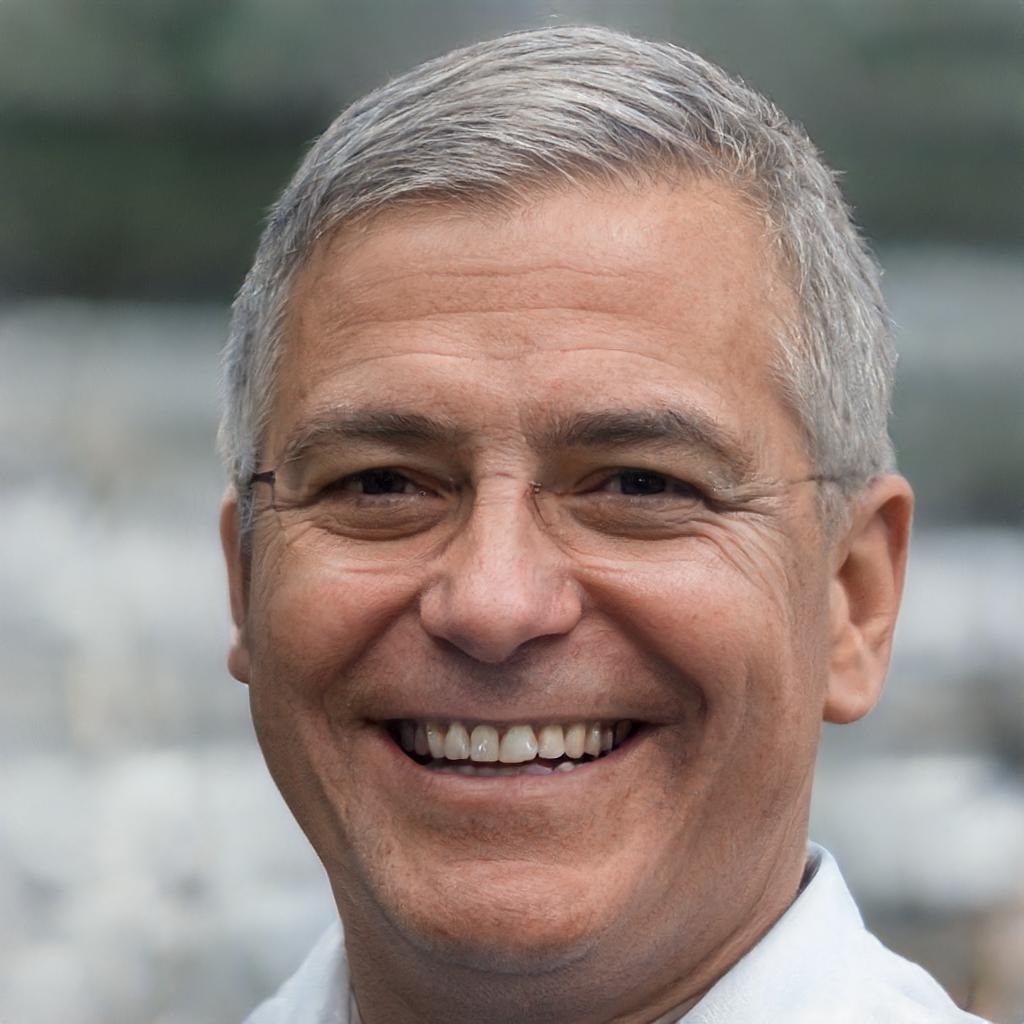} &
        \includegraphics[width=0.135\linewidth]{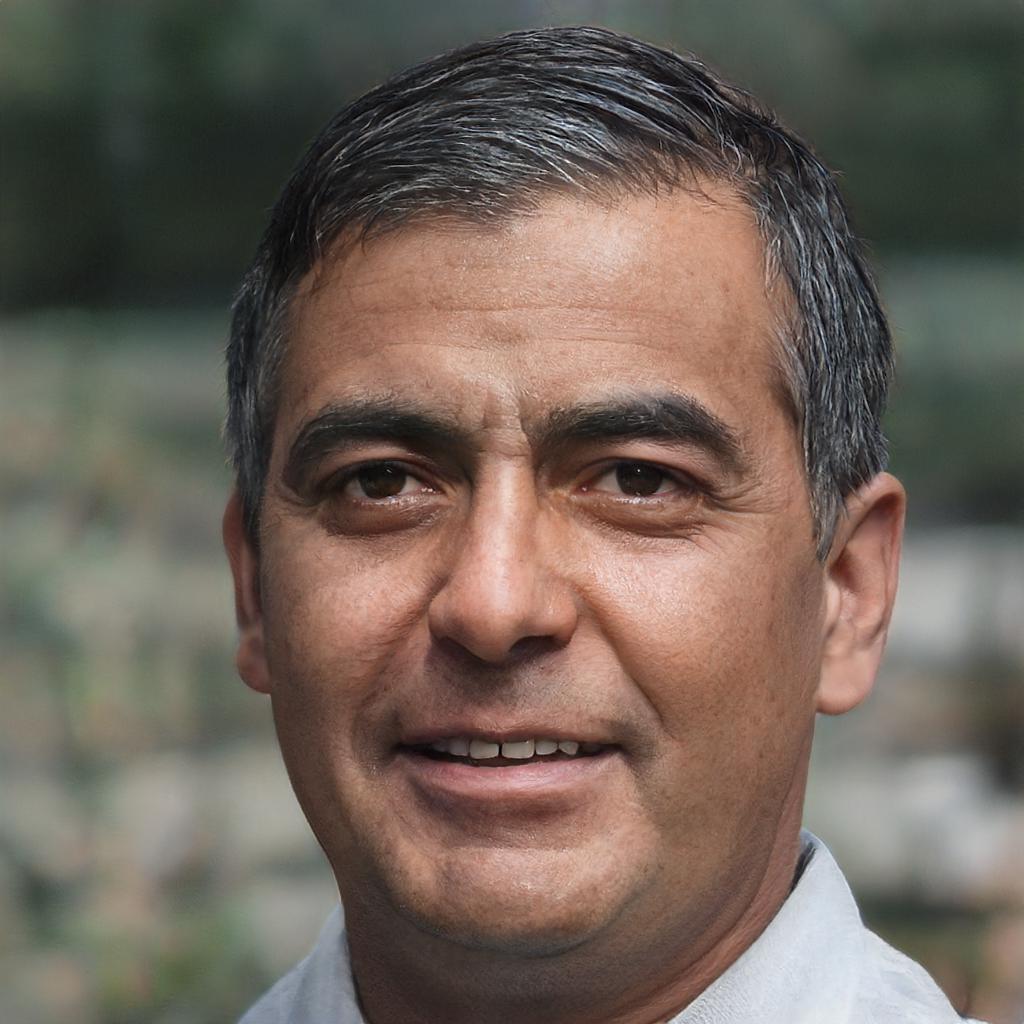} &
        \includegraphics[width=0.135\linewidth]{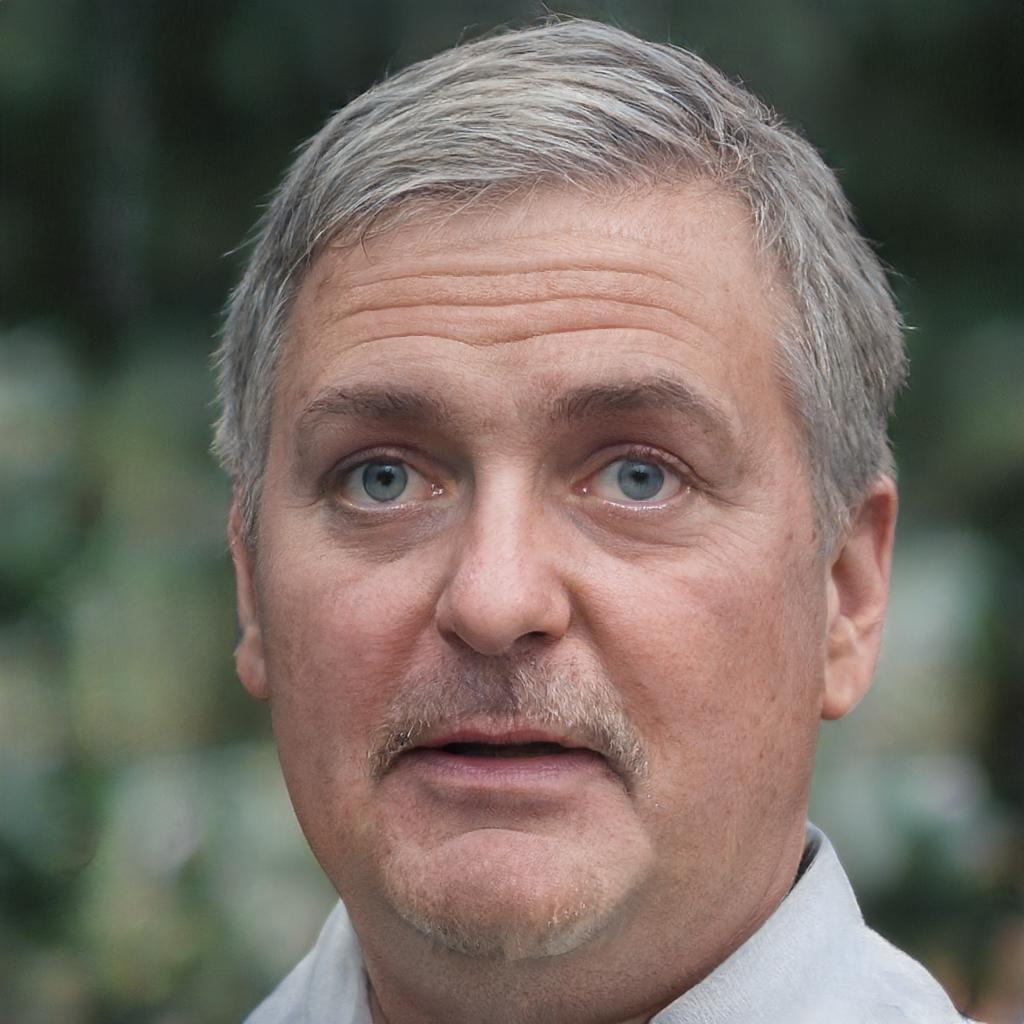} 
        \tabularnewline
        && \raisebox{0.35in}{\rotatebox[origin=t]{90}{\footnotesize StyleFusion}} &
        \includegraphics[width=0.135\linewidth]{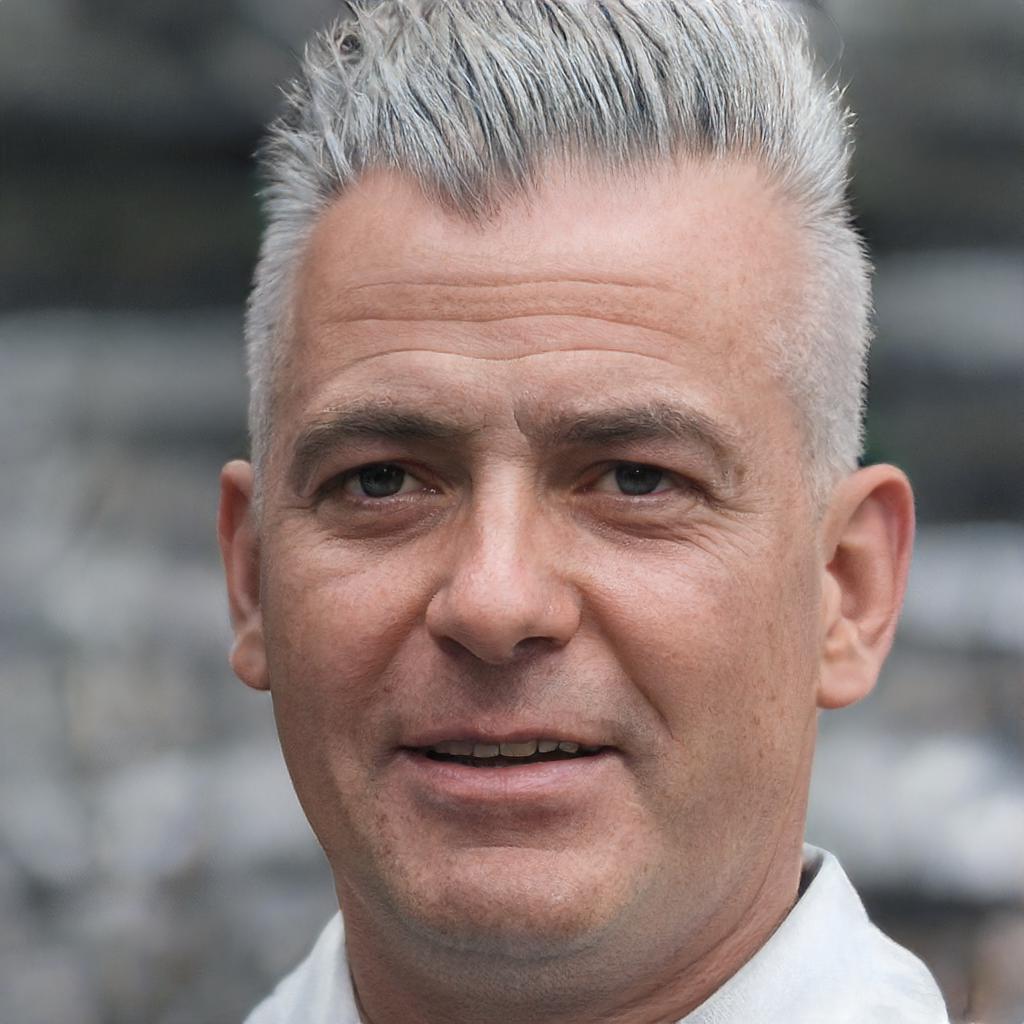} &
        \includegraphics[width=0.135\linewidth]{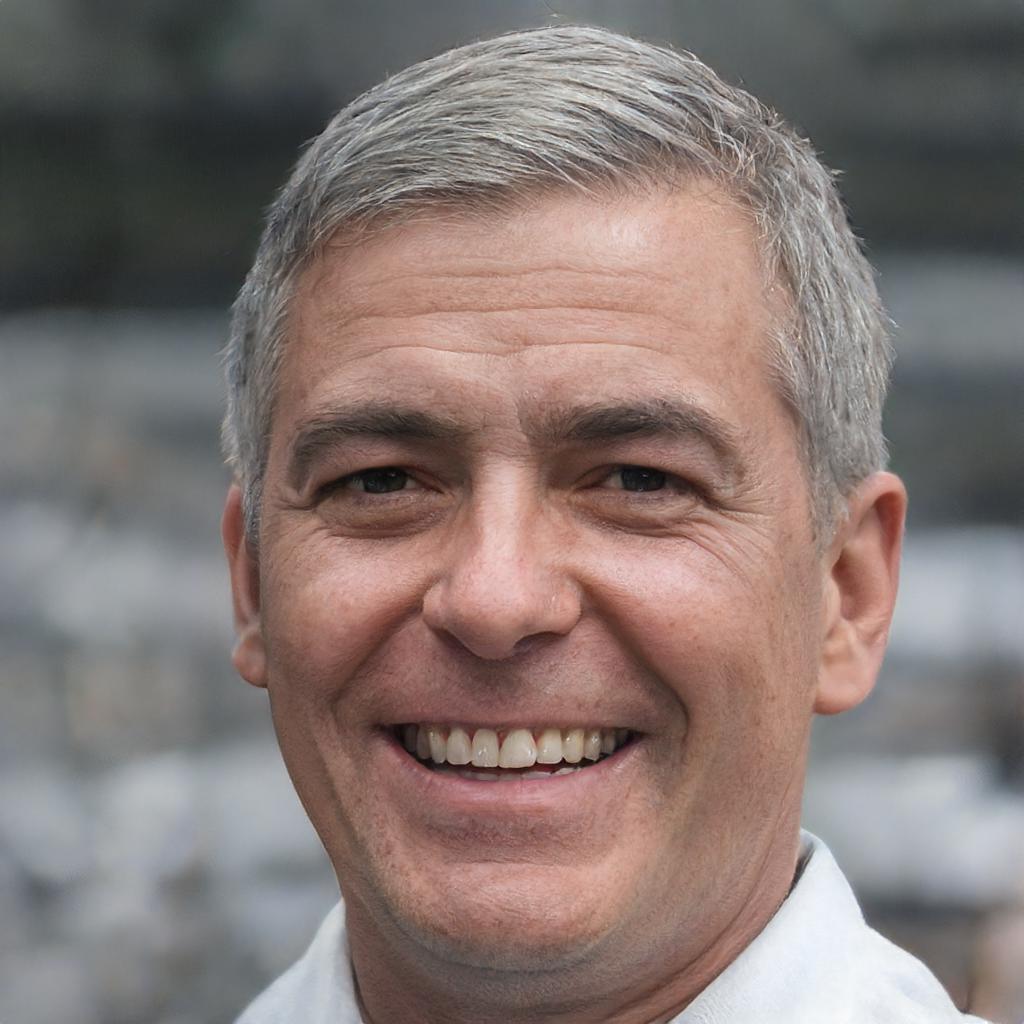} &
        \includegraphics[width=0.135\linewidth]{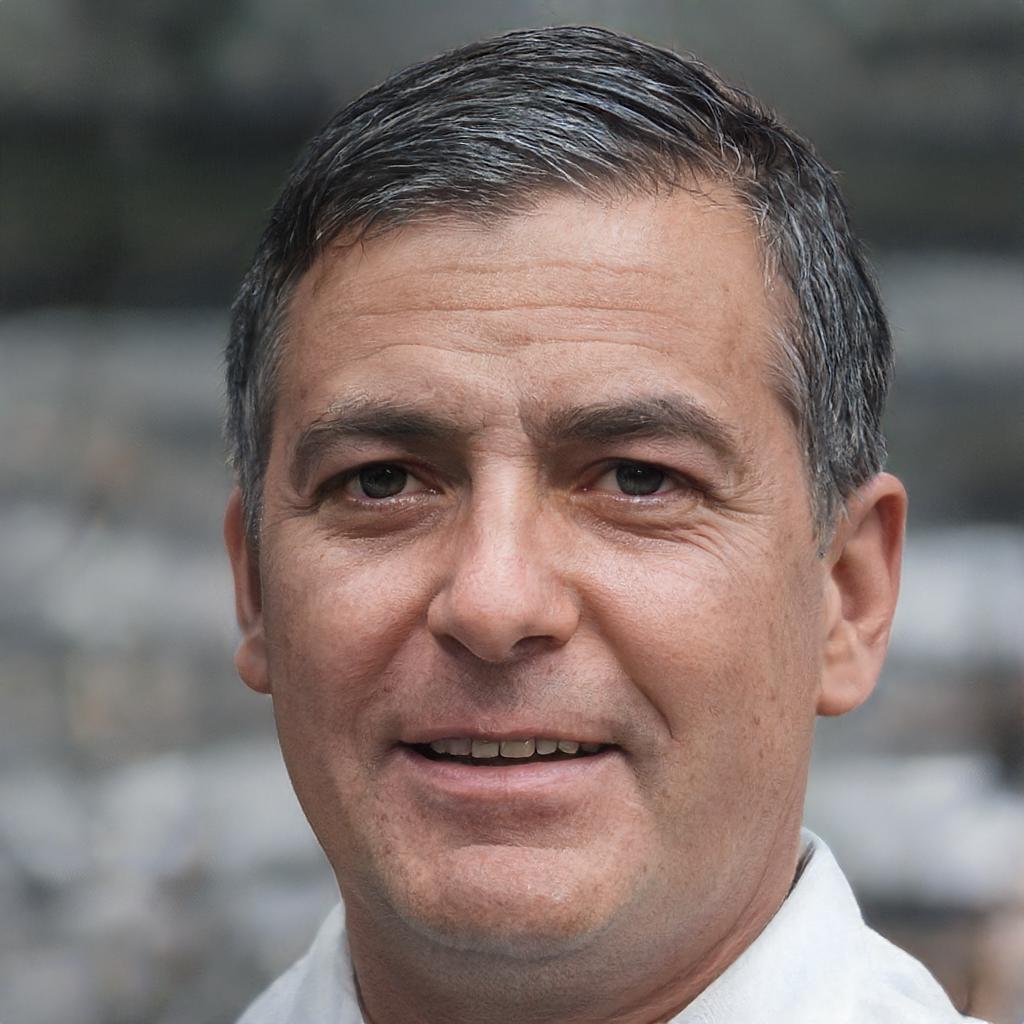} &
        \includegraphics[width=0.135\linewidth]{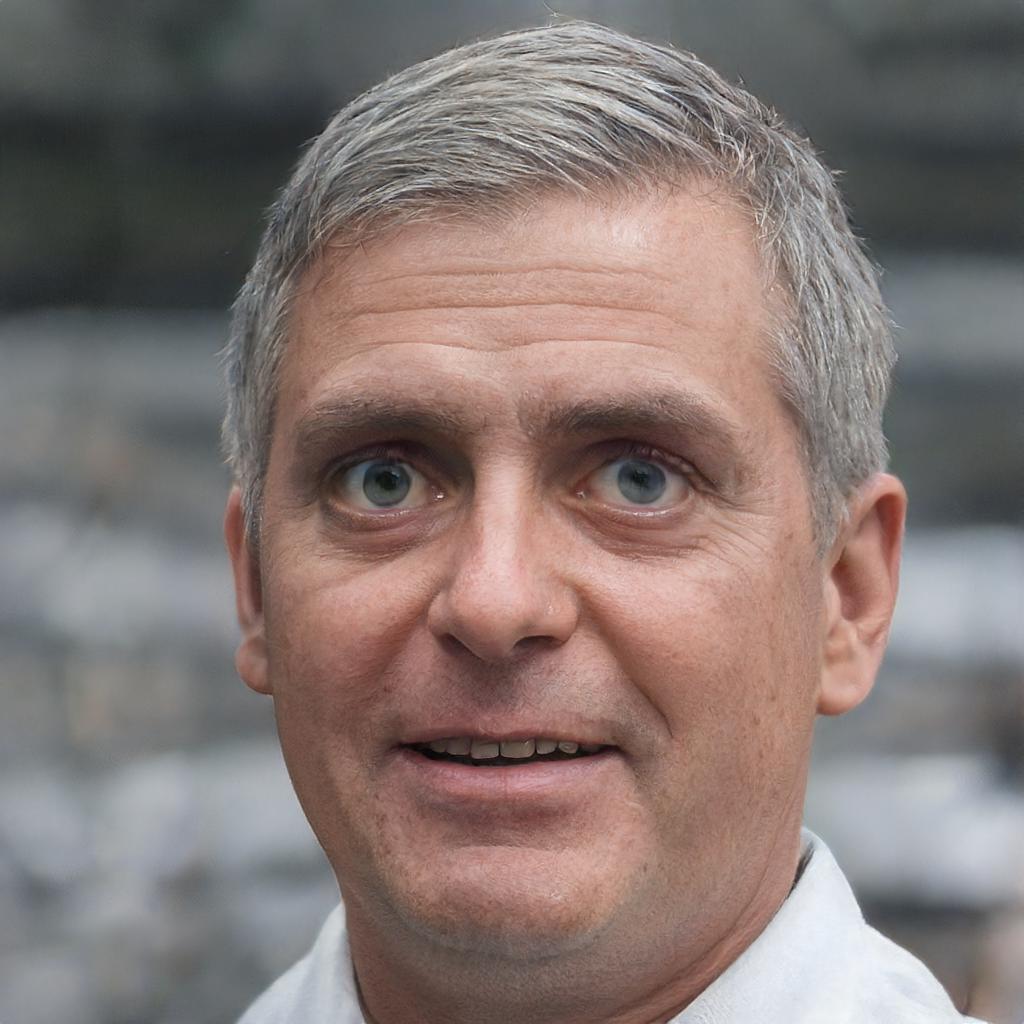} 
        \tabularnewline

        \includegraphics[width=0.135\linewidth]{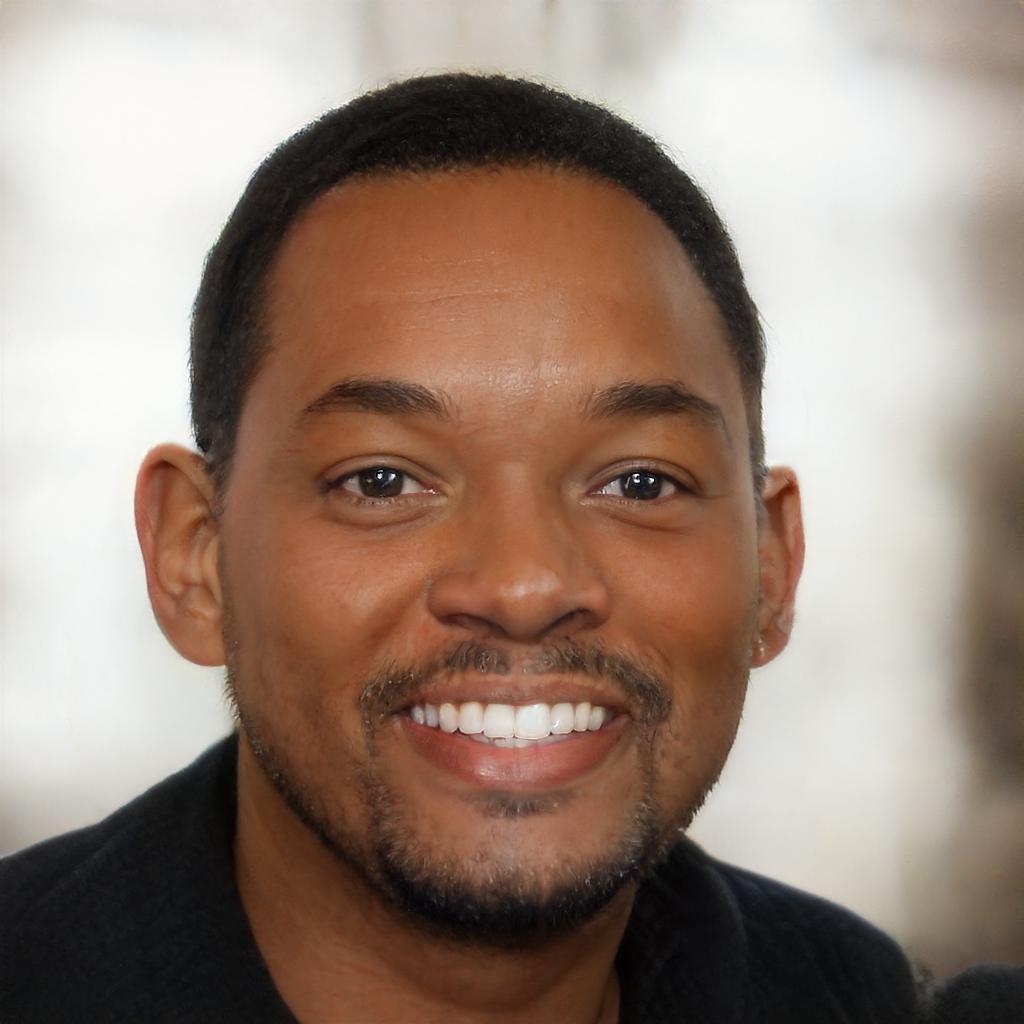} &&
        \raisebox{0.35in}{\rotatebox[origin=t]{90}{\footnotesize StyleGAN2}} &
        \includegraphics[width=0.135\linewidth]{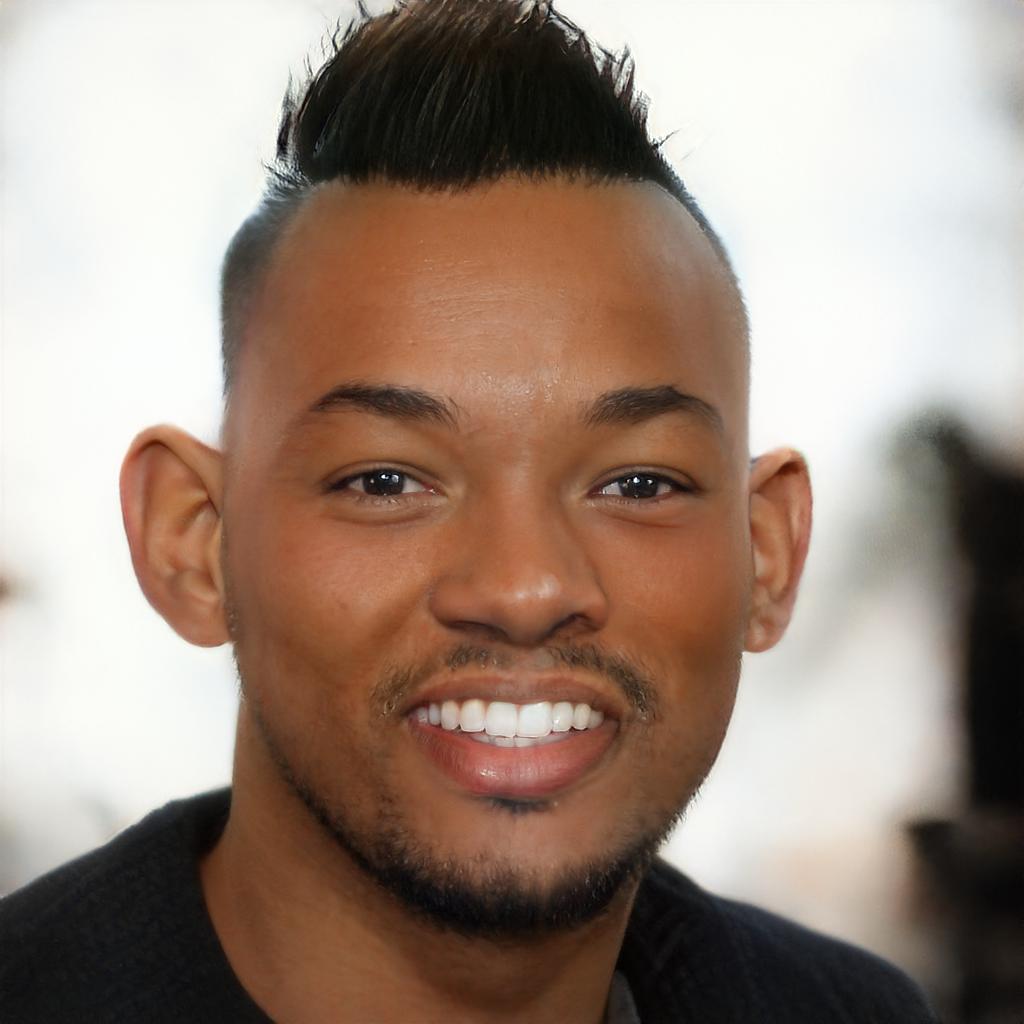} &
        \includegraphics[width=0.135\linewidth]{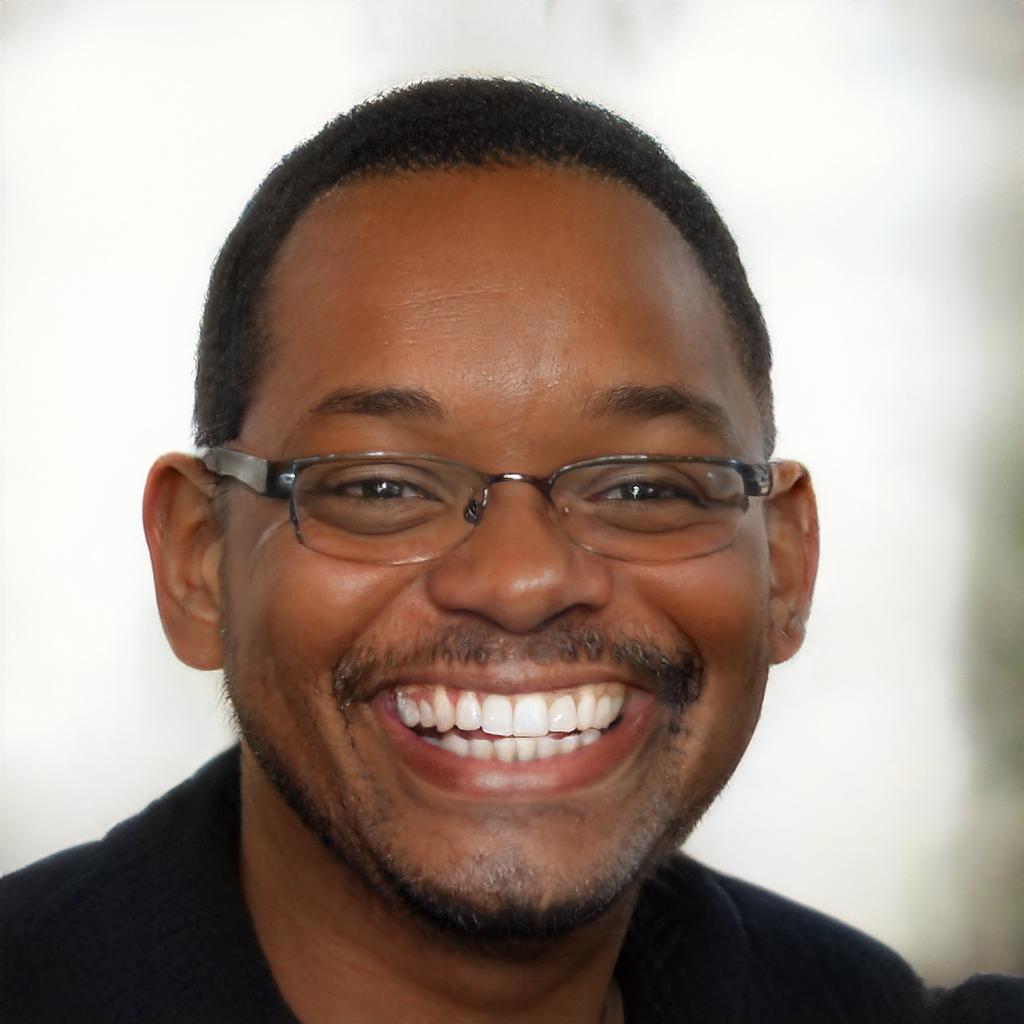} &
        \includegraphics[width=0.135\linewidth]{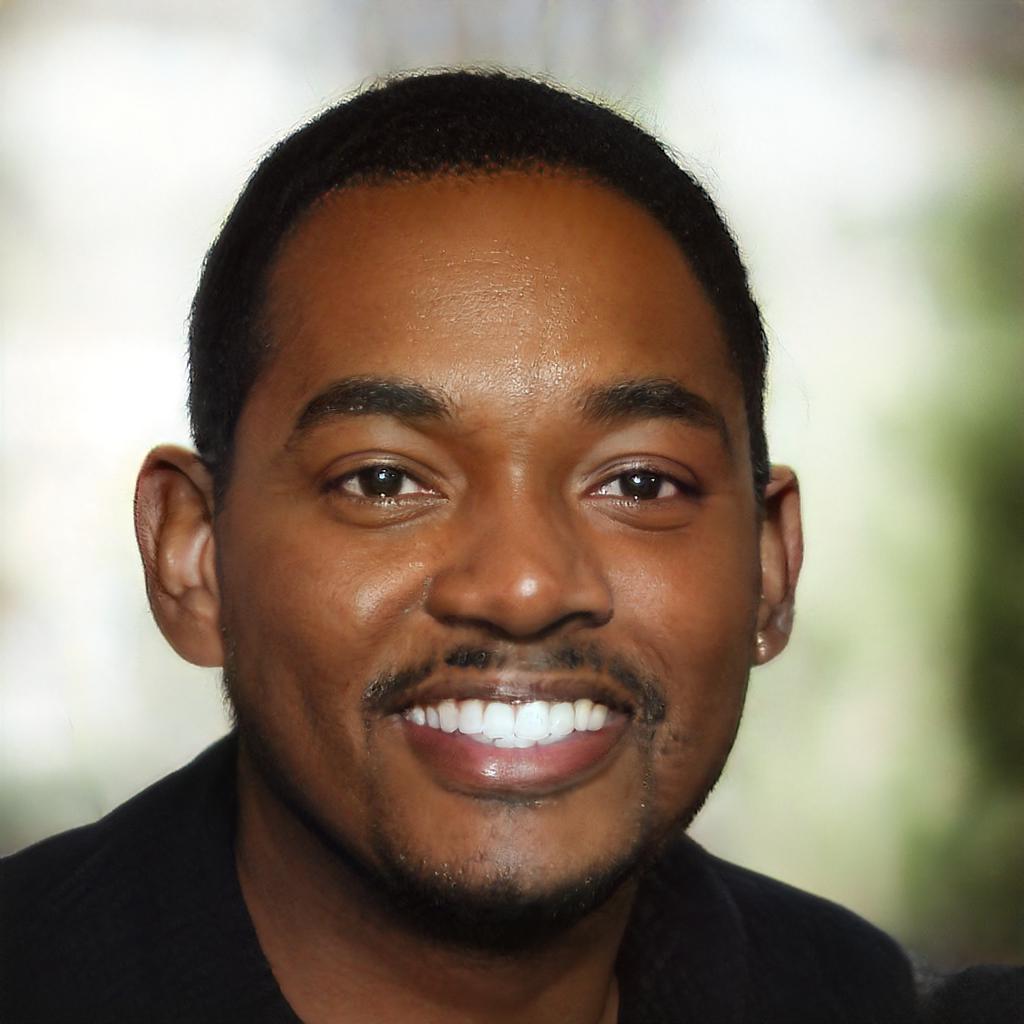} &
        \includegraphics[width=0.135\linewidth]{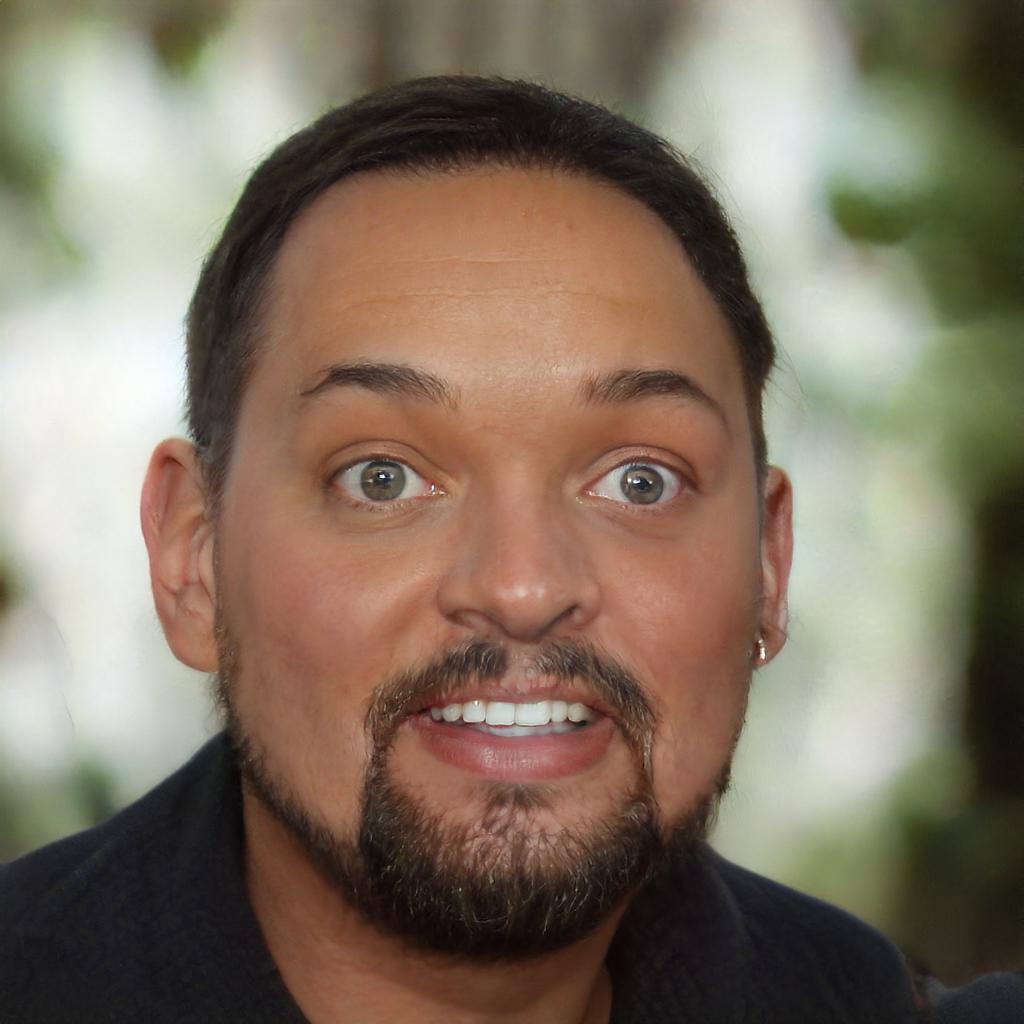} 
        \tabularnewline
        && \raisebox{0.35in}{\rotatebox[origin=t]{90}{\footnotesize StyleFusion}} &
        \includegraphics[width=0.135\linewidth]{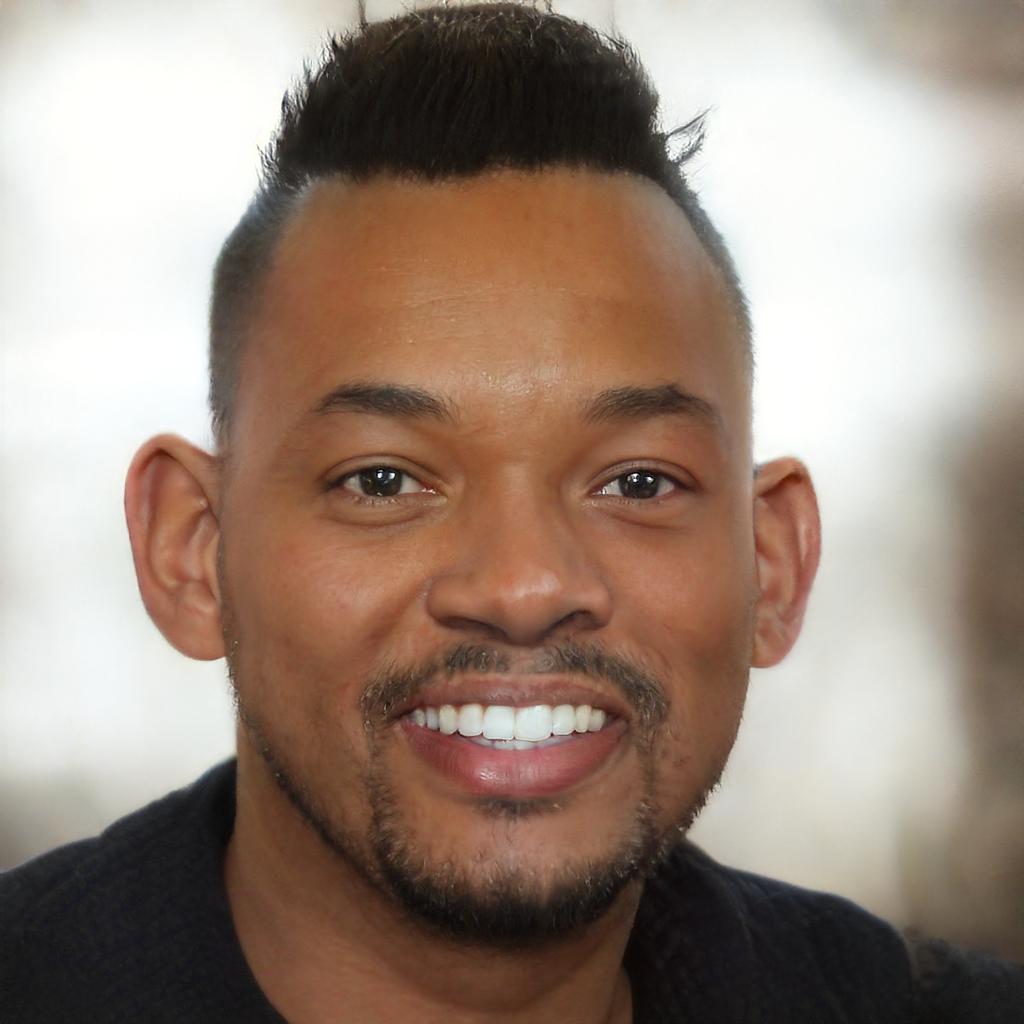} &
        \includegraphics[width=0.135\linewidth]{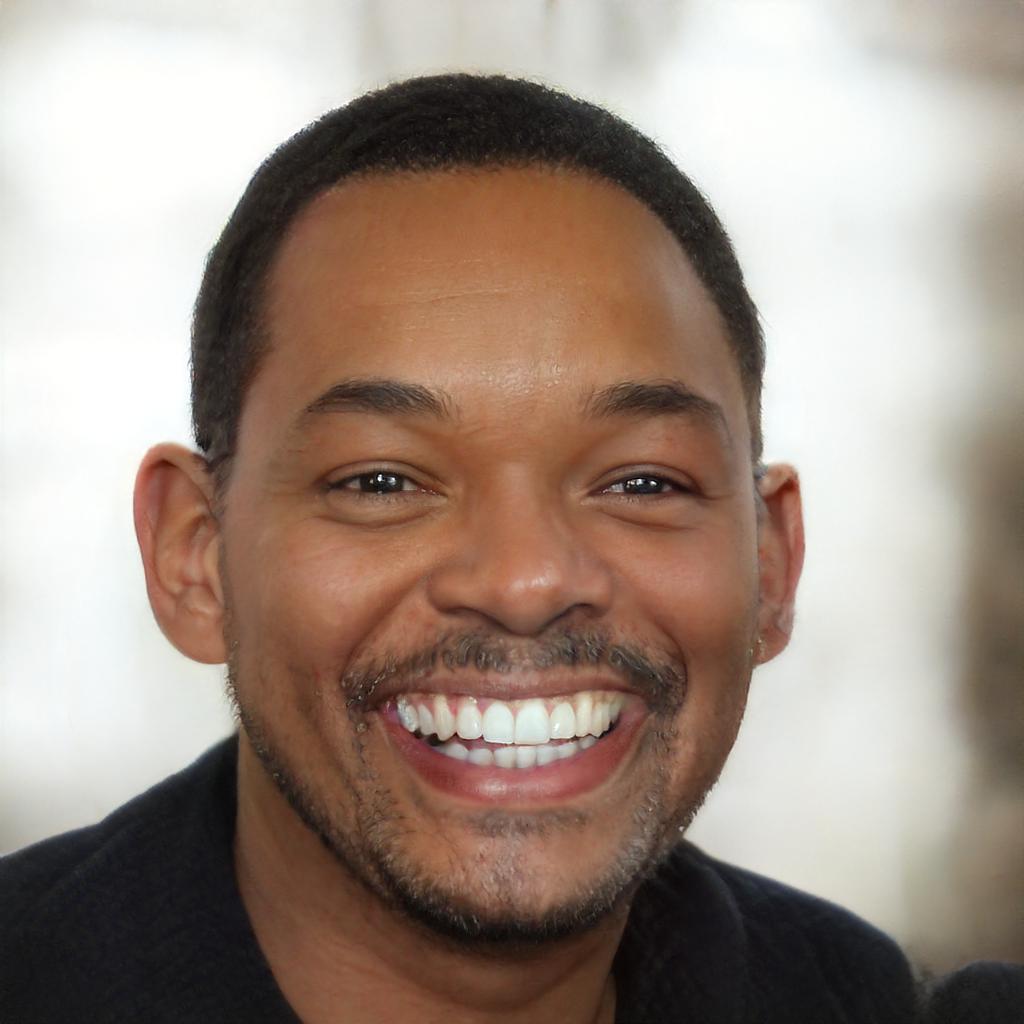} &
        \includegraphics[width=0.135\linewidth]{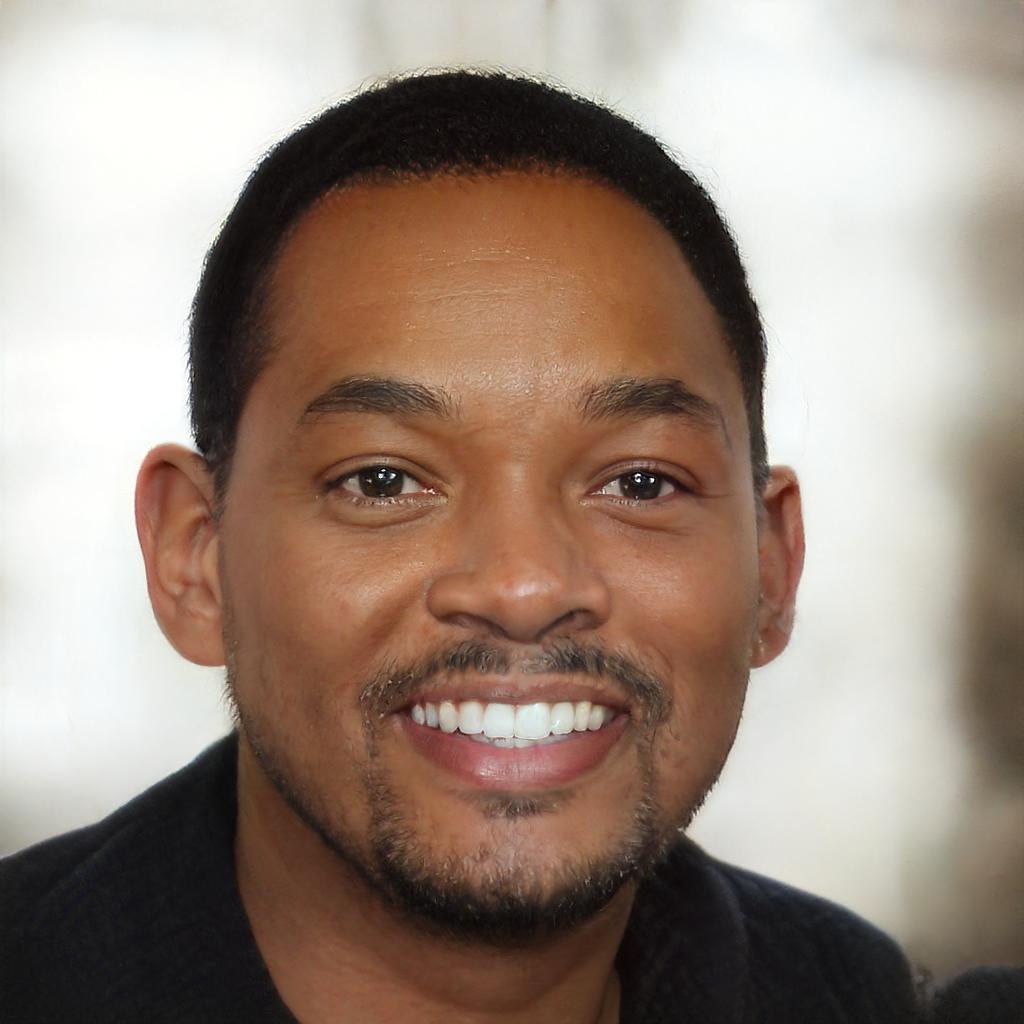} &
        \includegraphics[width=0.135\linewidth]{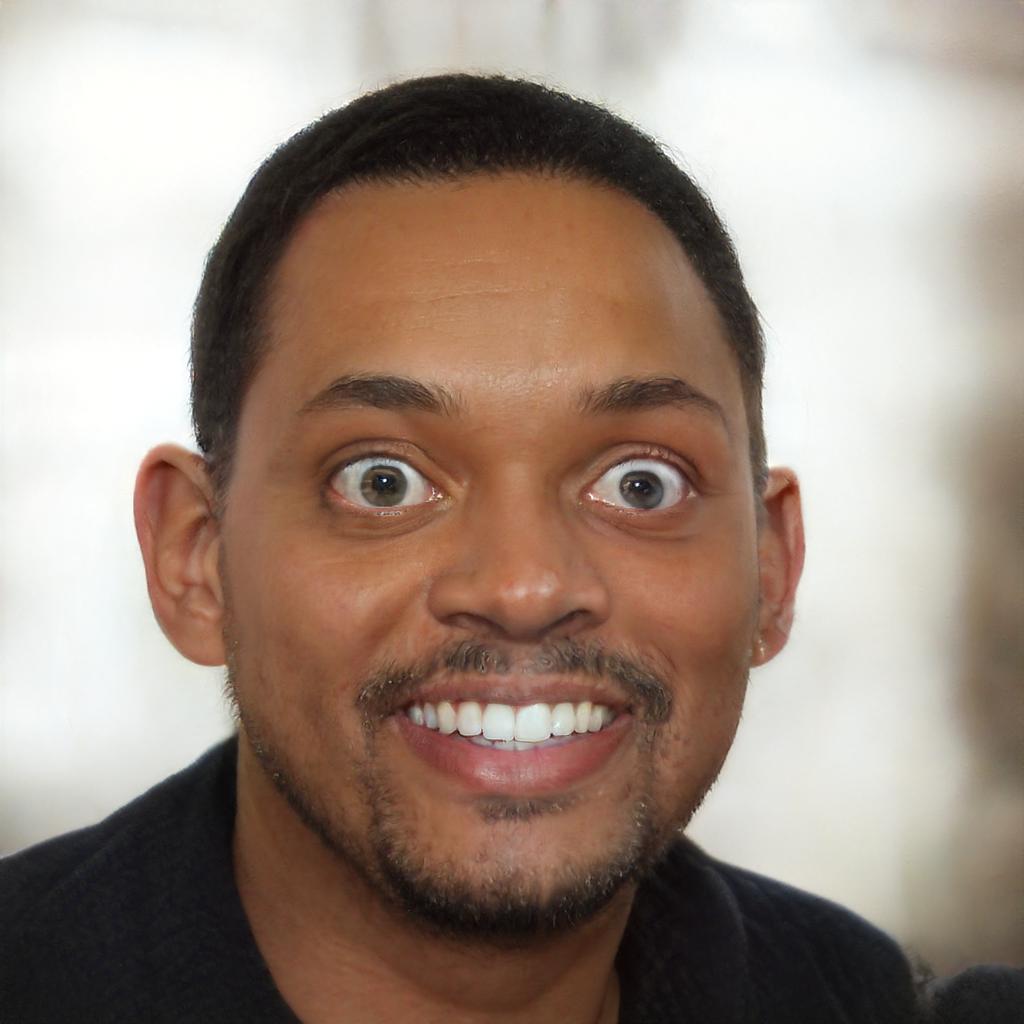} 
        \tabularnewline

        \includegraphics[width=0.135\linewidth]{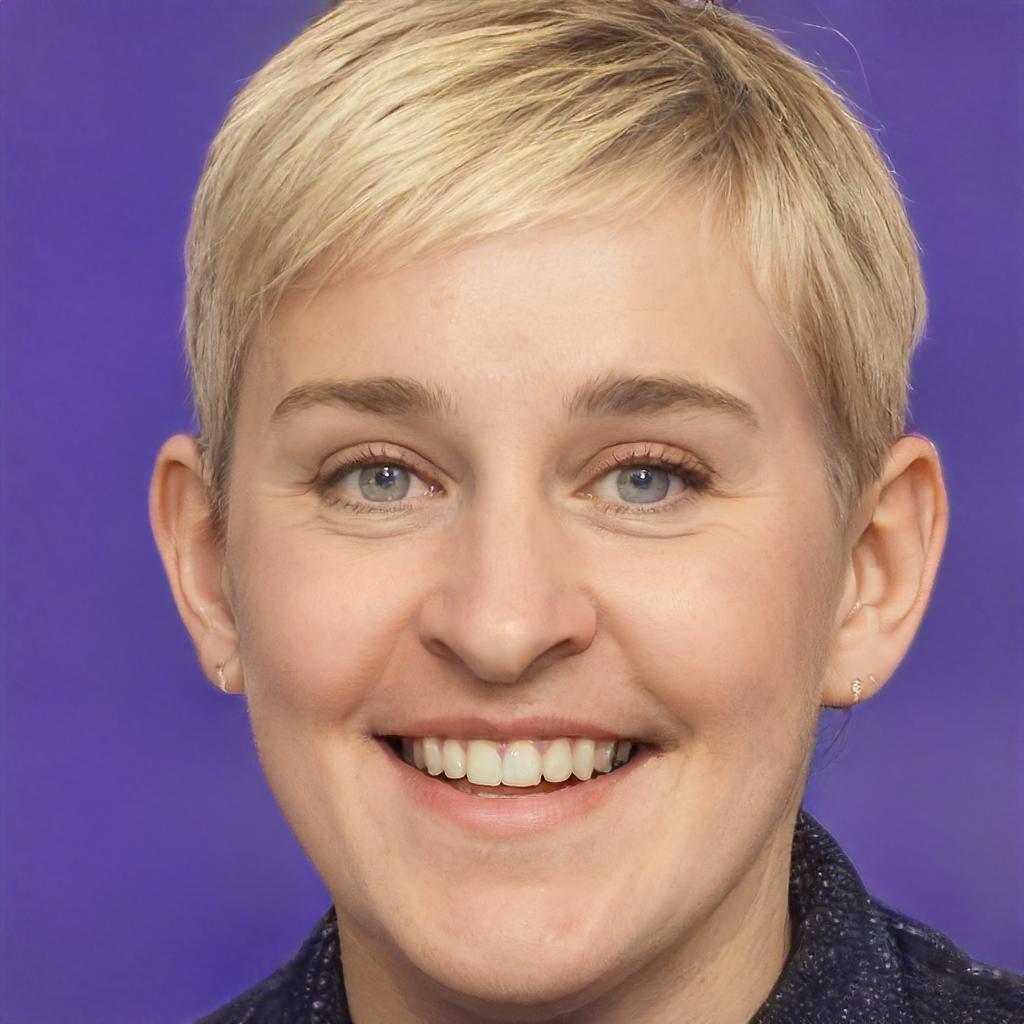} &&
        \raisebox{0.35in}{\rotatebox[origin=t]{90}{\footnotesize StyleGAN2}} &
        \includegraphics[width=0.135\linewidth]{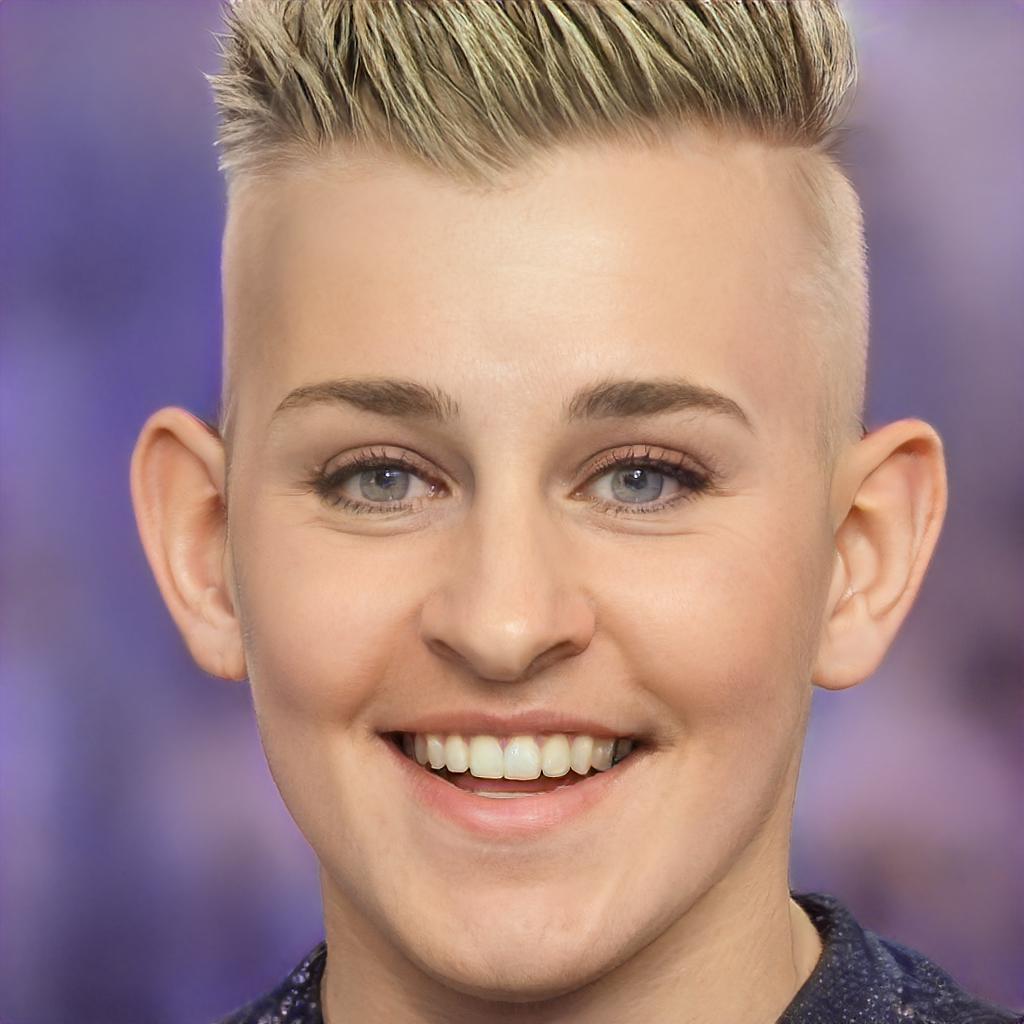} &
        \includegraphics[width=0.135\linewidth]{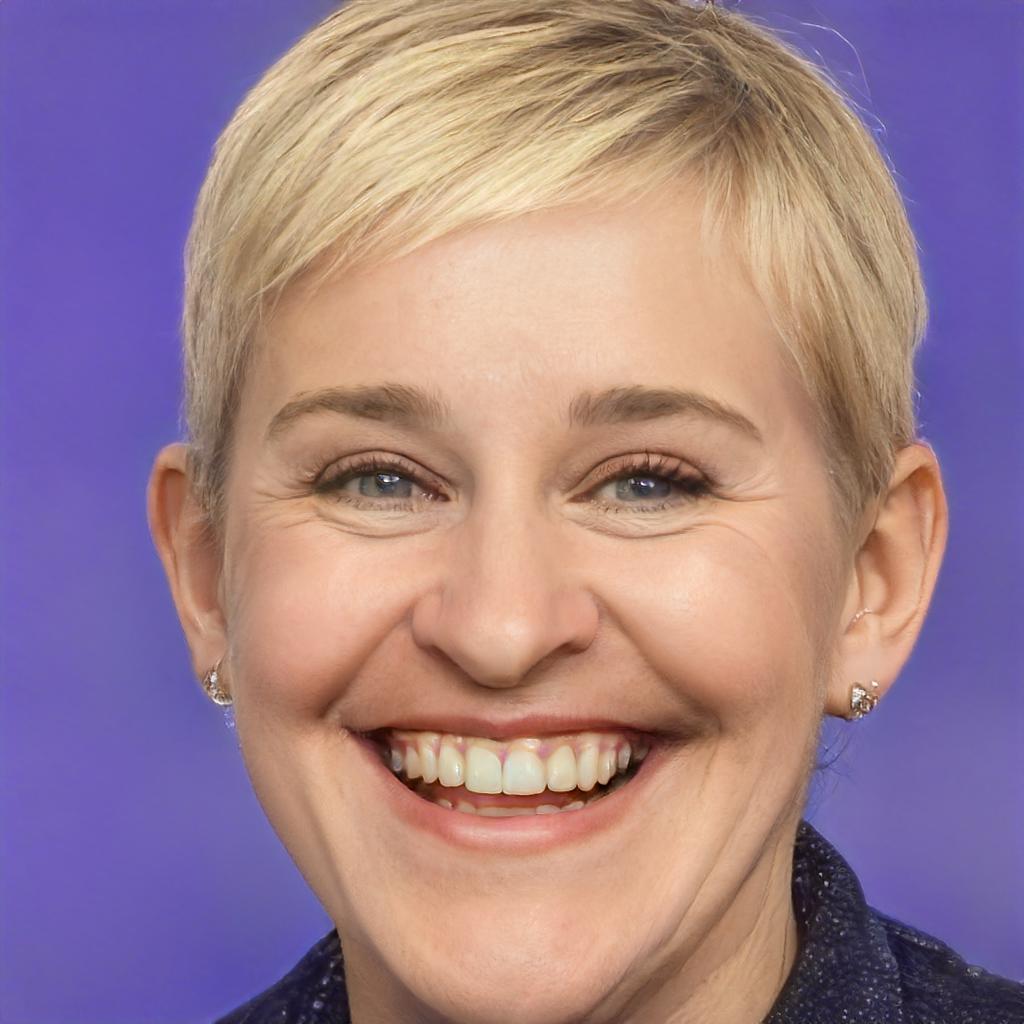} &
        \includegraphics[width=0.135\linewidth]{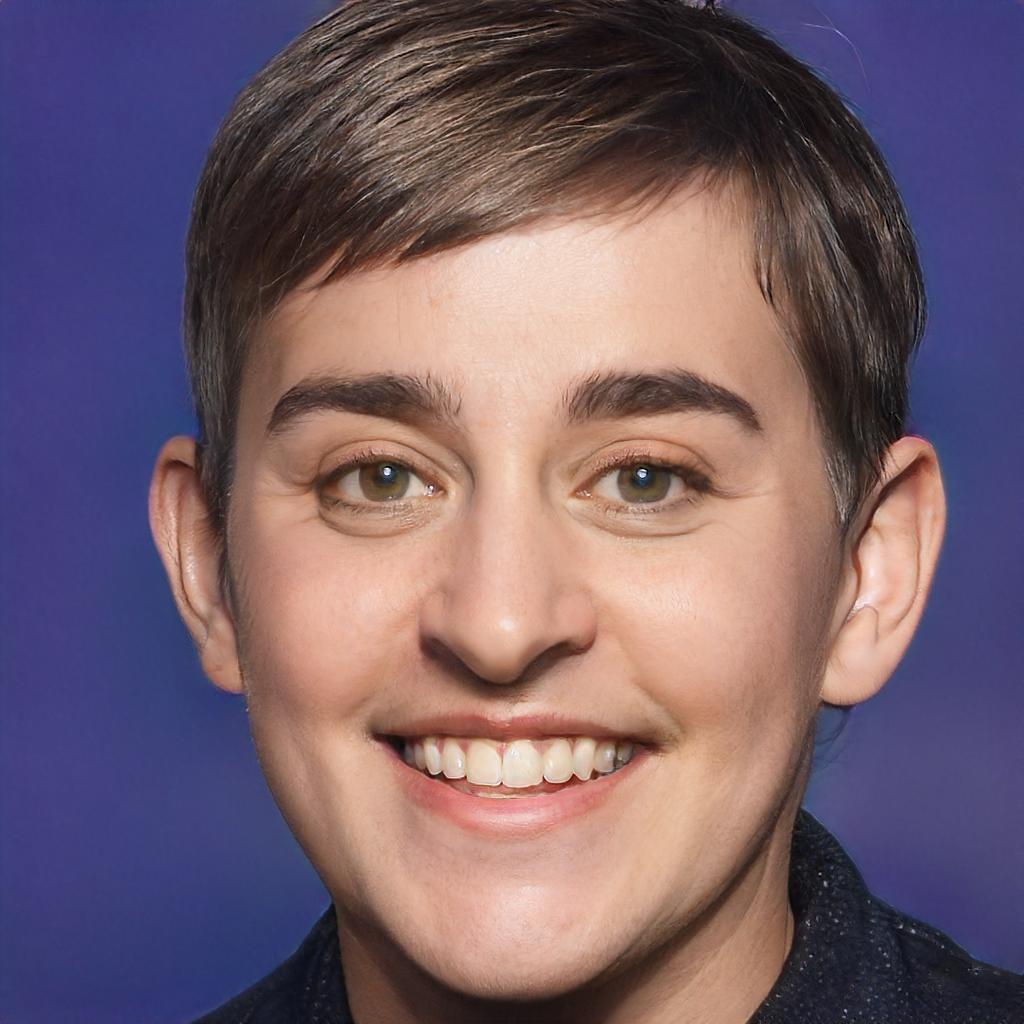} &
        \includegraphics[width=0.135\linewidth]{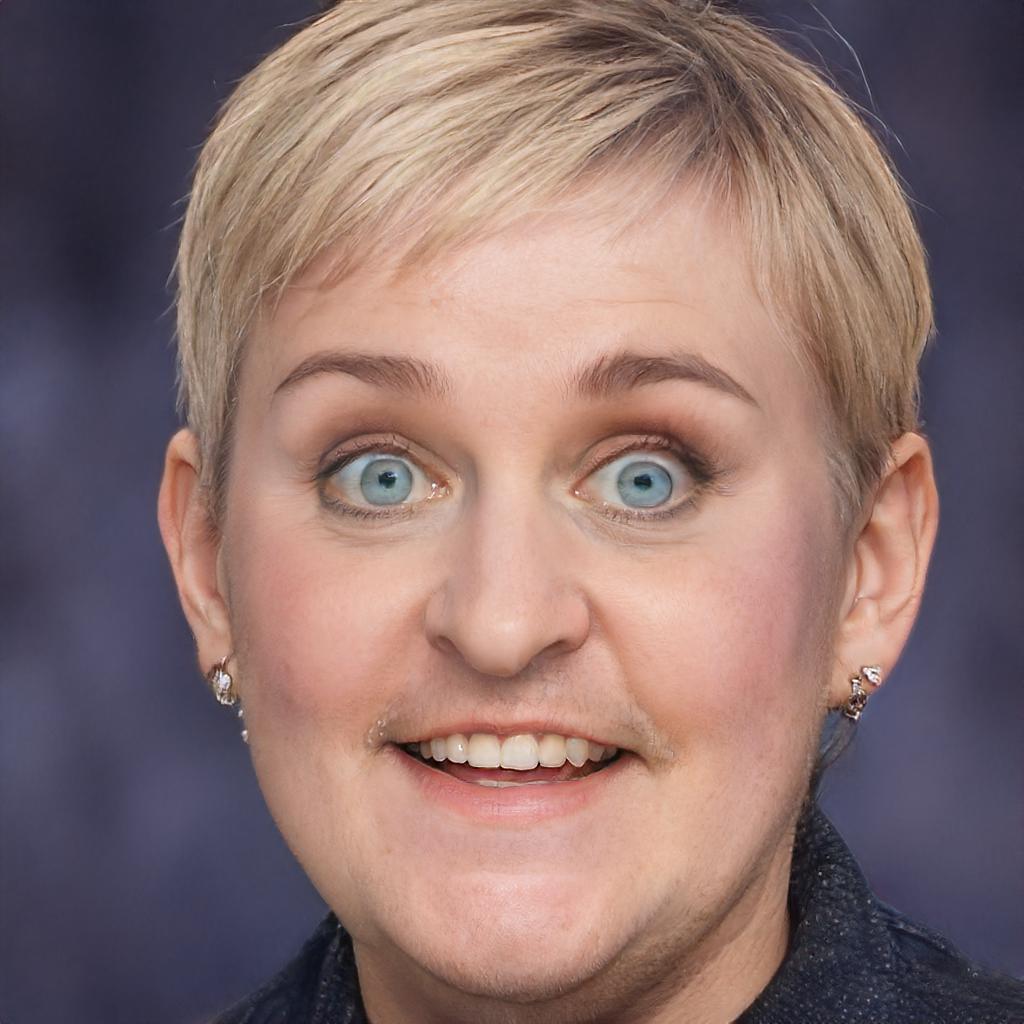} 
        \tabularnewline
        && \raisebox{0.35in}{\rotatebox[origin=t]{90}{\footnotesize StyleFusion}} &
        \includegraphics[width=0.135\linewidth]{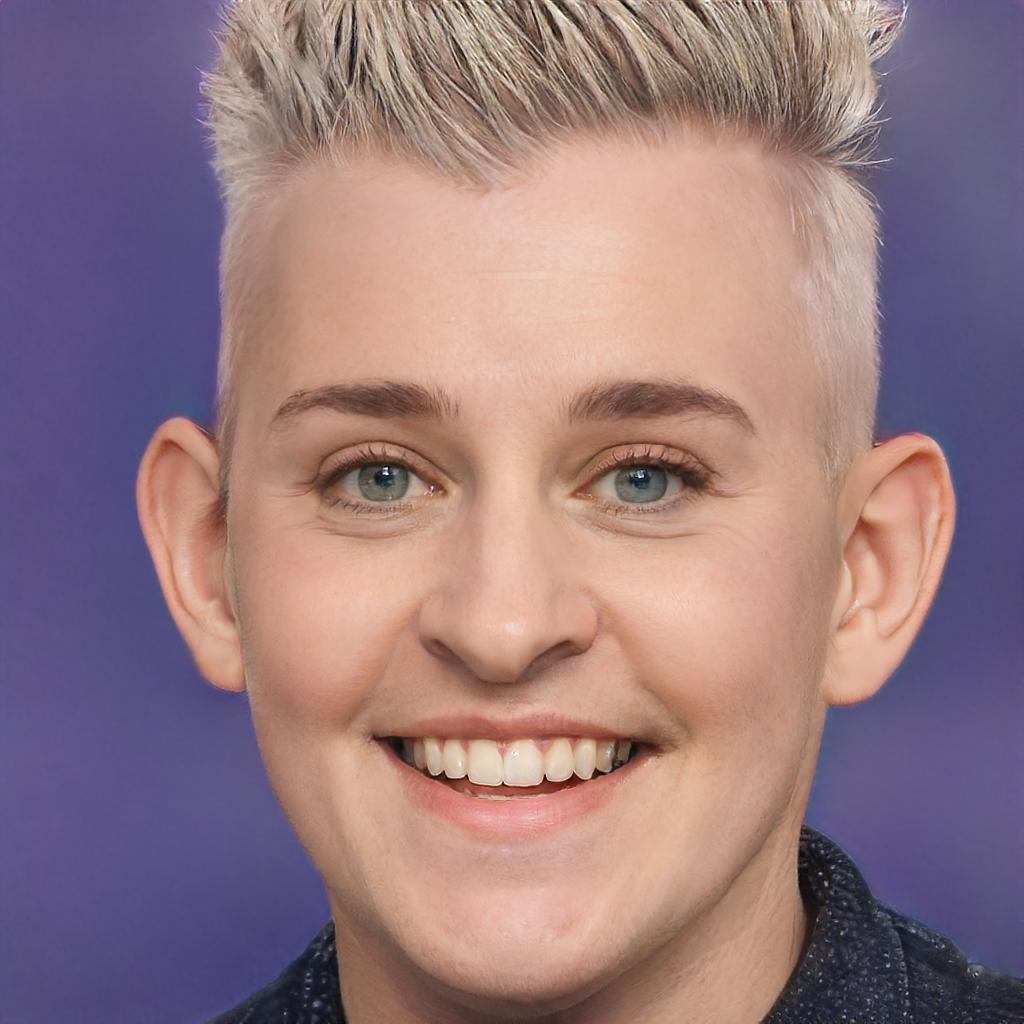} &
        \includegraphics[width=0.135\linewidth]{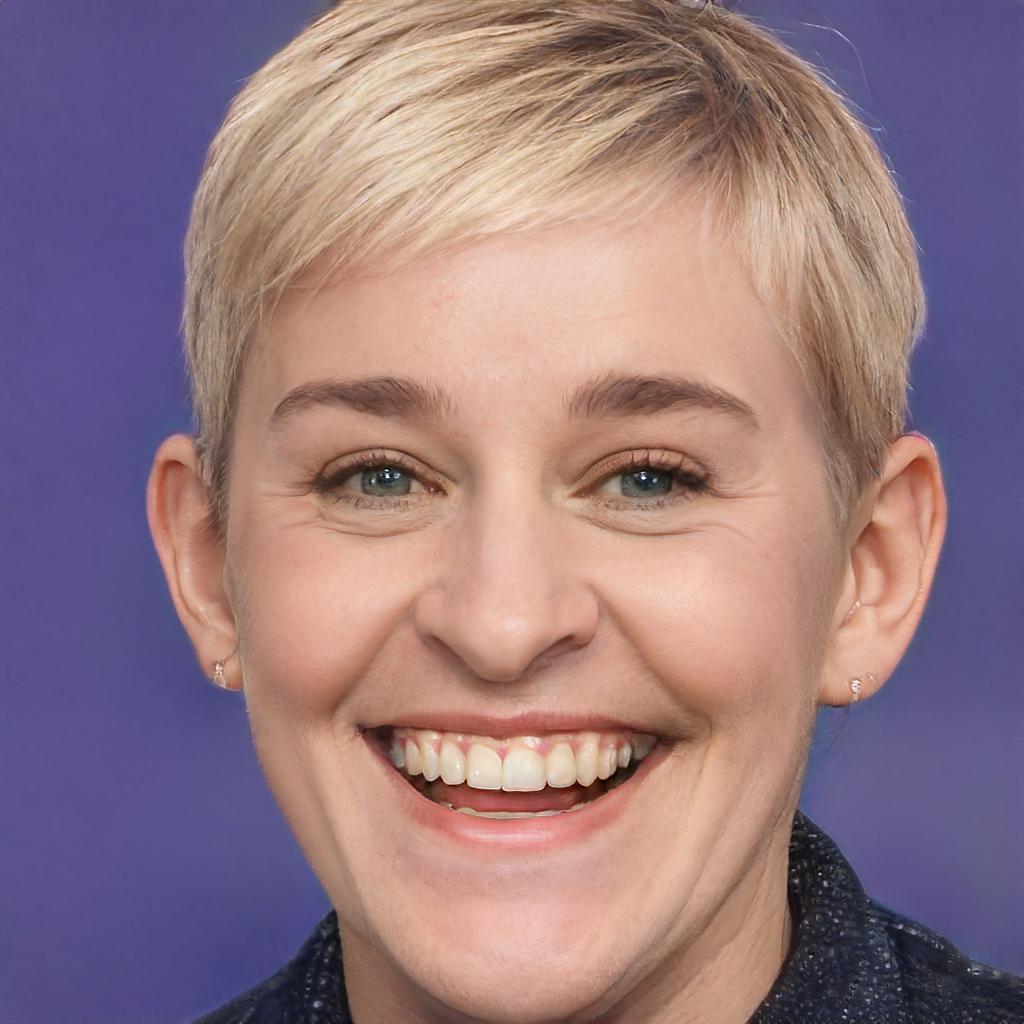} &
        \includegraphics[width=0.135\linewidth]{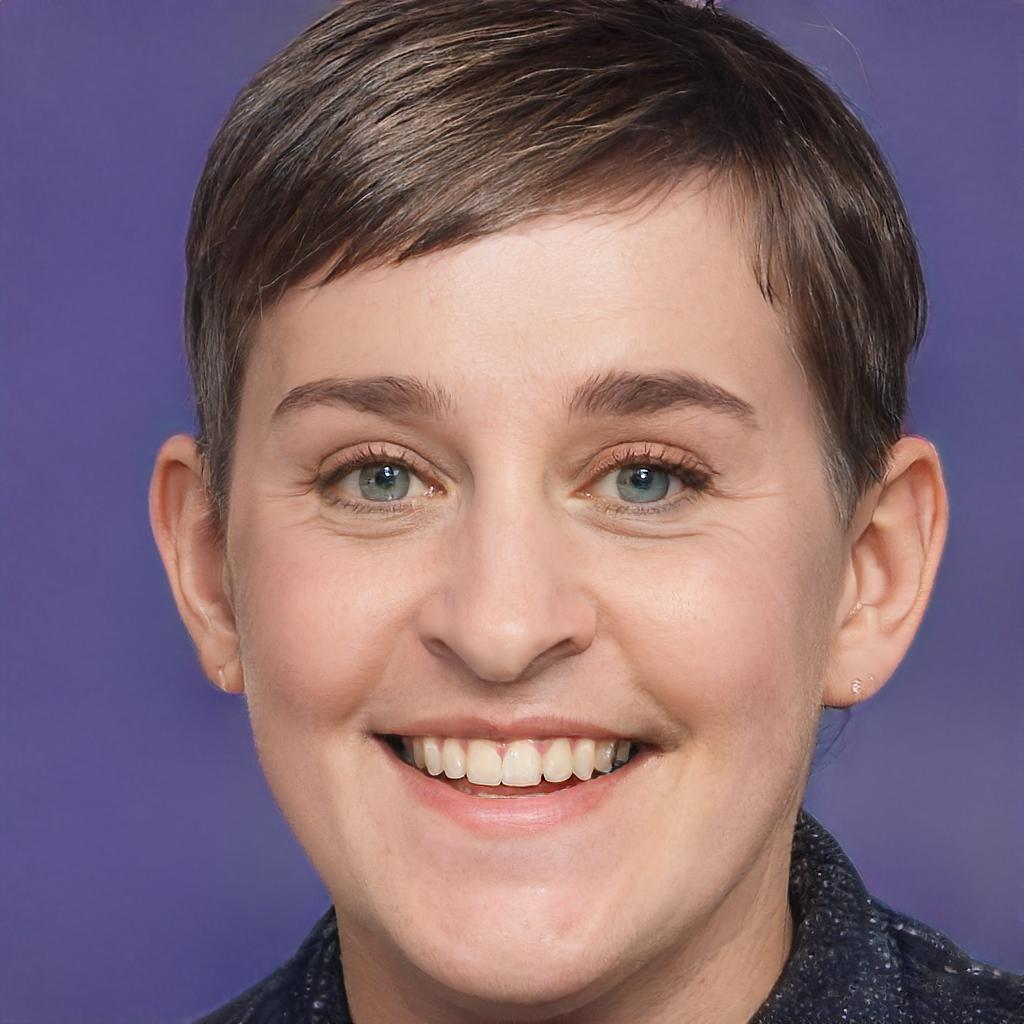} &
        \includegraphics[width=0.135\linewidth]{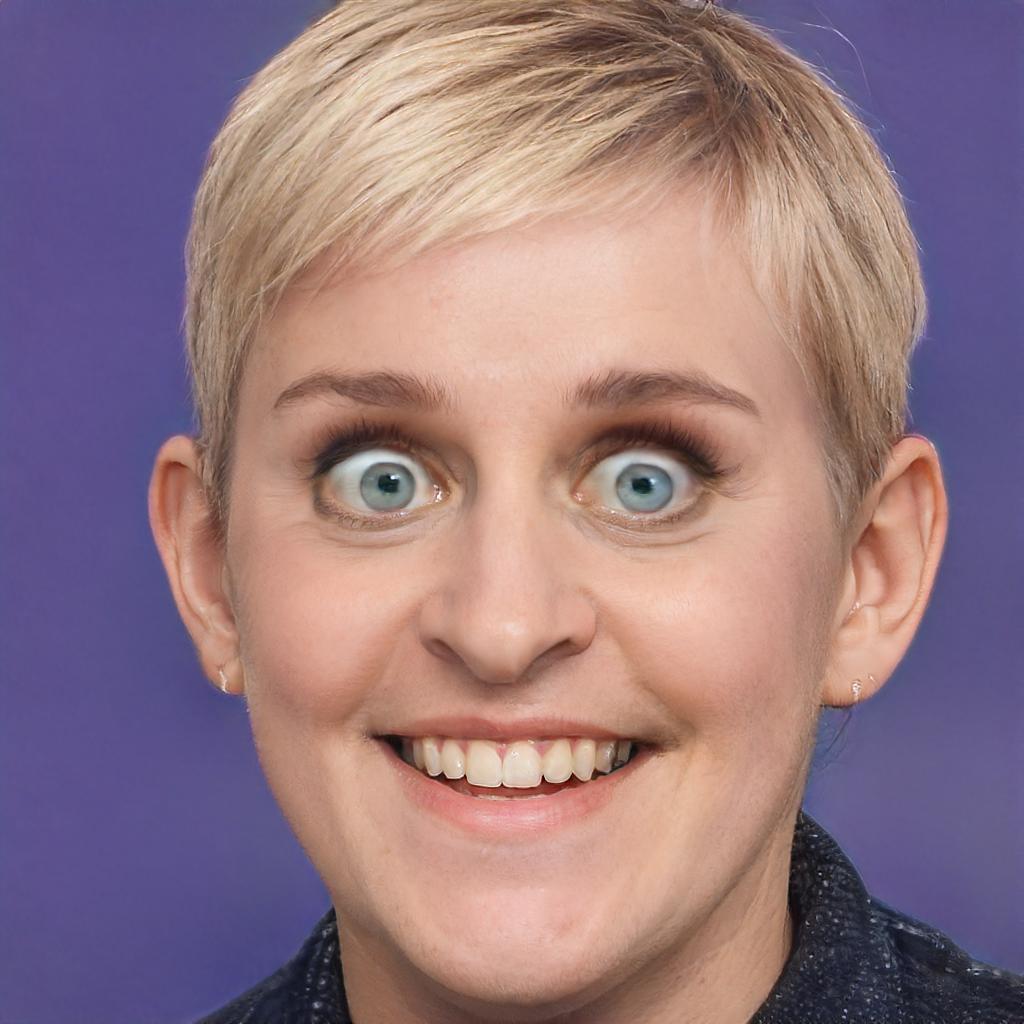} 
        \tabularnewline

        \includegraphics[width=0.135\linewidth]{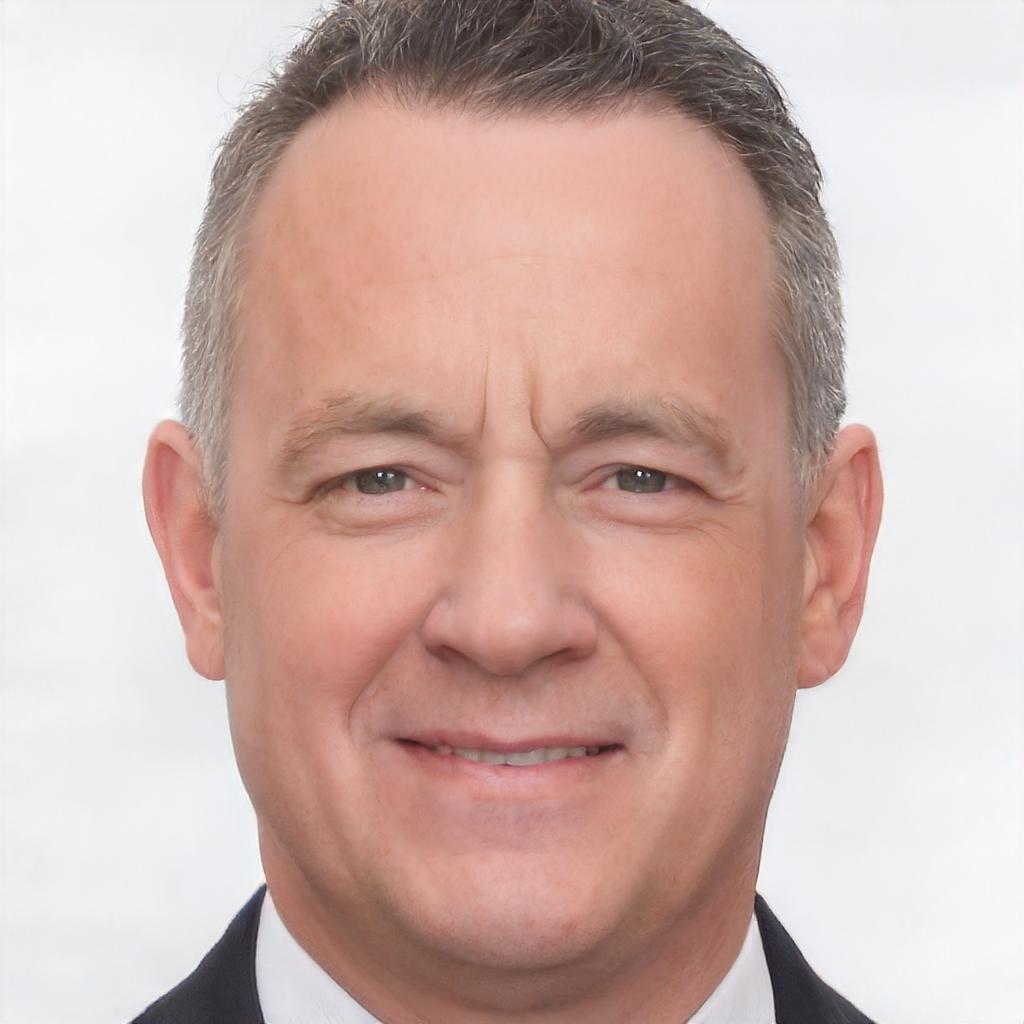} &&
        \raisebox{0.35in}{\rotatebox[origin=t]{90}{\footnotesize StyleGAN2}} &
        \includegraphics[width=0.135\linewidth]{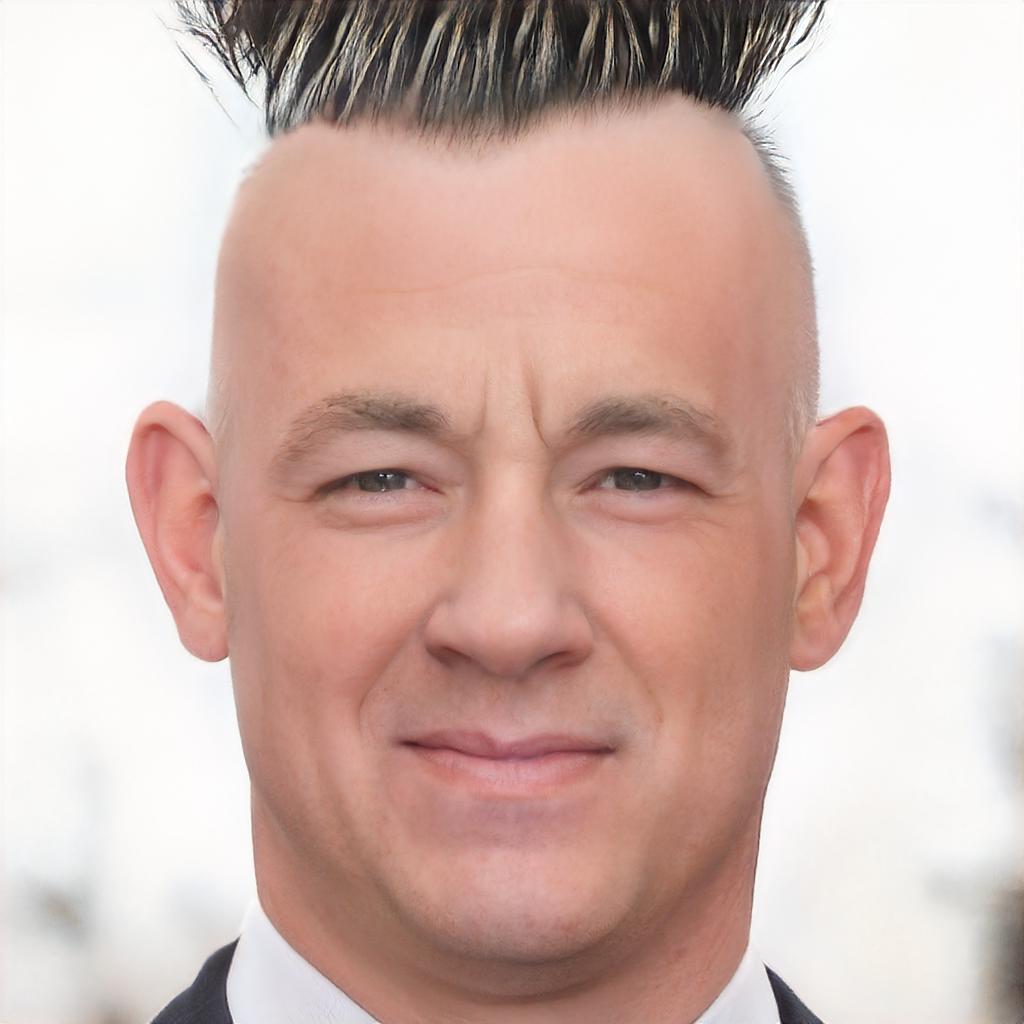} &
        \includegraphics[width=0.135\linewidth]{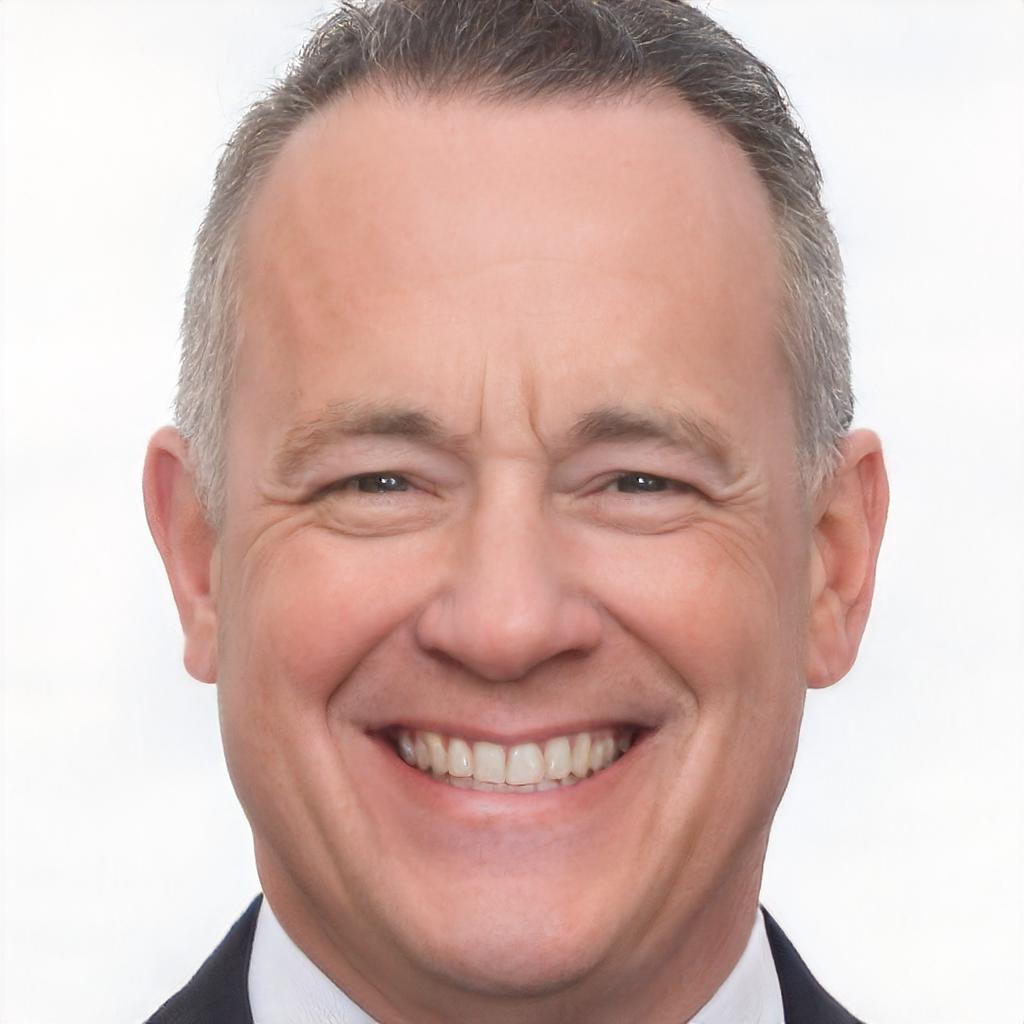} &
        \includegraphics[width=0.135\linewidth]{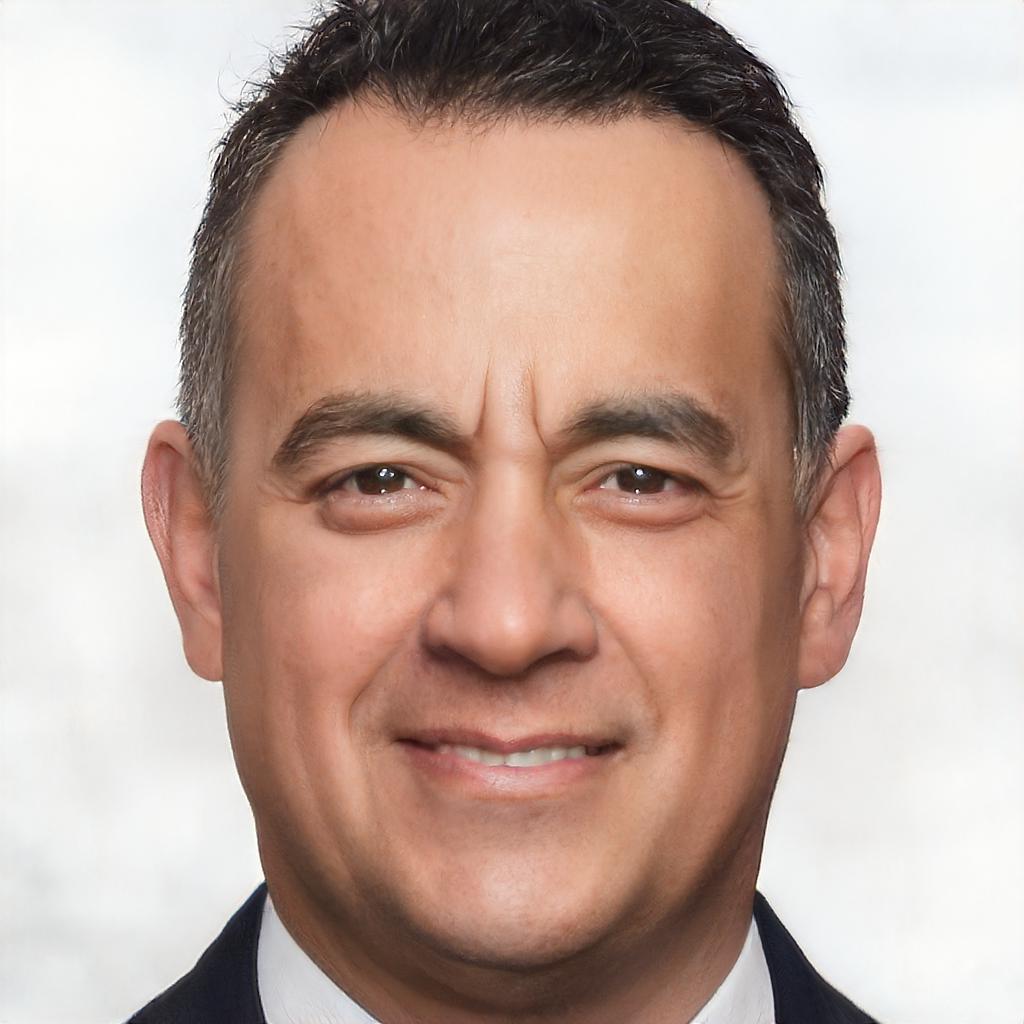} &
        \includegraphics[width=0.135\linewidth]{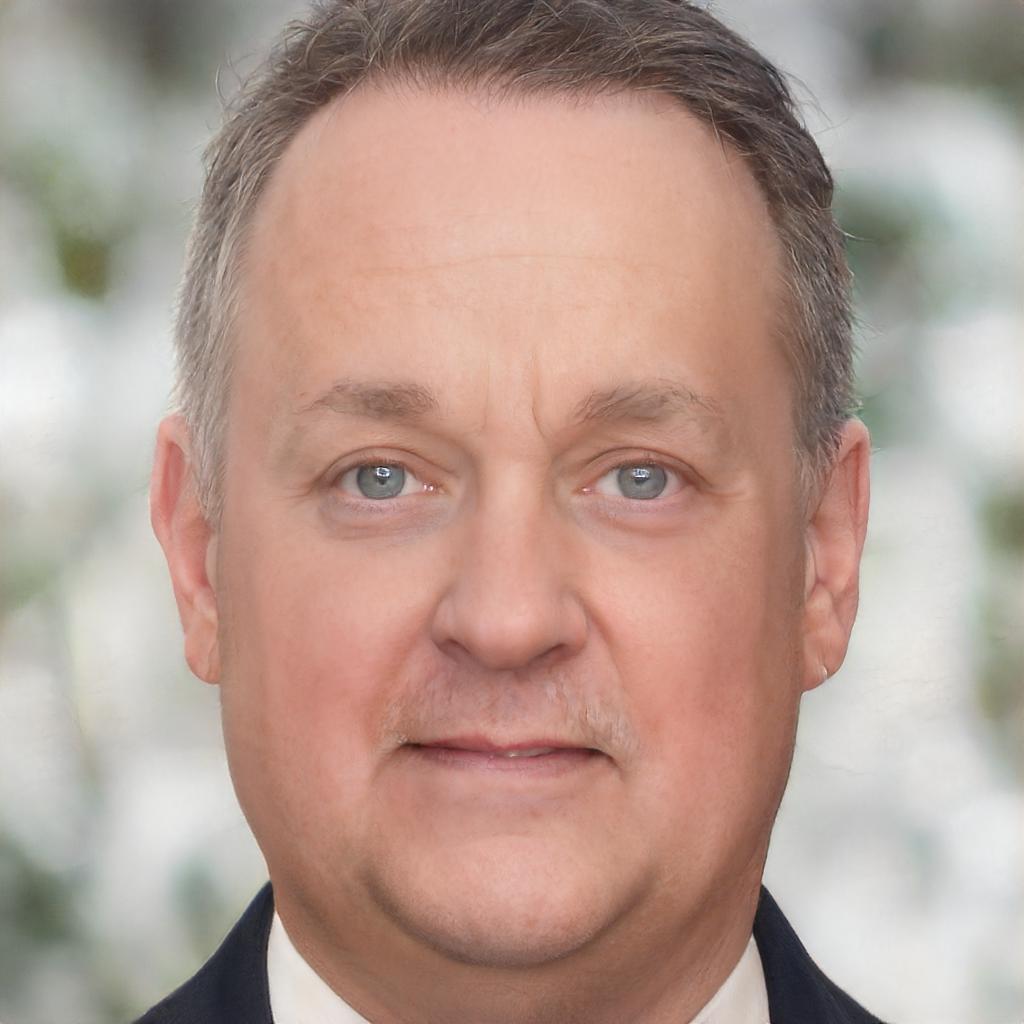} 
        \tabularnewline
        && \raisebox{0.35in}{\rotatebox[origin=t]{90}{\footnotesize StyleFusion}} &
        \includegraphics[width=0.135\linewidth]{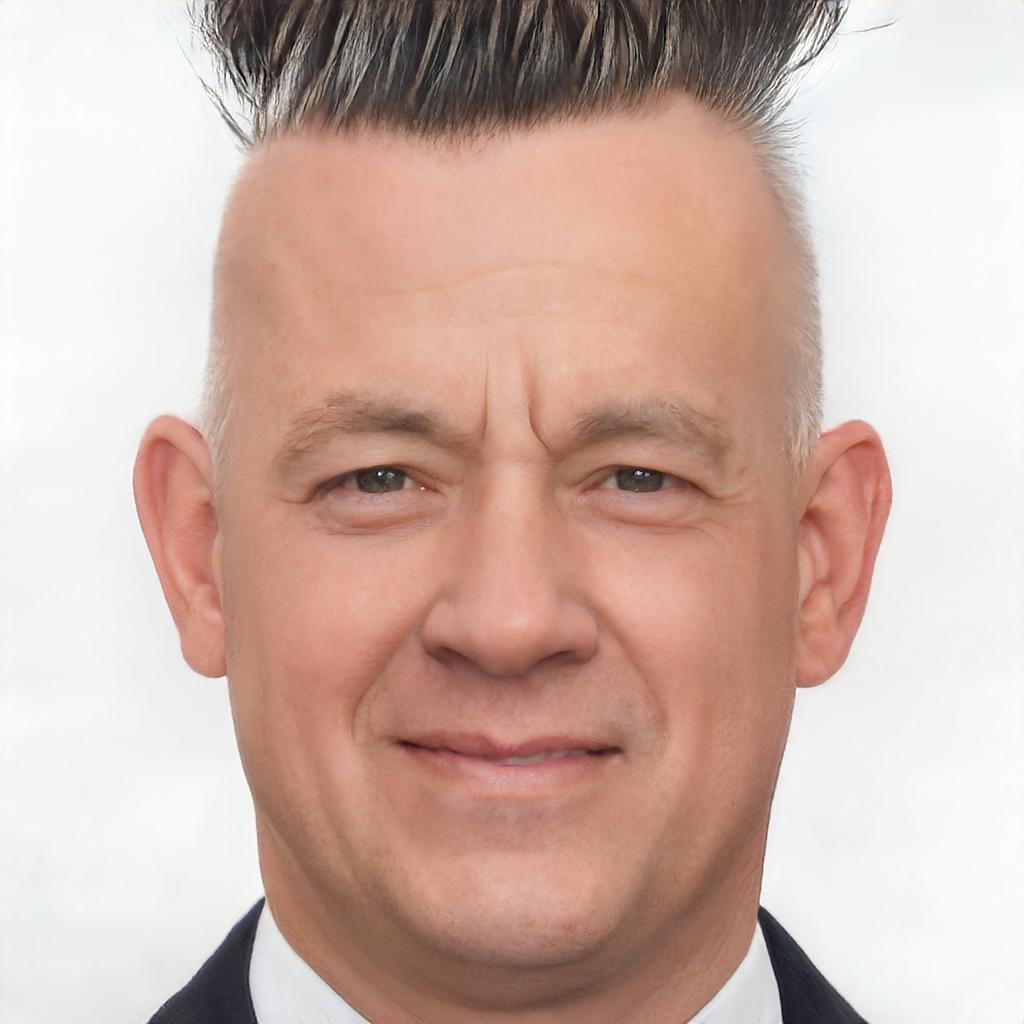} &
        \includegraphics[width=0.135\linewidth]{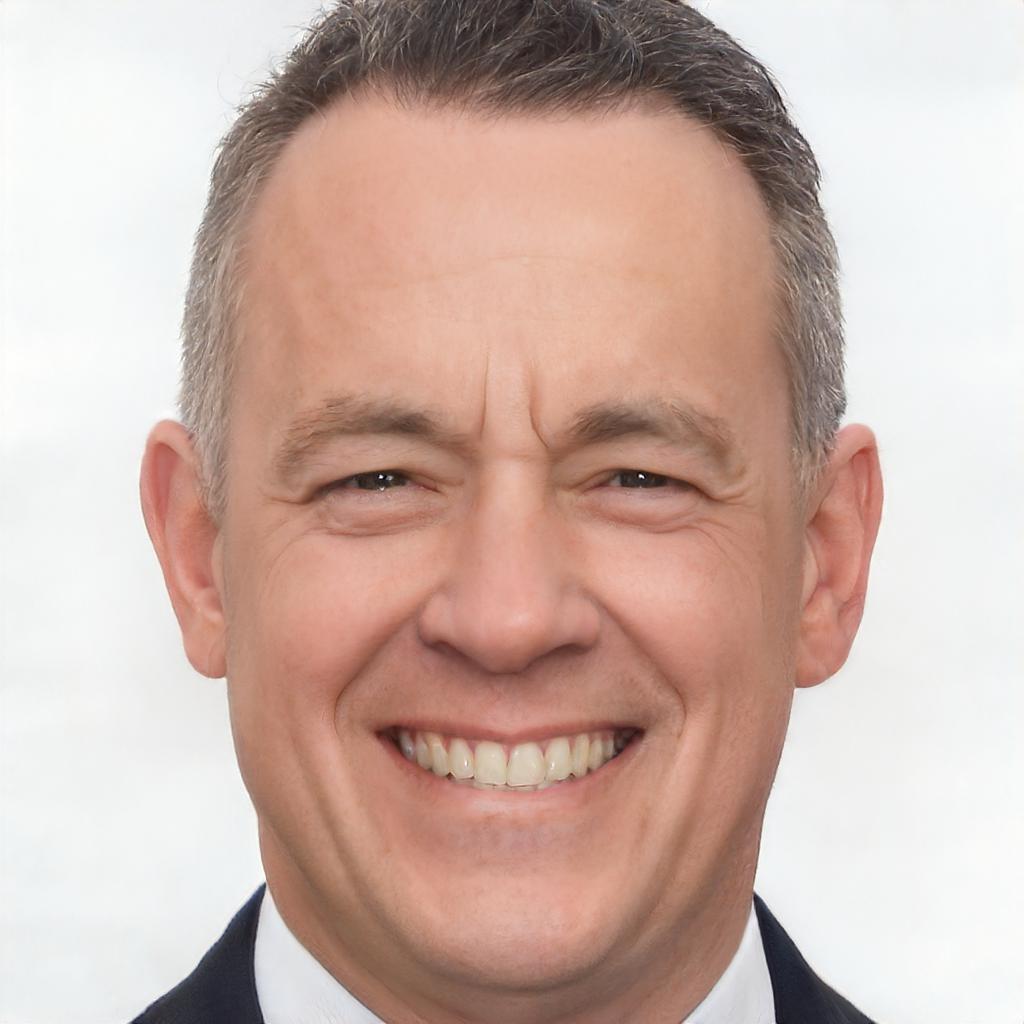} &
        \includegraphics[width=0.135\linewidth]{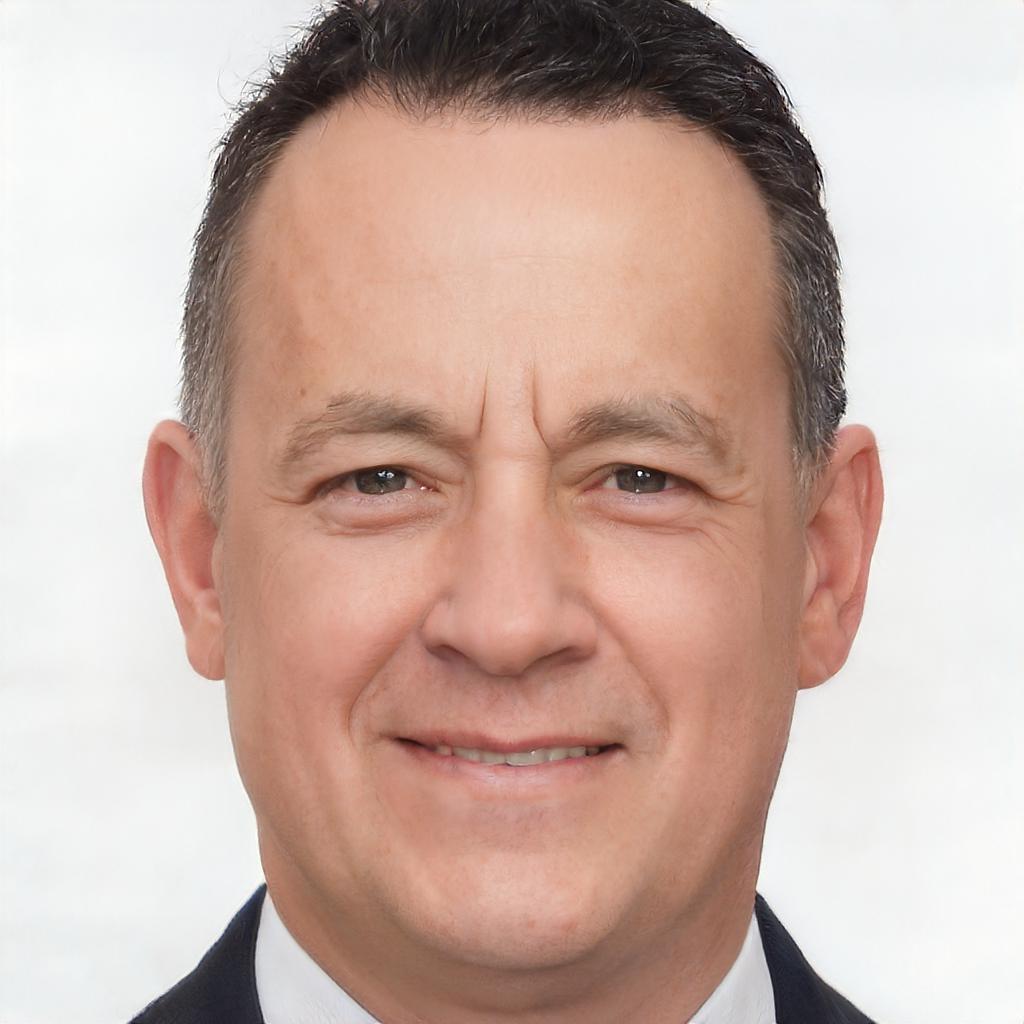} &
        \includegraphics[width=0.135\linewidth]{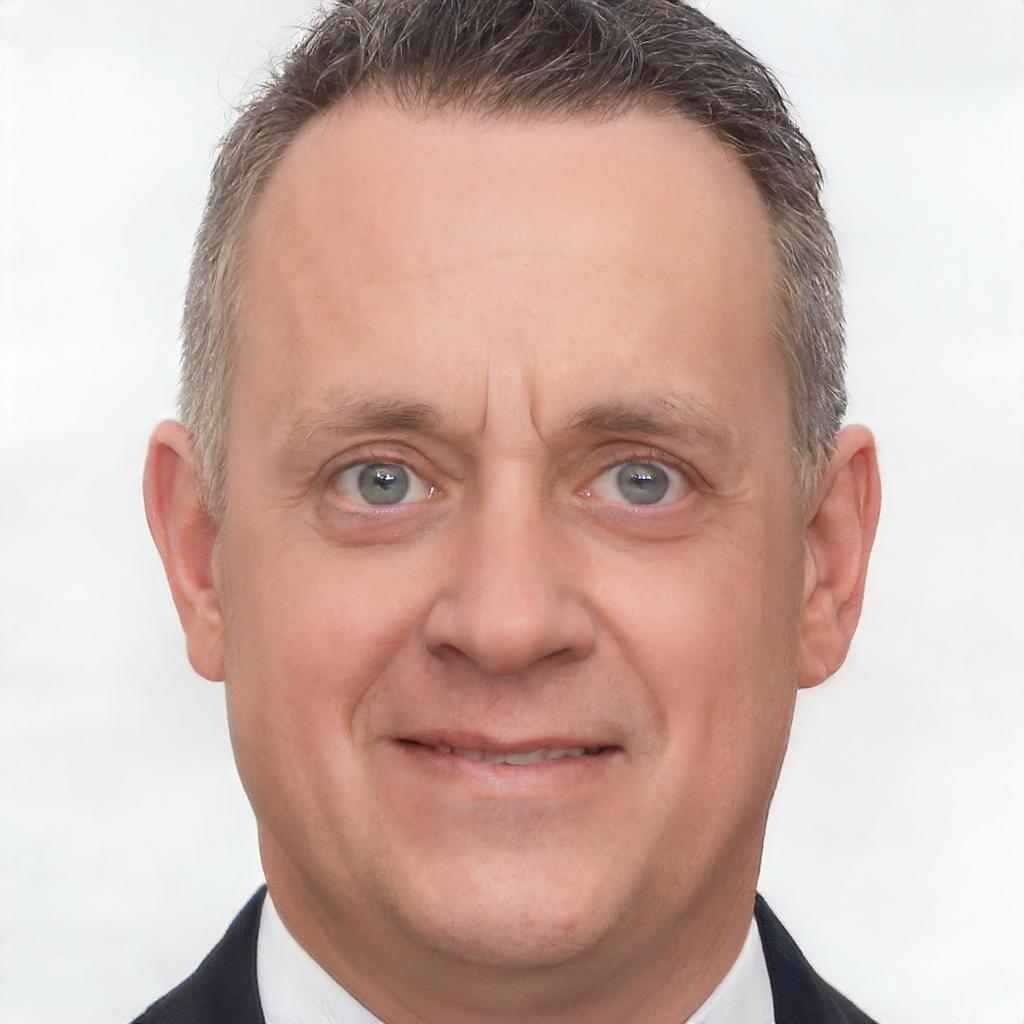} 
        \tabularnewline

        Input &&& Mohawk & Smile & Hair Albedo & Fearful Eyes
        \tabularnewline
        \end{tabular}
        }
    \vspace{0.1cm}
    \caption{Latent traversal editing techniques performed on real images with and without StyleFusion.}
    \label{fig:latent_editing_appendix_2}
\end{figure*}

\begin{figure*}
    \setlength{\tabcolsep}{1.5pt}
    \centering
    \small{
        \begin{tabular}{c c c c c c c}
        
        &&& StyleCLIP & InterFace & GANSpace & GANSpace \\
        
        \includegraphics[width=0.135\linewidth]{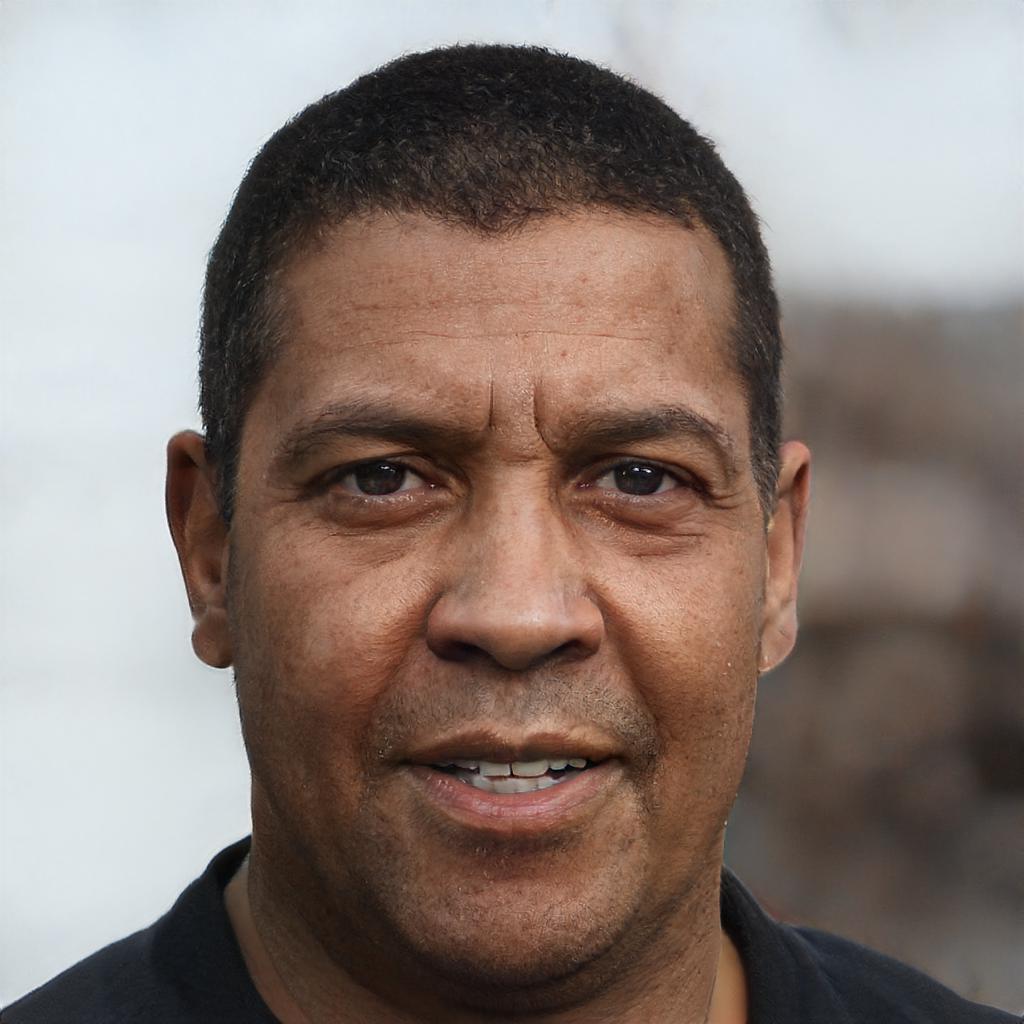} &&
        \raisebox{0.35in}{\rotatebox[origin=t]{90}{\footnotesize StyleGAN2}} &
        \includegraphics[width=0.135\linewidth]{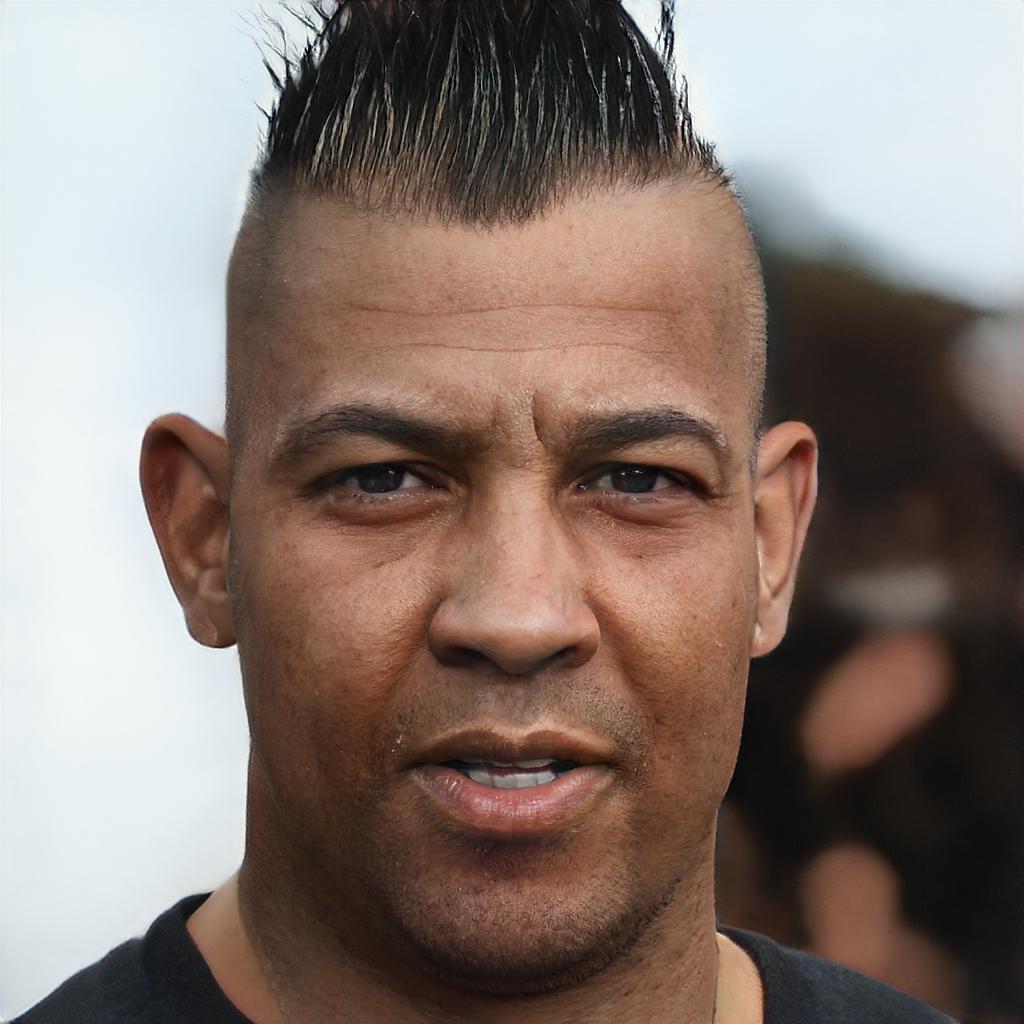} &
        \includegraphics[width=0.135\linewidth]{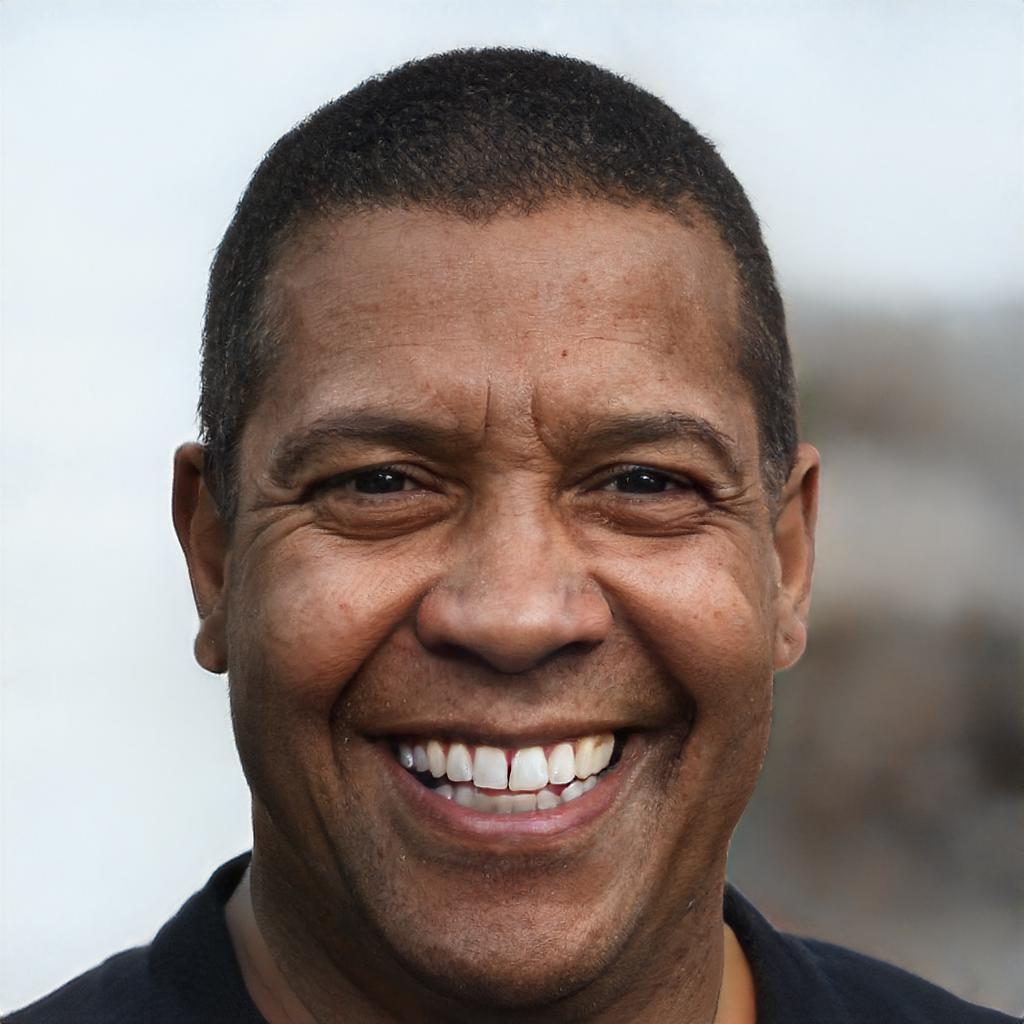} &
        \includegraphics[width=0.135\linewidth]{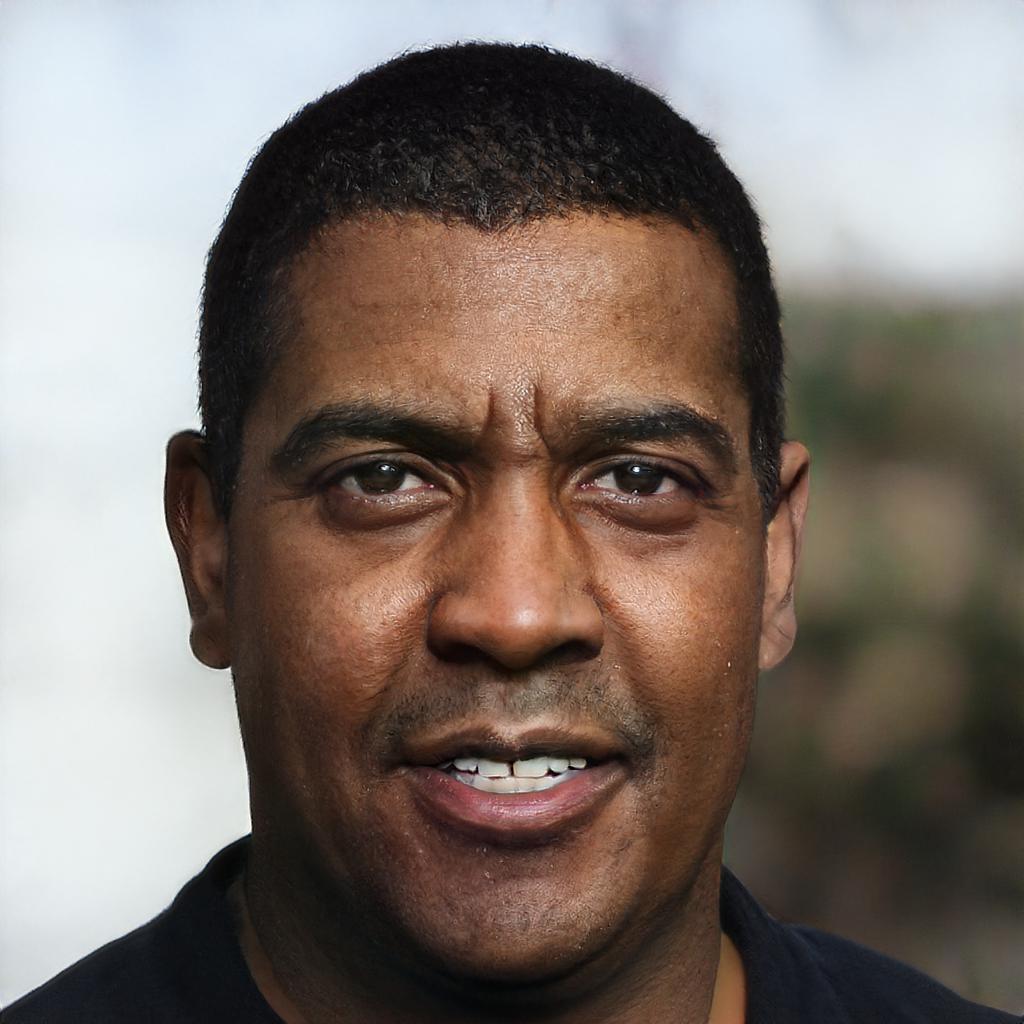} &
        \includegraphics[width=0.135\linewidth]{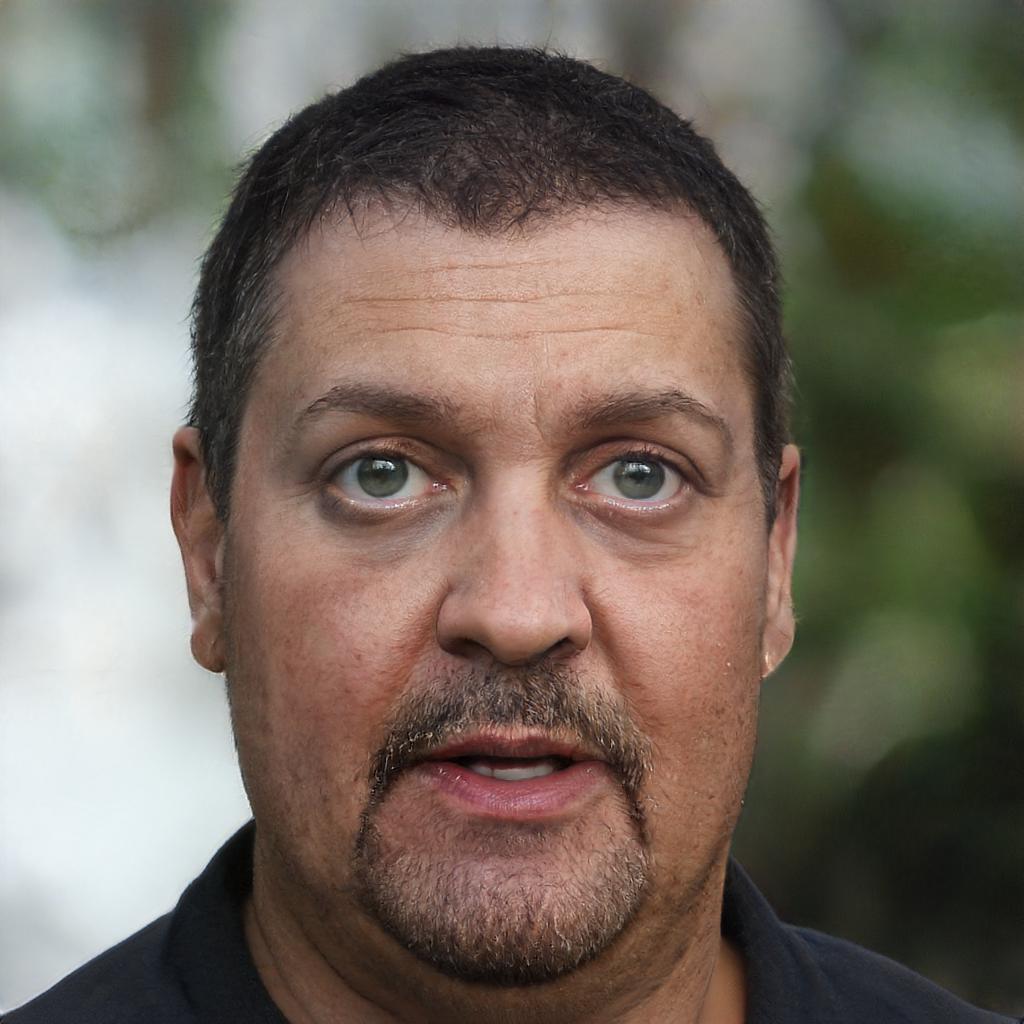} 
        \tabularnewline
        && \raisebox{0.35in}{\rotatebox[origin=t]{90}{\footnotesize StyleFusion}} &
        \includegraphics[width=0.135\linewidth]{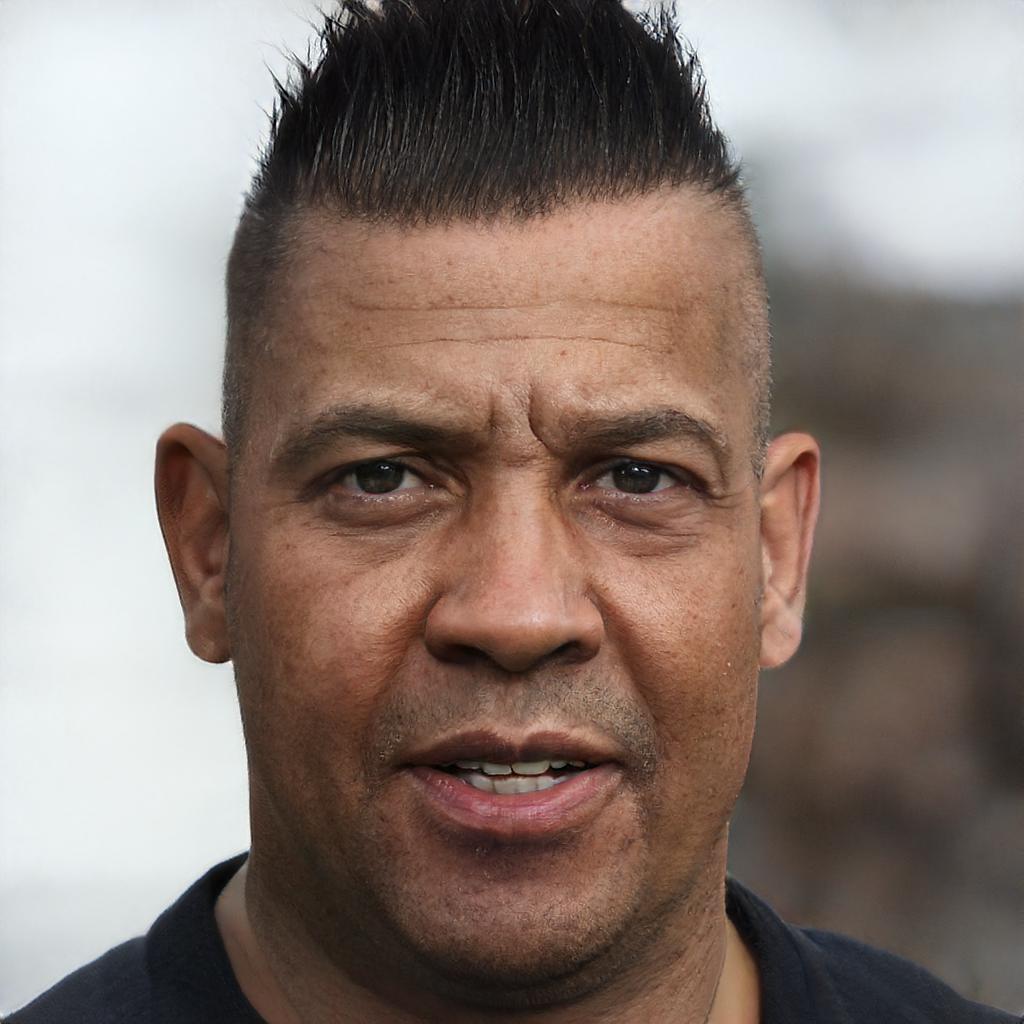} &
        \includegraphics[width=0.135\linewidth]{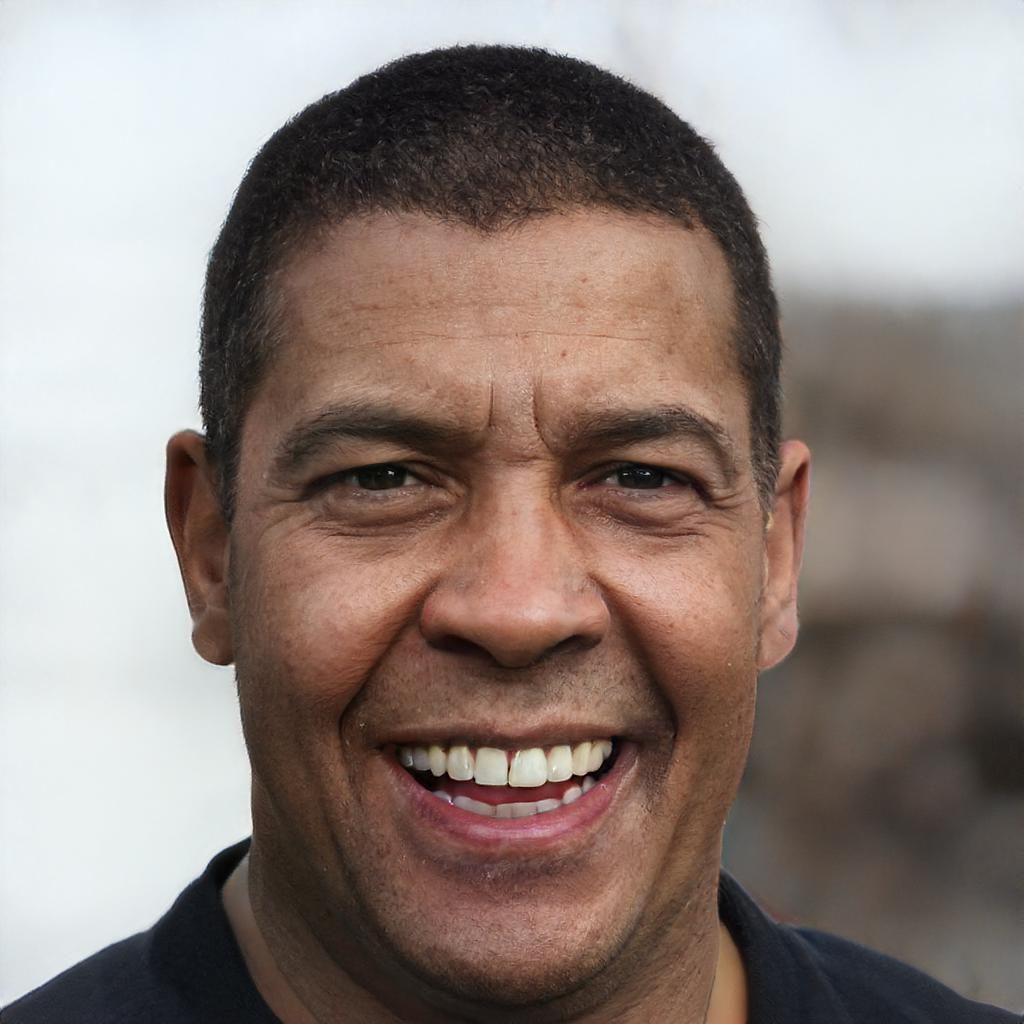} &
        \includegraphics[width=0.135\linewidth]{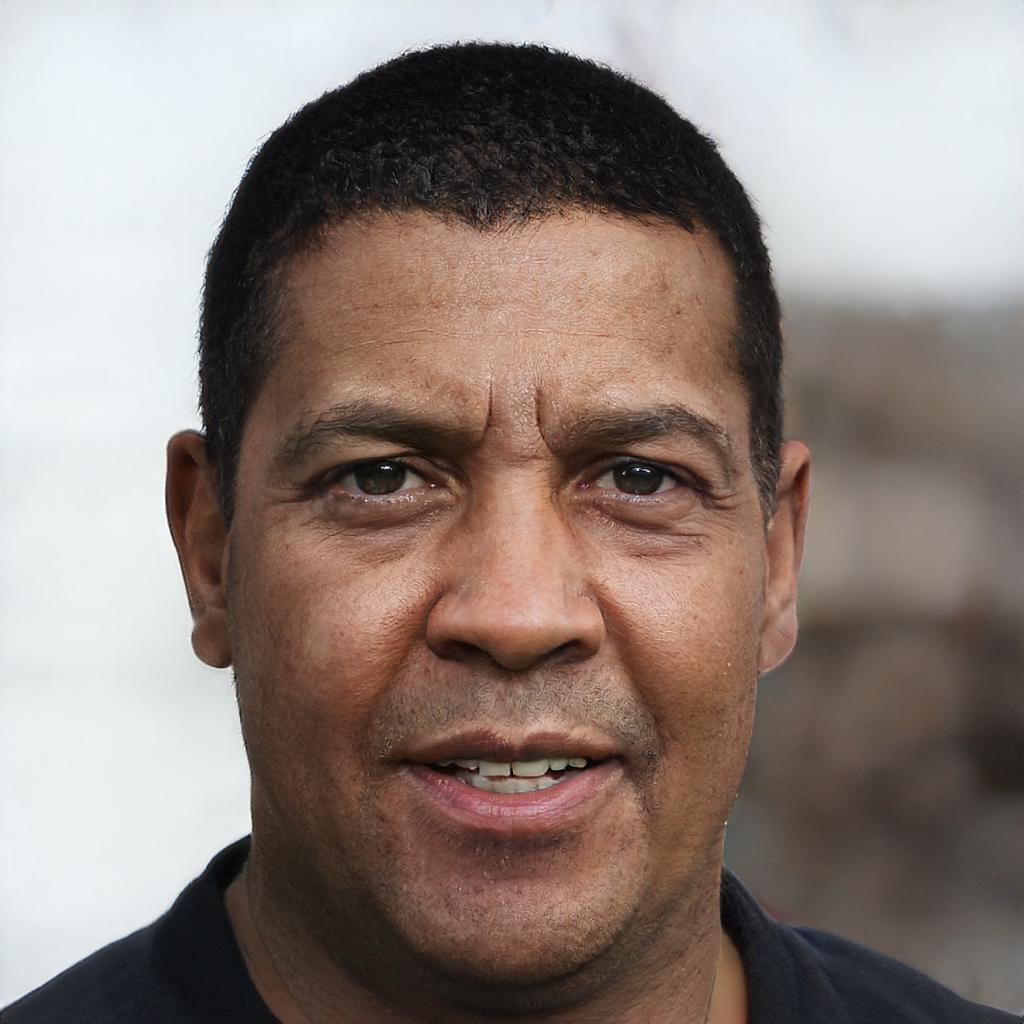} &
        \includegraphics[width=0.135\linewidth]{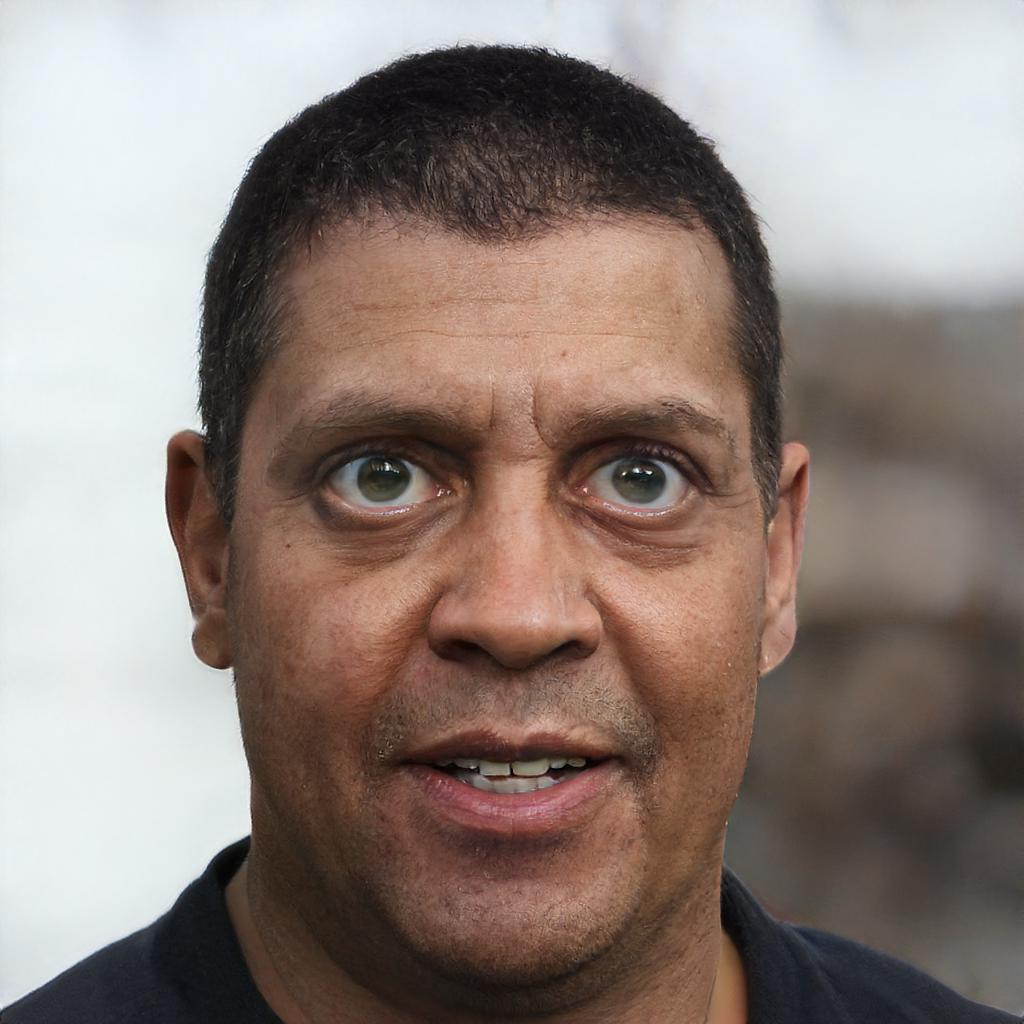} 
        \tabularnewline

        \includegraphics[width=0.135\linewidth]{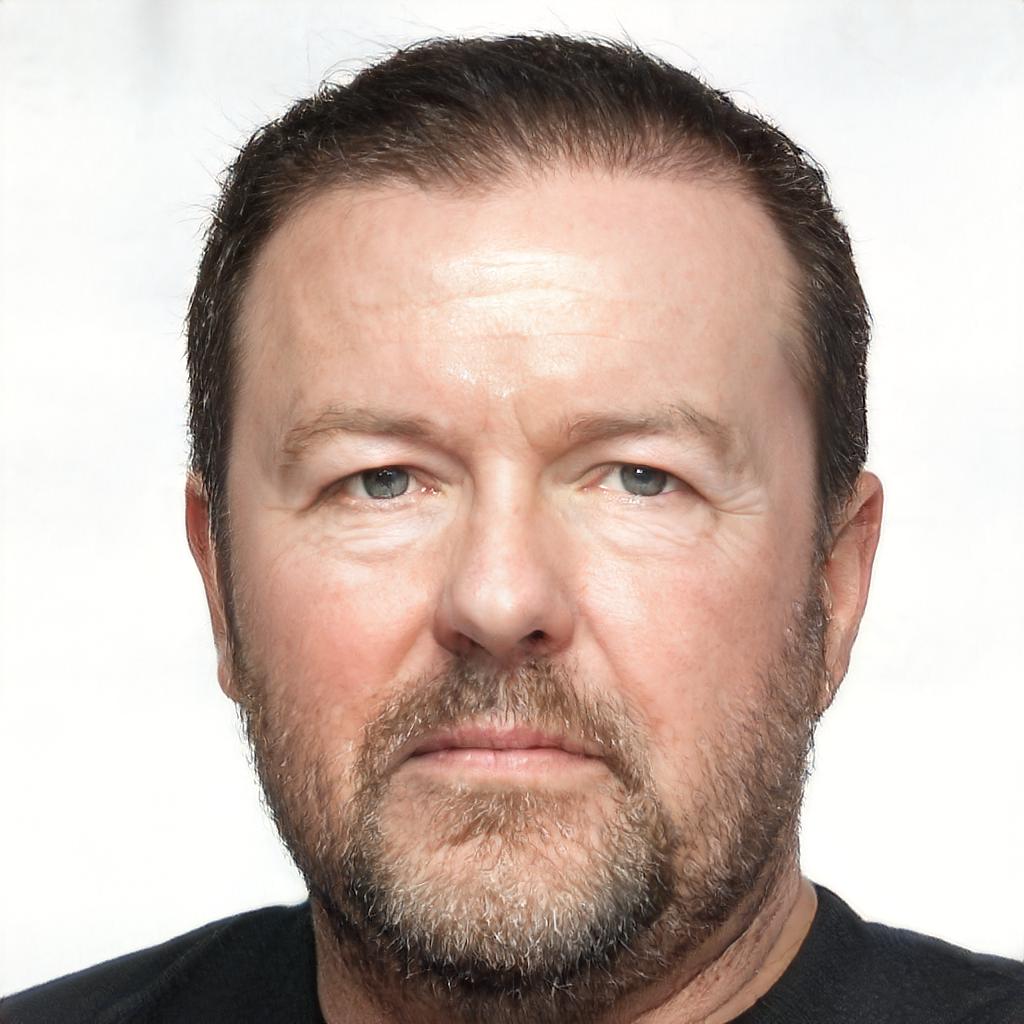} &&
        \raisebox{0.35in}{\rotatebox[origin=t]{90}{\footnotesize StyleGAN2}} &
        \includegraphics[width=0.135\linewidth]{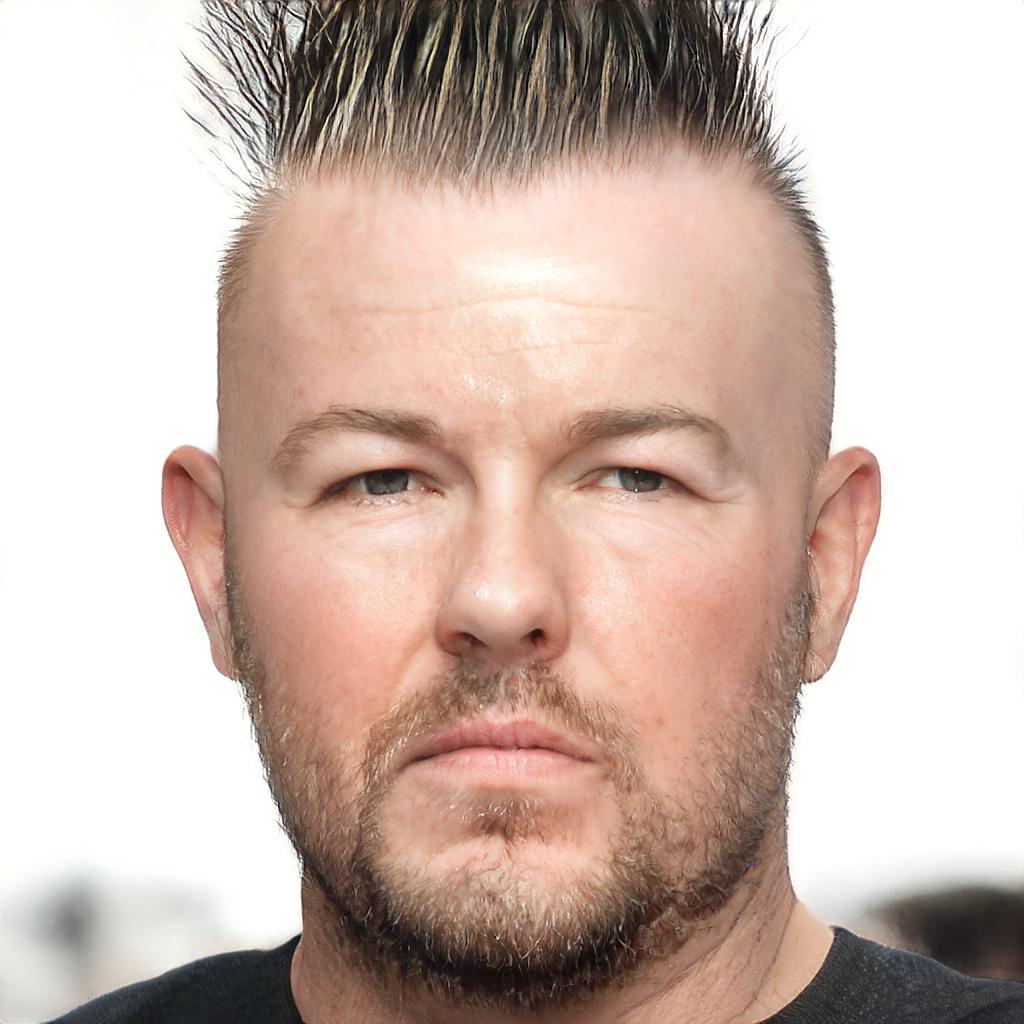} &
        \includegraphics[width=0.135\linewidth]{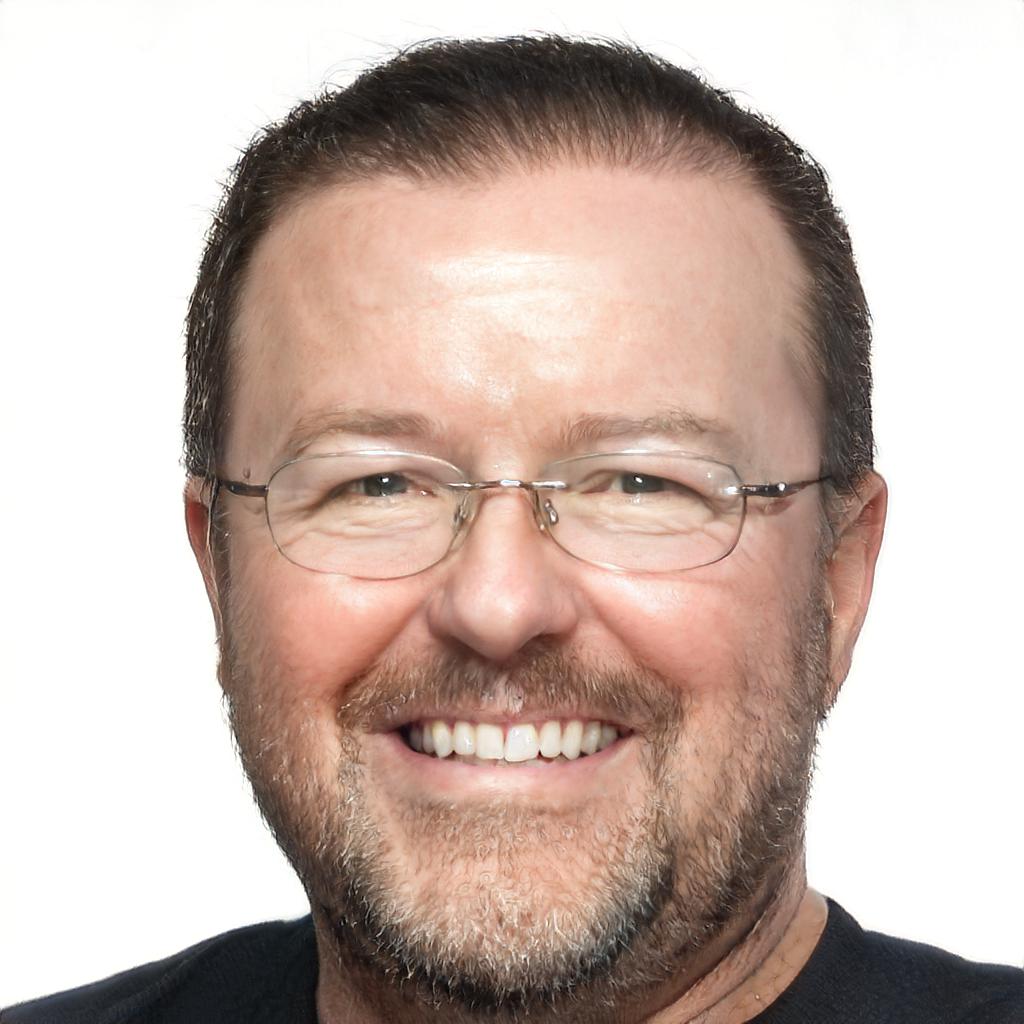} &
        \includegraphics[width=0.135\linewidth]{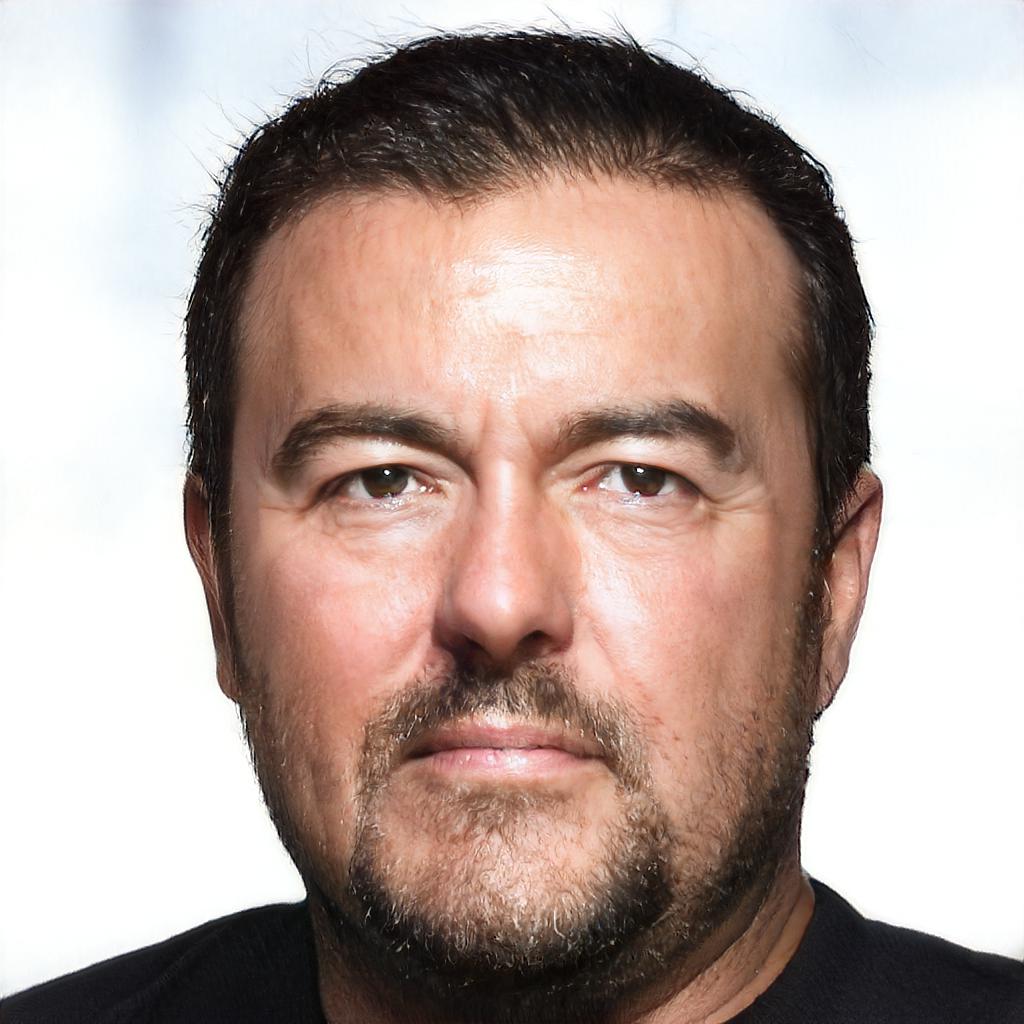} &
        \includegraphics[width=0.135\linewidth]{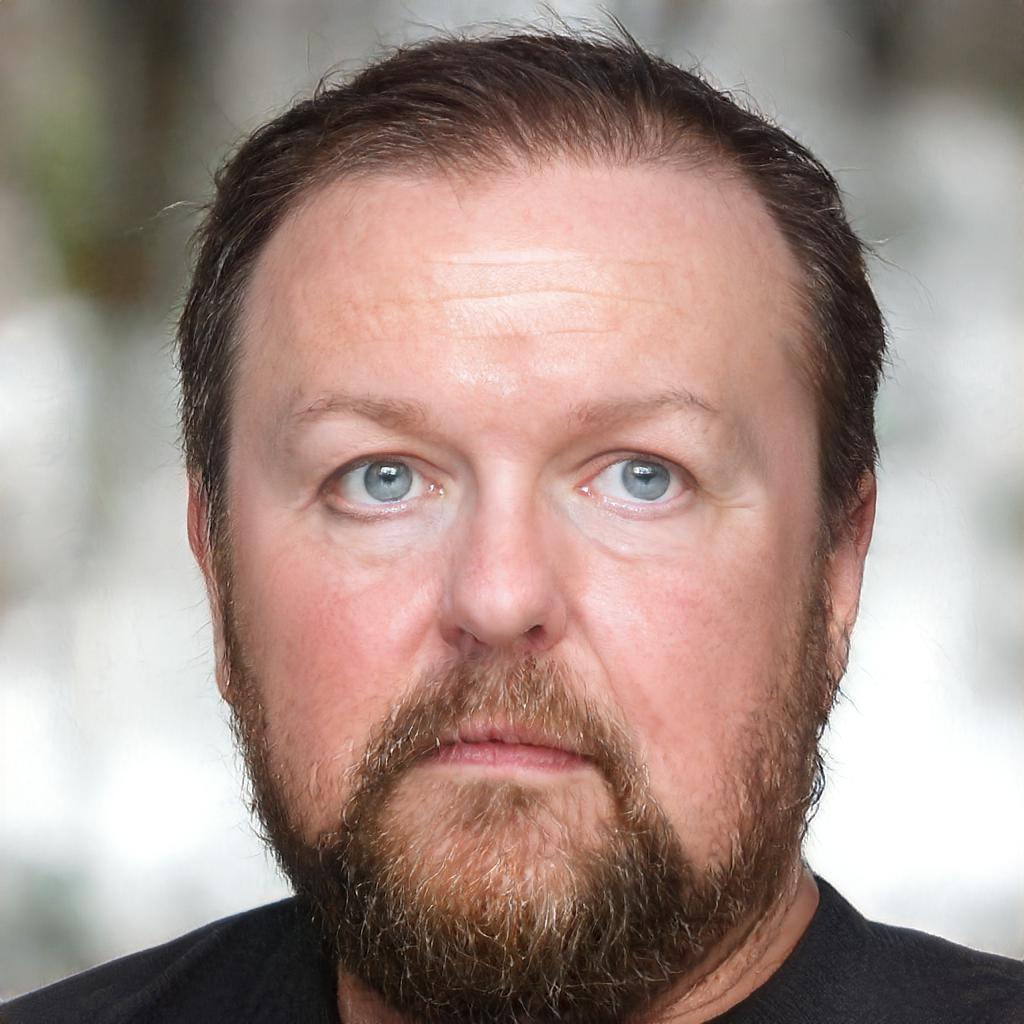} 
        \tabularnewline
        && \raisebox{0.35in}{\rotatebox[origin=t]{90}{\footnotesize StyleFusion}} &
        \includegraphics[width=0.135\linewidth]{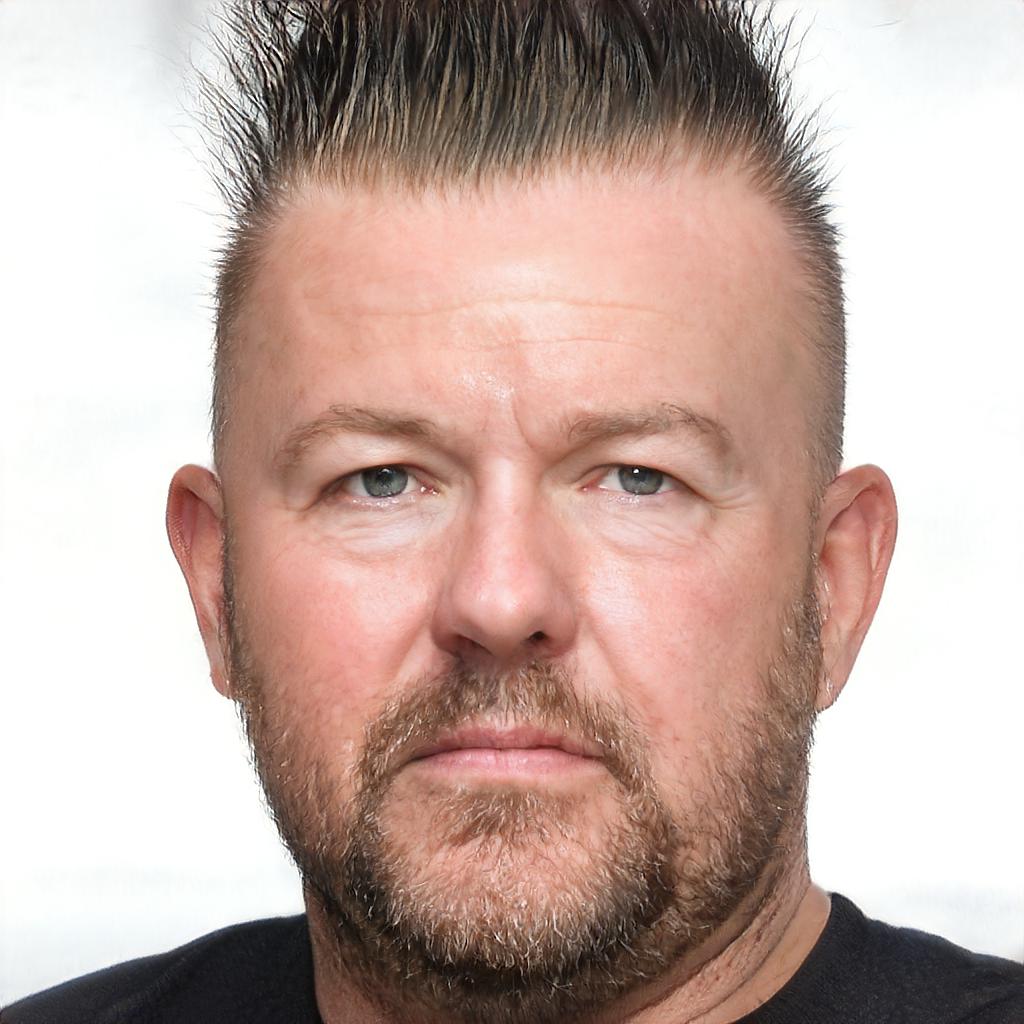} &
        \includegraphics[width=0.135\linewidth]{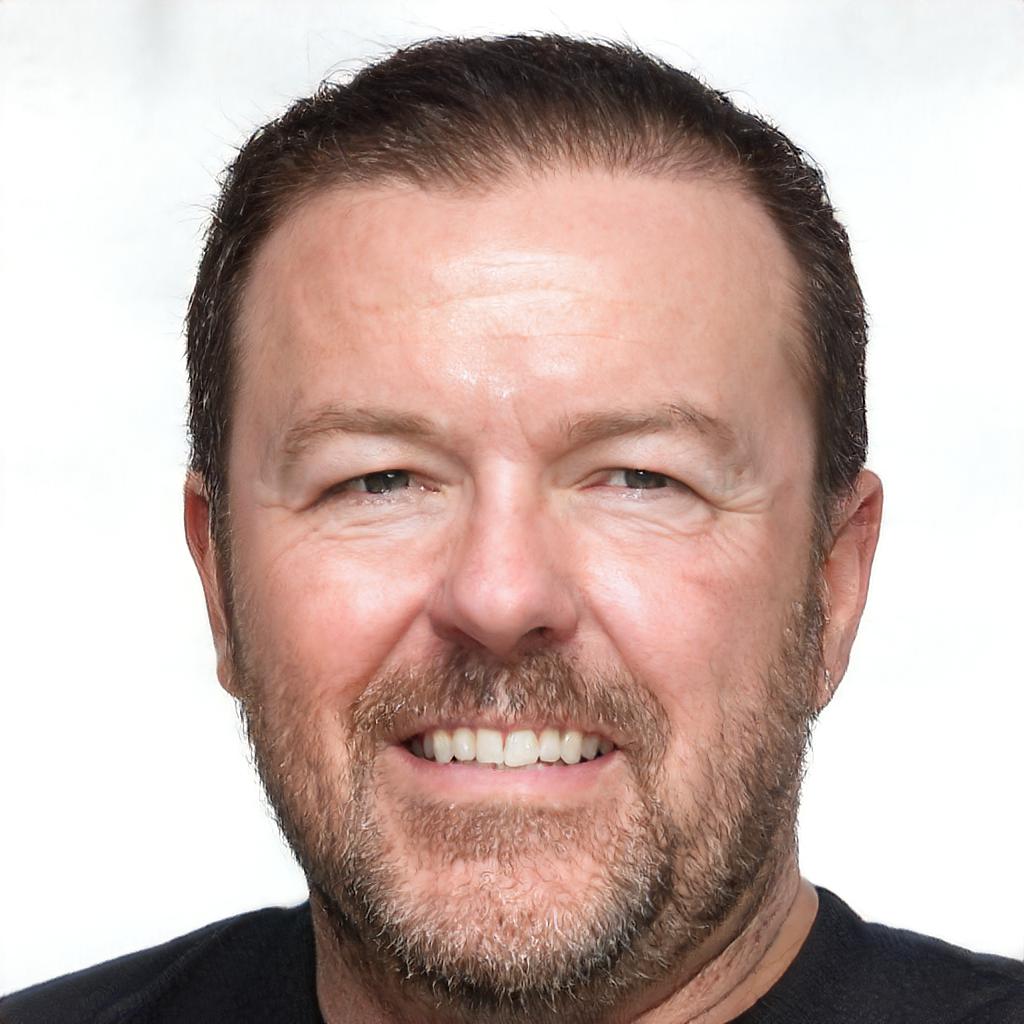} &
        \includegraphics[width=0.135\linewidth]{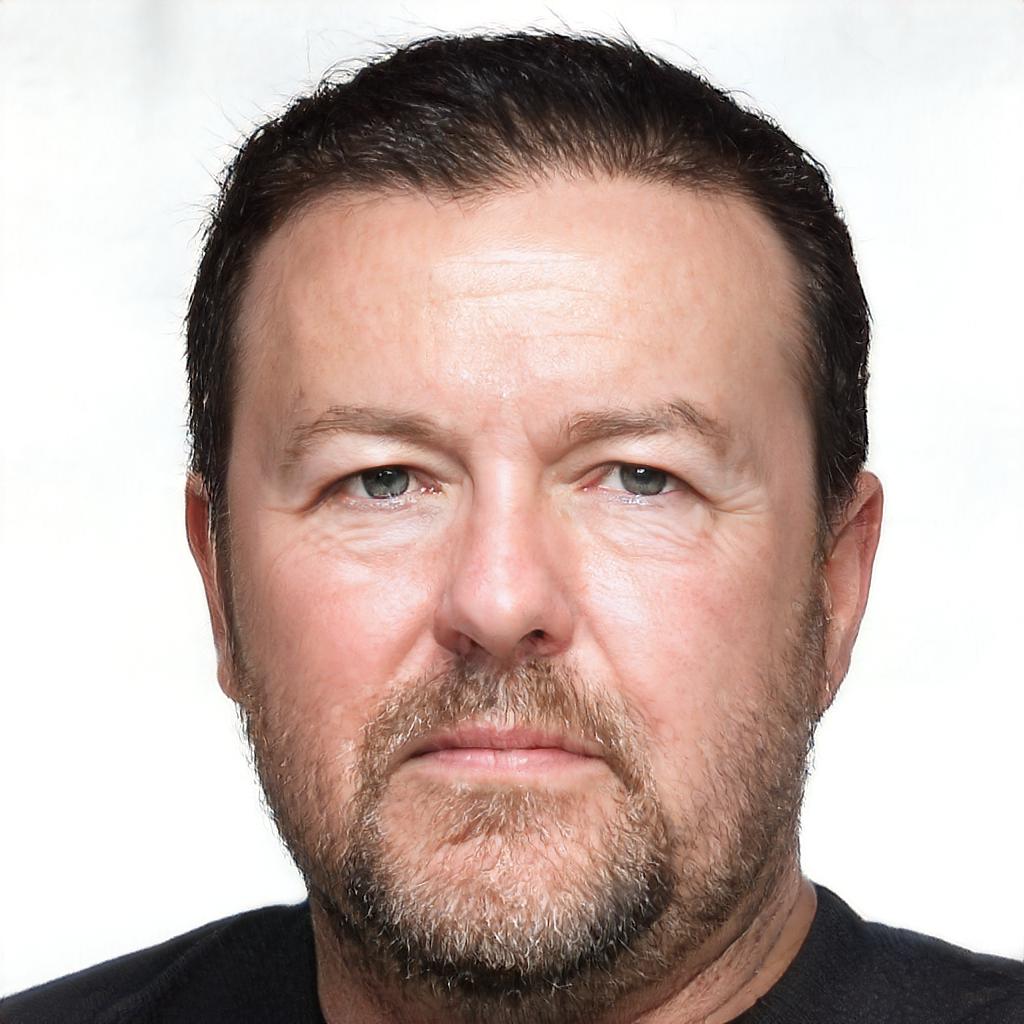} &
        \includegraphics[width=0.135\linewidth]{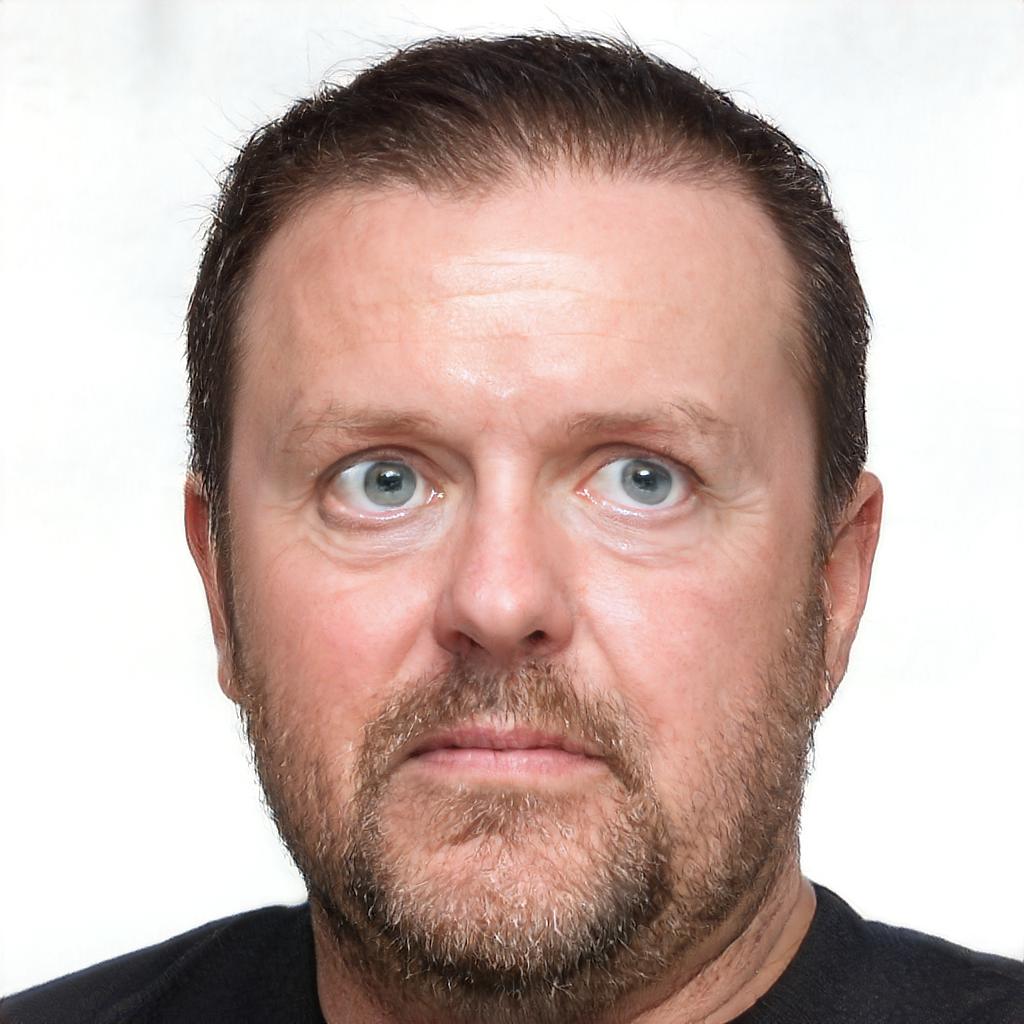} 
        \tabularnewline

        \includegraphics[width=0.135\linewidth]{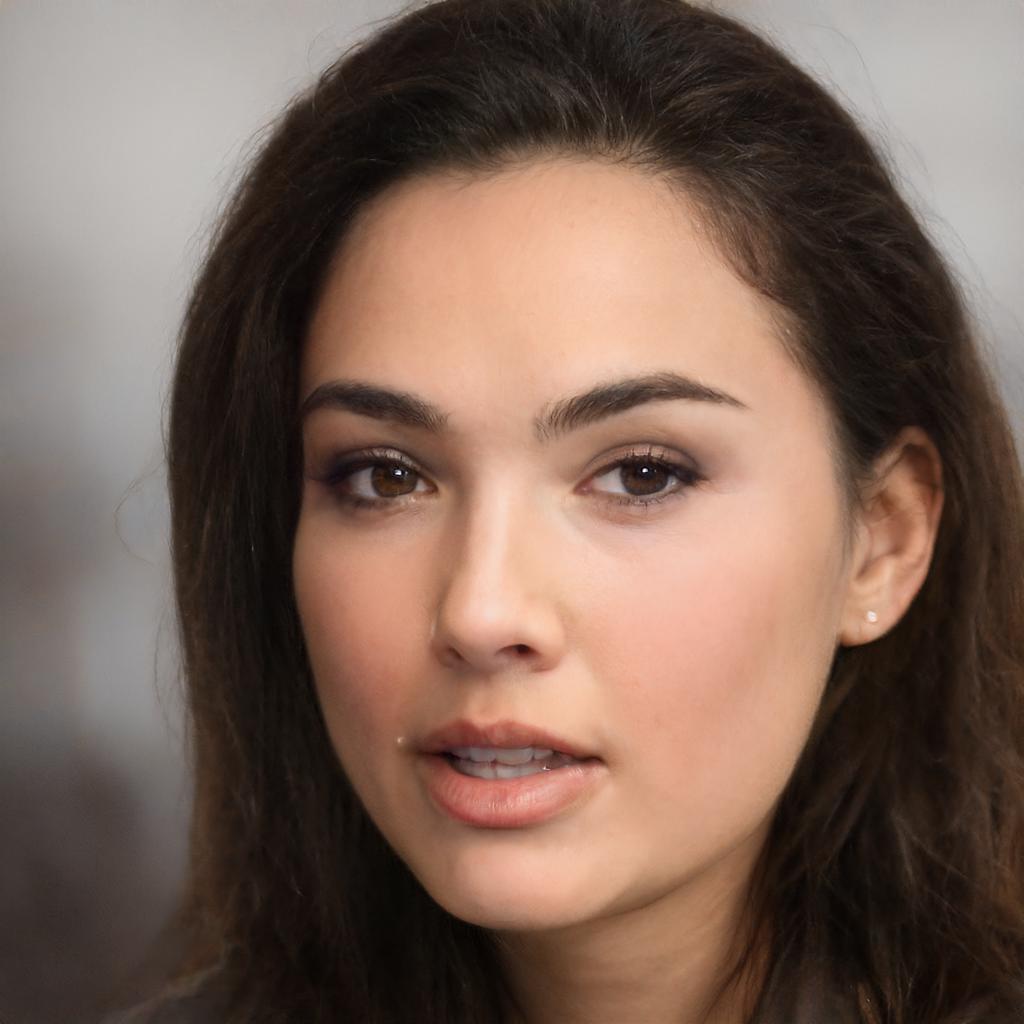} &&
        \raisebox{0.35in}{\rotatebox[origin=t]{90}{\footnotesize StyleGAN2}} &
        \includegraphics[width=0.135\linewidth]{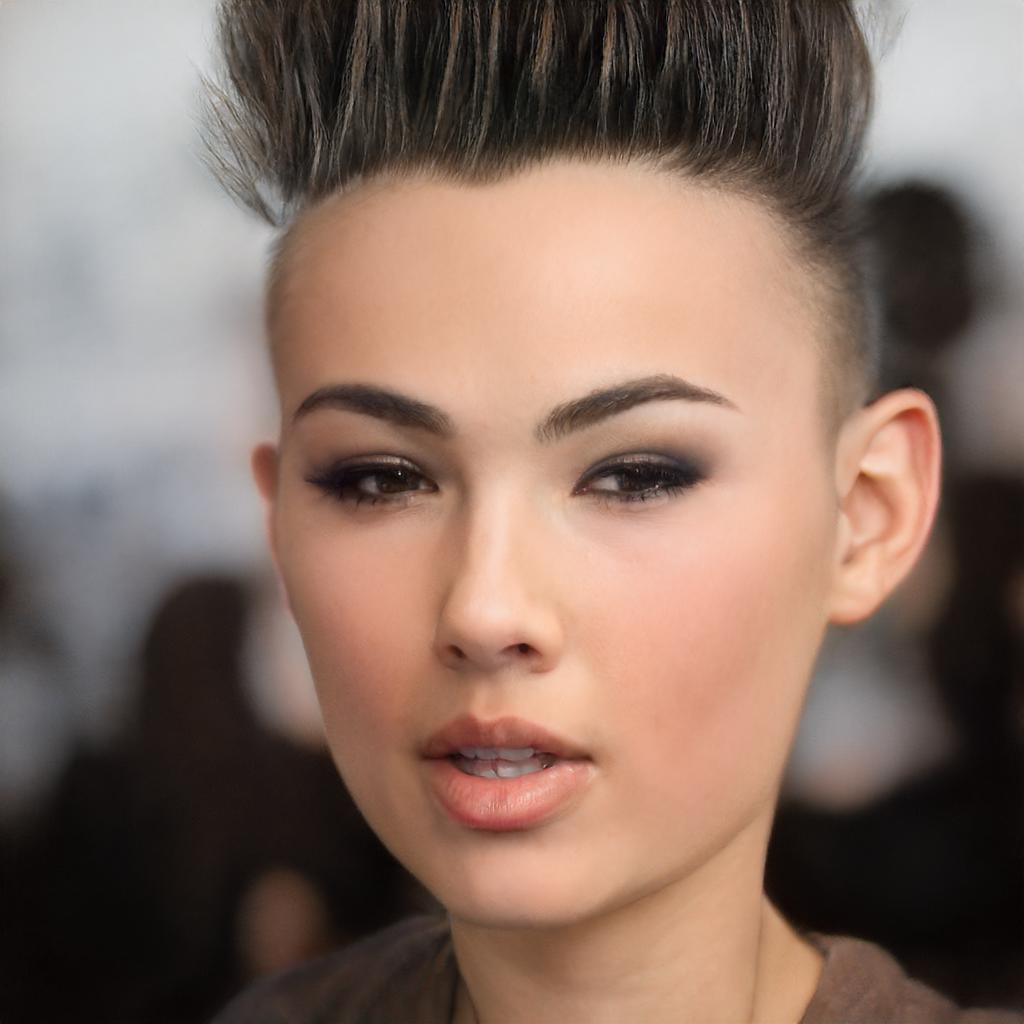} &
        \includegraphics[width=0.135\linewidth]{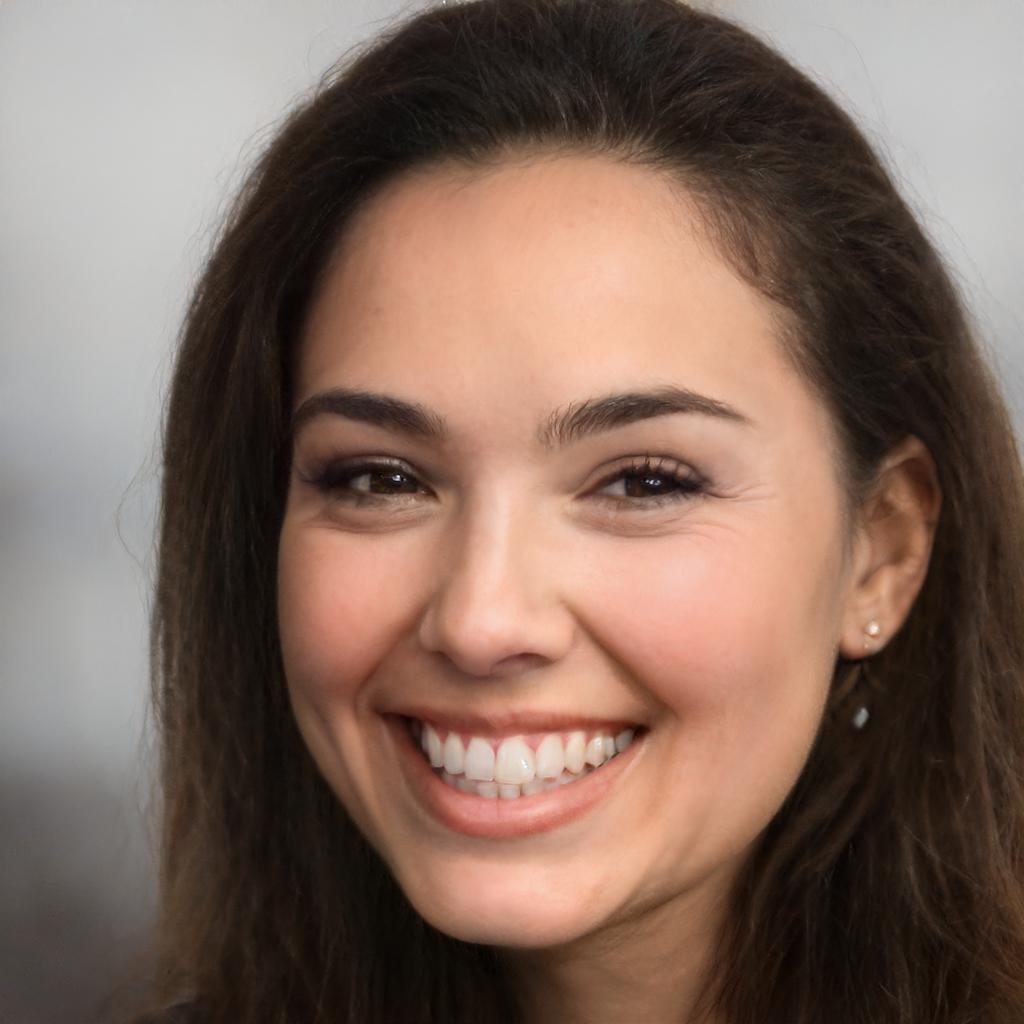} &
        \includegraphics[width=0.135\linewidth]{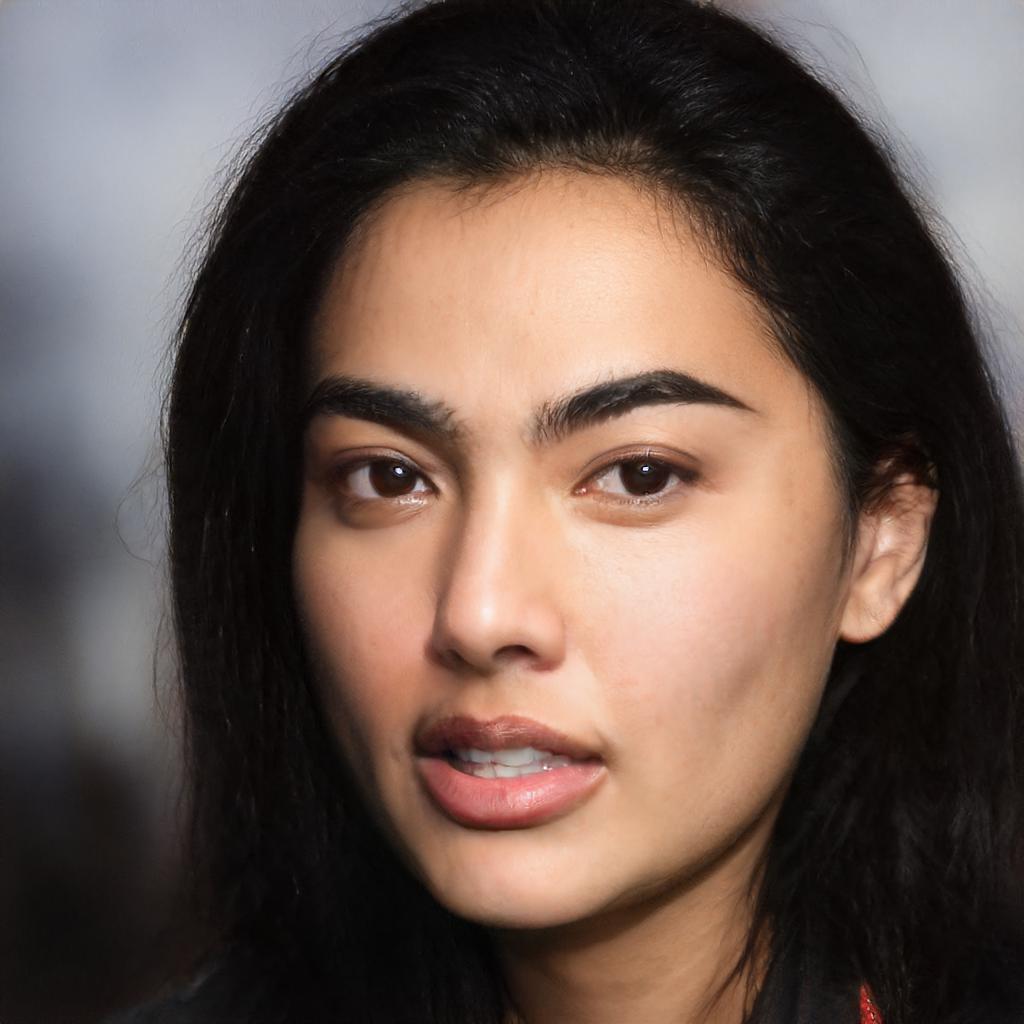} &
        \includegraphics[width=0.135\linewidth]{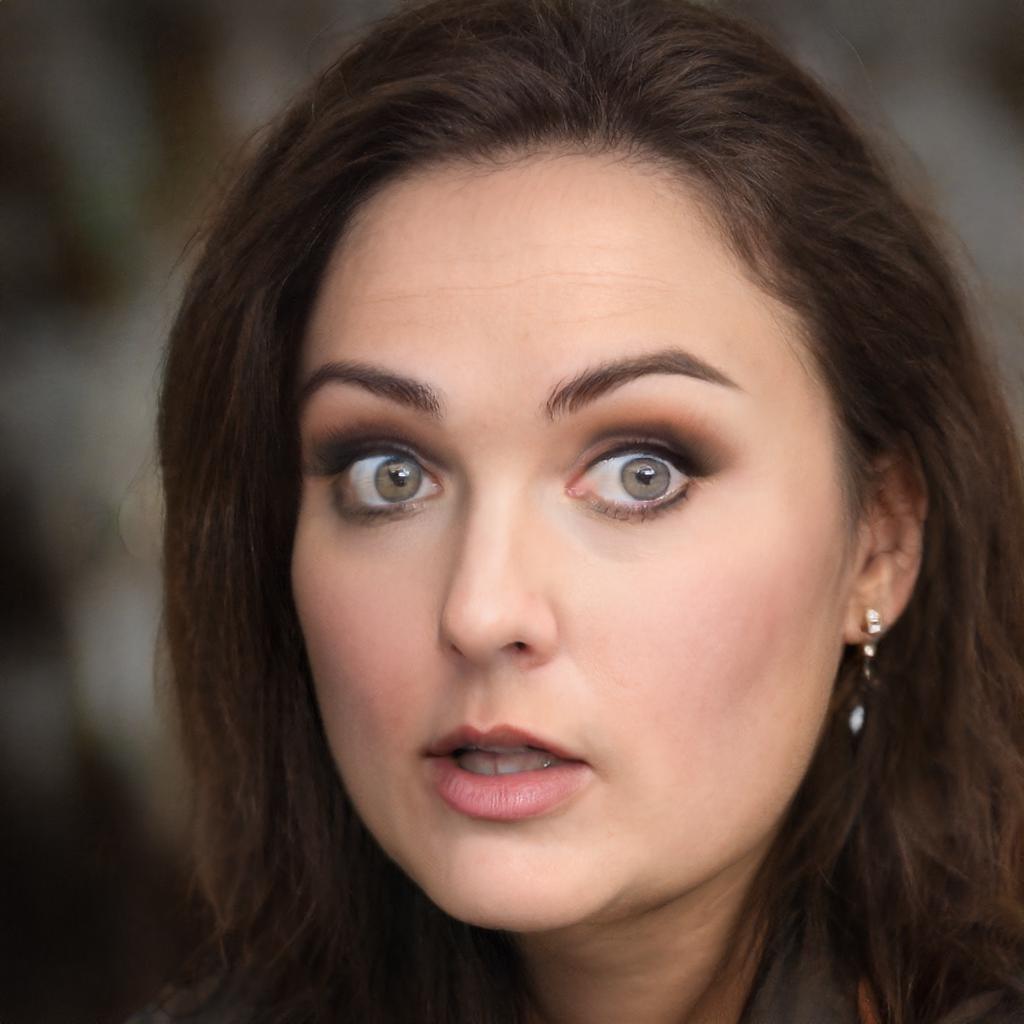} 
        \tabularnewline
        && \raisebox{0.35in}{\rotatebox[origin=t]{90}{\footnotesize StyleFusion}} &
        \includegraphics[width=0.135\linewidth]{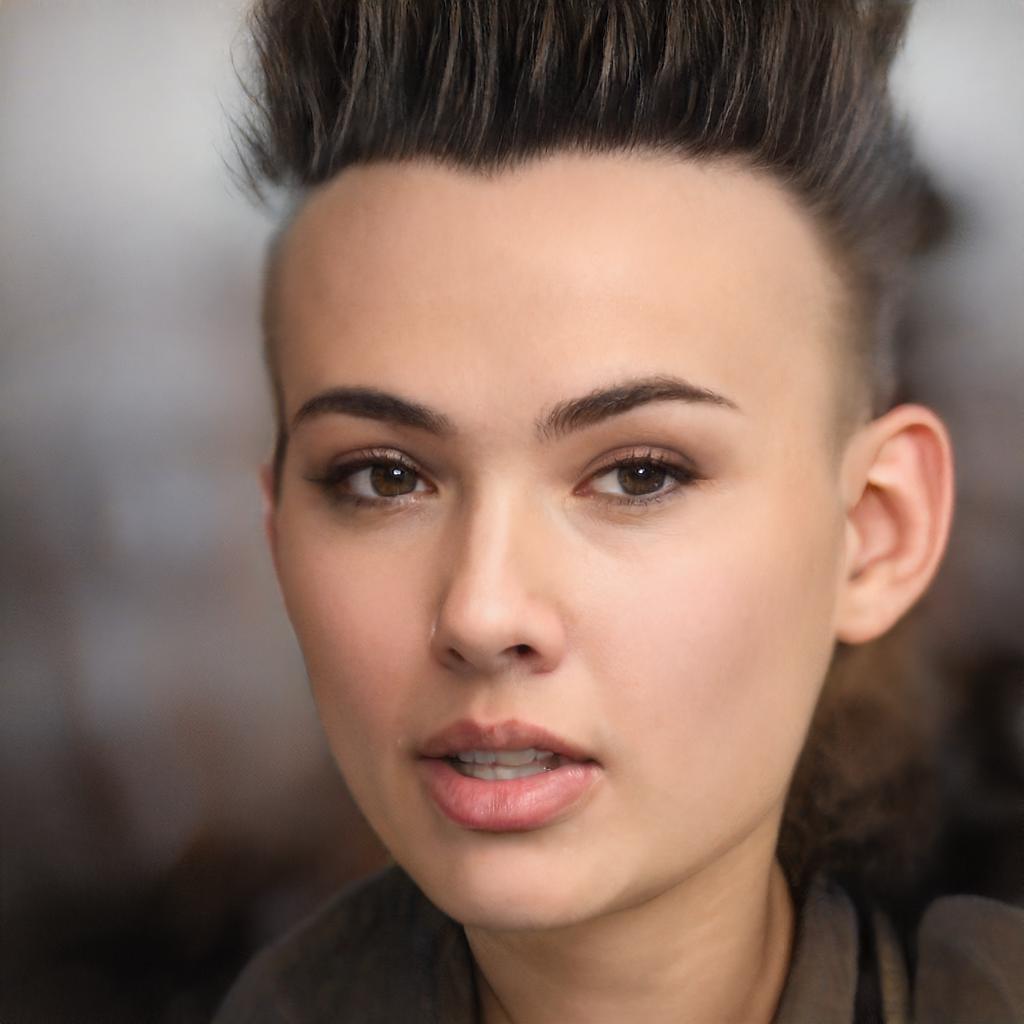} &
        \includegraphics[width=0.135\linewidth]{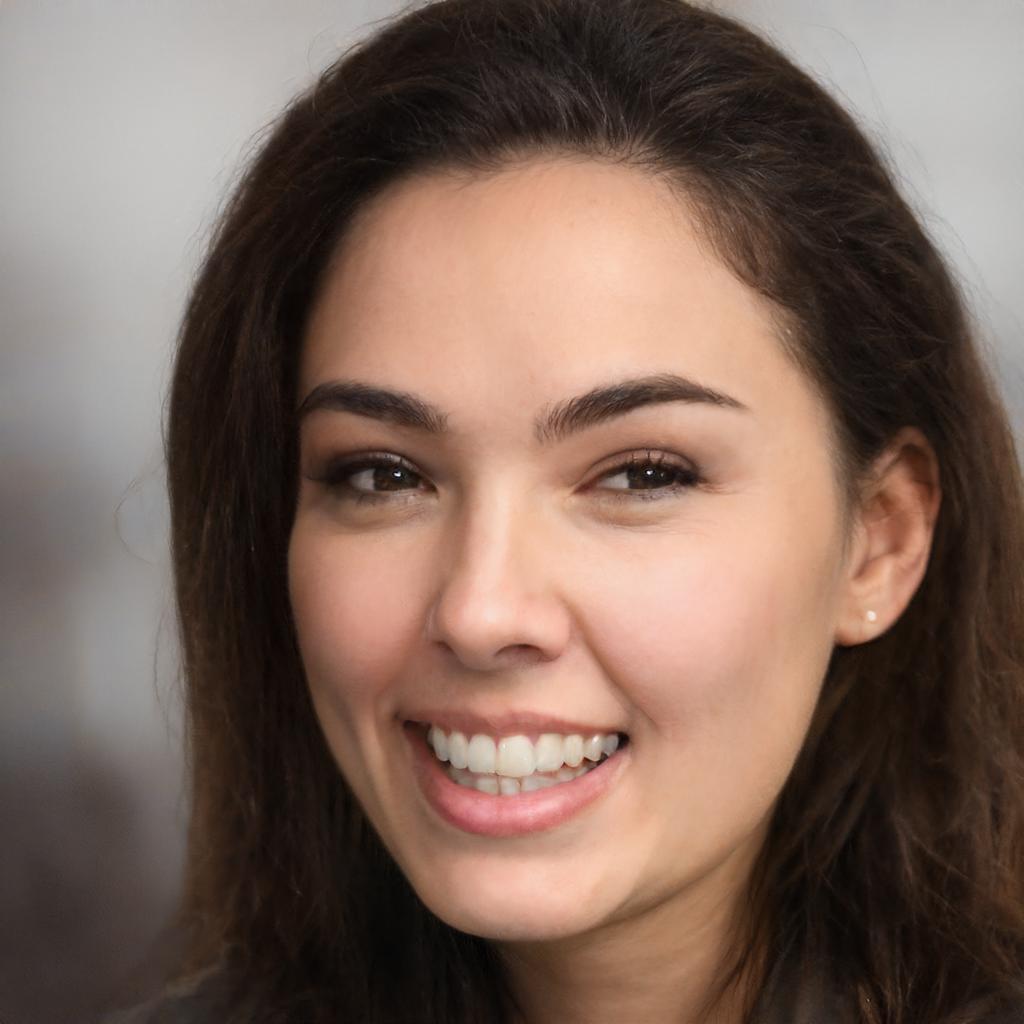} &
        \includegraphics[width=0.135\linewidth]{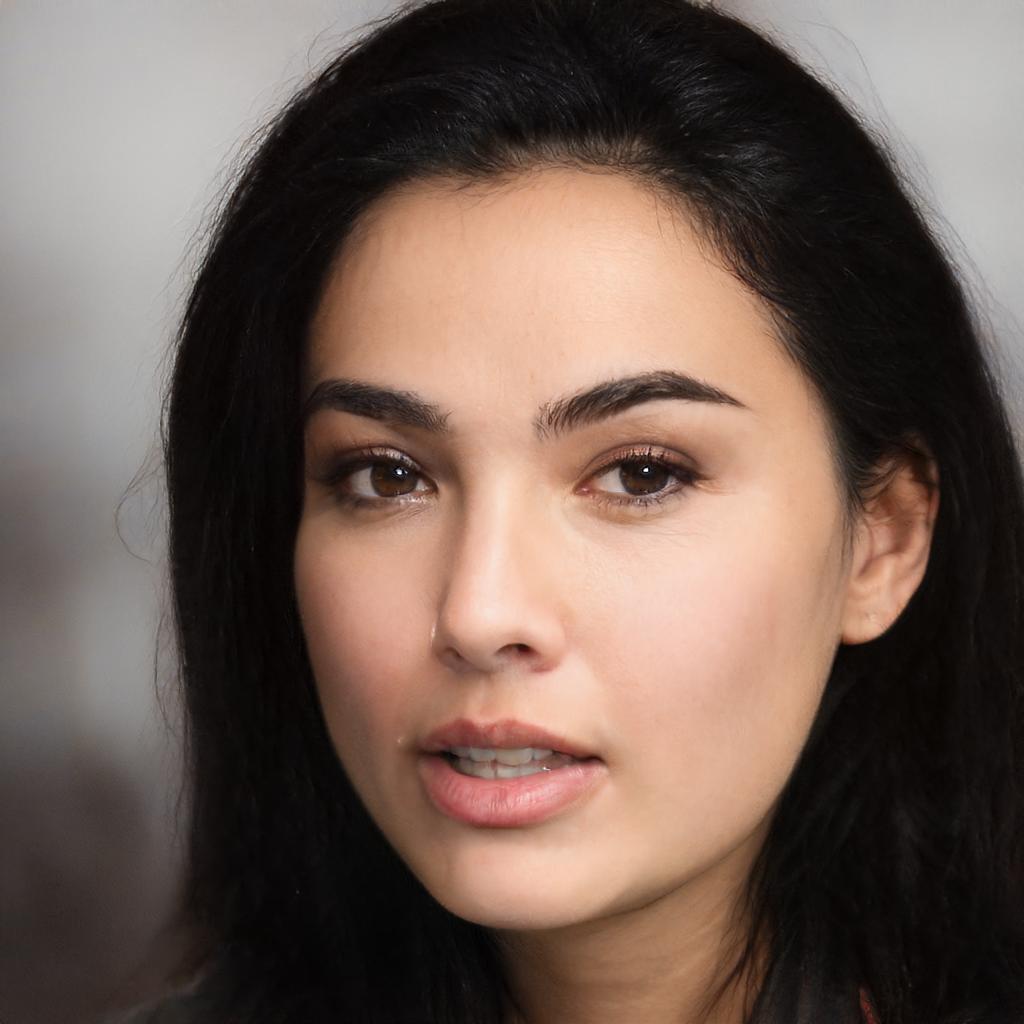} &
        \includegraphics[width=0.135\linewidth]{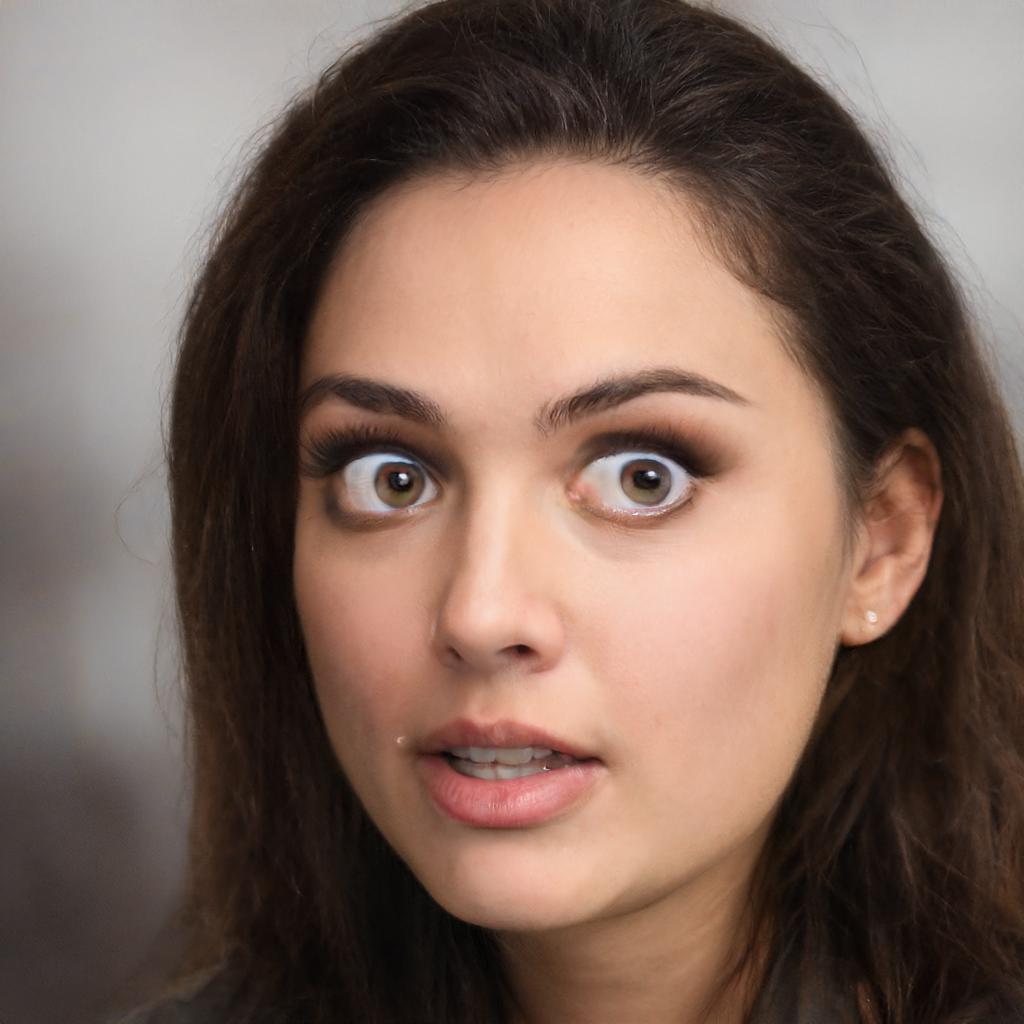} 
        \tabularnewline

        \includegraphics[width=0.135\linewidth]{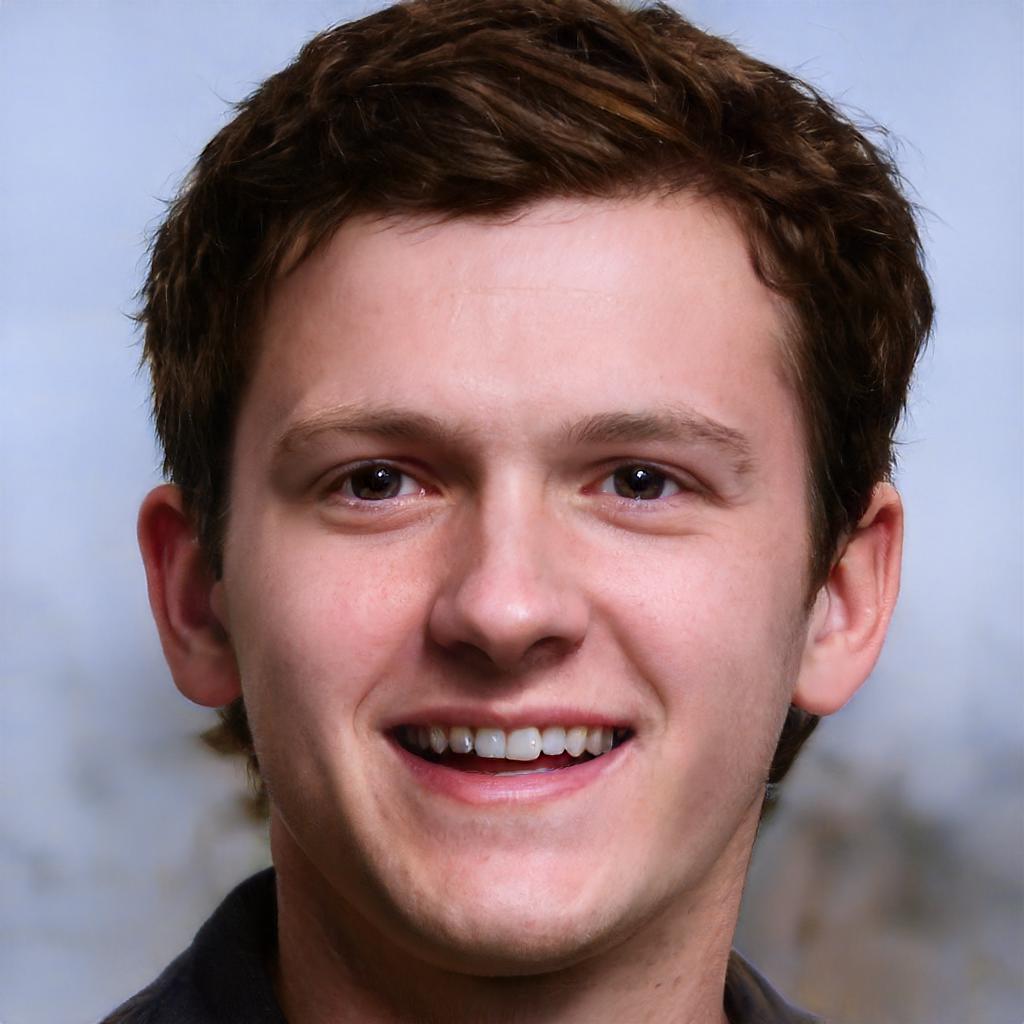} &&
        \raisebox{0.35in}{\rotatebox[origin=t]{90}{\footnotesize StyleGAN2}} &
        \includegraphics[width=0.135\linewidth]{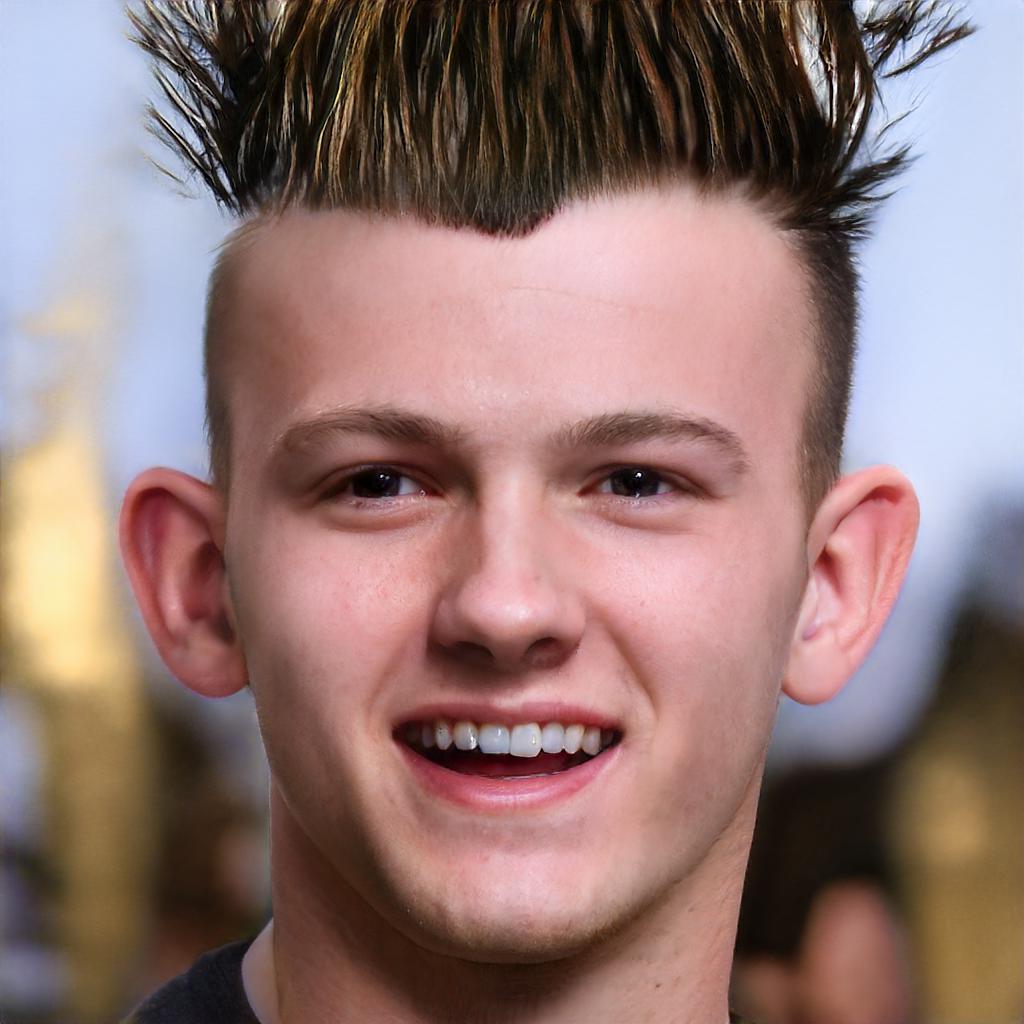} &
        \includegraphics[width=0.135\linewidth]{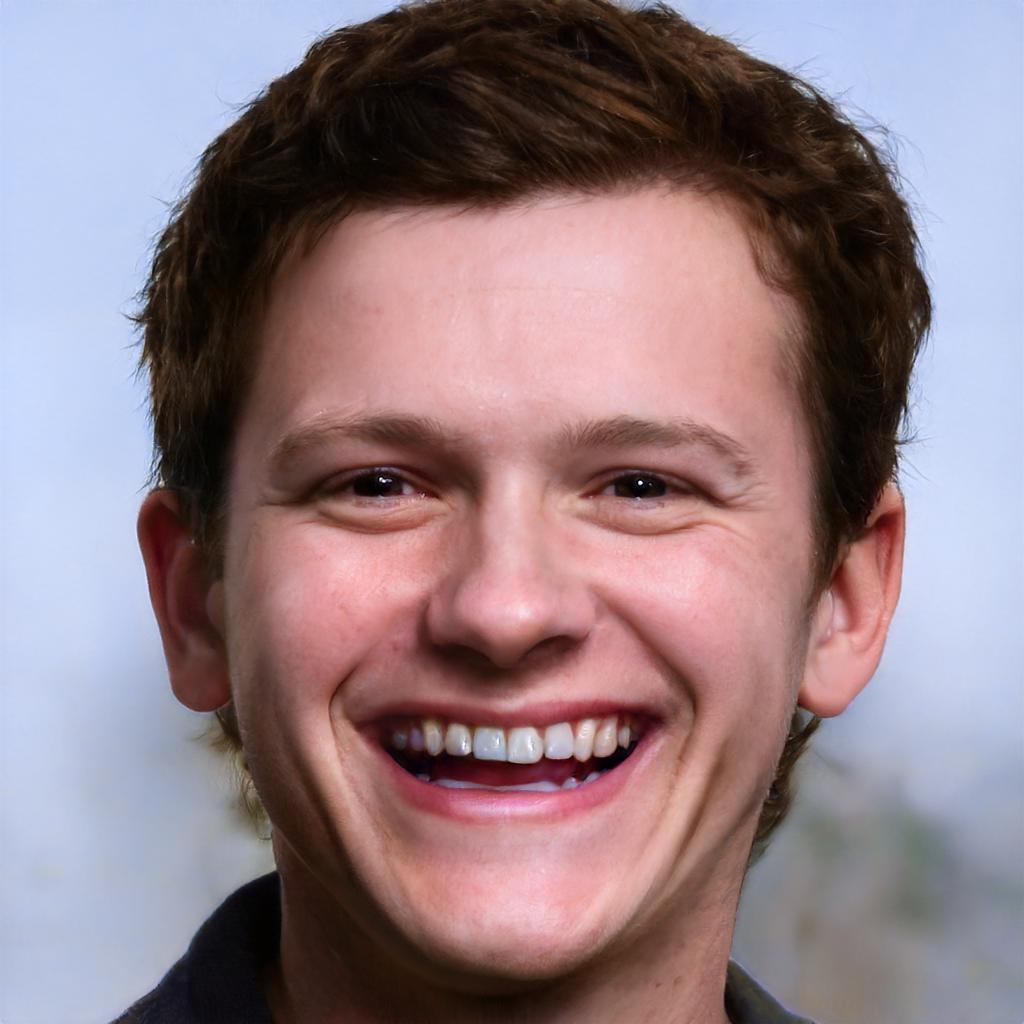} &
        \includegraphics[width=0.135\linewidth]{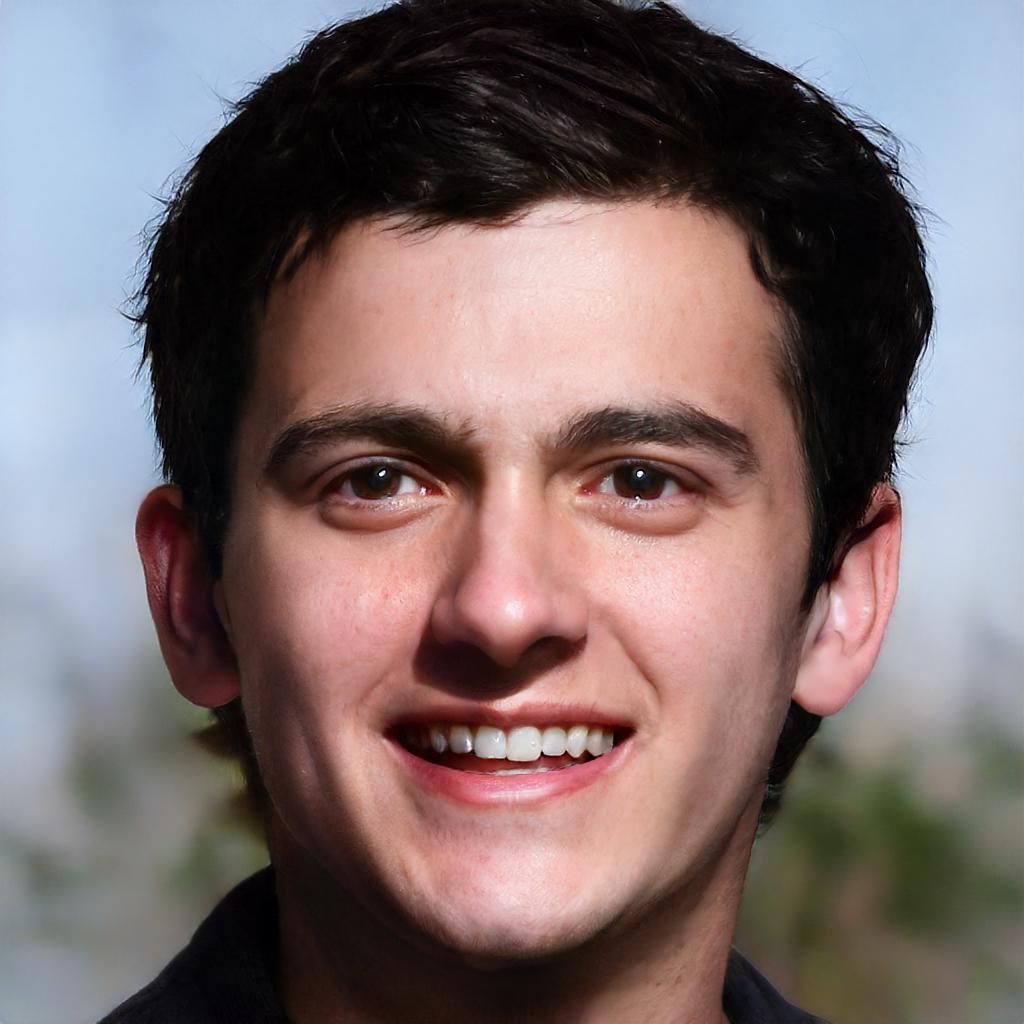} &
        \includegraphics[width=0.135\linewidth]{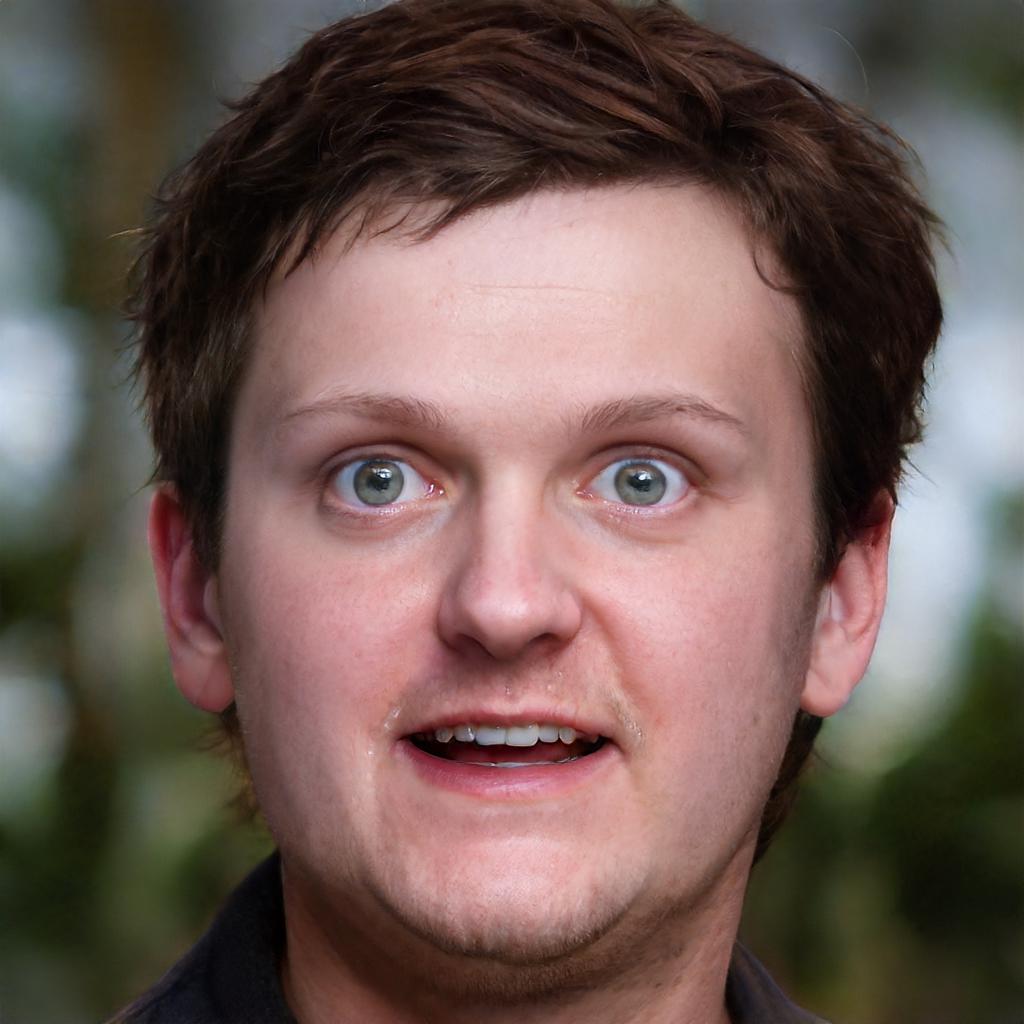} 
        \tabularnewline
        && \raisebox{0.35in}{\rotatebox[origin=t]{90}{\footnotesize StyleFusion}} &
        \includegraphics[width=0.135\linewidth]{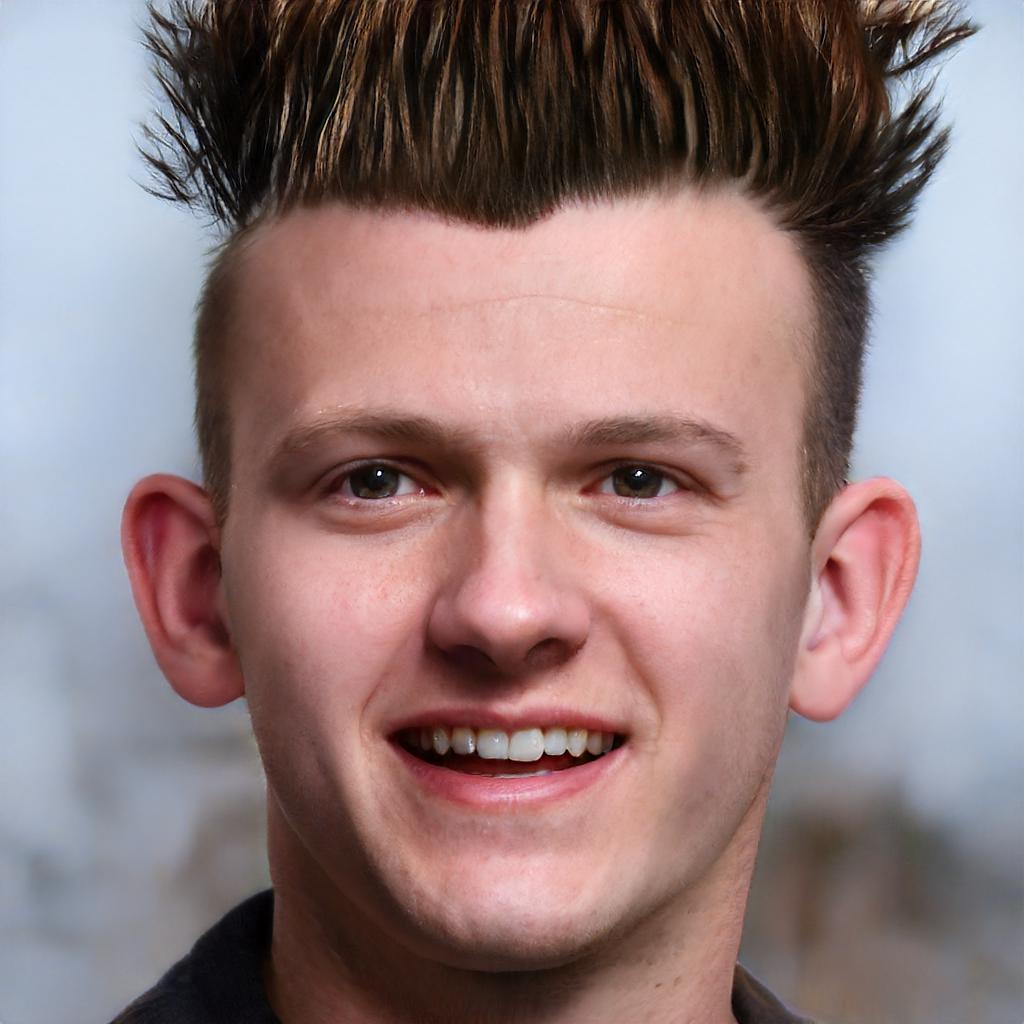} &
        \includegraphics[width=0.135\linewidth]{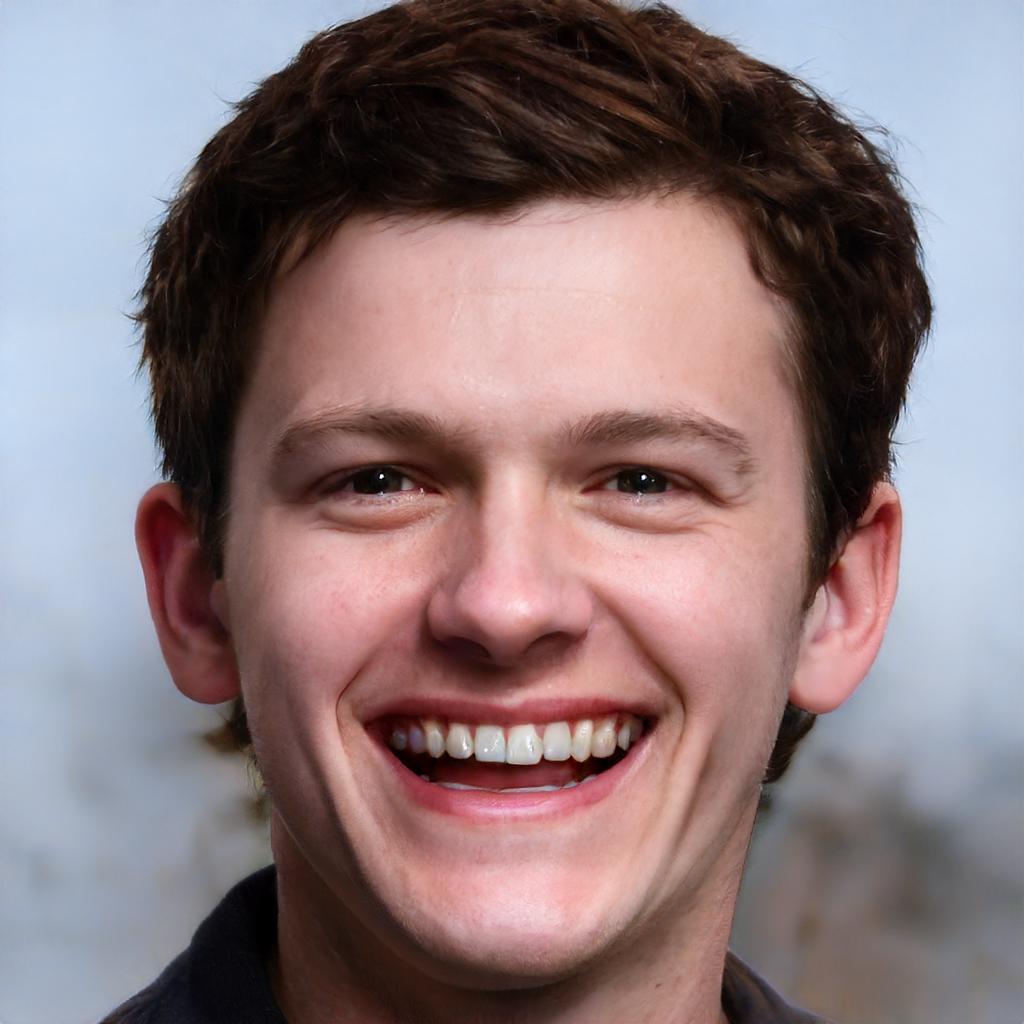} &
        \includegraphics[width=0.135\linewidth]{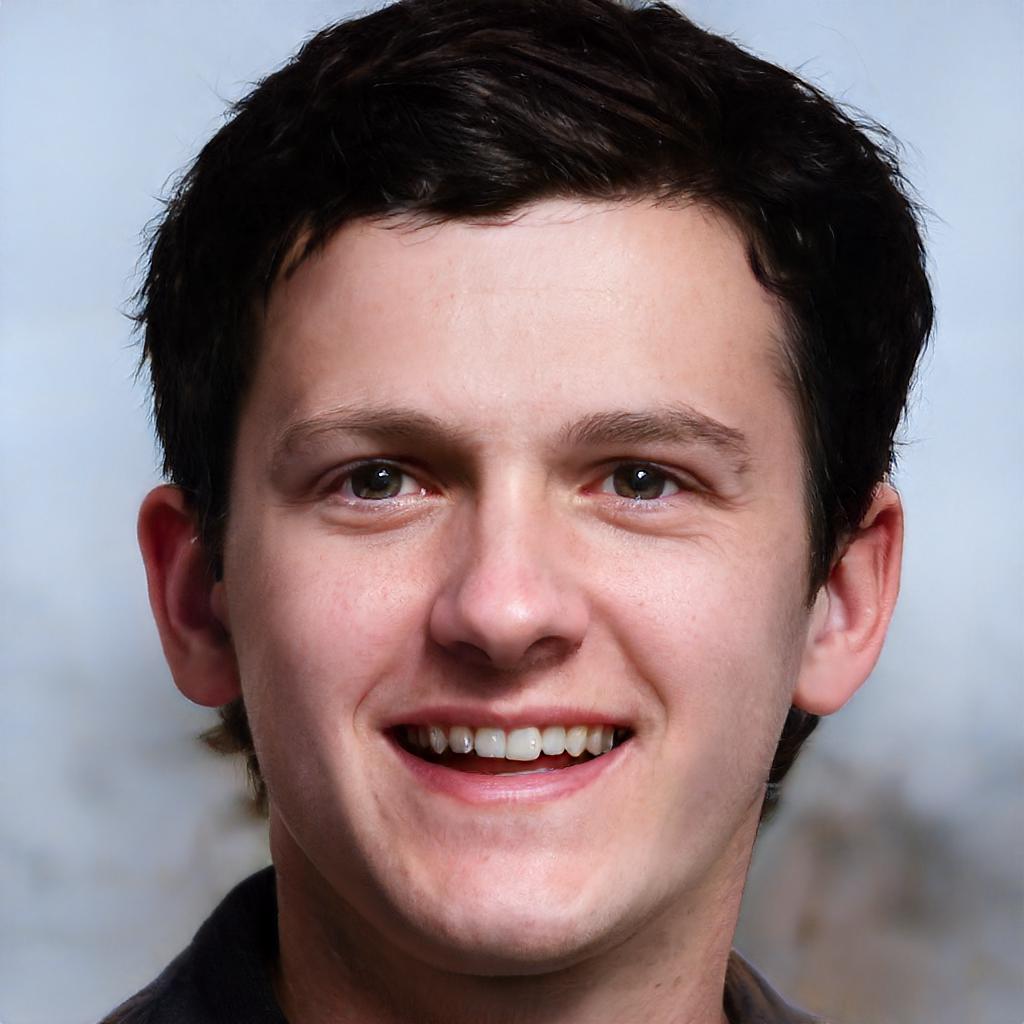} &
        \includegraphics[width=0.135\linewidth]{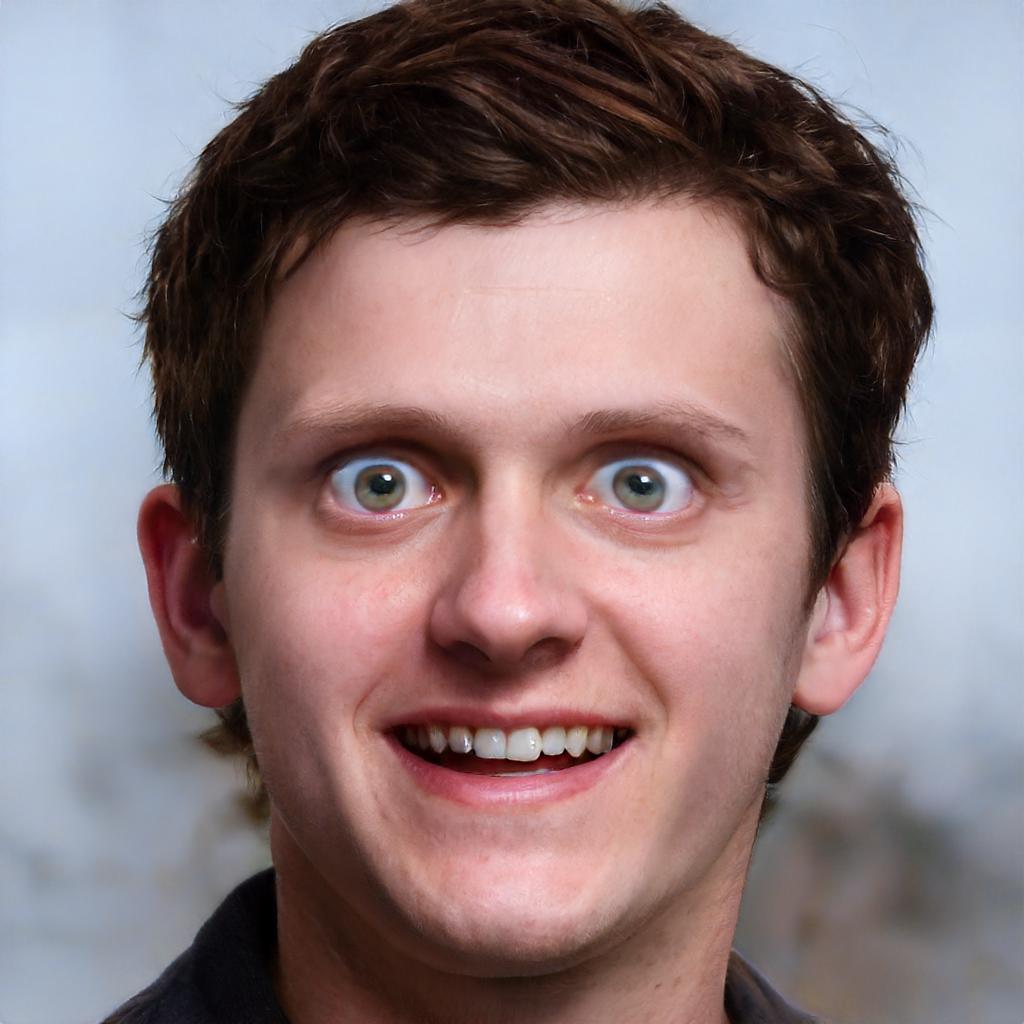} 
        \tabularnewline

        Input &&& Mohawk & Smile & Hair Albedo & Fearful Eyes
        \tabularnewline
        \end{tabular}
        }
    \vspace{0.1cm}
    \caption{Latent traversal editing techniques performed on real images with and without StyleFusion.}
    \label{fig:latent_editing_appendix_3}
\end{figure*}

The remainder of this document contains additional results, as follows:
\begin{enumerate}
    \item Figures~\ref{fig:local_control_faces2} and ~\ref{fig:local_control_faces_3} demonstrate StyleFusion's local control over the human facial domain. 
    \item Figure~\ref{fig:local_control_cars2} illustrates StyleFusion's local control over the cars domain. 
    \item Figures~\ref{fig:appendix_ffhq_all_1} and ~\ref{fig:appendix_ffhq_common_bg_hair} demonstrate StyleFusion's ability to control global features such as head pose and hair, respectively. 
    \item Figure~\ref{fig:appendix_car_all_1} demonstrate StyleFusion's global control over the car pose. 
    \item Figures~\ref{fig:appendix_church_common_1} and ~\ref{fig:appendix_church_common_2} illustrate StyleFusion's ability to generate multiple church buildings of the same style but different structure. 
    \item Figures~\ref{fig:appendix_cross_pose_hair1} and ~\ref{fig:appendix_cross_pose_hair2} demonstrates StyleFusion's ability to perform hair transform from a given reference image to a desired target image at varying poses.
    \item Figures~\ref{fig:latent_editing_appendix_1},~\ref{fig:latent_editing_appendix_2}, and~\ref{fig:latent_editing_appendix_3} demonstrates various latent space editing techniques performed with and without StyleFusion's learned disentanglement.
\end{enumerate}


\end{document}